# Theoretical Foundations of Adversarially Robust Learning

A DISSERTATION PRESENTED
BY
OMAR MONTASSER
TO
THE TOYOTA TECHNOLOGICAL INSTITUTE AT CHICAGO
CHICAGO, IL
MAY 2023

*Submitted in partial fulfillment of the requirements for the degree of Doctor of Philosophy in Computer Science.*

**Thesis Committee:**
Nathan Srebro (Thesis Advisor)
Avrim Blum
Steve Hanneke
Adam Tauman Kalai

# Theoretical Foundations of Adversarially Robust Learning

## Abstract

Despite extraordinary progress, current machine learning systems have been shown to be brittle against *adversarial examples*: seemingly innocuous but carefully crafted perturbations of test examples that cause machine learning predictors to misclassify. Can we learn predictors robust to adversarial examples? and how? There has been much empirical interest in this contemporary challenge in machine learning, and in this thesis, we address it from a theoretical perspective.

In this thesis, we explore what robustness properties can we hope to guarantee against adversarial examples and develop an understanding of how to algorithmically guarantee them. We illustrate the need to go beyond traditional approaches and principles such as empirical risk minimization and uniform convergence, and make contributions that can be categorized as follows: (1) introducing problem formulations capturing aspects of emerging practical challenges in robust learning, (2) designing new learning algorithms with provable robustness guarantees, and (3) characterizing the complexity of robust learning and fundamental limitations on the performance of any algorithm.

بِسْمِ ٱللَّهِ ٱلرَّحْمَٰنِ ٱلرَّحِيمِ

In memory of my dear mother, Nabila

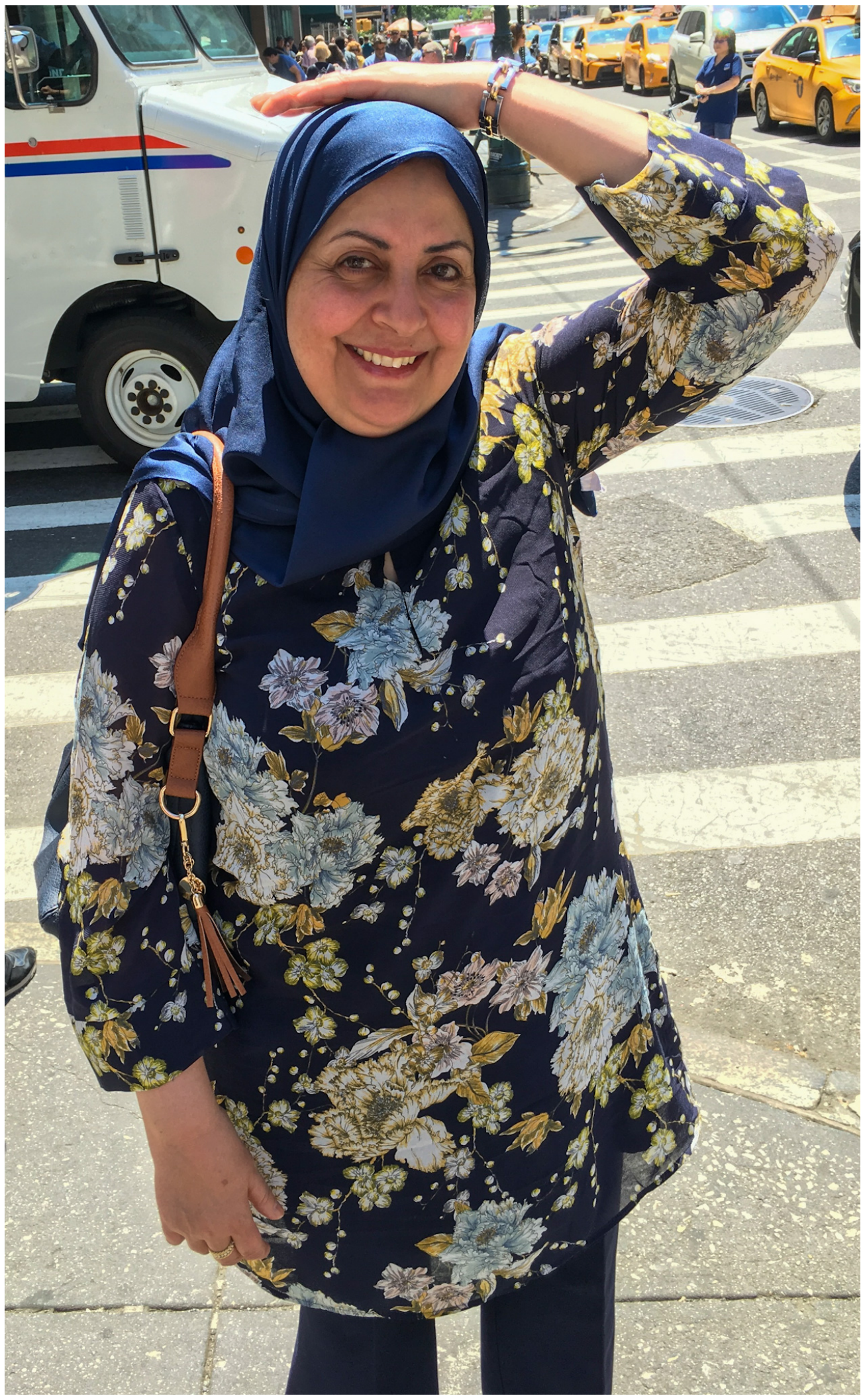

أبلِغ عَزيزًا فى ثنايا القلبِ مَنزله　　　أنى وإن كُنتُ لا ألقاهُ ألقاهُ
وإن طرفى موصولٌ برؤيتهِ　　　وإن تباعد عَن سُكناى سُكناهُ
يا ليته يعلمُ أنى لستُ أذكرهُ　　　وكيف أذكرهُ إذ لستُ أنساهُ
يا مَن توَّهم أنى لستُ أذكرهُ　　　واللهُ يعلم أنى لستُ أنساهُ
إن غابَ عنى فالروحُ مَسكنهُ　　　مَن يسكنُ الروح كيف القلبُ ينساهُ

تُنسب هذه الأبيات إلى على بن الجهم، و أبو الطيب المتنبى.

# Acknowledgments

I am grateful and thankful to many people who have helped and supported me in my PhD journey. My words will probably fall short in articulating this, but I will try my best.

First and foremost, I am extremely fortunate to have had Nati Srebro as my PhD advisor. Nati's endless enthusiasm and excitement about research is truly admirable, and continues to be a source of inspiration for me. Nati has shown me by example how one might pursue and strive for excellence in research. Beyond this, as I experienced firsthand, Nati genuinely cares about his students growth as independent researchers and is actively engaged in nurturing them in various forms. I am thankful for his support throughout the years. From giving me intellectual freedom to think about and work on what interests me, and helping me develop my taste in problems and formulating questions, to sharpening my thinking. Nati, thank you for everything: the exciting research meetings, the thrilling paper editing sessions, the improv classes, and the many doors you've helped open for me.

I have learned and grown a lot from working with Steve Hanneke. My PhD would certainly not be the same without him, and by far not as productive. Steve has taught me so much about learning theory. I am grateful for his generosity and patience with research meetings. Steve has a unique meeting style where we would meet together for at least two hours (sometimes more), and think and discuss problems and questions. These meetings were very effective and hugely contributed to my growth as a researcher.

One of the main factors that helped me decide to pursue my PhD at TTIC was Avrim Blum joining as Professor and Chief Academic Officer. Throughout my PhD, I greatly enjoyed our interactions which I learned a lot from. Having Avrim as a mentor and doing research with him has been rewarding in so many ways. In particular, closely seeing his style in advising and mentoring students inspires me. Avrim, thank you for your wonderful mentorship and for helping me with my many little and big questions.

I am grateful for the opportunity to work with Adam Kalai during my PhD. Adam's taste in problems, ingenuity, and attention to details are truly exemplary. Adam, thank you for your support and for teaching me great research practices.

I would also like to thank all of my other collaborators and co-authors: Saba Ahmadi, Ilias Diakonikolas, Surbhi Goel, Shafi Goldwasser, Yael Tauman Kalai, Pritish Kamath, Greg Shakhnarovich, Han Shao, Kevin Stangl, Hongyang Zhang. Working with you has been enjoyable, and taught me valuable lessons.

I would like to thank professors at TTIC: Karen Livescu, David McAllister, Greg Shakhnarovich, Madhur Tulsiani, Matthew Turk, and Matthew Walter. I have enjoyed interacting with you, and I have learned a lot from you inside and outside classes. I would also like to thank past research assistant professors at TTIC: Brian Bullins, Suriya Gunasekar, Sepideh Mahabadi, and Mesrob Ohannessian.

I am very thankful to the Machine Learning and Optimization reading group at TTIC and its participants throughout the years. This reading group has been one of the main weekly intellectual highlights during my PhD journey. Its stimulating discussions have benefited me greatly.

I am thankful to students at TTIC, past and present, for wonderful times throughout my PhD journey, inside and outside of TTIC. Sharing the PhD experience with you has made it more rewarding, as we could all closely relate and navigate the ups and downs together. I especially enjoyed our social activities and fun outings. Thank you for the adventurous bike trips, the thrilling ski trips, and the many other creative ways we have used our student funds. I especially want to thank: Anmol Kabra, Gene Li, Naren Manoj, Keziah Naggita, Rachit Nimavat, Max Ovsiankin, Ankita Pasad, Kumar Kshitij Patel, Kavya Ravichandran, Pedro Pamplona Savarese, Shashank Srivastava, Kevin Stangl, Akilesh Tangela, Shubham Toshniwal, and David Yunis.

I would like to thank past PhD students at TTIC senior to me, who generously shared with me, early in my PhD, their experiences and lessons learned. This was invaluable to me. I especially want to thank: Charalampos Angelidakis, Mrinalkanti Ghosh, Behnam Neyshabur, Shubhendu Trivedi, and Blake Woodworth.

I am grateful to officers and administrative staff at TTIC: Adam Bohlander, Erica Cocom, Chrissy Coleman, Jessica Jacobson, Brandie Jones, Mary Marre, and Amy Minick, who saved me a lot of time and made my life easy in so many ways.

I am thankful to the Simons institute and the orgaizers of the Foundations of Deep Learning program for a truly wonderful summer at Berkeley in 2019. Participating in this program has allowed me to meet and interact with many amazing researchers in the field. I would also like to thank Suriya Gunasekar and the Machine Learning Foundations group at MSR Redmond for hosting me for a (remote) internship during Summer 2020. It was a great opportunity that I learned a lot from.

Outside of academia, I owe many thanks to the Libyan community in Chicago at large. I have greatly enjoyed our occasional social gatherings and activities, which has provided me with much needed comfort while being away from home. I would like to especially thank Sufyan Zeinuba for being a wonderful friend in times of hardship and ease, and for the many fun adventures in Chicago and beyond. I would also like to thank Abdelrhman Suleimani for providing a lot of support and advice in my last year on various matters.

I would like to thank my dear friends Mohammad Bara and Mohammad Benhassan for their continuous support, for always being there in the good and the bad, and for the memorable fun trips. I also thank my dear friend Kareem Khalil, my companion in undergrad at Penn State and who is probably one of the main people that got me to seriously consider doing a PhD.

Aysha—I will thank you in person.

I would like to thank my bigger family in Libya: my grandmother, my aunts, my uncle, and my cousins. You have always provided me with so much unconditional love and support. I am truly grateful for you. I am especially grateful for my late aunt, Fatima, whom I dearly miss.

To my mother, father, and brothers, my biggest fans since day one, I am deeply grateful for your never wavering support and enthusiasm in whatever I venture in. I thank you for your patience for all the time I have spent away from you. There are no words I can write that can even begin to describe what I owe to you.

# Contents

# Listing of figures

# 1
# Introduction

Despite extraordinary progress, current machine learning (ML) systems have been shown to be brittle against different notions of robustness—such as adversarial examples and distributional shifts—in various applications (e.g., medical imaging and self-driving cars). The modern ever-growing reliance on ML necessitates algorithms and models that perform well against unknown unintended conditions that arise either adversarially or naturally.

Learning models and predictors robust to *adversarial examples* is a major contemporary challenge in ML. Adversarial examples can be thought of as carefully crafted perturbations of test examples that cause predictors to miss-classify. For example, adding imperceptible noise to images to fool image recognition systems (Szegedy et al., 2013, Biggio et al., 2013) or attaching small adversarial stickers to physical traffic signs to fool self-driving cars (Eykholt et al., 2018). There has been a significant interest lately in how deep neural networks are *not* robust to adversarial examples (Szegedy et al., 2013, Biggio et al., 2013, Goodfellow et al., 2015a), leading to an ongoing effort to devise methods for learning models that *are* adversarially robust.

This thesis makes contributions that address, from a theoretical perspective, this robustness challenge of adversarial examples—by exploring what robustness properties can we hope to guarantee and understanding how to guarantee them. The traditional theory of machine learning does not explain when is it possible to learn robustly, nor does it suggest algorith-mic principles that offer robustness guarantees. Developing such a theoretical understanding of relevant robustness properties has the potential to fundamentally change the modern practice of machine learning. To this end, this thesis introduces problem formulations that capture aspects of emerging practical challenges in adversarially robust learning. Following that, the results and contributions range from designing new learning algorithms with provable robustness guarantees, to understanding and characterizing inherent limitations on the performance of any algorithm.

The Problem. In adversarially robust learning, the main goal is to learn from training examples a model or a predictor $\hat{h}$ that *generalizes robustly* to future unseen test examples as measured by the *robust* risk: the probability that predictors $\hat{h}$ makes a classification mistake on a randomly drawn test example $x$ or on any perturbation of it in a generic perturbation set $\mathcal{U}(x)$. That is, $\mathcal{U}(x)$ represents the set of possible perturbations of instance $x$ that an adversary might attack with at test-time and we would like to give correct classifications under these attacks. For example, $\mathcal{U}$ can be perturbations of bounded $\ell_\infty$ norm which represents "imperceptible" image perturbations in practice (Goodfellow, Shlens, and Szegedy, 2015a).

Formally, given an instance space $\mathcal{X}$ and label space $\mathcal{Y} = \{\pm 1\}$, we would like to be robust against a perturbation set $\mathcal{U} : \mathcal{X} \to 2^{\mathcal{X}}$, where $\mathcal{U}(x) \subseteq \mathcal{X}$ represents the set of perturbations (adversarial examples) that can be chosen by an adversary at test-time. For example, $\mathcal{U}$ could be perturbations of distance at most $\gamma$ w.r.t. some metric $\rho$, such as the $\ell_\infty$ metric considered in many applications: $\mathcal{U}(x) = \{z \in \mathcal{X} : \|x - z\|_\infty \leq \gamma\}$. Our only (implicit) restriction on the specification of $\mathcal{U}$ is that $\mathcal{U}(x)$ should be nonempty for every $x$. For an unknown distribution $\mathcal{D}$ over $\mathcal{X} \times \mathcal{Y}$, we observe $m$ i.i.d. samples $S \sim \mathcal{D}^m$, and our goal is to learn a predictor $\hat{h} : \mathcal{X} \to \mathcal{Y}$ having small robust risk,

$$\mathrm{R}_{\mathcal{U}}(\hat{h}; \mathcal{D}) := \mathbb{E}_{(x,y)\sim\mathcal{D}} \left[ \sup_{z \in \mathcal{U}(x)} 1[\hat{h}(z) \neq y] \right].$$

## 1.1 Overview of Thesis Contributions and Structure

The central theme of this thesis is to

*develop theoretical foundations for robust machine learning.*

Toward this, we make contributions that can be categorized in the following ways: (1) introducing problem formulations capturing aspects of emerging practical challenges in robust learning, (2) designing new learning algorithms with provable robustness guarantees under different settings, and (3) characterizing the complexity of robust learning and fundamental limitations on the performance of any algorithm.

We next provide a high-level summary of the main contributions of this thesis.

Chapter 2: Background and Problem Setup. We provide formal definitions for the problem of adversarially robust learning (Definition 2.1 and Definition 2.2), and we review some background material such as PAC learning, VC dimension, and uniform convergence.

Chapter 3: Improper vs. Proper Learning. The common approach to adversarially robust learning is to pick a class $\mathcal{H}$ of predictors (e.g., neural networks with some architecture) and learn through minimization of the *empirical* robust risk over $\mathcal{H}$. That is, optimizing model parameters to fit the training data and its adversarial examples which is popularly known as *adversarial training* in practice (Madry et al., 2018). We refer to this as robust *empirical* risk minimization:

$$\hat{h} \in \mathsf{RERM}_{\mathcal{H}}(S) := \operatorname*{argmin}_{h\in\mathcal{H}} \frac{1}{|S|} \sum_{(x,y)\in S} \sup_{z\in\mathcal{U}(x)} \mathbb{1}[h(z) \neq y]. \tag{1.1}$$

This approach is justified theoretically by uniform convergence of the robust risk—i.e., the promise that, with large enough training data, any model in the class $\mathcal{H}$ that achieves small robust risk on the training data will have small robust risk on future test examples (Cullina et al., 2018a, Bubeck et al., 2019, Yin et al., 2019).

In Chapter 3, we surprisingly reveal, however, that optimizing model parameters to minimize the robust risk on the training data (as considered in adversarial training in practice) or any surrogate loss can *fail* spectacularly (Theorem 3.1). We show that this failure actually extends to a broader family of algorithms: any procedure that chooses a model from the model class $\mathcal{H}$ (also known as *proper* learning). Unlike in standard supervised learning where empirical risk minimization—i.e., optimizing model parameters to fit the training data—is always sufficient for (non-robust) generalization, our result suggests that we need new algorithmic principles that depart from empirical (robust) risk minimization, and from *proper* learning more generally, for provable robustness guarantees. To this end, our main contribution, Theorem 3.4, is a new *improper* learning algorithm which trains an *ensemble* of models on adaptively chosen subsets of the training data and makes predictions by taking their majority vote, which provably provides much stronger robustness guarantees.

Chapter 4: An Optimal Learner and a Characterization. When is adversarially robust learning possible and when is it not? and how can we learn optimally? Characterizing learnability is one of the most basic problems of statistical learning theory. Many fundamental learning problems can be surprisingly characterized by means of combinatorial complexity measures, which are often quantitatively insightful in that they provide tight bounds on the number of examples needed for learning and also insightful for algorithm design. For example, in standard supervised learning, the fundamental theorem of statistical learning (Vapnik and Chervonenkis, 1971, 1974, Blumer, Ehrenfeucht, Haussler, and Warmuth, 1989a, Ehrenfeucht, Haussler, Kearns, and Valiant, 1989) provides a complete understanding of *what* is learnable: model classes $\mathcal{H}$ with finite Vapnik-Chervonenkis (VC) dimension, and *how* to learn: by the generic principle of empirical risk minimization.

In Chapter 4, we put forward a theory that precisely characterizes the complexity of robust learnability. Interestingly, we identify a broad class of algorithms that are sub-optimal for adversarially robust learning in a strong negative sense (Theorem 4.1). To go beyond the limitations of this class, we contribute new algorithmic ideas that depart from traditional approaches and principles. In particular, we propose a structural characterization—based on the model class $\mathcal{H}$ and the perturbation set $\mathcal{U}$—that explains when adversarially robust learning is possible and when not (Theorem 4.3), and tightly characterizes the optimal sample complexity (Theorem 4.4). Beyond this, we derive a generic and statistically optimal learner for adversarially robust learning based on our structural characterization (Theorem 4.2).

Chapter 5: Boosting Robustness. Adversarially robust learning has proven to be quite challenging in practice, where current methods typically learn models with low error on i.i.d. test examples but robust only on a small fraction of them. For example, on two standard image classification benchmarks (`CIFAR10` and `ImageNet`), the highest achieved *robust* accuracy is about 66% and 38%, respectively (Croce et al., 2020). When and how is it possible to leverage existing methods and go beyond their limits?

In Chapter 5, we explore when and how it is possible to algorithmically *boost* robustness of existing algorithms. That is, given a *barely* robust learning algorithm $\mathbb{A}$ which can only learn models robust on say 10% fraction of the data distribution, we ask whether it is possible to *boost* the robustness of $\mathbb{A}$ and learn predictors with high *robust* accuracy, say 90%. Beyond contributing a precise mathematical formulation of barely robust learning (Definition 5.1), we contribute a novel and simple algorithm for *boosting* robustness (Theorem 5.3); thereby showing that barely robust learning implies strongly robust learning. We also show that our proposed notion of barely robust learning is *necessary* for strongly robust learning (Theorem 5.9), and that weaker relaxations of it do not imply strongly robust learning (Theorem 5.11). This shows a qualitative and quantitative *equivalence* between two seemingly different problems: barely robust learning and strongly robust learning. Our results offer a new perspective on adversarial robustness; revealing an interesting landscape for boosting robustness with connections to classical pioneering work on boosting accuracy (Kearns, 1988, Schapire, 1990, Freund, 1990, Freund and Schapire, 1997). In particular, we achieve even stronger quantitative boosting guarantees by composing our algorithm for boosting robustness with classical algorithms for boosting accuracy (Corollary 5.4), such as AdaBoost (Freund and Schapire, 1997).

### Reductions in Adversarial Learning

Motivated by practical considerations, a core theme in this thesis is to study and understand, from a theoretical perspective, what kind of accesses and reductions can allow for robust learning guarantees. By identifying and formulating appropriate modes of access to adversarial perturbations, this thesis takes first steps towards building machine learning systems robust to abstract perturbations such as "imperceptible" perturbations (Laidlaw, Singla, and Feizi, 2021), going beyond simple $\ell_p$ perturbations (Goodfellow, Shlens, and Szegedy, 2015a) popular in practice. I summarize below contributions towards designing modular and generic robust learning algorithms.

**Chapter 6: Reductions to Non-Robust Learners.** Many machine learning systems in practice perform standard supervised learning but lack robustness guarantees. It would be beneficial to provide wrapper procedures that can guarantee adversarial robustness in a black-box manner without needing to modify current systems internally. In Chapter 6, we formally study this question and propose a reduction algorithm (a wrapper procedure) that can robustly learn, under appropriate conditions, any model class $\mathcal{H}$ using any black-box non-robust learning algorithm $\mathbb{A}$ for $\mathcal{H}$ (Theorem 6.2) in the realizable setting. We further show that the oracle complexity (number of calls to $\mathbb{A}$) of this reduction is near-optimal (Theorem 6.6). Finally, for the agnostic setting, we present a different reduction algorithm that specifically requires an ERM oracle for $\mathcal{H}$, with slightly weaker quantitative guarantees (Theorem 6.7).

Chapter 7: Robustness to Unknown Perturbations. As an emerging direction from the work above, in Chapter 7, we initiate a quest to find the "right" model of access to adversarial perturbations $\mathcal{U}$ by considering different forms of access and studying the robustness guarantees achievable in each case. We propose generic robust learning algorithms that work for any perturbation set $\mathcal{U}$, when given access to specific "attack procedures" for $\mathcal{U}$ (Section 7.3). Interestingly, we use black-box *online learning* algorithms to achieve this, which we show is *necessary* (Theorem 7.10). Furthermore, our algorithms can achieve robustness guarantees against multiple perturbation sets concurrently (which is of practical interest: Kang et al., 2019a, Tramèr and Boneh, 2019, Maini et al., 2020) when given access to separate attack oracles for them.

Chapter 8: Computationally Efficient Robustness. Continuing with the theme of providing robustness guarantees for abstract perturbations $\mathcal{U}$, in Chapter 8, we explore characterizing which sets $\mathcal{U}$ admit *computationally efficient* robust learners. We focus on the fundamental problem of learning adversarially robust linear predictors. We propose an efficient algorithm for finding *robust* linear and kernel predictors for a broad range of perturbation sets $\mathcal{U}$, when the data is robustly separable with a linear predictor (Corollary 8.6). Interestingly, our algorithm is an instantiation of robust learning algorithms that require only an attack oracle for $\mathcal{U}$ (discussed above). We show that an attack oracle for $\mathcal{U}$ can be efficiently implemented using a separation oracle for $\mathcal{U}$ which separates between perturbations and non-perturbations (Lemma 8.9). We also show that an efficient approximate separation oracle for $\mathcal{U}$ is *necessary* to even compute the robust loss of a linear predictor (Theorem 8.10).

Chapter 9: Transductive Robustness. In many applications in practice, test examples are available in batches and an ML system is tasked with classifying them all at once, which is known as *transductive* learning (Vapnik, 1998). In Chapter 9, we propose a *simple* transductive learning algorithm that when given a set of labeled training examples and a set of unlabeled test examples (*both* sets possibly *adversarially perturbed*), it correctly labels the test examples with an error rate that is *exponentially* smaller than the best known guarantee in the inductive setting (see Section 9.3). This result highlights the (potentially) stronger robustness guarantees that transductive learning may offer.

Chapter 10: Beyond Perturbations. Unlike our in-depth exploration of adversarial perturbations in previous chapters, Chapter 10 studies learning guarantees when the test examples are *arbitrary*. This includes *both* the possibility that test examples are chosen by an *unrestricted* adversary (i.e., not limited to a specific perturbation set) or that they are drawn from a distribution different from the training distribution (sometimes called "covariate shift").

While it may seem impossible to learn with arbitrary test examples, we surprisingly show that it is ~~im~~possible when we classify test examples in *batches* ("transductive learning") and when we are allowed to *abstain* from classifying some examples ("selective classification") (Definition 10.1). We provide an efficient transductive selective classification algorithm that incurs low abstention and miss-classification error rates (Theorem 10.5 and Theorem 10.6). These are the first nontrivial guarantees for learning with arbitrary train and test distributions—no prior guarantees were known even for simple functions such as intervals on the line. Even the simple approach of training a classifier to distinguish unlabeled train vs. test examples may be adequate in some applications, though for theoretical guarantees one requires somewhat more sophisticated algorithms.

## 1.2 Bibliographical Remarks

The research presented in this thesis is based on published joint work with a few co-authors. Chapter 3 is based on Montasser, Hanneke, and Srebro (2019). Chapter 4 is based on Montasser, Hanneke, and Srebro (2022b). Chapter 5 is based on Blum, Montasser, Shakhnarovich, and Zhang (2022). Chapter 6 is based on Montasser, Hanneke, and Srebro (2020b) and Ahmadi, Blum, Montasser, and Stangl (2023). Chapter 7 is based on Montasser, Hanneke, and Srebro (2021). Chapter 8 is based on Montasser, Goel, Diakonikolas, and Srebro (2020a). Chapter 9 is based on Montasser, Hanneke, and Srebro (2022a). Chapter 10 is based on Goldwasser, Kalai, Kalai, and Montasser (2020).

# 2

# Background and Problem Setup

Given an instance space $\mathcal{X}$ and label space $\mathcal{Y} = \{\pm 1\}$, we formalize a perturbation set we would like to be robust against as a map $\mathcal{U} : \mathcal{X} \to 2^{\mathcal{X}}$, where $\mathcal{U}(x) \subseteq \mathcal{X}$ represents the set of perturbations (adversarial examples) that can be chosen by an adversary at test-time. For example, $\mathcal{U}$ could be perturbations of distance at most $\gamma$ w.r.t. some metric $\rho$, such as the $\ell_\infty$ metric considered in many applications: $\mathcal{U}(x) = \{z \in \mathcal{X} : \|x - z\|_\infty \leq \gamma\}$. Our only (implicit) restriction on the specification of $\mathcal{U}$ is that $\mathcal{U}(x)$ should be nonempty for every $x$. For an unknown distribution $\mathcal{D}$ over $\mathcal{X} \times \mathcal{Y}$, we observe $m$ i.i.d. samples $S \sim \mathcal{D}^m$, and our goal is to learn a predictor $\hat{h} : \mathcal{X} \to \mathcal{Y}$ having small robust risk,

$$\mathrm{R}_{\mathcal{U}}(\hat{h}; \mathcal{D}) := \mathbb{E}_{(x,y)\sim\mathcal{D}}\left[\sup_{z\in\mathcal{U}(x)} \mathbb{1}[\hat{h}(z) \neq y]\right]. \tag{2.1}$$

Given a hypothesis class $\mathcal{H} \subseteq \mathcal{Y}^{\mathcal{X}}$, our goal is to design a learning rule $\mathbb{A} : (\mathcal{X} \times \mathcal{Y})^* \to \mathcal{Y}^{\mathcal{X}}$ such that for any distribution $\mathcal{D}$ over $\mathcal{X} \times \mathcal{Y}$, the rule $\mathbb{A}$ will find a predictor that competes with the best predictor $h^* \in \mathcal{H}$ in terms of the robust risk using a number of samples that is independent of the distribution $\mathcal{D}$. The following definitions formalize the notion of robust PAC learning in the realizable and agnostic settings:*

*We implicitly suppose that the hypotheses $h$ in $\mathcal{H}$ and their losses $\sup_{z\in\mathcal{U}(x)} \mathbb{1}[h(z) \neq y]$ are measurable,

**Definition 2.1** (Agnostic Robust PAC Learnability). *For any* $\varepsilon, \delta \in (0,1)$, *the* sample complexity of agnostic robust $(\varepsilon,\delta)-$PAC learning of $\mathcal{H}$ with respect to adversary $\mathcal{U}$, *denoted* $\mathcal{M}^{\mathrm{ag}}_{\varepsilon,\delta}(\mathcal{H},\mathcal{U})$, *is defined as the smallest* $m \in \mathbb{N} \cup \{0\}$ *for which there exists a learning rule* $\mathbb{A} : (\mathcal{X} \times \mathcal{Y})^* \to \mathcal{Y}^{\mathcal{X}}$ *such that, for every data distribution* $\mathcal{D}$ *over* $\mathcal{X} \times \mathcal{Y}$, *with probability at least* $1-\delta$ *over* $S \sim \mathcal{D}^m$,

$$\mathrm{R}_{\mathcal{U}}(\mathbb{A}(S);\mathcal{D}) \leq \inf_{h\in\mathcal{H}} \mathrm{R}_{\mathcal{U}}(h;\mathcal{D}) + \varepsilon.$$

*If no such* $m$ *exists, define* $\mathcal{M}^{\mathrm{ag}}_{\varepsilon,\delta}(\mathcal{H},\mathcal{U}) = \infty$. *We say that* $\mathcal{H}$ *is robustly PAC learnable in the agnostic setting with respect to adversary* $\mathcal{U}$ *if* $\forall \varepsilon, \delta \in (0,1)$, $\mathcal{M}^{\mathrm{ag}}_{\varepsilon,\delta}(\mathcal{H},\mathcal{U})$ *is finite.*

**Definition 2.2** (Realizable Robust PAC Learnability). *For any* $\varepsilon, \delta \in (0,1)$, *the* sample complexity of realizable robust $(\varepsilon,\delta)$-PAC learning of $\mathcal{H}$ with respect to adversary $\mathcal{U}$, *denoted* $\mathcal{M}^{\mathrm{re}}_{\varepsilon,\delta}(\mathcal{H},\mathcal{U})$, *is defined as the smallest* $m \in \mathbb{N} \cup \{0\}$ *for which there exists a learning rule* $\mathbb{A} : (\mathcal{X} \times \mathcal{Y})^* \to \mathcal{Y}^{\mathcal{X}}$ *such that, for every data distribution* $\mathcal{D}$ *over* $\mathcal{X} \times \mathcal{Y}$ *where there exists a predictor* $h^* \in \mathcal{H}$ *with zero robust risk,* $\mathrm{R}_{\mathcal{U}}(h^*;\mathcal{D}) = 0$, *with probability at least* $1-\delta$ *over* $S \sim \mathcal{D}^m$,

$$\mathrm{R}_{\mathcal{U}}(\mathbb{A}(S);\mathcal{D}) \leq \varepsilon.$$

*If no such* $m$ *exists, define* $\mathcal{M}^{\mathrm{re}}_{\varepsilon,\delta}(\mathcal{H},\mathcal{U})) = \infty$. *We say that* $\mathcal{H}$ *is robustly PAC learnable in the realizable setting with respect to adversary* $\mathcal{U}$ *if* $\forall \varepsilon, \delta \in (0,1)$, $\mathcal{M}^{\mathrm{re}}_{\varepsilon,\delta}(\mathcal{H},\mathcal{U})$ *is finite.*

**Definition 2.3** (Proper & Improper Learnability). *We say that* $\mathcal{H}$ *is* properly *robustly PAC learnable (in the agnostic or realizable setting) if it can be learned as in Definitions 2.1 or 2.2 using a learning rule* $\mathbb{A} : (\mathcal{X} \times \mathcal{Y})^* \to \mathcal{H}$ *that always outputs a predictor in* $\mathcal{H}$. *We refer to learning using any learning rule* $\mathbb{A} : (\mathcal{X} \times \mathcal{Y})^* \to \mathcal{Y}^{\mathcal{X}}$, *as in the definitions above, as* improper *learning.*

Notation. We denote $\mathrm{err}(h;\mathcal{D}) = \mathbb{P}(h(x) \neq y)$, the (non-robust) error rate under the 0-1 loss, and $\hat{\mathrm{err}}(h;S) = \frac{1}{|S|}\sum_{(x,y)\in S} 1[h(x) \neq y]$ the empirical error rate. These agree with the robust variant when $\mathcal{U}(x) = \{x\}$, and so robust learnability agrees with standard supervised learning when $\mathcal{U}(x) = \{x\}$.

---

and that standard mild restrictions on $\mathcal{H}$ are imposed to guarantee measurability of empirical processes, so that the standard tools of VC theory apply. See Blumer, Ehrenfeucht, Haussler, and Warmuth (1989a), van der Vaart and Wellner (1996) for discussion of such measurability issues, which we will not mention again in the remainder of this thesis.

## 2.1 Standard PAC Learnability and the Vapnik-Chervonenkis (VC) Dimension

**Definition 2.4** (PAC Learnability). *A target hypothesis class $\mathcal{H} \subseteq \mathcal{Y}^{\mathcal{X}}$ is said to be PAC learnable if there exists a learning rule $\mathbb{A} : (\mathcal{X} \times \mathcal{Y})^* \to \mathcal{Y}^{\mathcal{X}}$ with sample complexity $m(\varepsilon, \delta) : (0,1)^2 \to \mathbb{N}$ such that: for any $\varepsilon, \delta \in (0,1)$, for any distribution $\mathcal{D}$ over $\mathcal{X} \times \mathcal{Y}$, and any target concept $h \in \mathcal{H}$ with zero risk, $\mathrm{err}_{\mathcal{D}}(h) = 0$, with probability at least $1 - \delta$ over $S \sim \mathcal{D}^{m(\varepsilon,\delta)}$,*

$$\mathrm{err}_{\mathcal{D}}(\mathbb{A}(S)) \triangleq \Pr_{(x,y)\sim\mathcal{D}} [\mathbb{A}(S)(x) \neq y] \leq \varepsilon.$$

We recall the Vapnik-Chervonenkis dimension (VC dimension) is defined as follows,

**Definition 2.5** (VC dimension). *We say that a sequence $\{x_1, \dots, x_k\} \in \mathcal{X}$ is shattered by $\mathcal{H}$ if $\forall y_1, \dots, y_k \in \mathcal{Y}, \exists h \in \mathcal{H}$ such that $\forall i \in [k], h(x_i) = y_i$. The VC dimension of $\mathcal{H}$ (denoted $\mathrm{vc}(\mathcal{H})$) is then defined as the largest integer $k$ for which there exists $\{x_1, \dots, x_k\} \in \mathcal{X}$ that is shattered by $\mathcal{H}$. If no such $k$ exists, then $\mathrm{vc}(\mathcal{H})$ is said to be infinite.*

In the standard PAC learning framework, we know that a hypothesis class $\mathcal{H}$ is PAC learnable if and only if the VC dimension of $\mathcal{H}$ is finite (Vapnik and Chervonenkis, 1971, 1974, Blumer, Ehrenfeucht, Haussler, and Warmuth, 1989a, Ehrenfeucht, Haussler, Kearns, and Valiant, 1989). In particular, $\mathcal{H}$ is properly PAC learnable with $\mathsf{ERM}_{\mathcal{H}}$ and therefore proper learning is sufficient for supervised learning.

Another important complexity measure that is utilized in the study of robust PAC learning is the notion of *dual* VC dimension, which we define below:

**Definition 2.6** (Dual VC dimension). *Given a hypothesis class $\mathcal{H} \subseteq \mathcal{Y}^{\mathcal{X}}$, consider a* dual space *$\mathcal{G}$: a set of functions $g_x : \mathcal{H} \to \mathcal{Y}$ defined as $g_x(h) = h(x)$, for each $h \in \mathcal{H}$ and each $x \in \mathcal{X}$. Then, the dual VC dimesion of $\mathcal{H}$ (denoted $\mathrm{vc}^*(\mathcal{H})$) is defined as the VC dimension of $\mathcal{G}$. In other words, $\mathrm{vc}^*(\mathcal{H}) = \mathrm{vc}(\mathcal{G})$ and it represents the largest set $\{h_1, \dots, h_k\}$ that is shattered by points in $\mathcal{X}$.*

If the VC dimension is finite, then so is the dual VC dimesion, and it can be bounded as $\mathrm{vc}^*(\mathcal{H}) < 2^{\mathrm{vc}(\mathcal{H})+1}$ (Assouad, 1983). Although this exponential dependence is tight for some classes, for many natural classes, such as linear predictors and some neural networks (see, e.g. Lemma 6.3), the primal and dual VC dimensions are equal, or at least polynomially related.

## 2.2 Uniform Convergence of the Robust Loss

Adversarially robust learning is a special case of Vapnik's "General Learning" (Vapnik, 1982), but can not, in general, be phrased in terms of supervised learning of some modified hypothesis class or loss. By relying on Vapnik's "General Learning" (Vapnik, 1982), we can identify a necessary and sufficient condition for uniform convergence of the robust loss. Specifically, denote by $\mathcal{L}^{\mathcal{U}}_{\mathcal{H}}$ the robust loss class of $\mathcal{H}$, defined as:

$$\mathcal{L}^{\mathcal{U}}_{\mathcal{H}} = \left\{ (x,y) \mapsto \sup_{z \in \mathcal{U}(x)} 1[h(z) \neq y] : h \in \mathcal{H} \right\}. \tag{2.2}$$

Based on Vapnik's "General Learning" (Vapnik, 1982), the VC dimension of the robust loss class $\mathcal{L}^{\mathcal{U}}_{\mathcal{H}}$, denoted $\text{vc}(\mathcal{L}^{\mathcal{U}}_{\mathcal{H}})$, is the quantity that controls uniform convergence of the robust loss. Formally, we have the following guarantee: for any $\varepsilon, \delta \in (0,1)$ and any distribution $\mathcal{D}$ over $\mathcal{X} \times \mathcal{Y}$, with probability at least $1-\delta$ over $S \sim \mathcal{D}^{m(\varepsilon,\delta)}$,

$$\forall h \in \mathcal{H} : |\mathrm{R}_{\mathcal{U}}(h;\mathcal{D}) - \mathrm{R}_{\mathcal{U}}(h;S)| \leq \varepsilon, \text{ where } m(\varepsilon,\delta) = O\left(\frac{\text{vc}(\mathcal{L}^{\mathcal{U}}_{\mathcal{H}}) + \log(1/\delta)}{\varepsilon^2}\right).$$

Uniform convergence of the robust loss gives us a sufficient condition for learnability. Namely, if the robust loss class $\mathcal{L}^{\mathcal{U}}_{\mathcal{H}}$ has finite VC dimension ($\text{vc}(\mathcal{L}^{\mathcal{U}}_{\mathcal{H}}) < \infty$), then $\mathcal{H}$ is robustly PAC learnable using $\mathsf{RERM}_{\mathcal{H}}$ (see Equation 3.2) with sample complexity $O\left(\frac{\text{vc}(\mathcal{L}^{\mathcal{U}}_{\mathcal{H}})+\log(1/\delta)}{\varepsilon^2}\right)$. One might then wish to relate the VC dimension of the hypothesis class, $\text{vc}(\mathcal{H})$, to the VC dimension of the robust loss class, $\text{vc}(\mathcal{L}^{\mathcal{U}}_{\mathcal{H}})$. It turns out that the relationship between the two depends on whether the set of the perturbations $\mathcal{U}$ is finite or not.

When $\mathcal{U}$ is finite, i.e., when $|\mathcal{U}| \triangleq \sup_{x \in \mathcal{X}} |\mathcal{U}(x)| < \infty$, we can bound from above the VC dimension of the robust loss class $\text{vc}(\mathcal{L}^{\mathcal{U}}_{\mathcal{H}})$, in terms of the VC dimension of the hypothesis class $\text{vc}(\mathcal{H})$ and log-cardinality of $\mathcal{U}$ as in the following lemma:

**Lemma 2.7** (VC Dimension of the Robust Loss Attias, Kontorovich, and Mansour (2022))**.** *For any class $\mathcal{H}$ and any $\mathcal{U}$ where $|\mathcal{U}| \triangleq \sup_{x \in \mathcal{X}} |\mathcal{U}(x)| < \infty$, it holds that* $\text{vc}(\mathcal{L}^{\mathcal{U}}_{\mathcal{H}}) \leq$

$O(\text{vc}(\mathcal{H}) \log |\mathcal{U}|),.$

*Proof.* Let $k = |\mathcal{U}|$. By finiteness of $\mathcal{U}$, observe that for any dataset $S \in (\mathcal{X} \times \mathcal{Y})^m$, each robust loss vector in the set of robust loss behaviors:

$$\Pi_{\mathcal{L}^{\mathcal{U}}_{\mathcal{H}}}(S) = \left\{ (f(x_1, y_1), \ldots, f(x_m, y_m)) : f \in \mathcal{L}^{\mathcal{U}}_{\mathcal{H}} \right\}$$

maps to a 0-1 loss vector on the *inflated set*

$$S_{\mathcal{U}} = \left\{ (z_1^1, y_1), \ldots, (z_1^k, y_1), (z_2^1, y_2), \ldots, (z_2^k, y_2), \ldots, (z_m^1, y_m), \ldots, (z_m^k, y_m) \right\},$$

$$\Pi_{\mathcal{H}}(S_{\mathcal{U}}) = \left\{ (h(z_1^1), \ldots, h(z_1^k), h(z_2^1), \ldots, h(z_2^k), \ldots, h(z_m^1), \ldots, h(z_m^k)) : h \in \mathcal{H} \right\}.$$

Therefore, it follows that $\left| \Pi_{\mathcal{L}^{\mathcal{U}}_{\mathcal{H}}}(S) \right| \leq |\Pi_{\mathcal{H}}(S_{\mathcal{U}})|$. Then, by applying the Sauer-Shelah lemma, it follows that $|\Pi_{\mathcal{H}}(S_{\mathcal{U}})| \leq O((mk)^{\text{vc}(\mathcal{H})})$. Then, by solving for $m$ such that $O((mk)^{\text{vc}(\mathcal{H})}) \leq 2^m$, we get that $\text{vc}(\mathcal{L}^{\mathcal{U}}_{\mathcal{H}}) \leq O(\text{vc}(\mathcal{H}) \log(k))$. □

However, when $\mathcal{U}$ is infinite, and this includes for example $\mathcal{U}$ being an $\ell_p$ ball with some radius (which is of practical interest), then $\text{vc}(\mathcal{L}^{\mathcal{U}}_{\mathcal{H}})$ can be arbitrarily larger than $\text{vc}(\mathcal{H})$. A construction witnessing this separation was first given in (Cullina, Bhagoji, and Mittal, 2018a, Theorem 3).

In Chapter 3, Section 3.2, Lemma 3.2, we independently give a different construction which appeared in (Montasser et al., 2019). Later on in Section 3.2, we build on this construction to show that there are classes $\mathcal{H}$ with $\text{vc}(\mathcal{H}) = 1$ that do NOT enjoy any uniform convergence guarantees for the robust loss, and $\mathsf{RERM}_{\mathcal{H}}$ might not ensure robust learning, while the problem is still robustly learnable with a different (improper, in our case) learning rule. Thus, while uniform convergence is sufficient for robust learnability, it is not actually necessary!

As mentioned earlier, for supervised learning finite VC dimension of the loss class (which is equal to the VC dimension of the hypothesis class) is also necessary for learning. For general learning, unlike supervised learning, the loss class having finite VC dimension, and uniform convergence over this class, is not, in general, necessary, and rules other than ERM might be needed for learning (e.g. Vapnik, 1982, Shalev-Shwartz, Shamir, Srebro, and Sridharan, 2009, Daniely, Sabato, Ben-David, and Shalev-Shwartz, 2015).

# 3

# VC Classes are Adversarially Robustly Learnable, but Only Improperly

## 3.1 Introduction

Learning predictors that are robust to adversarial perturbations is an important challenge in contemporary machine learning. There has been a lot of interest lately in how predictors learned by deep learning are *not* robust to adversarial examples (Szegedy, Zaremba, Sutskever, Bruna, Erhan, Goodfellow, and Fergus, 2013, Biggio, Corona, Maiorca, Nelson, Šrndić, Laskov, Giacinto, and Roli, 2013, Goodfellow, Shlens, and Szegedy, 2014), and there is an ongoing effort to devise methods for learning predictors that *are* adversarially robust. In this chapter, we consider the problem of learning, based on a (non-adversarial) i.i.d. sample, a predictor that is robust to adversarial examples at test time. We emphasize that this is distinct from the learning process itself being robust to an adversarial training set.

Given an instance space $\mathcal{X}$ and label space $\mathcal{Y} = \{\pm 1\}$, we formalize a perturbation set we would like to protect against as $\mathcal{U} : \mathcal{X} \to 2^{\mathcal{X}}$, where $\mathcal{U}(x) \subseteq \mathcal{X}$ represents the set of perturbations (adversarial examples) that can be chosen by the adversary at test time. For example, $\mathcal{U}$ could be perturbations of distance at most $\gamma$ w.r.t. some metric $\rho$, such as the $\ell_\infty$ metric considered in many applications: $\mathcal{U}(x) = \{z \in \mathcal{X} : \|x - z\|_\infty \leq \gamma\}$. Our only (implicit) restriction on the specification of $\mathcal{U}$ is that $\mathcal{U}(x)$ should be nonempty for every $x$. For a distribution $\mathcal{D}$ over $\mathcal{X} \times \mathcal{Y}$, we observe $m$ i.i.d. samples $S \sim \mathcal{D}^m$, and our goal is to learn a predictor $\hat{h} : \mathcal{X} \mapsto \mathcal{Y}$ having small robust risk,

$$\mathrm{R}_{\mathcal{U}}(\hat{h}; \mathcal{D}) := \mathbb{E}_{(x,y)\sim\mathcal{D}} \left[ \sup_{z \in \mathcal{U}(x)} \mathbb{1}[\hat{h}(z) \neq y] \right]. \tag{3.1}$$

The common approach to adversarially robust learning is to pick a hypothesis class $\mathcal{H} \subseteq \mathcal{Y}^{\mathcal{X}}$ (e.g. neural networks) and learn through robust *empirical* risk minimization:

$$\hat{h} \in \mathsf{RERM}_{\mathcal{H}}(S) := \operatorname*{argmin}_{h \in \mathcal{H}} \hat{\mathrm{R}}_{\mathcal{U}}(h; S). \tag{3.2}$$

where $\hat{\mathrm{R}}_{\mathcal{U}}(h; S) = \frac{1}{|S|} \sum_{(x,y)\in S} \sup_{z \in \mathcal{U}(x)} \mathbb{1}[h(z) \neq y]$. Most work on the problem has focused on computational approaches to solve this empirical optimization problem, or related problems of minimizing a robust version of some surrogate loss instead of the 0/1 loss (Madry, Makelov, Schmidt, Tsipras, and Vladu, 2017, Wong and Kolter, 2018, Raghunathan, Steinhardt, and Liang, 2018a,b). But of course our true objective is not the empirical robust risk $\hat{\mathrm{R}}_{\mathcal{U}}(h; S)$, but rather the population robust risk $\mathrm{R}_{\mathcal{U}}(h; \mathcal{D})$.

How can we ensure that $\mathrm{R}_{\mathcal{U}}(h; \mathcal{D})$ is small? All prior approaches that we are aware of for ensuring adversarially robust generalization are based on uniform convergence, i.e. showing that w.h.p. for all predictors $h \in \mathcal{H}$, the estimation error $|\mathrm{R}_{\mathcal{U}}(h; \mathcal{D}) - \hat{\mathrm{R}}_{\mathcal{U}}(h; S)|$ is small, perhaps for some surrogate loss (Bubeck, Price, and Razenshteyn, 2018, Cullina, Bhagoji, and Mittal, 2018c, Khim and Loh, 2018, Yin, Ramchandran, and Bartlett, 2018). Such approaches justify RERM, and in particular yield M-estimation type *proper* learning rules: we are learning a hypothesis class by choosing a predictor in the class that minimizes some empirical functional. For standard supervised learning we know that proper learning, and specifically ERM, is sufficient for learning, and so it is sensible to limit attention to such methods.

But it has also been observed in practice that the adversarial error does not generalize as well as the standard error, i.e. there can be a large gap between $\mathrm{R}_{\mathcal{U}}(h; \mathcal{D})$ and $\hat{\mathrm{R}}_{\mathcal{U}}(h; S)$ even when their non-robust versions are similar (Schmidt, Santurkar, Tsipras, Talwar, and Madry, 2018). This suggests that perhaps the robust risk does not concentrate as well as the standard risk, and so RERM in adversarially robust learning might not work as well as ERM in stan-dard supervised learning. Does this mean that such problems are not adversarially robustly learnable? Or is it perhaps that proper learners might not be sufficient?

In this chapter we aim to characterize which hypothesis classes are adversarially robustly learnable, and using what learning rules. That is, for a given hypothesis class $\mathcal{H} \subseteq \mathcal{Y}^{\mathcal{X}}$ and adversary $\mathcal{U}$, we ask whether it is possible, based on an i.i.d. sample to learn a predictor $h$ that has population robust risk almost as good as any predictor in $\mathcal{H}$ (see Definition 2.1 in Chapter 2). We discover a stark contrast between *proper* learning rules which output predictors in $\mathcal{H}$, and *improper* learning rules which are not constrained to predictors in $\mathcal{H}$. Our main results are:

- We show that there exists a perturbation set $\mathcal{U}$ and a hypothesis class $\mathcal{H}$ with finite VC dimension that *cannot* be robustly PAC learned with any *proper* learning rule (including RERM).
- We show that for any adversary $\mathcal{U}$ and any hypothesis class $\mathcal{H}$ with finite VC dimension, there exists an *improper* learning rule that can robustly PAC learn $\mathcal{H}$ (although with sample complexity that is sometimes exponential in the VC dimension).

Our results suggest that we should start considering *improper* learning rules to ensure adversarially robust generalization. They also demonstrate that previous approaches to adversarially robust generalization are not always sufficient, as all prior work we are aware of is based on uniform convergence of the robust risk, either directly for the loss of interest (Bubeck, Price, and Razenshteyn, 2018, Cullina, Bhagoji, and Mittal, 2018c) or some carefully constructed surrogate loss (Khim and Loh, 2018, Yin, Ramchandran, and Bartlett, 2018), which would still justify the use of M-estimation type proper learning. The approach of Attias, Kontorovich, and Mansour (2018) for the case where $|\mathcal{U}(x)| \leq k$ (i.e. finite number of perturbations) is most similar to ours, as it uses an improper learning rule, but their analysis is still based on uniform convergence and so would apply also to RERM (the improperness is introduced only for computational, not statistical, reasons). Also, in this specific case, our approach would give an improved sample complexity that scales only roughly logarithmically with $k$, as opposed to the roughly linear scaling in Attias, Kontorovich, and Mansour (2018)—see discussion at the end of Section 3.3 for details.

A related negative result was presented by Schmidt, Santurkar, Tsipras, Talwar, and Madry (2018), where they showed that there exists a family of distributions (namely, mixtures of two $d$-dimensional spherical Gaussians) where the sample complexity for standard learning is $O(1)$, but the sample complexity for adversarially robust learning is at least $\Omega(\frac{\sqrt{d}}{\log d})$. This an interesting instance where there is a large separation in sample complexity between standard learning and robust learning. But distribution-specific learning is known to be less easily characterizable, with the uniform convergence not being necessary for learning, and ERM not always being optimal, even for standard (non-robust) supervised learning. In this chapter we focus on "worst case" distribution-free robust learning, as in standard PAC learnability.

A different notion of robust learning was studied by Xu and Mannor (2012). They use empirical robustness as a design technique for learning rules, but their goal, and the guarantees they establish are on the standard non-robust population risk, and so do not inform us about robust learnability.

## 3.2 Sometimes There are no Proper Robust Learners

We start by showing that even for hypothesis classes with finite VC dimension, indeed even if $\mathrm{vc}(\mathcal{H}) = 1$, robust PAC learning might not be possible using *any* proper learning rule. In particular, even if there is a robust predictor in $\mathcal{H}$, and even with an unbounded number of samples, RERM (or any other M-estimator or other proper learning rules), will not ensure a low robust risk.

**Theorem 3.1.** *There exists a hypothesis class $\mathcal{H} \subseteq \mathcal{Y}^{\mathcal{X}}$ with $\mathrm{vc}(\mathcal{H}) \leq 1$ and a perturbation set $\mathcal{U}$ such that $\mathcal{H}$ is not properly robustly PAC learnable with respect to $\mathcal{U}$ in the realizable setting.*

This result implies that finite VC dimension of a hypothesis class $\mathcal{H}$ is not sufficient for robust PAC learning if we want to use *proper* learning rules. For the proofs in this section, we will fix an instance space $\mathcal{X} = \mathbb{R}^d$ equipped with a metric $\rho$, and a perturbation set $\mathcal{U} : \mathcal{X} \to 2^{\mathcal{X}}$ such that $\mathcal{U}(x) = \{z \in \mathcal{X} : \rho(x, z) \leq \gamma\}$ for all $x \in \mathcal{X}$ for some $\gamma > 0$. First, we prove a lemma that shows that there exists a hypothesis class $\mathcal{H}$ where there is an arbitrarily large gap between the VC dimension of $\mathcal{H}$ and the VC dimension of the robust loss class of $\mathcal{H}$,

**Lemma 3.2.** *Let $m \in \mathbb{N}$. Then, there exists $\mathcal{H} \subseteq \mathcal{Y}^{\mathcal{X}}$ such that $\mathrm{vc}(\mathcal{H}) \leq 1$ but $\mathrm{vc}(\mathcal{L}^{\mathcal{U}}_{\mathcal{H}}) \geq m$.*

*Proof.* Pick $m$ points $x_1, \dots, x_m$ in $\mathcal{X}$ such that for all $i, j \in [m], \mathcal{U}(x_i) \cap \mathcal{U}(x_j) = \emptyset$. In other words, we want the perturbation sets $\mathcal{U}(x_1), \dots, \mathcal{U}(x_m)$ to be mutually disjoint.

We will construct a hypothesis class $\mathcal{H}$ in the following iterative manner. Initialize set $\mathcal{Z} = \{x_1, \dots, x_m\}$. For each bit string $b \in \{0,1\}^m$, initialize $Z_b = \emptyset$. For each $i \in [m]$, if $b_i = 1$ then pick a point $z \in \mathcal{U}(x_i) \setminus \mathcal{Z}$ and add it to $Z_b$, i.e. $Z_b = Z_b \cup \{z\}$. Once we finish picking points based on all bits that are set to 1, we add $Z_b$ to $\mathcal{Z}$ (i.e. $\mathcal{Z} = \mathcal{Z} \cup Z_b$). We define $h_b : \mathcal{X} \to \mathcal{Y}$ as:

$$h_b(x) = \begin{cases} +1 & \text{if } x \notin Z_b \\ -1 & \text{if } x \in Z_b \end{cases}$$

Then, let $\mathcal{H} = \{h_b : b \in \{0,1\}^m\}$. We can think of each mapping $h_b$ as being characterized by a unique signature $Z_b$ that indicates the points that it labels with $-1$. These points are carefully picked such that, first, they are inside the perturbation sets of $x_1, \dots, x_m$; and second, no two mappings label the same point with $-1$, i.e. for any $b, b' \in \{0,1\}^m$, where $b \neq b'$, $Z_b \cap Z'_b = \emptyset$. Also, we make sure that all mappings in $\mathcal{H}$ label the set $\{x_1, \dots, x_m\}$ with $+1$.

Next, we proceed with proving two claims about $\mathcal{H}$. First, that $\text{vc}(\mathcal{H}) \leq 1$. Pick any two points $z_1, z_2 \in \mathcal{X}$. Consider the following cases. In case $z_1$ or $z_2$ is in $\mathcal{X} \setminus \mathcal{Z}$. Suppose W.L.O.G that $z_2 \in \mathcal{X} \setminus \mathcal{Z}$. Then we know that all mappings label $z_2$ in the same way with label $+1$, because for all $b \in \{0,1\}^m$, $z_2 \notin Z_b$. Therefore, we cannot shatter $z_1, z_2$ with $\mathcal{H}$. In case $z_1$ and $z_2$ are both in $\mathcal{Z}$. Since by our construction, $\mathcal{Z} = \cup_{b \in \{0,1\}^m} Z_b$ and $Z_b \cap Z'_b = \emptyset$ for any $b \neq b'$, we have two sub-cases. Either $z_1, z_2 \in Z_b$ for some $b \in \{0,1\}^m$, which means that the only labelings we can obtain are $(-1,-1)$ with $h_b$, and $(+1,+1)$ with $h'_b$ for any $b' \neq b$. Second case is that $z_1 \in Z_b$ and $z_2 \in Z_{b'}$ for $b \neq b', b, b' \in \{0,1\}^m$. By our construction, we know that we cannot label both points $z_1$ and $z_2$ with $(-1,-1)$, because they don't belong to the same set. Therefore, in both subcases, we cannot shatter $z_1, z_2$ with $\mathcal{H}$. This concludes that $\text{vc}(\mathcal{H}) \leq 1$.

Second, we will show that $\text{vc}(\mathcal{L}^{\mathcal{U}}_{\mathcal{H}}) \geq m$. Consider the set $S = \{(x_1, +), \dots, (x_m, +)\}$. We will show that $\mathcal{L}^{\mathcal{U}}_{\mathcal{H}}$ shatters $S$. Pick any labeling $y \in \{0,1\}^m$. Note that by construction of $\mathcal{H}$, $\exists h_b \in \mathcal{H}$ such that $b = y$. Then, for each $i \in [m]$, $\sup_{z \in \mathcal{U}(x_i)} 1[h_b(z) \neq +1] = b_i = y_i$. This shows that $\mathcal{L}^{\mathcal{U}}_{\mathcal{H}}$ shatters $S$, and therefore $\text{vc}(\mathcal{L}^{\mathcal{U}}_{\mathcal{H}}) \geq m$. □

The following lemma (proof provided in Appendix A.1) establishes that for any sample size $m \in \mathbb{N}$, there exists a hypothesis class $\mathcal{H}$ with $\text{vc}(\mathcal{H}) \leq 1$ such that any *proper* learning rule will fail in learning a robust classifier if it observes at most $m$ samples but not more.

**Lemma 3.3.** *Let $m \in \mathbb{N}$. Then, there exists $\mathcal{H} \subseteq \mathcal{Y}^{\mathcal{X}}$ with $\mathrm{vc}(\mathcal{H}) \leq 1$ such that for any proper learning rule $\mathcal{A} : (\mathcal{X} \times \mathcal{Y})^* \mapsto \mathcal{H}$,*

- *$\exists$ a distribution $\mathcal{D}$ over $\mathcal{X} \times \mathcal{Y}$ and a predictor $h^* \in \mathcal{H}$ where $\mathrm{R}_{\mathcal{U}}(h^*; \mathcal{D}) = 0$.*
- *With probability at least $1/7$ over $S \sim \mathcal{D}^m$, $\mathrm{R}_{\mathcal{U}}(\mathcal{A}(S); \mathcal{D}) > 1/8$.*

We now proceed with the proof of Theorem 3.1.

*Proof of Theorem 3.1.* Let $(X_m)_{m\in\mathbb{N}}$ be an infinite sequence of sets such that each set $X_m$ contains $3m$ distinct points from $\mathcal{X}$, where for any $x_i, x_j \in \cup_{m=1}^{\infty} X_m$ such that $x_i \neq x_j$ we have $\mathcal{U}(x_i) \cap \mathcal{U}(x_j) = \emptyset$. Foreach $m \in \mathbb{N}$, construct $\mathcal{H}_m$ on $X_m$ as in Lemma 3.3. We want to ensure that predictors in $\mathcal{H}_m$ are non-robust on the points in $X_{m'}$ for all $m' \neq m$, by doing the following adjustment for each $h_b \in \mathcal{H}_m$ (recall from Lemma 3.2 that each predictor has its own unique signature $Z_b$),

$$h_b(x) = \begin{cases} -1 & \text{if } x \in Z_b \text{ or } x \in X_{m'} \text{ for } m' \neq m \\ +1 & \text{otherwise} \end{cases}$$

Let $\mathcal{H} = \cup_{m=1}^{\infty} \mathcal{H}_m$. We will show that $\mathrm{vc}(\mathcal{H}) \leq 1$. Pick any two points $z_1, z_2 \in \mathcal{X}$. There are six cases to consider. In case both $z_1$ and $z_2$ are in $X_m$ for some $m \in \mathbb{N}$, then we only obtain the labelings $(+1, +1)$ (by predictors from $\mathcal{H}_m$) and $(-1, -1)$ (by predictors from $\mathcal{H}_{m'}$ with $m' \neq m$). In case both $z_1$ and $z_2$ are in $\mathcal{U}(X_m) \setminus X_m$, then they are not shattered by Lemma 3.2. In case $z_1 \in X_i$ and $z_2 \in X_j$ for $i \neq j$, then we can only obtain the labelings $(+1, -1)$ (by predictors in $\mathcal{H}_i$), $(-1, +1)$ (by predictors in $\mathcal{H}_j$), and $(-1, -1)$ (by predictors in $\mathcal{H}_k$ for $k \neq i, j$). In case $z_1 \in X_i$ and $z_2 \in \mathcal{U}(X_j) \setminus X_j$ for $j \neq i$, then we can't obtain the labeling $(+1, -1)$. In case $z_1 \in \mathcal{U}(X_i) \setminus X_i$ and $z_2 \in \mathcal{U}(X_j) \setminus X_j$ for $i \neq j$, then we can't obtain the labeling $(-1, -1)$. Finally, if either $z_1$ or $z_2$ is in $\mathcal{X}$ but not in $\cup_{m=1}^{\infty} X_m$ and not in $\cup_{m=1}^{\infty} \mathcal{U}(X_m)$, then all predictors label $z_1$ or $z_2$ with $+1$, and so we can't shatter them. This shows that $\mathrm{vc}(\mathcal{H}) \leq 1$.

By Lemma 3.3, it follows that for any proper learning rule $\mathcal{A} : (\mathcal{X} \times \mathcal{Y})^* \mapsto \mathcal{H}$ and for any $m \in \mathbb{N}$, we can construct a distribution $\mathcal{D}$ over $X_m \times \mathcal{Y}$ where there exists a predictor $h^* \in \mathcal{H}_m$ with $\mathrm{R}_{\mathcal{U}}(h^*; \mathcal{D}) = 0$, but with probability at least $1/7$ over $S \sim \mathcal{D}^m$, $\mathrm{R}_{\mathcal{U}}(\mathcal{A}(S); \mathcal{D}) >$ 1/8. This works because classifiers from classes $\mathcal{H}_{m'}$ where $m' \neq m$ make mistakes on points in $X_m$ and so they are non-robust. Thus, rule $\mathcal{A}$ will do worse if it picks predictors from these classes. This shows that the sample complexity to properly robustly PAC learn $\mathcal{H}$ is infinite. This conclude that $\mathcal{H}$ is not properly robustly PAC learnable. □

## 3.3 Finite VC Dimension is Sufficient for (Improper) Robust Learnability

In the previous section we saw that finite VC dimension is *not* sufficient for *proper* robust learnability. We now show that it *is* sufficient for *improper* robust learnability, thus (1) establishing that if $\mathcal{H}$ is learnable, it is also robustly learnable, albeit possibly with a higher sample complexity; and (2) unlike the standard supervised learning setting, to achieve learnability we might need to escape properness, as improper learning is necessary for some hypothesis classes.

We begin, in Section 3.3.1 with the realizable case, i.e. where there exists $h^* \in \mathcal{H}$ with zero robust risk. Then in Section 3.3.2 we turn to the agnostic setting, and observe that a version of a recent reduction by David, Moran, and Yehudayoff (2016) from agnostic to realizable learning applies also for robust learning. We thus establish agnostic robust learnability of finite VC classes by using this reduction and relying on the realizable learning result of Section 3.3.1.

### 3.3.1 Realizable Robust Learnability

We will in fact establish a bound in terms of the *dual VC dimension*. Formally, for each $x \in \mathcal{X}$, define a function $g_x : \mathcal{H} \to \mathcal{Y}$ such that $g_x(h) = h(x)$ for each $h \in \mathcal{H}$. Then the dual VC dimension of $\mathcal{H}$, denoted $\mathrm{vc}^*(\mathcal{H})$, is defined as the VC dimension of the set $\mathcal{G} = \{g_x : x \in \mathcal{X}\}$. This quantity is known to satisfy $\mathrm{vc}^*(\mathcal{H}) < 2^{\mathrm{vc}(\mathcal{H})+1}$ (Assouad, 1983), though for many spaces it satisfies $\mathrm{vc}^*(\mathcal{H}) = O(\mathrm{poly}(\mathrm{vc}(\mathcal{H})))$ or even, as is the case for linear separators, $\mathrm{vc}^*(\mathcal{H}) = O(\mathrm{vc}(\mathcal{H}))$.

**Theorem 3.4.** *For any $\mathcal{H}$ and $\mathcal{U}$, $\forall \varepsilon, \delta \in (0, 1/2)$,*

$$\mathcal{M}^{\mathrm{re}}_{\varepsilon,\delta}(\mathcal{H},\mathcal{U}) = O\left(\mathrm{vc}(\mathcal{H})\mathrm{vc}^*(\mathcal{H})\frac{1}{\varepsilon}\log\left(\frac{\mathrm{vc}(\mathcal{H})\mathrm{vc}^*(\mathcal{H})}{\varepsilon}\right) + \frac{1}{\varepsilon}\log\left(\frac{1}{\delta}\right)\right),$$

Since Assouad (1983) has shown $\mathrm{vc}^*(\mathcal{H}) < 2^{\mathrm{vc}(\mathcal{H})+1}$, this implies the following corollary.

**Corollary 3.5.** *For any $\mathcal{H}$ and $\mathcal{U}$,* $\forall \varepsilon, \delta \in (0, 1/2)$,

$$\mathcal{M}^{\mathrm{re}}_{\varepsilon,\delta}(\mathcal{H},\mathcal{U}) = 2^{O(\mathrm{vc}(\mathcal{H}))}\frac{1}{\varepsilon}\log\left(\frac{1}{\varepsilon}\right) + O\left(\frac{1}{\varepsilon}\log\left(\frac{1}{\delta}\right)\right).$$

Our approach to this proof is via *sample compression* arguments. Specifically, we make use of a lemma (Lemma A.1 in Appendix 3.3.2), which extends to the robust loss the classic compression-based generalization guarantees from the 0-1 loss. We now proceed with the proof of Theorem 3.4.

*Proof of Theorem 3.4.* The learning algorithm achieving this bound is a modification of a sample compression scheme recently proposed by Moran and Yehudayoff (2016), or more precisely, a variant of that method explored by Hanneke, Kontorovich, and Sadigurschi (2019). Our modification forces the compression scheme to also have zero empirical *robust* loss. Fix $\varepsilon, \delta \in (0, 1)$ and a sample size $m > 2\mathrm{vc}(\mathcal{H})$, and denote by $P$ any distribution with $\inf_{h\in\mathcal{H}} R_{\mathcal{U}}(h; P) = 0$.

By classic PAC learning guarantees (Vapnik and Chervonenkis, 1974, Blumer et al., 1989a), there is a positive integer $n = O(\mathrm{vc}(\mathcal{H}))$ with the property that, for any distribution $D$ over $\mathcal{X}\times\mathcal{Y}$ with $\inf_{h\in\mathcal{H}} \mathrm{err}(h; D) = 0$, for $n$ iid $D$-distributed samples $S' = \{(x'_1, y'_1), \dots, (x'_n, y'_n)\}$, with nonzero probability, every $h \in \mathcal{H}$ satisfying $\hat{\mathrm{err}}(h; S') = 0$ also has $\mathrm{err}(h; D) < 1/3$.

Fix a deterministic function $\mathsf{RERM}_{\mathcal{H}}$ mapping any labeled data set to a classifier in $\mathcal{H}$ robustly consistent with the labels in the data set, if a robustly consistent classifier exists (i.e., having zero $\hat{\mathrm{R}}_{\mathcal{U}}$ on the given data set). Suppose we are given training examples $S = \{(x_1, y_1), \dots, (x_m, y_m)\}$ as input to the learner. Under the assumption that this is an iid sample from a robustly realizable distribution, we suppose $\hat{R}_{\mathcal{U}}(\mathsf{RERM}_{\mathcal{H}}(S); S) = 0$, which should hold with probability one. Denote by $I(x) = \min\{i \in \{1, \dots, m\} : x \in \mathcal{U}(x_i)\}$ for every $x \in \bigcup_{i\leq m} \mathcal{U}(x_i)$. Before we can apply the compression approach, we first need to *inflate* the data set to a (potentially infinite) larger set, and then *discretize* it to again reduce it back to a finite sample size. Denote by $\hat{\mathcal{H}} = \{\mathsf{RERM}_{\mathcal{H}}(L) : L \subseteq S, |L| = n\}$. Note that $|\hat{\mathcal{H}}| \leq |\{L : L \subseteq S, |L| = n\}| = \binom{m}{n} \leq \left(\frac{em}{n}\right)^n$. Define an *inflated* data set $S_{\mathcal{U}} = \bigcup_{i\leq m}\{(x, y_{I(x)}) : x \in \mathcal{U}(x_i)\}$. As it is difficult to handle this potentially-infinite set in an algorithm, we consider a discretized version of it. Specifically, consider a *dual space* $\mathcal{G}$: a set of functions $g_{(x,y)} : \mathcal{H} \to \{0, 1\}$ defined as $g_{(x,y)}(h) = 1[h(x) \neq y]$, for each $h \in \mathcal{H}$ and each $(x, y) \in S_{\mathcal{U}}$. The VC di-mension of $\mathcal{G}$ is at most the *dual VC dimension* of $\mathcal{H}$: $\text{vc}^*(\mathcal{H})$, which is known to satisfy $\text{vc}^*(\mathcal{H}) < 2^{\text{vc}(\mathcal{H})+1}$ (Assouad, 1983). Now denote by $\hat{S}_{\mathcal{U}}$ a subset of $S_{\mathcal{U}}$ which includes exactly one $(x, y) \in S_{\mathcal{U}}$ for each distinct classification $\{g_{(x,y)}(h)\}_{h \in \hat{\mathcal{H}}}$ of $\hat{\mathcal{H}}$ realized by functions $g_{(x,y)} \in \mathcal{G}$. In particular, by Sauer's lemma (Vapnik and Chervonenkis, 1971, Sauer, 1972), $|\hat{S}_{\mathcal{U}}| \leq \left(\frac{e|\hat{\mathcal{H}}|}{\text{vc}^*(\mathcal{H})}\right)^{\text{vc}^*(\mathcal{H})}$, which for $m > 2\text{vc}(\mathcal{H})$ is at most $(e^2 m/\text{vc}(\mathcal{H}))^{\text{vc}(\mathcal{H})\text{vc}^*(\mathcal{H})}$. In particular, note that for any $T \in \mathbb{N}$ and $h_1, \ldots, h_T \in \hat{\mathcal{H}}$, if $\frac{1}{T}\sum_{t=1}^{T} 1[h_t(x) = y] > \frac{1}{2}$ for every $(x, y) \in \hat{S}_{\mathcal{U}}$, then $\frac{1}{T}\sum_{t=1}^{T} 1[h_t(x) = y] > \frac{1}{2}$ for every $(x, y) \in S_{\mathcal{U}}$ as well, which would further imply $\hat{R}_{\mathcal{U}}(\text{Majority}(h_1, \ldots, h_T); S) = 0$. We will next go about finding such a set of $h_t$ functions.

By our choice of $n$, we know that for any distribution $D$ over $\hat{S}_{\mathcal{U}}$, $n$ iid samples $S'$ sampled from $D$ would have the property that, with nonzero probability, all $h \in \mathcal{H}$ with $\hat{\text{err}}(h; S') = 0$ also have $\text{err}(h; D) < 1/3$. In particular, this implies at least that there *exists* a subset $S' \subseteq \hat{S}_{\mathcal{U}}$ with $|S'| \leq n$ such that every $h \in \mathcal{H}$ with $\hat{\text{err}}(h; S') = 0$ has $\text{err}(h; D) < 1/3$. For such a set $S'$, note that $\{(x_{I(x)}, y) : (x, y) \in S'\} \subseteq S$, and therefore there exists a set $L$ with $|L| = n$ and $\{(x_{I(x)}, y) : (x, y) \in S'\} \subseteq L \subseteq S$. Furthermore, since $x \in \mathcal{U}(x_{I(x)})$ for every $(x, y) \in S'$, we know $\hat{\text{err}}(\mathsf{RERM}_{\mathcal{H}}(L); S') = 0$, and hence $\text{err}(\mathsf{RERM}_{\mathcal{H}}(L); D) < 1/3$. Altogether, we have that, for any distribution $D$ over $\hat{S}_{\mathcal{U}}$, $\exists h_D \in \hat{\mathcal{H}}$ with $\text{err}(h_D; D) < 1/3$.

We will use the above $h_D$ as a *weak hypothesis* in a boosting algorithm. Specifically, we run the $\alpha$-Boost algorithm (Schapire and Freund, 2012, Section 6.4.2) with $\hat{S}_{\mathcal{U}}$ as its data set, using the above mapping to produce the weak hypotheses for the distributions $D_t$ produced on each round of the algorithm. As proven in (Schapire and Freund, 2012), for an appropriate a-priori choice of $\alpha$ in the $\alpha$-Boost algorithm, running this algorithm for $T = O(\log(|\hat{S}_{\mathcal{U}}|))$ rounds suffices to produce a sequence of hypotheses $\hat{h}_1, \ldots, \hat{h}_T \in \hat{\mathcal{H}}$ s.t.

$$\forall (x, y) \in \hat{S}_{\mathcal{U}}, \frac{1}{T}\sum_{i=1}^{T} 1[\hat{h}_i(x) = y] \geq \frac{5}{9}.$$

From this observation, we already have a sample complexity bound, only slightly worse than the claimed result. Specifically, the above implies that $\hat{h} = \text{Majority}(\hat{h}_1, \ldots, \hat{h}_T)$ satisfies $\hat{R}_{\mathcal{U}}(\hat{h}; S) = 0$. Note that each of these classifiers $\hat{h}_t$ is equal $\mathsf{RERM}_{\mathcal{H}}(L_t)$ for some $L_t \subseteq S$ with $|L_t| = n$. Thus, the classifier $\hat{h}$ is representable as the value of an (order-dependent) reconstruction function $\phi$ with a compression set size

$$nT = O(\text{vc}(\mathcal{H}) \log(|\hat{S}_{\mathcal{U}}|)) = O(\text{vc}(\mathcal{H})^2 \text{vc}^*(\mathcal{H}) \log(m/\text{vc}(\mathcal{H}))). \quad (3.3)$$

Thus, invoking Lemma A.1, if $m > c\mathrm{vc}(\mathcal{H})^2\mathrm{vc}^*(\mathcal{H})\log(\mathrm{vc}(\mathcal{H})\mathrm{vc}^*(\mathcal{H}))$ (for a sufficiently large numerical constant $c$), we have that with probability at least $1-\delta$,

$$R_{\mathcal{U}}(\hat{h};P) \leq O\left(\mathrm{vc}(\mathcal{H})^2\mathrm{vc}^*(\mathcal{H})\tfrac{1}{m}\log(m/\mathrm{vc}(\mathcal{H}))\log(m) + \tfrac{1}{m}\log(1/\delta)\right),$$

and setting this less than $\varepsilon$ and solving for a sufficient size of $m$ to achieve this yields a sample complexity bound, which is slightly larger than that claimed in Theorem 3.4. We next proceed to further refine this bound via a sparsification step. However, as an aside, we note that the above intermediate step will be useful in a discussion below, where the size of this compression scheme in the second expression in (3.3) offers an improvement over a result of Attias, Kontorovich, and Mansour (2018).

Via a technique of (Moran and Yehudayoff, 2016) we can further reduce the above bound. Specifically, since all of $\hat{h}_1, \ldots, \hat{h}_T$ are in $\mathcal{H}$, classic uniform convergence results of Vapnik and Chervonenkis (1971) imply that taking $N = O(\mathrm{vc}^*(\mathcal{H}))$ independent random indices $i_1, \ldots, i_N \sim \mathrm{Uniform}(\{1, \ldots, T\})$, we have $\sup\limits_{(x,y)\in\mathcal{X}\times\mathcal{Y}} \left| \frac{1}{N}\sum\limits_{j=1}^{N} 1[h_{i_j}(x) = y] - \frac{1}{T}\sum\limits_{i=1}^{T} 1[h_i(x) = y] \right| < \frac{1}{18}$. In particular, together with the above guarantee from $\alpha$-Boost, this implies that there exist indices $i_1, \ldots, i_N \in \{1, \ldots, T\}$ (which may be chosen deterministically) satisfying

$$\forall (x,y) \in \hat{S}_{\mathcal{U}},\ \tfrac{1}{T}\textstyle\sum_{j=1}^{N} 1[h_{i_j}(x) = y] \geq -\tfrac{1}{18} + \tfrac{1}{T}\textstyle\sum_{i=1}^{T} 1[h_i(x) = y] > -\tfrac{1}{18} + \tfrac{5}{9} = \tfrac{1}{2},$$

so that the majority vote predictor $\hat{h}'(x) = \mathrm{Majority}(\hat{h}_{i_1}, \ldots, \hat{h}_{i_N})$ satisfies $\hat{\mathrm{err}}(\hat{h}'; \hat{S}_{\mathcal{U}}) = 0$, and hence $\hat{R}_{\mathcal{U}}(\hat{h}'; S) = 0$. Since again, each $\hat{h}_{i_j}$ is the result of $\mathsf{RERM}_{\mathcal{H}}(L_{i_j})$ for some $L_{i_j} \subseteq S$ of size $n$, we have that $\hat{h}'$ can be represented as the value of an (order-dependent) reconstruction function $\varphi$ with a compression set size $nN = O(\mathrm{vc}(\mathcal{H}))\mathrm{vc}^*(\mathcal{H}))$. Thus, Lemma A.1 implies that, for $m \geq c\mathrm{vc}(\mathcal{H})\mathrm{vc}^*(\mathcal{H})$ (for an appropriately large numerical constant $c$), with probability at least $1-\delta$, $R_{\mathcal{U}}(\hat{h}'; P) \leq O\left(\mathrm{vc}(\mathcal{H})\mathrm{vc}^*(\mathcal{H})\frac{1}{m}\log(m) + \frac{1}{m}\log(1/\delta)\right)$. Setting this less than $\varepsilon$ and solving for a sufficient size of $m$ to achieve this yields the stated bound. □

### 3.3.2 Agnostic Robust Learnability

For the agnostic case, we can establish an upper bound via reduction to the realizable case, following an argument from David, Moran, and Yehudayoff (2016). Specifically, we have the following result.

**Theorem 3.6.** *For any $\mathcal{H}$ and $\mathcal{U}$, $\forall \varepsilon, \delta \in (0, 1/2)$,*

$$\mathcal{M}^{\mathrm{ag}}_{\varepsilon,\delta}(\mathcal{H},\mathcal{U}) = O\Big(\mathrm{vc}(\mathcal{H})\mathrm{vc}^*(\mathcal{H})\log(\mathrm{vc}(\mathcal{H})\mathrm{vc}^*(\mathcal{H}))\tfrac{1}{\varepsilon^2}\log^2\Big(\tfrac{\mathrm{vc}(\mathcal{H})\mathrm{vc}^*(\mathcal{H})}{\varepsilon}\Big) + \tfrac{1}{\varepsilon^2}\log\big(\tfrac{1}{\delta}\big)\Big)\,.$$

As above, since Assouad (1983) has shown $\mathrm{vc}^*(\mathcal{H}) < 2^{\mathrm{vc}(\mathcal{H})+1}$, this implies the following corollary.

**Corollary 3.7.** *For any $\mathcal{H}$ and $\mathcal{U}$, $\forall \varepsilon, \delta \in (0, 1/2)$,*

$$\mathcal{M}^{\mathrm{ag}}_{\varepsilon,\delta}(\mathcal{H},\mathcal{U}) = 2^{O(\mathrm{vc}(\mathcal{H}))}\frac{1}{\varepsilon^2}\log^2\left(\frac{1}{\varepsilon}\right) + O\left(\frac{1}{\varepsilon^2}\log\left(\frac{1}{\delta}\right)\right)\,.$$

We establish the theorem via a reduction to the realizable case, following an approach used by David, Moran, and Yehudayoff (2016), except here applied to the robust loss. The reduction is summarized in the following Theorem, whose proof can be found in Appendix A.3:

**Theorem 3.8.** *Denote $m_0 = \mathcal{M}^{\mathrm{re}}_{1/3,1/3}(\mathcal{H},\mathcal{U})$. Then*

$$\mathcal{M}^{\mathrm{ag}}_{\varepsilon,\delta}(\mathcal{H},\mathcal{U}) = O\left(\frac{m_0}{\varepsilon^2}\log^2\left(\frac{m_0}{\varepsilon}\right) + \frac{1}{\varepsilon^2}\log\left(\frac{1}{\delta}\right)\right)\,.$$

From this, Theorem 3.6 follows immediately by combining Theorem 3.8 with Theorem 3.4.

Bounded cardinality confusion sets: As noted in the proof of Theorem 3.4, the compression size (3.3) further implies an improvement over a theorem of Attias, Kontorovich, and Mansour (2018). Specifically, Attias et al. considered the case $\max_{x\in\mathcal{X}} |\mathcal{U}(x)| \leq k$ for some fixed $k \in \mathbb{N}$, and presented a learning rule establishing the sample complexity gurantee:

$$\mathcal{M}^{\mathrm{ag}}_{\varepsilon,\delta}(\mathcal{H},\mathcal{U}) = O\Big(\tfrac{\mathrm{vc}(\mathcal{H})k\log(k)}{\varepsilon^2} + \tfrac{1}{\varepsilon^2}\log\big(\tfrac{1}{\delta}\big)\Big)\,. \tag{3.4}$$

Their analysis proceeds by bounding the Rademacher complexity of the robust loss class of the convex hull of $\mathcal{H}$, which implies the sample complexity (3.4) can also be achieved by $\mathsf{RERM}_{\mathcal{H}}$ (they propose an alternative, improper, learning rule for computational reasons). But when $\max |\mathcal{U}(x)| \leq k$, the second expression in our (3.3) would be at most $O(\mathrm{vc}(\mathcal{H})\log(mk))$. Thus, following the compression argument as in the proof of Theorem 3.4 would yield the following sample complexity for our improper rule:

$$\mathcal{M}^{\mathrm{re}}_{\varepsilon,\delta}(\mathcal{H},\mathcal{U}) = O\Big(\tfrac{\mathrm{vc}(\mathcal{H})\log(k)}{\varepsilon}\log\Big(\tfrac{\mathrm{vc}(\mathcal{H})\log(k)}{\varepsilon}\Big) + \tfrac{\mathrm{vc}(\mathcal{H})}{\varepsilon}\log^2\Big(\tfrac{\mathrm{vc}(\mathcal{H})}{\varepsilon}\Big) + \tfrac{1}{\varepsilon}\log\big(\tfrac{1}{\delta}\big)\Big)\,,$$

and hence by Theorem 3.8:

$$\mathcal{M}^{\mathrm{ag}}_{\varepsilon,\delta}(\mathcal{H},\mathcal{U}) = O\!\left(\frac{\mathrm{vc}(\mathcal{H})\log(k)}{\varepsilon^2}\mathrm{polylog}\!\left(\frac{\mathrm{vc}(\mathcal{H})\log(k)}{\varepsilon}\right) + \frac{1}{\varepsilon^2}\log\!\left(\frac{1}{\delta}\right)\right).$$

In particular, our approach reduces the dependence on $k$ from $k\log(k)$ in (3.4) as obtained by Attias, Kontorovich, and Mansour (2018), to $\log(k)(\log\log(k))^3$. To do so, our approach *does* rely on improper learning, and our arguments are not valid for $\mathsf{RERM}_{\mathcal{H}}$. We do not know whether improperness is *required* to obtain this improvement, or whether in this case a polylog$k$ dependence is possible even with RERM or some other proper learning rule. It follows from the construction of our negative result for proper learning in Theorem 3.1, that at least a $\log(k)$ factor is sometimes necessary for proper learning (regardless of the VC dimension), whereas our Corollary 3.7 implies that improper learning can achieve a sample complexity that is entirely independent of $k$ (albeit with a worse dependence on the VC dimension).

## 3.4 Toward Necessary and Sufficient conditions for Robust Learnability

In the previous section, we saw that having finite VC dimension is *sufficient* for robust learnability. But a simple construction shows that it is not *necessary*: consider an infinite domain $\mathcal{X}$, the hypothesis class of all possible predictors $\mathcal{H} = \{-,+\}^{\mathcal{X}}$, and an all-powerful adversary specified by $\mathcal{U}(x) = \mathcal{X}$. In this case, the hypothesis minimizing the population robust risk $\mathrm{R}_{\mathcal{U}}(h;\mathcal{D})$ would always be the all-positive or the all-negative hypothesis, and so these are the only two hypothesis we should compete with. And so, even though $\mathrm{vc}(\mathcal{H}) = \infty$, a single example suffices to inform the learner of whether to produce the all-positive or all-negative function.

Can we then have a tight characterization of robust learnability? Is there a weaker notion that is both necessary and sufficient for learning? A simple complexity measure one might consider is the maximum number of points $x_1,\ldots,x_m$ such that the entire perturbation sets $\mathcal{U}(x_1),\ldots,\mathcal{U}(x_m)$ are shattered by $\mathcal{H}$. That is, such that $\forall y_1,\ldots,y_m \in \{+1,-1\}, \exists h \in \mathcal{H}, \forall i \forall x' \in \mathcal{U}(x_i),\ h(x') = y_i$. We denote this as $\dim_{\mathcal{U}\times}(\mathcal{H})$. When $\mathcal{U}(x)$ are balls around $x$, which is the typical case in metric-based robustness, this can be thought of shattering with a margin in input space. Indeed, for linear predictors and when $\mathcal{U}(x) = \{x' | \|x - x'\|_2 \leq \gamma\}$ is a Euclidean ball around $x$, $\dim_{\mathcal{U}\times}(\mathcal{H})$ exactly agrees with the fat shattering dimension at scale $\gamma$ (or the $VC_\gamma$ dimension).

While it is fairly obvious that $\dim_{\mathcal{U}\times}(\mathcal{H})$ provides a lower bound on the sample complexity of robust learning, and thus its finiteness is *necessary* for learning, we construct an example in Appendix A.4 showing that it is not *sufficient*. Specifically, there are classes where *no* points can be shattered in this way, and yet the classes are not robustly learnable. Formally,

**Proposition 3.9.** *There exist $\mathcal{X}$, $\mathcal{H}$, $\mathcal{U}$ such that* $\dim_{\mathcal{U}\times}(\mathcal{H}) = 0$ *but* $\mathcal{M}^{\mathrm{re}}_{\varepsilon,\delta}(\mathcal{H},\mathcal{U}) = \infty$.

We now attempt to refine the above measure, and introduce a weaker notion of robust shattering that that can still be used to lower bound the sample complexity for robust learnability. Given an adversary $\mathcal{U}$ and a hypothesis class $\mathcal{H}$, consider the following notion of $\mathcal{U}$-robust shattering,

**Definition 3.10** (Robust Shattering Dimension). *A sequence $x_1, \dots, x_m \in \mathcal{X}$ is said to be* $\mathcal{U}$-robustly shattered *by $\mathcal{H}$ if $\exists z_1^+, z_1^-, \dots, z_m^+, z_m^- \in \mathcal{X}$ with $x_i \in \mathcal{U}(z_i^+) \cap \mathcal{U}(z_i^-)\ \forall i \in [m]$, and $\forall y_1, \dots, y_m \in \{-,+\}$, $\exists h \in \mathcal{H}$ with $h(z') = y_i, \forall z' \in \mathcal{U}(z_i^{y_i}), \forall i \in [m]$. The* $\mathcal{U}$-robust shattering dimension $\dim_{\mathcal{U}}(\mathcal{H})$ *is defined as the largest $m$ for which there exist $m$ points $\mathcal{U}$-robustly shattered by $\mathcal{H}$.*

We have that $\dim_{\mathcal{U}\times}(\mathcal{H}) \leq \dim_{\mathcal{U}}(\mathcal{H}) \leq \mathrm{vc}(\mathcal{H})$, where the first inequality follows since disjoint robust shattering is a special case of robust shattering with $z_i^y = x_i$, and so $\dim_{\mathcal{U}}(\mathcal{H})$ is a plausible candidate for a necessary and sufficient dimension of robust learnability. The following theorem (proof provided in appendix A.4) establishes that the sample complexity of robust learnability is indeed lower bounded by the *$\mathcal{U}$-robust shattering dimension* $\dim_{\mathcal{U}}(\mathcal{H})$,

**Theorem 3.11.** *For any $\mathcal{X}$, $\mathcal{H}$, and $\mathcal{U}$,*
$\mathcal{M}^{\mathrm{re}}_{\varepsilon,\delta}(\mathcal{H},\mathcal{U}) = \Omega\Big(\frac{\dim_{\mathcal{U}}(\mathcal{H})}{\varepsilon} + \frac{1}{\varepsilon}\log\big(\frac{1}{\delta}\big)\Big)$ *and* $\mathcal{M}^{\mathrm{ag}}_{\varepsilon,\delta}(\mathcal{H},\mathcal{U}) = \Omega\Big(\frac{\dim_{\mathcal{U}}(\mathcal{H})}{\varepsilon^2} + \frac{1}{\varepsilon^2}\log\big(\frac{1}{\delta}\big)\Big)$.

Based on Corollary 3.5 and Theorem 3.11, for any adversary $\mathcal{U}$ and any hypothesis class $\mathcal{H}$, we have

$$\Omega\Big(\frac{\dim_{\mathcal{U}}(\mathcal{H})}{\varepsilon} + \frac{1}{\varepsilon}\log\big(\frac{1}{\delta}\big)\Big) \leq \mathcal{M}^{\mathrm{re}}_{\varepsilon,\delta}(\mathcal{H},\mathcal{U}) \leq 2^{O(\mathrm{vc}(\mathcal{H}))}\frac{1}{\varepsilon}\log\big(\frac{1}{\varepsilon}\big) + O\big(\frac{1}{\varepsilon}\log\big(\frac{1}{\delta}\big)\big)\,. \quad (3.5)$$

That is, the VC dimension is sufficient, and the robust shattering dimension is necessary for robust learnability. As discussed at the beginning of the Section, we know the VC dimension is *not* necessary and there can be an arbitrary large, even infinite, gap in the second inequality. We do not know whether the robust shattering dimension is also sufficient for learning, or whether there can also be a big gap in the first inequality. Establishing a complexity measure that characterizes robust learnability thus remains an open question so far. We will address this question next in Chapter 4.

## 3.5 Discussion and Open Directions

Perhaps one of the most interesting takeaways from this work is that we should start considering *improper* learning algorithms for adversarially robust learning. Even though our improper learning rule might not be practical, our results suggest to consider departing from robust empirical risk minimization and M-estimation (as in almost all published work), and considering *improper* learning rules such as bagging or other ensemble methods.

Although we settled the question of robust learnability of VC classes, there remains a large gap in the question of what is the optimal sample complexity for robust learning. Can the exponential dependence on $\text{vc}(\mathcal{H})$ in Corollaries 3.5 and 3.7 be improved to a linear dependence? Perhaps this is possible with a new analysis of our learning rule or a different *improper* learning rule. Since our learning rule and analysis stem from recent progress on compression schemes for VC classes (Moran and Yehudayoff, 2016), it is certainly possible that further progress on the celebrated open problem regarding the existence of $\text{vc}(\mathcal{H})$ compression schemes (Floyd and Warmuth, 1995, Warmuth, 2003) could also assist in progress on adversarially robust learning.

Our results demonstrate that there *exist* hypothesis classes with large gaps between what can be done with proper vs. improper robust learning. This means that when studying a particular class, such as classes corresponding to neural networks, one should consider the possibility that there *might* be such a gap and that improper learning *might* be necessary. It remains open to establish whether such gaps actually exist for specific interesting neural net classes (e.g., functions representable by a specific architecture, possibly with a bounded weight norm).

Throughout the paper we ignored computational considerations. Our learning rule can be viewed as an algorithm with black-box access to $\mathsf{RERM}_{\mathcal{H}}$, but making order $m^{\text{vc}(\mathcal{H})}$ such calls, and additionally requiring order $m^{\text{vc}(\mathcal{H})\text{vc}^*(\mathcal{H})}$ time and space to represent and update the distributions used by the boosting algorithm. Without significantly increasing the sample complexity, is it possible to robustly learn with an algorithm making only a polynomial (in $\text{vc}(\mathcal{H}), \text{vc}^*(\mathcal{H}), m$) number of calls to $\mathsf{RERM}_{\mathcal{H}}$ or even $\mathsf{ERM}_{\mathcal{H}}$, plus polynomial additional time and space? What about $\text{poly}(\text{vc}(\mathcal{H}), m)$? This question becomes even more interesting if there is such an algorithm that also only requires sample size $m = \text{poly}(\text{vc}(\mathcal{H}), 1/\varepsilon, \log(1/\delta))$, rather than the

$m = \text{poly}(\text{vc}(\mathcal{H}), \text{vc}^*(\mathcal{H}), 1/\varepsilon, \log(1/\delta))$ sufficient for our algorithm. We refer the reader to Chapter 6 for some initial progress in this direction. Would another type of oracle be useful? For example, can one devise efficient methods that rely on black-box access to ERM on the dual of the hypothesis class (i.e. finding an example that is correct for the largest number of hypotheses in a given finite set of hypotheses)? More ambitiously, one may ask whether efficient PAC learnability implies efficient robust PAC learnability, roughly translating to asking whether access to *any* (non-robust) learning rule is sufficient for efficient robust learning.

As a final remark, we note that our results easily extend to the multiclass setting ($|\mathcal{Y}| > 2$). In that case, by essentially the same algorithms and proofs, Theorems 3.4 and 3.6 (and Corollaries 3.5 and 3.7) will hold with $\text{vc}(\mathcal{H})$ replaced by the *graph dimension* (Natarajan, 1989, Ben-David et al., 1995, Daniely et al., 2015). The lower bound in Theorem 3.11 also holds, by the same arguments, but with $\dim_{\mathcal{U}}(\mathcal{H})$ generalized analogous to the *Natarajan dimension* (Natarajan, 1989): that is, in the definition of robust shattering, after "and", we now require $\forall i \exists y_{i,-}, y_{i,+} \in \mathcal{Y}$ s.t. $\forall b_1, \ldots, b_m \in \{-, +\}$, $\exists h \in \mathcal{H}$ with $h(z') = y_{i,b_i}$, $\forall z' \in \mathcal{U}(z_i^{b_i})$, $\forall i$. We leave as an open question whether one can also express an upper bound controlled by this quantity.

# 4

# A Characterization of Adversarially Robust Learning

## 4.1 Introduction

The aim of this chapter is to put forward a theory precisely characterizing the complexity of *robust* learnability. We know from Chapter 3 that finite VC dimension is *sufficient* for robust learnability (see Corollary 3.5 and Corollary 3.7), but we also know that its finiteness is not *necessary* (see Section 3.4). Furthermore, there is a (potentially) *infinite* gap between the established quantitative upper and lower bounds on the sample complexity of adversarially robust learning (see Equation 3.5), and we do not know of any *optimal* learners for this problem. In this chapter, we address the following fundamental questions:

*What model classes* $\mathcal{H}$ *are* robustly *learnable with respect to an* arbitrary *perturbation set* $\mathcal{U}$*?*
*Can we design generic optimal learners for adversarially robust learning?*

The problem of characterizing learnability is the most basic question of statistical learning theory. In classical (non-robust) supervised learning, the fundamental theorem of statistical learning (Vapnik and Chervonenkis, 1971, 1974, Blumer, Ehrenfeucht, Haussler, and Warmuth, 1989a, Ehrenfeucht, Haussler, Kearns, and Valiant, 1989) provides a complete un-derstanding of *what* is learnable: classes $\mathcal{H}$ with finite VC dimension, and *how* to learn: by the generic learner *empirical risk minimization* ($\mathsf{ERM}_{\mathcal{H}}$). We also know that $\mathsf{ERM}_{\mathcal{H}}$ is a near-optimal learner for $\mathcal{H}$ with sample complexity that is quantified tightly by the VC dimension of $\mathcal{H}$.

Problem setup. Given an instance space $\mathcal{X}$ and label space $\mathcal{Y} = \{\pm 1\}$, we consider robustly learning an arbitrary hypothesis class $\mathcal{H} \subseteq \mathcal{Y}^{\mathcal{X}}$ with respect to an arbitrary perturbation set $\mathcal{U} : \mathcal{X} \to 2^{\mathcal{X}}$, where $\mathcal{U}(x) \subseteq \mathcal{X}$ represents the set of perturbations that can be chosen by an adversary at test-time, as measured by the *robust risk* (see Equation 2.1).

We denote by $\mathrm{RE}(\mathcal{H},\mathcal{U})$ the set of distributions $\mathcal{D}$ over $\mathcal{X} \times \mathcal{Y}$ that are *robustly realizable*: $\exists h^* \in \mathcal{H}, \mathrm{R}_{\mathcal{U}}(h^*;\mathcal{D}) = 0$. A learner $\mathbb{A} : (\mathcal{X} \times \mathcal{Y})^* \to \mathcal{Y}^{\mathcal{X}}$ receives $n$ i.i.d. examples $S = \{(x_i, y_i)\}_{i=1}^n$ drawn from some unknown distribution $\mathcal{D} \in \mathrm{RE}(\mathcal{H},\mathcal{U})$, and outputs a predictor $\mathbb{A}(S)$. The worst-case expected *robust* risk of learner $\mathbb{A}$ with respect to $\mathcal{H}$ and $\mathcal{U}$ is defined as:

$$\mathcal{E}_n(\mathbb{A};\mathcal{H},\mathcal{U}) = \sup_{\mathcal{D}\in\mathrm{RE}(\mathcal{H},\mathcal{U})} \mathop{\mathbb{E}}_{S\sim\mathcal{D}^n} \mathrm{R}_{\mathcal{U}}(\mathbb{A}(S);\mathcal{D}). \tag{4.1}$$

The *minimax* expected robust risk of learning $\mathcal{H}$ with respect to $\mathcal{U}$ is defined as:

$$\mathcal{E}_n(\mathcal{H},\mathcal{U}) = \inf_{\mathbb{A}} \mathcal{E}_n(\mathbb{A};\mathcal{H},\mathcal{U}). \tag{4.2}$$

For any $\varepsilon \in (0,1)$, the *sample complexity of realizable robust $\varepsilon$-PAC learning of $\mathcal{H}$ with respect to $\mathcal{U}$*, denoted $\mathcal{M}^{\mathrm{re}}_{\varepsilon}(\mathcal{H},\mathcal{U})$, is defined as

$$\mathcal{M}^{\mathrm{re}}_{\varepsilon}(\mathcal{H},\mathcal{U}) = \min\left\{n \in \mathbb{N}\cup\{\infty\} : \mathcal{E}_n(\mathcal{H},\mathcal{U}) \leq \varepsilon\right\}. \tag{4.3}$$

$\mathcal{H}$ is robustly PAC learnable realizably with respect to $\mathcal{U}$ if $\forall_{\varepsilon\in(0,1)}$, $\mathcal{M}^{\mathrm{re}}_{\varepsilon}(\mathcal{H},\mathcal{U})$ is finite.

Related work and gaps. We showed in Chapter 3 that any class $\mathcal{H}$ with finite VC dimension is robustly PAC learnable with respect to any perturbation set $\mathcal{U}$; by establishing that $\mathcal{M}^{\mathrm{re}}_{\varepsilon}(\mathcal{H},\mathcal{U}) \leq \tilde{O}(\frac{2^{\mathrm{vc}(\mathcal{H})}}{\varepsilon})$, where $\mathrm{vc}(\mathcal{H})$ denotes the VC dimension of $\mathcal{H}$ (see Corollary 3.5). While this gives a *sufficient* condition for robust learnability, we also showed that finite VC dimension is *not* necessary for robust learnability (see Section 3.4), indicating a (potentially) infinite gap between the established upper and lower bounds on the sample complexity. We next provide simple motivating examples that highlight these gaps in this existing theory, and suggest that the learner witnessing the upper bound in Theorem 3.4 and Corollary 3.5 might be very sub-optimal:

**Example 4.1.** Consider an infinite domain $\mathcal{X}$, the hypothesis class of all possible predictors $\mathcal{H} = \mathcal{Y}^{\mathcal{X}}$, and an all-powerful perturbation set $\mathcal{U}(x) = \mathcal{X}$. In this case, the hypothesis minimizing the population robust risk $\mathrm{R}_{\mathcal{U}}(h; \mathcal{D})$ would always be the all-positive or the all-negative hypothesis, and so these are the only two hypotheses we should compete with. And so, even though $\mathrm{vc}(\mathcal{H}) = \infty$, a single example suffices to inform the learner of whether to produce the all-positive or all-negative function.

**Example 4.2.** A less extreme and more natural example is to take $\mathcal{X} = \mathbb{R}^{\infty}$ (an infinite dimensional space), and $\mathcal{H}$ the set of homogeneous halfspaces in $\mathcal{X}$, and a perturbation set $\mathcal{U}(x) = \{z \in \mathcal{X} : \langle x, v \rangle = \langle z, v \rangle \text{ for } v \in V\}$ where $V$ is the set of the first $d$ standard basis vectors. In this example, an adversary is allowed to *arbitrarily* corrupt all but $d$ features. Note that $\mathrm{vc}(\mathcal{H}) = \infty$ but we can robustly PAC learn $\mathcal{H}$ with $O(d)$ samples: simply project samples from $\mathcal{X}$ onto the subspace spanned by $V$ and learn a $d$-dimensional halfspace.

Main Results. In fact, even more strongly, we show in Theorem 4.1 that there are problem instances $(\mathcal{H}, \mathcal{U})$ that are *not* robustly learnable by the learner we proposed in Section 3.3 (Theorem 3.4 and Corollary 3.5), but *are* robustly learnable with a different generic learner. Beyond this, Theorem 4.1 actually illustrates, in a strong negative sense, the suboptimality of any *local* learner—a family of learners that we identify in this chapter—which informally *only* has access to labeled training examples and perturbations of the training examples, but otherwise does not know the perturbation set $\mathcal{U}$ (defined formally in Definition 4.1).

In this chapter, we adopt a *global* perspective on robust learning. In Section 4.3, we introduce a novel graph construction, the *global one-inclusion graph*, that in essence embodies the complexity of *robust* learnability. In Theorem 4.2, for any class $\mathcal{H}$ and perturbation set $\mathcal{U}$, we utilize the global one-inclusion graph to construct a generic *minimax optimal* learner $\mathbb{G}_{\mathcal{H},\mathcal{U}}$ satisfying $\mathcal{E}_{2n-1}(\mathbb{G}_{\mathcal{H},\mathcal{U}}; \mathcal{H}, \mathcal{U}) \leq 4 \cdot \mathcal{E}_n(\mathcal{H}, \mathcal{U})$. Our global one-inclusion graph utilizes the structure of the class $\mathcal{H}$ and the perturbation set $\mathcal{U}$ in a global manner by considering *all* datasets of size $n$ that are robustly realizable, where each dataset corresponds to a vertex in the graph. Edges in the graph correspond to pairs of datasets that agree on $n-1$ datapoints, disagree on the $n^{\text{th}}$ label, and *overlap* on the $n^{\text{th}}$ datapoint according to their $\mathcal{U}$ sets. We arrive at an optimal learner by orienting the edges of this graph to minimize a notion of *adversarial-out-degree* that corresponds to the average leave-one-out *robust* error. Our learner avoids the lower bound in Theorem 4.1 since it is *non-local* and utilizes the structure of $\mathcal{U}$ at *test-time*.

In Section 4.5, we introduce a new complexity measure denoted $\mathfrak{D}_{\mathcal{U}}(\mathcal{H})$ (defined in Equation 4.11) based on our global one-inclusion graph. We show in Theorem 4.3 that $\mathfrak{D}_{\mathcal{U}}(\mathcal{H})$ *qualitatively* characterizes robust learnability: a class $\mathcal{H}$ is robustly learnable with respect to $\mathcal{U}$ if and only if $\mathfrak{D}_{\mathcal{U}}(\mathcal{H})$ is finite. In Theorem 4.4, we show that $\mathfrak{D}_{\mathcal{U}}(\mathcal{H})$ tightly *quantifies* the sample complexity of robust learnability: $\Omega(\frac{\mathfrak{D}_{\mathcal{U}}(\mathcal{H})}{\varepsilon}) \leq \mathcal{M}^{\text{re}}_{\varepsilon}(\mathcal{H}, \mathcal{U}) \leq \tilde{O}(\frac{\mathfrak{D}_{\mathcal{U}}(\mathcal{H})}{\varepsilon})$. This closes the (potentially) *infinite* gap previously established in Chapter 3.

In Section 4.5.3, beyond the realizable setting, we show in Theorem 4.6 that our complexity measure $\mathfrak{D}_{\mathcal{U}}(\mathcal{H})$ bounds the sample complexity of *agnostic* robust learning: $\mathcal{M}^{\text{ag}}_{\varepsilon}(\mathcal{H}, \mathcal{U}) \leq \tilde{O}(\frac{\mathfrak{D}_{\mathcal{U}}(\mathcal{H})}{\varepsilon^2})$. This shows that $\mathfrak{D}_{\mathcal{U}}(\mathcal{H})$ tightly (up to log factors) characterizes the sample complexity of *agnostic* robust learning, since by definition, $\mathcal{M}^{\text{ag}}_{\varepsilon}(\mathcal{H}, \mathcal{U}) \geq \mathcal{M}^{\text{re}}_{\varepsilon}(\mathcal{H}, \mathcal{U})$.

## 4.2 Local Learners are Suboptimal

In this section, we identify a broad family of learners, which we term *local* learners, and show that such learners are *suboptimal* for adversarially robust learning. Informally, *local* learners *only* have access to labeled training examples and perturbations of the training examples, but otherwise do not know the perturbation set $\mathcal{U}$. More formally,

**Definition 4.1** (Local Learners). *For any class $\mathcal{H}$, a local learner $\mathbb{A}_{\mathcal{H}} : (\mathcal{X} \times \mathcal{Y} \times 2^{\mathcal{X}})^* \to \mathcal{Y}^{\mathcal{X}}$ for $\mathcal{H}$ takes as input a sequence $S_{\mathcal{U}} = \{(x_i, y_i, \mathcal{U}(x_i))\}_{i=1}^m \in \mathcal{X} \times \mathcal{Y} \times 2^{\mathcal{X}}$ consisting of labeled training examples and their corresponding perturbations according to some perturbation set $\mathcal{U}$, and outputs a predictor $f \in \mathcal{Y}^{\mathcal{X}}$. In other words, $\mathbb{A}$ has full knowledge of $\mathcal{H}$, but only* local *knowledge of $\mathcal{U}$ through the training examples.*

We note that the robust learner we proposed previously in Section 3.3, for example, *is* a local learner: for a given a class $\mathcal{H}$ and input $S_{\mathcal{U}} = \{(x_i, y_i, \mathcal{U}(x_i))\}_{i=1}^m$, the learner outputs a majority-vote over predictors in $\mathcal{H}$, that are carefully chosen based on the input $S_{\mathcal{U}}$. Moreover, adversarial training methods in practice (e.g., Madry, Makelov, Schmidt, Tsipras, and Vladu, 2018, Zhang, Yu, Jiao, Xing, Ghaoui, and Jordan, 2019b) are also examples of local learners (they only utilize the perturbations on the training examples). We provably show next that *local* learners are *not* optimal. We give a construction where it is not possible to robustly learn without taking advantage of the information about $\mathcal{U}$ at *test-time*.

**Theorem 4.1.** *There is an instance space $\mathcal{X}$ and a class $\mathcal{H}$, such that for any* local *learner $\mathbb{A}_{\mathcal{H}} : (\mathcal{X} \times \mathcal{Y} \times 2^{\mathcal{X}})^* \to \mathcal{Y}^{\mathcal{X}}$ and any sample size $m \in \mathbb{N}$, there exists a perturbation set $\mathcal{U}$ for which:*

1. *$\mathbb{A}_{\mathcal{H}}$ fails to robustly learn $\mathcal{H}$ with respect to $\mathcal{U}$ using $m$ samples.*
2. *There exists a non-local learner $\mathbb{G}_{\mathcal{H},\mathcal{U}} : (\mathcal{X} \times \mathcal{Y})^* \to \mathcal{Y}^{\mathcal{X}}$ which robustly learns $\mathcal{H}$ with respect to $\mathcal{U}$ with $0$ samples.*

This negative result highlights that there are limitations to what can be achieved with *local* learners. It also highlights the importance of utilizing the structure of the perturbation set $\mathcal{U}$ at test-time, which we also observe in Chapter 9 in the context of *transductive* robust learning where the learner receives a training set of $n$ labeled examples and a test set of $n$ unlabeled adversarial perturbations, and is asked to label the test set with few errors. In practice, randomized smoothing (Cohen, Rosenfeld, and Kolter, 2019) is an example of a non-local method in the sense that at prediction time, it uses the perturbation set to compute predictions.

*Proof of Theorem 4.1.* We begin with describing the instance space $\mathcal{X}$ and the class $\mathcal{H}$. Pick three infinite unique sequences $(x_n^+)_{n\in\mathbb{N}}$, $(x_n^-)_{n\in\mathbb{N}}$, and $(z_n)_{n\in\mathbb{N}}$ from $\mathbb{R}^2$ such that for each $n \in \mathbb{N} : x_n^+ = (n, 1), x_n^- = (n, -1), z_n = (n, 0)$, and let $\mathcal{X} = \cup_{n\in\mathbb{N}} \{x_n^+, x_n^-, z_n\}$. Consider the class $\mathcal{H}$ defined by

$$\mathcal{H} = \left\{ h_{\boldsymbol{y}} : \boldsymbol{y} \in \{\pm 1\}^{\mathbb{N}} \right\}, \text{ where } h_{\boldsymbol{y}}(z_n) = y_n \wedge h_{\boldsymbol{y}}(x_n^+) = +1 \wedge h_{\boldsymbol{y}}(x_n^-) = -1 \, (\forall n \in \mathbb{N}) \,. \tag{4.4}$$

Observe that all classifiers in $\mathcal{H}$ are constant on $(x_n^+)_{n\in\mathbb{N}}$ and $(x_n^-)_{n\in\mathbb{N}}$, but they shatter $(z_n)_{n\in\mathbb{N}}$. We will consider a *random* perturbation set $\mathcal{U} : \mathcal{X} \to 2^{\mathcal{X}}$ that is defined as follows:

$$\forall n \in \mathbb{N} : \begin{cases} \mathcal{U}(x_n^+) = \{x_n^+, z_n\} \text{ and } \mathcal{U}(x_n^-) = \{x_n^-\} \text{ and } \mathcal{U}(z_n) = \{x_n^+, x_n^-, z_n\} \text{ w.p. } \frac{1}{2}, \\ \mathcal{U}(x_n^+) = \{x_n^+\} \text{ and } \mathcal{U}(x_n^-) = \{x_n^-, z_n\} \text{ and } \mathcal{U}(z_n) = \{x_n^+, x_n^-, z_n\} \text{ w.p. } \frac{1}{2}. \end{cases} \tag{4.5}$$

For any sample size $m \in \mathbb{N}$, let $P$ be a uniform distribution on

$$\left\{(x_1^+, +1), (x_1^-, -1), \dots, (x_{3m}^+, +1), (x_{3m}^-, -1)\right\}.$$

Observe that for any randomized $\mathcal{U}$ (according to Equation 4.5), the distribution $P$ is *robustly realizable* with respect to $\mathcal{U}$: $\exists h \in \mathcal{H}, \mathrm{R}_{\mathcal{U}}(h; P) = 0$. Let $\mathbb{A}$ be an arbitrary *local* learner (see Definition 4.1), i.e., $\mathbb{A}$ has full knowledge of the class $\mathcal{H}$, but only partial knowledge of $\mathcal{U}$ through the training samples. Let $S \sim P^m$ be a fixed random set of training examples drawn from $P$. Then,

$$\begin{aligned}
\mathop{\mathbb{E}}_{\mathcal{U}} \mathrm{R}_{\mathcal{U}}(\mathbb{A}(S_{\mathcal{U}}); P) &= \mathop{\mathbb{E}}_{\mathcal{U}} \mathop{\mathbb{E}}_{(x,y)\sim P} 1[\exists z \in \mathcal{U}(x) : \mathbb{A}(S_{\mathcal{U}})(z) \neq y] \\
&\geq \Pr_{(x,y)\sim P}[(x,y) \notin S] \mathop{\mathbb{E}}_{\mathcal{U}} \mathop{\mathbb{E}}_{(x,y)\sim P}[1[\exists z \in \mathcal{U}(x) : \mathbb{A}(S_{\mathcal{U}})(z) \neq y] | (x,y) \notin S] \\
&= \Pr_{(x,y)\sim P}[(x,y) \notin S] \mathop{\mathbb{E}}_{(x,y)\sim P} \Pr_{\mathcal{U}}[\exists z \in \mathcal{U}(x) : \mathbb{A}(S_{\mathcal{U}})(z) \neq y] | (x,y) \notin S] \\
&\geq \frac{1}{3} \cdot \frac{1}{2} = \frac{1}{6}.
\end{aligned}$$

By law of total expectation, this implies that there exists a deterministic choice of $\mathcal{U}$ such that

$\mathbb{E}_{S\sim P^m} \mathrm{R}_{\mathcal{U}}(\mathbb{A}(S_{\mathcal{U}}); P) \geq \frac{1}{6}$. This establishes that $\mathbb{A}$ fails to robustly learn $\mathcal{H}$ with respect to $\mathcal{U}$ using $m$ samples. On the other hand, $\mathcal{H}$ is robustly learnable with respect to $\mathcal{U}$ with 0 samples by means of our non-local learner $\mathbb{G}_{\mathcal{H},\mathcal{U}}$ (see Section 4.4 and Theorem 4.2) which utilizes *full* knowledge of $\mathcal{U}$. In particular, 0 samples are needed, since the graph $G_{\mathcal{H}}^{\mathcal{U}}$ will contain *no* edges by the definition of $\mathcal{H}$ (Equation 4.4) and $\mathcal{U}$ (Equation 4.5). □

## 4.3 A Global One-Inclusion Graph

To go beyond the limitations of local learners from Section 4.2, in this section, we introduce: the *global one-inclusion graph*, the main mathematical object which allows us to adopt a global perspective on robust learning. Our global one-inclusion graph is inspired by the classical one-inclusion graph introduced by Haussler, Littlestone, and Warmuth (1994), which leads to an algorithm that is near-optimal for (non-robust) PAC learning, and has also been adapted and used in multi-class learning (Rubinstein, Bartlett, and Rubinstein, 2006, Daniely and Shalev-Shwartz, 2014, Brukhim, Carmon, Dinur, Moran, and Yehudayoff, 2022) and for learning partial concept classes* (Alon, Hanneke, Holzman, and Moran, 2021). Before introducing our global one-inclusion graph, to ease the readers, we begin first with describing the construction of the classical one-inclusion graph due to Haussler, Littlestone, and Warmuth (1994) and its use as a (non-robust) learner, and discuss its limitations for adversarially robust learning.

### 4.3.1 Background: Classical One-Inclusion Graphs

For a given class $\mathcal{H}$ and a finite dataset $X = \{x_1, \dots, x_n\} \subseteq \mathcal{X}$, the classical one-inclusion graph $G_{X,\mathcal{H}}$ consists of vertices $V = \{(h(x_1), \dots, h(x_n)) : h \in \mathcal{H}\}$ where each vertex $v = (h(x_1), \dots, h(x_n)) \in V$ is a *realizable* labeling of $X$, and two vertices $u, v \in V$ are connected with an edge if and only if they differ only in the labeling of a single $x_i \in X$. Haussler, Littlestone, and Warmuth (1994) showed that the edges in $G_{X,\mathcal{H}}$ can be oriented such that each vertex has out-degree at most $\mathrm{vc}(\mathcal{H})$. Now, how can the one-inclusion graph be used as a learner? Given a training set of examples $S = \{(x_1, y_1), \dots, (x_{n-1}, y_{n-1})\}$ and a test example $x_n$, we construct the one-inclusion graph on $\{x_1, \dots, x_{n-1}\} \cup \{x_n\}$ using the class $\mathcal{H}$ and orient it such that maximum out-degree is at most $\mathrm{vc}(\mathcal{H})$. Then, we use the orientation to predict the label of the test point $x_n$. Specifically, if there exists $h, h' \in \mathcal{H}$ such that $\forall 1 \leq i \leq n-1 : h(x_i) = h'(x_i)$ and $h(x_n) \neq h'(x_n)$ then we will have two vertices in the

*At a first glance, it might seem that adversarially robust learning can be viewed as a special case of learning partial concepts classes (Alon et al., 2021), but we would like to remark that this is *not* the case. The apparent similarity arises because it is possible to state the robust realizability assumption in the language of partial concept classes, as in the example mentioned in Alon et al. (2021) on learning linear separators with a margin, but this is just an assumption on the data-distribution. Specifically, a partial concept class learner is *only* guaranteed to make few errors on samples drawn from the distribution (see Definition 2 in Alon et al., 2021), and not on their adversarial perturbations: i.e., performance is still measured under 0-1 loss, not robust risk.

graph $v = (h(x_1), \ldots, h(x_{n-1}), h(x_n))$ and $u = (h(x_1), \ldots, h(x_{n-1}), h'(x_n))$ with an edge connecting them (because they differ only in the label of $x_n$), and we predict the label of $x_n$ that this edge is directed towards. Since each vertex has out-degree at most $\text{vc}(\mathcal{H})$, this implies that the average leave-one-out-error (which bounds the expected risk from above) is at most $\frac{\text{vc}(\mathcal{H})}{n}$.

What breaks in the adversarial learning setting? At test-time, we do not observe an i.i.d. test example $x \sim \mathcal{D}$ but rather only an adversarially chosen perturbation $z \in \mathcal{U}(x)$. This completely breaks the exchangeability analysis of the classical one-inclusion graph, because the training points are i.i.d. but the perturbation $z$ of the test point $x$ is not. Furthermore, the classical one-inclusion graph is *local* in the sense that it depends on the training data and the test point, and as such different perturbations $z, \tilde{z} \in \mathcal{U}(x)$ could very well lead to different graphs, different orientations, and ultimately different predictions for $z$ and $\tilde{z}$ which by definition imply that the prediction is not robust on $x$.

### 4.3.2 Our Global One-Inclusion Graph

We now describe the construction of the global one-inclusion graph. For any class $\mathcal{H}$, any perturbation set $\mathcal{U}$, and any dataset size $n \in \mathbb{N}$, denote by $G^{\mathcal{U}}_{\mathcal{H}} = (V_n, E_n)$ the global one-inclusion graph. In words, $V_n$ is the collection of all datasets of size $n$ that can be *robustly* labeled by class $\mathcal{H}$ with respect to perturbation set $\mathcal{U}$. Formally, each vertex $v \in V_n$ is represented as a *multiset* of labeled examples $(x, y)$ of size $n$:[†]

$$V_n = \{\{(x_1, y_1), \ldots, (x_n, y_n)\} : (\exists h \in \mathcal{H})\,(\forall i \in [n])\,(\forall z \in \mathcal{U}(x_i))\,, h(z) = y_i\}\,. \quad (4.6)$$

Two vertices (datasets) $u, v \in V_n$ are connected by an edge if and only if there is a unique labeled example $(x, y) \in v$ that does not appear in $u$ and there is a unique labeled example $(\tilde{x}, \tilde{y}) \in u$ that does not appear in $v$ satisfying: $y \neq \tilde{y}$ and $\mathcal{U}(x) \cap \mathcal{U}(\tilde{x}) \neq \emptyset$. Formally, $u, v \in V_n$ are connected by an edge if and only if their symmetric difference $u \Delta v = \{(x, y), (\tilde{x}, \tilde{y})\}$ where $y \neq \tilde{y}$ and $\mathcal{U}(x) \cap \mathcal{U}(\tilde{x}) \neq \emptyset$. Furthermore, we will label an edge by the perturbation

[†]Note that we allow a labeled example $(x, y)$ to appear more than once in a vertex $v$, hence the multiset representation.

$z \in \mathcal{U}(x) \cap \mathcal{U}(\tilde{x})$ that witnesses this edge:

$$E_n = \{(\{u, v\}, z) : u, v \in V_n \wedge u \Delta v = \{(x, y), (\tilde{x}, \tilde{y})\} \wedge y \neq \tilde{y} \wedge z \in \mathcal{U}(x) \cap \mathcal{U}(\tilde{x})\}. \tag{4.7}$$

For each vertex $v \in V_n$, denote by adv-deg$(v)$ the adversarial degree of $v$ which is defined as the number of elements $(x, y) \in v$ that witness an edge incident on $v$:

$$\text{adv-deg}(v) = |\{(x, y) \in v : \exists u \in V_n, z \in \mathcal{X} \text{ s.t. } (\{v, u\}, z) \in E_n \wedge (x, y) \in v \Delta u\}|. \tag{4.8}$$

We want to emphasize that our notion of *adversarial degree* is different from the standard notion of degree used in graph theory, and in particular different from the degree notion in the classical one-inclusion graph used above. Specifically, we do *not* count all edges incident on a vertex rather we count the number of datapoints $(x, y)$ in a vertex that witness an edge. This different notion of degree is more suitable for our purposes and is related to the average leave-one-out *robust* error.

### 4.3.3 From Orientations to Learners

An orientation $\mathcal{O} : E_n \to V_n$ of the global one-inclusion graph $G_{\mathcal{H}}^{\mathcal{U}} = (V_n, E_n)$ is a mapping that directs each edge $e = (\{u, v\}, z) \in E_n$ towards a vertex $\mathcal{O}(e) \in \{u, v\}$. Given an orientation $\mathcal{O} : E_n \to V_n$ of the global one-inclusion graph $G_{\mathcal{H}}^{\mathcal{U}}$, the adversarial out-degree of a vertex $v \in V$, denoted by adv-outdeg$(v; \mathcal{O})$, is defined as the number of elements $(x, y) \in v$ that witness an out-going edge incident on $v$ according to orientation $\mathcal{O}$:

$$\text{adv-outdeg}(v; \mathcal{O}) = \left| \left\{ (x, y) \in v \;\middle|\; \begin{array}{l} \exists u \in V_n, z \in \mathcal{X} \text{ s.t. } (\{v, u\}, z) \in E_n \wedge \\ (x, y) \in v \Delta u \wedge \mathcal{O}((\{v, u\}, z)) = u \end{array} \right\} \right|. \tag{4.9}$$

Why are we interested in orientations of the global one-inclusion graph $G_{\mathcal{H}}^{\mathcal{U}}$? We show next that every orientation of $G_{\mathcal{H}}^{\mathcal{U}}$ can be used to construct a learner, and that the expected robust risk of this learner is bounded from above by the maximum adversarial out-degree of the corresponding orientation. We will use this observation later in Section 4.4 to construct an optimal learner.

**Lemma 4.2.** *For any class $\mathcal{H}$, any perturbation set $\mathcal{U}$, and any $n > 1$, let $G^{\mathcal{U}}_{\mathcal{H}} = (V_n, E_n)$ be the global one-inclusion graph. Then, for any orientation $\mathcal{O} : E_n \to V_n$ of $G^{\mathcal{U}}_{\mathcal{H}}$, there exists a learner $\mathbb{A}_{\mathcal{O}} : (\mathcal{X} \times \mathcal{Y})^{n-1} \to \mathcal{Y}^{\mathcal{X}}$, such that*

$$\mathcal{E}_{n-1}(\mathbb{A}_{\mathcal{O}}; \mathcal{H}, \mathcal{U}) \leq \frac{\max_{v \in V_n} \text{adv-outdeg}(v; \mathcal{O})}{n}.$$

The proof is deferred to Appendix B.1. At a high-level, we can use an orientation $\mathcal{O}$ of $G^{\mathcal{U}}_{\mathcal{H}}$ to make predictions in the following way: upon receiving training examples $S$ and a (possibly adversarial) test instance $z$, we consider all possible natural datapoints $(x, y)$ of which $z$ is a perturbation of $x$ (i.e., $z \in \mathcal{U}(x)$) such that $S \cup \{(x, y)\}$ can be labeled robustly using class $\mathcal{H}$ with respect to $\mathcal{U}$ (note that these are all vertices in $G^{\mathcal{U}}_{\mathcal{H}}$ by definition), and if two different robust labelings of $z$ are possible, the orientation $\mathcal{O}$ determines which label to predict. This is defined explicitly in Algorithm 4.1.

**Algorithm 4.1:** Converting an Orientation $\mathcal{O}$ of $G^{\mathcal{U}}_{\mathcal{H}}$ to a Learner $\mathbb{A}_{\mathcal{O}}$.

**Input:** Training dataset $S = \{(x_1, y_1), \ldots, (x_{n-1}, y_{n-1})\} \in (\mathcal{X} \times \mathcal{Y})^{n-1}$, test instance $z \in \mathcal{X}$, and an orientation $\mathcal{O} : E_n \to V_n$ of $G^{\mathcal{U}}_{\mathcal{H}} = (E_n, V_n)$.

Let
$$P_+ = \{v \in V_n : \exists x \in \mathcal{X} \text{ s.t. } z \in \mathcal{U}(x) \wedge v = \{(x_1, y_1), \ldots, (x_{n-1}, y_{n-1}), (x, +1)\}\}.$$

Let
$$P_- = \{v \in V_n : \exists x \in \mathcal{X} \text{ s.t. } z \in \mathcal{U}(x) \wedge v = \{(x_1, y_1), \ldots, (x_{n-1}, y_{n-1}), (x, -1)\}\}.$$

**If** $\left(\exists_{y \in \{\pm 1\}}\right) \left(\exists_{v \in P_y}\right) \left(\forall_{u \in P_{-y}}\right) : \mathcal{O}((\{v, u\}, z)) = v$, **then** output label $y$.

**Otherwise**, output $+1$.

## 4.4 A Generic Minimax Optimal Learner

We now present an optimal robust learner based on our global one-inclusion graph from Section 4.3.

For any class $\mathcal{H}$, any perturbation set $\mathcal{U}$, and integer $n > 1$, let $G^{\mathcal{U}}_{\mathcal{H}} = (V_n, E_n)$ be the global one-inclusion graph (Equations 4.6 and 4.7). Let $\mathcal{O}^*$ be an orientation that minimizes the maximum adversarial out-degree of $G^{\mathcal{U}}_{\mathcal{H}}$:

$$\mathcal{O}^* \in \operatorname*{argmin}_{\mathcal{O}:E_n \to V_n} \max_{v \in V_n} \text{adv-outdeg}(v; \mathcal{O}). \tag{4.10}$$

Then, let $\mathbb{G}_{\mathcal{H},\mathcal{U}}$ be the learner implied by orientation $\mathcal{O}^*$ as described in Algorithm 4.1.

**Theorem 4.2.** *For any $\mathcal{H}, \mathcal{U}$, any $n \in \mathbb{N}$, learner $\mathbb{G}_{\mathcal{H},\mathcal{U}}$ described above satisfies for any learner $\mathbb{A}$:*

$$\mathcal{E}_{2n-1}(\mathbb{G}_{\mathcal{H},\mathcal{U}}; \mathcal{H}, \mathcal{U}) \leq 4 \cdot \mathcal{E}_n(\mathbb{A}; \mathcal{H}, \mathcal{U}), \text{ \& equivalently } \mathcal{M}^{\mathrm{re}}_{\varepsilon}(\mathbb{G}_{\mathcal{H},\mathcal{U}}; \mathcal{H}, \mathcal{U}) \leq 2 \cdot \mathcal{M}^{\mathrm{re}}_{\varepsilon/4}(\mathbb{A}; \mathcal{H}, \mathcal{U}) - 1.$$

Before proceeding to the proof of Theorem 4.2, we first prove a key Lemma which basically shows that we can use an arbitrary learner $\mathbb{A}$ to orient the edges in the global one-inclusion graph $G^{\mathcal{U}}_{\mathcal{H}}$, and that the maximum adversarial out-degree of the resultant orientation is upper bounded by the robust error rate of $\mathbb{A}$.

**Lemma 4.3** (Lowerbound on Error Rate of Learners). *Let $\mathbb{A} : (\mathcal{X} \times \mathcal{Y})^* \to \mathcal{Y}^{\mathcal{X}}$ be any learner, and $n \in \mathbb{N}$. Let $G^{\mathcal{U}}_{\mathcal{H}} = (V_{2n}, E_{2n})$ be the global one-inclusion graph as defined in Equation 4.6 and Equation 4.7. Then, there exists an orientation $\mathcal{O}_{\mathbb{A}} : E_{2n} \to V_{2n}$ of $G^{\mathcal{U}}_{\mathcal{H}}$ such that*

$$\mathcal{E}_n(\mathbb{A}; \mathcal{H}, \mathcal{U}) \geq \frac{1}{4} \frac{\max_{v \in V_{2n}} \text{adv-outdeg}(v; \mathcal{O}_{\mathbb{A}})}{2n}.$$

*Proof.* We begin with describing the orientation $\mathcal{O}_{\mathbb{A}}$ by orienting edges incident on each vertex $v \in V_{2n}$. Consider an arbitrary vertex $v = \{(x_1, y_1), \dots, (x_{2n}, y_{2n})\}$ and let $P_v$ be a uniform distribution over $(x_1, y_1), \dots, (x_{2n}, y_{2n})$. For each $1 \leq t \leq 2n$, let

$$p_t(v) = \Pr_{S \sim P_v^n} \left[ \exists z \in \mathcal{U}(x_t) : \mathbb{A}(S)(z) \neq y_t | (x_t, y_t) \notin S \right].$$

For each $(x_t, y_t) \in v$ that witnesses an edge, i.e. $\exists u \in V_{2n}, z \in \mathcal{X}$ s.t. $(\{v, u\}, z) \in E_{2n}$ and $(x_t, y_t) \in v \Delta u$, if $p_t < \frac{1}{2}$, then orient *all* edges incident on $(x_t, y_t)$ inward, otherwise orient them arbitrarily. Note that this might yield edges that are oriented outwards from both their endpoint vertices, in which case, we arbitrarily orient such an edge. Observe also that we will not encounter a situation where edges are oriented inwards towards both their endpoints (which is an invalid orientation). This is because for any two vertices $v, u \in V_{2n}$ such that $\exists z_0 \in \mathcal{X}$ where $(\{u, v\}, z_0) \in E_{2n}$ and $v \Delta u = \{(x_t, y_t), (\tilde{x}_t, -y_t)\}$, we cannot have $p_t(v) < \frac{1}{2}$ and $p_t(u) < \frac{1}{2}$, since

$$p_t(v) \geq \Pr_{S \sim P_v^m}\left[\mathbb{A}(S)(z_0) \neq y_t | (x_t, y_t) \notin S\right] \text{ and } p_t(u) \geq \Pr_{S \sim P_u^m}\left[\mathbb{A}(S)(z_0) \neq -y_t | (\tilde{x}_t, -y_t) \notin S\right],$$

and $P_v$ conditioned on $(x_t, y_t) \notin S$ is the same distribution as $P_u$ conditioned on $(\tilde{x}_t, -y_t) \notin S$. This concludes describing the orientation $\mathcal{O}_{\mathbb{A}}$. We now bound the adversarial out-degree of vertices $v \in V_{2n}$:

$$\begin{aligned}
\text{adv-outdeg}(v; \mathcal{O}_{\mathbb{A}}) &\leq \sum_{t=1}^{2n} 1\left[p_t \geq \frac{1}{2}\right] \leq 2 \sum_{t=1}^{2n} p_t = 2 \sum_{t=1}^{2n} \Pr_{S \sim P^n}\left[\exists z \in \mathcal{U}(x_t) : \mathbb{A}(S)(z) \neq y_t | (x_t, y_t) \notin S\right] \\
&= 2 \sum_{t=1}^{2n} \frac{\Pr_{S \sim P^n}\left[(\exists z \in \mathcal{U}(x_t) : \mathbb{A}(S)(z) \neq y_t) \wedge (x_t, y_t) \notin S\right]}{\Pr_{S \sim P^n}\left[(x_t, y_t) \notin S\right]} \\
&\leq 4 \sum_{t=1}^{2n} \Pr_{S \sim P^n}\left[(\exists z \in \mathcal{U}(x_t) : \mathbb{A}(S)(z) \neq y_t) \wedge (x_t, y_t) \notin S\right] = 4 \mathop{\mathbb{E}}_{S \sim P^n} \sum_{(x_t, y_t) \notin S} 1\left[\exists z \in \mathcal{U}(x_t) : \mathbb{A}(S)(z) \neq y_t\right] \\
&\leq 4 \mathop{\mathbb{E}}_{S \sim P^n} \sum_{t=1}^{2n} 1\left[\exists z \in \mathcal{U}(x_t) : \mathbb{A}(S)(z) \neq y_t\right] = 8n \mathop{\mathbb{E}}_{S \sim P^n} \mathrm{R}_{\mathcal{U}}(\mathbb{A}(S); P) \leq 8n \mathcal{E}_n(\mathbb{A}; \mathcal{H}, \mathcal{U}).
\end{aligned}$$

Since the above holds for any vertex $v \in V_{2n}$, by rearranging terms, we get $\mathcal{E}_n(\mathbb{A}; \mathcal{H}, \mathcal{U}) \geq \frac{1}{4} \frac{\max_{v \in V_{2n}} \text{adv-outdeg}(v; \mathcal{O}_{\mathbb{A}})}{2n}$. □

We are now ready to proceed with the proof of Theorem 4.2.

*Proof of Theorem 4.2.* By invoking Lemma 4.3, we have that for any learner $\mathbb{A}$,

$$\mathcal{E}_n(\mathbb{A}; \mathcal{H}, \mathcal{U}) \geq \frac{1}{4} \frac{\max_{v \in V_{2n}} \text{adv-outdeg}(v; \mathcal{O}_{\mathbb{A}})}{2n}.$$

By Equation 4.10, an orientation $\mathcal{O}^*$ has smaller maximum adversarial out-degree, thus

$$\frac{1}{4} \frac{\max_{v \in V_{2n}} \text{adv-outdeg}(v; \mathcal{O}_{\mathbb{A}})}{2n} \geq \frac{1}{4} \frac{\max_{v \in V_{2n}} \text{adv-outdeg}(v; \mathcal{O}^*)}{2n}.$$

By invoking Lemma 4.2, it follows that our optimal learner $\mathbb{G}_{\mathcal{H},\mathcal{U}}$ satisfies

$$\frac{1}{4}\frac{\max_{v\in V_{2n}} \text{adv-outdeg}(v;\mathcal{O}^*)}{2n} \geq \frac{\mathcal{E}_{2n-1}(\mathbb{G}_{\mathcal{H},\mathcal{U}};\mathcal{H},\mathcal{U})}{4}.$$

We arrive at the theorem statement by chaining the above inequalities and rearranging terms. □

## 4.5 A Complexity Measure and Sample Complexity Bounds

In Section 4.4, we showed how our global one-inclusion graph yields a near-optimal learner for adversarially robust learning. We now turn to characterizing adversarially robust learnability.

Across learning theory, many fundamental learning problems can be surprisingly characterized by means of combinatorial complexity measures. Such characterizations are often quantitatively insightful in that they provide tight bounds on the number of examples needed for learning, and also insightful for algorithm design. For example, for standard (non-robust) learning, the VC dimension characterizes what classes $\mathcal{H}$ are PAC learnable (Vapnik and Chervonenkis, 1971, 1974, Blumer, Ehrenfeucht, Haussler, and Warmuth, 1989a, Ehrenfeucht, Haussler, Kearns, and Valiant, 1989). For multi-class learning, there are characterizations based on the Natarjan and Graph dimensions, and the Daniely-Shalev-Shwartz (DS) (Natarajan, 1989, Daniely, Sabato, Ben-David, and Shalev-Shwartz, 2015, Brukhim, Carmon, Dinur, Moran, and Yehudayoff, 2022). For learning real-valued functions, the fat-shattering dimension plays a similar role (Alon, Ben-David, Cesa-Bianchi, and Haussler, 1997, Kearns and Schapire, 1994, Simon, 1997). The Littlestone dimension characterizes online learnability (Littlestone, 1987), and the star number characterizes the label complexity of active learning (Hanneke and Yang, 2015).

Shalev-Shwartz, Shamir, Srebro, and Sridharan (2010) showed that in Vapnik's "General Learning" problem (Vapnik, 1982), the loss class having finite VC dimension is sufficient but not, in general, necessary for learnability and asked whether there is another dimension that characterizes learnability in this setting. But recently, Ben-David, Hrubes, Moran, Shpilka, and Yehudayoff (2019) surprisingly exhibited a statistical learning problem that can not be characterized with a combinatorial VC-like dimension. In order to do so, they presented a *formal* definition of the notion of "dimension" or "complexity measure" (see Definition B.4), that all previously proposed dimensions in statistical learning theory comply with. This raises the following natural question:

*Is there a dimension that characterizes robust learnability, and if so, what is it?!*

### 4.5.1 A Dimension Characterizing Robust Learning

We present next a dimension for adversarially robust learnability, which is inspired by our global one-inclusion graph described in Section 4.3.

$$\mathfrak{D}_{\mathcal{U}}(\mathcal{H}) = \max \left\{ n \in \mathbb{N} \cup \{\infty\} \;\middle|\; \begin{array}{l} \exists \text{ a finite subgraph } G = (V, E) \text{ of } G^{\mathcal{U}}_{\mathcal{H}} = (V_n, E_n) \text{ s.t.} \\ \forall \text{ orientations } \mathcal{O} \text{ of } G, \exists v \in V \text{ where adv-outdeg}(v; \mathcal{O}) \geq \frac{n}{3}. \end{array} \right\}. \tag{4.11}$$

In Appendix B.4, we discuss how our dimension satisfies the formal definition proposed by (Ben-David et al., 2019). We now show that $\mathfrak{D}_{\mathcal{U}}(\mathcal{H})$ characterizes robust learnability *qualitatively* and *quantitatively*.

**Theorem 4.3** (Qualitative Characterization). *For any class $\mathcal{H}$ and any perturbation set $\mathcal{U}$, $\mathcal{H}$ is robustly PAC learnable with respect to $\mathcal{U}$* if and only if *$\mathfrak{D}_{\mathcal{U}}(\mathcal{H})$ is finite.*

**Theorem 4.4** (Quantitative Characterization). *For any class $\mathcal{H}$ and any perturbation set $\mathcal{U}$,*

$$\Omega\left(\frac{\mathfrak{D}_{\mathcal{U}}(\mathcal{H})}{\varepsilon}\right) \leq \mathcal{M}^{\text{re}}_{\varepsilon,\delta}(\mathcal{H},\mathcal{U}) \leq O\left(\frac{\mathfrak{D}_{\mathcal{U}}(\mathcal{H})}{\varepsilon}\log^2\frac{\mathfrak{D}_{\mathcal{U}}(\mathcal{H})}{\varepsilon} + \frac{\log(1/\delta)}{\varepsilon}\right).$$

Theorem 4.3 follows immediately from Theorem 4.4. To prove Theorem 4.4, we first prove the following key Lemma which provides upper and lower bounds on the *minimax* expected robust risk of learning a class $\mathcal{H}$ with respect to a perturbation set $\mathcal{U}$ (see Equation 4.2) as a function of our introduced dimension $\mathfrak{D}_{\mathcal{U}}(\mathcal{H})$. Theorem 4.4 follows from an argument to boost the robust risk and the confidence as appeared in Chapter 3, Section 3.3. The proofs are deferred to Appendix B.2.

**Lemma 4.4.** *For any class $\mathcal{H}$, any perturbation set $\mathcal{U}$, and any $\varepsilon \in (0,1)$,*

*1.* $\forall n > \mathfrak{D}_{\mathcal{U}}(\mathcal{H}) : \mathcal{E}_{n-1}(\mathcal{H},\mathcal{U}) \leq \frac{1}{3}$.

2. $\forall 2 \leq n \leq \frac{\mathfrak{D}_{\mathcal{U}}(\mathcal{H})}{2}: \mathcal{E}_{\frac{n}{\varepsilon}}(\mathcal{H},\mathcal{U}) \geq \frac{\varepsilon}{54}$.

### 4.5.2 Examples

We discuss a few ways of estimating or calculating our proposed dimension $\mathfrak{D}_{\mathcal{U}}(\mathcal{H})$.

**Proposition 4.5.** *For any class $\mathcal{H}$ and perturbation set $\mathcal{U}$:*

$$\mathfrak{D}_{\mathcal{U}}(\mathcal{H}) \leq \min\left\{\tilde{O}(\mathrm{vc}(\mathcal{H})\mathrm{vc}^*(\mathcal{H})), \tilde{O}(\mathrm{vc}(\mathcal{L}^{\mathcal{U}}_{\mathcal{H}}))\right\},$$

*where* $\mathrm{vc}^*(\mathcal{H})$ *denotes the* dual *VC dimension, and* $\mathrm{vc}(\mathcal{L}^{\mathcal{U}}_{\mathcal{H}}))$ *denotes the VC dimension of the robust-loss class* $\mathcal{L}^{\mathcal{U}}_{\mathcal{H}} = \left\{(x,y) \mapsto \sup_{z\in\mathcal{U}(x)} 1[h(z \neq y)] : h \in \mathcal{H}\right\}$.

*Proof.* Set $\varepsilon_0 = \frac{1}{3}$. We know from Theorem 4.4 that $\mathcal{M}^{\mathrm{re}}_{\varepsilon_0}(\mathcal{H},\mathcal{U}) \geq \Omega(\mathfrak{D}_{\mathcal{U}}(\mathcal{H}))$. We also know from Theorem 3.4 that $\mathcal{M}^{\mathrm{re}}_{\varepsilon_0}(\mathcal{H},\mathcal{U}) \leq \tilde{O}(\mathrm{vc}(\mathcal{H})\mathrm{vc}^*(\mathcal{H}))$. Finally, we know from (Theorem 1 in Cullina, Bhagoji, and Mittal, 2018a) that $\mathcal{M}^{\mathrm{re}}_{\varepsilon_0}(\mathcal{H},\mathcal{U}) \leq \tilde{O}(\mathrm{vc}(\mathcal{L}^{\mathcal{U}}_{\mathcal{H}}))$. Combining these together yields that stated bound. □

The dual VC dimension satisfies: $\mathrm{vc}^*(\mathcal{H}) < 2^{\mathrm{vc}(\mathcal{H})+1}$ (Assouad, 1983), and this exponential dependence is tight for some classes. For many natural classes, however, such as linear predictors and some neural networks, the primal and dual VC dimensions are equal, or at least polynomially related (see, e.g. Lemma 6.3). Using Proposition 4.5, we can conclude that for such classes $\mathfrak{D}_{\mathcal{U}}(\mathcal{H}) \leq \mathrm{poly}(\mathrm{vc}(\mathcal{H}))$, specifically, for $\mathcal{H}$ being linear predictors, $\mathfrak{D}_{\mathcal{U}}(\mathcal{H}) \leq \tilde{O}(\mathrm{vc}^2(\mathcal{H}))$. Furthermore, for $\mathcal{H}$ being linear predictors and $\mathcal{U} = \ell_p$ perturbations, we know that $\mathrm{vc}(\mathcal{L}^{\mathcal{U}}_{\mathcal{H}}) = O(\mathrm{vc}(\mathcal{H}))$ (Theorem 2 in Cullina, Bhagoji, and Mittal, 2018a), and so using Proposition 4.5 again, we get a tighter bound for these $\ell_p$ perturbations $\mathfrak{D}_{\mathcal{U}}(\mathcal{H}) \leq \tilde{O}(\mathrm{vc}(\mathcal{H}))$.

While Proposition 4.5 is certainly useful for estimating our dimension $\mathfrak{D}_{\mathcal{U}}(\mathcal{H})$, we get vacuous bounds when the VC dimension $\mathrm{vc}(\mathcal{H})$ is infinite. To this end, recall the $(\mathcal{H},\mathcal{U})$ examples in Example 4.1 and Example 4.2 mentioned in Section 4.1, which satisfy $\mathrm{vc}(\mathcal{H}) = \infty$. We can calculate $\mathfrak{D}_{\mathcal{U}}(\mathcal{H})$ for these examples differently. In particular, in Example 4.1, by definition, the global one-inclusion graph $G^{\mathcal{U}}_{\mathcal{H}} = (V_n, E_n)$ has no edges when $n > 1$ because $\mathcal{U}(x) = \mathcal{X}$ and thus $\mathfrak{D}_{\mathcal{U}}(\mathcal{H}) \leq 1$. In Example 4.2, we get that $\mathfrak{D}_{\mathcal{U}}(\mathcal{H}) \leq \tilde{O}(d)$ since we can robustly learn with $O(d)$ samples, but we can also calculate $\mathfrak{D}_{\mathcal{U}}(\mathcal{H})$ directly by constructing the global one-inclusion graph $G^{\mathcal{U}}_{\mathcal{H}} = (V_n, E_n)$ for $n > 3d$ and observing that we can orient $G^{\mathcal{U}}_{\mathcal{H}}$ such that the adversarial out-degree is at most $d$, which is possible because of the definition of $\mathcal{U}$.

### 4.5.3 Agnostic Robust Learnability

For the agnostic setting, we consider robust learnability with respect to arbitrary distributions $\mathcal{D}$ that are *not* necessarily robustly realizable, i.e., $\mathcal{D} \notin \mathrm{RE}(\mathcal{H}, \mathcal{U})$ (see Definition 2.1). As we did in Chapter 3, Subsection 3.3.2, we can establish an upper bound in the agnostic setting via reduction to the realizable case, following an argument from David, Moran, and Yehudayoff (2016).

**Theorem 4.6.** *For any class $\mathcal{H}$ and any perturbation set $\mathcal{U}$,*

$$\mathcal{M}^{\mathrm{ag}}_{\varepsilon,\delta}(\mathcal{H}, \mathcal{U}) = O\left(\frac{\mathfrak{D}_{\mathcal{U}}(\mathcal{H})}{\varepsilon^2} \log^2\left(\frac{\mathfrak{D}_{\mathcal{U}}(\mathcal{H})}{\varepsilon}\right) + \frac{1}{\varepsilon^2}\log\left(\frac{1}{\delta}\right)\right).$$

This is achieved by applying the agnostic-to-realizable reduction to the optimal learner $\mathbb{G}_{\mathcal{H},\mathcal{U}}$ that we get from orienting the graph $G^{\mathcal{U}}_{\mathcal{H}} = (V_{\mathfrak{D}_{\mathcal{U}}(\mathcal{H})+1}, E_{\mathfrak{D}_{\mathcal{U}}(\mathcal{H})+1})$. The reduction is stated abstractly in the following Lemma whose proof is provided in Appendix B.3.

**Lemma 4.5.** *For any well-defined realizable learner $\mathbb{A}$, there is an agnostic learner $\mathbb{B}$ such that*

$$\mathcal{M}^{\mathrm{re}}_{\varepsilon}(\mathbb{A}; \mathcal{H}, \mathcal{U}) \leq \mathcal{M}^{\mathrm{ag}}_{\varepsilon,\delta}(\mathbb{B}; \mathcal{H}, \mathcal{U}) \leq O\left(\frac{\mathcal{M}^{\mathrm{re}}_{1/3}(\mathbb{A}; \mathcal{H}, \mathcal{U})}{\varepsilon^2} \log^2\left(\frac{\mathcal{M}^{\mathrm{re}}_{1/3}(\mathbb{A}; \mathcal{H}, \mathcal{U})}{\varepsilon}\right) + \frac{1}{\varepsilon^2}\log\left(\frac{1}{\delta}\right)\right).$$

Theorem 4.6 immediately follows by combining Lemma 4.5 and Theorem 4.4.

## 4.6 Conjectures

While we have shown that our proposed dimension $\mathfrak{D}_{\mathcal{U}}(\mathcal{H})$ in Equation 4.11 characterizes robust learnability, we believe that there are other *equivalent* dimensions that are *simpler* to describe. For a more appealing dimension, we may take inspiration from recent progress on multi-class learning (Brukhim, Carmon, Dinur, Moran, and Yehudayoff, 2022), where it was shown that the DS dimension due to Daniely and Shalev-Shwartz (2014) characterizes multi-class learnability. For a class $\mathcal{H} \subseteq \mathcal{Y}^{\mathcal{X}}$ ($|\mathcal{Y}| > 2$), the DS dimension corresponds to the largest $n$ s.t. there exists points $P \in \mathcal{X}^n$ where the projection of $\mathcal{H}$ onto $P$ induces a one-inclusion hyper graph where every vertex has full-degree. This inspires the *full-degree* dimension of the global one-inclusion graph:

$$\mathfrak{F}\mathfrak{D}_{\mathcal{U}}(\mathcal{H}) = \max \left\{ n \in \mathbb{N} \cup \{\infty\} \,\middle|\, \begin{array}{l} \exists \text{ a finite subgraph } G = (V, E) \text{ of } G^{\mathcal{U}}_{\mathcal{H}} = (V_n, E_n) \text{ s.t. every} \\ \text{vertex has full-degree: } \forall v \in V, \text{adv-deg}(v; E) \geq n. \end{array} \right\}.$$

This complexity measure avoids orientations, and thus, it is perhaps simpler to verify "$\mathfrak{F}\mathfrak{D}_{\mathcal{U}}(\mathcal{H}) \geq d$" than "$\mathfrak{D}_{\mathcal{U}}(\mathcal{H}) \geq d$". Furthermore, when $\mathcal{U}(x) = \{x\}\, \forall x \in \mathcal{X}$, the full-degree dimension, $\mathfrak{F}\mathfrak{D}_{\mathcal{U}}(\mathcal{H})$, corresponds exactly to the VC dimension of $\mathcal{H}$, $\text{vc}(\mathcal{H})$.

**Conjecture 4.1.** *For any class $\mathcal{H}$ and perturbation set $\mathcal{U}$, $\mathcal{M}^{\text{re}}_{\varepsilon,\delta}(\mathcal{H}, \mathcal{U}) = \Theta_{\varepsilon,\delta}\left(\mathfrak{F}\mathfrak{D}_{\mathcal{U}}(\mathcal{H})\right)$.*

In Chapter 3, Section 3.4, we proposed the following combinatorial robust shattering dimension, denoted $\dim_{\mathcal{U}}(\mathcal{H})$, and showed that $\mathcal{M}^{\text{re}}_{\varepsilon,\delta}(\mathcal{H}, \mathcal{U}) \geq \Omega_{\varepsilon,\delta}(\dim_{\mathcal{U}}(\mathcal{H}))$ (Theorem 3.11).

**Definition 4.6** (Robust Shattering Dimension). *A sequence $z_1, \dots, z_k \in \mathcal{X}$ is said to be $\mathcal{U}$-*robustly shattered *by $\mathcal{H}$ if $\exists x_1^+, x_1^-, \dots, x_k^+, x_k^- \in \mathcal{X}$ s.t. $\forall i \in [k], z_i \in \mathcal{U}(x_i^+) \cap \mathcal{U}(x_i^-)$ and $\forall y_1, \dots, y_k \in \{\pm 1\} : \exists h \in \mathcal{H}$ such that $h(z') = y_i \forall z' \in \mathcal{U}(x_i^{y_i}) \forall 1 \leq i \leq k$. The $\mathcal{U}$-*robust shattering dimension $\dim_{\mathcal{U}}(\mathcal{H})$ *is defined as the largest $k$ for which there exist $k$ points $\mathcal{U}$-robustly shattered by $\mathcal{H}$.*

In regards to the relationship between the robust shattering dimension $\dim_{\mathcal{U}}(\mathcal{H})$ and the new dimension $\mathfrak{D}_{\mathcal{U}}(\mathcal{H})$ proposed in this chapter (Equation 4.11), we conjecture that $\mathfrak{D}_{\mathcal{U}}(\mathcal{H})$ can be arbitrarily larger than $\dim_{\mathcal{U}}(\mathcal{H})$. In other words, we conjecture that the robust shattering dimension $\dim_{\mathcal{U}}(\mathcal{H})$ does not characterize robust learnability.

**Conjecture 4.2.** $\forall n \in \mathbb{N}, \exists \mathcal{X}, \mathcal{H}, \mathcal{U}$, *such that* $\dim_{\mathcal{U}}(\mathcal{H}) = O(1)$ *but* $\mathfrak{D}_{\mathcal{U}}(\mathcal{H}) \geq n$.

We find this to be analogous to a separation in multi-class learnability, where the Natarajan dimension was shown to not characterize multi-class learnability (Brukhim et al., 2022).

Because in both graphs, the one-inclusion hyper graph and our global one-inclusion graph, the Natarajan and robust shattering dimensions represent a "cube" in their corresponding graph, while the DS dimension and our full-degree dimension represent a "pseudo-cube" in the terminology of Brukhim et al. (2022).

Another interesting and perhaps useful direction to explore is the relationship between our proposed complexity measure $\mathfrak{D}_{\mathcal{U}}(\mathcal{H})$ and the VC dimension. We believe that it is actually possible to orient the global one-inclusion graph such that the maximum adversarial out-degree is at most $O(\mathrm{vc}(\mathcal{H}))$.

**Conjecture 4.3.** *For any class $\mathcal{H}$ and perturbation set $\mathcal{U}$, $\mathfrak{D}_{\mathcal{U}}(\mathcal{H}) \leq O(\mathrm{vc}(\mathcal{H}))$.*

# 5

# Boosting Barely Robust Learners

## 5.1 Introduction

Adversarially robust learning has proven to be quite challenging in practice, where current adversarial learning methods typically learn predictors with low natural error but robust only on a small fraction of the data. For example, according to the RobustBench leaderboard (Croce, Andriushchenko, Sehwag, Debenedetti, Flammarion, Chiang, Mittal, and Hein, 2020), the highest achieved robust accuracy with respect to $\ell_\infty$ perturbations on CIFAR10 is $\approx 66\%$ and on ImageNet is $\approx 38\%$. Can we leverage existing methods and go beyond their limits? This motivates us to pursue the idea of *boosting* robustness, and study the following theoretical question:

*Can we boost* barely *robust learning algorithms to learn predictors with high* robust *accuracy?*

That is, given a *barely* robust learning algorithm $\mathbb{A}$ which can only learn predictors robust on say $\beta = 10\%$ fraction of the data distribution, we are asking whether it is possible to *boost* the robustness of $\mathbb{A}$ and learn predictors with high *robust* accuracy, say $90\%$. We want to emphasize that we are interested here in extreme situations when the robustness parameter $\beta \ll 1$. We are interested in generic boosting algorithms that take as input a black-box learner $\mathbb{A}$ and a specification of the perturbation set $\mathcal{U}$, and output a predictor with high robust accuracy by repeatedly calling $\mathbb{A}$.

In this chapter, by studying the question above, we offer a new perspective on adversarial robustness. Specifically, we discover a qualitative and quantitative *equivalence* between two seemingly unrelated problems: strongly robust learning and barely robust learning. We show that barely robust learning implies strongly robust learning through a novel algorithm for *boosting* robustness. As we elaborate below, our proposed notion of barely robust learning requires robustness with respect to a "larger" perturbation set. We also show that this is *necessary* for strongly robust learning, and that weaker relaxations of barely robust learning do not imply strongly robust learning.

### 5.1.1 Main Contributions

When formally studying the problem of boosting robustness, an important question emerges which is: what notion of "barely robust" learning is required for boosting robustness? As we shall show, this is not immediately obvious. One of the main contributions of this chapter is the following key definition of *barely robust* learners:

**Definition 5.1** (Barely Robust Learner). *Learner* $\mathbb{A}$ $(\beta, \varepsilon, \delta)$*-barely-robustly-learns a concept* $c : \mathcal{X} \to \mathcal{Y}$ *w.r.t.* $\mathcal{U}^{-1}(\mathcal{U})$ *if* $\exists m_{\mathbb{A}}(\beta, \varepsilon, \delta) \in \mathbb{N}$ *s.t. for any distribution* $D$ *over* $\mathcal{X}$ *satisfying* $\Pr_{x\sim D}\left[\exists z \in \mathcal{U}(x) : c(z) \neq c(x)\right] = 0$*, with prob. at least* $1-\delta$ *over* $S = \{(x_i, c(x_i))\}_{i=1}^m \sim D_c$ *and any internal randomness in the learner,* $\mathbb{A}$ *outputs a predictor* $\hat{h} = \mathbb{A}(S)$ *satisfying:*

$$\Pr_{x\sim D}\left[\forall \tilde{x} \in \mathcal{U}^{-1}(\mathcal{U})(x) : \hat{h}(\tilde{x}) = \hat{h}(x)\right] \geq \beta \;\; and \;\; \Pr_{x\sim D}\left[\hat{h}(x) \neq c(x)\right] \leq \varepsilon.$$

*Notice that we require* $\beta$*-robustness with respect to a "larger" perturbation set* $\mathcal{U}^{-1}(\mathcal{U})$*. Specifically,* $\mathcal{U}^{-1}(\mathcal{U})(x)$ *is the set of all* natural *examples* $\tilde{x}$ *that share an adversarial perturbation* $z$ *with* $x$ *(see Equation 5.1). E.g., if* $\mathcal{U}(x)$ *is an* $\ell_p$*-ball with radius* $\gamma$*, then* $\mathcal{U}^{-1}(\mathcal{U})(x)$ *is an* $\ell_p$*-ball with radius* $2\gamma$*.*

On the other hand, $(\varepsilon, \delta)$-robustly-learning a concept $c$ with respect to $\mathcal{U}$ is concerned with learning a predictor $\hat{h}$ from samples $S$ with small robust risk $\mathrm{R}_{\mathcal{U}}(\hat{h}; D_c) \leq \varepsilon$ with probability at least $1-\delta$ over $S \sim D_c^m$ (see Definition 5.2), where we are interested in robustness with re-spect to $\mathcal{U}$ and *not* $\mathcal{U}^{-1}(\mathcal{U})$. Despite this qualitative difference between *barely robust* learning and *strongly robust* learning, we provably show next that they are in fact *equivalent*.

Our main algorithmic result is $\beta$-RoBoost, an oracle-efficient boosting algorithm that boosts barely robust learners to strongly robust learners:

**Theorem 5.3.** *For any perturbation set $\mathcal{U}$, $\beta$-*RoBoost *$(\varepsilon, \delta)$-robustly-learns any target concept $c : \mathcal{X} \to \mathcal{Y}$ w.r.t. $\mathcal{U}$ using $T \leq \frac{\ln(2/\varepsilon)}{\beta}$ black-box oracle calls to any $(\beta, \frac{\beta\varepsilon}{2}, \frac{\delta}{2T})$-barely-robust learner $\mathbb{A}$ for $c$ w.r.t. $\mathcal{U}^{-1}(\mathcal{U})$, with sample complexity $\frac{4Tm_{\mathbb{A}}}{\varepsilon}$, where $m_{\mathbb{A}}$ is the sample complexity of learner $\mathbb{A}$.*

The result above shows that barely robust learning is *sufficient* for strongly robust learning. An important question remains, however: is our proposed notion of barely robust learning *necessary* for strongly robust learning? In particular, our proposed notion of barely robust learning requires $\beta$-robustness with respect to a "larger" perturbation set $\mathcal{U}^{-1}(\mathcal{U})$, instead of the actual perturbation set $\mathcal{U}$ that we care about. We provably show next that this is *necessary*.

**Theorem 5.9.** *For any $\mathcal{U}$, learner $\mathbb{B}$, and $\varepsilon \in (0, 1/4)$, if $\mathbb{B}$ $(\varepsilon, \delta)$-robustly-learns some unknown target concept $c$ w.r.t. $\mathcal{U}$, then there is a learner $\tilde{\mathbb{B}}$ that $(\frac{1-\varepsilon}{2}, 2\varepsilon, 2\delta)$-barely-robustly-learns $c$ w.r.t. $\mathcal{U}^{-1}(\mathcal{U})$.*

This still does *not* rule out the possibility that boosting robustness is possible even with the weaker requirement of $\beta$-robustness with respect to $\mathcal{U}$. But we show next that, indeed, barely robust learning with respect to $\mathcal{U}$ is *not sufficient* for strongly robust learning with respect to $\mathcal{U}$:

**Theorem 5.11.** *There is a space $\mathcal{X}$, a perturbation set $\mathcal{U}$, and a class of concepts $\mathcal{C}$ s.t. $\mathcal{C}$ is $(\beta = \frac{1}{2}, \varepsilon = 0, \delta)$-barely-robustly-learnable w.r.t $\mathcal{U}$, but $\mathcal{C}$ is* not *$(\varepsilon, \delta)$-robustly-learnable w.r.t. $\mathcal{U}$ for any $\varepsilon < 1/2$.*

Our results offer a new perspective on adversarially robust learning. We show that two seemingly unrelated problems: barely robust learning w.r.t. $\mathcal{U}^{-1}(\mathcal{U})$ and strongly robust learning w.r.t. $\mathcal{U}$, are in fact *equivalent*. The following corollary follows from Theorem 5.3 and Theorem 5.9.

**Corollary I.** *For any class $\mathcal{C}$ and any perturbation set $\mathcal{U}$, $\mathcal{C}$ is* strongly *robustly learnable with respect to $\mathcal{U}$ if and only if $\mathcal{C}$ is* barely *robustly learnable with respect to $\mathcal{U}^{-1}(\mathcal{U})$.*

We would like to note that in our treatment of boosting robustness, having a separate robustness parameter $\beta$ and a natural error parameter $\varepsilon$ allows us to consider regimes where $\beta < \frac{1}{2}$ and $\varepsilon$ is small. This models typical scenarios in practice where learning algorithms are able to learn predictors with reasonably low natural error but the predictors are only barely robust. More generally, this allows us to explore the relationship between $\beta$ and $\varepsilon$ in terms of boosting robustness (see Section 5.5 for a more elaborate discussion).

Landscape of Boosting Robustness. Our results reveal an interesting landscape for boosting robustness when put in context of prior work. When the robustness parameter $\beta > \frac{1}{2}$, it is known from prior work that $\beta$-robustness with respect to $\mathcal{U}$ *suffices* for boosting robustness (see e.g., Chapter 3 and Abernethy et al., 2021), which is witnessed by the $\alpha$-Boost algorithm (Schapire and Freund, 2012). When the robustness parameter $\beta \leq \frac{1}{2}$, our results show that boosting is still *possible*, but $\beta$-robustness with respect to $\mathcal{U}^{-1}(\mathcal{U})$ is *necessary* and we *cannot* boost robustness with $\beta$-robustness with respect to $\mathcal{U}$.

In fact, by combining our algorithm $\beta$-RoBoost with $\alpha$-Boost, we obtain an even *stronger* boosting result that only requires barely robust learners with a natural error parameter that does *not* scale with the targeted robust error. Beyond that, our results imply that we can even boost robustness with respect to $\mathcal{U}^{-1}(\mathcal{U})$. This is summarized in the following corollary which follows from Theorem 5.3, Lemma 5.6, and Theorem 5.9.

**Corollary II** (Landscape of Boosting Robustness). *Let $\mathcal{C}$ be a class of concepts. For fixed $\varepsilon_0, \delta_0 = (\frac{1}{3}, \frac{1}{3})$ and any target $\varepsilon < \varepsilon_0$ and $\delta > \delta_0$:*

1. *If $\mathcal{C}$ is $(\beta, \frac{\beta\varepsilon_0}{2}, \frac{\beta\delta_0}{\ln(2/\varepsilon_0)})$-barely-robustly-learnable w.r.t. $\mathcal{U}^{-1}(\mathcal{U})$, then $\mathcal{C}$ is $(\varepsilon_0, \delta_0)$-robustly-learnable w.r.t. $\mathcal{U}$.*

2. *If $\mathcal{C}$ is $(\varepsilon_0, \delta_0)$-robustly-learnable w.r.t. $\mathcal{U}$, then $\mathcal{C}$ is $(\varepsilon, \delta)$-robustly learnable w.r.t. $\mathcal{U}$.*

3. *If $\mathcal{C}$ is $(\varepsilon, \delta)$-robustly learnable w.r.t. $\mathcal{U}$, then $\mathcal{C}$ is $(\frac{1-\varepsilon}{2}, 2\varepsilon, 2\delta)$-barely-robustly-learnable w.r.t. $\mathcal{U}^{-1}(\mathcal{U})$.*

*In particular, $1 \Rightarrow 2 \Rightarrow 3$ reveals that we can also algorithmically boost robustness w.r.t. $\mathcal{U}^{-1}(\mathcal{U})$.*

Extra Features. In Theorem 5.8, we show that a variant of our boosting algorithm, $\beta$-URoBoost, can boost robustness using *unlabeled* data when having access to a barely robust learner $\mathbb{A}$ that is tolerant to small noise in the labels. In Appendix C.2, we discuss an idea of obtaining robustness at different levels of granularity through our boosting algorithm. Specifically, when $\mathcal{U}(x)$ is a metric-ball around $x$ with radius $\gamma$, we can learn a predictor $\hat{h}$ with different robustness levels: $\gamma, \frac{\gamma}{2}, \frac{\gamma}{4}, \ldots$, in different regions of the distribution.

### 5.1.2 Related Work

To put our work in context, the classic and pioneering works of (Kearns, 1988, Schapire, 1990, Freund, 1990, Freund and Schapire, 1997) explored the question of boosting the *accuracy* of *weak* learning algorithms, from accuracy slightly better than $\frac{1}{2}$ to arbitrarily high accuracy in the realizable PAC learning setting. Later works have explored boosting the accuracy in the agnostic PAC setting (see e.g., Kalai and Kanade, 2009). In this chapter, we are interested in the problem of boosting *robustness* rather than accuracy. In particular, boosting robustness of learners $\mathbb{A}$ that are highly accurate on *natural* examples drawn from the data distribution, but robust only on some $\beta$ fraction of data distribution. We consider this problem in the *robust realizable* setting, that is, when the unknown target concept $c$ has zero robust risk $\mathrm{R}_{\mathcal{U}}(c; D_c) = 0$.

If the robustness parameter $\beta > \frac{1}{2}$, then barely robust learning with respect to $\mathcal{U}$ implies strongly robust learning with respect to $\mathcal{U}$. We have observed and employed this result previously in Chapter 3 (see also related work by Abernethy et al., 2021). Essentially, in this case, we have *weak* learners with respect to the robust risk $\mathrm{R}_{\mathcal{U}}$, and then classical boosting algorithms such as the $\alpha$-Boost algorithm (Schapire and Freund, 2012, Section 6.4.2) can boost the robust risk. For a formal statement and proof of this result, see Lemma 5.6 and Appendix C.1. In contrast, in this chapter, we focus on boosting *barely* robust learners, i.e., mainly when the robustness parameter $\beta < 1/2$ which is a regime that is not addressed by classical boosting algorithms, hence our new $\beta$-RoBoost algorithm which actually works for any $0 < \beta \leq 1$.

In Chapter 3, we studied the problem of adversarially robust learning (as in Definition 5.2). We showed that if a hypothesis class $\mathcal{C}$ is PAC learnable non-robustly (i.e., $\mathcal{C}$ has finite VC dimension), then $\mathcal{C}$ is adversarially robustly learnable (Theorem 3.4). Later on, in Chapter 6, we studied a more constructive version of the same question: reducing strongly robust learning to *non-robust* PAC learning when given access to black-box *non-robust* PAC learners. This is different, however, from the question we study in this chapter. In particular, here we explore the relationship between strongly robust learning and barely robust learning, and we present a boosting algorithm, $\beta$-RoBoost, for learners that already have some non-trivial robustness guarantee $\beta > 0$.

## 5.2 Preliminaries

Let $\mathcal{X}$ denote the instance space and $\mathcal{Y}$ denote the label space. We would like to be robust with respect to a perturbation set $\mathcal{U} : \mathcal{X} \to 2^{\mathcal{X}}$, where $\mathcal{U}(x) \subseteq \mathcal{X}$ is the set of allowed adversarial perturbations that an adversary might replace $x$ with at test time. Denote by $\mathcal{U}^{-1}$ the inverse image of $\mathcal{U}$, where for each $z \in \mathcal{X}, \mathcal{U}^{-1}(z) = \{x \in \mathcal{X} : z \in \mathcal{U}(x)\}$. Observe that for any $x, z \in \mathcal{X}$ it holds that $z \in \mathcal{U}(x) \Leftrightarrow x \in \mathcal{U}^{-1}(z)$. Furthermore, when $\mathcal{U}$ is symmetric, where for any $x, z \in \mathcal{X}, z \in \mathcal{U}(x) \Leftrightarrow x \in \mathcal{U}(z)$, it holds that $\mathcal{U} = \mathcal{U}^{-1}$. For each $x \in \mathcal{X}$, denote by $\mathcal{U}^{-1}(\mathcal{U})(x)$ the set of all *natural* examples $\tilde{x}$ that share some adversarial perturbation $z$ with $x$, i.e.,

$$\mathcal{U}^{-1}(\mathcal{U})(x) = \cup_{z \in \mathcal{U}(x)} \mathcal{U}^{-1}(z) = \{\tilde{x} : \exists z \in \mathcal{U}(x) \cap \mathcal{U}(\tilde{x})\} . \tag{5.1}$$

For example, when $\mathcal{U}(x) = \mathrm{B}_\gamma(x) = \{z \in \mathcal{X} : \rho(x, z) \leq \gamma\}$ where $\gamma > 0$ and $\rho$ is some metric on $\mathcal{X}$ (e.g., $\ell_p$-balls), then $\mathcal{U}^{-1}(\mathcal{U})(x) = \mathrm{B}_{2\gamma}(x)$.

For any classifier $h : \mathcal{X} \to \mathcal{Y}$ and any $\mathcal{U}$, denote by $\mathrm{Rob}_{\mathcal{U}}(h)$ the *robust region* of $h$ with respect to $\mathcal{U}$ defined as:

$$\mathrm{Rob}_{\mathcal{U}}(h) \triangleq \{x \in \mathcal{X} : \forall z \in \mathcal{U}(x), h(z) = h(x)\} . \tag{5.2}$$

**Definition 5.2** (Strongly Robust Learner). *Learner $\mathbb{B}$ $(\varepsilon, \delta)$-robustly-learns a concept $c : \mathcal{X} \to \mathcal{Y}$ with respect to $\mathcal{U}$ if there exists $m(\varepsilon, \delta) \in \mathbb{N}$ s.t. for any distribution $D$ over $\mathcal{X}$ satisfying*

$\mathrm{Pr}_{x\sim D}[\exists z \in \mathcal{U}(x) : c(z) \neq c(x)] = 0$, *with probability at least* $1-\delta$ *over* $S = \{(x_i, c(x_i))\}_{i=1}^m \sim D_c$, $\mathbb{B}$ *outputs a predictor* $\hat{h} = \mathbb{B}(S)$ *satisfying:* $\mathrm{R}_{\mathcal{U}}(\hat{h}; D_c) = \mathrm{Pr}_{x\sim D}\left[\exists z \in \mathcal{U}(x) : \hat{h}(z) \neq c(x)\right] \leq \varepsilon$.

## 5.3 Boosting a Barely Robust Learner to a Strongly Robust Learner

We present our main result in this section: $\beta$-RoBoost is an algorithm for *boosting* the *robustness* of *barely* robust learners. Specifically, in Theorem 5.3, we show that given a *barely* robust learner $\mathbb{A}$ for some unknown target concept $c$ (according to Definition 5.1), it is *possible* to strongly robustly learn $c$ with $\beta$-RoBoost by making black-box oracle calls to $\mathbb{A}$.

**Algorithm 5.1:** $\beta$-RoBoost — Boosting *barely* robust learners.

**Input:** Sampling oracle for distribution $D_c$, black-box $(\beta, \varepsilon, \delta)$-barely-robust learner $\mathbb{A}$.

Set $T = \frac{\ln(2/\varepsilon)}{\beta}$, and $m = \max\left\{ m_{\mathbb{A}}(\beta, \frac{\beta\varepsilon}{2}, \frac{\delta}{2T}), 4\ln\left(\frac{2T}{\delta}\right)\right\}$.

**while** $1 \leq t \leq T$ **do**

- Call `RejectionSampling` on $h_1, \dots, h_{t-1}$ and $m$, and let $\tilde{S}_t$ be the returned dataset.
- **If** $\tilde{S}_t \neq \emptyset$**, then** call learner $\mathbb{A}$ on $\tilde{S}_t$ and let $h_t = \mathbb{A}(\tilde{S}_t)$ be its output. **Otherwise**, break.

**Output:** The cascade predictor defined as
$\mathrm{CAS}(h_1, \dots, h_T)(z) \triangleq G_{h_s}(z)$ where $s = \min\{1 \leq t \leq T : G_{h_t}(z) \neq \perp\}$, and *selective classifiers* $G_{h_t}(z) \triangleq \begin{cases} y, & \text{if } (\exists y \in \mathcal{Y})\ (\forall \tilde{x} \in \mathcal{U}^{-1}(z)) : h(\tilde{x}) = y; \\ \perp, & \text{otherwise.} \end{cases}$.

`RejectionSampling`(*predictors* $h_1, \dots, h_t$*, and sample size* $m$)**:**

- **for** $1 \leq i \leq m$ **do**
  - Draw samples $(x, y) \sim D$ until sampling an $(x_i, y_i)$ s.t.: $\forall_{t' \leq t} \exists_{z \in \mathcal{U}(x_i)} G_{h_{t'}}(z) = \perp$.
    `// sampling from the region of` $D$ `where` *all* `predictors` $h_1, \dots, h_t$ `are` *not* `robust.`
  - If this costs more than $\frac{4}{\varepsilon}$ samples from $D$, abort and return an empty dataset $\tilde{S} = \emptyset$.
    `// If the mass of the non-robust region is small, then we can safely terminate.`
- Output dataset $\tilde{S} = \{(x_1, y_1), \dots, (x_m, y_m)\}$.

**Theorem 5.3.** *For any perturbation set* $\mathcal{U}$*,* $\beta$-RoBoost $(\varepsilon, \delta)$*-robustly-learns any target concept* $c$ *w.r.t.* $\mathcal{U}$ *using* $T = \frac{\ln(2/\varepsilon)}{\beta}$ *black-box oracle calls to any* $(\beta, \frac{\beta\varepsilon}{2}, \frac{\delta}{2T})$*-barely-robust learner* $\mathbb{A}$ *for*

$c$ *w.r.t.* $\mathcal{U}^{-1}(\mathcal{U})$, *with total sample complexity*

$$m(\varepsilon, \delta) \leq \frac{4T \max\left\{m_{\mathbb{A}}(\beta, \frac{\beta\varepsilon}{2}, \frac{\delta}{2T}), 4\ln\left(\frac{2T}{\delta}\right)\right\}}{\varepsilon}.$$

In fact, we present next an even stronger result for boosting $(\beta, \varepsilon_0, \delta_0)$-barely-robust-learners with fixed error $\varepsilon_0 = \frac{\beta}{6}$ and confidence $\delta_0 = \frac{\beta}{6\ln(6)}$. This is established by combining two boosting algorithms: $\beta$-RoBoost from Theorem 5.3 and $\alpha$-Boost from earlier work (see e.g., Schapire and Freund, 2012) which is presented in Appendix C.1 for convenience. The main idea is to perform two layers of boosting. In the first layer, we use $\beta$-RoBoost to get a $(\frac{1}{3}, \frac{1}{3})$-robust-learner w.r.t. $\mathcal{U}$ from a $(\beta, \varepsilon_0, \delta_0)$-barely-robust-learner $\mathbb{A}$ w.r.t. $\mathcal{U}^{-1}(\mathcal{U})$. Then, in the second layer, we use $\alpha$-Boost to boost $\beta$-RoBoost from a $(\frac{1}{3}, \frac{1}{3})$-robust-learner to an $(\varepsilon, \delta)$-robust-learner w.r.t. $\mathcal{U}$.

**Corollary 5.4.** *For any perturbation set* $\mathcal{U}$, $\alpha$-Boost *combined with* $\beta$-RoBoost $(\varepsilon, \delta)$*-robustly-learn any target concept* $c$ *w.r.t.* $\mathcal{U}$ *using* $T = O(\log(m)(\log(1/\delta) + \log\log m)) \cdot \frac{1}{\beta}$ *black-box oracle calls to any* $(\beta, \frac{\beta}{6}, \frac{\beta}{6\ln(6)})$*-barely-robust-learner* $\mathbb{A}$ *for* $c$ *w.r.t.* $\mathcal{U}^{-1}(\mathcal{U})$, *with total sample complexity*

$$m(\varepsilon, \delta) = O\left(\frac{m_0}{\beta\varepsilon}\log^2\left(\frac{m_0}{\beta\varepsilon}\right) + \frac{\log(1/\delta)}{\varepsilon}\right), \text{ where } m_0 = \max\left\{m_{\mathbb{A}}\left(\beta, \frac{\beta}{6}, \frac{\beta}{6\ln(6)}\right), 4\ln\left(\frac{6\ln(6)}{\beta}\right)\right\}.$$

We begin with describing the intuition behind $\beta$-RoBoost, and then we will prove Theorem 5.3 and Corollary 5.4.

High-level Strategy. Let $D_c$ be the unknown distribution we want to robustly learn. Since $\mathbb{A}$ is a *barely* robust learner for $c$, calling learner $\mathbb{A}$ on an i.i.d. sample $S$ from $D_c$ will return a predictor $h_1$, where $h_1$ is robust only on a region $R_1 \subseteq \mathcal{X}$ of small mass $\beta > 0$ under distribution $D$, $\Pr_{x\sim D}[x \in R_1] \geq \beta$. We can trust the predictions of $h_1$ in the region $R_1$, but not in the complement region $\bar{R}_1$ where it is not robust. For this reason, we will use a *selective* classifier $G_{h_1}$ (see Equation 5.3) which makes predictions on *all* adversarial perturbations in region $R_1$, but *abstains* on adversarial perturbations not from $R_1$. In each round $t > 1$, the strategy is to focus on the region of distribution $D$ where *all* predictors $h_1, \ldots, h_{t-1}$ returned by $\mathbb{A}$ so far are *not* robust. By *rejection sampling*, $\beta$-RoBoost gives barely robust learner $\mathbb{A}$ a sample $\tilde{S}_t$ from this non-robust region, and then $\mathbb{A}$ returns a predictor $h_t$ with robustness at least $\beta$ in this region. Thus, in each round, we shrink by a factor of $\beta$ the mass of region $D$ where the predictors learned so far are not robust. After $T$ rounds, $\beta$-RoBoost outputs a *cascade* of *selective* classifiers $G_{h_1}, \dots, G_{h_T}$ where roughly each selective classifier $G_{h_t}$ is responsible for making predictions in the region where $h_t$ is robust.

As mentioned above, one of the main components in our boosting algorithm is *selective classifiers* that essentially abstain from predicting in the region where they are *not* robust. Formally, for any classifier $h : \mathcal{X} \to \mathcal{Y}$ and any $\mathcal{U}$, denote by $G_h : \mathcal{X} \to \mathcal{Y} \cup \{\perp\}$ a *selective* classifier defined as:

$$G_h(z) \triangleq \begin{cases} y, & \text{if } (\exists y \in \mathcal{Y})\,(\forall \tilde{x} \in \mathcal{U}^{-1}(z)) : h(\tilde{x}) = y; \\ \perp, & \text{otherwise.} \end{cases} \tag{5.3}$$

Before proceeding with the proof of Theorem 5.3, we prove the following key Lemma about *selective classifier* $G_h$ which states that $G_h$ will *not* abstain in the region where $h$ is robust, and whenever $G_h$ predicts a label for a perturbation $z$ then we are guaranteed that this is the same label that $h$ predicts on the corresponding natural example $x$ where $z \in \mathcal{U}(x)$.

**Lemma 5.5.** *For any distribution $D$ over $\mathcal{X}$, any $\mathcal{U}$, and any $\beta > 0$, given a classifier $h : \mathcal{X} \to \mathcal{Y}$ satisfying $\Pr_{x\sim D}\left[x \in \mathrm{Rob}_{\mathcal{U}^{-1}(\mathcal{U})}(h)\right] \geq \beta$, then the* selective *classifier $G_h : \mathcal{X} \to \mathcal{Y} \cup \{\perp\}$ (see Equation 5.3) satisfies:*

$$\Pr_{x\sim D}\left[\forall z \in \mathcal{U}(x) : G_h(z) = h(x)\right] \geq \beta \textit{ and } \Pr_{x\sim D}\left[\forall z \in \mathcal{U}(x) : G_h(z) = h(x) \vee G_h(z) = \perp\right] = 1.$$

*Proof.* Observe that, by the definition of the robust region of $h$ w.r.t. $\mathcal{U}^{-1}(\mathcal{U})$ (see Equation 5.2), for any $x \in \mathrm{Rob}_{\mathcal{U}^{-1}(\mathcal{U})}(h)$ the following holds: $(\forall z \in \mathcal{U}(x))\,(\forall \tilde{x} \in \mathcal{U}^{-1}(z)) : h(\tilde{x}) = h(x)$. By the definition of the *selective* classifier $G_h$ (see Equation 5.3), this implies that: $\forall z \in \mathcal{U}(x), G_h(z) = h(x)$.

Since $\Pr_{x\sim D}\left[x \in \mathrm{Rob}_{\mathcal{U}^{-1}(\mathcal{U})}(h)\right] \geq \beta$, the above implies $\Pr_{x\sim D}\left[\forall z \in \mathcal{U}(x) : G_h(z) = h(x)\right] \geq \beta$. Furthermore, for any $x \in \mathcal{X}$ and any $z \in \mathcal{U}(x)$, by definition of $\mathcal{U}^{-1}$, $x \in \mathcal{U}^{-1}(z)$. Thus, by definition of $G_h$ (see Equation 5.3), if $G_h(z) = y$ for some $y \in \mathcal{Y}$, then it holds that

$h(x) = y$. Combined with the above, this implies that

$$\Pr_{x \sim D}\left[\forall z \in \mathcal{U}(x) : G_h(z) = h(x) \vee G_h(z) = \perp\right] = 1.$$

$\square$

We are now ready to proceed with the proof of Theorem 5.3.

*Proof of Theorem 5.3.* Let $\mathcal{U}$ be an arbitrary adversary, and $\mathbb{A}$ a $(\beta, \varepsilon, \delta)$-barely-robust learner for some *unknown* target concept $c : \mathcal{X} \to \mathcal{Y}$ with respect to $\mathcal{U}^{-1}(\mathcal{U})$. We will show that $\beta$-RoBoost $(\varepsilon, \delta)$-robustly-learns $c$ with respect to $\mathcal{U}$. Let $D$ be some unknown distribution over $\mathcal{X}$ such that $\Pr_{x \sim D}\left[\exists z \in \mathcal{U}(x) : c(z) \neq c(x)\right] = 0$. Let $\varepsilon > 0$ be our target robust error, and $\varepsilon'$ be the error guarantee of learner $\mathbb{A}$ (which we will set later to be $\frac{\beta\varepsilon}{2}$).

Without loss of generality, suppose that $\beta$-RoBoost ran for $T = \frac{\ln(2/\varepsilon)}{\beta}$ rounds (Step 5 takes care of the scenario where progress is made faster). Let $h_1 = \mathbb{A}(\tilde{S}_1), \ldots, h_T = \mathbb{A}(\tilde{S}_T)$ be the predictors returned by learner $\mathbb{A}$ on rounds $1 \leq t \leq T$. For any $1 \leq t \leq T$ and any $x \in \mathcal{X}$, denote by $R_t$ the event that $x \in \mathrm{Rob}_{\mathcal{U}^{-1}(\mathcal{U})}(h_t)$, and by $\bar{R}_t$ the event that $x \notin \mathrm{Rob}_{\mathcal{U}^{-1}(\mathcal{U})}(h_t)$ (as defined in Equation 5.2). Observe that by properties of $\mathbb{A}$ (see Definition 5.1), we are guaranteed that

$$\forall 1 \leq t \leq T : \Pr_{x \sim D_t}\left[R_t\right] \geq \beta \text{ and } \Pr_{x \sim D_t}\left[h_t(x) \neq c(x)\right] \leq \varepsilon', \tag{5.4}$$

where $D_t$ is the distribution from which $\tilde{S}_t$ is drawn. In words, distribution $D_t$ is a conditional distribution focusing on the region of distribution $D$ where all predictors $h_1, \ldots, h_{t-1}$ are *non-robust*. Specifically, for any $x \sim D$, in case $x \in R_1 \cup \cdots \cup R_{t-1}$ (i.e., there is a predictor among $h_1, \ldots, h_{t-1}$ that is robust on $x$), then Lemma 5.5 guarantees that one of the selective classifiers $G_{h_1}, \ldots, G_{h_{t-1}}$ will not abstain on any $z \in \mathcal{U}(x)$: $\exists t' \leq t - 1, \forall z \in \mathcal{U}(x), G_{h_{t'}}(z) \neq \perp$. In case $x \in \bar{R}_1 \cap \cdots \cap \bar{R}_{t-1}$, then, by definition of $\bar{R}_1, \ldots, \bar{R}_{t-1}$, each of the selective classifiers $G_{h_1}, \ldots, G_{h_{t-1}}$ can be forced to abstain: $\forall t' \leq t - 1, \exists z \in \mathcal{U}(x), G_{h_{t'}}(z) = \perp$. Thus, in Step 7 of $\beta$-RoBoost, *rejection sampling* guarantees that $\tilde{S}_t$ is a sample drawn from distribution $D$ conditioned on the region $\bar{R}_{1:t-1} \triangleq \bar{R}_1 \cap \cdots \cap \bar{R}_{t-1}$.

Formally, distribution $D_t$ is defined such that for any measurable event $E$:

$$\Pr_{x \sim D_t}[E] \triangleq \Pr_{x \sim D}[E | \bar{R}_1 \cap \cdots \cap \bar{R}_{t-1}]. \tag{5.5}$$

Low Error on Natural Examples. Lemma 5.5 guarantees that whenever any of the *selective* classifiers $G_{h_1}, \ldots, G_{h_T}$ (see Equation 5.3) chooses to classify an instance $z$ at test-time their prediction will be correct with high probability. We consider two cases. First, in case of event $R_t$, $x \in \mathrm{Rob}_{\mathcal{U}^{-1}(\mathcal{U})}(h_t)$, and therefore, by Lemma 5.5, $\forall z \in \mathcal{U}(x) : G_{h_t}(z) = h_t(x)$. Thus, Equation 5.5 implies that $\forall 1 \leq t \leq T$:

$$\begin{aligned} &\Pr_{x \sim D}[R_t \wedge (\exists z \in \mathcal{U}(x) : (G_{h_t}(z) \neq \perp) \wedge (G_{h_t}(z) \neq c(x))) \,|\bar{R}_{1:t-1}] \\ &= \Pr_{x \sim D_t}[R_t \wedge (\exists z \in \mathcal{U}(x) : (G_{h_t}(z) \neq \perp) \wedge (G_{h_t}(z) \neq c(x)))] \\ &= \Pr_{x \sim D_t}[R_t \wedge (h_t(x) \neq c(x))]. \end{aligned} \tag{5.6}$$

Second, in case of the complement event $\bar{R}_t$, $x \notin \mathrm{Rob}_{\mathcal{U}^{-1}(\mathcal{U})}(h_t)$. Therefore, by Lemma 5.5, $\forall z \in \mathcal{U}(x)$, we have $G_{h_t}(z) = \perp$ or $G_{h_t}(z) = h_t(x)$. Thus,

$$\begin{aligned} &\Pr_{x \sim D}[\bar{R}_t \wedge (\exists z \in \mathcal{U}(x) : (G_{h_t}(z) \neq \perp) \wedge (G_{h_t}(z) \neq c(x))) \,|\bar{R}_{1:t-1}] \\ &= \Pr_{x \sim D_t}[\bar{R}_t \wedge (\exists z \in \mathcal{U}(x) : (G_{h_t}(z) \neq \perp) \wedge (G_{h_t}(z) \neq c(x)))] \leq \Pr_{x \sim D_t}[\bar{R}_t \wedge (h_t(x) \neq c(x))]. \end{aligned} \tag{5.7}$$

By law of total probability Equation 5.6, Equation 5.7, and Equation 5.4,

$$\begin{aligned} &\Pr_{x \sim D}[(R_t \vee \bar{R}_t) \wedge (\exists z \in \mathcal{U}(x) : (G_{h_t}(z) \neq \perp) \wedge (G_{h_t}(z) \neq c(x))) \,|\bar{R}_{1:t-1}] \\ &\quad \leq \Pr_{x \sim D_t}[R_t \wedge h_t(x) \neq c(x)] + \Pr_{x \sim D_t}[\bar{R}_t \wedge h_t(x) \neq c(x)] = \Pr_{x \sim D_t}[h_t(x) \neq c(x)] \leq \varepsilon'. \end{aligned} \tag{5.8}$$

Boosted Robustness. We claim that for each $1 \leq t \leq T : \Pr_{x \sim D}[\bar{R}_{1:t}] \leq (1-\beta)^t$. We proceed by induction on the number of rounds $1 \leq t \leq T$. In the base case, when $t = 1$, $D_1 = D$ and by Equation 5.4, we have $\Pr_{x \sim D}[R_1] \geq \beta$ and therefore $\Pr_{x \sim D}[\bar{R}_1] \leq 1 - \beta$.

When $t > 1$, again by Equation 5.4, we have that $\Pr_{x \sim D_t}[R_t] \geq \beta$ and therefore, by Equa-tion 5.5, $\Pr_{x\sim D}[\bar{R}_t|\bar{R}_{1:t-1}] = \Pr_{x\sim D_t}[\bar{R}_t] \le 1-\beta$. Finally, by the inductive hypothesis and Bayes' rule, we get

$$\Pr_{x\sim D}[\bar{R}_{1:t}] = \Pr_{x\sim D}[\bar{R}_t|\bar{R}_{1:t-1}] \Pr_{x\sim D}[\bar{R}_{1:t-1}] \le (1-\beta)(1-\beta)^{t-1} = (1-\beta)^t. \tag{5.9}$$

Analysis of Robust Risk. For each $1 \le t \le T$, let

$$A_t = \left\{x \in \bar{R}_t : \exists z \in \mathcal{U}(x) \text{ s.t. } G_{h_t(z)} \ne \perp \wedge \forall_{t'<t} G_{h_{t'}}(z) = \perp\right\}$$

denote the *non-robust* region of classifier $h_t$ where the *selective* classifier $G_{h_t}$ does not abstain but all selective classifiers $G_{h_1}, \ldots, G_{h_{t-1}}$ abstain. By the law of total probability, we can analyze the robust risk of the cascade predictor $\mathrm{CAS}(h_1, \ldots, h_T)$ by partitioning the space into the *robust* and *non-robust* regions of $h_1, \ldots, h_T$. Specifically, by the structure of the cascade predictor $\mathrm{CAS}(h_{1:T})$, each $x \sim D$ such that $\exists z \in \mathcal{U}(x)$ where $\mathrm{CAS}(h_{1:T})(z) \ne c(x)$ satisfies the following condition:

$$\exists_{z\in\mathcal{U}(x)} : \mathrm{CAS}(h_{1:T})(z) \ne c(x) \Rightarrow (\exists_{1\le t\le T}) \left(\exists_{z\in\mathcal{U}(x)}\right) : \forall_{t'<t} G_{h'_t}(z) = \perp \wedge G_{h_t}(z) \ne \perp \wedge G_{h_t}(z) \ne c(x).$$

Thus, each $x \sim D$ such that $\exists z \in \mathcal{U}(x)$ where $\mathrm{CAS}(h_{1:T})(z) \ne c(x)$ can be mapped to one (or more) of the following regions:

$$\underbrace{R_1 \vee A_1}_{G_{h_1}\text{ does not abstain}} \;\Big|\; \underbrace{\bar{R}_1 \wedge (R_2 \vee A_2)}_{G_{h_2}\text{ does not abstain}} \;\Big|\; \bar{R}_{1:2} \wedge (R_3 \vee A_3) \;\Big|\; \ldots \;\Big|\; \underbrace{\bar{R}_{1:T-1} \wedge (R_T \vee A_T)}_{G_{h_T}\text{ does not abstain}} \;\Big|\; \bar{R}_{1:T}.$$

We will now analyze the robust risk based on the above decomposition:

$$
\begin{aligned}
&\Pr_{x\sim D}\left[\exists z \in \mathcal{U}(x) : \mathrm{CAS}(h_{1:T})(z) \neq c(x)\right] \\
&\leq \sum_{t=1}^{T+1} \Pr_{x\sim D}\left[\bar{R}_{1:t-1} \wedge (R_t \vee A_t) \wedge (\exists z \in \mathcal{U}(x) : \mathrm{CAS}(h_{1:T})(z) \neq c(x))\right] \\
&\leq \sum_{t=1}^{T} \Pr_{x\sim D}\left[\bar{R}_{1:t-1} \wedge (R_t \vee A_t) \wedge (\exists z \in \mathcal{U}(x) : \mathrm{CAS}(h_{1:T})(z) \neq c(x))\right] + \Pr_{x\sim D}\left[\bar{R}_{1:T}\right] \\
&\overset{(i)}{\leq} \sum_{t=1}^{T} \Pr_{x\sim D}\left[\bar{R}_{1:t-1} \wedge (R_t \vee A_t) \wedge (\exists z \in \mathcal{U}(x) : G_{h_t}(z) \neq \perp \wedge G_{h_t}(z) \neq c(x))\right] + \Pr_{x\sim D}\left[\bar{R}_{1:T}\right] \\
&= \sum_{t=1}^{T} \Pr_{x\sim D}\left[\bar{R}_{1:t-1}\right] \Pr_{x\sim D}\left[(R_t \vee A_t) \wedge (\exists z \in \mathcal{U}(x) : G_{h_t}(z) \neq \perp \wedge G_{h_t}(z) \neq c(x)) \,|\bar{R}_{1:t-1}\right] + \Pr_{x\sim D}\left[\bar{R}_{1:T}\right] \\
&\overset{(ii)}{\leq} \sum_{t=1}^{T} (1-\beta)^{t-1} \varepsilon' + (1-\beta)^{T} = \varepsilon' \sum_{t=1}^{T} (1-\beta)^{t-1} = \varepsilon' \frac{1-(1-\beta)^T}{1-(1-\beta)} + (1-\beta)^T \leq \frac{\varepsilon'}{\beta} + (1-\beta)^T,
\end{aligned}
\tag{5.10}
$$

where inequality $(i)$ follows from the definitions of $\bar{R}_{1:t-1}$, $R_t$, and $A_t$, and inequality $(ii)$ follows from Equation 5.9 and Equation 5.8. It remains to choose $T$ and $\varepsilon'$ such that the robust risk is at most $\varepsilon$.

Sample and Oracle Complexity. It suffices to choose $T = \frac{\ln(2/\varepsilon)}{\beta}$ and $\varepsilon' = \frac{\varepsilon\beta}{2}$. We next analyze the sample complexity. Fix an arbitrary round $1 \leq t \leq T$. In order to obtain a *good* predictor $h_t$ from learner $\mathbb{A}$ satisfying Equation 5.4, we need to draw $m_{\mathbb{A}}(\beta, \beta\varepsilon/2, \delta/2T)$ samples from $D_t$. We do this by drawing samples from the original distribution $D$ and doing rejection sampling. Specifically, let $m = \max\left\{m_{\mathbb{A}}(\beta, \frac{\varepsilon\beta}{2}, \frac{\delta}{2T}), 4\ln\left(\frac{2T}{\delta}\right)\right\}$ (as defined in Step 1). Then, for each $1 \leq i \leq m$, let $X_{t,i}$ be a random variable counting the number of samples $(x, y)$ drawn from $D$ until a sample $(x, y) \in \bar{R}_{1:t-1}$ is obtained. Notice that $X_{t,i}$ is a geometric random variable with expectation $1/p_t$ where $p_t = \Pr_{x\sim D}\left[\bar{R}_{1:t}\right]$. Then, the expected number of samples drawn from $D$ in round $t$ is $\mathbb{E}\left[\sum_{i=1}^{m} X_{t,i}\right] = \frac{m}{p_t}$. By applying a standard concentra-tion inequality for the sums of i.i.d. geometric random variables (Brown, 2011), we get

$$\Pr\left[\sum_{i=1}^{m} X_{t,i} > 2\frac{m}{p_t}\right] \leq e^{-\frac{2m(1-1/2)^2}{2}} = e^{-\frac{m}{4}} \leq \frac{\delta}{2T},$$

where the last inequality follows from our choice of $m$. By a standard union bound, we get that with probability at least $1-\frac{\delta}{2}$, the total number of samples $\sum_{t=1}^{T}\sum_{i=1}^{m} X_{t,i} \leq \sum_{t=1}^{T} 2\frac{m}{p_t} = 2m\sum_{t=1}^{T}\frac{1}{p_t} \leq \frac{4mT}{\varepsilon}$. □

Before proceeding with the proof of Corollary 5.4, we state the following guarantee that $\alpha$-Boost provides for boosting weakly robust learners. Its proof is deferred to Appendix C.1.

**Lemma 5.6.** *For any perturbation set $\mathcal{U}$, $\alpha$-*Boost *$(\varepsilon, \delta)$-robustly-learns any target concept $c$ w.r.t. $\mathcal{U}$ using $T$ black-box oracle calls to any $(\frac{1}{3}, \frac{1}{3})$-robust-learner $\mathbb{B}$ for $c$ w.r.t. $\mathcal{U}$, with total sample complexity*

$$m(\varepsilon,\delta) = O\left(\frac{m_{\mathbb{B}}(1/3,1/3)}{\varepsilon}\log^2\left(\frac{m_{\mathbb{B}}(1/3,1/3)}{\varepsilon}\right) + \frac{\log(1/\delta)}{\varepsilon}\right),$$

*and oracle calls $T = O(\log(m)\,(\log(1/\delta) + \log\log m))$.*

We are now ready to proceed with the proof of Corollary 5.4.

*Proof of Corollary 5.4.* The main idea is to perform two layers of boosting. In the first layer, we use $\beta$-RoBoost to get a *weak* robust learner for $c$ w.r.t. $\mathcal{U}$ from a *barely* robust learner $\mathbb{A}$ for $c$ w.r.t. $\mathcal{U}^{-1}(\mathcal{U})$. Then, in the second layer, we use $\alpha$-Boost to boost $\beta$-RoBoost from a *weak* robust learner to a *strong* robust learner for $c$ w.r.t. $\mathcal{U}$.

Let $\mathbb{A}$ be a $(\beta, \frac{\beta}{6}, \frac{\beta}{6\ln(6)})$-barely-robust-learner $\mathbb{A}$ for $c$ w.r.t. $\mathcal{U}^{-1}(\mathcal{U})$. Let $(\varepsilon_0, \delta_0, T_0) = (\frac{1}{3}, \frac{1}{3}, \frac{\ln(6)}{\beta})$, and observe that $\mathbb{A}$ is a $(\beta, \frac{\beta\varepsilon_0}{2}, \frac{\delta_0}{2T_0})$-barely-robust-learner for $c$ w.r.t. $\mathcal{U}^{-1}(\mathcal{U})$. By Theorem 5.3, $\beta$-RoBoost $(\varepsilon_0, \delta_0)$-robustly-learns $c$ w.r.t. $\mathcal{U}$ using $T_0$ black-box oracle calls to $\mathbb{A}$, with sample complexity $m_0 = O\left(\frac{\max\left\{m_{\mathbb{A}}, 4\ln\left(\frac{6\ln(6)}{\beta}\right)\right\}}{\beta}\right)$. Finally, by Lemma 5.6, $\alpha$-Boost $(\varepsilon, \delta)$-robustly-learns $c$ w.r.t. $\mathcal{U}$ using $m(\varepsilon, \delta) = O\left(\frac{m_0}{\varepsilon}\log\left(\frac{m_0}{\varepsilon}\right) + \frac{\log(1/\delta)}{\varepsilon}\right)$ samples, and $O(\log m)$ black-box oracle calls to $\beta$-RoBoost. □

### 5.3.1 Boosting Robustness with Unlabeled Data

Prior work has shown that unlabeled data can improve adversarially robust generalization in practice (Alayrac, Uesato, Huang, Fawzi, Stanforth, and Kohli, 2019, Carmon, Raghunathan, Schmidt, Duchi, and Liang, 2019a), and there is also theoretical work quantifying the benefit of unlabeled data for robust generalization (Ashtiani, Pathak, and Urner, 2020). In this section, we highlight yet another benefit of unlabeled data for adversarially robust learning. Specifically, we show that it is *possible* to boost robustness by relying only on *unlabeled* data.

We will start with some intuition first. For an unknown distribution $D_c$, imagine having access to a *non-robust* classifier $h$ that makes no mistakes on natural examples, i.e., $\Pr_{x\sim D}[h(x) \neq c(x)] = 0$ but $\Pr_{x\sim D}[\exists z \in \mathcal{U}(x) : h(z) \neq h(x)] = 1$. Now, in order to learn a robust classifier, we can use $\beta$-RoBoost where in each round of boosting we sample unlabeled data from $D$ (label it with $h$) and call a barely robust learner $\mathbb{A}$ on this *pseudo-labeled* dataset.

This highlights that perhaps robustness can be boosted using only unlabeled data if we have access to a good *pseudo-labeler* $h$ that makes few mistakes on natural examples from $D$. But in case that $\Pr_{x\sim D}[h(x) \neq c(x)] = \varepsilon$ for some small $\varepsilon > 0$, it no longer suffices to use a barely robust learner $\mathbb{A}$ for $c$, but rather we need a more powerful learner that is tolerant to the noise in the labels introduced by $h$. Formally, we introduce the following noise-tolerant barely robust learner:

**Definition 5.7** (Noise-Tolerant Barely Robust Learner). *Learner $\mathbb{A}$ $(\eta, \beta, \varepsilon, \delta)$-barely-robustly-learns a concept $c : \mathcal{X} \to \mathcal{Y}$ w.r.t. $\mathcal{U}^{-1}(\mathcal{U})$ if there exists $m(\eta, \beta, \varepsilon, \delta) \in \mathbb{N}$ such that for any distribution $D$ over $\mathcal{X}$ satisfying $\Pr_{x\sim D}[\exists z \in \mathcal{U}(x) : c(z) \neq c(x)] = 0$ and any $h : \mathcal{X} \to \mathcal{Y}$ where $\Pr_{x\sim D}[h(x) \neq c(x)] \leq \eta$, w.p. at least $1-\delta$ over $S \sim D_h^m$, $\mathbb{A}$ outputs a predictor $\hat{h} = \mathbb{A}(S)$ satisfying:*

$$\Pr_{x\sim D}\left[x \in \mathrm{Rob}_{\mathcal{U}^{-1}(\mathcal{U})}(\hat{h})\right] \geq \beta \text{ and } \Pr_{x\sim D}\left[\hat{h}(x) \neq c(x)\right] \leq \Pr_{x\sim D}[h(x) \neq c(x)] + \varepsilon \leq \eta + \varepsilon.$$

In Theorem 5.8, we show that given a *noise-tolerant* barely robust learner $\mathbb{A}$ for some unknown target concept $c$ (according to Definition 5.7), it is *possible* to strongly robustly learn $c$ with $\beta$-URoBoost by making black-box oracle calls to $\mathbb{A}$.

**Algorithm 5.2:** $\beta$-URoBoost — Boosting Robustness with Unlabeled Data

**Input:** Sampling oracle for distribution $D_c$, black-box noise-tolerant barely-robust learner $\mathbb{A}$.

1 Draw $m = m_{\mathbb{A}}(\eta, \beta, \frac{\beta\varepsilon}{4}, \frac{\delta}{2})$ labeled samples $S = \{(x_1, y_1), \ldots, (x_m, y_m)\} \sim D_c$.

2 Call learner $\mathbb{A}$ on $S$ and let predictor $\hat{h} = \mathbb{A}(S)$ be its output.

3 Call $\beta$-RoBoost with access to labeled samples from $D_{\hat{h}}$ (i.e., $(x, \hat{h}(x)) \sim D_{\hat{h}}$), and black-box $(\eta, \beta, \frac{\varepsilon\beta}{4}, \frac{\delta}{2T})$-noise-tolerant-barely-robust-learner $\mathbb{A}$.

**Output:** The cascade predictor $\mathrm{CAS}(h_1, \ldots, h_T)$.

**Theorem 5.8.** *For any perturbation set $\mathcal{U}$, $\beta$-*URoBoost *$(\varepsilon, \delta)$-robustly-learns any target concept $c$ w.r.t. $\mathcal{U}$ using $T + 1 \leq \frac{\ln(2/\varepsilon)}{\beta} + 1$ black-box oracle calls to any $(\eta, \beta, \frac{\beta\varepsilon}{4}, \frac{\delta}{2T})$-barely-robust learner $\mathbb{A}$ for $c$ w.r.t. $\mathcal{U}^{-1}(\mathcal{U})$, with* labeled *sample complexity of $m_{\mathbb{A}}(\eta, \beta, \frac{\beta\varepsilon}{4}, \frac{\delta}{2})$ and* unlabeled *sample complexity of at most*

$$\frac{4T \max\left\{ m_{\mathbb{A}}(\eta, \beta, \frac{\beta\varepsilon}{4}, \frac{\delta}{4T}), 4\ln\left(\frac{4T}{\delta}\right) \right\}}{\varepsilon}.$$

*Proof.* Let $\mathcal{U}$ be an arbitrary perturbation set, and $\mathbb{A}$ a $(\eta, \beta, \varepsilon, \delta)$-barely-robust learner for some *unknown* target concept $c : \mathcal{X} \to \mathcal{Y}$ with respect to $\mathcal{U}^{-1}(\mathcal{U})$. We will show that $\beta$-URoBoost $(\varepsilon, \delta)$-robustly-learns $c$ with respect to $\mathcal{U}$. Let $D$ be some unknown distribution over $\mathcal{X}$ such that $\Pr_{x \sim D}\left[\exists z \in \mathcal{U}(x) : c(z) \neq c(x)\right] = 0$.

By Step 1 and Step 2 and the guarantee of learner $\mathbb{A}$ (see Definition 5.7), with probability at least $1 - \frac{\delta}{2}$ over $S^m \sim D_c$, it holds that

$$\Pr_{x \sim D}\left[\hat{h}(x) \neq c(x)\right] \leq \frac{\beta\varepsilon}{4}.$$

That is, with high probability, $\hat{h}$ is a predictor with low error on *natural* examples.

Pseudo labeling. In Step 3, $\beta$-URoBoost essentially runs $\beta$-RoBoost using unlabeled samples from $D$ that are labeled with the predictor $\hat{h}$. Thus, we can view this as robustly learning the concept $\hat{h}$ which is only an approximation of the true concept $c$ that we care about. Since the noise tolerance $\eta \geq \frac{\beta\varepsilon}{4}$, it follows by the guarantees of learner $\mathbb{A}$ (see Definition 5.7)

and Equation 5.10, that

$$\Pr_{x \sim D}\left[\exists z \in \mathcal{U}(x) : \mathrm{CAS}(h_{1:T})(z) \neq c(x)\right] \leq \frac{\Pr_{x \sim D}\left[\hat{h}(x) \neq c(x)\right]}{\beta} + \frac{\beta\varepsilon}{4\beta} + (1-\beta)^T \leq \frac{\varepsilon}{4} + \frac{\varepsilon}{4} + \frac{\varepsilon}{2} \leq \varepsilon.$$

□

## 5.4 The Necessity of Barely Robust Learning

We have established in Section 5.3 (Theorem 5.3) that our proposed notion of barely robust learning in Definition 5.1 *suffices* for strongly robust learning. But is our notion actually *necessary* for strongly robust learning? In particular, notice that in our proposed notion of barely robust learning in Definition 5.1, we require $\beta$-robustness with respect to a "larger" perturbation set $\mathcal{U}^{-1}(\mathcal{U})$, instead of the actual perturbation set $\mathcal{U}$ that we care about. Is this necessary? or can we perhaps boost robustness even with the weaker guarantee of $\beta$-robustness with respect to $\mathcal{U}$?

In this section, we answer this question in the negative. First, we provably show in Theorem 5.9 that strongly robust learning with respect to $\mathcal{U}$ implies barely robust learning with respect to $\mathcal{U}^{-1}(\mathcal{U})$. This indicates that our proposed notion of barely robust learning with respect to $\mathcal{U}^{-1}(\mathcal{U})$ (Definition 5.1) is *necessary* for strongly robust learning with respect to $\mathcal{U}$. Second, we provably show in Theorem 5.11 that barely robust learning with respect to $\mathcal{U}$ does *not* imply strongly robust learning with respect to $\mathcal{U}$ when the robustness parameter $\beta \leq \frac{1}{2}$ which is the main regime of interest that we study in this chapter.

**Theorem 5.9.** *For any $\mathcal{U}$, learner $\mathbb{B}$, and $\varepsilon \in (0, 1/4)$, if $\mathbb{B}$ $(\varepsilon, \delta)$-robustly learns some unknown target concept $c$ w.r.t. $\mathcal{U}$, then there is a learner $\tilde{\mathbb{B}}$ that $(\frac{1-\varepsilon}{2}, 2\varepsilon, 2\delta)$-barely-robustly-learns $c$ w.r.t. $\mathcal{U}^{-1}(\mathcal{U})$.*

We briefly describe the high-level strategy here. The main idea is to convert a strongly robust learner $\mathbb{B}$ with respect to $\mathcal{U}$ to a barely robust learner $\tilde{\mathbb{B}}$ with respect to $\mathcal{U}^{-1}(\mathcal{U})$. We do this with a simple *expansion* trick that modifies a predictor $h$ robust with respect to $\mathcal{U}$ to a predictor $g$ robust with respect to $\mathcal{U}^{-1}(\mathcal{U})$. For each label $y \in \mathcal{Y}$, we do this expansion conditional on the label to get a predictor $g_y$ that is robust w.r.t. $\mathcal{U}^{-1}(\mathcal{U})$ but only in the region of $\mathcal{X}$ where $h$ predicts the label $y$ robustly w.r.t. $\mathcal{U}$. This is described in the following key Lemma. We then use fresh samples to select predictor a $g_y$ whose label $y$ occurs more often.

**Lemma 5.10.** *For any distribution $D$ over $\mathcal{X}$ and any concept $c : \mathcal{X} \to \{\pm 1\}$, given a predictor $\hat{h} : \mathcal{X} \to \{\pm 1\}$ such that $\mathrm{R}_{\mathcal{U}}(\hat{h}; D_c) \leq \varepsilon$ for some $\varepsilon \in (0, 1/4)$, then for each $y \in \{\pm 1\}$, the predictor $g_y$ defined for each $x \in \mathcal{X}$ as*

$$g_y(x) \triangleq y \text{ iff } x \in \bigcup_{\tilde{x} \in \mathrm{Rob}_{\mathcal{U}}(\hat{h}) \wedge \hat{h}(\tilde{x}) = y} \mathcal{U}^{-1}(\mathcal{U}(\tilde{x})) \textit{ satisfies}$$

$$\Pr_{x \sim D}\left[g_y(x) \neq c(x)\right] \leq 2\varepsilon \textit{ and } \Pr_{x \sim D}\left[x \in \mathrm{Rob}_{\mathcal{U}^{-1}(\mathcal{U})}(g_y)\right] \geq (1-\varepsilon) \Pr_{x \sim D}\left[\hat{h}(x) = y \middle| x \in \mathrm{Rob}_{\mathcal{U}}(\hat{h})\right].$$

*Proof.* Without loss of generality, let $y = +1$. Let $x \in \mathrm{supp}(D)$ such that $x \in \mathrm{Rob}_{\mathcal{U}}(\hat{h})$. In case $\hat{h}(x) = +1$, then by definition of $g_+$, since $x \in \mathcal{U}^{-1}(\mathcal{U}(x))$, it holds that $g_+(x) = +1$. In case $\hat{h}(x) = -1$, then $\neg \exists \tilde{x} \in \mathrm{Rob}_{\mathcal{U}}(\hat{h})$ such that $\hat{h}(\tilde{x}) = +1$ and $\mathcal{U}(\tilde{x}) \cap \mathcal{U}(x) \neq \emptyset$, which implies that $x \notin \bigcup_{\tilde{x} \in \mathrm{Rob}_{\mathcal{U}}(\hat{h}) \wedge \hat{h}(\tilde{x}) = +1} \mathcal{U}^{-1}(\mathcal{U}(\tilde{x}))$, and therefore, $g_+(x) = -1$. This establishes that in the robust region of $\hat{h}$, $\mathrm{Rob}_{\mathcal{U}}(\hat{h})$, the predictions of $g_+$ on *natural* examples $x \sim D$ are equal to the predictions of $\hat{h}$. We will use this observation, in addition to the fact that the robust risk of $\hat{h}$ is small ($\Pr_{x \sim D}\left[\exists z \in \mathcal{U}(x) : \hat{h}(z) \neq c(x)\right] \leq \varepsilon$) to show that the error of $g_+$ on natural examples is small. Specifically, by law of total probability,

$$\begin{aligned}
\Pr_{x \sim D}\left[g_+(x) \neq c(x)\right] &= \Pr_{x \sim D}\left[g_+(x) \neq c(x) \wedge x \in \mathrm{Rob}_{\mathcal{U}}(\hat{h})\right] + \Pr_{x \sim D}\left[g_+(x) \neq c(x) \wedge x \notin \mathrm{Rob}_{\mathcal{U}}(\hat{h})\right] \\
&= \Pr_{x \sim D}\left[\hat{h}(x) \neq c(x) \wedge x \in \mathrm{Rob}_{\mathcal{U}}(\hat{h})\right] + \Pr_{x \sim D}\left[g_+(x) \neq c(x) \wedge x \notin \mathrm{Rob}_{\mathcal{U}}(\hat{h})\right] \\
&\leq \Pr_{x \sim D}\left[\hat{h}(x) \neq c(x) \wedge x \in \mathrm{Rob}_{\mathcal{U}}(\hat{h})\right] + \Pr_{x \sim D}\left[x \notin \mathrm{Rob}_{\mathcal{U}}(\hat{h})\right] \leq \varepsilon + \varepsilon = 2\varepsilon.
\end{aligned}$$

Finally, observe that for any $x \in \mathrm{Rob}_{\mathcal{U}}(\hat{h})$ such that $\hat{h}(x) = +1$, by definition of $g_+$, it holds that $x \in \mathrm{Rob}_{\mathcal{U}^{-1}(\mathcal{U})}(g_+)$, thus

$$\begin{aligned}\Pr_{x\sim D}\left[x \in \mathrm{Rob}_{\mathcal{U}^{-1}(\mathcal{U})}(g_+)\right] &\geq \Pr_{x\sim D}\left[x \in \mathrm{Rob}_{\mathcal{U}}(\hat{h}) \wedge \hat{h}(x) = +1\right] \\ &= \Pr_{x\sim D}\left[x \in \mathrm{Rob}_{\mathcal{U}}(\hat{h})\right] \Pr_{x\sim D}\left[\hat{h}(x) = +1 \middle| x \in \mathrm{Rob}_{\mathcal{U}}(\hat{h})\right] \\ &\geq (1-\varepsilon) \Pr_{x\sim D}\left[\hat{h}(x) = +1 \middle| x \in \mathrm{Rob}_{\mathcal{U}}(\hat{h})\right].\end{aligned}$$

□

We are now ready to proceed with the proof of Theorem 5.9.

*Proof of Theorem 5.9.* Let $\mathcal{U}$ be an arbitrary perturbation set, and $\mathbb{B}$ an $(\varepsilon, \delta)$-robust learner for some *unknown* target concept $c : \mathcal{X} \to \mathcal{Y}$ with respect to $\mathcal{U}$. We will construct another learner $\tilde{\mathbb{B}}$ that $(\beta, 2\varepsilon, 2\delta)$-barely-robustly-learns $c$ with respect to $\mathcal{U}^{-1}(\mathcal{U})$, with $\beta = (1-\varepsilon)/2$. Let $D$ be some unknown distribution over $\mathcal{X}$ that is robustly realizable: $\Pr_{x\sim D}\left[\exists z \in \mathcal{U}(x) : c(z) \neq c(x)\right] = 0$.

Description of $\tilde{\mathbb{B}}$. Sample $S \sim D_c^{m_{\mathbb{B}}(\varepsilon,\delta)}$, and run learner $\mathbb{B}$ on $S$. Let $\hat{h} = \mathbb{B}(S)$ be the predictor returned by $\mathbb{B}$. Let $\tilde{m} \geq \frac{64}{9}\ln(1/\delta)$. For each $1 \leq i \leq \tilde{m}$, consider the following process: draw an example $(x, y) \sim D_c$. If $x \in \mathrm{Rob}_{\mathcal{U}}(\hat{h})$ terminate, otherwise repeat the process again. Let $\tilde{S} = \{(x_1, y_1), \ldots, (x_{\tilde{m}}, y_{\tilde{m}})\}$ be the sample resulting from this process. Calculate $M_+ = \frac{1}{|\tilde{S}|}\sum_{x\in\tilde{S}} 1[\hat{h}(x) = +1]$. If $M_+ \geq 1/2$, output $g_+$, otherwise, output $g_-$ (as defined in Lemma 5.10).

Analysis. With probability at least $1-\delta$ over $S \sim D_c^m$, $\hat{h}$ has small robust risk: $\mathrm{R}_{\mathcal{U}}(\hat{h}; D_c) \leq \varepsilon$. Lemma 5.10 implies then that for each $y \in \{\pm 1\}$, $g_y$ satisfies:

$$\Pr_{x\sim D}\left[g_y(x) \neq c(x)\right] \leq 2\varepsilon \text{ and } \Pr_{x\sim D}\left[x \in \mathrm{Rob}_{\mathcal{U}^{-1}(\mathcal{U})}(g_y)\right] \geq (1-\varepsilon) \Pr_{x\sim D}\left[\hat{h}(x) = y \middle| x \in \mathrm{Rob}_{\mathcal{U}}(\hat{h})\right].$$

It remains to show that with probability at least $1-\delta$ over $\tilde{S} \sim D^{\tilde{m}}$, for $g_{\hat{y}}$ returned by $\tilde{\mathbb{B}}$:

$$\Pr_{x\sim D}\left[\hat{h}(x) = \hat{y} \middle| x \in \mathrm{Rob}_{\mathcal{U}}(\hat{h})\right] \geq \frac{1}{2}.$$

Observe that by the rejection sampling mechanism of $\tilde{\mathbb{B}}$, $\tilde{S}$ is a sample from the region of distribution $D$ where $\hat{h}$ is robust. Furthermore, we know that

$$\max\left\{\Pr_{x\sim D}\left[\hat{h}(x)=+1\middle|x\in \mathrm{Rob}_{\mathcal{U}}(\hat{h})\right], \Pr_{x\sim D}\left[\hat{h}(x)=-1\middle|x\in \mathrm{Rob}_{\mathcal{U}}(\hat{h})\right]\right\}\geq \frac{1}{2}.$$

Without loss of generality, suppose that $p = \Pr_{x\sim D}\left[\hat{h}(x)=+1\middle|x\in \mathrm{Rob}_{\mathcal{U}}(\hat{h})\right] \geq 1/2$. Then, the failure event is that $\tilde{\mathbb{B}}$ outputs $g_{-}$, i.e. the event that $M_{+} < \frac{1}{2}$. By a standard application of the Chernoff bound, we get that

$$\Pr_{\tilde{S}}\left[M_{+}<\frac{1}{2}\right]\leq e^{\frac{-\tilde{m}p\frac{1}{4}\left(\frac{1}{2}-\frac{1}{p}\right)^2}{2}}\leq e^{-\frac{9\tilde{m}}{64}}\leq \delta,$$

where the last inequality follows from the choice of $\tilde{m}$ in the description of $\tilde{\mathbb{B}}$.

Finally, to conclude, observe that the sample complexity of learner $\tilde{\mathbb{B}}$ is equal to $m_{\mathbb{B}}(\varepsilon,\delta)$ plus the number of samples drawn from $D$ to construct $\tilde{S}$. For each $1\leq i\leq \tilde{m}$, let $X_i$ be the number of samples drawn from $D$ until a sample from the robust region $\mathrm{Rob}_{\mathcal{U}}(\hat{h})$ was observed. Note that $X_i$ is a geometric random variable with mean at most $1/(1-\varepsilon)$. By a standard concentration inequality for the sums of i.i.d. geometric random variables (Brown, 2011),

$$\Pr\left[\sum_{i=1}^{\tilde{m}} X_i > 2\frac{\tilde{m}}{1-\varepsilon}\right]\leq e^{-\frac{\tilde{m}}{4}}\leq\delta,$$

where the last inequality follows from the choice of $\tilde{m}$ in the description of $\tilde{\mathbb{B}}$. Thus, with probability at least $1-\delta$, the total sample complexity is $m_{\mathbb{B}}(\varepsilon,\delta)+\frac{2\tilde{m}}{1-\varepsilon}$. This concludes that learner $\tilde{\mathbb{B}}$ $(\beta, 2\varepsilon, 2\delta)$-barely-robustly-learns $c$ w.r.t. $\mathcal{U}^{-1}(\mathcal{U})$, where $\beta=(1-\varepsilon)/2$. □

As mentioned earlier, Theorem 5.9 still leaves open the question of whether the *weaker* requirement of barely robust learning with respect to $\mathcal{U}$ suffices for strongly robust learning with respect to $\mathcal{U}$. We show next that this weaker requirement is *not* sufficient.

**Theorem 5.11.** *There is an instance space $\mathcal{X}$, a perturbation set $\mathcal{U}$, and a class $\mathcal{C}$ such that $\mathcal{C}$ is $(\beta=\frac{1}{2}, \varepsilon=0,\delta)$-barely-robustly-learnable with respect to $\mathcal{U}$, but $\mathcal{C}$ is* not *$(\varepsilon,\delta)$-robustly-learnable with respect to $\mathcal{U}$ for any $\varepsilon<1/2$.*

Before proceeding with the proof of Theorem 5.11, we briefly sketch the high-level argument. In order to show this impossibility result, we construct a collection of distributions and show that this collection is *barely* robustly learnable with respect to $\mathcal{U}$ with robustness parameter $\beta = \frac{1}{2}$ and natural error $\varepsilon = 0$ using a randomized predictor. We also show that it is not possible to robustly learn this collection with robust risk strictly smaller than 1/2. The second part is shown by relying on a necessary condition for strongly robust learning we proposed in Chapter 3 (Section 3.4, Theorem 3.11) which is the finiteness of the robust shattering dimension (Definition 3.10).

We are now ready to proceed with the proof of Theorem 5.11.

*Proof of Theorem 5.11.* Pick three infinite unique sequences $(x_n^+)_{n\in\mathbb{N}}$, $(x_n^-)_{n\in\mathbb{N}}$, and $(z_n)_{n\in\mathbb{N}}$ from $\mathbb{R}^2$ such that for each $n \in \mathbb{N} : x_n^+ = (n, 1), x_n^- = (n, -1), z_n = (n, 0)$, and let $\mathcal{X} = \cup_{n\in\mathbb{N}} \{x_n^+, x_n^-, z_n\}$. We now describe the construction of the perturbation set $\mathcal{U}$. For each $n \in \mathbb{N}$, let $\mathcal{U}(x_n^+) = \{x_n^+, z_n\}$, $\mathcal{U}(x_n^-) = \{x_n^-, z_n\}$, and $\mathcal{U}(z_n) = \{z_n, x_n^+, x_n^-\}$.

We now describe the construction of the concept class $\mathcal{C}$. For each $\boldsymbol{y} \in \{\pm 1\}^{\mathbb{N}}$ define $h_{\boldsymbol{y}} : \mathcal{X} \to \mathcal{Y}$ to be:

$$\forall n \in \mathbb{N} : h_{\boldsymbol{y}}(z_n) = y_n \wedge h_{\boldsymbol{y}}(x_n^+) = +1 \wedge h_{\boldsymbol{y}}(x_n^-) = -1. \tag{5.11}$$

Let $\mathcal{C} = \left\{h_{\boldsymbol{y}} : \boldsymbol{y} \in \{\pm 1\}^{\mathbb{N}}\right\}$. Observe that by construction of $\mathcal{U}$ and $\mathcal{C}$, $\mathcal{C}$ robustly shatters the sequence $(z_n)_{n\in\mathbb{N}}$ with respect to $\mathcal{U}$ (see Definition 4.6), and therefore, the robust shattering dimension of $\mathcal{C}$ with respect to $\mathcal{U}$, $\dim_{\mathcal{U}}(\mathcal{C})$, is infinite. Thus, Theorem 3.11 implies that $\mathcal{C}$ is not $(\varepsilon, \delta)$-*strongly*-robustly-learnable with respect to $\mathcal{U}$.

We will now show that there is a simple learner $\mathbb{A}$ that $(\beta, \varepsilon, \delta)$-*barely*-robustly-learns $\mathcal{C}$ with respect to $\mathcal{U}$, with robustness parameter $\beta = \frac{1}{2}$ and natural error $\varepsilon = 0$. Specifically, $\mathbb{A}$ samples a bitstring $\tilde{\boldsymbol{y}} \in \{\pm 1\}^{\mathbb{N}}$ uniformly at random, and outputs the classifier $h_{\tilde{\boldsymbol{y}}}$. Learner $\mathbb{A}$ will not require any data as input.

We now proceed with analyzing the performance of learner $\mathbb{A}$. Let $h_{\boldsymbol{y}} \in \mathcal{C}$ be some unknown target concept and $D$ be some unknown distribution over $\mathcal{X}$ that is robustly realizable:

$\Pr_{x\sim D}\left[\exists z \in \mathcal{U}(x) : h_{\boldsymbol{y}}(z) \neq y\right] = 0$. Since $D$ is robustly realizable, by construction of $\mathcal{U}$ and $\mathcal{C}$, this implies that

$$\forall n \in \mathbb{N} : D(z_n) = 0 \text{ and } D(x_n^{-y_n}) = 0. \tag{5.12}$$

This is because $\mathcal{U}(z_n) = \{z_n, x_n^+, x_n^-\}$ and Equation 5.11 implies that $h_{\boldsymbol{y}}$ is not robust on $z_n$ since $h_{\boldsymbol{y}}(x_n^+) \neq h_{\boldsymbol{y}}(x_n^-)$, also $\mathcal{U}(x_n^+) \cap \mathcal{U}(x_n^-) = \{z_n\}$ and since $h_{\boldsymbol{y}}(z_n) = y_n$ this implies that $h_{\boldsymbol{y}}$ is not robust on $x_n^{-y_n}$. Equation 5.12 and Equation 5.11 together imply that the random classifier $h_{\tilde{\boldsymbol{y}}} \in \mathcal{C}$ selected by learner $\mathbb{A}$ has *zero* error on *natural* examples: with probability 1 over $\tilde{\boldsymbol{y}}$, $\Pr_{x\sim D}\left[h_{\tilde{\boldsymbol{y}}}(x) \neq h_{\boldsymbol{y}}(x)\right] = 0$.

We now turn to analyzing the *robust* risk of learner $\mathbb{A}$,

$$\begin{aligned}
\mathop{\mathbb{E}}_{\tilde{\boldsymbol{y}}}\left[\mathop{\mathbb{E}}_{x\sim D}\left[1\left\{\exists z \in \mathcal{U}(x) : h_{\tilde{\boldsymbol{y}}}(z) \neq h_{\boldsymbol{y}}(x)\right\}\right]\right] &= \mathop{\mathbb{E}}_{x\sim D}\left[\mathop{\mathbb{E}}_{\tilde{\boldsymbol{y}}}\left[1\left\{\exists z \in \mathcal{U}(x) : h_{\tilde{\boldsymbol{y}}}(z) \neq h_{\boldsymbol{y}}(x)\right\}\right]\right] \\
&= \sum_{n\in\mathbb{N}} D(x_n^{y_n}) \mathop{\mathbb{E}}_{\tilde{\boldsymbol{y}}}\left[1\left\{\exists z \in \mathcal{U}(x_n^{y_n}) : h_{\tilde{\boldsymbol{y}}}(z) \neq h_{\boldsymbol{y}}(x_n^{y_n})\right\}\right] \\
&= \sum_{n\in\mathbb{N}} D(x_n^{y_n}) \mathop{\mathbb{E}}_{\tilde{\boldsymbol{y}}}\left[1\left\{h_{\tilde{\boldsymbol{y}}}(z_n) \neq h_{\boldsymbol{y}}(x_n^{y_n})\right\}\right] \\
&= \sum_{n\in\mathbb{N}} D(x_n^{y_n}) \mathop{\mathbb{E}}_{\tilde{\boldsymbol{y}}}\left[1\left\{\tilde{y}_n \neq y_n\right\}\right] = \sum_{n\in\mathbb{N}} D(x_n^{y_n})\frac{1}{2} = \frac{1}{2}.
\end{aligned}$$

This implies that in expectation over randomness of learner $\mathbb{A}$, it will be robust on half the mass of distribution $D$: $\mathbb{E}_{\tilde{\boldsymbol{y}}}\, \mathbb{E}_{x\sim D}\, 1[x \in \mathrm{Rob}_{\mathcal{U}}(h_{\tilde{\boldsymbol{y}}})] = \frac{1}{2}$. □

## 5.5 Discussion and Open Directions

In this paper, we put forward a theory for *boosting* adversarial robustness. We discuss below practical implications and outstanding directions that remain to be addressed.

Practical implications. Our algorithm $\beta$-RoBoost is generic and can be used with any black-box barely robust learner $\mathbb{A}$. In the context of deep learning and $\ell_p$ robustness, our results suggest the following: for targeted robustness of radius $\gamma$, use an adversarial learning method (e.g., Madry et al., 2018, Zhang et al., 2019b, Cohen et al., 2019) to learn a neural net $h^1_{\mathrm{NN}}$ predictor robust with radius $2\gamma$, then filter the training examples to include *only* the ones on which $h^1_{\mathrm{NN}}$ is *not* robust with radius $2\gamma$, and repeat the training process on the filtered examples to learn a second neural net, and so on. Finally, use the cascade of neural nets $\mathrm{CAS}(h^1_{\mathrm{NN}}, h^2_{\mathrm{NN}}, \dots)$ as defined in $\beta$-RoBoost to predict.

It would be interesting to empirically explore whether adversarial learning methods (e.g., Madry et al., 2018, Zhang et al., 2019b, Cohen et al., 2019) satisfy the *barely* robust learning condition: on each round of boosting, the learning algorithm can shrink the fraction of the training examples on which the predictor from the previous round is not robust on. This is crucial for progress.

Multiclass. We would like to emphasize that our theory for boosting *barely robust* learners extends seamlessly to multiclass learning problems. In particular, when the label space $|\mathcal{Y}| > 2$, we obtain the same guarantees in Theorem 5.3 using the same algorithm $\beta$-RoBoost. The other direction of converting an $(\varepsilon, \delta)$-robust-learner with respect to $\mathcal{U}$ to an $(\beta, 2\varepsilon, 2\delta)$-barely-robust-learner with respect to $\mathcal{U}^{-1}(\mathcal{U})$ also holds, using the same technique in Theorem 5.9, but now we get $\beta = \frac{1-\varepsilon}{|\mathcal{Y}|}$.

Relationship between robustness parameter $\beta$ and error parameter $\varepsilon$. To achieve robust risk at most $\varepsilon$ using a barely robust learner $\mathbb{A}$ with robustness parameter $\beta$, our algorithm $\beta$-RoBoost requires $\mathbb{A}$ to achieve a natural error of $\tilde{\varepsilon} = \frac{\beta\varepsilon_0}{2}$ for any constant $\varepsilon_0 < \frac{1}{2}$ (say $\tilde{\varepsilon} = \frac{\beta}{6}$) (see Corollary 5.4). It would be interesting to resolve whether requiring natural risk $\tilde{\varepsilon}$ that depends on $\beta$ is necessary, or whether it is possible to avoid dependence on $\beta$. Concretely, an open question here is: can we achieve robust risk at most $\varepsilon$ using a $(\beta, O(\varepsilon), \delta)$-barely-robust-learner instead of requiring $(\beta, O(\beta), \delta)$-barely-robust-learner? It actually suffices to answer the following: given a $(\beta, \frac{1}{3}, \delta)$-barely-robust-learner, is it possible to achieve *robust* risk at most $\frac{1}{3}$? A related question is whether it is possible to boost the error while maintaining robustness fixed at some level. For example, given a $(\beta, \frac{1}{3}, \delta)$-barely-robust-learner $\mathbb{A}$, is it possible to boost this to a $(\beta, \varepsilon, \delta)$-barely-robust-learner $\mathbb{B}$?

Agnostic setting. We focused only on boosting robustness in the *realizable* setting, where the target unknown concept $c$ and the unknown distribution $D$ satisfy $\mathrm{R}_{\mathcal{U}}(c; D_c) = 0$, and we showed an equivalence between barely-robust-learning and strongly-robust-learning in this setting. In terms of obtaining stronger guarantees, we remark that an agnostic-to-realizable reduction described in Chapter 3 (Subsection 3.3.2), implies, together with Theo-rem 5.3, that barely-robust-learning is equivalent to agnostic-robust-learning. This agnostic-to-realizable reduction is not oracle-efficient, however, so the interesting open question here is to explore oracle-efficient boosting algorithms in the agnostic setting. A key step in this direction is to identify reasonable definitions of *agnostic*-barely-robust learners that go beyond Definition 5.1.

Computational efficiency. Our algorithm $\beta$-RoBoost is guaranteed to terminate in at most $T = \frac{\ln(2/\varepsilon)}{\beta}$ rounds of boosting (see Theorem 5.3), because our rejection sampling mechanism (Step 8) will safely terminate when the mass of the non-robust region becomes small (mass at most $\varepsilon/2$). In each round $t > 1$ of boosting, rejection sampling (Step 7) in $\beta$-RoBoost needs to sample from the region of the distribution where all previous predictors $h_1, \dots, h_{t-1}$ are *non-robust*, and doing this computationally efficiently requires an efficient procedure to *certify* robustness of $h_i$ on perturbation set $\mathcal{U}^{-1}(\mathcal{U})(x)$. Certifying robustness of predictors is an active area of research (e.g., randomized smoothing with $\ell_2$ perturbations in Cohen et al., 2019), but is outside the scope of this work, since we make no assumptions on the form of the returned predictors $h_i$ or the perturbation sets $\mathcal{U}$ and $\mathcal{U}^{-1}(\mathcal{U})$. In general, it's hard to imagine a world where efficient robust learning is possible without efficient certification of robustness. In other words, efficient certification of robustness seems to be a "necessary" condition for efficient robust learning, not just for our proposed boosting algorithm, but for any algorithm that is devised for robust learning. Without efficient certification of robustness, we can't even compute the robust loss, so how can we even learn.

Comparing $\mathcal{U}^{-1}(\mathcal{U})$ vs. $\mathcal{U}$. As mentioned earlier in the introduction, if $\mathcal{U}$ is a metric-ball of radius $\gamma$, then $\mathcal{U}^{-1}(\mathcal{U})$ is a metric-ball of radius $2\gamma$. More generally, we think of robustness w.r.t. $\mathcal{U}^{-1}(\mathcal{U})$ as being robust to "twice" the adversary's power (which is represented by $\mathcal{U}$). For example, for a general vector space $\mathcal{X}$, if $\mathcal{U}(x) = x + B$, where $B$ is some arbitrary symmetric set (i.e. $v \in B$ if and only if $-v \in B$), then $\mathcal{U}^{-1}(\mathcal{U})(x) = x + B + B$. We do not yet know of examples of $\mathcal{U}$ where $\mathcal{U}^{-1}(\mathcal{U})$ is much "larger" than $\mathcal{U}$, but this is an interesting direction to explore. More generally, it would be interesting to explore whether barely-robust-learning w.r.t. $\mathcal{U}^{-1}(\mathcal{U})$ is *computationally* equivalent to strong-robust-learning w.r.t. $\mathcal{U}$.

## 5.6 Simple Experiments

We conduct simple experiments that illustrate the utility of our theoretical contribution for boosting robustness. These experiments demonstrate that our algorithm, $\beta$-RoBoost, can boost and improve the robustness of black-box learning algorithms. We describe the setup and the results below.

Datasets. A synthetic binary classification dataset (`make_moons` from scikit-learn), and `MNIST` (rescaled by dividing by 255, and converted to binary classification of odd vs. even).

Perturbation set $\mathcal{U}$. We consider $\ell_2$ perturbations of some radius $\gamma$. In the respective datasets, we computed the minimum distance between examples from different classes and chose a radius $\gamma$ that's smaller than this minimum distance.

Barely Robust Learner. We use an off-the-shelve Linear SVM solver (from scikit-learn) as a barely-robust-learner. Using linear predictors and $\ell_2$ perturbations simplifies the computation of the robust loss since it exactly corresponds to computing the margin loss (see Lemma 8.13 in Chapter 8). Formally, for any linear predictor $w$ and example $(x, y)$,

$$\sup_{\|\delta\|_2 \leq \gamma} 1[\mathrm{sign}(\langle w, x + \delta \rangle) \neq y] = 1\left[y \left\langle \frac{w}{\|w\|_2}, x \right\rangle \leq \gamma\right].$$

We ran our boosting algorithm, $\beta$-RoBoost, and compared it against the baseline of a single Linear SVM call. In our boosting algorithm, we run for as many rounds as possible until there are no more examples left in the training set to run LinearSVM on.

Results On `make_moons` with perturbation radius $\gamma = 0.1$, the baseline Linear SVM achieves a robust accuracy of 84.78%, while $\beta$-RoBoost (with 2 rounds of boosting) achieves robust accuracy of 89.86%. On `MNIST` with perturbation radius $\gamma = 0.5$, the baseline Linear SVM achieves a robust accuracy of 73.9%, while $\beta$-RoBoost (with 2 rounds of boosting) achieves robust accuracy of 80.05%. Finally, on `MNIST` with a bigger perturbation radius $\gamma = 1.0$, the baseline Linear SVM achieves a robust accuracy of 48.1%, while $\beta$-RoBoost (with 2 rounds of boosting) achieves robust accuracy of 70.12%.

We observe that $\beta$-RoBoost improves the robustness of Linear SVM. Notice that even in the regime where the baseline Linear SVM archives robust-accuracy < 50% (MNIST with perturbation radius 1.0), $\beta$-RoBoost can actually improve the robust-accuracy beyond 50%.

We include code to reproduce our `MNIST` experiments with perturbation radius $\gamma = 1.0$ in Appendix C.3.

# 6

# Reductions to Non-Robust Learners

## 6.1 Introduction

In Chapter 3, we showed that if a hypothesis class $\mathcal{C}$ is PAC learnable non-robustly (i.e., $\mathcal{C}$ has finite VC dimension), then $\mathcal{C}$ is also adversarially robustly learnable. The approach we took in Chapter 3 can be interpreted as using black-box access to an oracle $\mathsf{RERM}_{\mathcal{C}}$ minimizing the robust *empirical* risk:

$$\hat{h} \in \mathsf{RERM}_{\mathcal{C}}(S) \triangleq \underset{h\in\mathcal{C}}{\operatorname{argmin}} \frac{1}{m}\sum_{i=1}^{m} \sup_{z\in\mathcal{U}(x)} \mathbb{1}[h(z) \neq y]. \tag{6.1}$$

But this goes well beyond using just a *non-robust* learning algorithm, or even standard empirical risk minimization (ERM). In this chapter, we are interested in understanding further the relationship between adversarially robust learning and standard non-robust learning. We consider the following question:

*Can we learn adversarially robust predictors given only black-box access to a non-robust learner?*

That is, we are asking whether it is possible to reduce adversarially robust learning to standard non-robust learning. Since we have a plethora of algorithms devised for standard non-robust learning, it would be useful if we could design efficient *reduction* algorithms that leverage such non-robust learning algorithms in a black-box manner to learn *robustly*. That is, design generic wrapper methods that take as input a learning algorithm $\mathbb{A}$ and a specification of the perturbation set $\mathcal{U}$, and robustly learn by calling $\mathbb{A}$. Many systems in practice perform standard learning but with no robustness guarantees, and therefore, it would be beneficial to provide wrapper procedures that can guarantee adversarial robustness in a black-box manner without needing to modify current learning systems internally.

Related Work Recent work (Mansour, Rubinstein, and Tennenholtz, 2015, Feige, Mansour, and Schapire, 2015, 2018, Attias, Kontorovich, and Mansour, 2019) can be interpreted as giving reduction algorithms for adversarially robust learning. Specifically, Feige et al. (2015) gave a reduction algorithm that can robustly learn a *finite* hypothesis class $\mathcal{C}$ using black-box access to an ERM for $\mathcal{C}$. Later, Attias et al. (2019) improved this to handle *infinite* hypothesis classes $\mathcal{C}$. But their complexity and the number of calls to ERM depend super-linearly on the number of possible perturbations $|\mathcal{U}| = \sup_x |\mathcal{U}(x)|$, which is undesirable for most types of perturbations—we completely avoid a sample complexity dependence on $|\mathcal{U}|$, and reduce the oracle complexity to at most a poly-logarithmic dependence. Furthermore, their work assumes access specifically to an ERM procedure, which is a very specific type of learner, while we only require access to any method that PAC-learns $\mathcal{C}$ and whose image has bounded VC-dimension.

A related goal was explored by Salman, Sun, Yang, Kapoor, and Kolter (2020): They proposed a method to *robustify pre-trained predictors*. Their method takes as input a black-box *predictor* (not a learning algorithm) and a point $x$, and outputs a label prediction $y$ for $x$ and a radius $r$ such that the label $y$ is robust to $\ell_2$ perturbations of radius $r$. But this doesn't guarantee that the predictions $y$ are correct, nor that the radius $r$ would be what we desire, and even if the predictor was returned by a learning algorithm and has a very small non-robust error, we do not end up with any gurantee on the robust risk of the robustified predictor. In this chapter, we require black-box access to a *learning algorithm* (not just to a single predictor), but we output a predictor that *is* guaranteed to have *small* robust risk (if one exists in the class, see Definition 2.2). We also provide a general treatment for arbitrary adversaries $\mathcal{U}$, not just $\ell_p$ perturbations.

Efficient Reductions From a computational perspective, the relationship between standard non-robust learning and adversarially robust learning is not well-understood. It is natural to wonder whether there is a general efficient reduction for adversarially robust learning, using only non-robust learners. Recent work has provided strong evidence that this is not the case in general. Specifically, Bubeck, Lee, Price, and Razenshteyn (2019) showed that there exists a learning problem that can be learned efficiently non-robustly, but is computationally intractable to learn robustly (under plausible complexity-theoretic assumptions). In this chapter, we aim to understand when such efficient reductions *are* possible.

### 6.1.1 Main Results

When studying reductions of adversarially robust learning to non-robust learning, an important aspect emerges regarding the form of access that the reduction algorithm has to the perturbation set $\mathcal{U}$. How should we model access to the sets of adversarial perturbations represented by $\mathcal{U}$?

In Section 6.3, we study the setting where the reduction algorithm has explicit access/-knowledge of the possible adversarial perturbations induced by the perturbation set $\mathcal{U}$ on the *training examples*. We first show that there is an algorithm that can learn adversarially robust predictors with black-box oracle access to a non-robust algorithm:

**Theorem 6.2.** *For any perturbation set $\mathcal{U}$, Algorithm 6.1 robustly learns any target class $\mathcal{C}$ using any black-box non-robust PAC learner $\mathbb{A}$ for $\mathcal{C}$, with $O(\log^2 |\mathcal{U}|)$ oracle calls to $\mathbb{A}$ and sample complexity independent of $|\mathcal{U}|$.*

The oracle complexity dependence on $|\mathcal{U}|$, even if only logarithmic, might be disappointing, but we show it is unavoidable:

**Theorem 6.6.** *There exists a perturbation set $\mathcal{U}$ such that for any reduction algorithm $\mathbb{B}$, there exists a target class $\mathcal{C}$ and a PAC learner $\mathbb{A}$ for $\mathcal{C}$ such that $\Omega(\log |\mathcal{U}|)$ oracle queries to $\mathbb{A}$ are necessary to robustly learn $\mathcal{C}$.*

This tells us that only requiring a non-robust PAC learner $\mathbb{A}$ is not enough to avoid the $\log |\mathcal{U}|$ dependence, even with explicit knowledge of $\mathcal{U}$.

## 6.2 Preliminaries

Let $\mathcal{X}$ denote the instance space and $\mathcal{Y} = \{\pm 1\}$ denote the label space. Let $\mathcal{U} : \mathcal{X} \to 2^{\mathcal{X}}$ denote an arbitrary perturbation set. For any perturbation set $\mathcal{U}$, denote by $|\mathcal{U}| \triangleq \sup_x |\mathcal{U}(x)|$ the number of allowed adversarial perturbations. We refer the reader to Chapter 2 for formal definitions of: PAC learning (Definition 2.4), robust PAC learning (Definition 2.2), VC dimension (Definition 2.5), and the dual VC dimension (Definition 2.6); which we will use throughout this chapter.

We also formally define what we mean by a *reduction* algorithm:

**Definition 6.1** (Reduction Algorithm). *For a perturbation set $\mathcal{U}$, a reduction algorithm $\mathbb{B}_{\mathcal{U}}$ takes as input a black-box learning algorithm $\mathbb{A}$ and a training set $S \subseteq \mathcal{X} \times \mathcal{Y}$, and can use $\mathbb{A}$ by calling it $T$ times on inputs $\mathbb{B}_{\mathcal{U}}$ constructs each of size $m_0 \in \mathbb{N}$, and outputs a predictor $f \in \mathcal{Y}^{\mathcal{X}}$.*

We emphasize that $\mathbb{B}_{\mathcal{U}}$ is allowed to be adaptive in its calls to $\mathbb{A}$. That is, it can call $\mathbb{A}$ on one constructed data set, then construct another data set depending on the returned predictor, and call $\mathbb{A}$ on this new data set. Such adaptive use of the base learner $\mathbb{A}$ is central to boosting-type constructions.

We know that a hypothesis class $\mathcal{C}$ is PAC learnable if and only if its VC dimension is finite (Vapnik and Chervonenkis, 1971, 1974, Blumer et al., 1989a, Ehrenfeucht et al., 1989). And in this case, $\mathcal{C}$ is properly PAC learnable with $\mathsf{ERM}_{\mathcal{C}}$. We showed in Chapter 3 (Theorem 3.4) that if $\mathcal{C}$ is PAC learnable, then $\mathcal{C}$ is adversarially robustly PAC learnable with an improper learning rule that required a $\mathsf{RERM}_{\mathcal{C}}$ oracle (see Equation 6.1) and sample complexity of $\tilde{O}\left(\frac{\mathrm{vc}(\mathcal{C})\mathrm{vc}^*(\mathcal{C})}{\varepsilon}\right)$. In this chapter, we study whether it is possible to adversarially robustly PAC learn $\mathcal{C}$ using only a black-nox non-robust PAC learner $\mathbb{A}$ for $\mathcal{C}$. We will not require $\mathbb{A}$ to be "proper" (i.e. returns a predictor in $\mathcal{C}$), but we will rely on it returning a predictor in some, possibly much larger, class which still has finite VC-dimension. To this end, we denote by $\mathrm{vc}(\mathbb{A}) = \mathrm{vc}(\mathrm{im}(\mathbb{A}))$ and $\mathrm{vc}^*(\mathbb{A}) = \mathrm{vc}^*(\mathrm{im}(\mathbb{A}))$ the primal and dual VC dimension of the image of $\mathbb{A}$, i.e. the class $\mathrm{im}(\mathbb{A}) = \{\mathbb{A}(S) | S \in (\mathcal{X} \times \mathcal{Y})^*\}$ of the possible predictors $\mathbb{A}$ might return. For $\mathsf{ERM}$, or any other proper learner, $\mathrm{im}(\mathbb{A}) \subseteq \mathcal{C}$ and so $\mathrm{vc}(\mathbb{A}) \leq \mathrm{vc}(\mathcal{C})$ and $\mathrm{vc}^*(\mathbb{A}) \leq \mathrm{vc}^*(\mathcal{C})$.

## 6.3 Learning with Explicitly Specified Adversarial Perturbations

When studying reductions of adversarially robust PAC learning to non-robust PAC learning, an important aspect emerges regarding the form of access that the reduction algorithm has to the perturbation set $\mathcal{U}$. How should we model access to the sets of adversarial perturbations represented by $\mathcal{U}$?

In this section, we explore the setting where the reduction algorithm has explicit knowledge of the perturbation set $\mathcal{U}$. That is, the reduction algorithm knows the set of possible adversarial perturbations for each example in the training set. This is in accordance with what is typically considered in practice, where the perturbation set $\mathcal{U}$ (e.g. $\ell_\infty$ perturbations) is known to the algorithm, and this knowledge is used in adversarial training (see e.g., Madry, Makelov, Schmidt, Tsipras, and Vladu, 2018). Formally, we consider the following question:

> For any perturbation set $\mathcal{U}$, does there exist an algorithm that can learn a target class $\mathcal{C}$ *robustly* with respect to $\mathcal{U}$ given only a black-box non-robust PAC learner $\mathbb{A}$ for $\mathcal{C}$?

We give a positive answer to this question. In Theorem 6.2, we present an algorithm (Algorithm 6.1)—based on the $\alpha$-Boost algorithm (Schapire and Freund, 2012, Section 6.4.2) and our algorithm from Chapter 3 (see Theorem 3.4)— that can adversarially robustly PAC learn a target class $\mathcal{C}$ with only black-box oracle access to a PAC learner $\mathbb{A}$ for $\mathcal{C}$.

**Algorithm 6.1:** Robustify The Non-Robust

**Input:** Training dataset $S = \{(x_1, y_1), \dots, (x_m, y_m)\}$, black-box non-robust learner $\mathbb{A}$

Inflate dataset $S$ to $S_{\mathcal{U}} = \bigcup_{i \le m} \{(z, y_i) : z \in \mathcal{U}(x_i)\}$. `//` $S_{\mathcal{U}}$ `contains all possible perturbations of` $S$.

Set $m_0 = O(\mathrm{vc}(\mathbb{A})\mathrm{vc}(\mathbb{A})^* \log \mathrm{vc}(\mathbb{A})^*)$, and $T = O(\log |S_{\mathcal{U}}|)$.

**for** $1 \le t \le T$ **do**

- Set distribution $D_t$ on $S_{\mathcal{U}}$ as in the $\alpha$-Boost algorithm.
- Sample $S' \sim D_t^{m_0}$, and project $S'$ to dataset $L \subseteq S$ by replacing each perturbation $z$ with its corresponding example $x$.
- Call `ZeroRobustLoss` on $L$, and denote by $f_t$ its output predictor.

Sample $N_{\mathrm{co}} = O\left(\mathrm{vc}^*(\mathbb{A}) \log \mathrm{vc}^*(\mathbb{A})\right)$ i.i.d. indices $i_1, \dots, i_{N_{\mathrm{co}}} \sim \mathrm{Uniform}(\{1, \dots, T\})$.

(repeat previous step until $f = \mathrm{MAJ}(f_{i_1}, \dots, f_{i_{N_{\mathrm{co}}}})$ has $R_{\mathcal{U}}(f; S) = 0$)

**Output:** A majority-vote $\mathrm{MAJ}(f_{i_1}, \dots, f_{i_{N_{\mathrm{co}}}})$ predictor.

`ZeroRobustLoss(`*Dataset L, Learner* $\mathbb{A}$`)`:

- Inflate dataset $L$ to $L_{\mathcal{U}} = \bigcup_{(x,y) \in L} \{(z, y) : z \in \mathcal{U}(x)\}$, and set $T_L = O(\log |L_{\mathcal{U}}|)$
- Run $\alpha$-Boost with black-box access to $\mathbb{A}$ on $L_{\mathcal{U}}$ for $T_L$ rounds.
- Let $h_1, \dots, h_{T_L}$ denote the hypotheses produced by $\alpha$-Boost with $T_L$ oracle queries to $\mathbb{A}$.
- Sample $N = O\left(\mathrm{vc}^*(\mathbb{A})\right)$ i.i.d. indices $i_1, \dots, i_N \sim \mathrm{Uniform}(\{1, \dots, T_L\})$.
- (repeat previous step until $f = \mathrm{MAJ}(h_{i_1}, \dots, h_{i_N})$ has $R_{\mathcal{U}}(f; L) = 0$)
- **return** $f = \mathrm{MAJ}(h_{i_1}, \dots, h_{i_N})$

$\alpha$`-Boost(`*Dataset L, Learner* $\mathbb{A}$`)`:

- Initialize $D_1$ to be uniform over $L$, and set $T_L = O(\log |L|)$.
- **for** $1 \le t \le T_L$ **do**
  - Run $\mathbb{A}$ on $S' \sim D_t^{m_0}$, and denote by $h_t$ its output. (repeat until $\mathrm{err}_{D_t}(h_t) \le 1/3$)
  - Compute a new distribution $D_{t+1}$ by applying the following update for each $(x, y) \in L$:

$$D_{t+1}(x) = \frac{D_t(x)}{Z_t} \times \begin{cases} e^{-2\alpha} & \text{if } h_t(x) = y \\ 1 & \text{otherwise} \end{cases}$$

  where $Z_t$ is a normalization factor and $\alpha$ is set as in Lemma 6.4

- **return** $h_1, \dots, h_{T_L}$.

**Theorem 6.2.** *For any perturbation set* $\mathcal{U}$*, Algorithm 6.1 can* robustly PAC learn *any target class* $\mathcal{C}$ *using black-box oracle calls to any PAC learner* $\mathbb{A}$ *for* $\mathcal{C}$ *with:*

1. *Sample Complexity* $m = O\left(\frac{d{d^*}^2 \log^2 d^*}{\varepsilon} \log\left(\frac{d{d^*}^2 \log^2 d^*}{\varepsilon}\right) + \frac{\log(1/\delta)}{\varepsilon}\right)$,
2. *Oracle Complexity* $T = O\left((\log m + \log |\mathcal{U}|)^2 + \log(1/\delta)\right)$,

*where* $d = \text{vc}(\mathbb{A})$ *and* $d^* = \text{vc}^*(\mathbb{A})$ *are the primal and dual VC dimension of* $\mathbb{A}$.

Importantly, the sample complexity of Algorithm 6.1 is independent of the number of allowed perturbations $|\mathcal{U}|$, in contrast to work by Attias et al. (2019), that can be interpreted as giving a reduction with sample complexity $m \propto |\mathcal{U}| \log |\mathcal{U}|$, and oracle complexity $T \propto |\mathcal{U}| \log^2 |\mathcal{U}|$.

Before proceeding with the proof of Theorem 6.2, we briefly describe our strategy and its main ingredients. Given a dataset $S$ that is robustly realizable by some target concept $c \in \mathcal{C}$, we show that we can use the non-robust learner $\mathbb{A}$ to implement a `RERM` oracle that guarantees zero *empirical* robust loss on $S$ using `ZeroRobustLoss` in Algorithm 6.1. But what about the *population* robust loss? Our main goal to is adversarially robustly learn $\mathcal{C}$ and not just minimize the empirical robust loss. Fortunately, we show that the arguments on *robust* generalization based on sample compression in Chapter 3 (see Theorem 3.4) will still go through when we replace the $\texttt{RERM}_{\mathcal{C}}$ oracle with the `ZeroRobustLoss` procedure in Algorithm 6.1. This is achieved by showing that the image of `ZeroRobustLoss` has bounded VC dimension and *dual* VC dimension. The following lemma, whose proof is provided in Appendix D.1, bounds the *dual* VC dimension of the convex-hull of a class $\mathcal{H}$. This result might be of independent interest.

**Lemma 6.3.** *Let* $\text{co}^k(\mathcal{H}) = \{x \mapsto \text{MAJ}(h_1, \dots, h_k)(x) : h_1, \dots, h_k \in \mathcal{H}\}$*. Then, the dual VC dimension of* $\text{co}^k(\mathcal{H})$ *satisfies* $\text{vc}^*(\text{co}^k(\mathcal{H})) \leq O(d^* \log k)$*.*

In addition, we state two extra key lemmas that will be useful for us in the proof. First, Lemma 6.4 states that running $\alpha$-Boost on a dataset for enough rounds produces a sequence of predictors that achieve zero loss on the dataset (with a margin).

**Lemma 6.4** (see, e.g., Corollary 6.4 and Section 6.4.3 in Schapire and Freund (2012)). *Let* $S = \{(x_i, c(x_i))\}_{i=1}^m$ *be a dataset where* $c \in \mathcal{C}$ *is some target concept, and* $\mathbb{A}$ *an arbitrary PAC* *learner for* $\mathcal{C}$ *(for* $\varepsilon = 1/3$, $\delta = 1/3$*). Then, running* $\alpha$*-Boost (see description in Algorithm 6.1) on* $S$ *with black-box oracle access to* $\mathbb{A}$ *with* $\alpha = \frac{1}{2}\ln\left(1 + \sqrt{\frac{2\ln m}{T}}\right)$ *for* $T = \lceil 112\ln(m)\rceil = O(\log m)$ *rounds suffices to produce a sequence of hypotheses* $h_1, \ldots, h_T \in \mathrm{im}(\mathbb{A})$ *such that*

$$\forall (x, y) \in S, \frac{1}{T}\sum_{i=1}^{T} 1[h_i(x) = y] \geq \frac{5}{9}.$$

*In particular, this implies that the majority-vote* $\mathrm{MAJ}(h_1, \ldots, h_T)$ *achieves zero error on* $S$*.*

Second, Lemma 6.5 describes a sparsification technique due to Moran and Yehudayoff (2016) which allows us to control the complexity of the majority-vote predictors that we use in Algorithm 6.1.

**Lemma 6.5** (Sparsification of Majority Votes, Moran and Yehudayoff (2016))**.** *Let* $\mathcal{H}$ *be a hypothesis class with finite primal and dual VC dimension, and* $h_1, \ldots, h_T$ *be predictors in* $\mathcal{H}$*. Then, for any* $(\varepsilon, \delta) \in (0, 1)$*, with probability at least* $1 - \delta$ *over* $N = O\left(\frac{\mathrm{vc}^*(\mathcal{H}) + \log(1/\delta)}{\varepsilon^2}\right)$ *independent random indices* $i_1, \ldots, i_N \sim \mathrm{Uniform}(\{1, \ldots, T\})$*, we have:*

$$\forall (x, y) \in \mathcal{X} \times \mathcal{Y}, \left|\frac{1}{N}\sum_{j=1}^{N} 1[h_{i_j}(x) = y] - \frac{1}{T}\sum_{i=1}^{T} 1[h_i(x) = y]\right| < \varepsilon.$$

We are now ready to proceed with the proof of Theorem 6.2.

*Proof of Theorem 6.2.* Let $\mathcal{U}$ be an arbitrary perturbation set. Let $\mathcal{C}$ be a target class that is PAC learnable with some PAC learner $\mathbb{A}$. Let $\mathcal{H}$ denote the base class of hypotheses of learner $\mathbb{A}$. Let $d$ denote the VC dimension of $\mathcal{H}$, and $d^*$ denote the dual VC dimension of $\mathcal{H}$. Our proof is divided into two parts.

Zero Empirical Robust Loss. Let $L = \{(x_1, y_1), \ldots, (x_m, y_m)\}$ be a dataset that is *robustly* realizable with some target concept $c \in \mathcal{C}$; in other words, for each $(x, y) \in L$ and each $z \in \mathcal{U}(x)$, $c(z) = y$. We will show that we can use the non-robust learner $\mathbb{A}$ to guarantee zero *empirical* robust loss on $L$. This procedure is described in `ZeroRobustLoss` in Algorithm 6.1. We inflate dataset $L$ to include all possible perturbations under the perturbation set $\mathcal{U}$. Let $L_{\mathcal{U}} = \bigcup_{i \leq m}\{(z, y_i) : z \in \mathcal{U}(x_i)\}$ denote the inflated dataset. Observe that $|L_{\mathcal{U}}| \leq m|\mathcal{U}|$, since each point $x \in \mathcal{X}$ has at most $|\mathcal{U}|$ possible perturbations. We run the $\alpha$-Boost algorithm on the inflated dataset $L_{\mathcal{U}}$ with *black-box* access to PAC learner $\mathbb{A}$, where in each round of boosting $m_0$ samples are fed to $\mathbb{A}$ (where $m_0$ is chosen Step 2). By Lemma 6.4, running $\alpha$-Boost with $T = O(\log(|L_{\mathcal{U}}|))$ oracle calls to $\mathbb{A}$ suffices to produce a sequence of hypotheses $h_1, \dots, h_T \in \mathcal{H}$ such that

$$\forall (z, y) \in L_{\mathcal{U}}, \frac{1}{T}\sum_{i=1}^{T} 1[h_i(z) = y] \geq \frac{5}{9}.$$

Specifically, the above implies that a majority-vote over hypotheses $h_1, \dots, h_T$ achieves zero *robust* loss on dataset $L$, $\mathrm{R}_{\mathcal{U}}(\mathrm{MAJ}(h_1, \dots, h_T); L) = 0$. By Step 13 in `ZeroRobustLoss` in Algorithm 6.1 and Lemma 6.5 (with $\varepsilon = 1/18, \delta = 1/3$), we have that for $N = O(d^*)$, the sampled predictors $h_{i_1}, \dots, h_{i_N}$ satisfy

$$\forall (z, y) \in L_{\mathcal{U}}, \frac{1}{N}\sum_{j=1}^{N} 1[h_{i_j}(z) = y] > \frac{1}{T}\sum_{i=1}^{T} 1[h_i(z) = y] - \frac{1}{18} > \frac{5}{9} - \frac{1}{18} = \frac{1}{2}.$$

Therefore, the majority-vote over the sampled hypotheses $\mathrm{MAJ}(h_{i_1}, \dots, h_{i_N})$ achieves zero robust loss on $L$, $\mathrm{R}_{\mathcal{U}}(\mathrm{MAJ}(h_{i_1}, \dots, h_{i_N}); S) = 0$. Thus, we can implement a `RERM` oracle (see Equation 3.2) using the procedure `ZeroRobustLoss` in Algorithm 6.1. The sparsification step (Step 12) controls the complexity of the image of `ZeroRobustLoss`, i.e., the hypothesis class that is being implicitly used. Specifically, observe that the sparsified predictor $f = \mathrm{MAJ}(h_{i_1}, \dots, h_{i_N})$ lives in $\mathrm{co}^{O(d^*)}(\mathcal{H})$, which is the convex-hull of $\mathcal{H}$ that combines at most $O(d^*)$ predictors. To guarantee *robust* generalization in the next part, it suffices to bound the VC dimension and dual VC dimension of $\mathrm{co}^{O(d^*)}(\mathcal{H})$. By (Blumer et al., 1989a), the VC dimension of $\mathrm{co}^{O(d^*)}(\mathcal{H})$ is at most $O(dd^* \log d^*)$, and by Lemma 6.3, the dual VC dimension of $\mathrm{co}^{O(d^*)}(\mathcal{H})$ is at most $O(d^* \log d^*)$.

Robust Generalization through Sample Compression. This part builds on the approach we took in Chapter 3 (Theorem 3.4). Specifically, we observe that their proof works even if we replace the $\mathsf{RERM}_{\mathcal{C}}$ oracle they used, with our `ZeroRobustLoss` procedure in Algorithm 6.1 that is described above. We provide a self-contained analysis below.

Let $\mathcal{D}$ be an arbitrary distribution over $\mathcal{X} \times \mathcal{Y}$ that is robustly realizable with some concept

$c \in \mathcal{C}$, i.e., $\mathrm{R}_{\mathcal{U}}(c;\mathcal{D}) = 0$. Fix $\varepsilon, \delta \in (0,1)$ and a sample size $m$ that will be determined later. Let $S = \{(x_1, y_1), \dots, (x_m, y_m)\}$ be an i.i.d. sample from $\mathcal{D}$. We run the $\alpha$-Boost algorithm (see Algorithm 6.1) on the inflated dataset $S_{\mathcal{U}}$, this time with `ZeroRobustLoss` as the subprocedure. Specifically, on each round of boosting, $\alpha$-Boost computes an empirical distribution $D_t$ over $S_{\mathcal{U}}$ (according to Step 18). We draw $m_0 = O(dd^* \log d^*)$ samples $S'$ from $D_t$, and *project* $S'$ to a dataset $L_t \subset S$ by replacing each perturbation $(z, y) \in S'$ with its corresponding original point $(x, y) \in S$, and then we run `ZeroRobustLoss` on dataset $L_t$. The projection step is crucial for the proof to work, since we use a *sample compression* argument to argue about *robust* generalization, and the sample compression must be done on the *original* points that appeared in $S$ rather than the perturbations in $S_{\mathcal{U}}$.

By classic PAC learning guarantees (Vapnik and Chervonenkis, 1974, Blumer et al., 1989a), with $m_0 = O(\mathrm{vc}(\mathrm{co}^{O(d^*)}(\mathcal{H}))) = O(dd^* \log d^*)$, we are guaranteed uniform convergence of 0-1 risk over predictors in $\mathrm{co}^{O(d^*)}(\mathcal{H})$ (the effective hypothesis class used by `ZeroRobustLoss`). So, for any distribution $D$ over $\mathcal{X} \times \mathcal{Y}$ with $\inf_{c \in \mathcal{C}} \mathrm{err}(c;\mathcal{D}) = 0$, with nonzero probability over $S' \sim \mathcal{D}^{m_0}$, every $f \in \mathrm{co}^{O(d^*)}(\mathcal{H})$ satisfying $\mathrm{err}_{S'}(f) = 0$, also has $\mathrm{err}_D(f) < 1/3$. By the guarantee of `ZeroRobustLoss` (established above), we know that $f_t = \texttt{ZeroRobustLoss}(L_t, \mathbb{A})$ achieves zero robust loss on $L_t$, $\mathrm{R}_{\mathcal{U}}(f_t; L_t) = 0$, which by definition of the projection means that $\mathrm{err}_{S'}(f_t) = 0$, and thus $\mathrm{err}_{D_t}(f_t) < 1/3$. This allows us to use `ZeroRobustLoss` with $\alpha$-Boost to establish a *robust* generalization guarantee. Specifically, Lemma 6.4 implies that running the $\alpha$-Boost algorithm with $S_{\mathcal{U}}$ as its dataset for $T = O(\log(|S_{\mathcal{U}}|))$ rounds, using `ZeroRobustLoss` to produce the hypotheses $f_t \in \mathrm{co}^{O(d^*)}(\mathcal{H})$ for the distributions $D_t$ produced on each round of the algorithm, will produce a sequence of hypotheses $f_1, \dots, f_T \in \mathrm{co}^{O(d^*)}(\mathcal{H})$ such that:

$$\forall (z, y) \in S_{\mathcal{U}}, \frac{1}{T} \sum_{i=1}^{T} 1[f_i(z) = y] \geq \frac{5}{9}.$$

Specifically, this implies that the majority-vote over hypotheses $f_1, \dots, f_T$ achieves zero *robust* loss on dataset $S$, $\mathrm{R}_{\mathcal{U}}(\mathrm{MAJ}(f_1, \dots, f_T); S) = 0$. Note that each of these classifiers $f_t$ is equal to $\texttt{ZeroRobustLoss}(L_t, \mathbb{A})$ for some $L_t \subseteq S$ with $|L_t| = m_0$. Thus, the classifier $\mathrm{MAJ}(f_1, \dots, f_T)$ is representable as the value of an (order-dependent) reconstruction func-tion $\varphi$ with a compression set size

$$m_0 T = O(\mathrm{vc}(\mathrm{co}^{O(d^*)}(\mathcal{H}))\log(|S_\mathcal{U}|)) = O(dd^*\log d^*\,(\log m + \log|\mathcal{U}|)).$$

This is not enough, however, to obtain a sample complexity bound that is independent of $|\mathcal{U}|$. For that, we will sparsify the majority-vote as in Step 7 in Algorithm 6.1. Lemma 6.5 (with $\varepsilon = 1/18, \delta = 1/3$) guarantees that for $N_{\mathrm{co}} = O(d^*\log d^*)$, the sampled predictors $f_{i_1}, \dots, f_{i_{N_{\mathrm{co}}}}$ satisfy:

$$\forall (z,y) \in S_\mathcal{U}, \frac{1}{N_{\mathrm{co}}}\sum_{j=1}^{N_{\mathrm{co}}} 1[f_{i_j}(z) = y] > \frac{1}{T}\sum_{i=1}^{T} 1[f_i(z) = y] - \frac{1}{18} > \frac{5}{9} - \frac{1}{18} = \frac{1}{2},$$

so that the majority-vote achieves zero robust loss on $S$, $\mathrm{R}_\mathcal{U}(\mathrm{MAJ}(f_{i_1}, \dots, f_{i_{N_{\mathrm{co}}}}); S) = 0$. Since again, each $f_{i_j}$ is the result of ZeroRobustLoss$(L_t, \mathbb{A})$ for some $L_t \subseteq S$ with $|L_t| = m_0$, we have that the classifier $\mathrm{MAJ}(f_{i_1}, \dots, f_{i_{N_{\mathrm{co}}}})$ can be represented as the value of an (order-dependent) reconstruction function $\varphi$ with a compression set size $m_0 N_{\mathrm{co}} = O(dd^*\log d^* \cdot d^*\log d^*) = O(dd^{*2}\log^2(d^*))$. Lemma A.1 which extends to the robust loss the classic compression-based generalization guarantees from the 0-1 loss, implies that for $m \geq cdd^{*2}\log^2(d^*)$ (for an appropriately large numerical constant $c$), with probability at least $1-\delta$ over $S \sim \mathcal{D}^m$,

$$R_\mathcal{U}(\mathrm{MAJ}(f_{i_1}, \dots, f_{i_{N_{\mathrm{co}}}}); \mathcal{D}) \leq O\left(\frac{dd^{*2}\log^2(d^*)}{m}\log(m) + \frac{1}{m}\log(1/\delta)\right).$$

Setting this less than $\varepsilon$ and solving for a sufficient size of $m$ to achieve this yields the stated sample complexity bound.

Our oracle complexity $T$ (number of calls to $\mathbb{A}$) is at most

$O((\log|S_\mathcal{U}|)^2 + \log(1/\delta)) \leq O\left((\log m + \log|\mathcal{U}|)^2 + \log(1/\delta)\right)$, since `ZeroRobustLoss` in Algorithm 6.1 terminates in at most $O(\log|S_\mathcal{U}|)$ rounds each time it is invoked, and it is invoked at most $O(\log|S_\mathcal{U}|)$ times by the outer $\alpha$-Boost algorithm in Algorithm 6.1. Therefore, we have at most $O\left((\log m + \log|\mathcal{U}|)^2\right)$ geometric random variables that represent that number of times Step 19 is invoked, which is the step where learner $\mathbb{A}$ is called. The success probability of Step 19 is a constant (say 2/3), therefore the mean of the sum of the geometric random variables is $O\left((\log m + \log|\mathcal{U}|)^2\right)$. Since sums of geometric random variables concentrate around the mean (Brown, 2011), we get that with probability at least $1-\delta$, the total number of times Step 19 is executed is at most

$O\left((\log m + \log|\mathcal{U}|)^2 + \log(1/\delta)\right)$. This concludes the proof. □

## 6.4 A Lowerbound on the Oracle Complexity

The oracle complexity of Algorithm 6.1 depends on $\log|\mathcal{U}|$. Can this dependence be reduced or avoided? Unfortunately, we show in Theorem 6.6 that the dependence on $\log|\mathcal{U}|$ is unavoidable and *no* reduction algorithm can do better.

It is relatively easy to show, by relying on lower bounds from the boosting literature (Schapire and Freund, 2012, Section 13.2.2), that for any reduction algorithm, there exists a target class $\mathcal{C}$ with $\text{vc}(\mathcal{C}) \leq 1$ and a PAC learner $\mathbb{A}$ for $\mathcal{C}$ such that $\Omega(\log|\mathcal{U}|)$ oracle calls to this PAC learner are necessary to achieve zero *empirical* robust loss. But this is done by constructing a "crazy" improper learner with $\text{vc}(\mathbb{A}) \propto |\mathcal{U}| \gg \text{vc}(\mathcal{C})$.

Perhaps we can ensure better oracle complexity by requiring a more constrained PAC learner $\mathbb{A}$, e.g. with low VC dimension (as in our upper bound), or perhaps even a proper learner, or an ERM, or maybe where $\text{im}(\mathbb{A})$ (and so also $\mathcal{C}$) is finite. We next present a lower bound to show that none of these help improve the oracle complexity. Specifically, we will present a construction showing that for any reduction algorithm $\mathbb{B}$ there is a randomized target class $\mathcal{C}$ and a PAC learner $\mathbb{A}$ for $\mathcal{C}$ with $\text{vc}(\mathbb{A}) = 1$ where $\mathbb{B}$ needs to make $\Omega(\log|\mathcal{U}|)$ oracle calls to $\mathbb{A}$ to robustly learn $\mathcal{C}$. The idea here is that the target class $\mathcal{C}$ is chosen randomly after $\mathbb{B}$, and so $\mathbb{B}$ essentially knows nothing about $\mathcal{C}$ and needs to communicate with $\mathbb{A}$ in order to learn. As a reminder, a reduction algorithm has a budget of $T$ oracle calls to a non-robust learner $\mathbb{A}$, where each oracle call is constructed with $m_0$ points, or more generally a *distribution* over $\mathcal{X} \times \mathcal{Y}$. We next show that any successful reduction requires $T = \Omega(\log|\mathcal{U}|)$ for some non-robust learner $\mathbb{A}$.

**Theorem 6.6.** *For any sufficiently large integer $u$, if $|\mathcal{X}| \geq u^{10u}$, there exists a perturbation set $\mathcal{U}$ with $|\mathcal{U}| = u$ such that for any reduction algorithm $\mathcal{B}$ and for any $\varepsilon > 0$, there exists a target class $\mathcal{C}$ and a PAC learner $\mathcal{A}$ for $\mathcal{C}$ with* $\text{vc}(\mathbb{A}) = 1$ *such that, if the training sample has size at most $(1/8)|\mathcal{U}|^9$, then $\mathbb{B}$ needs to make $T \geq \frac{\log|\mathcal{U}|}{\log(2/\varepsilon)}$ oracle calls to $\mathbb{A}$ in order to robustly*

*learn $\mathcal{C}$.*

*Proof.* We begin with describing the construction of the perturbation set $\mathcal{U}$. Let $m \in \mathbb{N}$; we will construct $\mathcal{U}$ with $|\mathcal{U}| = 2^m$, supposing $|\mathcal{X}| \geq 2\binom{2^{10m}}{2^m} + 2^{10m}$. Let $Z = \{z_1, \dots, z_{2^{10m}}\} \subset \mathcal{X}$ be a set of $2^{10m}$ unique points from $\mathcal{X}$. For each subset $L \subset Z$ where $\left|L\right| = 2^m$, pick a unique pair $x_L^+, x_L^- \in \mathcal{X} \setminus Z$ and define $\mathcal{U}(x_L^+) = \mathcal{U}(x_L^-) = L$. That is, for every choice $L$ of $2^m$ perturbations from $Z$, there is a corresponding pair $x_L^+, x_L^-$ where $\mathcal{U}(x_L^+) = \mathcal{U}(x_L^-) = L$. For any point $x \in \mathcal{X} \setminus Z$ that is remaining, define $\mathcal{U}(x) = \{\}$.

Let $\mathbb{B}$ be an arbitrary reduction algorithm, and let $\varepsilon > 0$ be the error requirement. We will now describe the construction of the target class $\mathcal{C}$. The target class $\mathcal{C}$ will be constructed randomly. Namely, we will first define a labeling $\tilde{h} : Z \to \mathcal{Y}$ on the perturbations in $Z$ that is positive on the first half of $Z$ and negative on the second half of $Z$: $\tilde{h}(z_i) = +1$ if $i \leq \frac{2^{10m}}{2}$, and $\tilde{h}(z_i) = -1$ if $i > \frac{2^{10m}}{2}$. Divide the positive/negative halves into groups of size $2^m$:

$$\underbrace{\{\text{first } 2^m \text{ positives}\}}_{G_1^+}, \dots, \underbrace{\{\text{last } 2^m \text{ positives}\}}_{G_{2^{9m-1}}^+} \Bigg| \underbrace{\{\text{first } 2^m \text{ negatives}\}}_{G_1^-}, \dots, \underbrace{\{\text{last } 2^m \text{ negatives}\}}_{G_{2^{9m-1}}^-}.$$

Let $\varepsilon' = \varepsilon/2$. The target concept $h^* : \mathcal{X} \to \mathcal{Y}$ is generated by randomly flipping the labels of an $\varepsilon'$ fraction of the points in each group $G_1^+, \dots, G_{2^{9m-1}}^+$ from positive to negative and randomly flipping the labels of an $\varepsilon'$ fraction of the points in each group $G_1^-, \dots, G_{2^{9m-1}}^-$ from negative to positive. This defines $h^*$ on $Z$; then for every pair $x^+, x^- \in \mathcal{X} \setminus Z$ where $\mathcal{U}(x^+) = \mathcal{U}(x^-) \neq \{\}$, define $h^*(x^+) = +1$ and $h^*(x^-) = -1$. Once $h^*$ is generated, we define the distribution $D_{h^*}$ over $\mathcal{X} \times \mathcal{Y}$ that will be used in the lower bound by swapping the $\varepsilon'$ fractions of points with the flipped labels in each pair $(G_1^+, G_1^-), \dots, (G_{2^{9m-1}}^+, G_{2^{9m-1}}^-)$ which defines new positive/negative pairs:

$$(G(h^*)_1^+, G(h^*)_1^-), \dots, (G(h^*)_{2^{9m-1}}^+, G(h^*)_{2^{9m-1}}^-).$$

Let $x_i^+, _ = \mathcal{U}^{-1}(G(h^*)_i^+)$ and $_, x_i^- = \mathcal{U}^{-1}(G(h^*)_i^-)$ for each $i \in [2^{9m-1}]$ ($\mathcal{U}^{-1}$ returns a pair of points). Observe that by definition of $h^*$ on $\mathcal{X} \setminus Z$, we have that $h^*(x_i^+) = +1$ and $h^*(x_i^-) = -1$ since $h^*(z) = +1 \forall z \in G(h^*)_i^+$ and $h^*(z) = -1 \forall z \in G(h^*)_i^-$.

Let $D_{h^*}$ be a uniform distribution over $(x_1^+, +1), (x_1^-, -1), \dots, (x_{2^{9m-1}}^+, +1), (x_{2^{9m-1}}^-, -1)$.

Let $T \leq \frac{\log 2^m}{\log(1/\varepsilon')}$. Define a randomly-constructed target class $\mathcal{C} = \{h_1, \dots, h_T, h_{T+1}\}$

where $h_{T+1} = h^*$ and $h_1, h_2, \ldots, h_T$ are generated according the following process: If $t = 1$, then $h_1 := \tilde{h}$ (augmented to all of $\mathcal{X}$ by letting $\tilde{h}(x) = h^*(x)$ for all $x \in \mathcal{X} \setminus Z$). For $t \geq 2$, let $\mathrm{DIS}_{t-1} = \{z \in Z : h_{t-1}(z) \neq h^*(z)\}$, and construct $h_t$ by flipping a uniform randomly-selected $1 - \varepsilon'$ fraction of the labels of $h_{t-1}$ in $G_i^+ \cap \mathrm{DIS}_{t-1}$ and $1 - \varepsilon'$ fraction of the labels of $h_{t-1}$ in $G_i^- \cap \mathrm{DIS}_{t-1}$ for each $i \in [2^{9m-1}]$. Observe that by construction, $h_1, \ldots, h_T$ satisfy the property that they agree with $h^*$ on $\mathcal{X} \setminus Z$, i.e. $h_t(x) = h^*(x)$ for each $t \leq T$ and each $x \in \mathcal{X} \setminus Z$.

We now state a few properties of the randomly-constructed target class $\mathcal{C}$ that we will use in the remainder of the proof. First, observe that by definition of $\mathrm{DIS}_t$ for $t \leq T$, we have that $G_i^\pm \cap \mathrm{DIS}_T \subseteq G_i^\pm \cap \mathrm{DIS}_{T-1} \subseteq \cdots \subseteq G_i^\pm \cap \mathrm{DIS}_1$ for each $1 \leq i \leq 2^{9m-1}$. In addition,

$$|G_i^\pm \cap \mathrm{DIS}_t| \geq \varepsilon' |G_i^\pm \cap \mathrm{DIS}_{t-1}| \text{ for each } 1 \leq i \leq 2^{9m-1}.$$

By the random process generating $h^*$, we also know that $|G_i^\pm \cap \mathrm{DIS}_1| \geq \varepsilon' 2^m$. Combined with the above, this implies that:

$$|G_i^\pm \cap \mathrm{DIS}_T| \geq {\varepsilon'}^T 2^m \text{ for each } 1 \leq i \leq 2^{9m-1}.$$

So, for $T \leq \frac{\log 2^m}{\log(2/\varepsilon)}$, we are guaranteed that $|G_i^\pm \cap \mathrm{DIS}_T| \geq 1$ for each $1 \leq i \leq 2^{9m-1}$.

We now describe the construction of a PAC learner $\mathbb{A}$ with $\mathrm{vc}(\mathbb{A}) = 1$ for the randomly generated concept $h^*$ above; we assume that $\mathbb{A}$ knows $\mathcal{C}$ (but of course, $\mathbb{B}$ does not know $\mathcal{C}$).

---

**Algorithm 6.2:** Non-Robust PAC Learner $\mathbb{A}$

**Input:** Distribution $P$ over $\mathcal{X}$.
**Output:** $h_s$ for the *smallest* $s \in [T]$ with $\mathrm{err}_P(h_s, h^*) \leq \varepsilon$ (or outputting $h_{T+1} = h^*$ if no such $s$ exists).

---

First, we will show that $\mathrm{vc}(\mathbb{A}) = 1$. By definition of $\mathbb{A}$, it suffices to show that $\mathrm{vc}(\mathcal{C}) = \mathrm{vc}(\{h^*, h_1, \ldots, h_T\}) = 1$. By definition of $h^*$ and $h_1$, it is easy to see that there is a $z \in Z$ where $h^*(z) \neq h_1(z)$, and thus $\mathrm{vc}(\mathcal{C}) \geq 1$. Observe that by construction, each predictor in $h_1, \ldots, h_T$ operates as a threshold in each group $G_1^+, G_1^-, \ldots, G_{2^{9m-1}}^+, G_{2^{9m-1}}^-$ (ordered according to the order in which the labels are flipped in the $h_1, \ldots, h_T$ sequence). As a result, each $x \in \mathcal{X}$ has its label flipped at most once in the sequence $(h_1(x), \ldots, h_T(x), h^*(x))$. This is because once the ground-truth label of $x$, $h^*(x)$, is revealed by some $h_t$ (i.e., $h_t(x) = h^*(x)$), all subsequent predictors $h_{t'}$ satisfy $h_{t'}(x) = h^*(x)$. Thus, for any two points $z, z' \in \mathcal{X}$, the number of possible behaviors $\left|\{(h(z), h(z')) : h \in \mathcal{C}\}\right| \leq 3$. Therefore, $\mathcal{C}$ cannot shatter two points. This proves that $\text{vc}(\mathcal{C}) \leq 1$.

Analysis Suppose that we run the reduction algorithm $\mathbb{B}$ with non-robust learner $\mathbb{A}$ for $T$ rounds to obtain predictors $h_{s_1} = \mathbb{A}(P_1), \ldots, h_{s_T} = \mathbb{A}(P_T)$. We will show that $\Pr_{h^*}[s_T \leq T|S] > 0$, meaning that with non-zero probability learner $\mathbb{A}$ will not reveal the ground-truth hypothesis $h^*$. For $t \leq T$, let $E_t$ denote the event that $\text{err}_{P_t}(h_{s_{t-1}+1}, h^*) \leq \varepsilon$. When conditioning on $S, s_1, \ldots, s_{t-1}$, observe that by construction of the randomized hypothesis class $\mathcal{C}$, for each $i \leq 2^{9m-1}$ such that $\{(x_i^-, -1), (x_i^+, +1)\} \cap S = \emptyset$, and each $z \in G_i^{\pm} \cap \text{DIS}_{s_{t-1}} : \Pr_{h^*}\left[h^*(z) \neq h_{s_{t-1}+1}(z)|S, s_1, \ldots, s_{t-1}\right] \leq \varepsilon' = \varepsilon/2$. It follows then by the law of total probability that for any distribution $P_t$ constructed by $\mathbb{A}$:

$$\mathop{\mathbb{E}}_{h^*}\left[\text{err}_{P_t}(h_{s_{t-1}+1}, h^*)|S, s_1, \ldots, s_{t-1}\right] \leq \frac{\varepsilon}{2}.$$

By Markov's inequality, it follows that

$$\begin{aligned}\Pr_{h^*}\left[\bar{E}_t|S, s_1, \ldots, s_{t-1}\right] &= \Pr_{h^*}\left[\text{err}_{P_t}(h_{s_{t-1}+1}, h^*) > \varepsilon|S, s_1, \ldots, s_{t-1}\right] \\ &\leq \frac{\mathbb{E}_{h^*}\left[\text{err}_{P_t}(h_{s_{t-1}+1}, h^*)|S, s_1, \ldots, s_{t-1}\right]}{\varepsilon} \leq \frac{1}{2}.\end{aligned}$$

By law of total probability,

$$\Pr_{h^*}[s_T \leq T|S] \geq \Pr_{h^*}[E_1|S] \times \Pr_{h^*}[E_2|S, E_1] \times \cdots \times \Pr_{h^*}[E_T|S, E_1, \ldots, E_{T-1}] \geq \left(\frac{1}{2}\right)^T > 0.$$

To conclude the proof, we will show that if the reduction algorithm $\mathbb{B}$ sees at most $1/2$ of the support of distribution $D_{h^*}$ through a training set $S$ and makes only $T \leq \frac{\log 2^m}{\log(2/\varepsilon)}$ oracle calls to $\mathbb{A}$, then it will likely fail in robustly learning $h^*$. For each $i \leq 2^{9m-1}$, conditioned on the event that $\{(x_i^-, -1), (x_i^+, +1)\} \cap S = \emptyset$, and conditioned on $h_{s_1}, \ldots, h_{s_T}$, there is a $z \in Z$ that is equally likely to be in $\mathcal{U}(x_i^-)$ or $\mathcal{U}(x_i^+)$. To see why such a point exists, we first describe an equivalent distribution generating $h^*, h_1, \ldots, h_T$. For each $i \leq 2^{9m-1}$ randomly select a $2\varepsilon'$ fraction of points from $G_i^+$ and a $2\varepsilon'$ fraction of points from $G_i^-$. Then, randomly pair the points in each $2\varepsilon'$ fraction to get $\varepsilon' 2^m$ pairs $z_i, z_i'$ for each $G_i^{\pm}$. For each pair $z_i, z_i'$ flip a fair coin $c_i$: if $c_i = 1$, $z_i$'s label gets flipped and otherwise if $c_i = 0$ then $z_i'$'s label gets flipped. This is equivalent to generating $h^*$ by flipping the labels of a uniform randomly-selected $\varepsilon$ fraction of points in each $G_i^{\pm}$ as originally described, but is helpful book-keeping that simplifies our analysis. In addition, $h_1, \ldots, h_T$ can be generated in a similar fashion. Since $T \leq \frac{\log 2^m}{\log(2/\varepsilon)}$, we are guaranteed that $|G_i^{\pm} \cap \mathrm{DIS}_{s_T}| \geq 1$. By definition of $\mathrm{DIS}_{s_T}$, this implies that that there is a pair of points $z_i, z_i'$ in each $G_i^{\pm}$ where each $h_{s_t}(z_i) = h_{s_t}(z_i')$ for $t \leq T$ but $h^*(z_i) \neq h^*(z_i')$ (i.e., each $h_{s_t}$ never reveals the ground-truth label for at least one pair). And then in the end, if $\{(x_i^-, -1), (x_i^+, +1)\} \cap S = \emptyset$, $\mathbb{B}$ will make some prediction on $z_i$, and the posterior probability of it being wrong is $1/2$. More formally, for any training dataset $S \sim D_{h^*}^{|S|}$ where $|S| \leq 2^{9m-3}$, any $h_{s_1}, \ldots, h_{s_T}$ returned by $\mathbb{A}$ where $T \leq \frac{\log 2^m}{\log(2/\varepsilon)}$, and any predictor $f : \mathcal{X} \to \mathcal{Y}$ that is picked by $\mathbb{B}$:

$$\mathop{\mathbb{E}}_{h^*}\left[\mathrm{R}_{\mathcal{U}}(f; D_{h^*}) | S, h_{s_1}, \ldots, h_{s_T}\right] \geq \mathop{\mathbb{E}}_{h^*}\left[\frac{1}{2^{9m}} \sum_{\substack{(x,y)\notin S,\\ (x,y)\in \mathrm{supp}(D_{h^*})}} \sup_{z\in\mathcal{U}(x)} 1[f(z) \neq y] \middle| S, h_{s_1}, \ldots, h_{s_T}\right]$$

$$= \frac{1}{2^{9m}} \sum_{i=1}^{2^{9m-1}} \Pr_{h^*}\Big[\big((x_i^+, +1), (x_i^-, -1) \notin S\big) \;\wedge$$

$$\big(\exists z \in \mathcal{U}(x_i^+) \text{ s.t. } f(z) \neq +1 \vee \exists z \in \mathcal{U}(x_i^-) \text{ s.t. } f(z) \neq -1\big) \Big| S, h_{s_1}, \ldots, h_{s_T}\Big]$$

$$\geq \frac{2^{9m-1}}{2^{9m}} \frac{1}{2} = \frac{1}{4}.$$

This implies that, for any $\mathbb{B}$ limited to $n \leq 2^{9m-3}$ training examples and $T \leq \frac{m}{\log_2(2/\varepsilon)}$ queries, there exists a *deterministic* choice of $h^*$ and $h_1, \ldots, h_T$, and a corresponding learner $\mathbb{A}$ that is a PAC learner for $\{h^*\}$ using hypothesis class $\{h^*, h_1, \ldots, h_T\}$ of VC dimension 1, such that, for $S \sim D_{h^*}^n$, $\mathbb{E}_S[\mathrm{R}_{\mathcal{U}}(f; D_{h^*})] \geq \frac{1}{4}$. $\square$

## 6.5 Agnostic Guarantees

Thus far, in Section 6.3, we focused only on robust PAC learning in the realizable setting, where we assume there is a $h \in \mathcal{H}$ with zero robust error, i.e., when $\mathsf{OPT}_{\mathcal{H}} := \min_{h\in\mathcal{H}} \mathrm{R}_{\mathcal{U}}(h;\mathcal{D}) = 0$. It would be desirable to extend our results also to the agnostic setting, when $\mathsf{OPT}_{\mathcal{H}} > 0$ and we want to compete with $\mathsf{OPT}_{\mathcal{H}}$. We next present a different reduction algorithm with guarantees for the agnostic setting. This reduction algorithm specifically requires an $\mathsf{ERM}_{\mathcal{H}}$ oracle (and not just a PAC learner for $\mathcal{H}$). The algorithm is due to Feige, Mansour, and Schapire (2015), but the analysis and application are novel in this chapter.

**Algorithm 6.3:** Feige, Mansour, and Schapire (2015)

**Input:** weight update parameter $\eta > 0$, number of rounds $T$, training dataset $S = \{(x_1, y_1), \dots, (x_m, y_m)\}$, and $\mathsf{ERM}_{\mathcal{H}}$ oracle.

Set $w_1(z,(x,y)) = 1$ for each $(x,y) \in S, z \in \mathcal{U}(z)$.

Set $P^1(z,(x,y)) = \frac{w_1(z,(x,y))}{\sum_{z' \in \mathcal{U}(x)} w_1(z',(x,y))}$.

**for** *each* $t \leftarrow 1$ **to** $T$ **do**

  Call $\mathsf{ERM}_{\mathcal{H}}$ on the empirical weighted distribution:

  $h_t = \mathrm{argmin}_{h\in\mathcal{H}} \frac{1}{m} \sum_{(x,y)\in S} \sum_{z\in\mathcal{U}(x)} P^t(z,(x,y)) \mathbb{1}[h_t(z) \neq y]$.

  **for** *each* $(x,y) \in S$ *and* $z \in \mathcal{U}(x)$ **do**

    $w_{t+1}(z,(x,y)) = (1 + \eta \mathbb{1}[h_t(z) \neq y]) \cdot w_t(z,(x,y))$.

    $P^{t+1}(z,(x,y)) = \frac{w_t(z,(x,y))}{\sum_{z' \in \mathcal{U}(x)} w_t(z',(x,y))}$.

**Output:** The majority-vote predictor $\mathrm{MAJ}(h_1, \dots, h_T)$.

**Theorem 6.7.** *Consider an arbitrary perturbation set* $\mathcal{U}$*, target class* $\mathcal{H}$*, and* $\mathsf{ERM}_{\mathcal{H}}$ *oracle. Set* $T(\varepsilon) = \frac{32 \ln|\mathcal{U}|}{\varepsilon^2}$ *and* $m(\varepsilon,\delta) = O\left(\frac{\mathrm{vc}(\mathcal{H})(\ln|\mathcal{U}|)^2}{\varepsilon^4} \ln\left(\frac{\ln|\mathcal{U}|}{\varepsilon^2}\right) + \frac{\ln(1/\delta)}{\varepsilon^2}\right)$*. Then, for any distribution* $\mathcal{D}$ *over* $\mathcal{X} \times \mathcal{Y}$*, with probability at least* $1-\delta$ *over* $S \sim \mathcal{D}^{m(\varepsilon,\delta)}$*, running Algorithm 6.3 for* $T(\varepsilon)$ *rounds produces* $h_1, \dots, h_{T(\varepsilon)} \in \mathcal{H}$ *satisfying:*

$$\mathrm{R}_{\mathcal{U}}\left(\mathrm{MAJ}(h_1, \dots, h_{T(\varepsilon)}); \mathcal{D}\right) \leq 2\mathsf{OPT}_{\mathcal{H}} + \varepsilon.$$

Comparison with prior related work  Feige, Mansour, and Schapire (2015) only considered *finite* hypothesis classes $\mathcal{H}$ and provided generalization guarantees depending on $\log|\mathcal{H}|$, while we consider here infinite classes $\mathcal{H}$ with bounded VC dimension and provide tighter robust generalization bounds. The robust learning guarantee due to Attias, Kontorovich, and Mansour (2022, Theorem 2) assumes access to a $\mathsf{RERM}_{\mathcal{H}}$ oracle, which minimizes the robust loss on the training dataset and is not the same as a standard $\mathsf{ERM}_{\mathcal{H}}$ oracle. On the other hand, Theorem 6.7 and Algorithm 6.3 above can be interpreted as a reduction algorithm to implement $\mathsf{RERM}_{\mathcal{H}}$ using only an $\mathsf{ERM}_{\mathcal{H}}$ oracle in the challenging *non-realizable* setting. Finally, we remark that while the sample complexity dependence on $\mathrm{poly}(\ln|\mathcal{U}|)$ in Theorem 6.7 may be benign, it is an open question to improve this result and remove dependence on $|\mathcal{U}|$ (as was possible in the realizable setting in Section 6.3).

Before proceeding with the proof of Theorem 6.7, we describe at a high-level the proof strategy. The main insight is to solve a finite zero-sum game. In particular, our goal is to find a mixed-strategy over the hypothesis class that is approximately close to the value of the game:

$$\mathsf{OPT}_{S,\mathcal{H}} \triangleq \min_{h\in\mathcal{H}} \frac{1}{|S|} \sum_{(x,y)\in S} \sup_{z\in\mathcal{U}(x)} 1\left[h(z) \neq y\right].$$

We observe that Algorithm 6.3 due to (Feige et al., 2015) solves a similar finite zero-sum game (see Lemma 6.9), and then we relate it to the value of the game we are interested in (see Lemma 6.8). Combined together, this only establishes that we can minimize the robust loss on the empirical dataset using an $\mathsf{ERM}_{\mathcal{H}}$ oracle. We then appeal to uniform convergence guarantees for the robust loss (see Lemma 2.7) to show that, with large enough training data, our output predictor achieves robust risk that is close to the value of the game.

For any set arbitrary set $W$, we denote by $\Delta(W)$ the set of distributions over $W$. We use this notation below.

**Lemma 6.8.** *For any dataset $S = \{(x_1, y_1), \ldots, (x_m, y_m)\} \in (\mathcal{X} \times \mathcal{Y})^m$,*

$$\mathsf{OPT}_{S,\mathcal{H}} = \min_{h\in\mathcal{H}} \frac{1}{m} \sum_{i=1}^{m} \max_{z_i\in\mathcal{U}(x_i)} 1\left[h(z_i) \neq y_i\right] \geq \min_{Q\in\Delta(\mathcal{H})} \max_{\substack{P_1\in\Delta(\mathcal{U}(x_1)),\\ \ddots \\ P_m\in\Delta(\mathcal{U}(x_m))}} \frac{1}{m} \sum_{i=1}^{m} \mathop{\mathbb{E}}_{z_i\sim P_i} \mathop{\mathbb{E}}_{h\sim Q} 1\left[h(z_i) \neq y_i\right]$$

*Proof.* By definition of $\mathsf{OPT}_{S,\mathcal{H}}$, it follows that

$$\begin{aligned}
\mathsf{OPT}_{S,\mathcal{H}} &= \min_{h\in\mathcal{H}} \frac{1}{m}\sum_{i=1}^{m} \max_{z_i\in\mathcal{U}(x_i)} 1\left[h(z_i)\neq y_i\right] \\
&\geq \min_{h\in\mathcal{H}} \max_{z_1\in\mathcal{U}(x_1),\dots,z_m\in\mathcal{U}(x_m)} \frac{1}{m}\sum_{i=1}^{m} 1\left[h(z_i)\neq y_i\right] \\
&\geq \min_{Q\in\Delta(H)} \max_{z_1\in\mathcal{U}(x_1),\dots,z_m\in\mathcal{U}(x_m)} \frac{1}{m}\sum_{i=1}^{m} \mathop{\mathbb{E}}_{h\sim Q} 1\left[h(z_i)\neq y_i\right] \\
&\geq \min_{Q\in\Delta(\mathcal{H})} \max_{P_1\in\Delta(\mathcal{U}(x_1)),\dots,P_m\in\Delta(\mathcal{U}(x_m))} \frac{1}{m}\sum_{i=1}^{m} \mathop{\mathbb{E}}_{z_i\sim P_i}\mathop{\mathbb{E}}_{h\sim Q} 1\left[h(z_i)\neq y_i\right].
\end{aligned}$$

□

**Lemma 6.9** (Feige, Mansour, and Schapire (2015)). *For any data set* $S=\{(x_1,y_1),\dots,(x_m,y_m)\}\in(\mathcal{X}\times\mathcal{Y})^m$*, running Algorithm 6.3 for* $T$ *rounds produces a mixed-strategy* $\hat{Q}=\frac{1}{T}\sum_{t=1}^{T} h_t\in\Delta(\mathcal{H})$ *satisfying:*

$$\max_{\substack{P_1\in\Delta(\mathcal{U}(x_1)),\\ \dots,\\ P_m\in\Delta(\mathcal{U}(x_m))}} \frac{1}{m}\sum_{i=1}^{m}\mathop{\mathbb{E}}_{z_i\sim P_i}\frac{1}{T}\sum_{t=1}^{T} 1\left[h_t(z_i)\neq y_i\right] \leq \min_{Q\in\Delta(\mathcal{H})}\max_{\substack{P_1\in\Delta(\mathcal{U}(x_1)),\\ \dots,\\ P_m\in\Delta(\mathcal{U}(x_m))}} \frac{1}{m}\sum_{i=1}^{m}\mathop{\mathbb{E}}_{z_i\sim P_i}\mathop{\mathbb{E}}_{h\sim Q} 1\left[h(z_i)\neq y_i\right] + 2\sqrt{\frac{\ln|\mathcal{U}|}{T}}.$$

*Proof.* By the minimax theorem and (Feige, Mansour, and Schapire, 2015, Equation 3 and 9 in proof of Theorem 1), we have that

$$\begin{aligned}
&\max_{P_1\in\Delta(\mathcal{U}(x_1)),\dots,P_m\in\Delta(\mathcal{U}(x_m))} \sum_{i=1}^{m}\mathop{\mathbb{E}}_{z_i\sim P_i}\frac{1}{T}\sum_{t=1}^{T} 1\left[h_t(z_i)\neq y_i\right] \leq \\
&\qquad \min_{Q\in\Delta(\mathcal{H})}\max_{P_1\in\Delta(\mathcal{U}(x_1)),\dots,P_m\in\Delta(\mathcal{U}(x_m))} \mathop{\mathbb{E}}_{z_i\sim P_i}\mathop{\mathbb{E}}_{h\sim Q} 1\left[h(z_i)\neq y_i\right] + 2\frac{\sqrt{\mathcal{L}^* m\ln|\mathcal{U}|}}{T},
\end{aligned}$$

where $\mathcal{L}^*=\sum_{i=1}^{m}\max_{z\in\mathcal{U}(x_i)}\sum_{t=1}^{T} 1\left[h_t(z)\neq y\right]$. By observing that $\mathcal{L}^*\leq mT$ and dividing both sides of the inequality above by $m$, we arrive at the inequality stated in the lemma. □

We are now ready to proceed with the proof of Theorem 6.7.

*Proof of Theorem 6.7.* Let $S\sim\mathcal{D}^m$ be an iid sample from $\mathcal{D}$, where the size of the sample $m$

will be determined later. By invoking Lemma 6.9 and Lemma 6.8, we observe that running Algorithm 6.3 on $S$ for $T$ rounds, produces $h_1, \dots, h_T \in \mathcal{H}$ satisfying

$$\max_{\substack{P_1 \in \Delta(\mathcal{U}(x_1)),\\ \dots,\\ P_m \in \Delta(\mathcal{U}(x_m))}} \frac{1}{m}\sum_{i=1}^{m} \mathop{\mathbb{E}}_{z_i \sim P_i} \frac{1}{T}\sum_{t=1}^{T} 1\left[h_t(z_i) \neq y_i\right] \leq \mathsf{OPT}_{S,\mathcal{H}} + \frac{\varepsilon}{4}$$

Next, the average robust loss for the majority-vote predictor $\mathrm{MAJ}(h_1, \dots, h_T)$ can be bounded from above as follows:

$$\frac{1}{m}\sum_{i=1}^{m} \max_{z_i \in \mathcal{U}(x_i)} 1\left[\mathrm{MAJ}(h_1, \dots, h_T)(z_i) \neq y_i\right] \leq \frac{1}{m}\sum_{i=1}^{m} \max_{z_i \in \mathcal{U}(x_i)} 2 \mathop{\mathbb{E}}_{t \sim [T]} 1\left[h_t(z_i) \neq y_i\right]$$
$$= 2\frac{1}{m}\sum_{i=1}^{m} \max_{z_i \in \mathcal{U}(x_i)} \frac{1}{T}\sum_{t=1}^{T} 1\left[h_t(z_i) \neq y_i\right]$$
$$\leq 2 \max_{\substack{P_1 \in \Delta(\mathcal{U}(x_1)),\\ \dots,\\ P_m \in \Delta(\mathcal{U}(x_m))}} \frac{1}{m}\sum_{i=1}^{m} \mathop{\mathbb{E}}_{z_i \sim P_i} \frac{1}{T}\sum_{t=1}^{T} 1\left[h_t(z_i) \neq y_i\right] \leq 2\mathsf{OPT}_{S,\mathcal{H}} + \frac{\varepsilon}{2}.$$

Next, we invoke Lemma 2.7 to obtain a uniform convergence guarantee on the robust loss. In particular, we apply Lemma 2.7 on the *convex-hull* of $\mathcal{H}$: $\mathcal{H}^T = \{\mathrm{MAJ}(h_1, \dots, h_T) : h_1, \dots, h_T \in \mathcal{H}\}$. By a classic result due to Blumer et al. (1989a), it holds that $\mathrm{vc}(\mathcal{H}^T) = O(\mathrm{vc}(\mathcal{H}) T \ln T)$. Combining this with Lemma 2.7 and plugging-in the value of $T = \frac{32 \ln|\mathcal{U}|}{\varepsilon^2}$, we get that the VC dimension of the robust loss class of $\mathcal{H}^T$ is bounded from above by

$$\mathrm{vc}(\mathcal{L}^{\mathcal{U}}_{\mathcal{H}^T}) \leq O\left(\frac{\mathrm{vc}(\mathcal{H})(\ln|\mathcal{U}|)^2}{\varepsilon^2} \ln\left(\frac{\ln|\mathcal{U}|}{\varepsilon^2}\right)\right).$$

Finally, using Vapnik's "General Learning" uniform convergence (Vapnik, 1982), with probability at least $1-\delta$ over $S \sim \mathcal{D}^m$ where $m = O\left(\frac{\mathrm{vc}(\mathcal{H})(\ln|\mathcal{U}|)^2}{\varepsilon^4} \ln\left(\frac{\ln|\mathcal{U}|}{\varepsilon^2}\right) + \frac{\ln(1/\delta)}{\varepsilon^2}\right)$, it holds that

$$\forall f \in \mathcal{H}^T : \mathop{\mathbb{E}}_{(x,y)\sim\mathcal{D}}\left[\max_{z \in \mathcal{U}(x)} 1\left[f(z) \neq y\right]\right] \leq \frac{1}{m}\sum_{i=1}^{m} \max_{z_i \in \mathcal{U}(x_i)} 1\left[f(z_i) \neq y_i\right] + \frac{\varepsilon}{4}$$

This also applies to the particular output $\text{MAJ}(h_1, \ldots, h_T)$ of Algorithm 6.3, and thus

$$\begin{aligned}
&\mathop{\mathbb{E}}_{(x,y)\sim\mathcal{D}}\left[\max_{z\in\mathcal{U}(x)} 1\left[\text{MAJ}(h_1, \ldots, h_{T(\varepsilon)})(z) \neq y\right]\right] \\
&\leq \frac{1}{m}\sum_{i=1}^{m} \max_{z_i\in\mathcal{U}(x_i)} 1\left[\text{MAJ}(h_1, \ldots, h_T)(z_i) \neq y_i\right] + \frac{\varepsilon}{4} \\
&\leq 2\mathsf{OPT}_{S,\mathcal{H}} + \frac{\varepsilon}{2} + \frac{\varepsilon}{4}.
\end{aligned}$$

Finally, by applying a standard Chernoff-Hoeffding concentration inequality, we get that $\mathsf{OPT}_{S,\mathcal{H}} \leq \mathsf{OPT}_{\mathcal{H}} + \frac{\varepsilon}{8}$. Combining this with the above inequality concludes the proof. □

## 6.6 Discussion

Computational Efficiency  Although the sample complexity of Algorithm 6.1 is independent of $|\mathcal{U}|$, we showed that the $\log|\mathcal{U}|$ dependence in oracle complexity is unavoidable. This implies that the runtime of Algorithm 6.1 will be at best weakly polynomial and have at least a $\log|\mathcal{U}|$ dependence. But maybe this is not so bad, because it is equivalent to the number of bits required to represent the adversarial perturbations. This weak poly-time dependence is common in almost all optimization algorithms (gradient descent, interior point methods, etc). What is more concerning is the linear runtime and memory dependence on $|\mathcal{U}|$ that emerges from the explicit representation of the adversarial perturbations during training. In practice, many of the adversarial perturbations $\mathcal{U}$ are infinite, but specified implicitly, and not by enumerating over all possible perturbations (e.g. $\ell_p$ perturbations). This motivates the following next steps: What operations do we need to be able to implement efficiently on $\mathcal{U}$ in order to robustly learn? What access (oracle calls, or "interface") do we need to $\mathcal{U}$?

Sampling Oracle Over Perturbations  A reasonable form of access to $\mathcal{U}$ that is sufficient for implementing Algorithm 6.1 is a sampling oracle that takes as input a point $x$ and an energy function $E : \mathcal{X} \to \mathbb{R}$, and does the following:

1. Samples a perturbation $z$ from a distribution given by $p_x(z) \propto \exp(E(z)) * 1[z \in \mathcal{U}(x)]$. That is, the oracle samples from the set $\mathcal{U}(x)$ based on the weighting encoded in $E$.

2. Calculates $\Pr[z \in \mathcal{U}(x)]$ for the distribution given by $p(z) \propto \exp(E(z))$.

With such an oracle, Algorithm 6.1 can be implemented without the need to do explicit inflation of $S$ to $S_{\mathcal{U}}$, and can avoid the linear dependence on $|\mathcal{U}|$. This is because Algorithm 6.1 and its subprocedure `ZeroRobustLoss` just need to sample from distributions over the inflated set $S_{\mathcal{U}}$ that are constructed by $\alpha$-Boost (as required in Steps 5 and 17 in Algorithm 6.1). This can be simulated via a two-stage process where we maintain a conditional distribution over $S$ (the original points), and then draw perturbations using the sampling oracle. Specifically, to sample from a distribution $D_t$ that is constructed by $\alpha$-Boost, we use two energy functions $E_t^+(z) = -2\alpha \sum_{i \leq t} 1[g_t(z) = +1]$ and $E_t^-(z) = -2\alpha \sum_{i \leq t} 1[g_t(z) = -1]$, where $g_1, \dots, g_t$ represent the sequence of predictors produced during the first $t$ rounds of boosting (either $h_t$'s produced by non-robust learner $\mathbb{A}$ or $f_t$'s produced by `ZeroRobustLoss`). Using the sampling oracle, we can sample from $D_t$, by first sampling $(x, y)$ from $S$ based on the marginal estimates computed by the oracle (operation (b) described above) using energy function $E_t^y$, and then sampling $z$'s from their $\mathcal{U}(x)$ (operation (a) described above) using energy function $E_t^y$.

# 7

# Robustness to Unknown Perturbations

## 7.1 Introduction

Almost all prior work on adversarial robustness starts with specifying a perturbation set $\mathcal{U}$ we would like to be robust against. The type of perturbation sets we are truly interested in are often sets $\mathcal{U}$ that capture "natural" or "imperecptible" perturbations. But partially because of the need to specify $\mathcal{U}$ explicitly during training, simpler sets are often used, such as $\ell_p$-norm balls (Goodfellow, Shlens, and Szegedy, 2015a), or orbits with respect to translations and rotations (Engstrom, Tran, Tsipras, Schmidt, and Madry, 2019). Furthermore, training procedures are often specific to the perturbation set $\mathcal{U}$, or have the perturbation set "hard coded" inside them. Some methods rely on predictor implementations that need to "know" the specific perturbation set $\mathcal{U}$ at test-time (e.g., randomized smoothing Lécuyer, Atlidakis, Geambasu, Hsu, and Jana, 2019, Cohen, Rosenfeld, and Kolter, 2019, Salman, Li, Razenshteyn, Zhang, Zhang, Bubeck, and Yang, 2019), and some methods use "explicit" knowledge of $\mathcal{U}$ only during training-time (e.g., Wong and Kolter, 2018, Raghunathan, Steinhardt, and Liang, 2018a,b, Montasser, Hanneke, and Srebro, 2019).

*Can we design robust learning algorithms that do not require explicit knowledge of the adversarial perturbations $\mathcal{U}$? What reasonable models of access to, or interactions with, $\mathcal{U}$ could we rely on instead?*

| | Sample Complexity | Oracle Complexity | |
|---|---|---|---|
| Realizable | $\tilde{O}(\mathrm{lit}(\mathcal{H}))$ | $\tilde{O}(\mathrm{lit}(\mathcal{H}))$ | Theorem 7.4. |
| | $\tilde{O}(\mathrm{vc}(\mathcal{H})\mathrm{vc}^{*2}(\mathcal{H}))$ | $2^{\tilde{O}(\mathrm{vc}^2(\mathcal{H})\mathrm{vc}^{*2}(\mathcal{H}))}\mathrm{lit}(\mathcal{H})$ | Theorem 7.6. |
| | | $\Omega(\mathrm{lit}(\mathcal{H}))$ | Theorem 7.10. |
| Agnostic | $\tilde{O}(\mathrm{lit}(\mathcal{H}))$ | $\tilde{O}(\mathrm{lit}^2(\mathcal{H}))$ | Theorem 7.7. |
| | $\tilde{O}(\mathrm{vc}(\mathcal{H})\mathrm{vc}^{*2}(\mathcal{H}))$ | $2^{\tilde{O}(\mathrm{vc}^2(\mathcal{H})\mathrm{vc}^{*2}(\mathcal{H}))}\mathrm{lit}(\mathcal{H})$ | Theorem 7.8. |

**Table 7.1:** We show that a hypothesis class $\mathcal{H}$ is robustly learnable in the Perfect Attack Oracle model if and only if $\mathcal{H}$ is *online* learnable. We give upperbounds (corresponding to algorithms) in the realizable setting (Section 7.3.1) and the agnostic setting (Section 7.3.2), and lower bounds on the oracle complexity in the realizable setting (Section 7.3.3). Futhermore, our results show that sophisticated algorithms that leverage online learners can be favorable to more traditional online-to-batch conversion schemes in terms of their robust generalization guarantees. The $\tilde{O}$ notation hides logarithmic factors and dependence on error $\varepsilon$ and failure probability $\delta$, $\mathrm{vc}(\mathcal{H})$ and $\mathrm{vc}^*(\mathcal{H})$ denote the primal and dual VC dimension of $\mathcal{H}$, and $\mathrm{lit}(\mathcal{H})$ denotes the Littlestone dimension of $\mathcal{H}$.

In this chapter, we ask whether it is possible to develop generic learning algorithms with robustness guarantees, without knowing the perturbation set $\mathcal{U}$ a-priori. That is, we want to design general robust algorithmic frameworks that work for any perturbation set $\mathcal{U}$, given a *reasonable* form of access to $\mathcal{U}$, and avoid algorithms tailored to a specific $\mathcal{U}$ such as $\ell_\infty$ or $\ell_2$ perturbations. This is important if we want to be able to easily adapt our training procedures to different perturbation sets, or would like to build ML systems that are robust to fairly abstract perturbation sets $\mathcal{U}$ such as "images that are indistinguishable to the human eye" (see e.g., Laidlaw, Singla, and Feizi, 2020). In our frameworks, instead of redesigning or reprogramming the training algorithm, one would only need to implement or provide specific "attack procedures" for $\mathcal{U}$.

In this chapter, we consider robustly learning a hypothesis class $\mathcal{H} \subseteq \mathcal{Y}^{\mathcal{X}}$ (e.g., neural networks). The learning algorithm receives as input $m$ iid samples $S = \{(x_i, y_i)\}_{i=1}^m$ drawn from an unknown distribution $\mathcal{D}$ over $\mathcal{X} \times \mathcal{Y}$. A predictor $h : \mathcal{X} \to \mathcal{Y}$ is robustly correct on an example $(x, y)$ w.r.t. $\mathcal{U}$ if $\sup_{z\in\mathcal{U}(x)} 1[h(z) \neq y] = 0$. The learning algorithm has no *explicit* knowledge of $\mathcal{U}$, but instead is allowed the following forms of access:

Access to a (perfect) adversarial attack oracle In this model, the learning algorithm has access to a "mistake oracle", which we can also think of as a perfect attack oracle for $\mathcal{U}$. A perfect attack oracle for $\mathcal{U}$ receives as input a predictor $g : \mathcal{X} \to \mathcal{Y}$ and a labeled example $(x, y)$, and is asked to either assert that $g$ is robustly correct on $(x, y)$, or return a perturbation $z \in \mathcal{U}(x)$ that is miss-classified (see Definition 7.3). The learning algorithm can query the perfect attack oracle for $\mathcal{U}$ by calling it $T$ times with queries of the form: $(g_t, (x'_t, y'_t))$, where $g_t$ is a predictor and $(x'_t, y'_t)$ is a labeled example (not necessarily from the training set $S$). The goal of the learning algorithm is to output a predictor $\hat{h}$ with small *robust* risk (see Equation 2.1).

In Section 7.3, we present algorithms, guarantees on the required sample complexity and number of oracle accesses, and lower bounds on the required number of accesses, for robustly learning $\mathcal{H}$ in the Perfect Attack Oracle model. These results are summarized in Table 7.1.

To program such a perfect attack oracle, $\mathcal{U}$ still has to be specified inside it. And even for simple $\mathcal{U}$, a perfect attack oracle is generally intractable. Furthermore, practical attack engines used in training (e.g., PGD Madry, Makelov, Schmidt, Tsipras, and Vladu, 2018) are not perfect, and are not always guaranteed to find miss-classified adversarial perturbations even when they do exist. Can we still provide meaningful robustness guarantees if we only have access to an *imperfect* attack oracle?

Access to an (imperfect) adversarial attack oracle In this model, the learning algorithm has access to a possibly imperfect attacking algorithm $\mathbb{A}$ for $\mathcal{U}$. The learning algorithm can query $\mathbb{A}$ by calling it $T$ times with queries of the form: $(g_t, (x'_t, y'_t))$, where $g_t : \mathcal{X} \to \mathcal{Y}$ is a predictor and $(x'_t, y'_t)$ is a labeled example. The goal of the learning algorithm is to output a predictor $\hat{h}$ with small error w.r.t. future attacks from $\mathbb{A}$,

$$\mathrm{err}_{\mathbb{A}}(\hat{h}; \mathcal{D}) \triangleq \Pr_{\substack{(x,y)\sim\mathcal{D} \\ \text{randomness of } \mathbb{A}}} \left[ \hat{h}(\mathbb{A}(\hat{h}, (x, y))) \neq y \right]. \tag{7.1}$$

In Section 7.5, we give an algorithm with sample and oracle complexity of $\tilde{O}(\mathrm{lit}(\mathcal{H}))$ that guarantees small $\mathrm{err}_{\mathbb{A}}$ when then attacker $\mathbb{A}$ is "stationary", i.e., $\mathbb{A}$ doesn't learn or adapt over time.

But what happens if the adversary $\mathbb{A}$ changes over time? In the above model, the predictor $\hat{h}$ is fixed after training, and thus if the adversary $\mathbb{A}$ changes, e.g., by adapting to the returned predictor, or perhaps if we encounter an altogether different adversary than the one we accessed during training, this might result in a much higher error rate. Is it possible to continually adapt to changing adversaries in a meaningful way, ensuring strong robustness guarantees?

Interaction with an actual attacker In this online model, the learning algorithm $\mathbb{B}$ can monitor the behaviour of an actual attacker $\mathbb{A}$ and adapt accordingly. The attacker knows the current predictor $h$ used, as well as the perturbation set $\mathcal{U}$, and attempts attacks on an iid stream of samples $(x_t, y_t)$. Whenever the attacker succeeds in finding a perturbation $z_t \in \mathcal{U}(x_t)$ s.t. $h(x_t) \neq y_t$, it scores a "successful attack", but the perturbation $z_t$ is revealed to learner $\mathbb{B}$, who can also obtain the true label $y_t$, and learner $\mathbb{B}$ can update its predictor. The goal of the learner is to bound the total number of successful attacks.

Monitoring and adapting to an attacker might sometimes be possible and appropriate, e.g., when attacks to predictors can be detected in hindsight and when the predictor is running on the cloud or when predictor updates can be pushed to devices, which is becoming increasingly common. But beyond such scenarios, this online model is also useful as an analysis tool of the imperfect attack oracle model above, and our methods for the imperfect attack oracle model are based on this online model.

In Section 7.4, we show upper bounds and lower bounds on the the number of successful attacks in terms of the Littlestone dimension $\text{lit}(\mathcal{H})$, although our results leave open a possible exponential gap in the bound on the number of successful attacks (in a setting where the learner has access to infinitely many uncorrupted samples, i.e. knows the uncorrupted source distribution).

Practical Relevance Our goal is to understand how, from a theoretical perspective, it is possible to depart from assuming full and explicit knowledge of the perturbation set, and what types of other accesses and interactions could still enable adversarially robust learning. We obviously need some dependence on the perturbation set or possible attacks during training, and we are making the first steps in establishing what forms of access (beyond explicit exact knowledge) could be sufficient, and how they could be used, and what are the limits (lower bounds) on what might be possible using different access models. Some of the models above already capture approaches used in practice. But perhaps more importantly, we hope this study will lead to interest in defining "better" access models, and finding the "right" framework for adversarially robust learning, perhaps, by way of analogy, similar to how early studies of privacy in data analysis struggled with finding the "right" attack models and definitions.

Related Work Most prior work on adversarial robustness has focused on methods that are tailored to specific perturbation sets $\mathcal{U}$. For example, in randomized smoothing (Lécuyer, Atlidakis, Geambasu, Hsu, and Jana, 2019, Cohen, Rosenfeld, and Kolter, 2019, Salman, Li, Razenshteyn, Zhang, Zhang, Bubeck, and Yang, 2019), computing a prediction on a test-point $x$ requires sampling perturbations $z$ from a distribution $P$ over $\mathcal{U}(x)$, and returning the most likely prediction given by a learned predictor $f : \mathcal{X} \to \mathcal{Y}$. Distribution $P$ is chosen based on $\mathcal{U}$, for example, if $\mathcal{U}$ is $\ell_2$ perturbations, then $P$ is an isotropic Gaussian distribution $\mathcal{N}(x, \sigma^2 I)$. In addition, there are algorithms (certified defenses) that minimize some surrogate loss $\ell_{\mathcal{U}}$ where the construction of $\ell_{\mathcal{U}}$ depends on $\mathcal{U}$ (e.g., Wong and Kolter, 2018, Raghunathan, Steinhardt, and Liang, 2018a,b).

The adversarial training framework (Goodfellow, Shlens, and Szegedy, 2015a, Madry, Makelov, Schmidt, Tsipras, and Vladu, 2018) does not use explicit knowledge of $\mathcal{U}$, but only uses an attacking algorithm (e.g., FGSM or PGD) implemented for the perturbation set $\mathcal{U}$. However, no formal guarantees are known about adversarial training in terms of robust generalization. Specifically, it is not known whether adversarial training will yield predictors that generalize to future adversarial perturbations from $\mathcal{U}$, or even generalize to specific perturbations chosen by PGD or FGSM. It has been observed that common forms of adversarial training on deep neural nets do not generalize to future attacks from PGD (Schmidt, Santurkar, Tsipras, Talwar, and Madry, 2018). Our work can be seen as a theoretical study of such generic approaches, which leads to different, and considerably more sophisticated methods (yet at this stage, perhaps not easily implementable).

Towards our quest in this chapter for finding the right form of access to $\mathcal{U}$, we build on our algorithms in Chapter 3 and Chapter 6 by re-interpreting them in light of the questions we study in this chapter, but also extending them significantly in the following ways: we avoid using a *robust* empirical risk minimization $\mathsf{RERM}_{\mathcal{U}}$ oracle that requires explicit knowledge of $\mathcal{U}$ as used in Chapter 3 and use an online learning algorithm instead, and we carry-out a technical inflation procedure of the training sample to include perturbations by utilizing a perfect attack oracle for $\mathcal{U}$ without explicit knowledge of $\mathcal{U}$ as was done in Chapter 6. Furthermore, Chapter 6 considered robustly PAC learning $\mathcal{H}$ using only black-box access to a non-robust

PAC learner for $\mathcal{H}$ but allowed explicit knowledge of $\mathcal{U}$, their reduction makes oracle calls that depend on the bit complexity $\log|\mathcal{U}|$, and they show this is unavoidable. In this chapter, our algorithms can be viewed as black-box reductions that use an *online* learner for $\mathcal{H}$ (instead of just a PAC learner), furthermore, they do not require a explicit knowledge of $\mathcal{U}$ but only an attack oracle for $\mathcal{U}$. Our algorithms achieve the same sample complexity bound, but with number of calls to the online learner that is independent of $\log|\mathcal{U}|$ and only depends on the VC dimension $\text{vc}(\mathcal{H})$.

Ashtiani, Pathak, and Urner (2020) considered a weaker form of attacking algorithms—those that receive as input a black-box predictor—in a $\mathcal{U}$-specific learning framework and showed an upper bound of $\tilde{O}(\text{vc}(\mathcal{H}))$ on the sample complexity of robust PAC learning when the pair $(\mathcal{H}, \mathcal{U})$ admits a query efficient attacking algorithm. Their learning algorithm relies on a *robust* empirical risk minimization $\mathsf{RERM}_{\mathcal{U}}$ oracle that requires explicit knowledge of $\mathcal{U}$. In this chapter, we focus on modularity and avoid using a $\mathsf{RERM}$ oracle for $\mathcal{H}$, and use a black-box online learner $\mathbb{A}$ for $\mathcal{H}$.

Goldwasser, Kalai, Kalai, and Montasser (2020) considered classifying *arbitrary* test examples in a transductive selective classification setting. They gave an algorithm that takes as input: (a) training examples from a distribution $P$ over $\mathcal{X}$ labeled with some unknown function $h^*$ in a class $\mathcal{H}$ with finite VC dimension, and (b) a batch of arbitrary unlabeled test examples (possibly chosen by an unknown adversary), and outputs a selective predictor $\hat{f}$—which abstains from predicting on some examples—that has a low rejection rate w.r.t. $P$, and low error rate on the test examples. Selective predictor $\hat{f}$, however, can potentially abstain from classifying most test examples if they are adversarial. In this chapter, we consider classifying test examples $x \sim P$ or adversarial perturbations $z \in \mathcal{U}(x)$ where the perturbation set $\mathcal{U}$ is unknown, and output predictors that do not abstain but always provide a classification with low error rate. We do not require unlabeled test examples, but require black-box access to an attack oracle for $\mathcal{U}$.

## 7.2 Preliminaries

Let $\mathcal{X}$ denote the instance space, $\mathcal{Y} = \{\pm 1\}$ denote the label space, and $\mathcal{H} \subseteq \mathcal{Y}^{\mathcal{X}}$ denote a hypothesis class. We refer the reader to Chapter 2 for formal definitions of: adversarially robust PAC learning (Definition 2.2), VC dimension (Definition 2.5), and the dual VC di-mension (Definition 2.6); which we will use throughout this chapter.

Online Learnability and Littlestone Dimension  An online learning algorithm $\mathbb{A}$ is a (measurable) map $(\mathcal{X} \times \mathcal{Y})^* \to \mathcal{Y}^{\mathcal{X}}$. For a class $\mathcal{H} \subseteq \mathcal{Y}^{\mathcal{X}}$, the mistake bound of $\mathbb{A}$ is the maximum possible number of mistakes algorithm $\mathbb{A}$ makes on any sequence of examples labeled with some $h \in \mathcal{H}$:

$$M(\mathbb{A}, \mathcal{H}) := \sup_{x_1, x_2, \dots \in \mathcal{X}} \sup_{h \in \mathcal{H}} \sum_{t=1}^{\infty} 1\left[\mathbb{A}(\{(x_i, h(x_i))\}_{i=1}^{t-1})(x_t) \neq h(x_t)\right]. \tag{7.2}$$

We say that a class $\mathcal{H}$ is online learnable if there exists an online learning algorithm $\mathbb{A}$ such that $M(\mathbb{A}, \mathcal{H}) < \infty$. A class $\mathcal{H}$ is online learnable if and only if the Littlestone dimension of $\mathcal{H}$ denoted $\mathrm{lit}(\mathcal{H})$ is finite (Littlestone, 1987). Furthermore, Littlestone (1987) proposed the Standard Optimal algorithm ($\mathsf{SOA}$) and showed that $M(\mathsf{SOA}, \mathcal{H}) \leq \mathrm{lit}(\mathcal{H})$. We now briefly recall the definition of Littlestone dimension by introducing the notion of Littlestone trees:

**Definition 7.1** (Littlestone trees). *A Littlestone tree for $\mathcal{H}$ is a complete binary tree of depth $d \leq \infty$ whose internal nodes are labeled by instances from $\mathcal{X}$, and whose two edges connecting a node to its children are labeled with $+1$ and $-1$ such that every finite path emanating from the root is consistent with some concept in $\mathcal{H}$. That is, a Littlestone tree is a collection $\left\{x_{\boldsymbol{u}} : 0 \leq k < d, \boldsymbol{u} \in \{\pm 1\}^k\right\} \subseteq \mathcal{X}$ such that for every $\boldsymbol{y} \in \{\pm 1\}^d$, there exists $h \in \mathcal{H}$ such that $h(x_{\boldsymbol{y}_{1:k}}) = y_{k+1}$ for $0 \leq k < d$.*

**Definition 7.2** (Littlestone dimension). *The Littlestone dimension of $\mathcal{H}$, denoted $\mathrm{lit}(\mathcal{H})$, is the largest integer $d$ such that there exists a Littlestone tree for $\mathcal{H}$ of depth $d$ (see Definition 7.1). If no such $d$ exists, then $\mathrm{lit}(\mathcal{H})$ is said to be infinite.*

## 7.3 Access to a Perfect Attack Oracle

In this section, we study robust learning with algorithms that are only allowed access to a perfect attack oracle for $\mathcal{U}$ at training-time. Formally,

**Definition 7.3** (Perfect Attack Oracle). *Denote by $\mathsf{O}_{\mathcal{U}}$ a perfect attack oracle for $\mathcal{U}$. $\mathsf{O}_{\mathcal{U}}$ takes as input a predictor $f : \mathcal{X} \to \mathcal{Y}$ and an example $(x, y) \in \mathcal{X} \times \mathcal{Y}$ and either: (a) asserts that $f$ is robust on $(x, y)$ (i.e. $\forall z \in \mathcal{U}(x), f(z) = y$), or (b) returns a perturbation $z \in \mathcal{U}(x)$ such that $f(z) \neq y$.*[*]

In the Perfect Attack Oracle model, a learning algorithm $\mathbb{B}$ takes as input iid distributed training samples $S = \{(x_1, y_1), \ldots, (x_m, y_m)\}$ drawn from an unknown distribution $\mathcal{D}$ over $\mathcal{X} \times \mathcal{Y}$, and a black-box perfect attack oracle $\mathsf{O}_{\mathcal{U}}$. Learner $\mathbb{B}$ can query $\mathsf{O}_{\mathcal{U}}$ by calling it $T$ times with queries of the form: $(g_t, (x'_t, y'_t))$, where $g_t : \mathcal{X} \to \mathcal{Y}$ is a predictor and $(x'_t, y'_t)$ is a labeled example. The goal of learner $\mathbb{B}$ is to output a predictor $\hat{h} \in \mathcal{Y}^{\mathcal{X}}$ with small robust risk $\mathrm{R}_{\mathcal{U}}(\hat{h}; \mathcal{D}) \leq \varepsilon$ (see Equation 2.1). Learner $\mathbb{B}$ $(\varepsilon, \delta)$-robustly PAC learns $\mathcal{H}$ in the Perfect Attack Oracle model with oracle complexity $T(\varepsilon, \delta)$ if for any perturbation set $\mathcal{U}$, learner $\mathbb{B}$ $(\varepsilon, \delta)$-robustly PAC learns $\mathcal{H}$ with at most $T(\varepsilon, \delta)$ calls to $\mathsf{O}_{\mathcal{U}}$.

From a practical or engineering perspective, to establish robust generalization guarantees with respect to $\mathcal{U}$ in the Perfect Attack Oracle model, it suffices to build a perfect attack oracle for $\mathcal{U}$. Furthermore, to achieve robustness guarantees to multiple perturbation sets $\mathcal{U}_1, \ldots, \mathcal{U}_k$ concurrently, which is a goal of interest in practice (see e.g., Kang, Sun, Hendrycks, Brown, and Steinhardt, 2019a, Tramèr and Boneh, 2019, Maini, Wong, and Kolter, 2020), it suffices to *separately* build perfect attack oracles $\mathsf{O}_{\mathcal{U}_1}, \ldots, \mathsf{O}_{\mathcal{U}_k}$, and then implement a perfect attack oracle for the union $\cup_{i \leq k} \mathcal{U}_i$ by calling each attack oracle $\mathsf{O}_{\mathcal{U}_1}, \ldots, \mathsf{O}_{\mathcal{U}_k}$ separately.

**Questions.** *What hypothesis classes $\mathcal{H}$ are robustly PAC learnable in the Perfect Attack Oracle model? How can we learn a generic $\mathcal{H}$ using such access? With how many samples $m$ and oracle calls $T$?*

Summary of Results We begin in Section 7.3.1 with the realizable setting under which it is assumed that there is a predictor $h^* \in \mathcal{H}$ with zero robust risk, i.e. $\inf_{h \in \mathcal{H}} \mathrm{R}_{\mathcal{U}}(h; \mathcal{D}) = 0$.

[*] To be clear, we suppose $\mathsf{O}_{\mathcal{U}}$ acts as a function so that the $z$ it returns from calling $\mathsf{O}_{\mathcal{U}}(g, (x, y))$ is deterministic and oblivious to the history of interactions.

In Theorem 7.5, we give a simple algorithm (Algorithm E.3) CycleRobust that robustly learns $\mathcal{H}$ in the Perfect Attack Oracle model with sample complexity $m = O(\text{lit}(\mathcal{H}))$ and oracle complexity $T = O(\text{lit}^2(\mathcal{H}))$. In Theorem 7.6, we give an alternative algorithm (Algorithm E.3) RLUA to robustly learn $\mathcal{H}$ in the Perfect Attack Oracle model with reduced sample complexity $m = \tilde{O}(\text{vc}(\mathcal{H})\text{vc}^{*2}(\mathcal{H}))$ depending only the VC and dual VC dimension but at the cost of higher oracle complexity $T \approx 2^{\tilde{O}(\text{vc}^2(\mathcal{H})\text{vc}^{*2}(\mathcal{H}))}\text{lit}(\mathcal{H})$. Then, in Section 7.3.2, we extend our algorithmic results in Theorem 7.7 and Theorem 7.8 to the more general agnostic setting where we want to compete with the best attainable robust risk $\inf_{h\in\mathcal{H}} \text{R}_{\mathcal{U}}(h; \mathcal{D})$. Finally, in Section 7.3.3, we give a lower bound on the oracle complexity necessary to robustly learn in the Perfect Attack Oracle model. In Corollary 7.11, we show that for any class $\mathcal{H}$, the oracle complexity to robustly learn $\mathcal{H}$ is at least $\Omega(\log\log \text{lit}(\mathcal{H}))$. Furthermore, Corollary 7.12 gives a specific hypothesis class $\mathcal{H}$ with $\text{vc}(\mathcal{H}) = O(1) \ll \text{lit}(\mathcal{H})$ such that the oracle complexity to robustly learn $\mathcal{H}$ is at least $\Omega(\text{lit}(\mathcal{H}))$. These results are summarized in Table 7.1 on page 101.

Related Work Later on, in Chapter 8, we will gave an algorithm based on the Ellipsoid method to efficiently robustly learn halfspaces (linear predictors) in the Perfect Attack Oracle model in the realizable setting, for a broad range of perturbation sets $\mathcal{U}$ given access to a separation oracle for $\mathcal{U}$, with oracle complexity that depends on the bit complexity.

### 7.3.1 Algorithms and guarantees in the realizable setting

We begin in Theorem 7.4 and Theorem 7.5 with two simple algorithms: OnePassRobust (Algorithm E.1) and CycleRobust (Algorithm E.2), based on traditional online-to-batch conversion schemes, that robustly PAC learn a class $\mathcal{H}$ with sample complexity and oracle complexity depending on the Littlestone dimension $\text{lit}(\mathcal{H})$.

First, OnePassRobust (Algorithm E.1) runs an online learner $\mathbb{A}$ for $\mathcal{H}$ on a training set $S$ in a one-pass fashion, updating the online learner each time its not robustly correct on a training example using a Perfect Attack Oracle. This algorithm and analysis are based on the longest survivor technique (GALLANT, 1986). The proof is provided in Appendix E.1.1.

**Theorem 7.4.** *For any class $\mathcal{H}$,* OnePassRobust *(Algorithm E.1) robustly PAC learns $\mathcal{H}$ w.r.t. any $\mathcal{U}$ with:*

1. *Sample complexity* $m(\varepsilon,\delta) \leq 2^{\frac{\mathrm{lit}(\mathcal{H})}{\varepsilon}} \log\left(\frac{\mathrm{lit}(\mathcal{H})}{\delta}\right)$.
2. *Oracle complexity* $T(\varepsilon,\delta) \leq 2^{\frac{\mathrm{lit}(\mathcal{H})}{\varepsilon}} \log\left(\frac{\mathrm{lit}(\mathcal{H})}{\delta}\right)$.

Next, we can achieve slightly better sample complexity (at the expense of worse oracle complexity) using CycleRobust (Algorithm E.2), which cycles an online learner $\mathbb{A}$ for $\mathcal{H}$ on the training set $S$ until it robustly correctly classifies all training examples. To establish a robust generalization guarantee, we show that CycleRobust (Algorithm E.2) can be viewed as a stable compression scheme for the robust loss. This conversion technique and its connection to stable sample compression schemes have been recently studied in the standard 0-1 loss setting (Bousquet, Hanneke, Moran, and Zhivotovskiy, 2020). The proof is provided in Appendix E.1.2.

**Theorem 7.5.** *For any class $\mathcal{H}$,* CycleRobust *(Algorithm E.3) robustly PAC learns $\mathcal{H}$ w.r.t. any $\mathcal{U}$ with:*

1. *Sample complexity* $m(\varepsilon,\delta) = O\left(\frac{\mathrm{lit}(\mathcal{H})+\log(1/\delta)}{\varepsilon}\right)$.
2. *Oracle complexity* $T(\varepsilon,\delta) = m(\varepsilon,\delta)\mathrm{lit}(\mathcal{H})$.

*Furthermore, the output of* CycleRobust *achieves zero robust loss on the training sample.*

OnePassRobust (Algorithm E.1) and CycleRobust (Algorithm E.2) robustly PAC learn $\mathcal{H}$ in the Perfect Attack Oracle model with sample complexity and oracle complexity both depending on the Littlestone dimension $\mathrm{lit}(\mathcal{H})$. But are there robust learning algorithms with better sample complexity and/or oracle complexity? At least with explicit knowledge of $\mathcal{U}$, we know that we can robustly PAC learn $\mathcal{H}$ with $\tilde{O}(\mathrm{vc}(\mathcal{H})\mathrm{vc}^*(\mathcal{H}))$ sample complexity (Chapter 3, Theorem 3.4), which is much smaller than $\mathrm{lit}(\mathcal{H})$ for many natural classes (e.g., halfspaces). Can we obtain a similar sample complexity bound in the Perfect Attack Oracle model, where explicit knowledge of $\mathcal{U}$ is not allowed? We prove *yes* in Theorem 7.6. Specifically, we give an algorithm that can robustly PAC learn $\mathcal{H}$ in Perfect Attack Oracle model with sample complexity $\tilde{O}(\mathrm{vc}(\mathcal{H})\mathrm{vc}^{*2}(\mathcal{H}))$ independent of $\mathrm{lit}(\mathcal{H})$.

**Theorem 7.6.** *For any class $\mathcal{H}$ with $\mathrm{vc}(\mathcal{H}) = d$ and $\mathrm{vc}^*(\mathcal{H}) = d^*$, there exists a learning algorithm $\tilde{\mathbb{B}}$ that robustly PAC learns $\mathcal{H}$ w.r.t any $\mathcal{U}$ with:*

1. *Sample Complexity* $m(\varepsilon, \delta) = O\left(\frac{dd^{*2}\log^2 d^*}{\varepsilon}\log^2\left(\frac{dd^{*2}\log^2 d^*}{\varepsilon}\right) + \frac{1}{\varepsilon}\log(\frac{2}{\delta})\right)$.
2. *Oracle Complexity* $T_{\mathrm{RE}}(\varepsilon, \delta) = \left(2^{O(d^2 d^{*2}\log^2 d^*)}\mathrm{lit}(\mathcal{H}) + m(\varepsilon, \delta)\right)\left(\log\left(m(\varepsilon, \delta)\right)\left(\log\left(\frac{\log(m(\varepsilon,\delta))}{\delta}\right)\right)\right)$.

The full proof is deferred to Appendix E.1.3, but we briefly describe the main building blocks of this result. We will adapt our algorithm from Chapter 3 (see Theorem 3.4) and establish a robust generalization guarantee that depends only on $\mathrm{vc}(\mathcal{H})$ and $\mathrm{vc}^*(\mathcal{H})$. In particular, the learning algorithm of Chapter 3 (Theorem 3.4) required explicit knowledge of $\mathcal{U}$, this knowledge was used to implement a $\mathsf{RERM}_{\mathcal{U}}$ oracle for $\mathcal{H}$, and for a sample inflation and a discretization step which is crucial to establish robust generalization based on sample compression. As explicit knowledge is not allowed in the Perfect Attack Oracle model, we show that we can avoid these limitations and use only queries to $\mathsf{O}_{\mathcal{U}}$. Specifically, observe that CycleRobust (Algorithm E.2) implements a $\mathsf{RERM}_{\mathcal{U}}$ oracle for $\mathcal{H}$ using only black-box access to $\mathsf{O}_{\mathcal{U}}$, since by Theorem 7.5 the output of CycleRobust achieves zero robust loss on its input dataset $S$. Similarly, the discretization step can be carried using only queries to $\mathsf{O}_{\mathcal{U}}$, by constructing queries using the output predictors of CycleRobust to force the oracle to reveal perturbations of the empirical sample $S$, we leave further details to the proof. While this suffices to establish a result of robust PAC learning in the Perfect Attack Oracle model with sample complexity completely independent of $\mathrm{lit}(\mathcal{H})$ (Theorem E.2 and its proof in Appendix E.1.3), we can further improve the dependence on $\varepsilon$ and $\delta$ in the oracle complexity. To this end, we treat the algorithm (Algorithm E.3) RLUA from Theorem E.2 as a *weak* robust learner with fixed $\varepsilon_0$ and $\delta_0$ and boost its robust error guarantee to improve the oracle complexity and obtain the result in Theorem 7.6.

### 7.3.2 Algorithms and guarantees in the agnostic setting

We now consider the more general agnostic setting where we want to compete with the best attainable robust risk $\inf_{h\in\mathcal{H}} \mathrm{R}_{\mathcal{U}}(h; \mathcal{D})$. Mirroring the results from the realizable section, we begin in Theorem 7.7 with a simple algorithm that can only guarantee a robust error at most $2\inf_{h\in\mathcal{H}} \mathrm{R}_{\mathcal{U}}(h;\mathcal{D}) + \varepsilon$ with sample and oracle complexity depending on the Littlestone dimension $\mathrm{lit}(\mathcal{H})$. Then, in Theorem 7.8, we give a reduction to the realizable setting of Theorem 7.6, that agnostically robustly PAC learns $\mathcal{H}$ in the Perfect Attack Oracle model with sample complexity depending only on the $\mathrm{vc}(\mathcal{H})$ and $\mathrm{vc}^*(\mathcal{H})$.

**Theorem 7.7.** *For any class $\mathcal{H}$,* Weighted Majority *(Algorithm E.4) guarantees that for any perturbation set $\mathcal{U}$ and any distribution $\mathcal{D}$ over $\mathcal{X} \times \mathcal{Y}$, with sample complexity $m(\varepsilon,\delta) = O\left(\frac{\mathrm{lit}(\mathcal{H})\log(1/\varepsilon)+\log(1/\delta)}{\varepsilon^2}\right)$ and oracle complexity $T(\varepsilon,\delta) = O(m(\varepsilon,\delta)^2)$, with probability at least $1-\delta$ over $S \sim \mathcal{D}^{m(\varepsilon,\delta)}$,*

$$\mathrm{R}_{\mathcal{U}}(\mathsf{WM}(S, \mathsf{O}_{\mathcal{U}}); \mathcal{D}) \leq 2 \inf_{h\in\mathcal{H}} \mathrm{R}_{\mathcal{U}}(h;\mathcal{D}) + \varepsilon.$$

We briefly describe the main ingredients of this result. First, in Lemma E.3, we show that for any class $\mathcal{H}$ with finite cardinality, a variant of the Weighted Majority algorithm (Littlestone and Warmuth, 1994) presented in (Algorithm E.4) has a regret guarantee with respect to the robust loss. Then, in Lemma E.4, we extend this regret guarantee for infinite $\mathcal{H}$ using a technique due to Ben-David et al. (2009) for agnostic online learning. Finally, we apply a standard online-to-batch conversion (Cesa-Bianchi, Conconi, and Gentile, 2004) to convert the regret guarantee to a robust generalization guarantee. These helper lemmas and proofs are deferred to Appendix E.2.

Similarly to the realizable setting, we can establish an upper bound with sample complexity independent of $\mathrm{lit}(\mathcal{H})$. This is achieved via a reduction to the realizable setting (Theorem 7.6), following an argument of David, Moran, and Yehudayoff (2016) as done originally in Chapter 3, Subsection 3.3.2. The proof is deferred to Appendix E.2.

**Theorem 7.8** (Reduction to Realizable Setting). *For any class $\mathcal{H}$ with $\mathrm{vc}(\mathcal{H}) = d$ and $\mathrm{vc}^*(\mathcal{H}) = d^*$, there is a learning algorithm $\tilde{\mathbb{B}}$ that robustly* agnostically *PAC learns $\mathcal{H}$ w.r.t any $\mathcal{U}$ with:*

1. *Sample Complexity* $m(\varepsilon,\delta) = O\left(\frac{dd^{*2}\log^2 d^*}{\varepsilon^2}\log^2\left(\frac{dd^{*2}\log^2 d^*}{\varepsilon}\right) + \frac{1}{\varepsilon^2}\log(\frac{2}{\delta})\right)$.
2. *Oracle Complexity* $T(\varepsilon,\delta) = 2^{m(\varepsilon,\delta)}\mathrm{lit}(\mathcal{H}) + T_{\mathrm{RE}}(\varepsilon,\delta)$.

### 7.3.3 Lowerbound on Oracle Complexity

In Section 7.3.1 and Section 7.3.2, we have shown that it is possible to robustly PAC learn in the Perfect Attack Oracle model with sample complexity that is completely *independent* of the Littlestone dimension, and with oracle complexity that *depends* on the Littlestone dimension. It is natural to ask whether the oracle complexity can be improved. Perhaps, we can avoid dependence on Littlestone dimension altogether? In Theorem 7.10, we prove that the answer is *no*.

Specifically, we will first establish a lower bound in terms of another complexity measure known as the Threshold dimension of $\mathcal{H}$, denoted by $\text{Tdim}(\mathcal{H})$. Informally, $\text{Tdim}(\mathcal{H})$ is the largest number of thresholds that can be embedded in class $\mathcal{H}$ (see Definition 7.9 below). Importantly, the Threshold dimension of $\mathcal{H}$ is related to the Littlestone dimension of $\mathcal{H}$ and is known to satisfy: $\lfloor \log_2 \text{lit}(\mathcal{H}) \rfloor \leq \text{Tdim}(\mathcal{H}) \leq 2^{\text{lit}(\mathcal{H})}$ (Shelah, 1990, Hodges, 1997, Alon, Livni, Malliaris, and Moran, 2019). This relationship was recently used to establish that *private* PAC learnability implies online learnability (Alon, Livni, Malliaris, and Moran, 2019).

**Definition 7.9** (Threshold Dimension). *We say that a class $\mathcal{H} \subseteq \mathcal{Y}^{\mathcal{X}}$ contains k thresholds if $\exists x_1, \dots, x_k \in \mathcal{X}$ and $\exists h_1, \dots, h_k \in \mathcal{H}$ such that $h_i(x_j) = +1$ if and only if $j \leq i, \forall i, j \leq k$. The Threshold dimension of $\mathcal{H}$, $\text{Tdim}(\mathcal{H})$, is the largest integer k such that $\mathcal{H}$ contains k thresholds.*

**Theorem 7.10.** *For any class $\mathcal{H}$, there exists a distribution $\mathcal{D}$ over $\mathcal{X} \times \mathcal{Y}$, such that for any learner $\mathbb{B}$, there exists a perturbation set $\mathcal{U} : \mathcal{X} \to 2^{\mathcal{X}}$ where $\inf_{h \in \mathcal{H}} \text{R}_{\mathcal{U}}(h; \mathcal{D}) = 0$ and a perfect attack oracle $\mathsf{O}_{\mathcal{U}}$ such that $\mathbb{B}$ needs to make $\frac{\log_2(\text{Tdim}(\mathcal{H})-1)}{2}$ oracle queries to $\mathsf{O}_{\mathcal{U}}$ to robustly learn $\mathcal{D}$.*

The full proof is deferred to Appendix E.3, but we will provide some intuition behind the proof. The main idea is to use $h_1, \dots, h_{\text{Tdim}(\mathcal{H})}$ thresholds to construct $\mathcal{U}_1, \dots, \mathcal{U}_{\text{Tdim}(\mathcal{H})}$ perturbation sets. We will setup a single source distribution $\mathcal{D}$ that is known to the learner, but choose a random perturbation set from $\mathcal{U}_1, \dots, \mathcal{U}_{\text{Tdim}(\mathcal{H})}$. In order for the learner to *robustly* learn $\mathcal{D}$, it needs to figure out which perturbation set is picked, and that requires $\Omega(\log \text{Tdim}(\mathcal{H}))$ queries to the oracle $\mathsf{O}_{\mathcal{U}}$. Since $\text{Tdim}(\mathcal{H}) \geq \lfloor \log_2 \text{lit}(\mathcal{H}) \rfloor$, Theorem 7.10 implies the following corollaries.

**Corollary 7.11.** *For any class $\mathcal{H}$, the oracle complexity to robustly learn $\mathcal{H}$ in the Perfect Attack Oracle model is at least* $\Omega(\log\log \mathrm{lit}(\mathcal{H}))$.

**Corollary 7.12.** *For any $n \in \mathbb{N}$, the class $\mathcal{H}_n$ consisting of $n$ thresholds satisfies* $\mathrm{lit}(\mathcal{H}_n) = \log_2 \mathrm{Tdim}(\mathcal{H}_n) = \log_2 n$[†] *and* $\mathrm{vc}(\mathcal{H}_n) = O(1)$. *Thus, the oracle complexity to robustly learn $\mathcal{H}_n$ in the Perfect Attack Oracle model is* $\Omega(\mathrm{lit}(\mathcal{H}_n))$.

A couple of remarks are in order. First, the lower bound of $\Omega(\log\log \mathrm{lit}(\mathcal{H}))$ applies to any hypothesis class $\mathcal{H}$, but a stronger lower bound for the special case of thresholds can be shown where $\Omega(\mathrm{lit}(\mathcal{H}))$ oracle queries are needed. Second, observe that these lower bounds apply to learning algorithms that know the distribution $\mathcal{D}$, and so even with infinite sample complexity, it is not possible to have oracle complexity independent of Littlestone dimension.

### 7.3.4 Gaps and Open Questions

We have established that $\mathcal{H}$ is robustly PAC learnable in the Perfect Attack Oracle model if and only if $\mathcal{H}$ is online learnable. We provided a simple online-to-batch conversion scheme CycleRobust (Algorithm E.3) with sample and oracle complexity scaling with $\mathrm{lit}(\mathcal{H})$. Then, with a more sophisticated algorithm, RLUA (Algorithm E.3), we get an improved sample complexity depending only on $\mathrm{vc}(\mathcal{H})$ and $\mathrm{vc}^*(\mathcal{H})$, but at the expense of higher oracle complexity with an exponential dependence on $\mathrm{vc}(\mathcal{H})$ and $\mathrm{vc}^*(\mathcal{H})$ and linear dependence on $\mathrm{lit}(\mathcal{H})$. We also showed that for any class $\mathcal{H}$, an oracle complexity of $\log\log \mathrm{lit}(\mathcal{H})$ is unavoidable, and furthermore, exhibit a class $\mathcal{H}$ with $\mathrm{vc}(\mathcal{H}) = O(1)$ and $\mathrm{lit}(\mathcal{H}) \gg \mathrm{vc}(\mathcal{H})$ where an oracle complexity of $\Omega(\mathrm{lit}(\mathcal{H}))$ is unavoidable.

An interesting direction is to improve the oracle complexity to perhaps a polynomial dependence $\mathrm{poly}(\mathrm{vc}(\mathcal{H}), \mathrm{vc}^*(\mathcal{H}))\mathrm{lit}(\mathcal{H})$, or more ambitiously $\mathrm{poly}(\mathrm{vc}(\mathcal{H}))\mathrm{lit}(\mathcal{H})$. It would also be interesting to establish a finer characterization for the oracle complexity that is adaptive to the perturbation sets $\mathcal{U}$, perhaps depending on some notion measuring the complexity of $\mathcal{U}$. Also, can we strengthen the lower bound and show that for any $\mathcal{H}$, $\Omega(\mathrm{lit}(\mathcal{H}))$ oracle complexity is necessary to robustly learn $\mathcal{H}$ or is there another complexity measure that tightly characterizes the oracle complexity?

[†]We learned about this fact from the following talk: `https://youtu.be/NPpPiWYcmPk`

## 7.4 Bounding the Number of Successful Attacks

In Section 7.3, we considered having access to a *perfect* attack oracle $\mathsf{O}_{\mathcal{U}}$. But in many settings, our practical attack oracle attack engines, e.g., PGD (Madry, Makelov, Schmidt, Tsipras, and Vladu, 2018), are not perfect—they might not always find miss-classified adversarsial perturbations even when they do exist. Also, the perturbation set $\mathcal{U}$ might be fairly abstract, like "images indistinguishable to the human eye", and so there isn't really a perfect attack oracle, but rather just approximations to it. Can we still provide meaningful robustness guarantees even with *imperfect* attackers?

In this section, we introduce a model where we consider working with an actual adversary or attack algorithm that is possibly imperfect, and the goal is to bound the number of successful attacks. In this model, a learning algorithm $\mathbb{B}$ first receives as input iid distributed training samples $S = \{(x_1, y_1), \ldots, (x_m, y_m)\}$ drawn from an unknown distribution $\mathcal{D}$ over $\mathcal{X} \times \mathcal{Y}$. Then, the learning algorithm $\mathbb{B}$ makes predictions on examples $z_t \in \mathcal{U}(x'_t)$ where $z_t$ is an adversarial perturbation chosen by an adversary $\mathbb{A}$, and $(x'_t, y'_t)$ is an iid sample drawn from $\mathcal{D}$. The adversary $\mathbb{A}$ has access to the random sample $(x'_t, y'_t)$ and the predictor used by learner $\mathbb{B}$, $h_t = \mathbb{B}(S \cup \{(z_j, y_j)_{j=1}^{t-1}\})$, but learner $\mathbb{B}$ only sees the perturbation $z_t$. After learner $\mathbb{B}$ makes its prediction $\hat{y}_t = h_t(z_t)$, the true label $y_t$ is revealed to $\mathbb{B}$. The goal is to bound the number of successful adversarial attacks where $\hat{y}_t \neq y_t$. For a class $\mathcal{H}$ and a learner $\mathbb{B}$, the maximum number of successful attacks caused by an adversary $\mathbb{A}$ w.r.t. $\mathcal{U}$ on a distribution $\mathcal{D}$ satisfying $\inf_{h \in \mathcal{H}} \mathrm{R}_{\mathcal{U}}(h; \mathcal{D}) = 0$ is defined as

$$M_{\mathcal{U},\mathbb{A}}(\mathbb{B}, \mathcal{H}; \mathcal{D}) := \sum_{t=1}^{\infty} 1\left[\mathbb{B}(S \cup \{(z_i, y_i)\}_{i=1}^{t-1})(z_t) \neq y_t\right], \tag{7.3}$$

where $z_t = \mathbb{A}(\mathbb{B}(S \cup \{(z_i, y_i)\}_{i=1}^{t-1}), (x'_t, y'_t))$ and $\{(x'_t, y'_t)\}_{t=1}^{\infty}$ are iid samples from $\mathcal{D}$.

**Questions.** *Can we obtain upper bounds and lower bounds on the maximum number of successful attacks for generic classes $\mathcal{H}$? Can additional training samples from $\mathcal{D}$ help?*

First, we show that we can upper bound the maximum number of successful attacks on any online learner $\mathbb{B}$ for $\mathcal{H}$ by the online mistake bound of $\mathbb{B}$.

**Theorem 7.13** (Upper Bound). *For any class $\mathcal{H}$ and any online learner $\mathbb{B}$, for any perturbation set $\mathcal{U}$, adversary $\mathbb{A}$, and distribution $\mathcal{D}$, $M_{\mathcal{U},\mathbb{A}}(\mathbb{B}, \mathcal{H}; \mathcal{D}) \leq M(\mathbb{B}, \mathcal{H})$. In particular, the Standard Optimal Algorithm (*SOA*) has an attack bound of at most* $\mathrm{lit}(\mathcal{H})$.

*Proof.* The proof follows directly from the definition of the online mistake bound (see Equation 7.2) and the online attack bound (see Equation 7.3). □

Is this the best achievable upper bound on the number of successful attacks? Perhaps there are learning algorithms with an attack bound that is much smaller than $\mathrm{lit}(\mathcal{H})$, maybe an attack bound that scales with $\mathrm{vc}(\mathcal{H})$? We next answer this question in the negative. Using the same the lower bound construction from Theorem 7.10 in Section 7.3.3, we first establish a lower bound on the number of the successful attacks based on the Threshold dimension of $\mathcal{H}$ (see Definition 7.9) (proof deferred to Appendix E.4). We then utilize the relationship $\mathrm{Tdim}(\mathcal{H}) \geq \lfloor \log_2 \mathrm{lit}(\mathcal{H}) \rfloor$ to get the corollaries.

**Theorem 7.14** (Lower Bound). *For any class $\mathcal{H}$, there exists a distribution $\mathcal{D}$ over $\mathcal{X} \times \mathcal{Y}$, such that for any learner $\mathbb{B}$, there is a perturbation set $\mathcal{U}$ where $\inf_{h\in\mathcal{H}} \mathrm{R}_{\mathcal{U}}(h; \mathcal{D}) = 0$ and an adversary $\mathbb{A}$ that makes at least $\frac{\log_2(\mathrm{Tdim}(\mathcal{H})-1)}{2}$ successful attacks on learner $\mathbb{B}$.*

**Corollary 7.15.** *For any class $\mathcal{H}$, there is a distribution $\mathcal{D}$, such that for any learner $\mathbb{B}$, there is a perturbation set $\mathcal{U}$ and adversary $\mathbb{A}$ such that $M_{\mathcal{U},\mathbb{A}}(\mathbb{B}, \mathcal{H}; \mathcal{D}) \geq \Omega(\log\log \mathrm{lit}(\mathcal{H}))$.*

**Corollary 7.16.** *For any $n \in \mathbb{N}$, the class $\mathcal{H}_n$ consisting of $n$ thresholds satisfies* $\mathrm{lit}(\mathcal{H}_n) = \log_2 \mathrm{Tdim}(\mathcal{H}_n) = \log_2 n$ *and* $\mathrm{vc}(\mathcal{H}_n) = O(1)$. *Thus,* $\exists_{\mathcal{D}} \forall_{\mathbb{B}} \exists_{\mathcal{U},\mathbb{A}} M_{\mathcal{U},\mathbb{A}}(\mathbb{B}, \mathcal{H}_n; \mathcal{D}) \geq \Omega(\mathrm{lit}(\mathcal{H}_n))$.

We remark that these lower bounds hold even for learning algorithms $\mathbb{B}$ that perfectly know the source distribution $\mathcal{D}$. For the class $\mathcal{H}_n$ of $n$ thresholds, we cannot expect a learning algorithm that leverages extra training data that avoids the $\Omega(\mathrm{lit}(\mathcal{H}_n))$ lower bound. But it might be that for other classes $\mathcal{H}$, additional training data might help reduce the attack bound to $\log \mathrm{lit}(\mathcal{H})$ or $\log\log \mathrm{lit}(\mathcal{H})$.

Gaps and Open Questions We have only considered the realizable setting, where there is a predictor $h \in \mathcal{H}$ that is perfectly robust to the attacker $\mathbb{A}$. It would be interesting to extend the guarantees to the agnostic setting. Can we strengthen the lower bound and show that for any $\mathcal{H}$, an attack bound of $\Omega(\mathrm{lit}(\mathcal{H}))$ is unavoidable or is there another complexity measure that tightly characterizes the attack bound? Are there examples of classes $\mathcal{H}$ where collecting additional samples from $\mathcal{D}$ helps reduce the number of successful attacks?

## 7.5 Robust Generalization to Imperfect Attack Algorithms

In Section 7.4, given an online learning algorithm, we can guarantee a finite number of successful attacks from any attacking algorithm even if it was imperfect. But what if we want to work in a more traditional train-then-ship approach, where we first ensure adversarial robustness without releasing anything, and only then release? Can we provide any robust generalization guarantees when we only have access to an *imperfect* attacking algorithm such as PGD (Madry, Makelov, Schmidt, Tsipras, and Vladu, 2018) at training-time?

In this model, a learning algorithm $\mathbb{B}$ takes as input a black-box (possibly imperfect) attacker $\mathbb{A}$, and iid distributed training samples $S = \{(x_1, y_1), \dots, (x_m, y_m)\}$ from an unknown distribution $\mathcal{D}$ over $\mathcal{X} \times \mathcal{Y}$. Learner $\mathbb{B}$ can query $\mathbb{A}$ by calling it $T$ times with queries of the form: $(g_t, (x'_t, y'_t))$, where $g_t : \mathcal{X} \to \mathcal{Y}$ is a predictor and $(x'_t, y'_t)$ is a labeled example. The goal of learner $\mathbb{B}$ is to output a predictor $\hat{h} \in \mathcal{Y}^{\mathcal{X}}$ with small error w.r.t. future attacks from $\mathbb{A}$, $\mathrm{err}_{\mathbb{A}}(\hat{h}; \mathcal{D}) \leq \varepsilon$ (see Equation 7.1).

**Definition 7.17** (Robust PAC Learnability with Imperfect Attackers)**.** *Learner $\mathbb{B}$ $(\varepsilon, \delta)$-robustly PAC learns $\mathcal{H} \subseteq \mathcal{Y}^{\mathcal{X}}$ with sample complexity $m(\varepsilon, \delta) : (0,1)^2 \to \mathbb{N}$ and oracle complexity $T(\varepsilon, \delta) : (0,1)^2 \to \mathbb{N}$ if for any (possibly randomized and imperfect) attacker $\mathbb{A} : \mathcal{Y}^{\mathcal{X}} \times (\mathcal{X} \times \mathcal{Y}) \to \mathcal{X}$, any distribution $\mathcal{D}$ over $\mathcal{X} \times \mathcal{Y}$, with at most $T(\varepsilon, \delta)$ oracle calls to $\mathbb{A}$ and with probability at least $1 - \delta$ over $S \sim \mathcal{D}^{m(\varepsilon,\delta)}$ :* $\mathrm{err}_{\mathbb{A}}(\mathbb{B}(S, \mathbb{A})) \leq \inf_{h \in \mathcal{H}} \mathrm{err}_{\mathbb{A}}(h) + \varepsilon$.

In this model, access to an imperfect attacker $\mathbb{A}$ at training-time ensures robust generalization to this *specific* attacker $\mathbb{A}$ at test-time. This is a different guarantee from robust generalization to a perturbation set $\mathcal{U}$, because it might well be that that there is a stronger attack algorithm $\mathbb{A}'$ than $\mathbb{A}$ such that $\mathrm{err}_{\mathbb{A}'}(\hat{h}; \mathcal{D}) \gg \mathrm{err}_{\mathbb{A}}(\hat{h}; \mathcal{D})$. Furthermore, since the "strength" of the attack algorithm $\mathbb{A}$ is a function of the predictor $\hat{h}$ it is attacking, establishing a generalization guarantee w.r.t. $\mathbb{A}$ is not immediate, and does not follow for example from our results in Section 7.3.

We relate robust learnability in this model to the online model from Section 7.4. In Theorem 7.18, we observe that we can apply a standard online-to-batch conversion based on the longest survivor technique (GALLANT, 1986) to establish generalization guarantees with respect to future attacks made by $\mathbb{A}$. Specifically, we simply output a predictor $\hat{h}$ that has survived a sufficient number of attacks from $\mathbb{A}$. The full algorithm and proof are presented in Appendix E.5.

**Theorem 7.18.** *For any class $\mathcal{H}$, Algorithm E.6 robustly PAC learns $\mathcal{H}$ w.r.t. any (possibly randomized and imperfect) attacker $\mathbb{A}$ and any distribution $\mathcal{D}$ such that* $\inf_{h\in\mathcal{H}} \mathrm{err}_{\mathbb{A}}(h;\mathcal{D}) = 0$, *with sample complexity $m(\varepsilon,\delta)$ and oracle complexity $T(\varepsilon,\delta)$ at most* $O\left(\frac{\mathrm{lit}(\mathcal{H})\log(\mathrm{lit}(\mathcal{H})/\delta)}{\varepsilon}\right)$.

Gaps and Open Questions Currently we only provide generalization guarantees in the realizable setting. It would be interesting to extend our guarantees to the agnostic setting. Are there algorithms with better sample complexity perhaps depending only on the VC dimension? We established such a result in Section 7.3 with access to a perfect attack oracle, but the same approach does not go through when using an imperfect attacker. What about better oracle complexity? Can we obtain similar generalization guarantees for adaptive attacking algorithms that change over time? Can we obtain generalization guarantees against a family of attacking algorithms (e.g., first order attacks)?

## 7.6 Discussion and Open Directions

In this chapter, we consider robust learning with respect to *unknown* perturbation sets $\mathcal{U}$. We initiate the quest to find the "right" model of access to $\mathcal{U}$ by considering different forms of access and studying the robustness guarantees achievable in each case. One of the main takeaways from this chapter is that we need to be mindful about what form of access to $\mathcal{U}$ we are assuming, because the guarantees that can be achieved can be different. So knowledge about $\mathcal{U}$ should not be thought of as a free resource, but rather we should quantify the complexity of the information we are asking about $\mathcal{U}$.

In some ways, adversarial learning is an arms race, and Athalye, Carlini, and Wagner (2018) have illustrated that predictors trained to be secure against a specific attack, might be easily defeated by a different attack. Our *imperfect* attack oracle model in Section 7.5 certainly suffers from this problem. But, it can also be viewed as taking a step towards addressing it, as it provides a generic way of turning any new attack into a defence, and thus defending against it, and since this is done in a black-box manner, could at the very least hasten the development time needed to defend against a new attack. The online model in Section 7.4 in a sense does so explicitly, and can indeed handle arbitrary new attacks.

In Section 7.3 we establish robust generalization guarantees w.r.t. any attacking algorithm $\mathbb{A}$ for $\mathcal{U}$, but it requires a perfect adversarial attack oracle for $\mathcal{U}$: $\mathsf{O}_{\mathcal{U}}$, and in Section 7.5, we establish a generalization guarantee w.r.t. a *specific* attack algorithm $\mathbb{A}$ when given black-box access to $\mathbb{A}$ at training-time. These are in a sense two opposing ends of the spectrum. Are there other interesting models that provide weaker access than a perfect attack oracle $\mathsf{O}_{\mathcal{U}}$, but also provide a stronger guarantee than that of generalization to a particular attack? For example, under what conditions, can we generalize to a test-time attacker $\mathbb{A}_{\text{test}}$ that is different from the attacker $\mathbb{A}_{\text{train}}$ used at training-time. What if we are interested in providing guarantees to a family of test-time attackers (e.g. first-order algorithms), what form of access would be sufficient and necessary?

Under explicit knowledge of the perturbation set $\mathcal{U}$, in Chapter 3, we showed that we can robustly learn any hypothesis class $\mathcal{H}$ that is PAC learnable, i.e., finite $\text{vc}(\mathcal{H})$. Given only a perfect attack oracle to $\mathcal{U}$, we show in this chapter that we can robustly learn any hypothesis class $\mathcal{H}$ that is online learnable, i.e., finite $\text{lit}(\mathcal{H})$, and we give lower bounds showing that online learnability is necessary. Are there other reasonable models of access to $\mathcal{U}$ (weaker than explicit knowledge) that would allow us to robustly learn a broader family of hypothesis classes beyond those that are online learnable?

Our approach to robust learning in this chapter is modular, in particular, the *perfect* and *imperfect* attack oracles that we consider are independent of the hypothesis class $\mathcal{H}$ since they just receive a predictor $g : \mathcal{X} \to \mathcal{Y}$ as input. But how is $g$ provided? And do we expect the oracles to accept as input any predictor $g$ regardless of its complexity? A more careful look reveals that for the simple algorithm CycleRobust (Algorithm E.3) it suffices for the perfect attack oracle $\mathsf{O}_{\mathcal{U}}$ to accept only predictors $h \in \mathcal{H}$. This seems like it creates a dependency and breaks the modularity, but it does not, since, e.g., the oracle might be implemented in terms of a much larger and more generic class, such as neural nets with any architecture, as opposed to the specific architecture we are trying to learn, and which the oracle need not be aware of. But the more sophisticated algorithm RLUA (Algorithm E.3) requires calling the oracle on predictors outside $\mathcal{H}$ and so we do need the oracle to accept arbitrary predictors, or at least predictors from a much broader class than $\mathcal{H}$. How can this be translated to a computational rather than purely mathematical framework, and implemented in practice?

Are there sensible but generic assumptions on the perturbation set $\mathcal{U}$ that can lead to improved guarantees? Either assumptions that are on $\mathcal{U}$ separate from the class $\mathcal{H}$, i.e., that hold even if $\mathcal{H}$ is applied to one relabeling or permutation of $\mathcal{X}$ and $\mathcal{U}$ is applied to a different relabeling, or that rely on simple and generic relationships between $\mathcal{U}$ and $\mathcal{H}$.

8

# Computationally Efficient Robust Learning

## 8.1 Introduction

Thus far in this thesis, we have focused primarily on the statistical aspects of adversarially robust learning, including studying what learning rules should be used for robust learning and how much training data is needed to guarantee high robust accuracy. On the other hand, the computational aspects of adversarially robust PAC learning are less understood. In this chapter, we take a first step towards studying this broad algorithmic question with a focus on the fundamental problem of learning adversarially robust halfspaces. Halfspaces are binary predictors of the form $h_{\boldsymbol{w}}(\boldsymbol{x}) = \text{sign}(\langle \boldsymbol{w}, \boldsymbol{x} \rangle)$, where $\boldsymbol{w} \in \mathbb{R}^d$.

A first question to ask is whether efficient PAC learning implies efficient *robust* PAC learning, i.e., whether there is a general reduction that solves the adversarially robust learning problem. Recent work has provided strong evidence that this is not the case. Specifically, Bubeck, Lee, Price, and Razenshteyn (2019) showed that there exists a learning problem that can be learned efficiently non-robustly, but is computationally intractable to learn robustly (under plausible complexity-theoretic assumptions). There is also more recent evidence that suggests that this is also the case in the PAC model. Awasthi, Dutta, and Vijayaraghavan (2019) showed that it is computationally intractable to even weakly robustly learn degree-2 polynomial threshold functions (PTFs) with $\ell_\infty$ perturbations in the realizable setting, while PTFs of any constant degree are known to be efficiently PAC learnable non-robustly in the realizable setting. Gourdeau, Kanade, Kwiatkowska, and Worrell (2019) showed that there are hypothesis classes that are hard to robustly PAC learn, under the assumption that it is hard to non-robustly PAC learn.

The aforementioned discussion suggests that when studying robust PAC learning, we need to characterize which types of perturbation sets $\mathcal{U}$ admit *computationally efficient* robust PAC learners and under which noise assumptions. In the agnostic PAC setting, it is known that even weak (non-robust) learning of halfspaces is computationally intractable (Feldman, Gopalan, Khot, and Ponnuswami, 2006, Guruswami and Raghavendra, 2009, Diakonikolas, O'Donnell, Servedio, and Wu, 2011, Daniely, 2016). For $\ell_2$-perturbations, where $\mathcal{U}(x) = \{z : \|x - z\|_2 \leq \gamma\}$, it was recently shown that the complexity of proper learning is exponential in $1/\gamma$ (Diakonikolas, Kane, and Manurangsi, 2019b). In this chapter, we focus on the realizable case and the (more challenging) case of random label noise.

We can be more optimistic in the realizable setting. Halfspaces are efficiently PAC learnable non-robustly via Linear Programming Maass and Turán (1994), and under the margin assumption via the Perceptron algorithm Rosenblatt (1958). But what can we say about robustly PAC learning halfspaces? Given a perturbation set $\mathcal{U}$ and under the assumption that there is a halfspace $h_{\boldsymbol{w}}$ that robustly separates the data, can we efficiently learn a predictor with small robust risk?

Just as empirical risk minimization (ERM) is central for non-robust PAC learning, a core component of adversarially robust learning is minimizing the *robust* empirical risk on a dataset $S$,

$$\hat{h} \in \mathsf{RERM}_{\mathcal{U}}(S) \triangleq \operatorname*{argmin}_{h \in \mathcal{H}} \frac{1}{m} \sum_{i=1}^{m} \sup_{\boldsymbol{z} \in \mathcal{U}(\boldsymbol{x})} 1[h(\boldsymbol{z}) \neq y].$$

In this chapter, we provide necessary and sufficient conditions on perturbation sets $\mathcal{U}$, under which the robust empirical risk minimization ($\mathsf{RERM}$) problem is *efficiently* solvable in the realizable setting. We show that an efficient separation oracle for $\mathcal{U}$ yields an efficient solver for $\mathsf{RERM}_{\mathcal{U}}$, while an efficient *approximate* separation oracle for $\mathcal{U}$ is necessary for even computing the robust loss $\sup_{\boldsymbol{z} \in \mathcal{U}(\boldsymbol{x})} 1[h_{\boldsymbol{w}}(\boldsymbol{z}) \neq y]$ of a halfspace $h_{\boldsymbol{w}}$. In addition, we relax our realizability assumption and show that under random classification noise Angluin and Laird (1987), we can efficiently robustly PAC learn halfspaces with respect to any $\ell_p$ perturbation.

Main Results   Our main results for this chapter can be summarized as follows:

1. In the realizable setting, the class of halfspaces is efficiently robustly PAC learnable with respect to $\mathcal{U}$, given an efficient separation oracle for $\mathcal{U}$.

2. To even compute the robust risk with respect to $\mathcal{U}$ efficiently, an efficient *approximate* separation oracle for $\mathcal{U}$ is *necessary*.

3. In the random classification noise setting, the class of halfspaces is efficiently robustly PAC learnable with respect to any $\ell_p$ perturbation.

### 8.1.1 Related Work

Here we focus on recent work that is most closely related to the results of this chapter. Awasthi, Dutta, and Vijayaraghavan (2019) studied the tractability of RERM with respect to $\ell_\infty$ perturbations, obtaining efficient algorithms for halfspaces in the realizable setting, but showing that RERM for degree-2 polynomial threshold functions is computationally intractable (assuming NP $\neq$ RP). Gourdeau, Kanade, Kwiatkowska, and Worrell (2019) studied robust learnability of hypothesis classes defined over $\{0,1\}^n$ with respect to hamming distance, and showed that monotone conjunctions are robustly learnable when the adversary can perturb only $\mathcal{O}(\log n)$ bits, but are *not* robustly learnable even under the uniform distribution when the adversary can flip $\omega(\log n)$ bits.

In this chapter, we take a more general approach, and instead of considering specific perturbation sets, we provide methods in terms of oracle access to a separation oracle for the perturbation set $\mathcal{U}$, and aim to characterize which perturbation sets $\mathcal{U}$ admit tractable RERM.

In the non-realizable setting, the only prior work we are aware of is by Diakonikolas, Kane, and Manurangsi (2019b) who studied the complexity of robustly learning halfspaces in the agnostic setting under $\ell_2$ perturbations.

## 8.2 The Realizable Setting

Let $\mathcal{X} = \mathbb{R}^d$ be the instance space and $\mathcal{Y} = \{\pm 1\}$ be the label space. We consider halfspaces $\mathcal{H} = \{\boldsymbol{x} \mapsto \text{sign}(\langle \boldsymbol{w}, \boldsymbol{x} \rangle) : \boldsymbol{w} \in \mathbb{R}^d\}$. We refer the reader to Chapter 2 for a formal definition of adversarially robust PAC learning in the realizable setting (see Definition 2.2).

What is the sample complexity of robustly PAC learning halfspaces? Denote by $\mathcal{L}^{\mathcal{U}}_{\mathcal{H}}$ the robust loss class of $\mathcal{H}$,

$$\mathcal{L}^{\mathcal{U}}_{\mathcal{H}} = \left\{ (x, y) \mapsto \sup_{z \in \mathcal{U}(x)} 1[h(z) \neq y] : h \in \mathcal{H} \right\}.$$

Cullina, Bhagoji, and Mittal (2018b) showed that for any set $\mathcal{B} \subseteq \mathcal{X}$ that is nonempty, closed, convex, and origin-symmetric, and a perturbation set $\mathcal{U}$ that is defined as $\mathcal{U}(\boldsymbol{x}) = \boldsymbol{x} + \mathcal{B}$ (e.g., $\ell_p$-balls), the VC dimension of the robust loss of halfspaces $\mathrm{vc}(\mathcal{L}^{\mathcal{U}}_{\mathcal{H}})$ is at most the standard VC dimension $\mathrm{vc}(\mathcal{H}) = d + 1$. Based on Vapnik's "General Learning" Vapnik (1982), this implies that we have uniform convergence of robust risk with $m = \mathcal{O}(\frac{d+\log(1/\delta)}{\varepsilon^2})$ samples. Formally, for any $\varepsilon, \delta \in (0, 1)$ and any distribution $\mathcal{D}$ over $\mathcal{X} \times \mathcal{Y}$, with probability at least $1 - \delta$ over $S \sim \mathcal{D}^m$,

$$\forall \boldsymbol{w} \in \mathbb{R}^d, |\mathrm{R}_{\mathcal{U}}(h_{\boldsymbol{w}}; \mathcal{D}) - \mathrm{R}_{\mathcal{U}}(h_{\boldsymbol{w}}; S)| \leq \varepsilon. \tag{8.1}$$

In particular, this implies that for any perturbation set $\mathcal{U}$ that satisfies the conditions above, $\mathcal{H}$ is robustly PAC learnable w.r.t. $\mathcal{U}$ by minimizing the *robust* empirical risk on $S$,

$$\mathsf{RERM}_{\mathcal{U}}(S) = \operatorname*{argmin}_{\boldsymbol{w} \in \mathbb{R}^d} \frac{1}{m} \sum_{i=1}^{m} \sup_{\boldsymbol{z}_i \in \mathcal{U}(\boldsymbol{x}_i)} 1[y_i \langle \boldsymbol{w}, \boldsymbol{z}_i \rangle \leq 0]. \tag{8.2}$$

Thus, it remains to efficiently solve the $\mathsf{RERM}_{\mathcal{U}}$ problem. In the remainder of this section, we show necessary and sufficient conditions for minimizing the robust empirical risk $\mathsf{RERM}_{\mathcal{U}}$ on a dataset $S = \{(\boldsymbol{x}_1, y_1), \dots, (\boldsymbol{x}_m, y_m)\} \in (\mathcal{X} \times \mathcal{Y})^m$ in the realizable setting, i.e. when the dataset $S$ is *robustly* separable with a halfspace $h_{\boldsymbol{w}^*}$ where $\boldsymbol{w}^* \in \mathbb{R}^d$.

In Theorem 8.5, we show that an efficient separation oracle for $\mathcal{U}$ yields an efficient solver for $\mathsf{RERM}_{\mathcal{U}}$. While in Theorem 8.10, we show that an efficient *approximate* separation oracle for $\mathcal{U}$ is necessary for even computing the robust loss $\sup_{\boldsymbol{z} \in \mathcal{U}(\boldsymbol{x})} 1[h_{\boldsymbol{w}}(\boldsymbol{z}) \neq y]$ of a halfspace $h_{\boldsymbol{w}}$.

Note that the set of allowed perturbations $\mathcal{U}$ can be non-convex, and so it might seem difficult to imagine being able to solve the $\mathsf{RERM}_{\mathcal{U}}$ problem in full generality. But, it turns out that for halfspaces it suffices to consider only convex perturbation sets due to the following observation:

**Observation 8.1.** *Given a halfspace* $\boldsymbol{w} \in \mathbb{R}^d$ *and an example* $(\boldsymbol{x}, y) \in \mathcal{X} \times \mathcal{Y}$. *If* $\forall \boldsymbol{z} \in \mathcal{U}(\boldsymbol{x}), y\langle \boldsymbol{w}, \boldsymbol{z}\rangle > 0$, *then* $\forall z \in \mathsf{conv}(\mathcal{U}(\boldsymbol{x})), y\langle \boldsymbol{w}, \boldsymbol{z}\rangle > 0$. *And if* $\exists z \in \mathcal{U}(\boldsymbol{x}), y\langle \boldsymbol{w}, \boldsymbol{z}\rangle \leq 0$, *then* $\exists z \in \mathsf{conv}(\mathcal{U}(\boldsymbol{x})), y\langle \boldsymbol{w}, \boldsymbol{z}\rangle \leq 0$, *where* $\mathsf{conv}(\mathcal{U}(\boldsymbol{x}))$ *denotes the convex-hull of* $\mathcal{U}(\boldsymbol{x})$.

Observation 8.1 shows that for any dataset $S$ that is *robustly* separable w.r.t. $\mathcal{U}$ with a halfspace $\boldsymbol{w}^*$, $S$ is also robustly separable w.r.t. the convex hull $\mathsf{conv}(\mathcal{U})$ using the same halfspace $\boldsymbol{w}^*$, where $\mathsf{conv}(\mathcal{U})(x) = \mathsf{conv}(\mathcal{U}(x))$. Thus, in the remainder of this section we only consider perturbation sets $\mathcal{U}$ that are convex, i.e., for each $\boldsymbol{x}, \mathcal{U}(\boldsymbol{x})$ is convex.

**Definition 8.2.** *Denote by* $\mathsf{SEP}_{\mathcal{U}}$ *a separation oracle for* $\mathcal{U}$. $\mathsf{SEP}_{\mathcal{U}}(\boldsymbol{x}, \boldsymbol{z})$ *takes as input* $\boldsymbol{x}, \boldsymbol{z} \in \mathcal{X}$ *and either:*

- *asserts that* $\boldsymbol{z} \in \mathcal{U}(\boldsymbol{x})$, *or*
- *returns a separating hyperplane* $\boldsymbol{w} \in \mathbb{R}^d$ *such that* $\langle \boldsymbol{w}, \boldsymbol{z}'\rangle \leq \langle \boldsymbol{w}, \boldsymbol{z}\rangle$ *for all* $\boldsymbol{z}' \in \mathcal{U}(\boldsymbol{x})$.

**Definition 8.3.** *For any* $\eta > 0$, *denote by* $\mathsf{SEP}^{\eta}_{\mathcal{U}}$ *an* approximate *separation oracle for* $\mathcal{U}$. $\mathsf{SEP}^{\eta}_{\mathcal{U}}$ *takes as input* $\boldsymbol{x}, \boldsymbol{z} \in \mathcal{X}$ *and either:*

- *asserts that* $\boldsymbol{z} \in \mathcal{U}(\boldsymbol{x})^{+\eta} \stackrel{\text{def}}{=} \{\boldsymbol{z} : \exists \boldsymbol{z}' \in \mathcal{U}(\boldsymbol{x}) \text{ s.t. } \|\boldsymbol{z} - \boldsymbol{z}'\|_2 \leq \eta\}$, *or*
- *returns a separating hyperplane* $\boldsymbol{w} \in \mathbb{R}^d$ *such that* $\langle \boldsymbol{w}, \boldsymbol{z}'\rangle \leq \langle \boldsymbol{w}, \boldsymbol{z}\rangle + \eta$ *for all* $\boldsymbol{z}' \in \mathcal{U}(\boldsymbol{x})^{-\eta} \stackrel{\text{def}}{=} \{\boldsymbol{z}' : B(\boldsymbol{z}', \eta) \subseteq \mathcal{U}(\boldsymbol{x})\}$.

**Definition 8.4.** *Denote by* $\mathsf{MEM}_{\mathcal{U}}$ *a membership oracle for* $\mathcal{U}$. $\mathsf{MEM}_{\mathcal{U}}(\boldsymbol{x}, \boldsymbol{z})$ *takes as input* $\boldsymbol{x}, \boldsymbol{z} \in \mathcal{X}$ *and either:*

- *asserts that* $\boldsymbol{z} \in \mathcal{U}(\boldsymbol{x})$, *or*

- *asserts that* $\boldsymbol{z} \notin \mathcal{U}(\boldsymbol{x})$.

When discussing a separation or membership oracle for a fixed convex set $K$, we overload notation and write $\mathsf{SEP}_K$, $\mathsf{SEP}^{\eta}_K$, and $\mathsf{MEM}_K$ (in this case only one argument is required).

### 8.2.1 An efficient separation oracle for $\mathcal{U}$ is sufficient to solve $\mathsf{RERM}_{\mathcal{U}}$ efficiently

Let $\mathsf{Soln}^{\mathcal{U}}_S = \{\boldsymbol{w} \in \mathbb{R}^d : \forall(\boldsymbol{x}, y) \in S, \forall \boldsymbol{z} \in \mathcal{U}(\boldsymbol{x}), y\langle \boldsymbol{w}, \boldsymbol{z}\rangle > 0\}$ denote the set of valid solutions for $\mathsf{RERM}_{\mathcal{U}}(S)$ (see Equation 8.2). Note that $\mathsf{Soln}^{\mathcal{U}}_S$ is not empty since we are considering the realizable setting. Although the treatment we present here is for homogeneous halfspaces (where a bias term is not needed), the results extend trivially to the non-homogeneous case.

Below, we show that we can efficiently find a solution $\boldsymbol{w} \in \mathsf{Soln}^{\mathcal{U}}_S$ given access to a separation oracle for $\mathcal{U}$, $\mathsf{SEP}_{\mathcal{U}}$.

**Theorem 8.5.** *Let $\mathcal{U}$ be an arbitrary convex perturbation set. Given access to a separation oracle for $\mathcal{U}$, $\mathsf{SEP}_{\mathcal{U}}$ that runs in time $\mathsf{poly}(d, b)$. There is an algorithm that finds $\boldsymbol{w} \in \mathsf{Soln}^{\mathcal{U}}_S$ in $\mathsf{poly}(m, d, b)$ time where $b$ is an upper bound on the bit complexity of the valid solutions in $\mathsf{Soln}^{\mathcal{U}}_S$ and the examples and perturbations in $S$.*

Note that the polynomial dependence on $b$ in the runtime is unavoidable even in standard non-robust ERM for halfspaces, unless we can solve linear programs in strongly polynomial time, which is currently an open problem.

Theorem 8.5 implies that for a broad family of perturbation sets $\mathcal{U}$, halfspaces $\mathcal{H}$ are efficiently robustly PAC learnable with respect to $\mathcal{U}$ in the realizable setting, as we show in the following corollary:

**Corollary 8.6.** *Let $\mathcal{U} : \mathcal{X} \to 2^{\mathcal{X}}$ be a perturbation set such that $\mathcal{U}(\boldsymbol{x}) = \boldsymbol{x} + \mathcal{B}$ where $\mathcal{B}$ is nonempty, closed, convex, and origin-symmetric. Then, given access to an efficient separation oracle $\mathsf{SEP}_{\mathcal{U}}$ that runs in time $\mathsf{poly}(d, b)$, $\mathcal{H}$ is robustly PAC learnable w.r.t. $\mathcal{U}$ in the realizable setting in time $\mathsf{poly}(d, b, 1/\varepsilon, \log(1/\delta))$.*

*Proof.* This follows from the uniform convergence guarantee for the robust risk of halfspaces (see Equation (8.1)) and Theorem 8.5. □

This covers many types of perturbation sets that are considered in practice. For example, $\mathcal{U}$ could be perturbations of distance at most $\gamma$ w.r.t. some norm $\|\cdot\|$, such as the $\ell_\infty$ norm considered in many applications: $\mathcal{U}(\boldsymbol{x}) = \{\boldsymbol{z} \in \mathcal{X} : \|\boldsymbol{x} - \boldsymbol{z}\|_\infty \leq \gamma\}$. In addition, Theorem 8.5 also implies that we can solve the RERM problem for other natural perturbation sets such as translations and rotations in images (see, e.g., Engstrom et al. (2019)), and perhaps mixtures of perturbations of different types (see, e.g., Kang et al. (2019a)), as long we have access to efficient separation oracles for these sets.

Benefits of handling general perturbation sets $\mathcal{U}$: One important implication of Theorem 8.5 that highlights the importance of having a treatment that considers general perturbation sets (and not just $\ell_p$ perturbations for example) is the following: for any efficiently computable feature map $\phi : \mathbb{R}^r \to \mathbb{R}^d$, we can efficiently solve the robust empirical risk problem over the induced halfspaces $\mathcal{H}_\phi = \left\{\boldsymbol{x} \mapsto \text{sign}(\langle \boldsymbol{w}, \phi(\boldsymbol{x})\rangle) : \boldsymbol{w} \in \mathbb{R}^d\right\}$, as long as we have access to an efficient separation oracle for the image of the perturbations $\phi(\mathcal{U}(\boldsymbol{x}))$. Observe that in general $\phi(\mathcal{U}(\boldsymbol{x}))$ maybe non-convex and complicated even if $\mathcal{U}(\boldsymbol{x})$ is convex, however Observation 8.1 combined with the realizability assumption imply that it suffices to have an efficient separation oracle for the convex-hull $\mathsf{conv}(\phi(\mathcal{U}(\boldsymbol{x})))$.

Before we proceed with the proof of Theorem 8.5, we state the following requirements and guarantees for the Ellipsoid method which will be useful for us in the remainder of the section:

**Lemma 8.7** (see, e.g., Theorem 2.4 in Bubeck et al. (2015))**.** *Let $K \subseteq \mathbb{R}^d$ be a convex set, and $\mathsf{SEP}_K$ a separation oracle for $K$. Then, the Ellipsoid method using $\mathcal{O}(d^2 b)$ oracle queries to $\mathsf{SEP}_K$, will find a $\boldsymbol{w} \in K$, or assert that $K$ is empty. Furthermore, the total runtime is $\mathcal{O}(d^4 b)$.*

The proof of Theorem 8.5 relies on two key lemmas. First, we show that efficient robust certification yields an efficient solver for the RERM problem. Given a halfspace $\boldsymbol{w} \in \mathbb{R}^d$ and an example $(\boldsymbol{x}, y) \in \mathbb{R}^d \times \mathcal{Y}$, efficient robust certification means that there is an algorithm that can *efficiently* either: (a) assert that $\boldsymbol{w}$ is robust on $\mathcal{U}(\boldsymbol{x})$, i.e. $\forall \boldsymbol{z} \in \mathcal{U}(\boldsymbol{x}), y\langle \boldsymbol{w}, \boldsymbol{z}\rangle > 0$, or (b) return a perturbation $\boldsymbol{z} \in \mathcal{U}(\boldsymbol{x})$ such that $y\langle \boldsymbol{w}, \boldsymbol{z}\rangle \leq 0$.

**Lemma 8.8.** *Let* $\mathsf{CERT}_{\mathcal{U}}(\boldsymbol{w}, (\boldsymbol{x}, y))$ *be a procedure that either: (a) Asserts that* $\boldsymbol{w}$ *is robust on* $\mathcal{U}(\boldsymbol{x})$*, i.e.,* $\forall \boldsymbol{z} \in \mathcal{U}(\boldsymbol{x}), y\langle \boldsymbol{w}, \boldsymbol{z}\rangle > 0$*, or (b) Finds a perturbation* $\boldsymbol{z} \in \mathcal{U}(\boldsymbol{x})$ *such that* $y\langle \boldsymbol{w}, \boldsymbol{z}\rangle \leq 0$*. If* $\mathsf{CERT}_{\mathcal{U}}(\boldsymbol{w}, (\boldsymbol{x}, \boldsymbol{y}))$ *can be solved in* $\mathsf{poly}(d, b)$ *time, then there is an algorithm that finds* $\boldsymbol{w} \in \mathsf{Soln}_S^{\mathcal{U}}$ *in* $\mathsf{poly}(m, d, b)$ *time.*

*Proof.* Observe that $\mathsf{Soln}_S^{\mathcal{U}}$ is a convex set since

$$\begin{aligned}
\boldsymbol{w}_1, \boldsymbol{w}_2 \in \mathsf{Soln}_S^{\mathcal{U}} &\Rightarrow \forall (\boldsymbol{x}, y) \in S, \forall z \in \mathcal{U}(\boldsymbol{x}), y\langle \boldsymbol{w}_1, \boldsymbol{z}\rangle > 0 \\
&\text{and } y\langle \boldsymbol{w}_2, \boldsymbol{z}\rangle > 0 \\
&\Rightarrow \forall \alpha \in [0,1], \forall (\boldsymbol{x}, y) \in S, \forall \boldsymbol{z} \in \mathcal{U}(\boldsymbol{x}), \\
&y\langle \alpha \boldsymbol{w}_1 + (1-\alpha)\boldsymbol{w}_2, \boldsymbol{z}\rangle > 0 \\
&\Rightarrow \forall \alpha \in [0,1], \alpha \boldsymbol{w}_1 + (1-\alpha)\boldsymbol{w}_2 \in \mathsf{Soln}_S^{\mathcal{U}}.
\end{aligned}$$

Our goal is to find a $\boldsymbol{w} \in \mathsf{Soln}_S^{\mathcal{U}}$. Let $\mathsf{CERT}_{\mathcal{U}}(\boldsymbol{w}, (\boldsymbol{x}, y))$ be an efficient robust certifier that runs in $\mathsf{poly}(d, b)$ time. We will use $\mathsf{CERT}_{\mathcal{U}}(\boldsymbol{w}, (\boldsymbol{x}, y))$ to implement a separation oracle for $\mathsf{Soln}_S^{\mathcal{U}}$ denoted $\mathsf{SEP}_{\mathsf{Soln}_S^{\mathcal{U}}}$. Given a halfspace $\boldsymbol{w} \in \mathbb{R}^d$, we simply check if $\boldsymbol{w}$ is robustly correct on all datapoints by running $\mathsf{CERT}_{\mathcal{U}}(\boldsymbol{w}, (\boldsymbol{x}_i, y_i))$ on each $(\boldsymbol{x}_i, y_i) \in S$. If there is a point $(\boldsymbol{x}_i, y_i) \in S$ where $\boldsymbol{w}$ is not robustly correct, then we get a perturbation $\boldsymbol{z}_i \in \mathcal{U}(\boldsymbol{x}_i)$ where $y_i\langle \boldsymbol{w}, \boldsymbol{z}_i\rangle \leq 0$, and we return $-y_i\boldsymbol{z}_i$ as a separating hyperplane. Otherwise, we know that $\boldsymbol{w}$ is robustly correct on all datapoints, and we just assert that $\boldsymbol{w} \in \mathsf{Soln}_S^{\mathcal{U}}$.

Once we have a separation oracle $\mathsf{SEP}_{\mathsf{Soln}_S^{\mathcal{U}}}$, we can use the Ellipsoid method (see Lemma 8.7) to solve the $\mathsf{RERM}_{\mathcal{U}}(S)$ problem. More specifically, with a query complexity of $\mathcal{O}(d^2 b)$ to $\mathsf{SEP}_{\mathsf{Soln}_S^{\mathcal{U}}}$ and overall runtime of $\mathsf{poly}(m, d, b)$ (this depends on runtime of $\mathsf{CERT}_{\mathcal{U}}(\boldsymbol{w}, (\boldsymbol{x}, y))$), the Ellipsoid method will return a $\boldsymbol{w} \in \mathsf{Soln}_S^{\mathcal{U}}$. □

Next, we show that we can do efficient robust certification when given access to an efficient separation oracle for $\mathcal{U}$, $\mathsf{SEP}_{\mathcal{U}}$.

**Lemma 8.9.** *If we have an efficient separation oracle* $\mathsf{SEP}_{\mathcal{U}}$ *that runs in* $\mathsf{poly}(d, b)$ *time. Then, we can efficiently solve* $\mathsf{CERT}_{\mathcal{U}}(\boldsymbol{w}, (\boldsymbol{x}, y))$ *in* $\mathsf{poly}(d, b)$ *time.*

*Proof.* Given a halfspace $\boldsymbol{w} \in \mathbb{R}^d$ and $(\boldsymbol{x}, y) \in \mathbb{R}^d \times \mathcal{Y}$, we want to either: (a) assert that $\boldsymbol{w}$ is robust on $\mathcal{U}(\boldsymbol{x})$, i.e. $\forall \boldsymbol{z} \in \mathcal{U}(\boldsymbol{x}), y\langle \boldsymbol{w}, \boldsymbol{z}\rangle > 0$, or (b) find a perturbation $\boldsymbol{z} \in \mathcal{U}(\boldsymbol{x})$ such that $y\langle \boldsymbol{w}, \boldsymbol{z}\rangle \leq 0$. Let $\mathcal{M}(\boldsymbol{w}, y) = \{\boldsymbol{z}' \in \mathcal{X} : y\langle \boldsymbol{w}, \boldsymbol{z}'\rangle \leq 0\}$ be the set of all points that $\boldsymbol{w}$ mis-labels. Observe that by definition $\mathcal{M}(\boldsymbol{w}, y)$ is convex, and therefore $\mathcal{U}(\boldsymbol{x}) \cap \mathcal{M}(\boldsymbol{w}, y)$ is also convex. We argue that having an efficient separation oracle for $\mathcal{U}(\boldsymbol{x}) \cap \mathcal{M}(\boldsymbol{w}, y)$ suffices to solve our robust certification problem. Because if $\mathcal{U}(\boldsymbol{x}) \cap \mathcal{M}(\boldsymbol{w}, y)$ is not empty, then by definition, we can find a perturbation $\boldsymbol{z} \in \mathcal{U}(\boldsymbol{x})$ such that $y \langle \boldsymbol{w}, \boldsymbol{z} \rangle \leq 0$ with a separation oracle $\mathsf{SEP}_{\mathcal{U}(\boldsymbol{x}) \cap \mathcal{M}(\boldsymbol{w},y)}$ and the Ellipsoid method (see Lemma 8.7). If $\mathcal{U}(\boldsymbol{x}) \cap \mathcal{M}(\boldsymbol{w}, y)$ is empty, then by definition, $\boldsymbol{w}$ is robustly correct on $\mathcal{U}(\boldsymbol{x})$, and the Ellipsoid method will terminate and assert that $\mathcal{U}(\boldsymbol{x}) \cap \mathcal{M}(\boldsymbol{w}, y)$ is empty.

Thus, it remains to implement $\mathsf{SEP}_{\mathcal{U}(\boldsymbol{x}) \cap \mathcal{M}(\boldsymbol{w},y)}$. Given a point $\boldsymbol{z} \in \mathbb{R}^d$, we simply ask the separation oracle for $\mathcal{U}(\boldsymbol{x})$ by calling $\mathsf{SEP}_{\mathcal{U}}(\boldsymbol{x}, \boldsymbol{z})$ and the separation oracle for $\mathcal{M}(\boldsymbol{w}, y)$ by checking if $y \langle \boldsymbol{w}, \boldsymbol{z} \rangle \leq 0$. If $\boldsymbol{z} \notin \mathcal{U}(\boldsymbol{x})$ the we get a separating hyperplane $\boldsymbol{c}$ from $\mathsf{SEP}_{\mathcal{U}}$ and we can use it separate $\boldsymbol{z}$ from $\mathcal{U}(\boldsymbol{x}) \cap \mathcal{M}(\boldsymbol{w}, y)$. Similarly, if $\boldsymbol{z} \notin \mathcal{M}(\boldsymbol{w}, y)$, by definition, $\langle y\boldsymbol{w}, \boldsymbol{z} \rangle > 0$ and so we can use $y\boldsymbol{w}$ as a separating hyperplane to separate $\boldsymbol{z}$ from $\mathcal{U}(\boldsymbol{x}) \cap \mathcal{M}(\boldsymbol{w}, y)$. The overall runtime of this separation oracle is $\mathsf{poly}(d, b)$, and so we can efficiently solve $\mathsf{CERT}_{\mathcal{U}}(\boldsymbol{w}, (\boldsymbol{x}, y))$ in $\mathsf{poly}(d, b)$ time using the Ellipsoid method (Lemma 8.7). □

We are now ready to proceed with the proof of Theorem 8.5.

*Proof of Theorem 8.5.* We want to efficiently solve $\mathsf{RERM}_{\mathcal{U}}(S)$. Given that we have a separation oracle for $\mathcal{U}$, $\mathsf{SEP}_{\mathcal{U}}$ that runs in $\mathsf{poly}(d, b)$. By Lemma 8.9, we get an efficient robust certification procedure $\mathsf{CERT}_{\mathcal{U}}(\boldsymbol{w}, (\boldsymbol{x}, y))$. Then, by Lemma 8.8, we get an efficient solver for $\mathsf{RERM}_{\mathcal{U}}$. In particular, the runtime complexity is $\mathsf{poly}(m, d, b)$. □

### 8.2.2 An efficient approximate separation oracle for $\mathcal{U}$ is necessary for computing the robust loss

Our efficient algorithm for $\mathsf{RERM}_{\mathcal{U}}$ requires a separation oracle for $\mathcal{U}$. We now show that even efficiently computing the robust loss of a halfspace $(\boldsymbol{w}, b_0) \in \mathbb{R}^d \times \mathbb{R}$ on an example $(\boldsymbol{x}, y) \in \mathbb{R}^d \times \mathcal{Y}$ requires an efficient *approximate* separation oracle for $\mathcal{U}$.

**Theorem 8.10.** *Given a halfspace* $\boldsymbol{w} \in \mathbb{R}^d$ *and an example* $(\boldsymbol{x}, y) \in \mathbb{R}^d \times \mathcal{Y}$*, let* $\mathsf{EVAL}_{\mathcal{U}}((\boldsymbol{w}, b_0), (\boldsymbol{x}, y))$ *be a procedure that computes the robust loss* $\sup_{\boldsymbol{z} \in \mathcal{U}(\boldsymbol{x})} \mathbb{1}[y(\langle \boldsymbol{w}, \boldsymbol{z} \rangle + b_0) \leq 0]$ *in* $\mathsf{poly}(d, b)$ *time, then for any* $\gamma > 0$*, we can implement an efficient* $\gamma$*-approximate separation oracle* $\mathsf{SEP}^{\gamma}_{\mathcal{U}}(\boldsymbol{x}, \boldsymbol{z})$ *in* $\mathsf{poly}(d, b, \log(1/\gamma), \log(R))$ *time, where* $\mathcal{U}(\boldsymbol{x}) \subseteq B(0, R)$.

*Proof.* Let $\gamma > 0$. We will describe how to implement a $\gamma$-approximate separation oracle for $\mathcal{U}$ denoted $\mathsf{SEP}^{\gamma}_{\mathcal{U}}(\boldsymbol{x}, \boldsymbol{z})$. Fix the first argument to an arbitrary $\boldsymbol{x} \in \mathcal{X}$. Upon receiving a point $\boldsymbol{z} \in \mathcal{X}$ as input, the main strategy is to search for a halfspace $\boldsymbol{w} \in \mathbb{R}^d$ that can label all of $\mathcal{U}(\boldsymbol{x})$ with $+1$, and label the point $\boldsymbol{z}$ with $-1$. If $\boldsymbol{z} \notin \mathcal{U}(\boldsymbol{x})$ then there is a halfspace $\boldsymbol{w}$ that separates $\boldsymbol{z}$ from $\mathcal{U}(\boldsymbol{x})$ because $\mathcal{U}(\boldsymbol{x})$ is convex, but this is impossible if $\boldsymbol{z} \in \mathcal{U}(\boldsymbol{x})$. Since we are only concerned with implementing an *approximate* separation oracle, we will settle for a slight relaxation which is to either:

- assert that $\boldsymbol{z}$ is $\gamma$-close to $\mathcal{U}(\boldsymbol{x})$, i.e., $\boldsymbol{z} \in B(\mathcal{U}(\boldsymbol{x}), \gamma)$, or
- return a separating hyperplane $\boldsymbol{w}$ such that $\langle \boldsymbol{w}, \boldsymbol{z}' \rangle \leq \langle \boldsymbol{w}, \boldsymbol{z} \rangle$ for all $\boldsymbol{z}' \in \mathcal{U}(\boldsymbol{x})$.

Let $K = \{(\boldsymbol{w}, b_0) : \forall \boldsymbol{z}' \in \mathcal{U}(\boldsymbol{x}), \langle \boldsymbol{w}, \boldsymbol{z}' \rangle + b_0 > 0\}$ denote the set of halfspaces that label all of $\mathcal{U}(\boldsymbol{x})$ with $+1$. Since $\mathcal{U}(\boldsymbol{x})$ is nonempty, it follows by definition that $K$ is nonempty. To evaluate membership in $K$, given a query $\boldsymbol{w}_q, b_q$, we just make a call to $\mathsf{EVAL}_{\mathcal{U}}((\boldsymbol{w}_q, b_q), (\boldsymbol{x}, +))$. Let $\mathsf{MEM}_K(\boldsymbol{w}_q, b_q) = 1 - \mathsf{EVAL}_{\mathcal{U}}((\boldsymbol{w}_q, b_q), (\boldsymbol{x}, +))$. This can be efficiently computed in $\mathsf{poly}(d, b)$ time. Next, for any $\eta \in (0, 0.5)$, we can get an $\eta$-approximate separation oracle for $K$ denoted $\mathsf{SEP}^{\eta}_K$ (see Definition 8.3) using $\mathcal{O}(db \log(d/\eta))$ queries to the membership oracle $\mathsf{MEM}_K$ described above Lee et al. (2018). When queried with a halfspace $\tilde{\boldsymbol{w}} = (\boldsymbol{w}, b_0)$, $\mathsf{SEP}^{\eta}_K$ either:

- asserts that $\tilde{\boldsymbol{w}} \in K^{+\eta}$, or
- returns a separating hyperplane $\boldsymbol{c}$ such that $\langle \boldsymbol{c}, \tilde{\boldsymbol{w}}' \rangle \leq \langle \boldsymbol{c}, \tilde{\boldsymbol{w}} \rangle + \eta$ for all halfspaces $\tilde{\boldsymbol{w}}' \in K^{-\eta}$.

Observe that by definition, $K^{-\eta} \subseteq K \subseteq K^{+\eta}$. Furthermore, for any $\boldsymbol{w} \in K^{+\eta}$, by definition, $\exists \tilde{\boldsymbol{w}}' \in K$ such that $\|\tilde{\boldsymbol{w}} - \tilde{\boldsymbol{w}}'\|_2 \leq \eta$. Since, for each $\boldsymbol{z}' \in \mathcal{U}(\boldsymbol{x})$, by definition of $K$, we have $\langle \tilde{\boldsymbol{w}}', (\boldsymbol{z}', 1) \rangle = \langle \boldsymbol{w}', \boldsymbol{z}' \rangle + b_0 > 0$, it follows by Cauchy-Schwarz inequality that $(\forall \tilde{\boldsymbol{w}} \in K^{+\eta})\ (\forall \boldsymbol{z}' \in \mathcal{U}(\boldsymbol{x}))$:

$$\langle \tilde{\boldsymbol{w}}, (\boldsymbol{z}', 1) \rangle = \langle \tilde{\boldsymbol{w}} - \tilde{\boldsymbol{w}}', \boldsymbol{z}' \rangle + \langle \tilde{\boldsymbol{w}}', (\boldsymbol{z}', 1) \rangle > -\eta 2R. \tag{8.3}$$

Let $\mathsf{SEP}_K^{\gamma/4R}$ be a $\frac{\gamma}{4R}$-approximate separation oracle for $K$. Observe that if the distance between $\boldsymbol{z}$ and $\mathcal{U}(\boldsymbol{x})$ is greater than $\gamma$, it follows that there is $(\boldsymbol{w}, b_0)$ such that:

$$\langle \boldsymbol{w}, \boldsymbol{z} \rangle + b_0 \leq -\gamma/2 \text{ and } \langle \boldsymbol{w}, \boldsymbol{z}' \rangle + b_0 > 0 \, (\forall \boldsymbol{z}' \in \mathcal{U}(\boldsymbol{x})) \, .$$

By definition of $K$, this implies that $K \cap \{(\boldsymbol{w}, b_0) : \langle \boldsymbol{w}, \boldsymbol{z} \rangle + b_0 \leq -\gamma/2\}$ is not empty, which implies that the intersection $K^{+\frac{\gamma}{4R}} \cap \{(\boldsymbol{w}, b_0) : \langle \boldsymbol{w}, \boldsymbol{z} \rangle + b_0 \leq -\gamma/2\}$ is nonempty. We also have the contrapositive, which is, if the intersection $K^{+\frac{\gamma}{4R}} \cap \{(\boldsymbol{w}, b_0) : \langle \boldsymbol{w}, \boldsymbol{z} \rangle + b_0 \leq -\gamma/2\}$ is empty, then we know that $\boldsymbol{z} \in B(\mathcal{U}(\boldsymbol{x}), \gamma)$. To conclude the proof, we run the Ellipsoid method with the approximate separation oracle $\mathsf{SEP}_K^{\gamma/4R}$ to search over the restricted space $\{(\boldsymbol{w}, b_0) : \langle \boldsymbol{w}, \boldsymbol{z} \rangle + b_0 \leq -\gamma/2\}$. Restricting the space is easily done because we will use the query point $\boldsymbol{z}$ as the separating hyperplane. Either the Ellipsoid method will find $(\boldsymbol{w}, b_0) \in K^{+\frac{\gamma}{4R}} \cap \{(\boldsymbol{w}, b_0) : \langle \boldsymbol{w}, \boldsymbol{z} \rangle + b_0 \leq -\gamma/2\}$, in which case by Equation 8.3, $(\boldsymbol{w}, b_0)$ has the property that:

$$\langle \boldsymbol{w}, \boldsymbol{z} \rangle + b_0 \leq -\frac{\gamma}{2} \text{ and } \langle \boldsymbol{w}, \boldsymbol{z}' \rangle + b_0 > -\frac{\gamma}{2} \, (\forall \boldsymbol{z}' \in \mathcal{U}(\boldsymbol{x})) \, ,$$

and so we return $\boldsymbol{w}$ as a separating hyperplane between $\boldsymbol{z}$ and $\mathcal{U}(\boldsymbol{x})$. If the Ellipsoid terminates without finding any such $(\boldsymbol{w}, b_0)$, this implies that the intersection $K^{+\frac{\gamma}{4R}} \cap \{(\boldsymbol{w}, b_0) : \langle \boldsymbol{w}, \boldsymbol{z} \rangle + b_0 \leq -\gamma/2\}$ is empty, and therefore, by the contrapositive above, we assert that $\boldsymbol{z} \in B(\mathcal{U}(\boldsymbol{x}), \gamma)$. □

## 8.3 Random Classification Noise

In this section, we relax the realizability assumption to random classification noise (Angluin and Laird, 1987). The following definition formalizes adversarially robust learning in the random classification noise setting:

**Definition 8.11** (Robust PAC Learning with Random Classification Noise). *Let $h^* \in \mathcal{H}$ be an unknown halfspace. Let $\mathcal{D}_{\boldsymbol{x}}$ be an arbitrary distribution over $\mathcal{X}$ such that $\mathrm{R}_{\mathcal{U}}(h^*; \mathcal{D}_{h^*}) = 0$, and $\eta \leq 0 < 1/2$. A noisy example oracle, $\mathrm{EX}(h^*, \mathcal{D}_{\boldsymbol{x}}, \eta)$ works as follows: Each time $\mathrm{EX}(h^*, \mathcal{D}_{\boldsymbol{x}}, \eta)$ is invoked, it returns a labeled example $(\boldsymbol{x}, y)$, where $\boldsymbol{x} \sim \mathcal{D}_{\boldsymbol{x}}$, $y = h^*(x)$ with probability $1 - \eta$ and $y = -h^*(x)$ with probability $\eta$. Let $\mathcal{D}$ be the joint distribution on $(\boldsymbol{x}, y)$ generated by the above oracle.*

*We say $\mathcal{H}$ is robustly PAC learnable with respect to a perturbation set $\mathcal{U}$ in the* random classification noise *model, if $\exists m(\varepsilon, \delta, \eta) \in \mathbb{N} \cup \{0\}$ and a learning algorithm $\mathbb{A} : (\mathcal{X} \times \mathcal{Y})^* \to \mathcal{Y}^{\mathcal{X}}$, such that for every distribution $\mathcal{D}$ over $\mathcal{X} \times \mathcal{Y}$ (generated as above by a noisy oracle), with probability at least $1 - \delta$ over $S \sim \mathcal{D}^m$,*

$$\mathrm{R}_{\mathcal{U}}(\mathbb{A}(S); \mathcal{D}) \leq \eta + \varepsilon.$$

We show that for any perturbation set $\mathcal{U}$ that represents perturbations of bounded norm (i.e., $\mathcal{U}(x) = x + \mathcal{B}$, where $\mathcal{B} = \left\{\delta \in \mathbb{R}^d : \|\delta\|_p \leq \gamma\right\}, p \in [1, \infty]$), the class of halfspaces $\mathcal{H}$ is efficiently robustly PAC learnable with respect to $\mathcal{U}$ in the random classification noise model.

**Theorem 8.12.** *Let $\mathcal{U} : \mathcal{X} \to 2^{\mathcal{X}}$ be a perturbation set such that $\mathcal{U}(\boldsymbol{x}) = \boldsymbol{x} + \mathcal{B}$ where $\mathcal{B} = \left\{\delta \in \mathbb{R}^d : \|\delta\|_p \leq \gamma\right\}$ and $p \in [1, \infty]$. Then, $\mathcal{H}$ is robustly PAC learnable w.r.t $\mathcal{U}$ under random classification noise in time* $\mathsf{poly}(d, 1/\varepsilon, 1/\gamma, 1/(1-2\eta), \log(1/\delta))$.

The proof of Theorem 8.12 relies on the following key lemma. We show that the structure of the perturbations $\mathcal{B}$ allows us to relate the robust loss of a halfspace $\boldsymbol{w} \in \mathbb{R}^d$ with the $\gamma$-margin loss of $\boldsymbol{w}$. Before we state the lemma, recall that the dual norm of $\boldsymbol{w}$ denoted $\|\boldsymbol{w}\|_*$ is defined as $\sup\left\{\langle \boldsymbol{u}, \boldsymbol{w}\rangle : \|\boldsymbol{u}\| \leq 1\right\}$.

**Lemma 8.13.** *For any $\boldsymbol{w}, \boldsymbol{x} \in \mathbb{R}^d$ and any $y \in \mathcal{Y}$,*

$$\sup_{\delta \in \mathcal{B}} 1\{h_{\boldsymbol{w}}(\boldsymbol{x} + \delta) \neq y\} = 1\left\{y \left\langle \frac{\boldsymbol{w}}{\|\boldsymbol{w}\|_*}, \boldsymbol{x} \right\rangle \leq \gamma\right\}.$$

*Proof.* First observe that

$$\begin{aligned}\sup_{\delta \in \mathcal{B}} 1\{h_{\boldsymbol{w}}(\boldsymbol{x} + \delta) \neq y\} &= \sup_{\delta \in \mathcal{B}} 1\{y \langle \boldsymbol{w}, \boldsymbol{x} + \delta\rangle \leq 0\} \\ &= 1\left\{\inf_{\delta \in \mathcal{B}} y \langle \boldsymbol{w}, \boldsymbol{x} + \delta\rangle \leq 0\right\}.\end{aligned}$$

This holds because when $\inf_{\delta \in \mathcal{B}} y \langle \boldsymbol{w}, \boldsymbol{x} + \delta\rangle > 0$, by definition $\forall \delta \in \mathcal{B}, y \langle \boldsymbol{w}, \boldsymbol{x} + \delta\rangle > 0$, which implies that $\sup_{\delta \in \mathcal{B}} 1\{h_{\boldsymbol{w}}(\boldsymbol{x} + \delta) \neq y\} = 0$. For the other direction, when $\sup_{\delta \in \mathcal{B}} 1\{h_{\boldsymbol{w}}(\boldsymbol{x} + \delta) \neq y\} = 1$, by definition $\exists \delta \in \mathcal{B}$ such that $y \langle \boldsymbol{w}, \boldsymbol{x} + \delta\rangle \leq 0$, which implies that $\inf_{\delta \in \mathcal{B}} y \langle \boldsymbol{w}, \boldsymbol{x} + \delta\rangle \leq$

0. To conclude the proof, by definition of the set $\mathcal{B}$ and the dual norm $\|\cdot\|_*$, we have

$$\begin{aligned}\inf_{\partial \in \mathcal{B}} y \langle \boldsymbol{w}, \boldsymbol{x} + \partial \rangle &= y \langle \boldsymbol{w}, \boldsymbol{x} \rangle - \sup_{\partial \in \mathcal{B}} \langle -y\boldsymbol{w}, \partial \rangle \\ &= y \langle \boldsymbol{w}, \boldsymbol{x} \rangle - \|\boldsymbol{w}\|_* \gamma.\end{aligned}$$

□

Theorem 8.13 implies that for any distribution $\mathcal{D}$ over $\mathcal{X} \times \mathcal{Y}$, to solve the $\gamma$-robust learning problem

$$\operatorname*{argmin}_{w \in \mathbb{R}^d} \mathop{\mathbb{E}}_{(\boldsymbol{x},y)\sim\mathcal{D}} \left[ \sup_{\partial \in \mathcal{B}} 1\{h_{\boldsymbol{w}}(\boldsymbol{x} + \partial) \neq y\} \right], \tag{8.4}$$

it suffices to solve the $\gamma$-margin learning problem

$$\operatorname*{argmin}_{\|\boldsymbol{w}\|_*=1} \mathop{\mathbb{E}}_{(\boldsymbol{x},y)\sim\mathcal{D}} \left[ 1\{y \langle \boldsymbol{w}, \boldsymbol{x} \rangle \leq \gamma\} \right]. \tag{8.5}$$

We will solve the $\gamma$-margin learning problem in Equation (8.5) in the random classification noise setting using an appropriately chosen convex surrogate loss. Our convex surrogate loss and its analysis build on a convex surrogate that appears in the appendix of Diakonikolas et al. (2019a) for learning large $\ell_2$-margin halfspaces under random classification noise w.r.t. the 0-1 loss. We note that the idea of using a convex surrogate to (non-robustly) learn large margin halfspaces in the presence of random classification noise is implicit in a number of prior works, starting with Bylander (1994).

Our robust setting is more challenging for the following reasons. First, we are not interested in only ensuring small 0-1 loss, but rather ensuring small $\gamma$-margin loss. Second, we want to be able to handle all $\ell_p$ norms, as opposed to just the $\ell_2$ norm. As a result, our analysis is somewhat delicate.

Let

$$\varphi(s) = \begin{cases} \lambda(1 - \frac{s}{\gamma}), & s > \gamma \\ (1-\lambda)(1 - \frac{s}{\gamma}), & s \leq \gamma \end{cases}.$$

We will show that solving the following convex optimization problem:

$$\operatorname*{argmin}_{\|\boldsymbol{w}\|_* \le 1} G_\lambda^\gamma(\boldsymbol{w}) \stackrel{\text{def}}{=} \mathop{\mathbb{E}}_{(\boldsymbol{x},y)\sim\mathcal{D}} \left[\varphi(y \langle \boldsymbol{w}, \boldsymbol{x}\rangle)\right] \ , \tag{8.6}$$

where $\lambda = \frac{\varepsilon\gamma/2+\eta}{1+\varepsilon\gamma}$, suffices to solve the $\gamma$-margin learning problem in Equation (8.5). Intuitively, the idea here is that for $\lambda = 0$ , the $\varphi$ objective is exactly a scaled hinge loss, which gives a learning guarantee w.r.t to the $\gamma$-margin loss when there is no noise ($\eta = 0$). When the noise $\eta > 0$, we slightly adjust the slopes, such that even correct prediction encounters a loss. The choice of the slope is based on $\lambda$ which will depend on the noise rate $\eta$ and the $\varepsilon$-suboptimality that is required for Equation (8.5).

We can solve Equation (8.6) with a standard first-order method through samples using Stochastic Mirror Descent, when the dual norm $\|\cdot\|_*$ is an $\ell_q$-norm. We state the following properties of Mirror Descent we will require:

**Lemma 8.14** (see, e.g., Theorem 6.1 in Bubeck et al. (2015)). *Let $G(\boldsymbol{w}) \stackrel{\text{def}}{=} \mathbb{E}_{(\boldsymbol{x},y)\sim\mathcal{D}}\left[\ell(\boldsymbol{w},(\boldsymbol{x},y))\right]$ be a convex function that is L-Lipschitz w.r.t. $\|\boldsymbol{w}\|_q$ where $q > 1$. Then, using the potential function $\psi(\boldsymbol{w}) = \frac{1}{2}\|\boldsymbol{w}\|_q^2$, a suitable step-size $\eta$, and a sequence of iterates $\boldsymbol{w}^k$ computed by the following update:*

$$\boldsymbol{w}^{k+1} = \Pi_{B_q}^{\psi}\left(\nabla\psi^{-1}\left(\nabla\psi(\boldsymbol{w}^k) - \eta \boldsymbol{g}^k\right)\right) \ ,$$

*Stochastic Mirror Descent with $\mathcal{O}(L^2/(q-1)\varepsilon^2)$ stochastic gradients $\boldsymbol{g}$ of G, will find an $\varepsilon$-suboptimal point $\hat{w}$ such that $\|\hat{\boldsymbol{w}}\|_q \le 1$ and $G(\hat{\boldsymbol{w}}) \le \inf_{\boldsymbol{w}} G(\boldsymbol{w}) + \varepsilon$.*

**Remark 8.15.** *When $q = 1$, we will use the entropy potential function $\psi(\boldsymbol{w}) = \sum_{i=1}^d w_i \log w_i$. In this case, Stochastic Mirror Descent will require $\mathcal{O}(\frac{L^2 \log d}{\varepsilon^2})$ stochastic gradients.*

We are now ready to state our main result for this section:

**Theorem 8.16.** *Let $\mathcal{X} = \left\{\boldsymbol{x} \in \mathbb{R}^d : \|\boldsymbol{x}\|_p \le 1\right\}$. Let $\mathcal{D}$ be a distribution over $\mathcal{X} \times \mathcal{Y}$ such that there exists a halfspace $\boldsymbol{w}^* \in \mathbb{R}^d$ with $\mathbf{Pr}_{\boldsymbol{x}\sim\mathcal{D}_{\boldsymbol{x}}}\left[\left|\langle \boldsymbol{w}^*, \boldsymbol{x}\rangle\right| > \gamma\right] = 1$ and $y$ is generated by $h_{\boldsymbol{w}^*}(\boldsymbol{x}) := \operatorname{sign}(\langle \boldsymbol{w}^*, \boldsymbol{x}\rangle)$ corrupted by RCN with noise rate $\eta < 1/2$. An application of Stochastic Mirror Descent on $G_\lambda^\gamma(\boldsymbol{w})$, returns, with high probability, a halfspace $\boldsymbol{w}$ where $\|\boldsymbol{w}\|_q \le 1$ with $\gamma/2$-robust misclassification error $\mathbb{E}_{(\boldsymbol{x},y)\sim\mathcal{D}}\left[1\{y\langle \boldsymbol{w}, \boldsymbol{x}\rangle \le \gamma/2\}\right] \le \eta + \varepsilon$ in $\mathsf{poly}(d, 1/\varepsilon, 1/\gamma, 1/(1-2\eta))$ time.*

With Theorem 8.16, the proof of Theorem 8.12 immediately follows.

*Proof of Theorem 8.12.* This follows from Lemma 8.13 and Theorem 8.16. □

**Remark 8.17.** *In Theorem 8.16, we get a $\gamma/2$-robustness guarantee assuming $\gamma$-robust half-space $\boldsymbol{w}^*$ that is corrupted with random classification noise. This can be strengthened to get a guarantee of $(1-c)\gamma$-robustness for any constant $c > 0$.*

The rest of this section is devoted to the proof of Theorem 8.16. The high-level strategy is to show that an $\varepsilon'$-suboptimal solution to Equation (8.6) gives us an $\varepsilon$-suboptimal solution to Equation (8.5) (for a suitably chosen $\varepsilon'$). In Lemma 8.19, we bound from above the $\gamma/2$-margin loss in terms of our convex surrogate objective $G_\lambda^\gamma$, and in Lemma 8.20 we show that there are minimizers of our convex surrogate $G_\lambda^\gamma$ such that it is sufficiently small. These are the two key lemmas that we will use to piece everything together. The proofs of lemmas are deferred to Appendix F.1.

For any $\boldsymbol{w}, \boldsymbol{x} \in \mathbb{R}^d$, consider the contribution of the objective $G_\lambda^\gamma$ of $\boldsymbol{x}$, denoted by $G_\lambda^\gamma(\boldsymbol{w}, \boldsymbol{x})$. This is defined as $G_\lambda^\gamma(\boldsymbol{w}, \boldsymbol{x}) = \mathbb{E}_{y \sim \mathcal{D}_y(\boldsymbol{x})}\left[\varphi(y\langle \boldsymbol{w}, \boldsymbol{x}\rangle)\right] = \eta\varphi(-z) + (1-\eta)\varphi(z)$ where $z = h_{\boldsymbol{w}^*}(\boldsymbol{x})\langle \boldsymbol{w}, \boldsymbol{x}\rangle$. In the following lemma, we provide a decomposition of $G_\lambda^\gamma(\boldsymbol{w}, \boldsymbol{x})$ that will help us in proving Lemmas 8.19 and 8.20.

**Lemma 8.18.** *For any $\boldsymbol{w}, \boldsymbol{x} \in \mathbb{R}^d$, let $z = h_{\boldsymbol{w}^*}(\boldsymbol{x})\langle \boldsymbol{w}, \boldsymbol{x}\rangle$. Then, we have that:*

$$\begin{aligned} G_\lambda^\gamma(\boldsymbol{w}, \boldsymbol{x}) = (\eta - \lambda)\left(\frac{z}{\gamma}\right) + \lambda + \eta - 2\lambda\eta \\ + 1\{-\gamma \le z \le \gamma\}(1-\eta)(1-2\lambda)\left(1 - \frac{z}{\gamma}\right) \\ + 1\{z < -\gamma\}(1-2\lambda)\left(1 - 2\eta - \frac{z}{\gamma}\right). \end{aligned}$$

The following lemma allows us to bound from below our convex surrogate $\mathbb{E}_{\boldsymbol{x}\sim\mathcal{D}_{\boldsymbol{x}}}[G_\lambda^\gamma(\boldsymbol{w}, \boldsymbol{x})]$ in terms of the $\gamma/2$-margin loss of $\boldsymbol{w}$.

**Lemma 8.19.** *Assume that $\lambda$ is chosen such that $\lambda < 1/2$ and $\eta < \lambda$. Then, for any $\boldsymbol{w} \in \mathbb{R}^d$, $\mathbb{E}_{\boldsymbol{x}\sim\mathcal{D}_{\boldsymbol{x}}}[G_\lambda^\gamma(\boldsymbol{w}, \boldsymbol{x})] \ge \frac{\eta-\lambda}{\gamma} + \frac{1}{2}(1-2\lambda)(1-\eta)\,\mathbb{E}_x\left[1\{z \le \frac{\gamma}{2}\}\right] + \lambda + \eta - 2\lambda\eta$.*

We now show that there exist minimizers of the convex surrogate $G_\lambda^\gamma$ such that it is sufficiently small, which will be useful later in choosing the suboptimality parameter $\varepsilon'$.

**Lemma 8.20.** *Assume that $\lambda$ is chosen such that $\lambda < 1/2$ and $\eta < \lambda$. Then we have that*

$$\inf_{\boldsymbol{w}\in\mathbb{R}^d} \mathop{\mathbb{E}}_{\boldsymbol{x}\sim\mathcal{D}_{\boldsymbol{x}}}[G_\lambda^\gamma(\boldsymbol{w},\boldsymbol{x})] \le 2\eta(1-\lambda).$$

Using the above lemmas, we are now able to bound from above the $\gamma/2$-margin loss of a halfspace $\boldsymbol{w}$ that is $\varepsilon'$-suboptimal for our convex optimization problem (see Equation (8.6)).

**Lemma 8.21.** *For any $\varepsilon' \in (0,1)$ and any $\boldsymbol{w} \in \mathbb{R}^d$ such that $\mathbb{E}_{x\sim\mathcal{D}_x}[G_\lambda^\gamma(\boldsymbol{w},\boldsymbol{x})] \le \mathbb{E}_{x\sim\mathcal{D}_x}[G_\lambda^\gamma(\boldsymbol{w}^*,\boldsymbol{x})] + \varepsilon'$, the $\gamma/2$-missclassification error of $\boldsymbol{w}$ satisfies*

$$\mathop{\mathbb{E}}_{(\boldsymbol{x},y)\sim\mathcal{D}}[1\{y\langle\boldsymbol{w},\boldsymbol{x}\rangle \le \gamma/2\}] \le$$
$$\eta + \frac{2}{(1-2\lambda)}\left(\varepsilon' + (\lambda-\eta)\left(\frac{1}{\gamma}-1\right)\right) .$$

*Proof.* By Lemma 8.19 and Lemma 8.20, we have

$$\frac{\eta-\lambda}{\gamma} + \frac{1}{2}(1-2\lambda)(1-\eta)\mathop{\mathbb{E}}_{x}\left[1\left\{z \le \frac{\gamma}{2}\right\}\right] + \lambda + \eta - 2\lambda\eta \le 2\eta(1-\lambda) + \varepsilon'.$$

This implies

$$(1-\eta)\mathop{\mathbb{E}}_{x}\left[1\left\{z \le \frac{\gamma}{2}\right\}\right] \le \frac{2}{(1-2\lambda)}\left(\varepsilon' + (\lambda-\eta)\left(\frac{1}{\gamma}-1\right)\right).$$

Since $\mathbb{E}_{(\boldsymbol{x},y)\sim\mathcal{D}}[1\{y\langle\boldsymbol{w},\boldsymbol{x}\rangle \le \gamma/2\}] \le \eta + (1-\eta)\,\mathbb{E}_x\left[1\left\{z \le \frac{\gamma}{2}\right\}\right]$, we get the desired result. □

We are now ready to prove Theorem 8.16.

*Proof of Theorem 8.16.* Based on Lemma 8.21, we will choose $\lambda, \varepsilon'$ such that

$$\frac{2}{(1-2\lambda)}\left(\varepsilon' + (\lambda-\eta)\left(\frac{1}{\gamma}-1\right)\right) \le \varepsilon .$$

By setting $\varepsilon' = \lambda - \eta$, this condition reduces to

$$\frac{2(\lambda - \eta)}{(1 - 2\lambda)} \leq \gamma\varepsilon \ .$$

This implies that we need $\lambda \leq \frac{\varepsilon\gamma/2+\eta}{1+\varepsilon\gamma}$. We will choose $\lambda = \frac{\varepsilon\gamma/2+\eta}{1+\varepsilon\gamma}$. Note that our analysis relied on having $\lambda \leq 1/2$ and $\eta \leq \lambda$. These conditions combined imply that we should choose $\lambda$ such that $\eta \leq \lambda \leq 1/2$. Our choice of $\lambda = \frac{\varepsilon\gamma/2+\eta}{1+\varepsilon\gamma}$ satisfies these conditions, since

$$\frac{\varepsilon\gamma/2 + \eta}{1 + \varepsilon\gamma} - \eta = \frac{\varepsilon\gamma/2 + \eta - \eta(1 + \varepsilon\gamma)}{1 + \varepsilon\gamma} = \frac{\varepsilon\gamma(1/2 - \eta)}{1 + \varepsilon\gamma} \geq 0 \ ,$$

and

$$\frac{\varepsilon\gamma/2 + \eta}{1 + \varepsilon\gamma} - \frac{1}{2} = \frac{\varepsilon\gamma/2 + \eta - 1/2(1 + \varepsilon\gamma)}{1 + \varepsilon\gamma} = \frac{\eta - 1/2}{1 + \varepsilon\gamma} \leq 0 \ .$$

By our choice of $\lambda$, we have that $\varepsilon' = \lambda - \eta = \frac{\varepsilon\gamma(1/2-\eta)}{1+\varepsilon\gamma} = \frac{\varepsilon\gamma(1-2\eta)}{2(1+\varepsilon\gamma)}$. By the guarantees of Stochastic Mirror Descent (see Lemma 8.14), our theorem follows with $\mathcal{O}\left(\frac{1}{\varepsilon^2\gamma^2(1-2\eta)^2(q-1)}\right)$ samples for $q > 1$ and $\mathcal{O}\left(\frac{\log d}{\varepsilon^2\gamma^2(1-2\eta)^2}\right)$ samples for $q = 1$. □

## 8.4 Conclusion

In this chapter, we provide necessary and sufficient conditions for perturbation sets $\mathcal{U}$, under which we can efficiently solve the robust empirical risk minimization (RERM) problem. We give a polynomial time algorithm to solve RERM given access to a polynomial time separation oracle for $\mathcal{U}$. In addition, we show that an efficient *approximate* separation oracle for $\mathcal{U}$ is necessary for even computing the robust loss of a halfspace. As a corollary, we show that halfspaces are efficiently robustly PAC learnable for a broad range of perturbation sets. By relaxing the realizability assumption, we show that under random classification noise, we can efficiently robustly PAC learn halfspaces with respect to any $\ell_p$ perturbations. An interesting direction for future work is to understand the computational complexity of robustly PAC learning halfspaces under stronger noise models, including Massart noise and agnostic noise.

# 9

# Transductive Robustness

## 9.1 Introduction

Thus far, we have studied adversarially robust learning almost exclusively in the *inductive* setting, where the task is to learn, from (non-adversarial) training data, a *predictor* with small robust risk (Equation 2.1). In many applications in practice, however, test examples are available in batches and machine learning systems are tasked with classifying them all at once. *Transductive* learning refers to the learning setting where the goal is to classify a given unlabeled test set that is presented together with the training set (Vapnik, 1998).

In this chapter, we study adversarially robust learning in the *transductive* setting. In this problem, $n$ i.i.d. training examples $(\boldsymbol{x}, \boldsymbol{y}) \sim \mathcal{D}^n$ and $m$ separate i.i.d. test examples $(\tilde{\boldsymbol{x}}, \tilde{\boldsymbol{y}}) \sim \mathcal{D}^m$ are drawn from some unknown distribution $\mathcal{D}$. Then, based on all available information: $\boldsymbol{x}, \boldsymbol{y}, \tilde{\boldsymbol{x}}, \tilde{\boldsymbol{y}}$, distribution $\mathcal{D}$, perturbation set $\mathcal{U}$, and white-box access to the transductive learner $\mathbb{A} : (\mathcal{X} \times \mathcal{Y})^n \times \mathcal{X}^m \to \mathcal{Y}^m$, an adversary chooses adversarial perturbations of the test set $\tilde{z}_i \in \mathcal{U}(\tilde{x}_i) \forall i \in [m]$, which we henceforth denote by $\tilde{\boldsymbol{z}} \in \mathcal{U}(\tilde{\boldsymbol{x}})$. Finally, the transductive learner $\mathbb{A}$ receives as input the labeled training examples $(\boldsymbol{x}, \boldsymbol{y})$ and the perturbed test examples $\tilde{\boldsymbol{z}}$, and outputs a labeling for $\tilde{\boldsymbol{z}}$ which we denote by $\mathbb{A}(\boldsymbol{x}, \boldsymbol{y}, \tilde{\boldsymbol{z}}) \in \mathcal{Y}^{m}$*. The performance of $\mathbb{A}$ is

*Throughout the paper, we abuse notation and use $\mathbb{A}(\boldsymbol{x}, \boldsymbol{y}, \tilde{\boldsymbol{z}})(\tilde{z}_i)$ to refer to the $i^{\text{th}}$ entry in the vector $\mathbb{A}(\boldsymbol{x}, \boldsymbol{y}, \tilde{\boldsymbol{z}})$.

measured by the *transductive robust risk*[†]:

$$\mathrm{TR}_{\mathcal{U}}^{n,m}(\mathbb{A};\mathcal{D}) = \mathop{\mathbb{E}}_{\substack{(\boldsymbol{x},\boldsymbol{y})\sim\mathcal{D}^n \\ (\tilde{\boldsymbol{x}},\tilde{\boldsymbol{y}})\sim\mathcal{D}^m}} \left[ \sup_{\tilde{\boldsymbol{z}}\in\mathcal{U}(\tilde{\boldsymbol{x}})} \frac{1}{m}\sum_{i=1}^{m} 1\left\{\mathbb{A}(\boldsymbol{x},\boldsymbol{y},\tilde{\boldsymbol{z}})(\tilde{z}_i) \neq \tilde{y}_i\right\} \right]. \tag{9.1}$$

As we shall show, the transductive setting allows for much stronger results than what we currently know in the inductive adversarially robust setting.

How is this possible? In traditional (non-robust) learning, there are standard transductive-to-inductive and inductive-to-transductive reductions which establish that both settings are essentially equivalent statistically. However, in Section 9.4 we discuss how the inductive-to-transductive reduction breaks down for adverserially robust learning, opening the possibility that transductive robust learning might be inherently easier than inductive robust learning. This is good news, since inductive adversarially robust learning so far seems challenging. In particular, we showed in Chapter 3 (Theorem 3.4) that hypothesis classes with bounded VC dimension are adversarially robustly learnable, but existing inductive methods require sample complexity exponential in the VC dimension. In contrast, for transductive adversarially robust learning, we will present in this chapter a simple and straight-forward algorithm with sample complexity only *linear* in the VC-dimension!

So why are we interested in the transductive setting? First, if the adversarially robust transductive setting is indeed easier than its inductive counterpart, it is important to develop methods that take advantage of this setting, and could be applicable and beneficial when entire batches of test examples are processed concurrently. This is the first work, as far as we are aware, in this direction. Alternatively, perhaps advances in analyzing the transductive setting could potentially translate back to the inductive setting—although the standard reduction does not apply, we can still be hopeful we might close the gap through additional ideas.

Relaxed guarantees: choice of competitor  As with most learning theory gurantees, we will show how, given enough samples, we can approach the error of some reference competitor. The best we can hope for is to compete with $\mathsf{OPT}_{\mathcal{U}} = \inf_{h\in\mathcal{H}} \Pr_{(x,y)\sim\mathcal{D}}\left[\exists z\in\mathcal{U}(x) : h(z)\neq y\right]$, which is the smallest attainable robust risk against perturbation set $\mathcal{U}$—this is the best we could do even if we knew the source distribution. In this chapter, we consider a weaker goal

[†]Unless otherwise stated, in this chapter we fix the test set size $m = n$.

where we compete with the smallest attainable robust risk against a stronger adversary:

$$\mathsf{OPT}_{\mathcal{U}^{-1}(\mathcal{U})} = \inf_{h\in\mathcal{H}} \Pr_{(x,y)\sim\mathcal{D}} \left[\exists \tilde{x} \in \mathcal{U}^{-1}(\mathcal{U})(x) : h(\tilde{x}) \neq y\right], \tag{9.2}$$

where $\mathcal{U}^{-1}(z) = \{x \in \mathcal{X} : z \in \mathcal{U}(x)\}$ and $\mathcal{U}^{-1}(\mathcal{U})(x) = \cup_{z\in\mathcal{U}(x)}\mathcal{U}^{-1}(z) = \{\tilde{x} \in \mathcal{X} : \mathcal{U}(x) \cap \mathcal{U}(\tilde{x}) \neq \emptyset\}$.

In words, $\mathsf{OPT}_{\mathcal{U}^{-1}(\mathcal{U})}$ is the smallest attainable robust risk against the larger perturbation set $\mathcal{U}^{-1}(\mathcal{U})$. In particular, when $x \in \mathcal{U}(x), \mathcal{U}(x) \subseteq \mathcal{U}^{-1}(\mathcal{U})(x)$ and $\mathsf{OPT}_{\mathcal{U}} \leq \mathsf{OPT}_{\mathcal{U}^{-1}(\mathcal{U})}$. And what we will show is a transductive learner $\mathbb{A}$ with robust risk $\mathrm{TR}_{\mathcal{U}}(\mathbb{A}; \mathcal{D})$ which is competitive with the best robust risk $\mathsf{OPT}_{\mathcal{U}^{-1}(\mathcal{U})}$ against the larger perturbation set $\mathcal{U}^{-1}(\mathcal{U})$.

For example, consider $\mathcal{U}(x) = \mathrm{B}_{\gamma}(x) \triangleq \{z \in \mathcal{X} : \rho(x,z) \leq \gamma\}$ where $\gamma > 0$ and $\rho$ is some metric on $\mathcal{X}$ (e.g., $\ell_p$-balls). In this case, $\mathcal{U}^{-1}(\mathcal{U})(x) = \mathrm{B}_{2\gamma}(x)$. Furthermore, $\mathsf{OPT}_{\mathcal{U}}$ corresponds to optimal robust risk with radius $\gamma$, while $\mathsf{OPT}_{\mathcal{U}^{-1}(\mathcal{U})}$ corresponds to optimal robust risk with radius $2\gamma$. In this case, our guarantees will ensure robustness to perturbations within radius $\gamma$, that is almost as good as the best possible robust risk with radius $2\gamma$. In particular, our guarantees in the realizable setting ensure robustness to perturbations within radius $\gamma$ when the smallest robust risk with radius $2\gamma$ is zero, i.e., $\mathsf{OPT}_{\mathcal{U}^{-1}(\mathcal{U})} = 0$. By way of analogy, guarantees that are similar in spirit are common in the context of bi-criteria approximation algorithms for discrete optimization problems, e.g., the sparsest cut approximation algorithm due to Arora, Rao, and Vazirani (2004).

Main Results We shed some new light on the problem of adversarially robust learning by studying the transductive setting. We propose a *simple* transductive learning algorithm with robust learning guarantees that are stronger than the known inductive guarantees in some aspects, but weaker in other aspects. Specifically, our algorithm enjoys an improved robust error rate that is at most *linear* in the VC dimension and is adaptive to the complexity of the perturbation set $\mathcal{U}$, and is also robust to adversarial perturbations in the *training* data. This comes at the expense of competing with the more restrictive $\mathsf{OPT}_{\mathcal{U}^{-1}(\mathcal{U})}$, where the inductive guarantees compete with $\mathsf{OPT}_{\mathcal{U}}$.

Specifically, given a class $\mathcal{H}$ and a perturbation set $\mathcal{U}$, we present a simple tansductive learner $\mathbb{A} : (\mathcal{X} \times \mathcal{Y})^n \times \mathcal{X}^n \to \mathcal{Y}^n$ (see Section 9.3) such that for any distribution $\mathcal{D}$ over $\mathcal{X} \times \mathcal{Y}$:

$$\mathrm{TR}_{\mathcal{U}}(\mathbb{A};\mathcal{D}) \leq \frac{\mathrm{vc}(\mathcal{H})\log(2n)}{n} \qquad \text{(Realizable, } \mathsf{OPT}_{\mathcal{U}^{-1}(\mathcal{U})}=0\text{)}, \tag{9.3}$$

$$\mathrm{TR}_{\mathcal{U}}(\mathbb{A};\mathcal{D}) \leq 2\mathsf{OPT}_{\mathcal{U}^{-1}(\mathcal{U})} + O\left(\sqrt{\frac{\mathrm{vc}(\mathcal{H})}{n}}\right) \qquad \text{(Agnostic, } \mathsf{OPT}_{\mathcal{U}^{-1}(\mathcal{U})}>0\text{)}. \tag{9.4}$$

Our transductive learner $\mathbb{A}$ simply asks for any predictor $h \in \mathcal{H}$ that robustly and correctly labels the training examples $(\boldsymbol{x}, \boldsymbol{y})$ with respect to $\mathcal{U}^{-1}$ and robustly labels the test examples $\boldsymbol{z}$ with respect to $\mathcal{U}^{-1}$. In Section 9.3, we show that our transductive learner additionally enjoys the following properties:

1. Robustness guarantees against adversarial perturbations in the *training* data. These are the first non-trivial learning guarantees against adversarial perturbations in the training data, which has not been considered before in the literature to the best of our knowledge.

2. Adaptive robust error rates that are controlled by the complexity of $\mathcal{H}$ and the perturbation set $\mathcal{U}$ in the form of a new complexity measure that we introduce: the *relaxed $\mathcal{U}$-robust shattering dimension* $\mathrm{rdim}_{\mathcal{U}}(\mathcal{H})$ (see Definition 9.1). These are the first general robust learning guarantees that take the complexity of the perturbation set $\mathcal{U}$ into account.

Practical Implications In the context of deep learning and robustness to $\ell_p$ perturbations, and in scenarios where (adversarial) unlabeled test data is available in batches, our results suggest that to incur a low error rate on the test data it suffices to perform adversarial training (e.g., Madry, Makelov, Schmidt, Tsipras, and Vladu, 2018, Zhang, Yu, Jiao, Xing, Ghaoui, and Jordan, 2019b) to find network parameters that simultaneously: (a) robustly and correctly fit the labeled training data, and (b) robustly fit the unlabeled (adversarial) test data. This is in contrast with the inductive setting, where it is empirically observed that adversarial training does not always guarantee robust generalization (Schmidt, Santurkar, Tsipras, Talwar, and Madry, 2018). Compared with inductive learning, transductive learning offers a new perspective on adversarial robustness that highlights how unlabeled adversarial test data can inform local robustness, which perhaps is easier to achieve than global robustness.

## 9.2 Preliminaries

Let $\mathcal{X}$ denote the instance space and $\mathcal{Y} = \{\pm 1\}$. Let $\mathcal{H} \subseteq \mathcal{Y}^{\mathcal{X}}$ denote a hypothesis class and $\mathrm{vc}(\mathcal{H})$ denotes its VC dimension. Let $\mathcal{U} : \mathcal{X} \to 2^{\mathcal{X}}$ denote an arbitrary perturbation set such that for each $x \in \mathcal{X}, \mathcal{U}(x)$ is non-empty. Denote by $\mathcal{U}^{-1}$ the inverse image of $\mathcal{U}$, where for each $z \in \mathcal{X}, \mathcal{U}^{-1}(z) = \{x \in \mathcal{X} : z \in \mathcal{U}(x)\}$. Observe that for any $x, z \in \mathcal{X}$ it holds that $z \in \mathcal{U}(x) \Leftrightarrow x \in \mathcal{U}^{-1}(z)$. For an instance $x \in \mathcal{X}, \mathcal{U}^{-1}(\mathcal{U})(x)$ denotes the set of all *natural* examples $\tilde{x}$ that share some perturbation with $x$ according to $\mathcal{U}$, i.e., $\mathcal{U}^{-1}(\mathcal{U})(x) = \cup_{z \in \mathcal{U}(x)} \mathcal{U}^{-1}(z) = \{\tilde{x} \in \mathcal{X} : \mathcal{U}(x) \cap \mathcal{U}(\tilde{x}) \neq \emptyset\}$. For any sequence of labeled points $(\boldsymbol{x}, \boldsymbol{y}) \in (\mathcal{X} \times \mathcal{Y})^n$, any sequence of adversarial perturbations $\boldsymbol{z} \in \mathcal{X}^n$, and any predictor $h : \mathcal{X} \to \mathcal{Y}$ let $\mathrm{err}_{\boldsymbol{x},\boldsymbol{y}}(h) = \frac{1}{n} \sum_{i=1}^{n} 1\{h(x) \neq y\}$ denote the standard 0-1 error, and define:

$$\mathrm{R}_{\mathcal{U}^{-1}}(h; \boldsymbol{z}, \boldsymbol{y}) = \frac{1}{n} \sum_{i=1}^{n} 1\left\{\exists \tilde{x}_i \in \mathcal{U}^{-1}(z_i) : h(\tilde{x}_i) \neq y_i\right\}. \tag{9.5}$$

$$\mathrm{R}_{\mathcal{U}^{-1}}(h; \boldsymbol{z}) = \frac{1}{n} \sum_{i=1}^{n} 1\left\{\exists \tilde{x}_i \in \mathcal{U}^{-1}(z_i) : h(\tilde{x}_i) \neq h(z_i)\right\}. \tag{9.6}$$

Our transductive robust learning guarantees (presented in Section 9.3) are in fact in terms of an adaptive complexity measure—that is in general tighter than the VC dimension and takes into account the complexity of both $\mathcal{H}$ and $\mathcal{U}$—which we introduce next:

**Definition 9.1** (Relaxed Robust Shattering Dimension). *A sequence $z_1, \dots, z_k \in \mathcal{X}$ is said to be* relaxed $\mathcal{U}$-robustly shattered *by $\mathcal{H}$ if $\forall y_1, \dots, y_k \in \{\pm 1\} : \exists x_1^{y_1}, \dots, x_k^{y_k} \in \mathcal{X}$ and $\exists h \in \mathcal{H}$ such that $z_i \in \mathcal{U}(x_i^{y_i})$ and $h(\mathcal{U}(x_i^{y_i})) = y_i \forall 1 \leq i \leq k$. The* relaxed $\mathcal{U}$-robust shattering dimension $\mathrm{rdim}_{\mathcal{U}}(\mathcal{H})$ *is defined as the largest $k$ for which there exist $k$ points that are relaxed $\mathcal{U}$-robustly shattered by $\mathcal{H}$.*

The above complexity measure is inspired by the robust shattering dimension (Definition 3.10) that we introduced in Chapter 3 and showed to lower bound the sample complexity of robust learning in the inductive setting (Theorem 3.11). The definition of $\mathrm{rdim}_{\mathcal{U}}(\mathcal{H})$ implies the following:

**Remark 9.1.** *For any class $\mathcal{H}$ and any perturbation set $\mathcal{U}$,* $\mathrm{rdim}_{\mathcal{U}}(\mathcal{H}) \leq \mathrm{vc}(\mathcal{H})$.

## 9.3 Main Results

We obtain strong robust learning guarantees against worst-case adversarial perturbations of *both* the training data and the test data. Specifically, after training examples $(\boldsymbol{x}, \boldsymbol{y}) \sim \mathcal{D}^n$ and test examples $(\tilde{\boldsymbol{x}}, \tilde{\boldsymbol{y}}) \sim \mathcal{D}^n$ are drawn, an adversary perturbs both training and test examples by choosing adversarial perturbations $\boldsymbol{z} \in \mathcal{U}(\boldsymbol{x})$ and $\tilde{\boldsymbol{z}} \in \mathcal{U}(\tilde{\boldsymbol{x}})$. Our transductive learner observes as input $(\boldsymbol{z}, \boldsymbol{y})$ and $\tilde{\boldsymbol{z}}$, and outputs $\hat{h}(\tilde{\boldsymbol{z}}) \in \mathcal{Y}^n$ where $\hat{h} \in \Delta^{\mathcal{U}}_{\mathcal{H}}(\boldsymbol{z}, \boldsymbol{y}, \tilde{\boldsymbol{z}})$ defined as follows:

$$\Delta^{\mathcal{U}}_{\mathcal{H}}(\boldsymbol{z}, \boldsymbol{y}, \tilde{\boldsymbol{z}}) = \{h \in \mathcal{H} : \mathrm{R}_{\mathcal{U}^{-1}}(h; \boldsymbol{z}, \boldsymbol{y}) = 0 \wedge \mathrm{R}_{\mathcal{U}^{-1}}(h; \tilde{\boldsymbol{z}}) = 0\} \quad \text{(Realizable, } \mathsf{OPT}_{\mathcal{U}^{-1}(\mathcal{U})} = 0\text{)}. \tag{9.7}$$

$$\Delta^{\mathcal{U}}_{\mathcal{H}}(\boldsymbol{z}, \boldsymbol{y}, \tilde{\boldsymbol{z}}) = \operatorname*{argmin}_{h \in \mathcal{H}} \max\left\{\mathrm{R}_{\mathcal{U}^{-1}}(h; \boldsymbol{z}, \boldsymbol{y}),\ \mathrm{R}_{\mathcal{U}^{-1}}(h; \tilde{\boldsymbol{z}})\right\} \quad \text{(Agnostic, } \mathsf{OPT}_{\mathcal{U}^{-1}(\mathcal{U})} > 0\text{)}. \tag{9.8}$$

Our transductive learner simply asks for any predictor $h \in \mathcal{H}$ that robustly and correctly labels the training examples $(\boldsymbol{z}, \boldsymbol{y})$ with respect to $\mathcal{U}^{-1}$ and robustly labels the test examples $\boldsymbol{z}$ with respect to $\mathcal{U}^{-1}$. Observe that requiring robustness on $\boldsymbol{z}$ and $\tilde{\boldsymbol{z}}$ with respect to $\mathcal{U}^{-1}$ implies, by definition of $\mathcal{U}^{-1}$, that the i.i.d. examples $\boldsymbol{x}$ and $\tilde{\boldsymbol{x}}$ will be labeled in the same way as $\boldsymbol{z}$ and $\tilde{\boldsymbol{z}}$, even though the learner does not observe $\boldsymbol{x}$ and $\tilde{\boldsymbol{x}}$. This is the main insight that we rely on to obtain our transductive robust learning guarantees:

**Theorem 9.1** (Realizable). *For any $n \in \mathbb{N}$, $\delta > 0$, class $\mathcal{H}$, perturbation set $\mathcal{U}$, and distribution $\mathcal{D}$ over $\mathcal{X} \times \mathcal{Y}$ satisfying $\mathsf{OPT}_{\mathcal{U}^{-1}(\mathcal{U})} = 0$:*

$$\Pr_{\substack{(\boldsymbol{x},\boldsymbol{y}) \sim \mathcal{D}^n \\ (\tilde{\boldsymbol{x}},\tilde{\boldsymbol{y}}) \sim \mathcal{D}^n}} \left[\forall \boldsymbol{z} \in \mathcal{U}(\boldsymbol{x}), \forall \tilde{\boldsymbol{z}} \in \mathcal{U}(\tilde{\boldsymbol{x}}), \forall \hat{h} \in \Delta^{\mathcal{U}}_{\mathcal{H}}(\boldsymbol{z}, \boldsymbol{y}, \tilde{\boldsymbol{z}}) :\ \mathrm{err}_{\tilde{\boldsymbol{z}},\tilde{\boldsymbol{y}}}(\hat{h}) \leq \varepsilon\right] \geq 1 - \delta,$$

*where* $\varepsilon = \frac{\mathrm{rdim}_{\mathcal{U}^{-1}}(\mathcal{H}) \log(2n) + \log(1/\delta)}{n} \leq \frac{\mathrm{vc}(\mathcal{H}) \log(2n) + \log(1/\delta)}{n}$.

**Theorem 9.2** (Agnostic). *For any $n \in \mathbb{N}$, $\delta > 0$, class $\mathcal{H}$, perturbation set $\mathcal{U}$, and distribution*

$\mathcal{D}$ *over* $\mathcal{X} \times \mathcal{Y}$,

$$\Pr_{\substack{(\boldsymbol{x},\boldsymbol{y})\sim\mathcal{D}^n \\ (\tilde{\boldsymbol{x}},\tilde{\boldsymbol{y}})\sim\mathcal{D}^n}} \left[\forall \boldsymbol{z} \in \mathcal{U}(\boldsymbol{x}), \forall \tilde{\boldsymbol{z}} \in \mathcal{U}(\tilde{\boldsymbol{x}}), \forall \hat{h} \in \Delta^{\mathcal{U}}_{\mathcal{H}}(\boldsymbol{z}, \boldsymbol{y}, \tilde{\boldsymbol{z}}) : \ \mathrm{err}_{(\tilde{\boldsymbol{z}},\tilde{\boldsymbol{y}})}(\hat{h}) \leq \varepsilon\right] \geq 1 - \delta,$$

*where* $\varepsilon = \min\left\{2\mathsf{OPT}_{\mathcal{U}^{-1}(\mathcal{U})} + O\left(\sqrt{\frac{\mathrm{vc}(\mathcal{H})+\log(1/\delta)}{n}}\right), 3\mathsf{OPT}_{\mathcal{U}^{-1}(\mathcal{U})} + O\left(\sqrt{\frac{\mathrm{rdim}_{\mathcal{U}^{-1}}(\mathcal{H})\ln(2n)+\ln(1/\delta)}{n}}\right)\right\}.$

## 9.4 Transductive vs. Inductive

For purposes of the discussion below, let $\mathbb{A}_I : (\mathcal{X} \times \mathcal{Y})^* \to \mathcal{Y}^{\mathcal{X}}$ denote an inductive learner and $\mathbb{A}_T : (\mathcal{X} \times \mathcal{Y})^n \times \mathcal{X}^m \to \mathcal{Y}^m$ denote a trnasductive learner. The *inductive* robust risk of $\mathbb{A}_I$ is defined as

$$\mathrm{IR}^n_{\mathcal{U}}(\mathbb{A}; \mathcal{D}) = \mathop{\mathbb{E}}_{(\boldsymbol{x},\boldsymbol{y})\sim\mathcal{D}^n} \mathrm{R}_{\mathcal{U}}(\mathbb{A}(\boldsymbol{x}, \boldsymbol{y}); \mathcal{D}) = \mathop{\mathbb{E}}_{(\boldsymbol{x},\boldsymbol{y})\sim\mathcal{D}^n} \mathop{\mathbb{E}}_{(x,y)\sim\mathcal{D}} \sup_{z\in\mathcal{U}(x)} \mathbb{1}\left\{\mathbb{A}(\boldsymbol{x}, \boldsymbol{y})(z) \neq y\right\}.$$

For standard (non-robust) supervised learning, i.e., when $\mathcal{U}(x) = \{x\}$, there isn't much difference between the transductive and inductive settings in terms of statistical performance—an observation which has been employed in designing and analyzing inductive learning algorithms by relying on the transductive setting (Vapnik and Chervonenkis, 1974). We can always take an inductive learner $\mathbb{A}_I$ and use it transductively as $\mathbb{A}_T$ defined as

$$\forall i \in [m] : \mathbb{A}_T(\boldsymbol{x}, \boldsymbol{y}, \tilde{\boldsymbol{x}})(\tilde{x}_i) = \mathbb{A}_I(\boldsymbol{x}, \boldsymbol{y})(\tilde{x}_i), \tag{9.9}$$

and so $\mathrm{TR}^{n,m}(\mathbb{A}_T; \mathcal{D}) \leq \mathbb{E}\left[\frac{1}{n}\sum_{i=1}^n \mathbb{1}\left\{\mathbb{A}_I(\boldsymbol{x}, \boldsymbol{y})(\tilde{x}_i) \neq \tilde{y}_i\right\}\right] = \mathrm{IR}^n(\mathbb{A}_I; \mathcal{D})$.

In the other direction, given a transductive learner $\mathbb{A}_T$, if it's guarantee doesn't depend on the test set size $m$ (i.e., holds even when $m = 1$), we can consider an inductive learner $\mathbb{A}_I$ that outputs a predictor which just runs the transductive learner at test-time defined as

$$\forall x \in \mathcal{X} : \mathbb{A}_I(\boldsymbol{x}, \boldsymbol{y})(x) = \mathbb{A}_T(\boldsymbol{x}, \boldsymbol{y}, x), \tag{9.10}$$

ensuring $\mathrm{IR}^n(\mathbb{A}_I; \mathcal{D}) = \mathrm{TR}^{n,1}(\mathbb{A}_T; \mathcal{D})$.

More generally, if the transductive learner $\mathbb{A}_T$ does rely on having multiple test examples, e.g., $m = n$ as in our case, we can randomly split the training set, using some of the training examples as test examples:

$$\forall x \in \mathcal{X} : \mathbb{A}_I(\boldsymbol{x}, \boldsymbol{y})(x) = \mathbb{A}_T(\boldsymbol{x}', \boldsymbol{y}', \boldsymbol{x}'' \cup \{x\})(x), \tag{9.11}$$

where $\boldsymbol{x}'$ and $\boldsymbol{x}''$ are random disjoint subsets of $\boldsymbol{x}$ of size $\lfloor \frac{n}{2} \rfloor$ and $\lfloor \frac{n}{2} \rfloor - 1$, and $\boldsymbol{x}'' \cup \{x\}$ is a random permutation of the concatenation. This ensures

$$\begin{aligned} \mathrm{IR}^n(\mathbb{A}_I; \mathcal{D}) &= \mathbb{E}\left[1\left\{\mathbb{A}_T(\boldsymbol{x}', \boldsymbol{y}', \boldsymbol{x}'' \cup \{x\})(x) \neq y\right\}\right] = \mathop{\mathbb{E}}_{\substack{(\boldsymbol{x},\boldsymbol{y}) \sim \mathcal{D}^{n/2} \\ (\tilde{\boldsymbol{x}},\tilde{\boldsymbol{y}}) \sim \mathcal{D}^{n/2}}} \mathop{\mathbb{E}}_{i \sim \mathrm{Unif}[n/2]} \left[1\left\{\mathbb{A}_T(\boldsymbol{x}, \boldsymbol{y}, \tilde{\boldsymbol{x}})(\tilde{x}_i) \neq \tilde{y}_i\right\}\right] \\ &= \mathop{\mathbb{E}}_{\substack{(\boldsymbol{x},\boldsymbol{y}) \sim \mathcal{D}^{n/2} \\ (\tilde{\boldsymbol{x}},\tilde{\boldsymbol{y}}) \sim \mathcal{D}^{n/2}}} \frac{2}{n} \sum_{i=1}^{\frac{n}{2}} 1\left\{\mathbb{A}_T(\boldsymbol{x}, \boldsymbol{y}, \tilde{\boldsymbol{x}})(\tilde{x}_i) \neq \tilde{y}_i\right\} = \mathrm{TR}^{\frac{n}{2}, \frac{n}{2}}(\mathbb{A}_T; \mathcal{D}). \end{aligned}$$

Why is the robust setting different? We can still reduce transductive to inductive just the same. Given an inductive learner $\mathbb{A}_I$, the construction of Equation 9.9 is still valid, and we have $\mathrm{TR}^{n,m}_{\mathcal{U}}(\mathbb{A}_T; \mathcal{D}) \leq \mathrm{IR}^n_{\mathcal{U}}(\mathbb{A}_I; \mathcal{D})$.

But what happens in the reverse direction? If transductive learner $\mathbb{A}_T$ doesn't rely on having multiple test examples, i.e., its guarantee doesn't depend on $m$ and is valid even if $m = 1$, the construction in Equation 9.10 can still be used, and we have $\mathrm{IR}^n_{\mathcal{U}}(\mathbb{A}_I; \mathcal{D}) = \mathrm{TR}^{n,1}_{\mathcal{U}}(\mathbb{A}_T; \mathcal{D})$. This reduction has the potential of aiding in designing robust inductive learning methods, but it relies on the transductive method not depending on the number of test examples, or equivalently referred to as 1-point transductive learners (e.g., the one-inclusion graph prediction algorithm due to Haussler, Littlestone, and Warmuth, 1994)[‡]. Unfortunately, this is not the case for our transductive learner $\mathbb{A}_T$ (presented in Section 9.3) which requires $m = n$.

Trying to apply the other reduction from Equation 9.11 and its analysis in the transductive setting, we would need to implement: $\mathbb{A}_T(\boldsymbol{x}', \boldsymbol{y}', \boldsymbol{z}'' \cup \{z\})$. To apply our transductive learner $\mathbb{A}_T$, we would need not the subset of training points $\boldsymbol{x}''$, but rather their perturbations $\boldsymbol{z}''$. But how could we obtain this? This is not given to us. If $x \in \mathcal{U}(x)$, we can try using $z''_i = x''_i$, i.e.

[‡]We refer the reader to Chapter 4 for a more in-depth exploration of 1-point transductive learners for adversarially robust learning, which led us to a minimax optimal learner in the inductive setting (see Section 4.4).

$\mathbb{A}_T(\boldsymbol{x}', \boldsymbol{y}', \boldsymbol{x}'' \cup \{z\})$, for which we get the following inductive error:

$$\mathrm{IR}^n_{\mathcal{U}}(\mathbb{A}_I; \mathcal{D}) = \mathop{\mathbb{E}}_{\substack{(\boldsymbol{x},\boldsymbol{y})\sim\mathcal{D}^{n/2} \\ (\tilde{\boldsymbol{x}},\tilde{\boldsymbol{y}})\sim\mathcal{D}^{n/2}}} \left[ \mathop{\mathbb{E}}_{i\sim\mathrm{Unif}[n/2]} \sup_{\tilde{z}_i\in\mathcal{U}(\tilde{x}_i)} 1\left\{\mathbb{A}_T(\boldsymbol{x}, \boldsymbol{y}, \tilde{\boldsymbol{x}}_{1:i}, \tilde{z}_i, \tilde{\boldsymbol{x}}_{i+1:\frac{n}{2}})(\tilde{z}_i) \neq y_i\right\} \right].$$

But the right hand side here is only a (loose) upper bound on $\mathrm{TR}^{n/2,n/2}_{\mathcal{U}}(\mathbb{A}_T; \mathcal{D})$. Specifically, in the above, the supremum representing the adversary, comes in after knowing which one of the $\frac{n}{2}$ points we are evaluating on. On the other hand, in a true transductive setting, i.e., in the definition of $\mathrm{TR}^{n/2,n/2}_{\mathcal{U}}$, the adversary needs to commit to a perturbation that would be bad (for us) on all $\frac{n}{2}$ of the points. In a sense, the fact that the adversary must perturb all points, and we can leverage knowledge of these perturbations, is what restricts the power of the adversary in the transductive setting, and allows us to better protect against adversarial attacks affecting many examples (we might still get a few examples wrong, but that's OK).

Proper vs. Improper Another issue where we see a difference between inductive and transductive adversarially robust learning is with regards to whether the learning can be proper. In Chapter 3, we showed that learning some VC classes in the inductive setting necessarily requires improper learning. Specifically, there are classes $\mathcal{H}$ with constant VC dimension that are not robustly PAC learnable with any inductive proper learner $\mathbb{A}_I : (\mathcal{X}\times\mathcal{Y})^* \to \mathcal{H}$ which is constrained to outputting a predictor in $\mathcal{H}$ (Theorem 3.1). Even in the case of robust realizability with respect to $\mathcal{U}^{-1}(\mathcal{U})$, i.e., $\mathsf{OPT}_{\mathcal{U}^{-1}(\mathcal{U})} = 0$, we can still adapt the construction of Theorem 3.1 to conclude that improper learning is needed in the inductive setting, whereas for transductive learning, our learner from Section 9.3 is proper. But this isn't surprising, and also in the standard (non-robust) setting we can expect differences in properness between transductive and inductive.

We mentioned that any transductive non-robust learner can be transformed to an inductive learner, using the reduction in Equation 9.11. But even if the transductive learner is proper, the resulting inductive learner is not. And furthermore, any improper transductive learner, whether non-robust or robust, can be transformed to a proper transductive learner. Specifically, for any transductive learner $\mathbb{A}_T$ and any input $(\boldsymbol{x}, \boldsymbol{y}, \tilde{\boldsymbol{z}}) \in (\mathcal{X} \times \mathcal{Y})^n \times \mathcal{X}^m$, we can project the labeling $\mathbb{A}_T(\boldsymbol{x}, \boldsymbol{y}, \tilde{\boldsymbol{z}})$ to the closest *proper* labeling in the set $\Gamma_{\mathcal{H}}(\tilde{\boldsymbol{z}}) = \{(h(\tilde{\boldsymbol{z}})) : h \in \mathcal{H}\}$. In the realizable setting, when $\exists h \in \mathcal{H}$ s.t. $\mathrm{R}_{\mathcal{U}}(h; \mathcal{D}) = 0$, we are guar-anteed that whenever $\mathbb{A}(\boldsymbol{x}, \boldsymbol{y}, \tilde{\boldsymbol{z}})$ has $\varepsilon$ error then the *proper* labeling has at most $2\varepsilon$ error. In the agnostic setting, we are guaranteed that the *proper* labeling will incur robust error at most $3 \inf_{h \in \mathcal{H}} \mathrm{R}_{\mathcal{U}}(h; \mathcal{D}) + 2\varepsilon$ whenever $\mathbb{A}(\boldsymbol{x}, \boldsymbol{y}, \tilde{\boldsymbol{z}})$ has robust error of $\inf_{h \in \mathcal{H}} \mathrm{R}_{\mathcal{U}}(h; \mathcal{D}) + \varepsilon$. We therefore see that in the transductive setting, improperness can never buy a significant advantage, and we should not be surprised that learning that must be improper in the inductive setting can be proper in the transductive one.

## 9.5 Proofs

We start with proving a helpful lemma that extends the classic Sauer-Shelah-Perles lemma for the robust setting.

**Lemma 9.2** (Sauer's lemma for $\mathrm{rdim}_{\mathcal{U}}(\mathcal{H})$, Definition 9.1). *For any class $\mathcal{H}$, any perturbation set $\mathcal{U}$, and any sequence of points $z_1, \dots, z_n \in \mathcal{X}$,*

$$\left|\Pi_{\mathcal{H}}^{\mathcal{U}}(z_1, \dots, z_n)\right| \triangleq \left|\left\{ (h(z_1), \dots, h(z_n)) \Big|_{z_i \in \mathcal{U}(x_i) \wedge h(\mathcal{U}(x_i)) = h(z_i) \forall 1 \le i \le n}^{\exists x_1, \dots, x_n \in \mathcal{X}, \exists h \in \mathcal{H}:} \right\}\right|$$
$$\le \binom{n}{\le \mathrm{rdim}_{\mathcal{U}}(\mathcal{H})} \triangleq \sum_{i=0}^{\mathrm{rdim}_{\mathcal{U}}(\mathcal{H})} \binom{n}{i}.$$

*Proof.* The proof will follow a standard argument that is used to prove Sauer-Shela-Perles lemma (see e.g., Shalev-Shwartz and Ben-David, 2014). Specifically, it suffices to prove the following stronger claim:

$$\left|\Pi_{\mathcal{H}}^{\mathcal{U}}(z_1, \dots, z_n)\right| \le \left|\{S \subseteq \{z_1, \dots, z_n\} : S \text{ is relaxed } \mathcal{U}\text{-robustly shattered by } \mathcal{H}\}\right|. \tag{9.12}$$

This is because

$$\left|\{S \subseteq \{z_1, \dots, z_n\} : S \text{ is relaxed } \mathcal{U}\text{-robustly shattered by } \mathcal{H}\}\right| \le \binom{n}{\le \mathrm{rdim}_{\mathcal{U}}(\mathcal{H})}.$$

We will prove Equation 9.12 by induction on $n$. When $n = 1$, both sides of Equation 9.12 either evaluate to 1 or 2 (the empty set is always considered to be relaxed $\mathcal{U}$-robustly shattered by $\mathcal{H}$). When $n > 1$, assume that Equation 9.12 holds for sequences of length $k < n$. Let

$C = \{z_1, \dots, z_n\}$ and $C' = \{z_2, \dots, z_n\}$. Consider the following two sets:

$$Y_0 = \left\{ (y_2, \dots, y_n) : (+1, y_2, \dots, y_n) \in \Pi^{\mathcal{U}}_{\mathcal{H}}(z_1, \dots, z_n) \vee (-1, y_2, \dots, y_n) \in \Pi^{\mathcal{U}}_{\mathcal{H}}(z_1, \dots, z_n) \right\},$$

and

$$Y_1 = \left\{ (y_2, \dots, y_n) : (+1, y_2, \dots, y_n) \in \Pi^{\mathcal{U}}_{\mathcal{H}}(z_1, \dots, z_n) \wedge (-1, y_2, \dots, y_n) \in \Pi^{\mathcal{U}}_{\mathcal{H}}(z_1, \dots, z_n) \right\}.$$

Observe that $\left|\Pi^{\mathcal{U}}_{\mathcal{H}}(z_1, \dots, z_n)\right| = |Y_0| + |Y_1|$. Additionally, note that by definition of $Y_0$, $Y_0 \subseteq \Pi^{\mathcal{U}}_{\mathcal{H}}(z_2, \dots, z_n)$. Thus, by the inductive assumption,

$$\begin{aligned} |Y_0| \leq \left|\Pi^{\mathcal{U}}_{\mathcal{H}}(z_2, \dots, z_n)\right| &\leq |\{S \subseteq C' : S \text{ is relaxed } \mathcal{U}\text{-robustly shattered by } \mathcal{H}\}| \\ &= |\{S \subseteq C : z_1 \notin S \wedge S \text{ is relaxed } \mathcal{U}\text{-robustly shattered by } \mathcal{H}\}|. \end{aligned}$$

Next, define $\mathcal{H}' \subseteq \mathcal{H}$ to be

$$\mathcal{H}' = \left\{ h \in \mathcal{H} : \exists h' \in \mathcal{H}, \boldsymbol{x}_{2:n}, \tilde{\boldsymbol{x}}_{2:n} \in \mathcal{U}^{-1}(\boldsymbol{z}_{2:n}) \text{ s.t. } h(\mathcal{U}(x_1)) = -h'(\mathcal{U}(x_1)) \wedge h(\mathcal{U}(\boldsymbol{x}_{2:n})) = h'(\mathcal{U}(\tilde{\boldsymbol{x}}_{2:n})) \right\}.$$

Observe that if a set $S \subseteq C'$ is relaxed $\mathcal{U}$-robustly shattered by $\mathcal{H}'$, then $S \cup \{z_1\}$ is also relaxed $\mathcal{U}$-robustly shattered by $\mathcal{H}'$ and vice versa. Observe also that, by definition, $Y_1 = \Pi^{\mathcal{U}}_{\mathcal{H}'}(z_2, \dots, z_n)$. By applying the inductive assumption on $\mathcal{H}'$ and $C'$ we obtain that

$$\begin{aligned} |Y_1| = \left|\Pi^{\mathcal{U}}_{\mathcal{H}'}(z_2, \dots, z_n)\right| &\leq |\{S \subseteq C' : S \text{ is relaxed } \mathcal{U}\text{-robustly shattered by } \mathcal{H}'\}| \\ &= |\{S \subseteq C' : S \cup \{z_1\} \text{ is relaxed } \mathcal{U}\text{-robustly shattered by } \mathcal{H}'\}| \\ &= |\{S \subseteq C : z_1 \in S \wedge S \text{ is relaxed } \mathcal{U}\text{-robustly shattered by } \mathcal{H}'\}| \\ &\leq |\{S \subseteq C : z_1 \in S \wedge S \text{ is relaxed } \mathcal{U}\text{-robustly shattered by } \mathcal{H}\}|. \end{aligned}$$

Overall, we have shown that

$$\begin{aligned}\left|\Pi^{\mathcal{U}}_{\mathcal{H}}(z_1,\dots,z_n)\right| &= |Y_0| + |Y_1| \\ &\leq |\{S \subseteq C : z_1 \notin S \wedge S \text{ is relaxed } \mathcal{U}\text{-robustly shattered by } \mathcal{H}\}| \\ &\quad + |\{S \subseteq C : z_1 \in S \wedge S \text{ is relaxed } \mathcal{U}\text{-robustly shattered by } \mathcal{H}\}| \\ &= |\{S \subseteq \{z_1,\dots,z_n\} : S \text{ is relaxed } \mathcal{U}\text{-robustly shattered by } \mathcal{H}\}|,\end{aligned}$$

which concludes our proof. □

### 9.5.1 Realizable Setting

*Proof of Theorem 9.1.* It suffices to show that

$$\Pr_{\substack{(\boldsymbol{x},\boldsymbol{y})\sim\mathcal{D}^n\\(\tilde{\boldsymbol{x}},\tilde{\boldsymbol{y}})\sim\mathcal{D}^n}}\left[\exists \boldsymbol{z}\in\mathcal{U}(\boldsymbol{x}), \exists\tilde{\boldsymbol{z}}\in\mathcal{U}(\tilde{\boldsymbol{x}}), \exists\hat{h}\in\Delta^{\mathcal{U}}_{\mathcal{H}}(\boldsymbol{z},\boldsymbol{y},\tilde{\boldsymbol{z}}) : \mathrm{err}_{\tilde{\boldsymbol{z}},\tilde{\boldsymbol{y}}}(\hat{h}) > \varepsilon\right] \leq \delta.$$

Observe that since $\mathsf{OPT}_{\mathcal{U}^{-1}(\mathcal{U})} = \inf_{h\in\mathcal{H}} \Pr_{(x,y)\sim\mathcal{D}}\left[\exists z\in\mathcal{U}(x), \exists\tilde{x}\in\mathcal{U}^{-1}(z) : h(\tilde{x})\neq y\right] = 0$, it holds by definition of $\Delta^{\mathcal{U}}_{\mathcal{H}}$ (see Equation 9.7) that the set $\Delta^{\mathcal{U}}_{\mathcal{H}}(\boldsymbol{z},\boldsymbol{y},\tilde{\boldsymbol{z}})$ is non-empty with probability 1.

We will first start with a standard observation stating that sampling two iid sequences of length $n$, $(\boldsymbol{x},\boldsymbol{y})\sim\mathcal{D}^n$ and $(\tilde{\boldsymbol{x}},\tilde{\boldsymbol{y}})\sim\mathcal{D}^n$, is equivalent to sampling a single iid sequence of length $2n$, $(\boldsymbol{x},\boldsymbol{y})\sim\mathcal{D}^{2n}$, and then randomly splitting it into two sequences of length $n$ (using a permutation $\sigma$ of $\{1,\dots,2n\}$ sampled uniformly at random). Thus, it follows that

$$\Pr_{\substack{(\boldsymbol{x},\boldsymbol{y})\sim\mathcal{D}^n\\(\tilde{\boldsymbol{x}},\tilde{\boldsymbol{y}})\sim\mathcal{D}^n}}\left[\exists \boldsymbol{z}\in\mathcal{U}(\boldsymbol{x}), \exists\tilde{\boldsymbol{z}}\in\mathcal{U}(\tilde{\boldsymbol{x}}), \exists\hat{h}\in\Delta^{\mathcal{U}}_{\mathcal{H}}(\boldsymbol{z},\boldsymbol{y},\tilde{\boldsymbol{z}}) : \mathrm{err}_{\tilde{\boldsymbol{z}},\tilde{\boldsymbol{y}}}(\hat{h}) > \varepsilon\right] = \mathop{\mathbb{E}}_{(\boldsymbol{x},\boldsymbol{y})\sim\mathcal{D}^{2n}}\left[\Pr_{\sigma}\left[E_{\sigma,\boldsymbol{z}} | (\boldsymbol{x},\boldsymbol{y})\right]\right],$$

where $\sigma$ is a permutation of $[2n]$ sampled uniformly at random and $E_{\sigma,\boldsymbol{z}}$ is defined as:

$$E_{\sigma,\boldsymbol{z}} = \left\{\exists \boldsymbol{z}_{\sigma(1:2n)}\in\mathcal{U}(\boldsymbol{x}_{\sigma(1:2n)}), \exists\hat{h}\in\Delta^{\mathcal{U}}_{\mathcal{H}}(\boldsymbol{z}_{\sigma(1:n)},\boldsymbol{y}_{\sigma(1:n)},\boldsymbol{z}_{\sigma(n+1:2n)}) : \mathrm{err}_{\boldsymbol{z}_{\sigma(n+1:2n)},\boldsymbol{y}_{\sigma(n+1:2n)}}(\hat{h}) > \varepsilon\right\}.$$

High error on $\boldsymbol{z}$'s implies high error on $\boldsymbol{x}$'s. It suffices to show that for any $(\boldsymbol{x},\boldsymbol{y})\sim\mathcal{D}^{2n}$ such that $\exists h^*\in\mathcal{H}$ with $h^*(\mathcal{U}^{-1}(\mathcal{U})(\boldsymbol{x})) = \boldsymbol{y}$ (which occurs with probability one): $\Pr_\sigma\left[E_{\sigma,\boldsymbol{z}}|(\boldsymbol{x},\boldsymbol{y})\right] \leq \delta$. To this end, we will start by showing that the event $E_{\sigma,\boldsymbol{z}}$ implies the

following event $E_{\sigma,\boldsymbol{x}}$:

$$E_{\sigma,\boldsymbol{x}} = \left\{ \exists \boldsymbol{z}_{\sigma(1:2n)} \in \mathcal{U}(\boldsymbol{x}_{\sigma(1:2n)}), \exists \hat{h} \in \Delta^{\mathcal{U}}_{\mathcal{H}}(\boldsymbol{z}_{\sigma(1:n)}, \boldsymbol{y}_{\sigma(1:n)}, \boldsymbol{z}_{\sigma(n+1:2n)}) : \mathrm{err}_{\boldsymbol{x}_{\sigma(n+1:2n)}, \boldsymbol{y}_{\sigma(n+1:2n)}}(\hat{h}) > \varepsilon \right\}.$$

In words, in case there are adversarial perturbations $\boldsymbol{z}_{\sigma(1:2n)} \in \mathcal{U}(\boldsymbol{x}_{\sigma(1:2n)})$ and a predictor $\hat{h} \in \Delta^{\mathcal{U}}_{\mathcal{H}}(\boldsymbol{z}_{\sigma(1:n)}, \boldsymbol{y}_{\sigma(1:n)}, \boldsymbol{z}_{\sigma(n+1:2n)})$ with many mistakes on the adversarial perturbations: $\mathrm{err}_{\boldsymbol{z}_{\sigma(n+1:2n)}, \boldsymbol{y}_{\sigma(n+1:2n)}}(\hat{h}) > \varepsilon$, then this implies that $\hat{h}$ makes many mistakes on the original non-adversarial test sequence: $\mathrm{err}_{\boldsymbol{x}_{\sigma(n+1:2n)}, \boldsymbol{y}_{\sigma(n+1:2n)}}(\hat{h}) > \varepsilon$. This is because for any $\boldsymbol{z}_{\sigma(1:2n)} \in \mathcal{U}(\boldsymbol{x}_{\sigma(1:2n)})$, by definition of $\Delta^{\mathcal{U}}_{\mathcal{H}}$ (see Equation 9.7), any $\hat{h} \in \Delta^{\mathcal{U}}_{\mathcal{H}}(\boldsymbol{z}_{\sigma(1:n)}, \boldsymbol{y}_{\sigma(1:n)}, \boldsymbol{z}_{\sigma(n+1:2n)})$ robustly labels the perturbations $\boldsymbol{z}_{\sigma(1:2n)}$: $\hat{h}(\mathcal{U}^{-1}(\boldsymbol{z}_{\sigma(1:2n)})) = \hat{h}(\boldsymbol{z}_{\sigma(1:2n)})$. That is,

$$(\forall 1 \leq i \leq 2n) \left(\forall \tilde{x} \in \mathcal{U}^{-1}(z_{\sigma(i)})\right) : \hat{h}(\tilde{x}) = \hat{h}(z_{\sigma(i)}).$$

By definition of $\mathcal{U}^{-1}$, it holds that $\boldsymbol{x}_{\sigma(1:2n)} \in \mathcal{U}^{-1}(\boldsymbol{z}_{\sigma(1:2n)})$. Thus, it follows that $\hat{h}(\boldsymbol{z}_{\sigma(n+1:2n)}) = \hat{h}(\boldsymbol{x}_{\sigma(n+1:2n)})$, and therefore, event $E_{\sigma,\boldsymbol{z}}$ implies event $E_{\sigma,\boldsymbol{x}}$.

Finite robust labelings on $\boldsymbol{x}$'s Based on the above, it suffices now to show that: $\Pr_\sigma\left[E_{\sigma,\boldsymbol{x}} | (\boldsymbol{x}, \boldsymbol{y})\right] \leq \delta$. To this end, we will show that for any permutation $\sigma$, any $\boldsymbol{z}_{\sigma(1:2n)} \in \mathcal{U}(\boldsymbol{x}_{\sigma(1:2n)})$, and any $\hat{h} \in \Delta^{\mathcal{U}}_{\mathcal{H}}(\boldsymbol{z}_{\sigma(1:n)}, \boldsymbol{y}_{\sigma(1:n)}, \boldsymbol{z}_{\sigma(n+1:2n)})$ it holds that the labeling $\hat{h}(\boldsymbol{x}_{\sigma(1:2n)})$ is included in a finite set of possible behaviors $\Pi^{\mathcal{U}}_{\mathcal{H}}$ defined on the entire sequence $\boldsymbol{x} = (x_1, \ldots, x_{2n})$ by:

$$\Pi^{\mathcal{U}}_{\mathcal{H}}(x_1, \ldots, x_{2n}) = \left\{ (h(x_1), h(x_2), \ldots, h(x_{2n})) \middle| \substack{\exists z_1 \in \mathcal{U}(x_1), \ldots, z_{2n} \in \mathcal{U}(x_{2n}), \\ \exists h \in \mathcal{H} : h(\mathcal{U}^{-1}(z_i)) = h(x_i) \forall 1 \leq i \leq 2n} \right\}$$

Consider an arbitrary permutation $\sigma$ and an arbitrary $\boldsymbol{z}_{\sigma(1:2n)} \in \mathcal{U}(\boldsymbol{x}_{\sigma(1:2n)})$. For any $\hat{h} \in \Delta^{\mathcal{U}}_{\mathcal{H}}(\boldsymbol{z}_{\sigma(1:n)}, \boldsymbol{y}_{\sigma(1:n)}, \boldsymbol{z}_{\sigma(n+1:2n)})$, by definition of $\Delta^{\mathcal{U}}_{\mathcal{H}}$, it holds that $\hat{h}\left(\mathcal{U}^{-1}(\boldsymbol{z}_{\sigma(1:n)})\right) = \boldsymbol{y}_{\sigma(1:n)}$ and $\hat{h}(\mathcal{U}^{-1}(\boldsymbol{z}_{\sigma(n+1:2n)})) = \hat{h}(\boldsymbol{z}_{\sigma(n+1:2n)})$. Therefore, $\boldsymbol{z}_{\sigma(1:2n)} \in \mathcal{U}(\boldsymbol{x}_{\sigma(1:2n)})$ and the predictor $\hat{h} \in \mathcal{H}$ are witnesses that satisfy the following:

$$(\forall 1 \leq i \leq 2n) : z_{\sigma(i)} \in \mathcal{U}(x_{\sigma(i)}) \wedge \hat{h}(\mathcal{U}^{-1}(z_{\sigma(i)})) = \hat{h}(x_{\sigma(i)}).$$

Thus, by definition of $\Pi^{\mathcal{U}}_{\mathcal{H}}$, it holds that $\hat{h}(\boldsymbol{x}_{\sigma(1:2n)}) \in \Pi^{\mathcal{U}}_{\mathcal{H}}(x_1, \ldots, x_{2n})$. This allows us to establish that the event $E_{\sigma,\boldsymbol{x}}$ implies the event that there exists a labeling $\hat{h}(\boldsymbol{x}_{\sigma(1:2n)}) \in \Pi^{\mathcal{U}}_{\mathcal{H}}(x_1, \ldots, x_{2n})$ that achieves zero loss on the training examples $\mathrm{err}_{\boldsymbol{x}_{\sigma(1:n)}, \boldsymbol{y}_{\sigma(1:n)}}(\hat{h}) = 0$, but makes error more than $\varepsilon$ on the test examples $\mathrm{err}_{\boldsymbol{x}_{\sigma(n+1:2n)},\boldsymbol{y}_{\sigma(n+1:2n)}}(\hat{h}) > \varepsilon$. Specifically,

$$\begin{aligned}\Pr_{\sigma}\left[E_{\sigma,\boldsymbol{x}}\right] &\leq \Pr_{\sigma}\left[\exists \hat{h} \in \Pi_{\mathcal{H}}^{\mathcal{U}}(x_1,\ldots,x_{2n}) : \mathrm{err}_{\boldsymbol{x}_{\sigma(1:n)},\boldsymbol{y}_{\sigma(1:n)}}(\hat{h}) = 0 \wedge \mathrm{err}_{\boldsymbol{x}_{\sigma(n+1:2n)},\boldsymbol{y}_{\sigma(n+1:2n)}}(\hat{h}) > \varepsilon\right] \\ &\overset{(i)}{\leq} \left|\Pi_{\mathcal{H}}^{\mathcal{U}}(x_1,\ldots,x_{2n})\right| 2^{\lceil -\varepsilon n\rceil} \overset{(ii)}{\leq} (2n)^{\mathrm{rdim}_{\mathcal{U}^{-1}}(\mathcal{H})}\, 2^{\lceil -\varepsilon n\rceil},\end{aligned}$$

where inequality $(i)$ follows from applying a union bound over labelings $\hat{h} \in \Pi_{\mathcal{H}}^{\mathcal{U}}(x_1,\ldots,x_{2n})$, and observing that for any such fixed $\hat{h}$:

$$\Pr_{\sigma}\left[\mathrm{err}_{\boldsymbol{x}_{\sigma(1:n)},\boldsymbol{y}_{\sigma(1:n)}}(\hat{h}) = 0 \wedge \mathrm{err}_{\boldsymbol{x}_{\sigma(n+1:2n)},\boldsymbol{y}_{\sigma(n+1:2n)}}(\hat{h}) > \varepsilon\right] \leq 2^{-\lceil \varepsilon n\rceil}.$$

To see this, suppose that $s = \sum_{i=1}^{2n} \mathbb{1}\{\hat{h}(x_i) \neq y_i\} \geq \lceil \varepsilon n\rceil$ (otherwise, the probability of the event above is zero). Now, when sampling a random permutation $\sigma$, the chance that all of the mistakes fall into the test split is at most $2^{-s} \leq 2^{-\lceil \varepsilon n\rceil}$. Because if we pair the $s$ mistakes and any $s$ out of the $2n - s$ non-mistakes while fixing the remaining non-mistakes to be in the training split, then the chance that all the $s$ mistakes appear in the test split is at most $2^{-s}$.

Finally, inequality $(ii)$ follows from applying Sauer's lemma on our introduced relaxed notion of robust shattering dimension (Definition 9.1). Setting $(2n)^{\mathrm{rdim}_{\mathcal{U}^{-1}}(\mathcal{H})}\, 2^{\lceil -\varepsilon n\rceil} \leq \delta$ and solving for $\varepsilon$ yields the stated bound. □

### 9.5.2 Agnostic Setting

*Proof of Theorem 9.2.* Let $n \in \mathbb{N}$. For notational brevity, we write $\mathsf{OPT} = \mathsf{OPT}_{\mathcal{U}^{-1}(\mathcal{U})}$. We will assume that $\mathsf{OPT}$ (see Equation 9.2) is attained by some predictor $h^* \in \mathcal{H}$[§]. For this fixed $h^*$, observe that by a standard Hoeffding bound, for $\varepsilon_0 = \sqrt{\frac{\ln(2/\delta)}{2n}}$, it holds that

$$\Pr_{\substack{(\boldsymbol{x},\boldsymbol{y})\sim\mathcal{D}^n \\ (\tilde{\boldsymbol{x}},\tilde{\boldsymbol{y}})\sim\mathcal{D}^n}}\left[\left(\mathrm{R}_{\mathcal{U}^{-1}(\mathcal{U})}(h^*;\boldsymbol{x},\boldsymbol{y}) \leq \mathsf{OPT} + \varepsilon_0\right) \wedge \left(\mathrm{R}_{\mathcal{U}^{-1}(\mathcal{U})}(h^*;\tilde{\boldsymbol{x}},\tilde{\boldsymbol{y}}) \leq \mathsf{OPT} + \varepsilon_0\right)\right] \geq 1 - \delta.$$

By definition of $\mathrm{R}_{\mathcal{U}^{-1}(\mathcal{U})}$ (see Equation 9.5), this implies that

$$\Pr_{\substack{(\boldsymbol{x},\boldsymbol{y})\sim\mathcal{D}^n \\ (\tilde{\boldsymbol{x}},\tilde{\boldsymbol{y}})\sim\mathcal{D}^n}}\left[\left(\forall \boldsymbol{z} \in \mathcal{U}(\boldsymbol{x}) : \mathrm{R}_{\mathcal{U}^{-1}}(h^*;\boldsymbol{z},\boldsymbol{y}) \leq \mathsf{OPT} + \varepsilon_0\right) \wedge \left(\forall \tilde{\boldsymbol{z}} \in \mathcal{U}(\tilde{\boldsymbol{x}}) : \mathrm{R}_{\mathcal{U}^{-1}}(h^*;\tilde{\boldsymbol{z}}) \leq \mathsf{OPT} + \varepsilon_0\right)\right] \geq 1-\delta.$$

[§]Otherwise, we can always choose a predictor $h^* \in \mathcal{H}$ attaining $\mathsf{OPT} + \varepsilon'$ for any small $\varepsilon' > 0$.

This implies that

$$\Pr_{\substack{(\boldsymbol{x},\boldsymbol{y})\sim\mathcal{D}^n\\(\tilde{\boldsymbol{x}},\tilde{\boldsymbol{y}})\sim\mathcal{D}^n}}\left[\forall \boldsymbol{z}\in\mathcal{U}(\boldsymbol{x}),\forall\tilde{\boldsymbol{z}}\in\mathcal{U}(\tilde{\boldsymbol{x}}):\min_{h\in\mathcal{H}}\max\left\{\mathrm{R}_{\mathcal{U}^{-1}}(h;\boldsymbol{z},\boldsymbol{y}),\mathrm{R}_{\mathcal{U}^{-1}}(h;\tilde{\boldsymbol{z}})\right\}\leq \mathsf{OPT}+\varepsilon_0\right]\geq 1-\delta.$$

By Equation 9.8, we have

$$\Pr_{\substack{(\boldsymbol{x},\boldsymbol{y})\sim\mathcal{D}^n\\(\tilde{\boldsymbol{x}},\tilde{\boldsymbol{y}})\sim\mathcal{D}^n}}\left[\forall \boldsymbol{z}\in\mathcal{U}(\boldsymbol{x}),\forall\tilde{\boldsymbol{z}}\in\mathcal{U}(\tilde{\boldsymbol{x}}),\forall\hat{h}\in\Delta^{\mathcal{U}}_{\mathcal{H}}(\boldsymbol{z},\boldsymbol{y},\tilde{\boldsymbol{z}}):\max\left\{\mathrm{R}_{\mathcal{U}^{-1}}(\hat{h};\boldsymbol{z},\boldsymbol{y}),\mathrm{R}_{\mathcal{U}^{-1}}(\hat{h};\tilde{\boldsymbol{z}})\right\}\leq \mathsf{OPT}+\varepsilon_0\right]\geq 1-\delta.$$

That is, we have established that with high probability over the drawings of $(\boldsymbol{x},\boldsymbol{y}),(\tilde{\boldsymbol{x}},\tilde{\boldsymbol{y}})\sim\mathcal{D}^n$: for any adversarial perturbations $\boldsymbol{z}\in\mathcal{U}(\boldsymbol{x}),\tilde{\boldsymbol{z}}\in\mathcal{U}(\tilde{\boldsymbol{x}})$, and any predictor $\hat{h}\in\Delta^{\mathcal{U}}_{\mathcal{H}}(\boldsymbol{z},\boldsymbol{y},\tilde{\boldsymbol{z}})$:

1. $\hat{h}$ achieves low robust error on the training examples: $\mathrm{R}_{\mathcal{U}^{-1}}(\hat{h};\boldsymbol{z},\boldsymbol{y})\leq\mathsf{OPT}+\varepsilon_0$.
2. $\hat{h}$ is robust (but not necessarily correct) on many of the test examples: $\mathrm{R}_{\mathcal{U}^{-1}}(\hat{h};\tilde{\boldsymbol{z}})\leq\mathsf{OPT}+\varepsilon_0$.

VC Guarantee Next, to show that $\hat{h}$ achieves low error on the test examples $\tilde{\boldsymbol{z}}$, we will combine the properties above with a standard guarantee in the transductive setting from VC theory, which states that for $\varepsilon=O\left(\sqrt{\frac{\mathrm{vc}(\mathcal{H})+\log(1/\delta)}{n}}\right)$:

$$\Pr_{\substack{(\boldsymbol{x},\boldsymbol{y})\sim\mathcal{D}^n\\(\tilde{\boldsymbol{x}},\tilde{\boldsymbol{y}})\sim\mathcal{D}^n}}\left[\forall h\in\mathcal{H}:\left|\mathrm{err}_{\boldsymbol{x},\boldsymbol{y}}(h)-\mathrm{err}_{\tilde{\boldsymbol{x}},\tilde{\boldsymbol{y}}}(h)\right|\leq\varepsilon\right]\geq 1-\delta.$$

Thus, for $\varepsilon=O\left(\sqrt{\frac{\mathrm{vc}(\mathcal{H})+\log(1/\delta)}{n}}\right)$:

$$\Pr_{\substack{(\boldsymbol{x},\boldsymbol{y})\sim\mathcal{D}^n\\(\tilde{\boldsymbol{x}},\tilde{\boldsymbol{y}})\sim\mathcal{D}^n}}\left[\begin{array}{l}\forall \boldsymbol{z}\in\mathcal{U}(\boldsymbol{x}),\forall\tilde{\boldsymbol{z}}\in\mathcal{U}(\tilde{\boldsymbol{x}}),\forall\hat{h}\in\Delta^{\mathcal{U}}_{\mathcal{H}}(\boldsymbol{x},\boldsymbol{y},\boldsymbol{z}):\\ \max\left\{\mathrm{R}_{\mathcal{U}^{-1}}(\hat{h};\boldsymbol{z},\boldsymbol{y}),\mathrm{R}_{\mathcal{U}^{-1}}(\hat{h};\tilde{\boldsymbol{z}})\right\}\leq\mathsf{OPT}+\varepsilon\wedge\left|\mathrm{err}_{\boldsymbol{x},\boldsymbol{y}}(\hat{h})-\mathrm{err}_{\tilde{\boldsymbol{x}},\tilde{\boldsymbol{y}}}(\hat{h})\right|\leq\varepsilon\end{array}\right]\geq 1-2\delta.$$

Finally, observe that for any predictor $\hat{h} \in \Delta^{\mathcal{U}}_{\mathcal{H}}(\boldsymbol{z}, \boldsymbol{y}, \tilde{\boldsymbol{z}})$ satisfying $\max\left\{\mathrm{R}_{\mathcal{U}^{-1}}(\hat{h}; \boldsymbol{z}, \boldsymbol{y}), \mathrm{R}_{\mathcal{U}^{-1}}(\hat{h}; \tilde{\boldsymbol{z}})\right\} \leq \mathsf{OPT} + \varepsilon$ and $|\operatorname{err}_{\boldsymbol{x},\boldsymbol{y}}(\hat{h}) - \operatorname{err}_{\tilde{\boldsymbol{x}},\tilde{\boldsymbol{y}}}(\hat{h})| \leq \varepsilon$, we can deduce that:

- $\operatorname{err}_{\boldsymbol{x},\boldsymbol{y}}(\hat{h}) \leq \mathsf{OPT} + \varepsilon$ (since $\mathrm{R}_{\mathcal{U}^{-1}}(\hat{h}; \boldsymbol{z}, \boldsymbol{y}) \leq \mathsf{OPT} + \varepsilon$ and $\boldsymbol{x} \in \mathcal{U}^{-1}(\boldsymbol{z})$). Therefore, $\operatorname{err}_{\tilde{\boldsymbol{x}},\tilde{\boldsymbol{y}}}(\hat{h}) \leq \mathsf{OPT} + 2\varepsilon$ (since $|\operatorname{err}_{\boldsymbol{x},\boldsymbol{y}}(\hat{h}) - \operatorname{err}_{\tilde{\boldsymbol{x}},\tilde{\boldsymbol{y}}}(\hat{h})| \leq \varepsilon$).
- Since $\operatorname{err}_{\tilde{\boldsymbol{x}},\tilde{\boldsymbol{y}}}(\hat{h}) \leq \mathsf{OPT} + 2\varepsilon$ and $\mathrm{R}_{\mathcal{U}^{-1}}(\hat{h}; \boldsymbol{z}) \leq \mathsf{OPT} + \varepsilon$, this implies that $\operatorname{err}_{\boldsymbol{z},\tilde{\boldsymbol{y}}}(\hat{h}) \leq 2\mathsf{OPT} + 3\varepsilon$.

A refined bound  We will show that

$$\Pr_{\substack{(\boldsymbol{x},\boldsymbol{y})\sim\mathcal{D}^n \\ (\tilde{\boldsymbol{x}},\tilde{\boldsymbol{y}})\sim\mathcal{D}^n}}\Bigg[\forall \boldsymbol{z} \in \mathcal{U}(\boldsymbol{x}), \forall \tilde{\boldsymbol{z}} \in \mathcal{U}(\tilde{\boldsymbol{x}}), \forall \hat{h} \in \Delta^{\mathcal{U}}_{\mathcal{H}}(\boldsymbol{z}, \boldsymbol{y}, \tilde{\boldsymbol{z}}) : \\ \max\left\{\mathrm{R}_{\mathcal{U}^{-1}}(\hat{h}; \boldsymbol{z}, \boldsymbol{y}), \mathrm{R}_{\mathcal{U}^{-1}}(\hat{h}; \tilde{\boldsymbol{z}})\right\} \leq \mathsf{OPT} + \tilde{\varepsilon} \wedge \left|\operatorname{err}_{\boldsymbol{x},\boldsymbol{y}}(\hat{h}) - \operatorname{err}_{\tilde{\boldsymbol{x}},\tilde{\boldsymbol{y}}}(\hat{h})\right| \leq \tilde{\varepsilon}\Bigg] \geq 1 - \delta,$$

for a smaller $\tilde{\varepsilon}$ that scales with $\mathsf{OPT}$ and $\operatorname{rdim}_{\mathcal{U}^{-1}}(\mathcal{H})$ (instead of $\operatorname{vc}(\mathcal{H})$). To this end, it suffices to show that for any fixed $(\boldsymbol{x}, \boldsymbol{y}) \sim \mathcal{D}^{2n}$:

$$\Pr_{\sigma}\left[\begin{array}{l} \forall \boldsymbol{z}_{\sigma(1:2n)} \in \mathcal{U}(\boldsymbol{x}_{\sigma(1:2n)}), \forall \hat{h} \in \Delta^{\mathcal{U}}_{\mathcal{H}}(\boldsymbol{z}_{\sigma(1:n)}, \boldsymbol{y}_{\sigma(1:n)}, \boldsymbol{z}_{\sigma(n+1:2n)}) : \\ \max\left\{\mathrm{R}_{\mathcal{U}^{-1}}(\hat{h}; \boldsymbol{z}_{\sigma(1:n)}, \boldsymbol{y}_{\sigma(1:n)}), \mathrm{R}_{\mathcal{U}^{-1}}(\hat{h}; \boldsymbol{z}_{\sigma(n+1,2n)})\right\} \leq \mathsf{OPT} + \tilde{\varepsilon} \wedge \left|\operatorname{err}_{\boldsymbol{x},\boldsymbol{y}}(\hat{h}) - \operatorname{err}_{\tilde{\boldsymbol{x}},\tilde{\boldsymbol{y}}}(\hat{h})\right| \leq \tilde{\varepsilon} \end{array}\right] \geq 1 - \delta,$$

where $\sigma$ is a permutation of $\{1, 2, 3, \ldots, 2n\}$ sampled uniformly at random.
We will show that for any permutation $\sigma$, any $\boldsymbol{z}_{\sigma(1:2n)} \in \mathcal{U}(\boldsymbol{x}_{\sigma(1:2n)})$, and any $\hat{h} \in \Delta^{\mathcal{U}}_{\mathcal{H}}(\boldsymbol{z}_{\sigma(1:n)}, \boldsymbol{y}_{\sigma(1:n)}, \boldsymbol{z}_{\sigma(n+1:2n)})$ it holds that the labeling $\hat{h}(\boldsymbol{x}_{\sigma(1:2n)})$ is included in a finite set of possible behaviors $\Pi^{\mathcal{U}}_{\mathcal{H}}$ defined on the entire sequence $\boldsymbol{x} = (x_1, \ldots, x_{2n})$ by:

$$\Pi^{\mathcal{U}}_{\mathcal{H}}(x_1, \ldots, x_{2n}) = \left\{ (h(x_1), h(x_2), \ldots, h(x_{2n})) \,\middle|\, \begin{array}{c} \exists I \subseteq [2n], |I| \geq (1 - \mathsf{OPT} - \varepsilon_0)2n, \\ \forall i \in I, \exists z_i \in \mathcal{U}(x_i), \\ \exists h \in \mathcal{H} : h(\mathcal{U}^{-1}(z_i)) = h(x_i) \forall i \in I \end{array} \right\}$$

When it holds that $\mathrm{R}_{\mathcal{U}^{-1}}(\hat{h}; \boldsymbol{z}_{\sigma(1:n)}, \boldsymbol{y}_{\sigma(1:n)}) \leq \mathsf{OPT} + \varepsilon_0$ and $\mathrm{R}_{\mathcal{U}^{-1}}(\hat{h}; \boldsymbol{z}_{\sigma(n+1:2n)}) \leq \mathsf{OPT} + \varepsilon_0$, by definition of $\mathrm{R}_{\mathcal{U}^{-1}}$ (see Equation 9.5), it follows that the predictor $\hat{h} \in \mathcal{H}$ and $\boldsymbol{z}_{\sigma(1:2n)} \in$

$\mathcal{U}(\boldsymbol{x}_{\sigma(1:2n)})$ are witnesses that satisfy the following:

$$(\exists I \subseteq [2n], |I| \geq (1 - \mathsf{OPT} - \varepsilon)2n)\ (\forall i \in I) : z_{\sigma(i)} \in \mathcal{U}(x_{\sigma(i)}) \wedge \hat{h}(\mathcal{U}^{-1}(z_{\sigma(i)})) = \hat{h}(x_{\sigma(i)}).$$

Thus, by definition of $\Pi^{\mathcal{U}}_{\mathcal{H}}$, it holds that $\hat{h}(\boldsymbol{x}_{\sigma(1:2n)}) \in \Pi^{\mathcal{U}}_{\mathcal{H}}(x_1, \ldots, x_{2n})$. Then, observe that for any $\tilde{\varepsilon} > 0$

$$\begin{aligned}
&\Pr_{\sigma}\left[\exists \hat{h} \in \Pi^{\mathcal{U}}_{\mathcal{H}}(x_1, \ldots, x_{2n}) : \left|\mathrm{err}_{\boldsymbol{x}_{\sigma(1:n)}, \boldsymbol{y}_{\sigma(1:n)}}(\hat{h}) - \mathrm{err}_{\boldsymbol{x}_{\sigma(n+1:2n)}, \boldsymbol{y}_{\sigma(n+1:2n)}}(\hat{h})\right| > \tilde{\varepsilon}\right] \\
&\overset{(i)}{\leq} \left|\Pi^{\mathcal{U}}_{\mathcal{H}}(x_1, \ldots, x_{2n})\right| e^{-\tilde{\varepsilon}^2 n} \overset{(ii)}{\leq} \binom{2n}{\leq (\mathsf{OPT} + \varepsilon_0)2n} \binom{(1 - \mathsf{OPT} - \varepsilon_0)2n}{\leq \mathrm{rdim}_{\mathcal{U}^{-1}}(\mathcal{H})} e^{-\tilde{\varepsilon}^2 n} \\
&\overset{(iii)}{\leq} 2^{H(\mathsf{OPT}+\varepsilon_0)2n} \left((1 - \mathsf{OPT} - \varepsilon_0)2n\right)^{\mathrm{rdim}_{\mathcal{U}^{-1}}(\mathcal{H})} e^{-\tilde{\varepsilon}^2 n},
\end{aligned}$$

where inequality $(i)$ follows from applying a union bound over labelings $\hat{h} \in \Pi^{\mathcal{U}}_{\mathcal{H}}(x_1, \ldots, x_{2n})$, and observing that for any such fixed $\hat{h}$:

$$\Pr_{\sigma}\left[\left|\mathrm{err}_{\boldsymbol{x}_{\sigma(1:n)}, \boldsymbol{y}_{\sigma(1:n)}}(\hat{h}) - \mathrm{err}_{\boldsymbol{x}_{\sigma(n+1:2n)}, \boldsymbol{y}_{\sigma(n+1:2n)}}(\hat{h})\right| > \tilde{\varepsilon}\right] \leq e^{-\tilde{\varepsilon}^2 n}.$$

Inequality $(ii)$ follows from the definition of $\Pi^{\mathcal{U}}_{\mathcal{H}}$ and applying Sauer's lemma on our introduced relaxed notion of robust shattering dimension (Lemma 9.2). Inequality $(iii)$ follows from bounds on the binomial coefficients, where $H$ is the entropy function.

Setting $2^{H(\mathsf{OPT}+\varepsilon_0)2n} \left((1 - \mathsf{OPT} - \varepsilon_0)2n\right)^{\mathrm{rdim}_{\mathcal{U}^{-1}}(\mathcal{H})} e^{-\tilde{\varepsilon}^2 n}$ less than $\frac{\delta}{2}$ and solving for $\tilde{\varepsilon}$ yields:

$$\begin{aligned}
\tilde{\varepsilon} &\leq \sqrt{2\ln(2)H(\mathsf{OPT} + \varepsilon_0) + \frac{\mathrm{rdim}_{\mathcal{U}^{-1}}(\mathcal{H}) \ln\left((1 - \mathsf{OPT} - \varepsilon_0)2n\right) + \ln(1/\delta)}{n}} \\
&\leq \sqrt{2\ln(2)H(\mathsf{OPT} + \varepsilon_0)} + \sqrt{\frac{\mathrm{rdim}_{\mathcal{U}^{-1}}(\mathcal{H}) \ln\left((1 - \mathsf{OPT} - \varepsilon_0)2n\right) + \ln(1/\delta)}{n}} \\
&\leq \mathsf{OPT} + \varepsilon_0 + \sqrt{\frac{\mathrm{rdim}_{\mathcal{U}^{-1}}(\mathcal{H}) \ln(2n) + \ln(1/\delta)}{n}}
\end{aligned}$$

Combining both events from above, we get that $\mathrm{err}_{\tilde{\boldsymbol{z}}, \tilde{\boldsymbol{y}}}(\hat{h}) \leq \mathrm{err}_{\tilde{\boldsymbol{x}}, \tilde{\boldsymbol{y}}}(\hat{h}) + \mathrm{R}_{\mathcal{U}^{-1}}(\hat{h}; \tilde{\boldsymbol{z}}) \leq \mathrm{err}_{\boldsymbol{x}, \boldsymbol{y}}(\hat{h}) + \tilde{\varepsilon} + \mathrm{R}_{\mathcal{U}^{-1}}(\hat{h}; \tilde{\boldsymbol{z}}) \leq \mathsf{OPT} + \varepsilon_0 + \tilde{\varepsilon} + \mathsf{OPT} + \varepsilon_0 = 2\mathsf{OPT} + 2\varepsilon_0 + \tilde{\varepsilon} \leq 3\mathsf{OPT} + 3\varepsilon_0 + \sqrt{\frac{\mathrm{rdim}_{\mathcal{U}^{-1}}(\mathcal{H}) \ln(2n) + \ln(1/\delta)}{n}}$. □

## 9.6 Discussion and Open Questions

Related Work  Adversarially robust learning has been mainly studied in the inductive setting (see e.g., Schmidt, Santurkar, Tsipras, Talwar, and Madry, 2018, Cullina, Bhagoji, and Mittal, 2018a, Khim and Loh, 2018, Bubeck, Lee, Price, and Razenshteyn, 2019, Yin, Ramchandran, and Bartlett, 2019, Montasser, Hanneke, and Srebro, 2019). This includes studying what learning rules should be used for robust learning and how much training data is needed to guarantee low robust error. It is now known that any hypothesis class $\mathcal{H}$ with finite VC dimension is robustly learnable, though sometimes improper learning is necessary and the best known upper bound on the robust error is exponential in the VC dimension (see results in Chapter 3).

In transductive learning, the learner is given unlabeled test examples to classify all at once or in batches, rather than individually (Vapnik, 1998). Without robustness guarantees, it is known that ERM is nearly minimax optimal in the transductive setting (Vapnik and Chervonenkis, 1974, Blumer, Ehrenfeucht, Haussler, and Warmuth, 1989a, Tolstikhin and Lopez-Paz, 2016). In particular, additional unlabeled test data does not offer any help from a minimax perspective. More recently, Goldwasser, Kalai, Kalai, and Montasser (2020) gave a transductive learning algorithm that takes as input labeled training examples from a distribution $\mathcal{D}$ and *arbitrary* unlabeled test examples (chosen by an unbounded adversary, not necessarily according to perturbation set $\mathcal{U}$). For classes $\mathcal{H}$ of bounded VC dimension, their algorithm guarantees low error rate on the test examples but it might *abstain* from classifying some (or perhaps even all) of them. This is different from the guarantees we present in this chapter, where we restrict the adversary to choose from a perturbation set $\mathcal{U}$ but we do *not* abstain from classifying.

On the empirical side, recently Wu, Yuan, and Wu (2020) proposed a method that leverages unlabeled test data for adversarial robustness in the context of deep neural networks. However, Chen, Guo, Wu, Li, Lao, Liang, and Jha (2021) later proposed an empirical attack that breaks their defense. Furthermore, Chen, Guo, Wu, Li, Lao, Liang, and Jha (2021) proposed another empirical transductive defense but with no theoretical guarantees. In semi-supervised learning, recent work by Alayrac, Uesato, Huang, Fawzi, Stanforth, and Kohli (2019), Carmon, Raghunathan, Schmidt, Duchi, and Liang (2019a) has shown that (non-adversarial) unlabeled test data can improve adversarially robust generalization in practice, and there is also theoretical work quantifying the benefit of unlabeled data for robust generalization (Ashtiani, Pathak, and Urner, 2020).

Open Questions Can we design transductive learners that compete with $\mathsf{OPT}_{\mathcal{U}}$ instead of $\mathsf{OPT}_{\mathcal{U}^{-1}(\mathcal{U})}$? We note that this will likely require more sophistication, in the sense that we can construct classes $\mathcal{H}$ with $\mathrm{vc}(\mathcal{H}) = 1$ (similar construction in Theorem 3.1) and distributions $\mathcal{D}$ where $\mathsf{OPT}_{\mathcal{U}} = 0$ but $\mathsf{OPT}_{\mathcal{U}^{-1}(\mathcal{U})} = 1$, and moreover, our simple transductive learner fails and finding a robust labeling on the test examples no longer suffices.

At the expense of competing with $\mathsf{OPT}_{\mathcal{U}^{-1}(\mathcal{U})}$, can we obtain stronger robust learning guarantees in the inductive setting, similar to the transductive guarantees established in this chapter? As we discuss in Section 9.4, we can not obtain such guarantees by directly reducing to the transductive problem, and we need improper learning because proper learning will not work.

# 10

# Beyond Perturbations: Learning Guarantees with Arbitrary Adversarial Test Examples

## 10.1 Introduction

Thus far in this thesis, we focused exclusively on adversarial learning when the adversary is restricted to some perturbation set $\mathcal{U}$, which specifies the possible perturbations that the adversary can attack with at test-time. We used this generic framing to capture interesting challenges in practice such as learning predictors robust to imperceptible $\ell_\infty$ perturbations. However, perturbations do not cover all types of test examples, and in this chapter we explore robust learning guarantees beyond perturbations.

Consider binary classification where test examples are not from the training distribution. Specifically, consider learning a binary function $f : \mathcal{X} \to \{0, 1\}$ where training examples are assumed to be i.i.d. from a distribution $P$ over $\mathcal{X}$, while the test examples are *arbitrary*. This includes both the possibility that test examples are chosen by an *unrestricted* adversary or that they are drawn from a distribution $Q \neq P$ (sometimes called "covariate shift"). For a disturbing example of covariate shift, consider learning to classify abnormal lung scans. A system trained on scans prior to 2019 may miss abnormalities due to COVID-19 since there were none in the training data. Fang et al. (2020) find noticeable signs of COVID in many scans. As a troubling adversarial example, consider explicit content detectors which are trained to classify normal vs. explicit images. Adversarial spammers synthesize endless variations of explicit images that evade these detectors for purposes such as advertising and phishing, this includes using conspicuous image distortion techniques (e.g., overlaying a large colored rectangle on an image) which goes beyond imperceptible perturbations (Yuan et al., 2019).

In general, there are several reasons why learning with arbitrary test examples is actually impossible. First of all, one may not be able to predict the labels of test examples that are far from training examples, as illustrated by the examples in group (1) of Figure 10.1. Secondly, as illustrated by group (2), given any classifier $h$, an adversary or test distribution $Q$ may concentrate on or near an error. High error rates are thus unavoidable since an adversary can simply repeat any single erroneous example they can find. This could also arise naturally, as in the COVID example, if $Q$ contains a concentration of new examples near one another–individually they appear "normal" (but are suspicious as a group). This is true even under the standard *realizable* assumption that the target function $f \in \mathcal{H}$ is in a known class $\mathcal{H}$ of bounded VC dimension $d = \text{vc}(\mathcal{H})$.

As we now argue, learning with arbitrary test examples requires *selective classifiers* and *transductive learning*, which have each been independently studied extensively. We refer to the combination as classification with *redaction*, a term which refers to the removal/obscuring of certain information when documents are released. A *selective classifier* (SC) is one which is allowed to abstain from predicting on some examples. In particular, it specifies both a classifier $h$ and a subset $S \subseteq \mathcal{X}$ of examples to classify, and rejects the rest. Equivalently, one can think of a SC as $h|_S : \mathcal{X} \to \{0, 1, ?\}$ where ? indicates $x \notin S$, abstinence.

$$h|_S(x) := \begin{cases} h(x) & \text{if } x \in S \\ ? & \text{if } x \notin S. \end{cases}$$

We say the learner *classifies* $x$ if $x \in S$ and otherwise it rejects $x$. Following standard terminology, if $x \notin S$ (i.e., $h|_S(x) = ?$) we say the classifier *rejects* $x$ (the term is not meant to indicate anything negative about the example $x$ but merely that its classification may be unreliable). We sat that $h|_S$ *misclassifies* or *errs* on $x$ if $h|_S(x) = 1 - f(x)$. There is a long literature on SCs, starting with the work of Chow (1957) on character recognition. In standard classification, *transductive learning* refers to the simple learning setting where the goal is to classify a given unlabeled test set that is presented together with the training examples (see e.g., Vapnik, 1998). We will also consider the generalization error of the learned classifier.

This raises the question: *When are unlabeled test examples available in advance?* In some applications, test examples are classified all at once (or in batches). Otherwise, redaction can also be beneficial *in retrospect*. For instance, even if image classifications are necessary immediately, an offensive image detector may be run daily with rejections flagged for inspection; and images may later be blocked if they are deemed offensive. Similarly, if a group of unusual lung scans showing COVID were detected after a period of time, the recognition of the new disease could be valuable even in hindsight. Furthermore, in some applications, one cannot simply label a sample of test examples. For instance, in learning to classify messages on an online platform, test data may contain both public and private data while training data may consist only of public messages. Due to privacy concerns, labeling data from the actual test distribution may be prohibited.

It is clear that a SC is necessary to guarantee few test misclassifications, e.g., if $P$ is concentrated on a single point $x$, rejection is necessary to guarantee few errors on arbitrary test points. However, no prior guarantees (even statistical guarantees) were known even for learning elementary classes such as intervals or halfspaces with arbitrary $P \neq Q$. This is because learning such classes is impossible without unlabeled examples.

To illustrate how redaction (transductive SC) is useful, consider learning an interval $[a, b]$ on $\mathcal{X} = \mathbb{R}$ with arbitrary $P \neq Q$. This is illustrated below with (blue) dots indicating test examples:

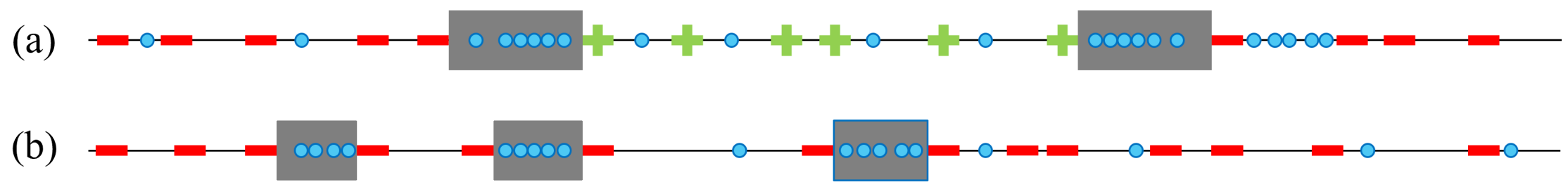

With positive training examples as in (a), one can guarantee 0 test errors by rejecting the two (grey) regions adjacent to the positive examples. When there are no positive training examples,[*] as in (b), one can guarantee $\leq k$ test errors by rejecting any region with $> k$ test exam-ples and no training examples; and predicting negative elsewhere. Of course, one can guarantee 0 errors by rejecting everywhere, but that would mean rejecting even future examples distributed like $P$. While our error objective will be an $\varepsilon$ test error rate, our rejection objective will be more subtle since we cannot absolutely bound the test rejection rate. Indeed, as illustrated above, in some cases one should reject many test examples.

[*]Learning with an all-negative training set (trivial in standard learning) is a useful "anomaly detection" set-

Note that our redaction model assumes that the target function $f$ remains the same at train and test times. This assumption holds in several (but not all) applications of interest. *Label shift* problems where $f$ changes from train to test is also important but not addressed here. Our focus is primarily the well-studied realizable setting, where $f \in \mathcal{H}$, though we analyze an agnostic setting as well.

A NOTE OF CAUTION. Inequities may be caused by using training data that differs from the test distribution on which the classifier is used. For instance, in classifying a person's gender from a facial image, Buolamwini and Gebru (2018) have demonstrated that commercial classifiers are highly inaccurate on dark-skinned faces, likely because they were trained on light-skinned faces. In such cases, it is preferable to collect a more diverse training sample even if it comes at greater expense, or in some cases to abstain from using machine learning altogether. In such cases, $PQ$ learning should *not* be used, as an unbalanced distribution of rejections can also be harmful.[†]

### 10.1.1 Redaction Model and Guarantees

Our goal is to learn a target function $f \in \mathcal{H}$ of VC dimension $d$ with training distribution $P$ over $\mathcal{X}$. In the redaction model, the learner first chooses $h \in \mathcal{H}$ based on $n$ i.i.d. training examples $\boldsymbol{x} \sim \mathcal{X}^n$ and their labels $f(\boldsymbol{x}) = \big(f(x_1), f(x_2), \ldots, f(x_n)\big) \in \{0,1\}^n$. In other words, it trains a standard binary classifier. Next, a "white box" *adversary* selects $n$ arbitrary test examples $\tilde{\boldsymbol{x}} \in \mathcal{X}^n$ based on all information including $\boldsymbol{x}, f, h, P$ and the learning algorithm. Using the unlabeled test examples (and the labeled training examples), the learner finally outputs $S \subseteq \mathcal{X}$. Errors are those test examples in $S$ that were misclassified, i.e., $h|_S(x) = 1 - f(x)$.

ting in adversarial learning, e.g., when one aims to classify abnormal lung scans when none are present at train-time.

[†]We are grateful to an anonymous reviewer who pointed out that gender classification is an example of when *not* to use $PQ$ learning.

Rather than jumping straight into the transductive setting, we first describe the simpler generalization setting. We define the *PQ* model in which $\tilde{\boldsymbol{x}} \sim Q^n$ are drawn i.i.d. by *nature*, for an arbitrary distribution $Q$. While it will be easier to quantify generalization error and rejections in this simpler model, the *PQ* model does not permit a white-box adversary to choose test examples based on $h$. To measure performance here, define rejection and error rates for distribution $D$, respectively:

$$\text{rej}_D(S) := \mathbf{Pr}_{x\sim D}[x \notin S] \tag{10.1}$$

$$\text{err}_D(h|_S) := \mathbf{Pr}_{x\sim D}[h(x) \neq f(x) \wedge x \in S] \tag{10.2}$$

We write $\text{rej}_D$ and $\text{err}_D$ when $h$ and $S$ are clear from context. We extend the definition of PAC learning to $P \neq Q$ as follows:

**Definition 10.1** (PQ learning). *Learner $L$ $(\varepsilon, \delta, n)$-PQ-learns $\mathcal{H}$ if for any distributions $P, Q$ over $\mathcal{X}$ and any $f \in \mathcal{H}$, its output $h|_S = L(\boldsymbol{x}, f(\boldsymbol{x}), \tilde{\boldsymbol{x}})$ satisfies*

$$\mathbf{Pr}_{\boldsymbol{x}\sim P^n, \tilde{\boldsymbol{x}}\sim Q^n}\left[\text{rej}_P + \text{err}_Q \leq \varepsilon\right] \geq 1 - \delta.$$

*$L$ PQ-learns $\mathcal{H}$ if $L$ runs in polynomial time and if there is a polynomial $p$ such that $L$ $(\varepsilon, \delta, n)$-PQ-learns $\mathcal{H}$ for every $\varepsilon, \delta > 0, n \geq p(1/\varepsilon, 1/\delta)$.*

Now, at first it may seem strange that the definition bounds $\text{rej}_P$ rather than $\text{rej}_Q$, but as mentioned $\text{rej}_Q$ cannot be bound absolutely. Instead, it can be bound relative to $\text{rej}_P$ and the *total variation distance* (also called statistical distance) $|P - Q|_{\mathsf{TV}} \in [0, 1]$, as follows:

$$\text{rej}_Q \leq \text{rej}_P + |P - Q|_{\mathsf{TV}}.$$

This new perspective, of bounding the rejection probability of $P$, as opposed to $Q$, facilitates the analysis. Of course when $P = Q$, $|P - Q|_{\mathsf{TV}} = 0$ and $\text{rej}_Q = \text{rej}_P$, and when $P$ and $Q$ have disjoint supports (no overlap), then $|P - Q|_{\mathsf{TV}} = 1$ and the above bound is vacuous. We also discuss tighter bounds relating $\text{rej}_Q$ to $\text{rej}_P$.

Algorithms and guarantees. We provide two redactive learning algorithms: a supervised algorithm called Rejectron, and an unsupervised algorithm URejectron. Rejectron takes as input $n$ labeled training data $(\boldsymbol{x}, \boldsymbol{y}) \in \mathcal{X}^n \times \{0,1\}^n$ and $n$ test data $\tilde{\boldsymbol{x}} \in \mathcal{X}^n$ (and an error parameter $\varepsilon$). It can be implemented efficiently using any $\mathsf{ERM}_{\mathcal{H}}$ oracle that outputs a function $c \in \mathcal{H}$ of minimal error on any given set of labeled examples. It is formally presented in Algorithm 10.1. At a high level, it chooses $h = \mathsf{ERM}(\boldsymbol{x}, \boldsymbol{y})$ and chooses $S$ in an iterative manner. It starts with $S = \mathcal{X}$ and then iteratively chooses $c \in \mathcal{H}$ that disagrees significantly with $h|_S$ on $\tilde{\boldsymbol{x}}$ but agrees with $h|_S$ on $\boldsymbol{x}$; it then rejects all $x$'s such that $c(x) \neq h(x)$. As we show in Lemma 10.4, choosing $c$ can be done efficiently given oracle access to $\mathsf{ERM}_{\mathcal{H}}$.

Theorem 10.5 shows that Rejectron PQ-learns any class $\mathcal{H}$ of bounded VC dimension $d$, specifically with $\varepsilon = \tilde{O}(\sqrt{d/n})$. (The $\tilde{O}$ notation hides logarithmic factors including the dependence on the failure probability $\delta$.) This is worse than the standard $\varepsilon = \tilde{O}(d/n)$ bound of supervised learning when $P = Q$, though Theorem 10.7 shows this is necessary with an $\Omega(\sqrt{d/n})$ lower-bound for $P \neq Q$.

Our unsupervised learning algorithm URejectron, formally presented in Algorithm 10.2, computes $S$ only from unlabeled training and test examples, and has similar guarantees (Theorem 10.9). The algorithm tries to distinguish training and test examples and then rejects whatever is almost surely a test example. More specifically, as above, it chooses $S$ in an iterative manner, starting with $S = \mathcal{X}$. It (iteratively) chooses *two* functions $c, c' \in \mathcal{H}$ such that $c|_S$ and $c'|_S$ have high disagreement on $\tilde{\boldsymbol{x}}$ and low disagreement on $\boldsymbol{x}$, and rejects all $x$'s on which $c|_S, c'|_S$ disagree. As we show in Lemma G.6, choosing $c$ and $c'$ can be done efficiently given a (stronger) $\mathsf{ERM}_{\mathsf{DIS}}$ oracle for the class DIS of disagreements between $c, c' \in \mathcal{H}$. We emphasize that URejectron can also be used for multi-class learning as it does not use training labels, and can be paired with any classifier trained separately. This advantage of URejectron over Rejectron comes at the cost of requiring a stronger base classifier to be used for ERM, and may lead to examples being unnecessarily rejected.

In Figure 10.1 we illustrate our algorithms for the class $\mathcal{H}$ of halfspaces. A natural idea would be to train a halfspace to distinguish unlabeled train and test examples—intuitively, one can safely reject anything that is clearly distinguishable as test without increasing $\mathrm{rej}_P$. However, this on its own is insufficient. For example, see group (2) in Figure 10.1, which cannot be distinguished from training data by a halfspace. This is precisely why having test examples is absolutely necessary. It allows us to use an $\mathsf{ERM}_{\mathcal{H}}$ oracle to PQ-learn $\mathcal{H}$.

Transductive analysis. A similar analysis of Rejectron in a transductive setting gives error and rejection bounds directly on the test examples. The bounds here are with respect to a stronger white-box adversary who need not even choose a test set $\tilde{\boldsymbol{x}}$ i.i.d. from a distribution. Such an adversary chooses the test set with knowledge of $P, f, h$ and $\boldsymbol{x}$. In particular, first $h$ is chosen based on $\boldsymbol{x}$ and $\boldsymbol{y}$; then the adversary chooses the test set $\tilde{\boldsymbol{x}}$ based on all available information; and finally, $S$ is chosen. We introduce a novel notion of *false rejection*, where we reject a test example that was in fact chosen from $P$ and not modified by an adversary. Theorem 10.6 gives bounds that are similar in spirit to Theorem 10.5 but for the harsher transductive setting.

Agnostic bounds. Thus far, we have considered the realizable setting where the target $f \in \mathcal{H}$. In agnostic learning (Kearns et al., 1992), there is an arbitrary distribution $\mu$ over $\mathcal{X} \times \{0, 1\}$ and the goal is to learn a classifier that is nearly as accurate as the best classifier in $\mathcal{H}$. In our setting, we assume that there is a known $\eta \geq 0$ such that the train and test distributions $\mu$ and $\tilde{\mu}$ over $\mathcal{X} \times \{0, 1\}$ satisfy that there is some function $f \in \mathcal{H}$ that has error at most $\eta$ with respect to both $\mu$ and $\tilde{\mu}$. Unfortunately, we show that in such a setting one cannot guarantee less than $\Omega(\sqrt{\eta})$ errors and rejections, but we show that Rejectron nearly achieves such guarantees.

Experiments. As a proof of concept, we perform simple controlled experiments on the task of handwritten letter classification using lower-case English letters from the EMNIST dataset (Cohen et al., 2017), and the task of handwritten digit classification from the MNIST dataset (LeCun and Cortes, 2010). In one setup, to mimic a spamming adversary, after a classifier $h$ is trained, test examples are identified on which $h$ errs and are repeated many times in the test set. Existing SC algorithms (no matter how robust) will fail on such an example since they all choose $S$ without using unlabeled test examples—as long as an adversary can find even a single erroneous example, it can simply repeat it. In a second setup, we consider a natural test distribution which consists of a mix of lower- and upper-case letters, while the training set was only lower-case letters. In a third setup, we consider classifying digits as even vs. odd, where the training distribution is only supported on digits in $\{0, \ldots, 5\}$ and a natural test distribution supported on all 10 digits: $\{0, \ldots, 9\}$. The simplest version of URejectron achieves high accuracy while rejecting mostly adversarial or capital letters (EMNIST), and digits in

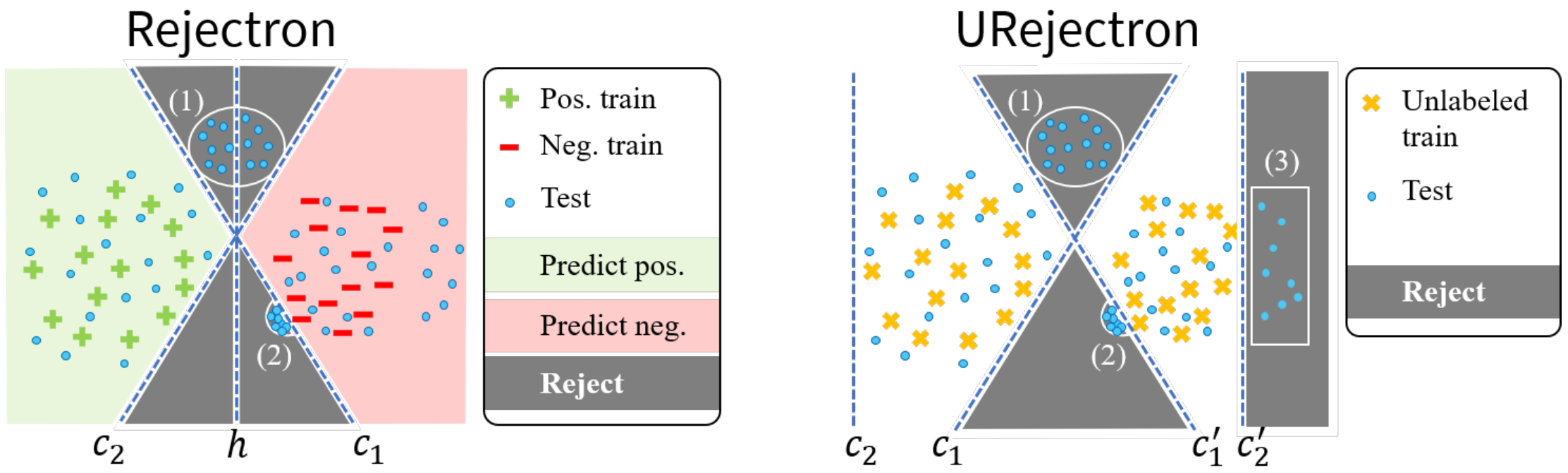

**Figure 10.1:** Our algorithm (and unsupervised variant) for learning $\mathcal{H}$=halfspaces. Rejectron (left) first trains $h$ on labeled training data, then finds other candidate classifiers $c_1, c_2$, such that $h$ and $c_i$ have high disagreement on $\tilde{\boldsymbol{x}}$ and low disagreement on $\boldsymbol{x}$, and rejects examples where $h$ and $c_i$ disagree. URejectron (right) aims to distinguish *unlabeled train* and test examples using pairs of classifiers $c_i, c'_i$ that agree on training data but disagree on many tests. Both reject: (1) clearly unpredictable examples which are very far from train and (2) a suspiciously dense cluster of tests which might all be positive despite being close to negatives. URejectron also rejects (3).

$\{6, \ldots, 9\}$ (MNIST).

## 10.2 Preliminaries and Notation

Henceforth, we assume a fixed class $\mathcal{H}$ of $c : \mathcal{X} \to \mathcal{Y}$ from domain $\mathcal{X}$ to $\mathcal{Y} = \{0, 1\}$,[‡] and let $d$ be the VC dimension of $\mathcal{H}$. Let $\log(x) = \log_2(x)$ denote the base-2 logarithm and $\ln(x)$ the natural logarithm. The set of functions from $\mathcal{X}$ to $\mathcal{Y}$ is denoted by $\mathcal{Y}^{\mathcal{X}}$. Let the set of subsets of $\mathcal{X}$ be denoted by $2^{\mathcal{X}}$. Finally, $[n]$ denotes $\{1, 2, \ldots, n\}$ for any natural number $n \in \mathbb{N}$.

## 10.3 Learning with Redaction

We now describe the two settings for SC. We use the same algorithm in both settings, so it can be viewed as two justifications for the same algorithm. The PQ model provides guarantees with respect to future examples from the test distribution, while the transductive model provides guarantees with respect to arbitrary test examples chosen by an all-powerful adversary. Interestingly, the transductive analysis is somewhat simpler and is used in the PQ analysis.

[‡]For simplicity, the theoretical model is defined for binary classification, though our experiments illustrate a multi-class application. To avoid measure-theoretic issues, we assume $\mathcal{X}$ is countably infinite or finite.

### 10.3.1 PQ Learning

In the *PQ* setting, an SC learner $h|_S = L(\boldsymbol{x}, f(\boldsymbol{x}), \tilde{\boldsymbol{x}})$ is given $n$ labeled examples $\boldsymbol{x} = (x_1, \ldots, x_n)$ drawn i.i.d. $\boldsymbol{x} \sim P^n$, labels $f(\boldsymbol{x}) = (f(x_1), \ldots, f(x_n))$ for some unknown $f \in \mathcal{H}$, and $n$ unlabeled examples $\tilde{\boldsymbol{x}} \sim Q^n$. $L$ outputs $h : \mathcal{X} \to \mathcal{Y}$ and $S \subseteq \mathcal{X}$. The adversary (or nature) chooses $Q$ based only on $f$, $P$ and knowledge of the learning algorithm $L$. The definition of PQ learning is given in Definition 10.1. Performance is measured in terms of $\text{err}_Q$ on future examples from $Q$ and $\text{rej}_P$ (rather than the more obvious $\text{rej}_Q$). Rejection rates on $P$ (and $Q$) can be estimated from held out data, if so desired. The quantities $\text{rej}_P$, $\text{rej}_Q$ can be related and a small $\text{rej}_P$ implies few rejections on future examples from $Q$ wherever it "overlaps" with $P$ by which we mean $Q(x) \le \Lambda \cdot P(x)$ for some constant $\Lambda$.

**Lemma 10.2.** *For any $S \subseteq \mathcal{X}$ and distributions $P, Q$ over $\mathcal{X}$:*

$$\text{rej}_Q(S) \le \text{rej}_P(S) + |P - Q|_{\mathsf{TV}}. \tag{10.3}$$

*Further, for any* $\Lambda \ge 0$,

$$\mathbf{Pr}_{x\sim Q}\big[x \notin S \textit{ and } Q(x) \le \Lambda P(x)\big] \le \Lambda\, \text{rej}_P(S). \tag{10.4}$$

*Proof.* For eq. (10.3), note that one can sample a point from $\tilde{x} \sim Q$ by first sampling $x \sim P$ and then changing it with probability $|P - Q|_{\mathsf{TV}}$. This follows from the definition of total variation distance. Thus, the probability that $\tilde{x}$ is rejected is at most the probability $x$ is rejected plus the probability $x \ne \tilde{x}$, establishing eq. (10.3). To see eq. (10.4), note

$$\mathbf{Pr}_{x\sim Q}\big[x \notin S \text{ and } Q(x) \le \Lambda P(x)\big] = \sum_{x\in\bar{S}: Q(x)\le \Lambda P(x)} Q(x) \le \sum_{x\in\bar{S}: Q(x)\le\Lambda P(x)} \Lambda P(x).$$

Clearly the above is at most $\sum_{x\in\bar{S}} \Lambda P(x) = \Lambda\, \text{rej}_P$. □

If $\text{rej}_P = 0$ then all $x \sim Q$ that lie in $P$'s support would necessarily be classified (i.e., $x \in S$). Note that the bound eq. (10.3) can be quite loose and a tight bound is given in Section G.7.

It is also worth mentioning that a PQ-learner can also be used to guarantee $\text{err}_P + \text{rej}_P \le \varepsilon$ meaning that it has *accuracy* $\mathbf{Pr}_P[h|_S(x) = f(x)] \ge 1 - \varepsilon$ with respect to $P$ (like a normal PAC learner) but is also simultaneously robust to $Q$. The following claim shows this and an additional property that PQ learners can be made robust with respect to any polynomial number of different $Q$'s.

**Claim 10.3.** *Let $f \in \mathcal{H}, \varepsilon, \delta > 0, n, k \geq 1$ and $P, Q_1, \ldots, Q_k$ be distributions over $\mathcal{X}$. Given a $\left(\frac{\varepsilon}{k+1}, \delta, n\right)$-PQ-learner $L$, $\boldsymbol{x} \sim P^n, f(\boldsymbol{x})$, and additional unlabeled samples $\boldsymbol{z} \sim P^n, \tilde{\boldsymbol{x}}_1 \sim Q_1^n, \ldots, \tilde{\boldsymbol{x}}_k \sim Q_k^n$, one can generate $\tilde{\boldsymbol{x}} \in \mathcal{X}^n$ such that $h|_S = L(\boldsymbol{x}, f(\boldsymbol{x}), \tilde{\boldsymbol{x}})$ satisfies,*

$$\mathbf{Pr}\left[\text{rej}_P + \text{err}_P + \sum_i \text{err}_{Q_i} \leq \varepsilon\right] \geq 1 - \delta.$$

*Proof of Claim 10.3.* Let $Q = \frac{1}{k+1}(P + Q_1 + \cdots + Q_k)$ be the blended distribution. Give $n$ samples from $P$ and each $Q_i$, one can straightforwardly construct $n$ i.i.d. samples $\tilde{\boldsymbol{x}} \sim Q$. Running $L(\boldsymbol{x}, f(\boldsymbol{x}), \tilde{\boldsymbol{x}})$ gives the guarantee that with prob. $\geq 1 - \delta$, $(k+1)(\text{rej}_P + \text{err}_Q) \leq \varepsilon$ which implies the claim since $(k+1)\,\text{err}_Q = \text{err}_P + \sum \text{err}_{Q_i}$. □

### 10.3.2 Transductive Setting with White-Box Adversary

In the *transductive* setting, there is no $Q$ and instead empirical analogs $\text{err}_{\boldsymbol{x}}$ and $\text{rej}_{\boldsymbol{x}}$ of error and rejection rates are defined as follows, for arbitrary $\boldsymbol{x} \in \mathcal{X}^n$:

$$\text{err}_{\boldsymbol{x}}(h|_S, f) := \frac{1}{n}\left|\{i \in [n] : f(x_i) \neq h(x_i) \text{ and } x_i \in S\}\right| \tag{10.5}$$

$$\text{rej}_{\boldsymbol{x}}(S) := \frac{1}{n}\left|\{i \in [n] : x_i \notin S\}\right| \tag{10.6}$$

Again, $h, f$ and $S$ may be omitted when clear from context.

In this setting, the learner first chooses $h$ using only $\boldsymbol{x} \sim P^n$ and $f(\boldsymbol{x})$. Then, a *true* test set $\boldsymbol{z} \sim P^n$ is drawn. Based on all available information ($\boldsymbol{x}, \boldsymbol{z}, f, h$, and the code for learner $L$) the adversary modifies any number of examples from $\boldsymbol{z}$ to create *arbitrary* test set $\tilde{\boldsymbol{x}} \in \mathcal{X}^n$. Finally, the learner chooses $S$ based on $\boldsymbol{x}, f(\boldsymbol{x})$, and $\tilde{\boldsymbol{x}}$. Performance is measured in terms of $\text{err}_{\tilde{\boldsymbol{x}}} + \text{rej}_{\boldsymbol{z}}$ rather than $\text{err}_Q + \text{rej}_P$, because $\boldsymbol{z} \sim P^n$. One can bound $\text{rej}_{\tilde{\boldsymbol{x}}}$ in terms of $\text{rej}_{\boldsymbol{z}}$ for any $\boldsymbol{z}, \tilde{\boldsymbol{x}} \in \mathcal{X}^n$ and $S \subseteq \mathcal{X}$, as follows:

$$\text{rej}_{\tilde{\boldsymbol{x}}} \leq \text{rej}_{\boldsymbol{z}} + \Delta(\boldsymbol{z}, \tilde{\boldsymbol{x}}), \quad \text{where} \quad \Delta(\boldsymbol{z}, \tilde{\boldsymbol{x}}) := \frac{1}{n}\left|\{i \in [n] : z_i \neq \tilde{x}_i\}\right|. \tag{10.7}$$

The hamming distance $\Delta(\boldsymbol{z}, \tilde{\boldsymbol{x}})$ is the transductive analog of $|P-Q|_{\mathsf{TV}}$. The following bounds the "false rejections," those unmodified examples that are rejected:

$$\frac{1}{n}\left|\{i \in [n] : \tilde{x}_i \notin S \text{ and } \tilde{x}_i = z_i\}\right| \leq \text{rej}_{\boldsymbol{z}}(S). \tag{10.8}$$

Both eqs. (10.7) and (10.8) follow by definition of $\text{rej}_{(\cdot)}$.

White-box adversaries The all-powerful transductive adversary is sometimes called "white box" in the sense that it can choose its examples while looking "inside" $h$ rather than using $h$ as a black box. While it cannot choose $\tilde{\boldsymbol{x}}$ with knowledge of $S$, it can know what $S$ will be as a function of $\tilde{\boldsymbol{x}}$ if the learner is deterministic, as our algorithms are. Also, we note that the generalization analysis may be extended to a white-box model where the adversary chooses $Q$ knowing $h$, but it is cumbersome even to denote probabilities over $\tilde{\boldsymbol{x}} \sim Q^n$ when $Q$ itself can depend on $\boldsymbol{x} \sim P^n$.

## 10.4 Algorithms and Guarantees

We assume that we have a deterministic oracle $\mathsf{ERM} = \mathsf{ERM}_{\mathcal{H}}$ which, given a set of labeled examples from $\mathcal{X} \times \mathcal{Y}$, outputs a classifier $c \in \mathcal{H}$ of minimal error. Algorithm 10.1 describes our algorithm Rejectron. It takes as input a set of labeled training examples $(\boldsymbol{x}, \boldsymbol{y})$, where $\boldsymbol{x} \in \mathcal{X}^n$ and $\boldsymbol{y} \in \mathcal{Y}^n$, and a set of test examples $\tilde{\boldsymbol{x}} \in \mathcal{X}^n$ along with an error parameter $\varepsilon > 0$ that trades off errors and rejections. A value for $\varepsilon$ that theoretically balances these is in Theorems 10.5 and 10.6.

**Lemma 10.4** (Computational efficiency). *For any $\boldsymbol{x}, \tilde{\boldsymbol{x}} \in \mathcal{X}^n$, $\boldsymbol{y} \in \mathcal{Y}^n$, $\varepsilon > 0$ and $\Lambda \in \mathbb{N}$,* $\mathsf{Rejectron}(\boldsymbol{x}, \boldsymbol{y}, \tilde{\boldsymbol{x}}, \varepsilon, \Lambda)$ *outputs $S_{T+1}$ for $T \leq \lfloor 1/\varepsilon \rfloor$. Further, each iteration can be implemented using one call to* $\mathsf{ERM}$ *on at most $(\Lambda + 1)n$ examples and $O(n)$ evaluations of classifiers in $\mathcal{H}$.*

*Proof.* To maximize $s_t$ using the ERM oracle for $\mathcal{H}$, construct a dataset consisting of each training example, labeled by $h$, repeated $\Lambda$ times, and each test example in $\tilde{x}_i \in S_t$, labeled $1 - h(\tilde{x}_i)$, included just once. Running $\mathsf{ERM}$ on this artificial dataset returns a classifier of

**Algorithm 10.1:** Rejectron

**Input:** train $\boldsymbol{x} \in \mathcal{X}^n$, labels $\boldsymbol{y} \in \mathcal{Y}^n$, test $\tilde{\boldsymbol{x}} \in \mathcal{X}^n$, error $\varepsilon \in [0, 1]$, weight $\Lambda = n + 1$.

$h := \mathsf{ERM}(\boldsymbol{x}, \boldsymbol{y})$ # assume black box oracle ERM to minimize errors.

**for** $t = 1, 2, 3, \ldots$ **do**

  $S_t := \{x \in \mathcal{X} : h(x) = c_1(x) = \ldots = c_{t-1}(x)\}$ # So $S_1 = \mathcal{X}$.

  Choose $c_t \in \mathcal{H}$ to maximize $s_t(c) := \text{err}_{\tilde{\boldsymbol{x}}}(h|_{S_t}, c) - \Lambda \cdot \text{err}_{\boldsymbol{x}}(h, c)$ over $c \in \mathcal{H}$.

   # Lemma 10.4 shows how to maximize $s_t$ using ERM (err is defined in eq. (10.5)).

  If $s_t(c_t) \leq \varepsilon$, then stop and return $h|_{S_t}$.

**Output:** The selective classifier $h|_{S_t}$.

**Figure 10.2:** The Rejectron algorithm takes labeled training examples and unlabeled test examples as input, and it outputs a selective classifier $h|_S$ that predicts $h(x)$ for $x \in S$ (and rejects all $x \notin S$). Parameter $\varepsilon$ controls the trade-off between errors and rejections and can be set to $\varepsilon = \tilde{\Theta}(\sqrt{d/n})$ to balance the two. The weight $\Lambda$ parameter is set to its default value of $n + 1$ for realizable (noiseless) learning but should be lower for agnostic learning.

minimal error on it. But the number of errors of classifier $c$ on this artificial dataset is:

$$\Lambda \sum_{i \in [n]} |c(x_i) - h(x_i)| + \sum_{i : \tilde{x}_i \in S_t} |c(\tilde{x}_i) - (1 - h(\tilde{x}_i))| =$$
$$\Lambda \sum_{i \in [n]} |c(x_i) - h(x_i)| + \sum_{i : \tilde{x}_i \in S_t} 1 - |c(\tilde{x}_i) - h(\tilde{x}_i)|,$$

which is equal to $\left|\{i \in [n] : \tilde{x}_i \in S_t\}\right| - n s_t(c)$. Hence $c$ minimizes error on this artificial dataset if and only if it maximizes $s_t$ of the algorithm.

Next, let $T$ be the number of iterations of the algorithm Rejectron, so its output is $h|_{S_{T+1}}$. We must show that $T \leq \lfloor 1/\varepsilon \rfloor$. To this end, note that by definition, for every $t \in [T]$ it holds that $S_{t+1} \subseteq S_t$, and moreover,

$$\frac{1}{n}\left|\{i \in [n] : \tilde{x}_i \in S_t\}\right| - \frac{1}{n}\left|\{i \in [n] : \tilde{x}_i \in S_{t+1}\}\right| = \text{err}_{\tilde{\boldsymbol{x}}}(h|_{S_t}, c_t) \geq s_t(c_t) > \varepsilon. \quad (10.9)$$

Hence, the fraction of additional rejected test examples in each iteration $t \in [T]$ is greater than $\varepsilon$, and hence $T < 1/\varepsilon$. Since $T$ is an integer, this means that $T \leq \lfloor 1/\varepsilon \rfloor$.

For efficiency, of course each $S_t$ is not explicitly stored since even $S_1 = \mathcal{X}$ could be infinite. Instead, note that to execute the algorithm, we only need to maintain: (a) the subset of indices $Z_t = \{j \in [n] \mid \tilde{x}_j \in S_t\}$ of test examples which are in the prediction set, and (b) the classifiers $h, c_1, \ldots, c_T$. Also note that updating $Z_t$ from $Z_{t-1}$ requires evaluating $c_t$ at most $n$ times. In this fashion, membership in $S_t$ and $S = S_{T+1}$ can be computed efficiently and output in a succinct manner. $\square$

Note that since we assume ERM is deterministic, the Rejectron algorithm is also deterministic. This efficient reduction to ERM, together with the following imply that Rejectron is a PQ learner:

**Theorem 10.5** (PQ guarantees). *For any $n \in \mathbb{N}, \delta > 0, f \in \mathcal{H}$ and distributions $P, Q$ over $\mathcal{X}$:*

$$\mathbf{Pr}_{\boldsymbol{x}\sim P^n, \tilde{\boldsymbol{x}}\sim Q^n}[\text{err}_Q \leq 2\varepsilon^* \ \wedge \ \text{rej}_P \leq \varepsilon^*] \geq 1 - \delta,$$

*where* $\varepsilon^* = \sqrt{\frac{8d \ln 2n}{n}} + \frac{8 \ln 16/\delta}{n}$ *and* $h|_S = \mathsf{Rejectron}(\boldsymbol{x}, f(\boldsymbol{x}), \tilde{\boldsymbol{x}}, \varepsilon^*)$.

More generally, Theorem G.5 shows that, by varying parameter $\varepsilon$, one can achieve any trade-off between $\text{err}_Q \leq O(\varepsilon)$ and $\text{rej}_P \leq \tilde{O}(\frac{d}{n\varepsilon})$. The analogous transductive guarantee is:

**Theorem 10.6** (Transductive). *For any $n \in \mathbb{N}, \delta > 0, f \in \mathcal{H}$ and dist. $P$ over $\mathcal{X}$:*

$$\mathbf{Pr}_{\boldsymbol{x},\boldsymbol{z}\sim P^n}\left[\forall \tilde{\boldsymbol{x}} \in \mathcal{X}^n : \ \text{err}_{\tilde{\boldsymbol{x}}}(h|_S) \leq \varepsilon^* \ \wedge \ \text{rej}_{\boldsymbol{z}}(S) \leq \varepsilon^*\right] \geq 1 - \delta,$$

*where* $\varepsilon^* = \sqrt{\frac{2d}{n}\log 2n + \frac{1}{n}\log\frac{1}{\delta}}$ *and* $h|_S = \mathsf{Rejectron}(\boldsymbol{x}, f(\boldsymbol{x}), \tilde{\boldsymbol{x}}, \varepsilon^*)$.

One thinks of $\boldsymbol{z}$ as the real test examples and $\tilde{\boldsymbol{x}}$ as an arbitrary adversarial modification, not necessarily iid. Equation (10.8) means that this implies $\leq \varepsilon^*$ errors on unmodified examples. As discussed earlier, the guarantee above holds for *any* $\tilde{\boldsymbol{x}}$ chosen by a white-box adversary, which may depend on $\boldsymbol{x}$ and $f$, and thus on $h$ (since $h = \mathsf{ERM}(\boldsymbol{x}, f(\boldsymbol{x}))$ is determined by $\boldsymbol{x}$ and $f$). More generally, Theorem G.2 shows that, by varying parameter $\varepsilon$, one can trade-off $\text{err}_{\tilde{\boldsymbol{x}}} \leq \varepsilon$ and $\text{rej}_{\boldsymbol{z}} \leq \tilde{O}(\frac{d}{n\varepsilon})$.

We note that Theorems 10.5 and 10.6 generalize in a rather straightforward manner to the case in which an adversary can inject additional training examples to form $\boldsymbol{x}' \supseteq \boldsymbol{x}$ which contains $\boldsymbol{x}$. Such an augmentation reduces the version space, i.e., the set of $h \in \mathcal{H}$ consistent with $f$ on $\boldsymbol{x}'$, but of course $f$ still remains in this set. The analysis remains essentially unchanged as long as $\boldsymbol{x}'$ contains $\boldsymbol{x}$ and $\boldsymbol{x}$ consists of $n$ examples. The bounds remain the same in terms of $n$, but Rejectron should be run with $\Lambda$ larger than the number of examples in $\boldsymbol{x}'$ in this case to ensure that each $c_t$ has zero training error.

Here we give the proof sketch of Theorem 10.6, since it is slightly simpler than Theorem 10.5. Full proofs are in Section G.1.

*Proof sketch for Theorem 10.6.* To show $\text{err}_{\tilde{\boldsymbol{x}}} \leq \varepsilon^*$, fix any $f, \boldsymbol{x}, \tilde{\boldsymbol{x}}$. Since $h = \mathsf{ERM}(\boldsymbol{x}, f(\boldsymbol{x}))$ and $f \in \mathcal{H}$, this implies that $h$ has zero training error, i.e., $\text{err}_{\boldsymbol{x}}(h, f) = 0$. Hence $s_t(f) = \text{err}_{\tilde{\boldsymbol{x}}}(h|_{S_t}, f)$ and the algorithm cannot terminate with $\text{err}_{\tilde{\boldsymbol{x}}}(h|_{S_t}, f) > \varepsilon$ since it could have selected $c_t = f$.

To prove $\text{rej}_{\boldsymbol{z}} \leq \varepsilon^*$, observe that Rejectron never rejects any training $\boldsymbol{x}$. This follows from the fact that $\Lambda > n$, together with the fact that $h(x_i) = f(x_i)$ for every $i \in [n]$ which follows, in turn, from the facts that $f \in \mathcal{H}$ and $h = \mathsf{ERM}(\boldsymbol{x}, f(\boldsymbol{x}))$. Now $\boldsymbol{x}$ and $\boldsymbol{z}$ are identically distributed. By a generalization-like bound (Lemma G.1), with probability $\geq 1 - \delta$ there is no classifier for which selects all of $\boldsymbol{x}$ and yet rejects with probability greater than $\varepsilon^*$ on $\boldsymbol{z}$ for $T \leq 1/\varepsilon^*$ (by Lemma 10.4). □

Unfortunately, the above bounds are worse than standard $\tilde{O}(d/n)$ VC-bounds for $P = Q$, but the following lower-bound shows that $\tilde{O}(\sqrt{d/n})$ is tight for some class $\mathcal{H}$.

**Theorem 10.7** (PQ lower bound). *There exists a constant $K > 0$ such that: for any $d \geq 1$, there is a concept class $\mathcal{H}$ of VC dimension $d$, distributions $P$ and $Q$, such that for any $n \geq 2d$ and learner $L : \mathcal{X}^n \times \mathcal{Y}^n \times \mathcal{X}^n \to \mathcal{Y}^{\mathcal{X}} \times 2^{\mathcal{X}}$, there exists $f \in \mathcal{H}$ with*

$$\mathbb{E}_{\substack{\boldsymbol{x} \sim P^n \\ \tilde{\boldsymbol{x}} \sim Q^n}} \left[\text{rej}_P + \text{err}_Q\right] \geq K\sqrt{\frac{d}{n}}, \quad \textit{where} \quad h|_S = L(\boldsymbol{x}, f(\boldsymbol{x}), \tilde{\boldsymbol{x}}).$$

Note that since $P$ and $Q$ are fixed, independent of the learner $L$, the unlabeled test examples from $Q$ are not useful for the learner as they could simulate as many samples from $Q$ as they would like on their own. Thus, the lower bound holds even given $n$ training examples and $m$ unlabeled test examples, for arbitrarily large $m$.

Theorem 10.7 implies that the learner needs at least $n = \Omega(d/\varepsilon^2)$ labeled training examples to get the $\varepsilon$ error plus rejection guarantee. However, it leaves open the possibility that many fewer than $m = \tilde{O}(d/\varepsilon^2)$ test examples are needed. We give a lower bound in the transductive case which shows that both $m, n$ must be at least $\Omega(d/\varepsilon^2)$:

**Algorithm 10.2:** URejectron

**Input:** train $\boldsymbol{x} \in \mathcal{X}^n$, test $\tilde{\boldsymbol{x}} \in \mathcal{X}^n$, error $\varepsilon \in [0,1]$, weight $\Lambda = n+1$.

```
for t = 1, 2, 3, ... do
    S_t := {x ∈ X : c_1(x) = c'_1(x) ∧ ··· ∧ c_{t-1}(x) = c'_{t-1}(x)}   # So S_1 = X.
    Choose c_t, c'_t ∈ H to maximize s_t(c, c') := err_x̃(c'|_{S_t}, c) − Λ · err_x(c', c).
        # Lemma G.6 shows how to maximize s_t using ERM_DIS (DIS is defined in
      eq. (10.10)).
    If s_t(c_t, c'_t) ≤ ε, then stop and return S_t.
```

**Output:** The set $S_t$.

**Figure 10.3:** The URejectron unsupervised algorithm takes unlabeled training examples and unlabeled test examples as input, and it outputs a set $S \subseteq \mathcal{X}$ where classification should take place.

**Theorem 10.8** (Transductive lower bound). *There exists a constant $K > 0$ such that: for any $d \geq 1$ there exists a concept class of VC dimension $d$ where, for any $m, n \geq 4d$ there exists a distribution $P$, and an adversary $\mathcal{A} : \mathcal{X}^{n+m} \to \mathcal{X}^m$, such that for any learner $L : \mathcal{X}^n \times \mathcal{Y}^n \times \mathcal{X}^m \to \mathcal{Y}^{\mathcal{X}} \times 2^{\mathcal{X}}$ there is a function $f \in \mathcal{H}$ such that:*

$$\mathbb{E}_{\substack{\boldsymbol{x}\sim P^n \\ \boldsymbol{z}\sim P^m}}[\mathrm{rej}_{\boldsymbol{z}} + \mathrm{err}_{\tilde{\boldsymbol{x}}}] \geq K\sqrt{\frac{d}{\min\{m,n\}}}$$

*where $\tilde{\boldsymbol{x}} = \mathcal{A}(\boldsymbol{x}, \boldsymbol{z})$ and $h|_S = L(\boldsymbol{x}, f(\boldsymbol{x}), \tilde{\boldsymbol{x}})$.*

This $\Omega(\sqrt{d/\min\{m,n\}})$ lower bound implies that one needs both $\Omega(d/\varepsilon^2)$ training and test examples to guarantee $\varepsilon$ error plus rejections. This is partly why, for simplicity, aside from the Theorem 10.8, our analysis takes $m = n$. The proofs of these two lower bounds are in Section G.6.

Unsupervised selection algorithm. Our unsupervised selection algorithm URejectron is described in Algorithm 10.2. It takes as input only train and test examples $\boldsymbol{x}, \tilde{\boldsymbol{x}} \in \mathcal{X}^n$ along with an error parameter $\varepsilon$ recommended to be $\tilde{\Theta}(\sqrt{d/n})$, and it outputs a set $S$ of the selected elements. URejectron requires a more powerful black-box ERM—we show that URejectron can be implemented efficiently if one can perform ERM with respect to the family of binary classifiers that are disagreements (xors) between two classifiers. For classifiers $c, c' : \mathcal{X} \to \mathcal{Y}$, define $\mathsf{dis}_{c,c'} : \mathcal{X} \to \{0,1\}$ and $\mathsf{DIS}$ as follows:

$$\mathsf{dis}_{c,c'}(x) := \begin{cases} 1 & \text{if } c(x) \neq c'(x) \\ 0 & \text{otherwise} \end{cases} \quad \text{and} \quad \mathsf{DIS} := \{\mathsf{dis}_{c,c'} : c, c' \in \mathcal{H}\}. \tag{10.10}$$

Lemma G.6 shows how URejectron is implemented efficiently with an $\mathsf{ERM}_{\mathsf{DIS}}$ oracle.

Also, we show nearly identical guarantees to those of Theorem 10.6 for URejectron:

**Theorem 10.9** (Unsupervised). *For any $n \in \mathbb{N}$, any $\delta \geq 0$, and any distribution $P$ over $\mathcal{X}$:*

$$\mathbf{Pr}_{\boldsymbol{x},\boldsymbol{z} \sim P^n} \left[ \forall f \in \mathcal{H}, \tilde{\boldsymbol{x}} \in \mathcal{X}^n : \ (\mathrm{err}_{\tilde{\boldsymbol{x}}}(h|_S) \leq \varepsilon^*) \wedge \left(\mathrm{rej}_{\boldsymbol{z}}(S) \leq \varepsilon^*\right) \right] \geq 1 - \delta,$$

*where* $\varepsilon^* = \sqrt{\frac{2d}{n} \log 2n} + \frac{1}{n} \log \frac{1}{\delta}$, $S = \mathsf{URejectron}(\boldsymbol{x}, \tilde{\boldsymbol{x}}, \varepsilon^*)$ *and* $h = \mathsf{ERM}_{\mathcal{H}}(\boldsymbol{x}, f(\boldsymbol{x}))$.

The proof is given in Section G.2 and follows from Theorem G.7 which shows that by varying parameter $\varepsilon$, one can achieve any trade-off $\mathrm{err}_{\tilde{\boldsymbol{x}}} \leq \varepsilon$ and $\mathrm{rej}_{\boldsymbol{z}} \leq \tilde{O}(\frac{d}{n\varepsilon})$. Since one runs URejectron without labels, it has guarantees with respect to any empirical risk minimizer $h$ which may be chosen separately, and its output is also suitable for a multi-class problem.

Massart noise. We also consider two non-realizable models. First, we consider the Massart noise model, where there is an arbitrary (possibly adversarial) noise rate $\eta(x) \leq \eta$ chosen for each example. We show that Rejectron is a PQ learner in the Massart noise model with $\eta < 1/2$, assuming an ERM oracle and an additional $N = \tilde{O}\left(\frac{dn^2}{\delta^2(1-2\eta)^2}\right)$ examples from $P$. See Section G.3 for details.

A semi-agnostic setting. We also consider the following semi-agnostic model. For an arbitrary distribution $D$ over $\mathcal{X} \times \mathcal{Y}$, again with $\mathcal{Y} = \{0,1\}$, the analogous notions of rejection and error are:

$$\mathrm{rej}_D(S) := \mathbf{Pr}_{(x,y)\sim D}[x \notin S] \quad \text{and} \quad \mathrm{err}_D(h|_S) := \mathbf{Pr}_{(x,y)\sim D}[h(x) \neq y \wedge x \in S]$$

In standard agnostic learning with respect to $D$, we suppose there is some classifier $f \in \mathcal{H}$ with error $\mathrm{err}_D(f) \leq \eta$ and we aim to find a classifier whose generalization error is not much greater than $\eta$. In that setting, one can of course choose $\eta_{\text{opt}} := \min_{f\in\mathcal{H}} \text{err}_D(f)$. For well-fitting models, where there is some classifier with very low error, $\eta$ may be small.

To prove any guarantees in our setting, the test distribution must somehow be related to the training distribution. To tie together the respective training and test distributions $\mu, \tilde{\mu}$ over $\mathcal{X} \times \mathcal{Y}$, we suppose we know $\eta$ such that both $\text{err}_\mu(f) \leq \eta$ and $\text{err}_{\tilde{\mu}}(f) \leq \eta$ for some $f \in \mathcal{H}$. Even with these conditions, Lemma G.13 shows that one cannot simultaneously guarantee error rate on $\tilde{\mu}$ and rejection rate on $\mu$ less than $\sqrt{\eta/8}$, and Theorem G.14 shows that our Rejectron algorithm achieves a similar upper bound. This suggests that PQ-learning (i.e., adversarial SC) may be especially challenging in settings where ML is not able to achieve low error $\eta$.

## 10.5 Experiments

Rather than classifying sensitive attributes such as explicit images, we perform simple experiments on handwritten letter classification from the popular EMNIST dataset (Cohen et al., 2017) and on handwritten digit classification from the popular MNIST dataset (LeCun and Cortes, 2010).

### 10.5.1 EMNIST Experiments

The training data consisted of the eight lowercase letters *a d e h l n r t*, chosen because they each had more than 10,000 instances. From each letter, 3,000 instances of each letter were reserved for use later, leaving 7,000 examples, each constituting 56,000 samples from $P$.

We then considered two test distributions, $Q_{\text{adv}}, Q_{\text{nat}}$ representing adversarial and natural settings. $Q_{\text{adv}}$ consisted of a mix of 50% samples from $P$ (the 3,000 reserved instances per lower-case letter mentioned above) and 50% samples from an adversary that used a classifier $h$ as a black box. To that, we added 3,000 adversarial examples for each letter selected as follows: the reserved 3,000 letters were labeled by $h$ and the adversary selected the first misclassified instance for each letter. Misclassified examples are shown in Figure 10.5. It made 3,000 imperceptible modifications of each of the above instances by changing the intensity value of a single pixel by at most 4 (out of 256). The result was 6,000 samples per letter constituting 48,000 samples from $Q_{\text{adv}}$.

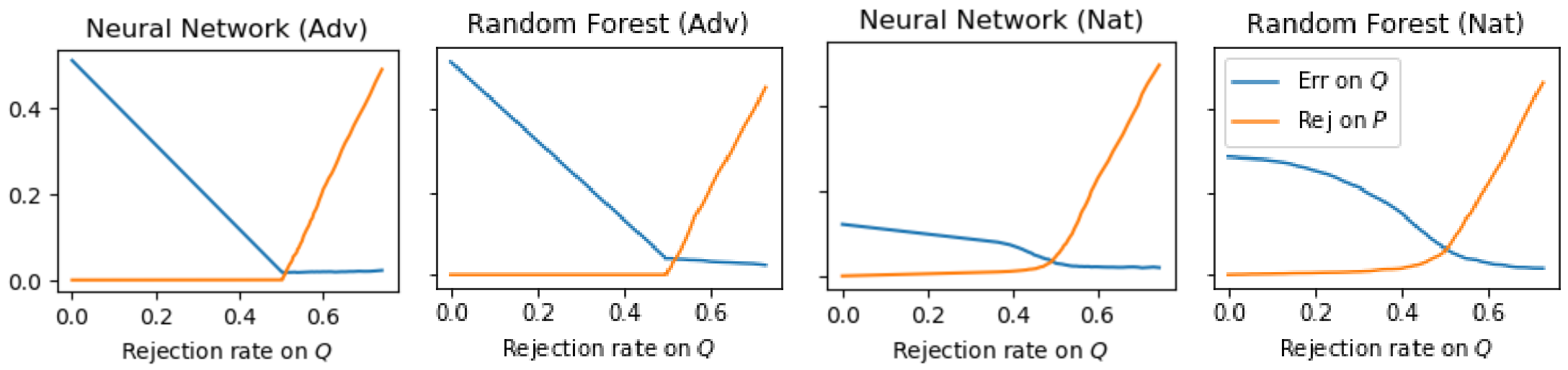

**Figure 10.4:** EMNIST Experiments: Trade-offs between rejection rate on $P$ and error rate on $Q$. The error on $Q$ (in blue) is the fraction of errors *among selected examples* (unlike $\mathrm{err}_Q$ in our analysis).

For $Q_{\text{nat}}$, the test set also consisted of 6,000 samples per letter, with 3,000 reserved samples from $P$ as above. In this case, the remaining half of the letters were simply upper-case[§] versions of the letters $A\ D\ E\ H\ L\ N\ R\ T$, taken from the EMNIST dataset (case information is also available in that dataset). Again the dataset size is 48,000. We denote this test distribution by $Q_{\text{nat}}$.

In Figure 10.4, we plot the trade-off between the rejection rate on $P$ and the error rate on $Q$ of the URejectron algorithm. Since this is a multi-class problem, we implement the most basic form of the URejectron algorithm, with $T = 1$ iterations. Instead of fixing parameter $\Lambda$, we simply train a predictor $h^{\text{Dis}}$ to distinguish between examples from $P$ and $Q$, and train a classifier $h$ on $P$. We trained two models, a random forest (with default parameters from scikit-learn Pedregosa et al., 2011) and a neural network. Complete details are provided at the end of this section. We threshold the prediction scores of distinguisher $h^{\text{Dis}}$ at various values. For each threshold $\tau$, we compute the fraction of examples from $P$ that are rejected (those with prediction score less than $\tau$), and similarly for $Q$, and the error rate of classifier $h$ on examples from $Q$ that are *not* rejected (those with prediction score at least $\tau$). We see in Figure 10.4 that for a suitable threshold $\tau$ our URejectron algorithm achieves both low rejection rate on $P$ and low error rate on $Q$. Thus on these problems the simple algorithm suffices.

We compare to the state-of-the-art SC algorithm SelectiveNet (Geifman and El-Yaniv, 2019). We ran it to train a selective neural network classifier on $P$. SelectiveNet performs exceptionally on $Q_{\text{nat}}$, achieving low error and reject almost exclusively upper-case letters. It of course errs on all adversarial examples from $Q_{\text{adv}}$, as will all existing SC algorithms (no matter how robust) since they all choose $S$ without using unlabeled test examples.

[§]In some datasets, letter classes consist of a mix of upper- and lower-case, while in others they are assigned different classes (EMNIST has both types of classes). In our experiments, they belong to the same class.

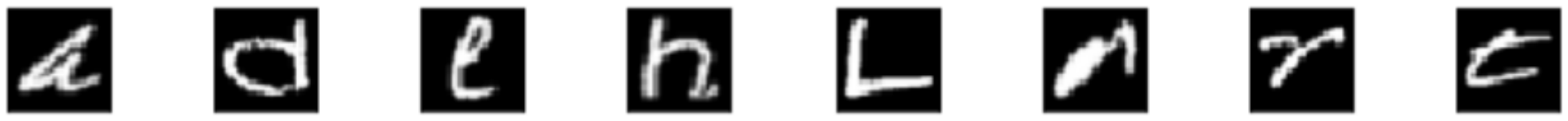

**Figure 10.5:** Adversarial choices of *a d e h l n r t*, misclassified by the Random Forest classifier.

Models  A Random Forest Classifier $h_{\text{RF}}$ from Scikit-Learn (default parameters including 100 estimators) (Pedregosa et al., 2011) and a simple neural network $h_{\text{NN}}$ consisting of two convolutional layers followed by two fully connected layers[¶] were fit on a stratified 90%/10% train/test split. The network parameters are trained with SGD with momentum (0.9), weight decay ($2 \times 10^{-4}$), batch size (128), for 85 epochs with a learning rate of 0.1, that is decayed it by a factor of 10 on epochs 57 and 72. $h_{\text{RF}}$ had a 3.6% test error rate on $P$, and $h_{\text{NN}}$ had a 1.3% test error rate on $P$.

SelectiveNet  SelectiveNet requires a target coverage hyperparameter which in our experiments is fixed to 0.7. We use an open-source PyTorch implementation of SelectiveNet that is available on GitHub [‖], with a VGG 16 architecure (Simonyan and Zisserman, 2015). To accommodate the VGG 16 architecure without changes, we pad all images with zeros (from 28x28 to 32x32), and duplicate the channels (from 1 to 3). SelectiveNet achieves rejection rates of 21.08% ($P$), 45.89% ($Q_{\text{nat}}$), and 24.04% ($Q_{\text{adv}}$), and error rates of 0.02% ($P$), 0.81% ($Q_{\text{nat}}$), and 76.78% ($Q_{\text{adv}}$).

### 10.5.2 An MNIST Experiment

Classifying digits as even vs. odd.  We use MNIST (LeCun and Cortes, 2010) as the data source, where $P$ is a distribution supported only on digits from $\{0, 1, 2, 3, 4, 5\}$, and $Q$ is a distribution supported on all 10 digits: $\{0, 1, \ldots, 9\}$. The PQ dataset for the experiments below was generated as follows: (1) Ptrain includes 20,000 digits in [0-5] from MNIST training split, (2) Qtrain includes 10,000 digits in [0-5] and 10,000 digits in [6-9] from MNIST training split, (3) Ptest includes 4,000 digits in [0-5] from MNIST test split,

[¶] https://github.com/pytorch/examples/blob/master/mnist/main.py

[‖] https://github.com/pranaymodukuru/pytorch-SelectiveNet

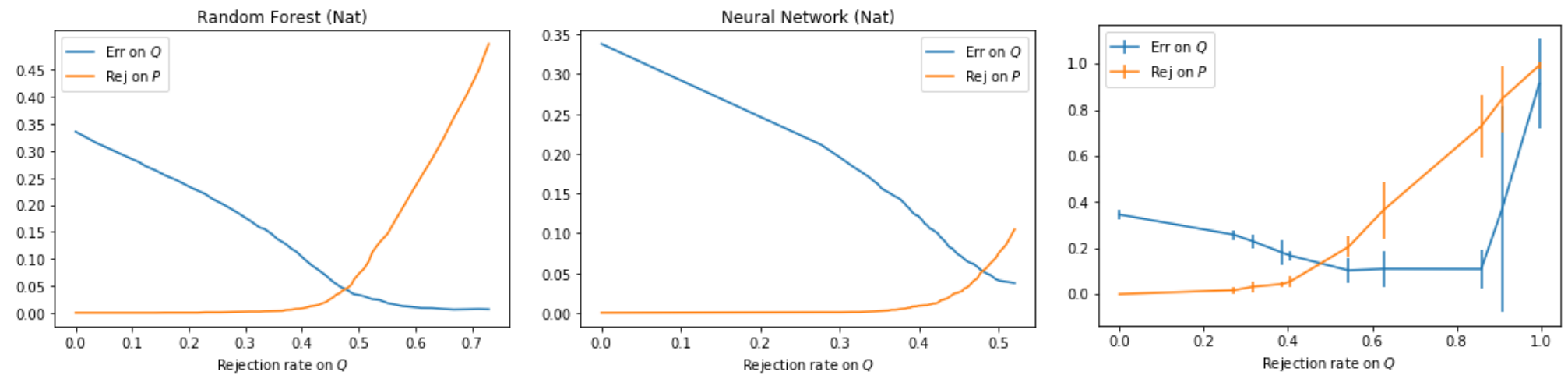

**(a)** URejectron (Random Forest) **(b)** URejectron (Neural Network) **(c)** SelectiveNet

**Figure 10.6:** MNIST Experiments: Trade-offs between rejection rate on $P$ (supported on digits $0, \ldots, 5$) and error rate on $Q$ (supported on digits $0, \ldots, 9$). The task is to classify digits as even vs. odd.

(4) Qtest includes 2,000 digits in [0-5] and 2,000 digits in [6-9] from MNIST test split. These four datasets are disjoint, i.e., they don't overlap in datapoints. We label all datapoints in these datasets as even or odd accordingly. Ptrain was used to train: a random forest classifier $h_{\mathrm{RF}}$, a neural network classifier $h_{\mathrm{NN}}$, and SelectiveNet (same configurations as described previously in Subsection 10.5.1). Additionally, Ptrain and Qtrain were both used to train random forest and neural network predictors to distinguish between $P$ and $Q$ (as part of URejectron). All evaluations were conducted on Ptest and Qtest.

In Figure 10.6, we plot the trade-off between the rejection rate on $P$ and the error rate on $Q$ of the URejectron algorithm, using a random forest $h_{\mathrm{RF}}$ or a neural network $h_{\mathrm{NN}}$. We also compare to the trade-off between the rejection rate on $P$ and the error rate on $Q$ when using SelectiveNet, with a coverage hyperparameter grid of $\{0.1, 0.2, \ldots, 1.0\}$**. We observe that URejectron exhibits a better trade-off curve between the rejection rate on $P$ and the error rate on $Q$ compared with SelectiveNet.

## 10.6 Related Work

The redaction model combines SC and transductive learning, which have each been extensively studied, separately. We first discuss prior work on these topics, which (with the notable exception of online SC) has generally been considered when test examples are from the same distribution as training examples.

**We did 6 runs for each choice of coverage hyperparameter, and plotted the mean and stadard deviation of $\mathrm{err}_Q$, $\mathrm{rej}_P$.

Selective classification. Selective classification go by various names including "classification with a reject option" and "reliable learning." To the best of our knowledge, prior work has not considered SC using unlabeled samples from $Q \neq P$. Early learning theory work by Rivest and Sloan (1988) required a guarantee of 0 test errors and few rejections. However, Kivinen (1990) showed that, for this definition, even learning rectangles under uniform distributions $P = Q$ requires exponential number of examples (as cited by Hopkins, Kane, and Lovett (2019) which like much other work therefore makes further assumptions on $P$ and $Q$). Most of this work assumes the same training and test distributions, without adversarial modification. Kanade, Kalai, and Mansour (2009) give a SC reduction to an agnostic learner (similar in spirit to our reduction to ERM) but again for the case of $P = Q$.

A notable exception is the work in *online* SC, where an *arbitrary sequence* of examples is presented one-by-one with immediate error feedback. This work includes the "knows-what-it-knows" algorithm (Li, Littman, Walsh, and Strehl, 2011), and (Sayedi, Zadimoghaddam, and Blum, 2010) exhibit an interesting trade-off between the number of mistakes and the number of rejections in such settings. However, basic classes such as intervals on the line are impossible to learn in these harsh online formulations. Interestingly, our division into labeled train and unlabeled test seems to make the problem easier than in the harsh online model.

Transductive (and semi-supervised) learning. In transductive learning, the classifier is given test examples to classify all at once or in batches, rather than individually (e.g., Vapnik, 1998). Performance is measured with respect to the test examples. It is related to *semi-supervised learning*, where unlabeled examples are given but performance is measured with respect to future examples from the same distribution. Here, since the assumption is that training and test examples are iid, it is generally the case that the unlabeled examples greatly outnumber the training examples, since otherwise they would provide limited additional value.

We now discuss related work which considers $Q \neq P$, but where classifiers must predict everywhere without the possibility of outputting ?.

Robustness to Adversarial Examples. As we have elaborated extensively in previous chapters, there is ongoing effort to devise methods for learning predictors that are robust to adversarial examples (Szegedy et al., 2013, Biggio et al., 2013, Goodfellow et al., 2015b) at test time. This line of work typically assumes that the adversarial examples are perturbations of honest examples chosen from $P$. The main objective is to learn a classifier that has high robust accuracy, meaning that with high probability, the classifier will answer correctly even if the test point was an adversarially perturbed example. Empirical work has mainly focused on training deep learning based classifiers to be more robust (e.g., Madry et al., 2018, Wong and Kolter, 2018, Zhang et al., 2019a). Kang et al. (2019b) consider the fact that perturbations may not be known in advance, and some work (e.g., Pang et al., 2018) addresses the problem of identifying adversarial examples. We emphasize that as opposed to this line of work, in this chapter we consider *arbitrary* test examples and use SC.

Detecting adversarial examples has been studied in practice, but Carlini and Wagner (2017) study ten proposed heuristics and are able to bypass all of them. Our algorithms in this chapter also require a sufficiently large set of unlabeled test examples. The use of unlabeled data for improving robustness has also been empirically explored recently (e.g., Carmon et al., 2019b, Stanforth et al., 2019, Zhai et al., 2019).

In work on real-world adversarial images, Yuan et al. (2019) find adversaries using highly visible transformations rather than imperceptible perturbations. They categorize seven major types of such transformations and write:

> "Compared with the adversarial examples studied by the ongoing adversarial learning, such adversarial explicit content does not need to be optimized in a sense that the perturbation introduced to an image remains less perceivable to humans.... today's cybercriminals likely still rely on a set of predetermined obfuscation techniques... not gradient descent."

Covariate Shift. The literature on learning with covariate shift is too large to survey here, see, e.g., the book by Quionero-Candela et al. (2009) and the references therein. To achieve guarantees, it is often assumed that the support of $Q$ is contained in the support of $P$. Like our work, many of these approaches use unlabeled data from $Q$ (e.g., Huang et al., 2007, Ben-David and Urner, 2012). Ben-David and Urner (2012) show that learning with covariate-shift is intractable, in the worst case, without such assumptions. In this chapter we overcome this negative result, and obtain guarantees for arbitrary $Q$, using SC. In summary, prior work on covariate shift that guarantees low test/target error requires strong assumptions regarding the distributions. This motivates our model of covariate shift with rejections.

Follow-up work. Recently, Kalai and Kanade (2021) presented a different PQ learning algorithm called *Slice-and-Dice*, which does not use an ERM oracle but instead uses a weaker oracle to a "reliable" learner (a model of learning with one-sided noise Kalai et al., 2012). Furthermore, they showed a computational equivalence between the two models: PQ learning and reliable learning.

## 10.7 Conclusion

The fundamental theorem of statistical learning states that an ERM algorithm for class $\mathcal{H}$ is asymptotically nearly optimal requiring $\tilde{\Theta}(d/n)$ labeled examples for learning arbitrary distributions when $P = Q$ (see, e.g., Shalev-Shwartz and Ben-David, 2014). This chapter can be viewed as a generalization of this theorem to the case where $P \neq Q$, obtaining $\tilde{\Theta}(\sqrt{d/n})$ rates. When $P = Q$, unlabeled samples from $Q$ are readily available by ignoring labels of some training data, but unlabeled test samples are necessary when $P \neq Q$. No prior such guarantee was known for arbitrary $P \neq Q$, even for simple classes such as intervals, perhaps because it may have seemed impossible to guarantee anything meaningful in the general case.

The practical implications are that, to address learning in the face of adversaries beyond perturbations (or drastic covariate shift), unlabeled examples and abstaining from classifying may be necessary. In this model, the learner can beat an unbounded white-box adversary. Even the simple approach of training a classifier to distinguish unlabeled train vs. test examples may be adequate in some applications, though for theoretical guarantees one requires somewhat more sophisticated algorithms.

# 11
# Open Problems

Besides the open directions dispersed throughout the thesis, we outline below additional open problems that might be of interest to the reader.

**Question 1.** *Are there universal oracle-efficient and sample-efficient learning algorithms that can robustly learn a class $\mathcal{H}$ with respect to a perturbation set $\mathcal{U}$, when only given access to a* $\mathsf{RERM}_{\mathcal{H}}$ *(as defined in Equation 3.2)? Is it possible to achieve this with sample and oracle complexity that are polynomial in* $\text{vc}(\mathcal{H})$*?*

Positive answers to this question will have interesting theoretical and practical implications. From a practical perspective, we can think of adversarial training methods (e.g., Madry et al., 2018) as heuristics that implement a $\mathsf{RERM}_{\mathcal{H}}$ oracle, and thus designing simple methods with provable robustness guarantees that operate only by interacting with a $\mathsf{RERM}_{\mathcal{H}}$ oracle might lead to further principled advances in state-of-the-art adversarial robustness. Additionally, we can compose such algorithms with our results from Chapters 6 and 7, to yield improved and more practical reduction algorithms.

We think a potential candidate algorithm that might answer the question above is *bagging*: given a training set $S$, randomly sampling with replacement subsets $P \subseteq S$ (of a certain size), calling $\mathsf{RERM}_{\mathcal{H}}(P)$, and then outputting a majority-vote over the returned predictors.

**Question 2.** *Given a class $\mathcal{H}$ and a perturbation set $\mathcal{U}$, can we characterize necessary and sufficient conditions for* proper *robust learning?*

From Chapter 3, we understand that *sometimes* proper learning algorithms provably *fail* in adversarially robust learning. But maybe for many natural settings encountered in practice such as $\mathcal{H}$ being a class of neural networks and $\mathcal{U}$ being an $\ell_p$-ball, just calling $\mathsf{RERM}_{\mathcal{H}}$ once or some other form of proper learning suffices for robust learning. When focusing specifically on robust learning guarantees with a single $\mathsf{RERM}_{\mathcal{H}}$ call, we can show that having finite VC dimension of the robust loss class $\text{vc}(\mathcal{L}_{\mathcal{H}}^{\mathcal{U}})$ is both necessary and sufficient (see Chapter 2 for discussion). However, in general, it is difficult to calculate $\text{vc}(\mathcal{L}_{\mathcal{H}}^{\mathcal{U}})$ because of the subtle interaction between $\mathcal{H}$ and $\mathcal{U}$. Thus, one can explore natural conditions on $\mathcal{H}$ and $\mathcal{U}$ under which it is possible to bound $\text{vc}(\mathcal{L}_{\mathcal{H}}^{\mathcal{U}})$ from above by some polynomial function in the standard VC dimension $\text{vc}(\mathcal{H})$.

# A

# VC Classes are Adversarially Robustly Learnable, but Only Improperly

## A.1 Auxilliary Proofs Related to Proper Robust Learnability

*Proof of Lemma 3.3.* This proof follows standard lower bound techniques that use the probabilistic method (Shalev-Shwartz and Ben-David, 2014, Chapter 5). Let $m \in \mathbb{N}$. Construct $\mathcal{H}_0$ as before, according to Lemma 3.2, on $3m$ points $x_1, \dots, x_{3m}$. By construction, we know that $\mathcal{L}^{\mathcal{U}}_{\mathcal{H}_0}$ shatters the set $C = \{(x_1, +1), \dots, (x_{3m}, +1)\}$. We will only keep a subset $\mathcal{H}$ of $\mathcal{H}_0$ that includes classifiers that are robustly correct only on subsets of size $2m$, i.e. $\mathcal{H} = \{h_b \in \mathcal{H}_0 : \sum_{i=1}^{3m} b_i = m\}$. Let $\mathcal{A} : (\mathcal{X} \times \mathcal{Y})^* \to \mathcal{H}$ be an arbitrary proper learning rule. The main idea here is to construct a family of distributions that are supported only on $2m$ points of $C$, which would force rule $\mathcal{A}$ to choose which points it can afford to be not correctly robust on. If rule $\mathcal{A}$ observes only $m$ points, it can't do anything better than guessing which of the remaining $2m$ points of $C$ are actually included in the support of the distribution.

Consider a family of distributions $\mathcal{D}_1, \dots, \mathcal{D}_T$ where $T = \binom{3m}{2m}$, each distribution $\mathcal{D}_i$ is uniform over only $2m$ points in $C$. For every distribution $\mathcal{D}_i$, by construction of $\mathcal{H}$, there exists a classifier $h^* \in \mathcal{H}$ such that $\mathrm{R}_{\mathcal{U}}(h^*; \mathcal{D}_i) = 0$. This satisfies the first requirement.

For the second requirement, we will use the probabilistic method to show that there exists a distribution $\mathcal{D}_i$ such that $\mathbb{E}_{S\sim\mathcal{D}_i^m}\left[\mathrm{R}_{\mathcal{U}}(A(S);\mathcal{D}_i)\right] \geq 1/4$, and finish the proof using a variant of Markov's inequality.

Pick an arbitrary sequence $S \in C^m$. Consider a uniform weighting over the distributions $\mathcal{D}_1, \dots, \mathcal{D}_T$. Denote by $E_S$ the event that $S \subset \mathrm{supp}(\mathcal{D}_i)$ for a distribution $\mathcal{D}_i$ that is picked uniformly at random. We will lower bound the expected robust loss of the classifier that rule $\mathcal{A}$ outputs, namely $\mathcal{A}(S) \in \mathcal{H}$, given the event $E_S$,

$$\mathbb{E}_{\mathcal{D}_i}[\mathrm{R}_{\mathcal{U}}(\mathcal{A}(S);\mathcal{D}_i)|E_S] = \mathbb{E}_{\mathcal{D}_i}\left[\mathbb{E}_{(x,y)\sim\mathcal{D}_i}\left[\sup_{z\in\mathcal{U}(x)} \mathbb{1}[\mathcal{A}(S)(z) \neq y]\right]\middle| E_S\right]. \tag{A.1}$$

We can lower bound the robust loss of the classifier $\mathcal{A}(S)$ by conditioning on the event that $(x,y) \notin S$ denoted $E_{(x,y)\notin S}$,

$$\mathop{\mathbb{E}}_{(x,y)\sim\mathcal{D}_i}\left[\sup_{z\in\mathcal{U}(x)} \mathbb{1}[\mathcal{A}(S)(z) \neq y]\right] \geq \mathop{\mathbb{P}}_{(x,y)\sim\mathcal{D}_i}[E_{(x,y)\notin S}] \mathop{\mathbb{E}}_{(x,y)\sim\mathcal{D}_i}\left[\sup_{z\in\mathcal{U}(x)} \mathbb{1}[\mathcal{A}(S)(z) \neq y]|E_{(x,y)\notin S}\right].$$

Since $|S| = m$, and $\mathcal{D}_i$ is uniform over its support of size $2m$, we have $\mathbb{P}_{(x,y)\sim\mathcal{D}_i}[E_{(x,y)\notin S}] \geq 1/2$. This allows us to get a lower bound on (A.1),

$$\mathbb{E}_{\mathcal{D}_i}\left[\mathrm{R}_{\mathcal{U}}(\mathcal{A}(S);\mathcal{D}_i)|E_S\right] \geq \frac{1}{2}\mathbb{E}_{\mathcal{D}_i}\left[\mathop{\mathbb{E}}_{(x,y)\sim\mathcal{D}_i}\left[\sup_{z\in\mathcal{U}(x)} \mathbb{1}[\mathcal{A}(S)(z) \neq y]\middle|E_{(x,y)\notin S}\right]\middle|E_S\right]. \tag{A.2}$$

Since $\mathcal{A}(S) \in \mathcal{H}$, by construction of $\mathcal{H}$, we know that there are at least $m$ points in $C$ where $\mathcal{A}(S)$ is not robustly correct. We can unroll the expectation over $\mathcal{D}_i$ as follows

$$\begin{aligned}
&\mathbb{E}_{\mathcal{D}_i}\left[\mathop{\mathbb{E}}_{(x,y)\sim\mathcal{D}_i}\left[\sup_{z\in\mathcal{U}(x)} \mathbb{1}[\mathcal{A}(S)(z) \neq y]|E_{(x,y)\notin S}\right]\middle|E_S\right] \\
&\geq \frac{1}{m}\sum_{(x,y)\notin S}\mathbb{E}_{\mathcal{D}_i}[\mathbb{1}_{(x,y)\in\mathrm{supp}(\mathcal{D}_i)}|E_S]\sup_{z\in\mathcal{U}(x)}\mathbb{1}[\mathcal{A}(S)(z) \neq y] \geq \frac{1}{m}\sum_{(x,y)\notin S}\frac{1}{2}\sup_{z\in\mathcal{U}(x)} \mathbb{1}[\mathcal{A}(S)(z) \neq y] \geq \frac{1}{2}.
\end{aligned}$$

Thus, it follows by (A.2) that $\mathbb{E}_{\mathcal{D}_i}\left[\mathrm{R}_{\mathcal{U}}(\mathcal{A}(S);\mathcal{D}_i)|E_S\right] \geq \frac{1}{4}$. Now, by law of total expectation,

$$\mathbb{E}_{\mathcal{D}_i}\left[\mathbb{E}_{S\sim\mathcal{D}_i^m}[\mathrm{R}_{\mathcal{U}}(\mathcal{A}(S);\mathcal{D}_i)]\right] = \mathbb{E}_{S\sim\mathcal{D}_i^m}[\mathbb{E}_{\mathcal{D}_i}[\mathrm{R}_{\mathcal{U}}(\mathcal{A}(S);\mathcal{D}_i)|E_S]] \geq \tfrac{1}{4}.$$

Since the expectation over $\mathcal{D}_1, \ldots, \mathcal{D}_T$ is at least $1/4$, this implies that there exists a distribution $\mathcal{D}_i$ such that $\mathbb{E}_{S\sim\mathcal{D}_i^m}\left[\mathrm{R}_{\mathcal{U}}(\mathcal{A}(S); \mathcal{D}_i)\right] \geq 1/4$. Using a variant of Markov's inequality, for any random variable $Z$ taking values in $[0,1]$, and any $a \in (0,1)$, we have $\mathbb{P}[Z > 1-a] \geq \frac{\mathbb{E}[Z]-(1-a)}{a}$. For $Z = \mathrm{R}_{\mathcal{U}}(\mathcal{A}(S); \mathcal{D}_i)$ and $a = 7/8$, we get $\mathbb{P}_{S\sim\mathcal{D}_i^m}\left[\mathrm{R}_{\mathcal{U}}(\mathcal{A}(S); \mathcal{D}_i) > \frac{1}{8}\right] \geq \frac{1/4-1/8}{7/8} = \frac{1}{7}$. □

## A.2 Auxilliary Proofs Related to Realizable Robust Learnability

The following lemma extends the classic compression-based generalization guarantees from the 0-1 loss to also hold for the robust loss. It is used in the proof of Theorem 3.4. Generally, it is also possible to extend other generalization guarantees for compression schemes to the robust loss, such as improved bounds for permutation-invariant compression schemes, or convergence guarantees for the agnostic case (as discussed in Section 3.3.2).

**Lemma A.1.** *For any $k \in \mathbb{N}$ and fixed function $\varphi : (\mathcal{X} \times \mathcal{Y})^k \to \mathcal{Y}^{\mathcal{X}}$, for any distribution $P$ over $\mathcal{X}\times\mathcal{Y}$ and any $m \in \mathbb{N}$, for $S = \{(x_1, y_1), \ldots, (x_m, y_m)\}$ iid $P$-distributed random variables, with probability at least $1-\delta$, if $\exists i_1, \ldots, i_k \in \{1, \ldots, m\}$ s.t. $\hat{R}_{\mathcal{U}}(\varphi((x_{i_1}, y_{i_1}), \ldots, (x_{i_k}, y_{i_k})); S) = 0$, then*

$$R_{\mathcal{U}}(\varphi((x_{i_1}, y_{i_1}), \ldots, (x_{i_k}, y_{i_k})); P) \leq \frac{1}{m-k}(k\ln(m) + \ln(1/\delta)).$$

*Proof.* For completeness, we include a brief proof, which merely notes that the classic argument of (Littlestone and Warmuth, 1986, Floyd and Warmuth, 1995) establishing generalization guarantees for sample compression schemes under the 0-1 loss remains valid under the robust loss.

For any indices $i_1, \ldots, i_k \in \{1, \ldots, m\}$,

$$\begin{aligned}
&\mathbb{P}\Big(\hat{\mathrm{R}}_{\mathcal{U}}(\varphi(\{(x_{i_j}, y_{i_j})\}_{j=1}^k); S) = 0 \text{ and } \mathrm{R}_{\mathcal{U}}(\varphi(\{(x_{i_j}, y_{i_j})\}_{j=1}^k); P) > \varepsilon\Big) \\
&\leq \mathbb{E}\Big[\mathbb{P}\Big(\hat{\mathrm{R}}_{\mathcal{U}}(\varphi(\{(x_{i_j}, y_{i_j})\}_{j=1}^k); S \setminus \{(x_{i_j}, y_{i_j})\}_{j=1}^k) = 0 \Big| \{(x_{i_j}, y_{i_j})\}_{j=1}^k\Big) \times \\
&\qquad\qquad \mathbb{1}\Big[\mathrm{R}_{\mathcal{U}}(\varphi(\{(x_{i_j}, y_{i_j})\}_{j=1}^k); P) > \varepsilon\Big]\Big] \\
&< (1-\varepsilon)^{m-k},
\end{aligned}$$

and a union bound over all $m^k$ possible choices of $i_1, \ldots, i_k$ implies a probability at most $m^k(1-\varepsilon)^{m-k} \leq m^k e^{-\varepsilon(m-k)}$ that there exist $i_1, \ldots, i_k$ with $\mathrm{R}_{\mathcal{U}}(\varphi(\{(x_{i_j}, y_{i_j})\}_{j=1}^k); \mathcal{P}_{XY}) > \varepsilon$ and yet $\hat{\mathrm{R}}_{\mathcal{U}}(\varphi(\{(x_{i_j}, y_{i_j})\}_{j=1}^k); S) = 0$. This is at most $\delta$ for a choice of $\varepsilon = \frac{1}{m-k}(k\ln(m) + \ln(1/\delta))$. □

## A.3 Proof of Agnostic Robust Learnability

*Proof of Theorem 3.8.* The argument follows closely a proof of an analogous result by David, Moran, and Yehudayoff (2016) for non-robust learning. Denote by $\mathbb{A}$ the optimal realizable-case learner achieving sample complexity $\mathcal{M}^{\mathrm{re}}_{1/3,1/3}(\mathcal{H}, \mathcal{U})$, and denote $m_0 = \mathcal{M}^{\mathrm{re}}_{1/3,1/3}(\mathcal{H}, \mathcal{U})$, as above.

Then, in the agnostic case, given a data set $S \sim \mathcal{D}^m$, we first do robust-ERM to find a maximal-size subsequence $S'$ of the data where the robust loss can be zero: that is, $\inf_{h\in\mathcal{H}} \hat{\mathrm{R}}_{\mathcal{U}}(h; S') = 0$. Then for any distribution $D$ over $S'$, there exists a sequence $S_D \in (S')^{m_0}$ such that $h_D := \mathbb{A}(S_D)$ has $\mathrm{R}_{\mathcal{U}}(h_D; D) \leq 1/3$; this follows since, by definition of $\mathcal{M}^{\mathrm{re}}_{1/3,1/3}(\mathcal{H}, \mathcal{U})$, there is a $1/3$ chance that $\hat{S}$ a random draw from $D^{m_0}$ yields $\mathrm{R}_{\mathcal{U}}(\mathbb{A}(\hat{S}); D) \leq 1/3$, so at least one such $S_D$ exists. We use this to define a weak robust-learner for distributions $D$ over $S'$: i.e., for any $D$, the weak learner chooses $h_D$ as its weak hypothesis.

Now we run the $\alpha$-Boost boosting algorithm (Schapire and Freund, 2012, Section 6.4.2) on data set $S'$, but using the robust loss rather than 0-1 loss. That is, we start with $D_1$ uniform on $S'$.* Then for each round $t$, we get $h_{D_t}$ as a weak robust classifier with respect to $D_t$, and for each $(x, y) \in S'$ we define a distribution $\mathcal{D}_{t+1}$ over $S'$ satisfying

$$D_{t+1}(\{(x,y)\}) \propto D_t(\{(x,y)\}) \exp\{-2\alpha 1[\forall x' \in \mathcal{U}(x), h_{D_t}(x') = y]\},$$

where $\alpha$ is a parameter we can set. Following the argument from Schapire and Freund (2012, Section 6.4.2), after $T$ rounds we are guaranteed

$$\min_{(x,y)\in S'} \frac{1}{T}\sum_{t=1}^{T} 1[\forall x' \in \mathcal{U}(x), h_{D_t}(x') = y] \geq \frac{2}{3} - \frac{2}{3}\alpha - \frac{\ln(|S'|)}{2\alpha T},$$

*We ignore the possibility of repeats; for our purposes we can just remove any repeats from $S'$ before this boosting step.

so we will plan on running until round $T = 1 + 48\ln(|S'|)$ with value $\alpha = 1/8$ to guarantee

$$\min_{(x,y)\in S'} \frac{1}{T}\sum_{t=1}^{T} 1[\forall x' \in \mathcal{U}(x), h_{D_t}(x') = y] > \frac{1}{2},$$

so that the classifier $\hat{h}(x) := 1\left[\frac{1}{T}\sum_{t=1}^{T} h_{D_t}(x) \geq \frac{1}{2}\right]$ has $\hat{\mathrm{R}}_{\mathcal{U}}(\hat{h}; S') = 0$.

Furthermore, note that, since each $h_{D_t}$ is given by $\mathbb{A}(S_{D_t})$, where $S_{D_t}$ is an $m_0$-tuple of points in $S'$, the classifier $\hat{h}$ is specified by an ordered sequence of $m_0 T$ points from $S$. Altogether, $\hat{h}$ is a function specified by an ordered sequence of $m_0 T$ points from $S$, and which has

$$\hat{\mathrm{R}}_{\mathcal{U}}(\hat{h}; S) \leq \min_{h\in\mathcal{H}} \hat{\mathrm{R}}_{\mathcal{U}}(h; S).$$

Similarly to the realizable case (see the proof of Lemma A.1), uniform convergence guarantees for sample compression schemes (see Graepel, Herbrich, and Shawe-Taylor, 2005) remain valid for the robust loss, by essentially the same argument; the essential argument is the same as in the proof of Lemma A.1 except using Hoeffding's inequality to get concentration of the empirical robust risks for each fixed index sequence, and then a union bound over the possible index sequnces as before. We omit the details for brevity. In particular, denoting $T_m = 1 + 48\ln(m)$, for $m > m_0 T_m$, with probability at least $1 - \delta/2$,

$$\mathrm{R}_{\mathcal{U}}(\hat{h}; \mathcal{D}) \leq \hat{\mathrm{R}}_{\mathcal{U}}(\hat{h}; S) + \sqrt{\frac{m_0 T_m \ln(m) + \ln(2/\delta)}{2m - 2m_0 T_m}}.$$

Let $h^* = \mathrm{argmin}_{h\in\mathcal{H}} \mathrm{R}_{\mathcal{U}}(h; \mathcal{D})$ (supposing the min is realized, for simplicity; else we could take an $h^*$ with very-nearly minimal risk). By Hoeffding's inequality, with probability at least $1 - \delta/2$,

$$\hat{\mathrm{R}}_{\mathcal{U}}(h^*; S) \leq \mathrm{R}_{\mathcal{U}}(h^*; \mathcal{D}) + \sqrt{\frac{\ln(2/\delta)}{2m}}.$$

By the union bound, if $m \geq 2m_0 T_m$, with probability at least $1-\delta$,

$$\begin{aligned}
\mathrm{R}_{\mathcal{U}}(\hat{h};\mathcal{D}) &\leq \min_{h\in\mathcal{H}} \hat{\mathrm{R}}_{\mathcal{U}}(h;S) + \sqrt{\frac{m_0 T_m \ln(m) + \ln(2/\delta)}{m}} \\
&\leq \hat{\mathrm{R}}_{\mathcal{U}}(h^*;S) + \sqrt{\frac{m_0 T_m \ln(m) + \ln(2/\delta)}{m}} \\
&\leq \mathrm{R}_{\mathcal{U}}(h^*;\mathcal{D}) + 2\sqrt{\frac{m_0 T_m \ln(m) + \ln(2/\delta)}{m}}.
\end{aligned}$$

Since $T_m = O(\log(m))$, the above is at most $\varepsilon$ for an appropriate choice of sample size $m = O\big(\frac{m_0}{\varepsilon^2}\log^2\big(\frac{m_0}{\varepsilon}\big) + \frac{1}{\varepsilon^2}\log\big(\frac{1}{\delta}\big)\big)$. □

## A.4 Auxilliary Proofs Related to Necessary Conditions for Robust Learnability

*Proof of Proposition 3.9.* Let $\mathcal{X} = \mathbb{R}^d$ equipped with a metric $\rho$, and $\mathcal{U} : \mathcal{X} \to 2^{\mathcal{X}}$ such that $\mathcal{U}(x) = \{z \in \mathcal{X} : \rho(x,z) \leq \gamma\}$ for all $x \in \mathcal{X}$ for some $\gamma > 0$. Consider two infinite sequences of points $(x_m)_m \in \mathbb{N}$ and $(z_m)_m \in \mathbb{N}$ such that for any $i \neq j, \mathcal{U}(x_i) \cap \mathcal{U}(x_j) = \emptyset$, $\mathcal{U}(x_i) \cap \mathcal{U}(z_j) = \emptyset, \mathcal{U}(x_j) \cap \mathcal{U}(z_i) = \emptyset$, but $\mathcal{U}(x_i) \cap \mathcal{U}(z_i) = u_i$. In other words, we want the $\gamma$-balls of pairs with different indices to be mutually disjoint, and the $\gamma$-balls for a pair with the same index to intersect at a single point (this is possible because we are considering closed balls).

Next, we proceed with the construction of $\mathcal{H}$. For each bit string $b \in \{0,1\}^{\mathbb{N}}$, we will define a predictor $h_b : \mathcal{X} \to \mathcal{Y}$ just on the $\gamma$-balls of the points $x_1, z_1, x_2, z_2, \ldots$ (it labels the rest of the $\mathcal{X}$ space with $+1$). Foreach $i \in \mathbb{N}$, if $b_i = 0$, set

$$h_b\Big(\mathcal{U}(x_i)\Big) = +1 \quad \text{and} \quad h_b\Big(\mathcal{U}(z_i) \setminus \mathcal{U}(x_i)\Big) = -1$$

and if $b_i = 1$, set

$$h_b\Big(\mathcal{U}(x_i) \setminus \mathcal{U}(z_i)\Big) = -1 \quad \text{and} \quad h_b\Big(\mathcal{U}(z_i)\Big) = +1$$

Let $\mathcal{H} = \{h_b : b \in \{0,1\}^{\mathbb{N}}\}$. Notice that $\dim_{\mathcal{U}\times}(\mathcal{H}) = 0$, because there is no single $\gamma$-ball that is labeled in both ways ($+1$ and $-1$). By construction of $\mathcal{H}$, all classifiers $h_b \in \mathcal{H}$ behave the same way on all points in $\mathcal{X}$, except at points in the intersections $u_1, u_2, \ldots$ which get shattered. However, the $\mathcal{U}$-robust shattering dimension (see definition 3.10) is infinite in this construction ($\dim_{\mathcal{U}}(\mathcal{H}) = \infty$), which by Theorem 3.11 (see below) implies that $\mathcal{M}^{\mathrm{re}}_{\varepsilon,\delta}(\mathcal{H},\mathcal{U}) = \infty$. □

*Proof Sketch of Theorem 3.11.* We first start with the realizable case. The proof follows a standard argument from (Mohri et al., 2018, Chapter 3). Let $\mathrm{d} = \dim_{\mathcal{U}}(\mathcal{H})$, and fix $x_1, \ldots, x_{\mathrm{d}}$ a sequence $\mathcal{U}$-robustly shattered by $\mathcal{H}$, and let $z_1^+, z_1^-, \ldots, z_{\mathrm{d}}^+, z_{\mathrm{d}}^- \in \mathcal{X}$ be as in definition 3.10; in particular, note that any $y, y'$ and any $i \neq j$ necessarily have $z_i^y \neq z_j^{y'}$. For each $\mathbf{y} = (y_1, \ldots, y_{\mathrm{d}}) \in \{+1, -1\}^{\mathrm{d}}$, let $h^{\mathbf{y}} \in \mathcal{H}$ be such that $\forall i \in [m], \forall z' \in \mathcal{U}(z_i^{y_i}), h^{\mathbf{y}}(z') = y_i$. Let $\mathcal{D}$ be a distribution over $\{1, 2, \ldots, \mathrm{d}\}$ such that $\mathbb{P}_{i\sim\mathcal{D}}[i = 1] = 1 - 8\varepsilon$ and $\mathbb{P}_{i\sim\mathcal{D}}[i = 1] = 8\varepsilon/(\mathrm{d}-1)$ for $2 \leq i \leq d$. Now choose $\mathbf{y} \sim \mathrm{Uniform}(\{+1, -1\}^{\mathrm{d}})$, and let $\mathcal{D}_{\mathbf{y}}$ be the induced distribution over $\mathcal{X} \times \mathcal{Y}$ such that

$$\mathbb{P}_{(x,y)\sim\mathcal{D}_{\mathbf{y}}}\left[(x,y) = (z_1^{y_1}, y_1)\right] = 1 - 8\varepsilon \text{ and } \mathbb{P}_{(x,y)\sim\mathcal{D}_{\mathbf{y}}}\left[(x,y) = (z_i^{y_i}, y_i)\right] = 8\varepsilon/(\mathrm{d}-1)$$

for $2 \leq i \leq \mathrm{d}$.

Note that by construction we have $\mathrm{R}_{\mathcal{U}}(h^{\mathbf{y}}; \mathcal{D}) = 0$. Now, consider an arbitrary learning rule $\mathcal{A} : (\mathcal{X} \times \mathcal{Y})^* \mapsto \mathcal{Y}^{\mathcal{X}}$. We will assume that $\mathcal{A}$ always gets the prediction of $z_1^{y_1}$ correct. Let $I = \{2, \ldots, d\}$ and let $\mathcal{S}$ be the set of all sequences of size $m$ containing at most $(\mathrm{d}-1)/2$ elements from $I$. Fix an arbitrary sequence $S \in \mathcal{S}$. Denote by $S_{\mathbf{y}} = ((z_i^{y_i}, y_i) : i \in \mathcal{S})$ the sequence of examples induced by the indices sequence $S$. Then,

$$\begin{aligned}
\mathbb{E}_{\mathbf{y}}\left[\mathrm{R}_{\mathcal{U}}(\mathcal{A}(S_{\mathbf{y}}); \mathcal{D}_{\mathbf{y}})\right] &\geq \mathbb{E}_{\mathbf{y}}\left[\sum_{i \notin S} \mathbb{P}_{\mathcal{D}_{\mathbf{y}}}(z_i^{y_i}) \sup_{z' \in \mathcal{U}(z_i^{y_i})} 1[\mathcal{A}(S_{\mathbf{y}})(z') \neq y_i]\right] \\
&\geq \frac{\mathrm{d}-1}{2} \times \frac{8\varepsilon}{\mathrm{d}-1} \times \mathbb{E}_{\mathbf{y}}\left[\sup_{z' \in \mathcal{U}(z)} 1[\mathcal{A}(S_{\mathbf{y}})(z') \neq y]\right] \\
&= \frac{\mathrm{d}-1}{2} \times \frac{8\varepsilon}{\mathrm{d}-1} \times \frac{1}{2} \\
&= 2\varepsilon
\end{aligned}$$

Since the inequality above holds for any sequence $S \in \mathcal{S}$, it follows that

$$\mathbb{E}_{S\sim\mathcal{D}^m}[\mathbb{E}_{\mathbf{y}}\left[\mathrm{R}_{\mathcal{U}}(\mathcal{A}(S_{\mathbf{y}});\mathcal{D}_{\mathbf{y}})1_{S\in\mathcal{S}}\right]] = \mathbb{E}_{\mathbf{y}}[\mathbb{E}_{S\sim\mathcal{D}^m}\left[\mathrm{R}_{\mathcal{U}}(\mathcal{A}(S_{\mathbf{y}});\mathcal{D}_{\mathbf{y}})|E_{S\in\mathcal{S}}\right]] \geq 2\varepsilon$$

Which implies that there exists $\mathbf{y}_0$ such that $\mathbb{E}_{S\sim\mathcal{D}^m}\left[\mathrm{R}_{\mathcal{U}}(\mathcal{A}(S_{\mathbf{y}_0});\mathcal{D}_{\mathbf{y}_0})|E_{S\in\mathcal{S}}\right] \geq 2\varepsilon$. Since $\mathbb{P}_{\mathcal{D}}[i \in I] \leq 8\varepsilon$, the robust risk $\mathrm{R}_{\mathcal{U}}(\mathcal{A}(S_{\mathbf{y}_0}));\mathcal{D}_{\mathbf{y}}) \leq 8\varepsilon$. Then, by law of total expectation, we have

$$\begin{aligned}2\varepsilon &\leq \mathbb{E}_{S\sim\mathcal{D}^m}\left[\mathrm{R}_{\mathcal{U}}(\mathcal{A}(S_{\mathbf{y}_0});\mathcal{D}_{\mathbf{y}_0})|E_{S\in\mathcal{S}}\right]\\ &\leq 8\varepsilon\mathbb{P}_{S\sim\mathcal{D}^m}\left[\mathrm{R}_{\mathcal{U}}(\mathcal{A}(S_{\mathbf{y}_0});\mathcal{D}_{\mathbf{y}}) \geq \varepsilon|E_{S\in\mathcal{S}}\right] + \varepsilon(1 - \mathbb{P}_{S\sim\mathcal{D}^m}\left[\mathrm{R}_{\mathcal{U}}(\mathcal{A}(S_{\mathbf{y}_0});\mathcal{D}_{\mathbf{y}_0}) \geq \varepsilon|\, E_{S\in\mathcal{S}}\right])\end{aligned}$$

By collecting terms, we obtain that $\mathbb{P}_{S\sim\mathcal{D}^m}\left[\mathrm{R}_{\mathcal{U}}(\mathcal{A}(S_{\mathbf{y}_0});\mathcal{P}_{XY}) \geq \varepsilon|\, E_{S\in\mathcal{S}}\right] \geq 1/7$. Then, by law of total probability, the probability over all sequences (not necessarily in $\mathcal{S}$) can be lower bounded,

$$\mathbb{P}_{S\sim\mathcal{D}^m}\left[\mathrm{R}_{\mathcal{U}}(\mathcal{A}(S_{\mathbf{y}_0});\mathcal{D}_{\mathbf{y}_0}) \geq \varepsilon\right] \geq \mathbb{P}[E_{S\in\mathcal{S}}]\mathbb{P}_{S\sim\mathcal{D}^m}\left[\mathrm{R}_{\mathcal{U}}(\mathcal{A}(S_{\mathbf{y}_0});\mathcal{D}_{\mathbf{y}_0}) \geq \varepsilon|E_{S\in\mathcal{S}}\right] \geq \frac{1}{7}\mathbb{P}[E_{S\in\mathcal{S}}]$$

By a standard application of Chernoff bounds, for $\varepsilon = \frac{\mathrm{d}-1}{32m}$ and $\delta \leq 1/100$, we get that $\mathbb{P}[E_{S\in\mathcal{S}}] \geq 7\delta$ and by the above this concludes that $\mathbb{P}_{S\sim\mathcal{D}^m}\left[\mathrm{R}_{\mathcal{U}}(\mathcal{A}(S_{\mathbf{y}_0});\mathcal{D}_{\mathbf{y}_0}) \geq \varepsilon\right] \geq \delta$. This establishes that

$$\mathcal{M}^{\mathrm{re}}_{\varepsilon,\delta}(\mathcal{H},\mathcal{U}) \geq \Omega\left(\frac{\mathrm{d}}{\varepsilon}\right)$$

To finish the proof, we need to show that

$$\mathcal{M}^{\mathrm{re}}_{\varepsilon,\delta}(\mathcal{H},\mathcal{U}) \geq \Omega\left(\frac{1}{\varepsilon}\log\left(\frac{1}{\delta}\right)\right)$$

For this just consider a distribution $P_1$ with mass $1-\varepsilon$ on $(z_1^+, +1)$ and mass $\varepsilon$ on $(z_2^+, +1)$, and another distribution $P_2$ with mass $1-\varepsilon$ on $(z_1^+, +1)$ and mass $\varepsilon$ on $(z_2^-, -1)$. If $m \leq (1/2\varepsilon)\ln(1/\delta)$, with probability at least $\delta$, we will only observe $m$ samples of $(z_1^+, +1)$, and thus learning rule $\mathcal{A}$ will make a mistake on $x_2$ (which is in $\mathcal{U}(z_2^+) \cap \mathcal{U}(z_2^-)$) with probability at least $1/2$, therefore having error at least $\varepsilon/2$. By combining both parts, we arrive at the theorem statement.

For the agnostic case, we briefly describe the construction. The remainder of the proof more or less follows a standard argument, for instance see Anthony and Bartlett (1999, Chapter 5). Let $\mathrm{d} = \dim_{\mathcal{U}}(\mathcal{H})$, and fix $x_1, \ldots, x_{\mathrm{d}}$ a sequence $\mathcal{U}$-robustly shattered by $\mathcal{H}$, and let $z_1^+, z_1^-, \ldots, z_{\mathrm{d}}^+, z_{\mathrm{d}}^- \in \mathcal{X}$ be as in definition 3.10; in particular, note that any $y, y'$ and any $i \neq j$ necessarily have $z_i^y \neq z_j^{y'}$. For $b \in \{0,1\}^{\mathrm{d}}$, define distribution $\mathcal{D}_b$ as follows, for $i \in [\mathrm{d}]$:

- If $b_i = 0$, then set $\mathbb{P}_{\mathcal{D}_b}((z_i^+, +1)) = (1-\alpha)/(2\mathrm{d})$ and $\mathbb{P}_{\mathcal{D}_b}((z_i^-, -1)) = (1+\alpha)/(2\mathrm{d})$.
- If $b_i = 1$, then set $\mathbb{P}_{\mathcal{D}_b}((z_i^+, +1)) = (1+\alpha)/(2\mathrm{d})$ and $\mathbb{P}_{\mathcal{D}_b}((z_i^-, -1)) = (1-\alpha)/(2\mathrm{d})$.

where ) $< \alpha < 1$ is appropriately chosen based on $\varepsilon$ and $\delta$. □

# B

# A Characterization of Adversarially Robust Learnability

## B.1 Converting an Orientation to a Learner (Proof of Lemma 4.2)

*Proof.* Let $\mathcal{O} : E_n \to V_n$ be an arbitrary orientation of $G^{\mathcal{U}}_{\mathcal{H}}$. We will show that orientation $\mathcal{O}$ implies a learner $\mathbb{A}_{\mathcal{O}} : (\mathcal{X} \times \mathcal{Y})^{n-1} \to \mathcal{Y}^{\mathcal{X}}$ with an expected robust risk $\mathcal{E}_{n-1}$ that is upper bounded by the maximum adversarial out-degree of orientation $\mathcal{O}$.

We begin with describing the learner $\mathbb{A}_{\mathcal{O}}$. For each input $((x_1, y_1), \dots, (x_{n-1}, y_{n-1}), z) \in (\mathcal{X} \times \mathcal{Y})^{n-1} \times \mathcal{X}$ define $\mathbb{A}_{\mathcal{O}}((x_1, y_1), \dots, (x_{n-1}, y_{n-1}))(z)$ as follows. Consider the set of vertices $v \in V$ that have the multiset $\{(x_1, y_1), \dots, (x_{n-1}, y_{n-1})\}$ and perturbation $z$ with a positive label

$$P_+ = \{v \in V : \exists x \in \mathcal{X} \text{ s.t. } z \in \mathcal{U}(x) \wedge v = \{(x_1, y_1), \dots, (x_{n-1}, y_{n-1}), (x, +1)\}\},$$

and the set of vertices $v \in V$ that have the multiset $\{(x_1, y_1), \dots, (x_{n-1}, y_{n-1})\}$ and perturbation $z$ with a negative label

$$P_- = \{v \in V : \exists x \in \mathcal{X} \text{ s.t. } z \in \mathcal{U}(x) \wedge v = \{(x_1, y_1), \dots, (x_{n-1}, y_{n-1}), (x, -1)\}\}.$$

We define $\mathbb{A}_{\mathcal{O}}((x_1,y_1),\ldots,(x_{n-1},y_{n-1}))(z)$ as a function of $P_+$, $P_-$, and the orientation $\mathcal{O}$:

$$\mathbb{A}_{\mathcal{O}}((x_1,y_1),\ldots,(x_{n-1},y_{n-1}))(z) = \begin{cases} y & \text{if } \left(\exists_{y\in\{\pm 1\}}\right)\left(\exists_{v\in P_y}\right)\left(\forall_{u\in P_{-y}}\right) : \mathcal{O}((\{v,u\},z)) = v. \\ +1 & \text{if } P_+ \neq \emptyset \wedge P_- = \emptyset. \\ -1 & \text{if } P_+ = \emptyset \wedge P_- \neq \emptyset. \\ +1 & \text{otherwise.} \end{cases}$$

Note that $\mathbb{A}_{\mathcal{O}}$ is well-defined. Specifically, observe that when $P_+ \neq \emptyset$ and $P_- \neq \emptyset$, by definition of $P_+$ and $P_-$ and Equation 4.7, vertices from $P_+$ and $P_-$ form a complete bipartite graph. That is, for each $v \in P_+$ and each $u \in P_-$, $(\{u,v\},z) \in E$. This implies that there exists at most one label: either $y = +1$ or $y = -1$ such that there is a vertex $v \in P_y$ where all edges $(\{v,u\},z) \in E$ for $u \in P_{-y}$ are incident on $v$ according to orientation $\mathcal{O}$: $(\exists! y \in \{\pm 1\})\left(\exists v \in P_y\right)\left(\forall u \in P_{-y}\right) : \mathcal{O}((\{v,u\},z)) = v$.

We now proceed with bounding from above the expected robust risk $\mathcal{E}_{n-1}$ of learner $\mathbb{A}_{\mathcal{O}}$ by the maximum adversarial out-degree of orientation $\mathcal{O}$. Consider an arbitrary multiset $S = \{(x_1,y_1),\ldots,(x_n,y_n)\} \in (\mathcal{X}\times\mathcal{Y})^n$ that is *robustly* realizable with respect to $(\mathcal{H},\mathcal{U})$. By definition $S \in V_n$, i.e., $S$ is a vertex in $G^{\mathcal{U}}_{\mathcal{H}}$. By definition of adversarial out-degree (see Equation 4.9), there exists $T \subseteq S$ where $|T| = n - \text{adv-outdeg}(S;\mathcal{O})$ such that for each $(x,y) \in T$ and for each $z \in \mathcal{U}(x)$: vertex $S$ will satisfy the condition that if there is any other vertex $u \in V_n$ where $(\{S,u\},z)$ is an edge: $(\{S,u\},z) \in E_n$, the orientation of edge $(\{S,u\},z)$ is towards $S$: $\mathcal{O}((\{S,u\},z)) = S$. Thus by definition of $\mathbb{A}_{\mathcal{O}}$ above, $\mathbb{A}_{\mathcal{O}}\left(S \setminus \{(x,y)\},z\right) = y$. This implies that

$$\frac{1}{n}\sum_{i=1}^{n} 1\left[\exists z \in \mathcal{U}(x_i) : \mathbb{A}_{\mathcal{O}}\left(S \setminus \{(x_i,y_i)\}\right)(z) \neq y_i\right] = \frac{\text{adv-outdeg}(S;\mathcal{O})}{n}.$$

To conclude, by definition of $\mathcal{E}_{n-1}(\mathbb{A}_{\mathcal{O}};\mathcal{H},\mathcal{U})$ (see Equation 4.1),

$$\begin{aligned}
\mathcal{E}_{n-1}(\mathbb{A}_{\mathcal{O}};\mathcal{H},\mathcal{U}) &= \sup_{\mathcal{D}\in\mathrm{RE}(\mathcal{H},\mathcal{U})} \mathop{\mathbb{E}}_{S\sim\mathcal{D}^{n-1}} \mathrm{R}_{\mathcal{U}}(\mathbb{A}_{\mathcal{O}}(S);\mathcal{D}) \\
&= \sup_{\mathcal{D}\in\mathrm{RE}(\mathcal{H},\mathcal{U})} \mathop{\mathbb{E}}_{S\sim\mathcal{D}^{n-1}} \mathop{\mathbb{E}}_{(x,y)\sim\mathcal{D}} \mathbb{1}\left\{\exists z\in\mathcal{U}(x) : \mathbb{A}_{\mathcal{O}}(S)(z)\neq y\right\} \\
&= \sup_{\mathcal{D}\in\mathrm{RE}(\mathcal{H},\mathcal{U})} \mathop{\mathbb{E}}_{S\sim\mathcal{D}^{n}} \frac{1}{n}\sum_{i=1}^{n} \mathbb{1}\left\{\exists z\in\mathcal{U}(x_i) : \mathbb{A}_{\mathcal{O}}(S\setminus\{(x_i,y_i)\})(z)\neq y_i\right\} \\
&\leq \frac{\max_{v\in V_n}\text{adv-outdeg}(v;\mathcal{O})}{n}.
\end{aligned}$$

□

## B.2 A Quantitative Characterization of Robust Learnability (Lemmas and Proofs for Theorem 4.4)

**Lemma B.1** (Rado's Selection Principle (Rado, 1949, Wolk, 1965)). *Let $I$ be an arbitrary index set, and let $\{X_i : i \in I\}$ be a family of non-empty finite sets. For each finite subset $A$ of $I$, let $f_A$ be a choice function whose domain is $A$ and such that $f_A(i) \in X_i$ for each $i \in A$. Then, there exists a choice function $f$ whose domain is $I$ with the following property: to every finite subset $A$ of $I$ there corresponds a finite set $B$, $A \subseteq B \subseteq I$, with $f(i) = f_B(i)$ for each $i \in A$.*

**Lemma B.2** (Refined Lowerbound on Error Rate of Learners). *For any integer $n \geq 2$, let $G^{\mathcal{U}}_{\mathcal{H}} = (V_{2n}, E_{2n})$ be the global one-inclusion graph as defined in Equation 4.6 and Equation 4.7. Then, for any learner $\mathbb{A} : (\mathcal{X}\times\mathcal{Y})^* \to \mathcal{Y}^{\mathcal{X}}$ and any $\varepsilon \in (0,1)$, there exists an orientation $\mathcal{O}_{\mathbb{A}} : E_{2n} \to V_{2n}$ of $G^{\mathcal{U}}_{\mathcal{H}}$ such that*

$$\mathcal{E}_{\frac{n}{\varepsilon}}(\mathbb{A};\mathcal{H},\mathcal{U}) \geq \frac{\varepsilon}{6}\cdot\frac{\max_{v\in V_{2n}}\text{adv-outdeg}(v;\mathcal{O}_{\mathbb{A}})-1}{2n-1}.$$

*Proof of Lemma B.2.* Set $m = \frac{n}{\varepsilon}$. We begin with describing the orientation $\mathcal{O}_{\mathbb{A}}$ by orienting edges incident on each vertex $v \in V_{2n}$. Consider an arbitrary vertex $v = \{(x_1,y_1),\dots,(x_{2n},y_{2n})\}$. Without loss of generality, let $P_v$ be a distribution over $\{(x_1,y_1),\dots,(x_{2n},y_{2n})\}$, defined as

$$P_v(\{(x_1,y_1)\}) = 1-\varepsilon \text{ and } P_v(\{(x_t,y_t)\}) = \frac{\varepsilon}{2n-1}\ \forall 2\leq t\leq 2n.$$

For each $1 \leq t \leq 2n$, let

$$p_t(v) = \Pr_{S \sim P_v^m} \left[ \exists z \in \mathcal{U}(x_t) : \mathbb{A}(S)(z) \neq y_t | (x_t, y_t) \notin S \right].$$

For each $1 \leq t \leq 2n$ such that $(x_t, y_t) \in v$ witnesses an edge, i.e. $\exists u \in V_{2n}, z \in \mathcal{X}$ s.t. $(\{v, u\}, z) \in E_{2n}$ and $(x_t, y_t) \in v \Delta u$, if $p_t < \frac{1}{2}$, then orient *all* edges incident on $(x_t, y_t)$ inward, otherwise orient them arbitrarily. Note that this might yield edges that are oriented outwards from both their endpoint vertices, in which case, we arbitrarily orient such an edge. Observe also that we will not encounter a situation where edges are oriented inwards towards both their endpoints (which is an invalid orientation). This is because for any two vertices $v, u \in V_{2n}$ such that $\exists z_0 \in \mathcal{X}$ where $(\{u, v\}, z_0) \in E_{2n}$ and $v \Delta u = \{(x_t, y_t), (\tilde{x}_t, -y_t)\}$, we can not have $p_t(v) < \frac{1}{2}$ and $p_t(u) < \frac{1}{2}$, since

$$p_t(v) \geq \Pr_{S \sim P_v^m} \left[ \mathbb{A}(S)(z_0) \neq y_t | (x_t, y_t) \notin S \right] \text{ and } p_t(u) \geq \Pr_{S \sim P_u^m} \left[ \mathbb{A}(S)(z_0) \neq -y_t | (\tilde{x}_t, -y_t) \notin S \right],$$

and $P_v$ conditioned on $(x_t, y_t) \notin S$ is the same distribution as $P_u$ conditioned on $(\tilde{x}_t, -y_t) \notin S$. This concludes describing the orientation $\mathcal{O}_{\mathbb{A}}$. We now bound from above the adversarial out-degree of vertices $v \in V_{2n}$ with respect to the orientation $\mathcal{O}_{\mathbb{A}}$:

$$
\begin{aligned}
\text{adv-outdeg}(v; \mathcal{O}_{\mathbb{A}}) &\leq \sum_{t=1}^{2n} 1\left[p_t \geq \frac{1}{2}\right] \leq 1 + \sum_{t=2}^{2n} 1\left[p_t \geq \frac{1}{2}\right] \leq 1 + 2\sum_{t=2}^{2n} p_t \\
&= 1 + 2\sum_{t=2}^{2n} \Pr_{S\sim P^m}\left[\exists z \in \mathcal{U}(x_t) : \mathbb{A}(S)(z) \neq y_t | (x_t, y_t) \notin S\right] \\
&= 1 + 2\sum_{t=2}^{2n} \frac{\Pr_{S\sim P^m}\left[(\exists z \in \mathcal{U}(x_t) : \mathbb{A}(S)(z) \neq y_t) \wedge (x_t, y_t) \notin S\right]}{\Pr_{S\sim P^m}\left[(x_t, y_t) \notin S\right]} \\
&\overset{(i)}{\leq} 1 + 2 \cdot 3\sum_{t=2}^{2n} \Pr_{S\sim P^m}\left[(\exists z \in \mathcal{U}(x_t) : \mathbb{A}(S)(z) \neq y_t) \wedge (x_t, y_t) \notin S\right] \\
&= 1 + 6\sum_{t=2}^{2n} \mathop{\mathbb{E}}_{S\sim P^m}\left[1\left[\exists z \in \mathcal{U}(x_t) : \mathbb{A}(S)(z) \neq y_t\right] 1\left[(x_t, y_t) \notin S\right]\right] \\
&= 1 + 6\mathop{\mathbb{E}}_{S\sim P^m}\left[\sum_{t=2}^{2n} 1\left[\exists z \in \mathcal{U}(x_t) : \mathbb{A}(S)(z) \neq y_t\right] 1\left[(x_t, y_t) \notin S\right]\right] \\
&\leq 1 + 6\mathop{\mathbb{E}}_{S\sim P^m}\left[\sum_{t=2}^{2n} 1\left[\exists z \in \mathcal{U}(x_t) : \mathbb{A}(S)(z) \neq y_t\right]\right] \\
&= 1 + 6 \cdot \frac{2n-1}{\varepsilon}\mathop{\mathbb{E}}_{S\sim P^m}\left[\frac{\varepsilon}{2n-1}\sum_{t=2}^{2n} 1\left[\exists z \in \mathcal{U}(x_t) : \mathbb{A}(S)(z) \neq y_t\right]\right] \\
&\leq 1 + 6 \cdot \frac{2n-1}{\varepsilon}\mathop{\mathbb{E}}_{S\sim P^m} \mathrm{R}_{\mathcal{U}}(\mathbb{A}(S); P) \\
&\leq 1 + 6 \cdot \frac{2n-1}{\varepsilon}\mathcal{E}_m(\mathbb{A}; \mathcal{H}, \mathcal{U}),
\end{aligned}
$$

where inequality $(i)$ follows from the following:

$$
\Pr_{S\sim P^m}\left[(x_t, y_t) \notin S\right] = \left(1 - \frac{\varepsilon}{2n-1}\right)^m \geq 1 - m \cdot \frac{\varepsilon}{2n-1} = 1 - \frac{n}{\varepsilon}\frac{\varepsilon}{2n-1} \geq 1 - \frac{n}{2n-1} \geq \frac{1}{3},
$$

for $n \geq 2$.

Since the above holds for any vertex $v \in V_{2n}$, by rearranging terms, we get $\mathcal{E}_m(\mathbb{A}; \mathcal{H}, \mathcal{U}) \geq \frac{\varepsilon}{6}\frac{\max_{v\in V_{2n}} \text{adv-outdeg}(v; \mathcal{O}_{\mathbb{A}}) - 1}{2n-1}$. □

*Proof of Lemma 4.4.* We will first start with the upper bound. Let $n > \mathfrak{D}_{\mathcal{U}}(\mathcal{H})$ and let $G^{\mathcal{U}}_{\mathcal{H}} = (V_n, E_n)$ be the (possibly infinite) one-inclusion graph. Then, by definition of $\mathfrak{D}_{\mathcal{U}}(\mathcal{H})$, for every finite subgraph $G = (V, E)$ of $G^{\mathcal{U}}_{\mathcal{H}}$ there exists an orientation $\mathcal{O}_E : E \to V$ such that every vertex in the subgraph has adversarial out-degree at most $\frac{n}{3}$: $\forall v \in V$, $\text{adv-outdeg}(v; \mathcal{O}_E) \leq \frac{n}{3}$.

We next invoke Lemma B.1 where $E_n$ represents our family of non-empty finite sets, and for each finite subset $E \subseteq E_n$, we let the orientation $\mathcal{O}_E$ (from above) represent the choice function. Then, Lemma B.1 implies that there exists an orientation $\mathcal{O} : E_n \to V_n$ of $G^{\mathcal{U}}_{\mathcal{H}}$ (i.e., an orientation of the entire global one-inclusion graph) with the following property: for each finite subset $A$ of $E_n$, there corresponds a finite set $E$ satisfying $A \subseteq E \subseteq E_n$ and $\mathcal{O}(e) = \mathcal{O}_E(e)$ for each $e \in A$. This implies that orientation $\mathcal{O}$ satisfies the property that $\forall v \in V_n$, $\text{adv-outdeg}(v; \mathcal{O}) \leq \frac{n}{3}$. Because, if not, then we can find a subgraph $G = (E, V)$ where $\mathcal{O}_E$ (from above) violates the adversarial out-degree upper bound of $\frac{n}{3}$ and that leads to a contradiction.

Now, we use orientation $\mathcal{O}$ of $G^{\mathcal{U}}_{\mathcal{H}}$ (which has adversarial out-degree at most $\frac{n}{3}$) to construct a learner $\mathbb{A}_{\mathcal{O}} : (\mathcal{X} \times \mathcal{Y})^{n-1} \times \mathcal{X} \to \mathcal{Y}$ as in Lemma 4.2. Then, Lemma 4.2 implies that

$$\mathcal{E}_{n-1}(\mathcal{H}, \mathcal{U}) \leq \mathcal{E}_{n-1}(\mathbb{A}_{\mathcal{O}}; \mathcal{H}, \mathcal{U}) \leq \frac{1}{3}.$$

We now turn to the lower bound. Let $2 \leq n \leq \frac{\mathfrak{D}_{\mathcal{U}}(\mathcal{H})}{2}$, $\varepsilon \in (0, 1)$, and let $G^{\mathcal{U}}_{\mathcal{H}} = (V_{2n}, E_{2n})$ be the (possibly infinite) one-inclusion graph. Since $2n \leq \mathfrak{D}_{\mathcal{U}}(\mathcal{H})$, by definition of $\mathfrak{D}_{\mathcal{U}}(\mathcal{H})$, it follows that there exists a finite subgraph $G = (V, E)$ of $G^{\mathcal{U}}_{\mathcal{H}} = (V_{2n}, E_{2n})$ such that

$$\forall \text{ orientations } \mathcal{O} : E \to V \text{ of subgraph } G, \max_{v \in V} \text{adv-outdeg}(v; \mathcal{O}) \geq \frac{2n}{3}. \qquad \text{(B.1)}$$

Now, let $\mathbb{A} : (\mathcal{X} \times \mathcal{Y})^* \to \mathcal{Y}^{\mathcal{X}}$ be an arbitrary learner. We invoke Lemma B.2, which is a refined statement of Lemma 4.3 that takes $\varepsilon$ into account, to orient the subgraph $G$ using learner $\mathbb{A}$. Lemma B.2 and Equation B.1 above imply that

$$\mathcal{E}_{\frac{n}{\varepsilon}}(\mathcal{H},\mathcal{U}) \geq \mathcal{E}_{\frac{n}{\varepsilon}}(\mathbb{A};\mathcal{H},\mathcal{U}) \geq \frac{\varepsilon}{6}\frac{\max_{v\in V}\text{adv-outdeg}(v;\mathcal{O}_{\mathbb{A}})-1}{2n-1} \geq \frac{\varepsilon}{6}\frac{(2n)/3-1}{2n-1}$$
$$= \frac{\varepsilon}{18}\frac{2n-3}{2n-1} = \frac{\varepsilon}{18}\left(1-\frac{2}{2n-1}\right) \geq \frac{\varepsilon}{54},$$

for $n \geq 2$. □

We are now ready to proceed with the proof of Theorem 4.4.

*Proof of Theorem 4.4.* We begin with proving the upper bound. Let $m_0 = \mathfrak{D}_{\mathcal{U}}(\mathcal{H})$. By Lemma 4.4, there exists a learner $\mathbb{A}$ from orienting the global one-inclusion graph $G^{\mathcal{U}}_{\mathcal{H}} = (V_{m_0+1}, E_{m_0+1})$ that satisfies worst-case expected risk

$$\mathcal{E}_{m_0}(\mathbb{A};\mathcal{H},\mathcal{U}) \leq \frac{1}{3}. \tag{B.2}$$

Let $\mathcal{D} \in \text{RE}(\mathcal{H},\mathcal{U})$ be some unknown *robustly realizable* distribution. Fix $\varepsilon, \delta \in (0,1)$ and a sample size $m(\varepsilon,\delta)$ that will be determined later. Let $S = \{(x_1,y_1),\dots,(x_m,y_m)\}$ be an i.i.d. sample from $\mathcal{D}$. Our strategy is to use $\mathbb{A}$ above as a *weak* robust learner and boost its confidence and robust error guarantee.

Weak Robust Learner. Observe that, by Equation B.2, for any empirical distribution $D$ over $S$, $\mathbb{E}_{S'\sim D^{m_0}} \mathrm{R}_{\mathcal{U}}(\mathbb{A}(S');D) \leq 1/3$. This implies that for any empirical distribution $D$ over $S$, there exists at least one sequence $S_D \in (S)^{m_0}$ such that $h_D := \mathbb{A}(S_D)$ satisfies $\mathrm{R}_{\mathcal{U}}(h_D;D) \leq 1/3$. We use this to define a weak robust-learner for distributions $D$ over $S$: i.e., for any $D$, the weak learner chooses $h_D$ as its weak hypothesis.

Boosting. Now we run the $\alpha$-Boost boosting algorithm (Schapire and Freund, 2012, Section 6.4.2) on data set $S$, but using the robust loss rather than 0-1 loss. That is, we start with $D_1$ uniform on $S$. Then for each round $t$, we get $h_{D_t}$ as a weak robust classifier with respect to

$D_t$, and for each $(x, y) \in S$ we define a distribution $D_{t+1}$ over $S$ satisfying

$$D_{t+1}(\{(x,y)\}) = \frac{D_t(\{(x,y)\})}{Z_t} \times \begin{cases} e^{-2\alpha} & \text{if } 1[\forall z \in \mathcal{U}(x) : h_{D_t}(z) = y] = 1 \\ 1 & \text{otherwise} \end{cases}$$

where $Z_t$ is a normalization factor, $\alpha$ is a parameter that will be determined below. Following the argument from (Schapire and Freund, 2012, Section 6.4.2), after $T$ rounds we are guaranteed

$$\min_{(x,y)\in S} \frac{1}{T}\sum_{t=1}^{T} 1[\forall z \in \mathcal{U}(x) : h_{D_t}(z) = y] \geq \frac{2}{3} - \frac{2}{3}\alpha - \frac{\ln(|S|)}{2\alpha T},$$

so we will plan on running until round $T = 1 + 48\ln(|S|)$ with value $\alpha = 1/8$ to guarantee

$$\min_{(x,y)\in S} \frac{1}{T}\sum_{t=1}^{T} 1[\forall z \in \mathcal{U}(x) : h_{D_t}(z) = y] > \frac{1}{2},$$

so that the majority-vote classifier $\mathrm{MAJ}(h_{D_1}, \ldots, h_{D_T})$ achieves *zero* robust loss on the empirical dataset $S$, $\mathrm{R}_{\mathcal{U}}(\mathrm{MAJ}(h_{D_1}, \ldots, h_{D_T}); S) = 0$.

Furthermore, note that, since each $h_{D_t}$ is given by $\mathbb{A}(S_{D_t})$, where $S_{D_t}$ is an $m_0$-tuple of points in $S$, the classifier $\mathrm{MAJ}(h_{D_1}, \ldots, h_{D_T})$ is specified by an ordered sequence of $m_0 \cdot T$ points from $S$. Thus, the classifier $\mathrm{MAJ}(h_1, \ldots, h_T)$ is representable as the value of an (order-dependent) reconstruction function $\varphi$ with a compression set size $m_0 T = m_0 O(\log m)$. Now, invoking Lemma A.1, we get the following *robust* generalization guarantee: with probability at least $1 - \delta$ over $S \sim \mathcal{D}^m$,

$$\mathrm{R}_{\mathcal{U}}(\mathrm{MAJ}(h_1, \ldots, h_T); \mathcal{D}) \leq O\left(\frac{m_0 \log^2 m}{m} + \frac{\log(1/\delta)}{m}\right),$$

and setting this less than $\varepsilon$ and solving for a sufficient size of $m$ yields the stated sample complexity bound.

We now turn to proving the lower bound. Let $n_0 = \frac{\mathfrak{D}_{\mathcal{U}}(\mathcal{H})}{2}$, by invoking Lemma 4.4, we get that $\mathcal{E}_{n_0/\varepsilon} \geq \Omega(\varepsilon)$. Then, by Equation 4.3, this implies that $\mathcal{M}_\varepsilon(\mathcal{H}, \mathcal{U}) \geq \Omega(1/\varepsilon)n_0 \geq$

$\Omega(1/\varepsilon)\mathfrak{D}_{\mathcal{U}}(\mathcal{H})$. □

## B.3 Agnostic Robust Learnability (Proofs for Section 4.5.3)

**Definition B.3** (Agnostic Robust PAC Learnability). *For any* $\varepsilon, \delta \in (0,1)$, *the* sample complexity of agnostic robust $(\varepsilon, \delta)$−PAC learning of $\mathcal{H}$ with respect to perturbation set $\mathcal{U}$, *denoted* $\mathcal{M}^{\mathrm{ag}}_{\varepsilon,\delta}(\mathcal{H},\mathcal{U})$, *is defined as the smallest* $m \in \mathbb{N} \cup \{0\}$ *for which there exists a learner* $\mathbb{A} : (\mathcal{X} \times \mathcal{Y})^* \to \mathcal{Y}^{\mathcal{X}}$ *such that, for every data distribution* $\mathcal{D}$ *over* $\mathcal{X} \times \mathcal{Y}$, *with probability at least* $1-\delta$ *over* $S \sim \mathcal{D}^m$,

$$\mathrm{R}_{\mathcal{U}}(\mathbb{A}(S); \mathcal{D}) \le \inf_{h\in\mathcal{H}} \mathrm{R}_{\mathcal{U}}(h; \mathcal{D}) + \varepsilon.$$

*If no such* $m$ *exists, define* $\mathcal{M}^{\mathrm{ag}}_{\varepsilon,\delta}(\mathcal{H},\mathcal{U}) = \infty$. *We say that* $\mathcal{H}$ *is robustly PAC learnable in the agnostic setting with respect to perturbation set* $\mathcal{U}$ *if* $\forall \varepsilon, \delta \in (0,1)$, $\mathcal{M}^{\mathrm{ag}}_{\varepsilon,\delta}(\mathcal{H},\mathcal{U})$ *is finite.*

*Proof of Lemma 4.5.* The argument follows closely a proof of an analogous result by (David, Moran, and Yehudayoff, 2016) for non-robust learning, and (Montasser, Hanneke, and Srebro, 2019) for robust learning. Denote by $\mathbb{A}$ a realizable learner with sample complexity $\mathcal{M}^{\mathrm{re}}_{1/3}(\mathcal{H},\mathcal{U})$, and denote $m_0 = \mathcal{M}^{\mathrm{re}}_{1/3}(\mathcal{H},\mathcal{U})$.

Description of agnostic learner $\mathbb{B}$. Given a data set $S \sim \mathcal{D}^m$ where $\mathcal{D}$ is some unknown distribution, we first do robust-ERM to find a maximal-size subsequence $S'$ of the data where the robust loss can be zero: that is, $\inf_{h\in\mathcal{H}} \hat{\mathrm{R}}_{\mathcal{U}}(h; S') = 0$. Then for any distribution $D$ over $S'$, there exists a sequence $S_D \in (S')^{m_0}$ such that $h_D := \mathbb{A}(S_D)$ has $\mathrm{R}_{\mathcal{U}}(h_D; D) \le 1/3$; this follows since, by definition of $\mathcal{M}^{\mathrm{re}}_{1/3}(\mathcal{H},\mathcal{U})$, $\mathcal{E}_{m_0}(\mathbb{A}; \mathcal{H},\mathcal{U}) \le 1/3$ so at least one such $S_D$ exists. We use this to define a weak robust-learner for distributions $D$ over $S'$: i.e., for any $D$, the weak learner chooses $h_D$ as its weak hypothesis.

Now we run the $\alpha$-Boost boosting algorithm (Schapire and Freund, 2012, Section 6.4.2) on data set $S'$, but using the robust loss rather than 0-1 loss. That is, we start with $D_1$ uniform on $S'$.* Then for each round $t$, we get $h_{D_t}$ as a weak robust classifier with respect to $D_t$, and

*We ignore the possibility of repeats; for our purposes we can just remove any repeats from $S'$ before this boosting step.

for each $(x,y) \in S'$ we define a distribution $D_{t+1}$ over $S'$ satisfying

$$D_{t+1}(\{(x,y)\}) \propto D_t(\{(x,y)\}) \exp\{-2\alpha 1[\forall x' \in \mathcal{U}(x), h_{D_t}(x') = y]\},$$

where $\alpha$ is a parameter we can set. Following the argument from (Schapire and Freund, 2012, Section 6.4.2), after $T$ rounds we are guaranteed

$$\min_{(x,y)\in S'} \frac{1}{T}\sum_{t=1}^{T} 1[\forall x' \in \mathcal{U}(x), h_{D_t}(x') = y] \geq \frac{2}{3} - \frac{2}{3}\alpha - \frac{\ln(|S'|)}{2\alpha T},$$

so we will plan on running until round $T = 1 + 48\ln(|S'|)$ with value $\alpha = 1/8$ to guarantee

$$\min_{(x,y)\in S'} \frac{1}{T}\sum_{t=1}^{T} 1[\forall x' \in \mathcal{U}(x), h_{D_t}(x') = y] > \frac{1}{2},$$

so that the classifier $\hat{h}(x) := 1\left[\frac{1}{T}\sum_{t=1}^{T} h_{D_t}(x) \geq \frac{1}{2}\right]$ has $\hat{\mathrm{R}}_{\mathcal{U}}(\hat{h}; S') = 0$.

Furthermore, note that, since each $h_{D_t}$ is given by $\mathbb{A}(S_{D_t})$, where $S_{D_t}$ is an $m_0$-tuple of points in $S'$, the classifier $\hat{h}$ is specified by an ordered sequence of $m_0 T$ points from $S$. Altogether, $\hat{h}$ is a function specified by an ordered sequence of $m_0 T$ points from $S$, and which has

$$\hat{\mathrm{R}}_{\mathcal{U}}(\hat{h}; S) \leq \min_{h\in\mathcal{H}} \hat{\mathrm{R}}_{\mathcal{U}}(h; S).$$

Similarly to the realizable case (see the proof of Lemma A.1), uniform convergence guarantees for sample compression schemes (see Graepel, Herbrich, and Shawe-Taylor, 2005) remain valid for the robust loss, by essentially the same argument; the essential argument is the same as in the proof of Lemma A.1 except using Hoeffding's inequality to get concentration of the empirical robust risks for each fixed index sequence, and then a union bound over the possible index sequences as before. We omit the details for brevity. In particular, denoting $T_m = 1 + 48\ln(m)$, for $m > m_0 T_m$, with probability at least $1 - \delta/2$,

$$\mathrm{R}_{\mathcal{U}}(\hat{h}; \mathcal{D}) \leq \hat{\mathrm{R}}_{\mathcal{U}}(\hat{h}; S) + \sqrt{\frac{m_0 T_m \ln(m) + \ln(2/\delta)}{2m - 2m_0 T_m}}.$$

Let $h^* = \mathrm{argmin}_{h\in\mathcal{H}} \mathrm{R}_{\mathcal{U}}(h; \mathcal{D})$ (supposing the min is realized, for simplicity; else we could take an $h^*$ with very-nearly minimal risk). By Hoeffding's inequality, with probability at least $1 - \delta/2$,

$$\hat{\mathrm{R}}_{\mathcal{U}}(h^*; S) \leq \mathrm{R}_{\mathcal{U}}(h^*; \mathcal{D}) + \sqrt{\frac{\ln(2/\delta)}{2m}}.$$

By the union bound, if $m \geq 2m_0 T_m$, with probability at least $1 - \delta$,

$$\begin{aligned}
\mathrm{R}_{\mathcal{U}}(\hat{h}; \mathcal{D}) &\leq \min_{h \in \mathcal{H}} \hat{\mathrm{R}}_{\mathcal{U}}(h; S) + \sqrt{\frac{m_0 T_m \ln(m) + \ln(2/\delta)}{m}} \\
&\leq \hat{\mathrm{R}}_{\mathcal{U}}(h^*; S) + \sqrt{\frac{m_0 T_m \ln(m) + \ln(2/\delta)}{m}} \\
&\leq \mathrm{R}_{\mathcal{U}}(h^*; \mathcal{D}) + 2\sqrt{\frac{m_0 T_m \ln(m) + \ln(2/\delta)}{m}}.
\end{aligned}$$

Since $T_m = O(\log(m))$, the above is at most $\varepsilon$ for an appropriate choice of sample size $m = O\big(\frac{m_0}{\varepsilon^2} \log^2\big(\frac{m_0}{\varepsilon}\big) + \frac{1}{\varepsilon^2} \log\big(\frac{1}{\delta}\big)\big)$. This concludes the upper bound on $\mathcal{M}^{\mathrm{ag}}_{\varepsilon,\delta}(\mathbb{B}; \mathcal{H}, \mathcal{U})$, and the lower bound trivially holds from the definition of $\mathbb{B}$. □

## B.4 Finite Character Property

Ben-David, Hrubes, Moran, Shpilka, and Yehudayoff (2019) gave a *formal* definition of the notion of "dimension" or "complexity measure", that all previously proposed dimensions in statistical learning theory comply with. In addition to characterizing learnability, a dimension should satisfy the *finite character* property:

**Definition B.4** (Finite Character). *A dimension characterizing learnability can be abstracted as a function $F$ that maps a class $\mathcal{H}$ to $\mathbb{N} \cup \{\infty\}$ and satisfies the* finite character *property: For every $d \in \mathbb{N}$ and $\mathcal{H}$, the statement "$F(\mathcal{H}) \geq d$" can be demonstrated by a finite set $X \subseteq \mathcal{X}$ of domain points, and a finite set of hypotheses $H \subseteq \mathcal{H}$. That is, "$F(\mathcal{H}) \geq d$" is equivalent to the existence of a bounded first order formula $\varphi(\mathcal{X}, \mathcal{H})$ in which all the quantifiers are of the form: $\exists x \in \mathcal{X}, \forall x \in \mathcal{X}$ or $\exists h \in \mathcal{H}, \forall h \in \mathcal{H}$.*

For example, the property "$\mathrm{vc}(\mathcal{H}) \geq d$" is a finite character property since it can be verified with a finite set of points $x_1 \dots, x_d \in \mathcal{X}$ and a finite set of classifiers $h_1, \dots, h_{2^d} \in \mathcal{H}$ that shatter these points, and a predicate $E(x, h) \equiv x \in h$ (i.e., the value $h(x)$). In our case, in addition to having a domain $\mathcal{X}$ and a hypothesis class $\mathcal{H}$, we also have a relation $\mathcal{U}$. In Claim B.5, we argue that our dimension $\mathfrak{D}_{\mathcal{U}}(\mathcal{H})$ satisfies Definition B.4, though unlike VC dimension, we do need $\forall$ quantifiers. Furthermore, we provably *cannot* verify the statement $\mathfrak{D}_{\mathcal{U}}(\mathcal{H}) \geq d$ by evaluating the predicate $E(x, h)$ on finitely many $x$'s and $h$'s, but we can verify it using a predicate $P_{\mathcal{U}}(x, h) \equiv \forall z \in \mathcal{U}(x) : h(z) = h(x)$ that evaluates the *robust* behavior of $h$ on $x$ w.r.t. $\mathcal{U}$. The proof is deferred to Appendix B.4.

**Claim B.5.** $\mathfrak{D}_{\mathcal{U}}(\mathcal{H})$ *satisfies the finite character property of Definition B.4.*

*Proof.* By the definition of $\mathfrak{D}_{\mathcal{U}}(\mathcal{H})$ in Equation 4.11, to demonstrate that $\mathfrak{D}_{\mathcal{U}}(\mathcal{H}) \geq d$, it suffices to present a finite subgraph $G = (V, E)$ of $G^{\mathcal{U}}_{\mathcal{H}} = (V_d, E_d)$ where every orientation $\mathcal{O} : E \to V$ has adversarial out-degree at least $\frac{n}{3}$. Since $V$ is, by definition, a finite collection of datasets robustly realizable with respect to $(\mathcal{H}, \mathcal{U})$ this means that we can demonstrate that $\mathfrak{D}_{\mathcal{U}}(\mathcal{H}) \geq d$ with a finite set $X \subseteq \mathcal{X}$ and a finite set of hypotheses $H \subseteq \mathcal{H}$ that can construct the finite collection $V$.

Note that in our case, we do not only have $\mathcal{X}$ and $\mathcal{H}$, but also a set relation $\mathcal{U}$ that specifies for each $x \in \mathcal{X}$ its corresponding set of perturbations $\mathcal{U}(x)$. We can still express $\mathfrak{D}_{\mathcal{U}}(\mathcal{H}) \geq d$ with a bounded formula using only quantifiers over $\mathcal{H}$ and $\mathcal{X}$, though unlike in the case of VC dimension, we do also need $\forall$ quantifiers. Furthermore, we provably cannot verify the formula by evaluating $h(x)$ only on finitely many $x \in \mathcal{X}, h \in \mathcal{H}$, since $\mathcal{U}(x)$ can be *infinite*. But, we can verify it given access to a predicate $P_{\mathcal{U}}(h, x) \equiv \forall z \in \mathcal{U}(x) : h(z) = h(x)$. □

# C

# Boosting Barely Robust Learners

## C.1 Boosting a Weakly Robust Learner to a Strongly Robust Learner

**Algorithm C.1:** $\alpha$-Boost — Boosting *weakly* robust learners

**Input:** Training dataset $S = \{(x_1, y_1), \ldots, (x_m, y_m)\}$, black-box *weak* robust learner $\mathbb{B}$.

1 Set $m_0 = m_{\mathbb{B}}(1/3, 1/3)$.
2 Initialize $D_1$ to be uniform over $S$, and set $T = O(\log m)$.
3 **for** $1 \leq t \leq T$ **do**
4 Sample $S_t \sim D_t^{m_0}$, call learner $\mathbb{B}$ on $S_t$, and denote by $h_t$ its output predictor. Repeat this step until $\mathrm{R}_{\mathcal{U}}(h_t; D_t) \leq 1/3$.
5 Compute a new distribution $D_{t+1}$ by applying the following update for each $(x, y) \in S$:

$$D_{t+1}(\{(x,y)\}) = \frac{D_t(\{(x,y)\})}{Z_t} \times \begin{cases} e^{-2\alpha}, & \text{if } \mathbb{1}[\forall z \in \mathcal{U}(x) : h_{t-1}(z) = y] = 1; \\ 1, & \text{otherwise,} \end{cases}$$

where $Z_t$ is a normalization factor and $\alpha = 1/8$.

**Output:** A majority-vote classifier $\mathrm{MAJ}(h_1, \ldots, h_T)$.

*Proof of Lemma 5.6.* Let $\mathbb{B}$ be a *weak* robust learner with fixed parameters $(\varepsilon_0, \delta_0) = (1/3, 1/3)$ for some *unknown* target concept $c$ with respect to $\mathcal{U}$. Let $D$ be some unknown distribution over $\mathcal{X}$ such that $\Pr_{x\sim D}[\exists z \in \mathcal{U}(x) : c(z) \neq c(x)] = 0$. By Definition 5.2, with fixed sample complexity $m_0 = m_{\mathbb{B}}(1/3, 1/3)$, for any distribution $\tilde{D}$ over $\mathcal{X}$ such that $\Pr_{x\sim\tilde{D}}[\exists z \in \mathcal{U}(x) : c(z) \neq c(x)] = 0$, with probability at least $1/3$ over $S \sim \tilde{D}_c^{m_0}$, $\mathrm{R}_{\mathcal{U}}(\mathbb{B}(S); \tilde{D}_c) \leq 1/3$.

We will now boost the confidence and robust error guarantee of the *weak* robust learner $\mathbb{B}$ by running boosting with respect to the *robust* loss (rather than the standard 0-1 loss). Specifically, fix $(\varepsilon, \delta) \in (0, 1)$ and a sample size $m(\varepsilon, \delta)$ that will be determined later. Let $S = \{(x_1, y_1), \ldots, (x_m, y_m)\}$ be an i.i.d. sample from $D_c$. Run the $\alpha$-Boost algorithm on dataset $S$ using $\mathbb{B}$ as the weak robust learner for a number of rounds $T$ that will be determined below. On each round $t$, $\alpha$-Boost computes an empirical distribution $D_t$ over $S$ by applying the following update for each $(x, y) \in S$:

$$D_t(\{(x,y)\}) = \frac{D_{t-1}(\{(x,y)\})}{Z_{t-1}} \times \begin{cases} e^{-2\alpha}, & \text{if } 1[\forall z \in \mathcal{U}(x) : h_{t-1}(z) = y] = 1; \\ 1, & \text{otherwise,} \end{cases}$$

where $Z_{t-1}$ is a normalization factor, $\alpha$ is a parameter that will be determined below, and $h_{t-1}$ is the *weak* robust predictor outputted by $\mathbb{B}$ on round $t-1$ that satisfies $\mathrm{R}_{\mathcal{U}}(h_{t-1}; D_{t-1}) \leq 1/3$. Once $D_t$ is computed, we sample $m_0$ examples from $D_t$ and run *weak* robust learner $\mathbb{B}$ on these examples to produce a hypothesis $h_t$ with robust error guarantee $\mathrm{R}_{\mathcal{U}}(h_t; D_t) \leq 1/3$. This step has failure probability at most $\delta_0 = 1/3$. We will repeat it for at most $\lceil \log(2T/\delta) \rceil$ times, until $\mathbb{B}$ succeeds in finding $h_t$ with robust error guarantee $\mathrm{R}_{\mathcal{U}}(h_t; D_t) \leq 1/3$. By a union bound argument, we are guaranteed that with probability at least $1 - \delta/2$, for each $1 \leq t \leq T$, $\mathrm{R}_{\mathcal{U}}(h_t; D_t) \leq 1/3$. Following the argument from Schapire and Freund (2012, Section 6.4.2), after $T$ rounds we are guaranteed

$$\min_{(x,y)\in S} \frac{1}{T} \sum_{t=1}^{T} 1[\forall z \in \mathcal{U}(x) : h_t(z) = y] \geq \frac{2}{3} - \frac{2}{3}\alpha - \frac{\ln(|S|)}{2\alpha T},$$

so we will plan on running until round $T = 1 + 48\ln(|S|)$ with value $\alpha = 1/8$ to guarantee

$$\min_{(x,y)\in S} \frac{1}{T}\sum_{t=1}^{T} 1[\forall z \in \mathcal{U}(x) : h_t(z) = y] > \frac{1}{2},$$

so that the majority-vote classifier $\mathrm{MAJ}(h_1, \ldots, h_T)$ achieves *zero* robust loss on the empirical dataset $S$, $\mathrm{R}_{\mathcal{U}}(\mathrm{MAJ}(h_1, \ldots, h_L); S) = 0$.

Note that each of these classifiers $h_t$ is equal to $\mathbb{B}(S'_t)$ for some $S'_t \subseteq S$ with $|S'_t| = m_0$. Thus, the classifier $\mathrm{MAJ}(h_1, \ldots, h_T)$ is representable as the value of an (order-dependent) reconstruction function $\varphi$ with a compression set size $m_0 T = m_0 O(\log m)$. Now, invoking Lemma A.1, with probability at least $1 - \delta/2$,

$$\mathrm{R}_{\mathcal{U}}(\mathrm{MAJ}(h_1, \ldots, h_T); \mathcal{D}) \leq O\left(\frac{m_0 \log^2 m}{m} + \frac{\log(2/\delta)}{m}\right),$$

and setting this less than $\varepsilon$ and solving for a sufficient size of $m$ yields the stated sample complexity bound. □

### C.2 Robustness at Different Levels of Granularity

For concreteness, throughout the rest of this section, we consider robustness with respect to metric balls $\mathrm{B}_\gamma(x) = \{z \in \mathcal{X} : \rho(x, z) \leq \gamma\}$ where $\rho$ is some metric on $\mathcal{X}$ (e.g., $\ell_\infty$ metric), and $\gamma > 0$ is the perturbation radius. Achieving small robust risk with respect to a *fixed* perturbation set $\mathrm{B}_\gamma$ is the common goal studied in adversarially robust learning. What we studied so far in this work is learning a predictor $\hat{h}$ robust to $\mathcal{U} = \mathrm{B}_\gamma$ perturbations as measured by the robust risk: $\Pr_{(x,y)\sim D}\left[\exists z \in \mathrm{B}_\gamma(x) : \hat{h}(z) \neq y\right]$, when given access to a learner $\mathbb{A}$ barely robust with respect to $\mathcal{U}^{-1}(\mathcal{U}) = \mathrm{B}_{2\gamma}$.

Our original approach to boosting robustness naturally leads us to an alternate interesting idea: learning a cascade of robust predictors with different levels of granularity. This might be desirable in situations where it is difficult to robustly learn a distribution $D_c$ over $\mathcal{X} \times \mathcal{Y}$ with robustness granularity $\gamma$ everywhere, and thus, we settle for a weaker goal which is first learning a robust predictor $h_1$ with granularity $\gamma$ on say $\beta$ mass of $D$, and then recursing on the conditional distribution of $D$ where $h_1$ is not $\gamma$-robust and learning a robust predictor $h_2$ with granularity $\gamma/2$, and so on. That is, we are adaptively learning a sequence of predictors $h_1, \ldots, h_T$ where each predictor $h_t$ is robust with granularity $\frac{\gamma}{2^t}$. Furthermore, if we are guaranteed that in each round we make progress on some $\beta$ mass then it follows that

$$\Pr\left[\cup_{t=1}^{T} \mathrm{Rob}_{\gamma/2^{t-1}}(h_t)\right] = 1 - \Pr\left[\cap_{t=1}^{T} \overline{\mathrm{Rob}}_{\gamma/2^{t-1}}(h_t)\right] \geq 1 - (1-\beta)^T,$$

and the cascade predictor $\mathrm{CAS}(h_{1:T})$ has the following robust risk guarantee

$$\sum_{t=1}^{T} \Pr\left[\bar{R}_{1:t-1} \wedge \left(\exists z \in \mathrm{B}_{\gamma/2^t}(x) : \forall_{t'<t} G_{h_{t'}}(z) = \perp \wedge G_{h_t}(z) = 1 - c(x)\right)\right] + \Pr[\bar{R}_{1:T}] \leq \frac{\varepsilon}{\beta} + (1-\beta)^T.$$

In words, the cascade predictor $\mathrm{CAS}(h_{1:T})$ offers robustness at different granularities. That is, for $x \sim D$ such that $x \in R_1$, $\mathrm{CAS}(h_{1:T})$ is guaranteed to be robust on $x$ with granularity $\gamma$, and for $x \sim D$ such that $x \in \bar{R}_1 \cap R_2$, $\mathrm{CAS}(h_{1:T})$ is guaranteed to be robust on $x$ with granularity $\gamma/2$, and so on.

**Applications.** We give a few examples where this can be useful. Consider using SVMs as barely robust learners. SVMs are known to be margin maximizing learning algorithms, which is equivalent to learning linear predictors robust to $\ell_2$ perturbations. In our context, by combining SVMs with our boosting algorithm, we can learn a cascade of linear predictors each with a maximal margin on the conditional distribution.

## C.3 Python Code

```python
# MNIST odd vs. even
from torchvision import datasets, transforms
from sklearn.preprocessing import StandardScaler
from sklearn.pipeline import make_pipeline
from sklearn.svm import LinearSVC

def flatten(X):
    return X.reshape(X.shape[0], -1)
```

```python
train_set = datasets.MNIST('./', train=True, download=True, transform =
    transforms.Compose([transforms.ToTensor()]))
test_set = datasets.MNIST('./', train=False, download=True, transform =
    transforms.Compose([transforms.ToTensor()]))

X_train = train_set.train_data.numpy()
y_train = train_set.train_labels.numpy()
X_test = test_set.test_data.numpy()
y_test = test_set.test_labels.numpy()
# Convert to binary labels
y_train_binary = 2*np.array([y % 2 for y in y_train])-1
y_test_binary = 2*np.array([y % 2 for y in y_test])-1

#flatten data
f_X_tr = flatten(X_train) / 255.0
f_X_te = flatten(X_test) / 255.0

# sample size, and robustness radius
m, gamma, T = 10000, 1.0, 2

# fit a linear classifier
print('===================================================')
print('Baseline Linear Classifier:')
base_clf = LinearSVC(C=10e-7, fit_intercept=False, loss='hinge', tol=1e
    -5)
base_clf.fit(f_X_tr, y_train_binary)
print("0-1 Accuracy - Testing:", 100*base_clf.score(f_X_te, y_test_binary
    ))

# robust accuracy
y_margin = y_test_binary*(base_clf.decision_function(f_X_te) / np.linalg.
    norm(base_clf.coef_))
robust_accuracy = len(y_margin[y_margin >= gamma]) / len(y_margin)
print('Robust Accuracy - Test:', 100*robust_accuracy)
```

```
y_margin = y_train_binary*(base_clf.decision_function(f_X_tr) / np.linalg
    .norm(base_clf.coef_))
robust_accuracy = len(y_margin[y_margin >= gamma]) / len(y_margin)
print('===================================================')

# boosting rounds
print('\nBoosting Robustness')
clf = []
for t in range(1,T+1):
    print("round ", t)
    # SVM
    lin_clf = LinearSVC(C=10e-7, fit_intercept=False, loss='hinge', tol=1
        e-5)
    lin_clf.fit(f_X_tr, y_train_binary)
    clf.append(lin_clf)

    if t < T:
        # Compute margin on fresh sample
        margin = lin_clf.decision_function(f_X_tr) / np.linalg.norm(
            lin_clf.coef_)
        amargin = abs(margin)
        print("Fraction of Training Data with Robustness at least 2gamma:
             ", 100*float(len(amargin[amargin >= 2*gamma])/len(amargin)))
        print(len(amargin[amargin < 2*gamma]), " samples with small
            margin")

        # Update / Filter training data
        f_X_tr = f_X_tr[amargin < 2*gamma]
        y_train_binary = y_train_binary[amargin < 2*gamma]

# Evaluation
print('\nEvaluation - Test')
print('number of test samples', len(y_test_binary))
```

```
adv_mistake = 0
total = len(y_test_binary)
for t in range(0,T):
    print('round ', t)
    amargin = abs(clf[t].decision_function(f_X_te) / np.linalg.norm(clf[t
        ].coef_))
    y_margin = y_test_binary*(clf[t].decision_function(f_X_te) / np.
        linalg.norm(clf[t].coef_))
    adv_mistake += len(y_test_binary[y_margin < -1*gamma])
    print('adversarial mistakes', len(y_test_binary[y_margin < -1*gamma])
        )

    f_X_te = f_X_te[(y_margin < 2*gamma) & (y_margin >= -1*gamma)]
    y_test_binary = y_test_binary[(y_margin < 2*gamma) & (y_margin >= -1*
        gamma)]
    print('abstained on data', len(y_test_binary), len(f_X_te))

    # last round
    if t == T-1 and len(f_X_te)>0:
        y_margin = y_test_binary*(clf[t].decision_function(f_X_te) / np.
            linalg.norm(clf[t].coef_))
        print('classifications in final round', len(y_margin))
        ee = len(y_test_binary[(y_margin < gamma)])
        print('extra mistakes', ee)
        adv_mistake += ee

print('total number of adv. mistakes', adv_mistake)
print('===================================================')
print('Cascade Robust Accuracy:', 100*float((total - adv_mistake)/total)
    )
print('===================================================')
```

# D

# Reductions to Non-Robust Learners

## D.1 Bounding the dual VC dimension of the convex-hull of a class $\mathcal{H}$

*Proof of Lemma 6.3.* Consider a *dual space* $\bar{\mathcal{G}}$: a set of functions $\bar{g}_x : \text{co}^k(\mathcal{H}) \to \mathcal{Y}$ defined as $\bar{g}_x(f) = f(x)$ for each $f = \text{MAJ}(h_1, \dots, h_k) \in \text{co}^k(\mathcal{H})$ and each $x \in \mathcal{X}$. It follows by definition of dual VC dimension that $\text{vc}(\bar{\mathcal{G}}) = \text{vc}^*(\text{co}^k(\mathcal{H}))$. Similarly, define another dual space $\mathcal{G}$: a set of functions $g : \mathcal{H} \to \mathcal{Y}$ defined as $g(x) = h(x)$ for each $h \in \mathcal{H}$ and each $x \in \mathcal{X}$. We know that $\text{vc}(\mathcal{G}) = \text{vc}^*(\mathcal{H}) = d^*$. Observe that by definition of $\mathcal{G}$ and $\bar{\mathcal{G}}$, we have that for each $x \in \mathcal{X}$ and each $f = \text{MAJ}(h_1, \dots, h_k) \in \text{co}^k(\mathcal{H})$,

$$\bar{g}_x(f) = f(x) = \text{MAJ}(h_1, \dots, h_k)(x) = \text{sign}\left(\sum_{i=1}^{k} h_i(x)\right) = \text{sign}\left(\sum_{i=1}^{k} g_x(h_i)\right).$$

By the Sauer-Shelah Lemma applied to dual class $\mathcal{G}$, for any set $H = \{h_1, \dots, h_n\} \subseteq \mathcal{H}$, the number of possible behaviors

$$|\mathcal{G}|_H| := |\{(g_x(h_1), \dots, g_x(h_n)) : x \in \mathcal{X}\}| \leq \binom{n}{\leq d^*}. \tag{D.1}$$

Consider a set $F = \{f_1, \dots, f_m\} \subseteq \text{co}^k(\mathcal{H})$, the number of possible behaviors can be upper-

bounded as follows:

$$
\begin{aligned}
\left|\bar{\mathcal{G}}|_F\right| &= \left|\{(\bar{g}_x(f_1), \ldots, \bar{g}_x(f_m)) : x \in \mathcal{X}\}\right| \\
&= \left|\left\{(\bar{g}_x(\mathrm{MAJ}(h_1^1, \ldots, h_1^k)), \ldots, \bar{g}_x(\mathrm{MAJ}(h_m^1, \ldots, h_m^k))) : x \in \mathcal{X}\right\}\right| \\
&= \left|\left\{\left(\mathrm{sign}\left(\textstyle\sum_{i=1}^k g_x(h_i)\right), \ldots, \mathrm{sign}\left(\textstyle\sum_{i=1}^k g_x(h_i)\right)\right) : x \in \mathcal{X}\right\}\right| \\
&\overset{(i)}{\leq} \left|\left\{(g_x(h_1^1), \ldots, g_x(h_1^k), g_x(h_2^1), \ldots, g_x(h_2^k), \ldots, g_x(h_m^1), \ldots, g_x(h_m^k)) : x \in \mathcal{X}\right\}\right| \\
&\overset{(ii)}{\leq} \binom{mk}{\leq d^*},
\end{aligned}
$$

where $(i)$ follows from observing that each expanded vector $(g_x(h_i^1), \ldots, g_x(h_i^k))_{i=1}^m \in \mathcal{Y}^{mk}$ can map to at most one vector $\left(\mathrm{sign}\left(\sum_{i=1}^k g_x(h_i)\right), \ldots, \mathrm{sign}\left(\sum_{i=1}^k g_x(h_i)\right)\right) \in \mathcal{Y}^m$, and $(ii)$ follows from Equation D.1. Observe that if $\left|\bar{\mathcal{G}}|_F\right| < 2^m$, then by definition, $F$ is not shattered by $\bar{\mathcal{G}}$, and this implies that $\mathrm{vc}(\bar{\mathcal{G}}) < m$. Thus, to conclude the proof, we need to find the smallest $m$ such that $\binom{mk}{\leq d^*} < 2^m$. It suffices to check that $m = O(d^* \log k)$ satisfies this condition. $\square$

# E

# Robustness to Unknown Perturbations

## E.1 The Perfect Attack Oracle model (Proofs for Realizable Guarantees)

### E.1.1 Proof of Theorem 7.4

*Proof of Theorem 7.4.* Let $\mathcal{U}$ be an arbitrary adversary and $\mathsf{O}_{\mathcal{U}}$ its corresponding mistake oracle. Let $\mathcal{C} \subseteq \mathcal{Y}^{\mathcal{X}}$ be an arbitrary target class, and $\mathbb{A}$ an online learner for $\mathcal{C}$ with mistake bound $M_{\mathbb{A}} = \mathrm{lit}(\mathcal{H}) < \infty$. We assume w.l.o.g. that the online learner $\mathbb{A}$ is conservative, meaning that it does not update its state unless it makes a mistake. Algorithm E.1 in essence is a standard conversion of a learner in the mistake bound model to a learner in the PAC model (see e.g., Balcan, 2010):

**Algorithm E.1:** OnePassRobust

**Input:** $S = \{(x_1, y_1), \ldots, (x_m, y_m)\}$, $\varepsilon$, $\delta$, black-box access to a an online learner $\mathbb{A}$, black-box access to a mistake oracle $\mathsf{O}_{\mathcal{U}}$

1 Initialize $h_0 = \mathbb{A}(\emptyset)$.

2 **for** $i \leq m$ **do**

3 Certify the robustness of $h$ on $(x_i, y_i)$ by asking the mistake oracle $\mathsf{O}_{\mathcal{U}}$.

4 If $h_t$ is not robust on $(x_i, y_i)$, update $h_t$ by running $\mathbb{A}$ on $(z, y_i)$, where $z$ is the perturbation returned by $\mathsf{O}_{\mathcal{U}}$.

5 Break when $h_t$ is robustly correct on a consecutive sequence of length $\frac{1}{\varepsilon} \log\left(\frac{M_{\mathbb{A}}}{\delta}\right)$.

**Output:** $h_t$.

Analysis Let $\mathcal{D}$ be an arbitrary distribution over $\mathcal{X} \times \mathcal{Y}$ that is robustly realizable with some concept $c \in \mathcal{C}$,i.e., $\mathrm{R}_{\mathcal{U}}(c; \mathcal{D}) = 0$. Fix $\varepsilon, \delta \in (0, 1)$ and a sample size $m = 2\frac{M_{\mathbb{A}}}{\varepsilon} \log\left(\frac{M_{\mathbb{A}}}{\delta}\right)$.

Since online learner $\mathbb{A}$ has a mistake bound of $M_{\mathbb{A}}$, Algorithm E.1 will terminate in at most $\frac{M_{\mathbb{A}}}{\varepsilon} \log\left(\frac{M_{\mathbb{A}}}{\delta}\right)$ steps of certification, which of course is an upperbound on the number of calls to the mistake oracle $\mathsf{O}_{\mathcal{U}}$, and the number of calls to the online learner $\mathbb{A}$.

It remains to show that the output of Algorithm E.1, the final predictor $h$, has low robust risk $\mathrm{R}_{\mathcal{U}}(h; \mathcal{D})$. Throughout the runtime of Algorithm E.1, the online learner can generate a sequence of at most $M_{\mathbb{A}} + 1$ predictors. There's the initial predictor from Step 1, plus the $M_{\mathbb{A}}$ updated predictors corresponding to potential updates by online learner $\mathbb{A}$. Observe that the probability that the final $h$ has robust risk more than $\varepsilon$

$$\Pr_{S\sim\mathcal{D}^m}[\mathrm{R}_{\mathcal{U}}(h; \mathcal{D}) > \varepsilon] \leq \Pr_{S\sim\mathcal{D}^m}\left[\exists j \in [M_{\mathbb{A}}+1] \text{ s.t. } \mathrm{R}_{\mathcal{U}}(h_j; \mathcal{D}) > \varepsilon\right] \leq (M_{\mathbb{A}}+1)(1-\varepsilon)^{\frac{1}{\varepsilon}\log\left(\frac{M_{\mathbb{A}}+1}{\delta}\right)} \leq \delta.$$

Therefore, with probability at least $1 - \delta$ over $S \sim \mathcal{D}^m$, Algorithm E.1 outputs a predictor $h$ with robust risk $\mathrm{R}_{\mathcal{U}}(h; \mathcal{D}) \leq \varepsilon$. Thus, Algorithm E.1 robustly PAC learns $\mathcal{C}$ w.r.t. adversary $\mathcal{U}$. □

### E.1.2 Lemma and Proof of Theorem 7.5

**Algorithm E.2:** CycleRobust

**Input:** Training dataset $L = \{(x_1, y_1), \ldots, (x_m, y_m)\}$, black-box online learner $\mathbb{A}$ for $\mathcal{H}$, black-box perfect attack oracle $\mathsf{O}_{\mathcal{U}}$.

```
Initialize Z = {}, and initialize ĥ = A(Z).
Set FullRobustPass = False.
while FullRobustPass is False do
    Set FullRobustPass = True.
    for 1 ≤ i ≤ m do
        Certify the robustness of ĥ on (x_i, y_i) by sending the query (ĥ, (x_i, y_i)) to the
          perfect attack oracle O_U.
        if ĥ is not robustly correct on (x_i, y_i) then
            Let z be the perturbation returned by O_U where ĥ(z) ≠ y_i. Add (z, y_i) to
              the set Z.
            Update ĥ by running A on example (z, y_i), or equivalently, set ĥ = A(Z).
            Set FullRobustPass = False.
```

**Output:** Predictor $\hat{h}$.

**Lemma E.1** (Robust Generalization with Stable Sample Compression). *Let $(\kappa, \rho)$ be a stable sample compression scheme of size $k$ for $\mathcal{H}$ with respect to the robust loss $\sup_{z \in \mathcal{U}(x)} 1[h(z) \neq y]$. Then, for any distribution $\mathcal{D}$ over $\mathcal{X} \times \mathcal{Y}$ such that $\inf_{h \in \mathcal{H}} \mathrm{R}_{\mathcal{U}}(h; \mathcal{D}) = 0$, any integer $m > 2k$, and any $\delta \in (0, 1)$, with probability at least $1 - \delta$ over $S = \{(x_1, y_1), \ldots, (x_m, y_m)\}$ iid $\mathcal{D}$-distributed random variables,*

$$\mathrm{R}_{\mathcal{U}}(\rho(\kappa(S)); \mathcal{D}) \leq \frac{2}{m - 2k} \left( k \ln(4) + \ln\left(\frac{1}{\delta}\right)\right).$$

*Proof.* The argument follows an analogous proof from Bousquet et al. (2020) for the 0-1 loss. We observe that the same argument applies to the robust loss, and we provide an explicit proof for completeness. Split the $m$ samples of $S$ into $2k$ sets $S_1, \ldots, S_{2k}$ each of size $\frac{m}{2k}$. Observe that the $k$ compression points chosen by $\kappa$, $\kappa(S)$, are in at most $k$ of these sets $S_{i_1^*}, \ldots, S_{i_k^*}$ where $i_1^*, \ldots, i_k^* \in \{1, \ldots, 2k\}$. Stability of $(\kappa, \rho)$ implies that $\rho(\kappa(\cup_{j=1}^{k} S_{i_j^*})) =$ $\rho(\kappa(S))$. Since by definition of $(\kappa, \rho)$, the robust risk $\mathrm{R}_{\mathcal{U}}(\rho(\kappa(S)); S) = 0$, it follows that $\mathrm{R}_{\mathcal{U}}(\rho(\kappa(\cup_{j=1}^{k} S_{i_j^*}))); S) = 0$. This implies that $\rho(\kappa(\cup_{j=1}^{k} S_{i_j^*}))$ is robustly correct on the remaining sets $\cup_{j \notin \{i_1^*, \dots, i_k^*\}} S_j$.

Observe that the event that $\mathrm{R}_{\mathcal{U}}(\rho(\kappa(S)); \mathcal{D}) > \varepsilon$ implies the event that there exists $i_1, \dots, i_k \in \{1, \dots, 2k\}$ such that $\mathrm{R}_{\mathcal{U}}(\rho(\kappa(\cup_{j=1}^{k} S_{i_j})); \mathcal{D}) > \varepsilon$ and $\rho(\kappa(\cup_{j=1}^{k} S_{i_j}))$ robustly correct on $\cup_{j \notin \{i_1, \dots, i_k\}} S_j$. Thus,

$$\begin{aligned}
&\Pr_{S \sim \mathcal{D}^m} \left[\mathrm{R}_{\mathcal{U}}(\rho(\kappa(S)); \mathcal{D}) > \varepsilon\right] \\
&\leq \Pr_{S \sim \mathcal{D}^m} \left[\exists i_1, \dots, i_k : \mathrm{R}_{\mathcal{U}}(\rho(\kappa(\cup_{j=1}^{k} S_{i_j})); \mathcal{D}) > \varepsilon \wedge \mathrm{R}_{\mathcal{U}}(\rho(\kappa(\cup_{j=1}^{k} S_{i_j})); \cup_{j \notin \{i_1, \dots, i_k\}} S_j) = 0\right] \\
&\overset{(i)}{\leq} \binom{2k}{k} \Pr_{S \sim \mathcal{D}^m} \left[\mathrm{R}_{\mathcal{U}}(\rho(\kappa(\cup_{j=1}^{k} S_{i_j})); \mathcal{D}) > \varepsilon \wedge \mathrm{R}_{\mathcal{U}}(\rho(\kappa(\cup_{j=1}^{k} S_{i_j})); \cup_{j \notin \{i_1, \dots, i_k\}} S_j) = 0\right] \\
&\overset{(ii)}{\leq} \binom{2k}{k} (1 - \varepsilon)^{m/2} < 4^k e^{-\varepsilon m/2},
\end{aligned}$$

where inequality $(i)$ follows from a union bound, and inequality $(ii)$ follows from observing that the $\frac{m}{2}$ samples in $\cup_{j \notin \{i_1, \dots, i_k\}} S_j$ are independent of $\rho(\kappa(\cup_{j=1}^{k} S_{i_j})$. Setting $4^k e^{-\varepsilon m/2} = \delta$ and solving for $\varepsilon$ yields the stated bound. $\square$

*Proof of Theorem 7.5.* Let $\mathbb{A} : (\mathcal{X} \times \mathcal{Y})^* \to \mathcal{Y}^{\mathcal{X}}$ be a conservative online learner for $\mathcal{H}$ with mistake bound equal to $\mathrm{lit}(\mathcal{H})$. Let $\mathcal{U} : \mathcal{X} \to 2^{\mathcal{X}}$ be an arbitrary adversarial set that is unknown to the learning algorithm and $\mathsf{O}_{\mathcal{U}}$ a black-box perfect attack oracle for $\mathcal{U}$. Let $\mathcal{D}$ be an arbitrary distribution over $\mathcal{X} \times \mathcal{Y}$ that is robustly realizable with some concept $h^* \in \mathcal{H}$, i.e., $\mathrm{R}_{\mathcal{U}}(h^*; \mathcal{D}) = 0$. Fix $\varepsilon, \delta \in (0, 1)$ and a sample size $m$ that will be determined later. Let $S = \{(x_1, y_1), \dots, (x_m, y_m)\}$ be an iid sample from $\mathcal{D}$. Our proof will be divided into two main parts.

Zero Empirical Robust Loss Observe that the output of CycleRobust (Algorithm E.3): $\hat{h} = \mathbb{B}(S, \mathsf{O}_{\mathcal{U}}, \mathbb{A})$, achieves zero robust loss on the training data, $\mathrm{R}_{\mathcal{U}}(\hat{h}; S) = 0$. This follows because whenever CycleRobust (Algorithm E.3) terminates, Steps 4-11 imply that it made a full pass on dataset $S$ without encountering any example $(x_i, y_i)$ where predictor $\hat{h}$ is not robustly correct. Furthermore, since conservative online learner $\mathbb{A}$ has a finite mistake bound of $\mathrm{lit}(\mathcal{H})$, it implies that the number of full passes (execution of Step 3) Algorithm E.3 makes over $S$ is at most $\text{lit}(\mathcal{H})$, and in each pass $m$ oracle queries to $\mathsf{O}_{\mathcal{U}}$ are made. Thus, with at most $m\text{lit}(\mathcal{H})$ oracle queries to $\mathsf{O}_{\mathcal{U}}$, CycleRobust (Algorithm E.3) outputs a predictor $\hat{h}$ with zero robust loss on $S$, $\text{R}_{\mathcal{U}}(\hat{h}; S) = 0$.

Robust Generalization through Stable Sample Compression CycleRobust (Algorithm E.3) can be viewed as a stable compression scheme for the robust loss. Specifically, the output of the compression function $\kappa(S, \mathsf{O}_{\mathcal{U}}, \mathbb{A})$ is an order-dependent sequence that contains all examples $(x_i, y_i)$ on which $\hat{h}$ was not robustly correct while cycling through dataset $S$ (Steps 6-7), since $\mathbb{A}$ has a finite mistake bound of $\text{lit}(\mathcal{H})$, it follows that $|\kappa(S, \mathsf{O}_{\mathcal{U}}, \mathbb{A})| \leq \text{lit}(\mathcal{H})$. The reconstruction function $\rho$ simply runs CycleRobust (Algorithm E.3) on the compressed dataset $S' = \kappa(S, \mathsf{O}_{\mathcal{U}}, \mathbb{A})$. The fact that $\mathbb{A}$ is a conservative online learner implies that $\hat{h} = \mathbb{B}(S, \mathsf{O}_{\mathcal{U}}, \mathbb{A}) = \mathbb{B}(S', \mathsf{O}_{\mathcal{U}}, \mathbb{A})$. Since $\text{R}_{\mathcal{U}}(\hat{h}; S) = 0$, this establishes that $(\kappa, \rho)$ is a sample compression scheme for the robust loss. Furthermore, since $\mathbb{A}$ is a conservative online learner, observe that for any $S''$ such that $\kappa(S, \mathsf{O}_{\mathcal{U}}, \mathbb{A}) \subseteq S'' \subseteq S$ it holds that $\kappa(S, \mathsf{O}_{\mathcal{U}}, \mathbb{A}) = \kappa(S'', \mathsf{O}_{\mathcal{U}}, \mathbb{A})$. That is, removing any of the examples from $S$ on which $\hat{h}$ was robustly correct in Step 6 will not change the output of the compression function $\kappa$. Thus, the pair $(\kappa, \rho)$ is a stable sample compression scheme for the robust loss of size $\text{lit}(\mathcal{H})$. To conclude the proof, Lemma E.1 guarantees that for a sample size $m(\varepsilon, \delta) = O\left(\frac{\text{lit}(\mathcal{H}) + \log(1/\delta)}{\varepsilon}\right)$, the robust risk $\text{R}_{\mathcal{U}}(\hat{h}; \mathcal{D}) \leq \varepsilon$. □

### E.1.3 Auxiliary Lemmas and Proof of Theorem 7.6

**Theorem E.2** (Weak Robust Learner). *For any class $\mathcal{H}$ with $\text{vc}(\mathcal{H}) = d$ and $\text{vc}^*(\mathcal{H}) = d$,* RLUA *(Algorithm E.3) robustly PAC learns $\mathcal{H}$ w.r.t any $\mathcal{U}$ with:*

1. *Sample Complexity* $m(\varepsilon, \delta) = O\left(\frac{dd^{*2} \log^2 d^*}{\varepsilon} \log\left(\frac{dd^{*2} \log^2 d^*}{\varepsilon}\right) + \frac{\log(1/\delta)}{\varepsilon}\right)$.
2. *Oracle Complexity* $T(\varepsilon, \delta) = O\left(m(\varepsilon, \delta)^{dd^{*2} \log^2 d^*} + m(\varepsilon, \delta)^{dd^* \log d^*} \text{lit}(\mathcal{H})\right)$.

*Proof.* Let $\mathbb{A} : (\mathcal{X} \times \mathcal{Y})^* \to \mathcal{Y}^{\mathcal{X}}$ be an online learner for $\mathcal{H}$ with mistake bound $M(\mathbb{A}, \mathcal{H}) = O(\text{lit}(\mathcal{H}))$. We do not require $\mathbb{A}$ to be "proper" (i.e. returns a predictor in $\mathcal{H}$), but we will

**Algorithm E.3:** Robust Learner against Unknown Adversaries (RLUA)

**Input:** Training dataset $S = \{(x_1, y_1), \ldots, (x_m, y_m)\}$, black-box online learner $\mathbb{A}$ for $\mathcal{H}$, black-box perfect attack oracle $\mathsf{O}_{\mathcal{U}}$.

Set $n = O(\mathrm{vc}(\mathbb{A}))$. Foreach $L \subset S$ such that $|L| = n$, run `CycleRobust` on $(L, \mathbb{A}, \mathsf{O}_{\mathcal{U}})$, and denote by $\hat{\mathcal{H}}$ the resulting set of predictors.

Call `Discretizer` on $(S, \hat{\mathcal{H}}, \mathsf{O}_{\mathcal{U}})$, and denote by $\hat{S}_{\mathcal{U}}$ its output.

Initialize $D_1$ to be uniform over $\hat{S}_{\mathcal{U}}$, and set $T = O(\log |S_{\mathcal{U}}|)$.

**for** $1 \leq t \leq T$ **do**

 Sample $S' \sim D_t^n$, and project $S'$ to dataset $L_t \subseteq S$ by replacing each perturbation $z$ with its corresponding example $x$.

 Call `CycleRobust` on $(L_t, \mathbb{A}, \mathsf{O}_{\mathcal{U}})$, and denote by $f_t$ its output predictor.

 Compute a new distribution $D_{t+1}$ by applying the following update for each $(z, y) \in \hat{S}_{\mathcal{U}}$:

$$D_{t+1}(\{(z,y)\}) = \frac{D_t(\{(z,y)\})}{Z_t} \times \begin{cases} e^{-2\alpha} & \text{if } f_t(z) = y \\ 1 & \text{otherwise} \end{cases}$$

 where $Z_t$ is a normalization factor and $\alpha$ is set as in Lemma 6.4.

Sample $N = O(\mathrm{vc}^*(\mathbb{A}))$ i.i.d. indices $i_1, \ldots, i_N \sim \mathrm{Uniform}(\{1, \ldots, T\})$.

 (repeat previous step until $g = \mathrm{MAJ}(f_{i_1}, \ldots, f_{i_N})$ satisfies $\mathrm{R}_{\mathcal{U}}(g; S) = 0$)

**Output:** A majority-vote $\mathrm{MAJ}(f_{i_1}, \ldots, f_{i_N})$ predictor.

`Discretizer`(*Dataset S, Predictors $\hat{\mathcal{H}}$, Oracle* $\mathsf{O}_{\mathcal{U}}$):

 Initialize **for** $(x, y) \in S$ **do**

  Initialize $P = \{(x, y)\}$. Let $f_P^y : \mathcal{X} \to \mathcal{Y}$ be a predictor of the form: $f_P^y(x') = y$ if and only if $\left(\exists_{(z,y) \in P}\right) \left(\forall_{h \in \hat{\mathcal{H}}}\right) 1_{[h(z) \neq y]} = 1_{[h(x') \neq y]}$.

  Send the query $(f_P^y, (x, y))$ to the perfect attack oracle $\mathsf{O}_{\mathcal{U}}$.

  **while** $f_P^y$ *is not robustly correct on* $(x, y)$ **do**

   Let $z$ be the perturbation returned by $\mathsf{O}_{\mathcal{U}}$ where $f_P^y(z) \neq y$.

   Append $(z, y)$ to the set $P$.

   Send an updated query $(f_P^y, (x, y))$ to the perfect attack oracle $\mathsf{O}_{\mathcal{U}}$.

rely on it returning a predictor in some, possibly much larger, class which still has finite VC-dimension. To this end, we denote by $\mathrm{vc}(\mathbb{A}) = \mathrm{vc}(\mathrm{im}(\mathbb{A}))$ and $\mathrm{vc}^*(\mathbb{A}) = \mathrm{vc}^*(\mathrm{im}(\mathbb{A}))$ the primal and dual VC dimension of the image of $\mathbb{A}$, i.e. the class $\mathrm{im}(\mathbb{A}) = \{\mathbb{A}(S) | S \in (\mathcal{X} \times \mathcal{Y})^*\}$ of the possible hypothesis $\mathbb{A}$ might return. We will first prove a sample and oracle complexity bound stated in terms of $\mathrm{vc}(\mathbb{A})$ and $\mathrm{vc}^*(\mathbb{A})$, and later, at the end of the proof, we will use a result due to (Hanneke et al., 2021) to bound $\mathrm{vc}(\mathbb{A})$ and $\mathrm{vc}^*(\mathbb{A})$ in terms of $d = \mathrm{vc}(\mathcal{H})$ and $d^* = \mathrm{vc}(\mathcal{H})$ for a specific online learner $\mathbb{A}$.

Let $\mathcal{U} : \mathcal{X} \to 2^{\mathcal{X}}$ be an arbitrary adversary that is unknown to the learner. Let $\mathcal{D}$ be an arbitrary distribution over $\mathcal{X} \times \mathcal{Y}$ that is robustly realizable with some concept $h^* \in \mathcal{H}$, i.e., $\mathrm{R}_{\mathcal{U}}(h^*; \mathcal{D}) = 0$. Fix $\varepsilon, \delta \in (0,1)$ and a sample size $m$ that will be determined later. Let $S = \{(x_1, y_1), \ldots, (x_m, y_m)\}$ be an iid sample from $\mathcal{D}$.

Zero Empirical Robust Loss. Let $L \subseteq S$. Let $\mathbb{A}_{\mathrm{cyc}}$ be CycleRobust (Algorithm E.3) from Theorem 7.5. By Theorem 7.5, running $\mathbb{A}_{\mathrm{cyc}}$ on input $L$ with black-box access to $\mathsf{O}_{\mathcal{U}}$ and black-box access to $\mathbb{A}$, guarantees that the output $\hat{h} = \mathbb{A}_{\mathrm{cyc}}(L, \mathsf{O}_{\mathcal{U}}, \mathbb{A})$ satisfies $\mathrm{R}_{\mathcal{U}}(\hat{h}; L) = 0$ with at most $|L|\ \mathrm{lit}(\mathcal{H})$ oracle queries to $\mathsf{O}_{\mathcal{U}}$.

Discretization Before we can apply the compression approach, we will inflate dataset $S$ to a (potentially infinite) larger dataset $S_{\mathcal{U}} = \bigcup_{i \le m} \{(z, y_i) : z \in \mathcal{U}(x_i)\}$ that includes all possible adversarial perturbations under $\mathcal{U}$. There are two challenges that need to be addressed. First, $S_{\mathcal{U}}$ can be potentially infinite, and so we would need to discretize it somehow. Second, the learner does not know $\mathcal{U}$ and so the inflation can be carried only through interaction with the perfect attack oracle $\mathsf{O}_{\mathcal{U}}$. Denote by $\hat{\mathcal{H}} = \{\mathbb{A}_{\mathrm{cyc}}(L) : L \subseteq S, |L| = n\}$ where $n = O(\mathrm{vc}(\mathbb{A}))$. Think of $\hat{\mathcal{H}}$ as the effective hypothesis class that is used by our robust learning algorithm $\mathcal{B}$ that we are constructing. Note that $|\hat{\mathcal{H}}| \le |\{L : L \subseteq S, |L| = n\}| = \binom{m}{n} \le \left(\frac{em}{n}\right)^n$. We will now apply classic tools from VC theory to argue that there is a finite number of behaviors when projecting the infinite unknown set $S_{\mathcal{U}}$ onto $\hat{\mathcal{H}}$. Specifically, consider a *dual class* $\mathcal{G}$: a set of functions $g_{(x,y)} : \hat{\mathcal{H}} \to \{0,1\}$ defined as $g_{(x,y)}(h) = 1[h(x) \neq y]$, for each $h \in \hat{\mathcal{H}}$ and each $(x,y) \in S_{\mathcal{U}}$. The VC dimension of $\mathcal{G}$ is at most the *dual VC dimension* of $\hat{\mathcal{H}}$: $\mathrm{vc}^*(\hat{\mathcal{H}})$, which is at most $\mathrm{vc}^*(\mathbb{A})$ since $\hat{\mathcal{H}} \subseteq \mathrm{im}(\mathbb{A})$. The set of behaviors when projecting

$S_{\mathcal{U}}$ onto $\hat{\mathcal{H}}$ is defined as follows:

$$S_{\mathcal{U}}|_{\hat{\mathcal{H}}} = \left\{ \left( g_{(z,y)}(h_1), \dots, g_{(z,y)}(h_{|\hat{\mathcal{H}}|}) \right) : (z,y) \in S_{\mathcal{U}} \right\}.$$

Now denote by $\hat{S}_{\mathcal{U}}$ a subset of $S_{\mathcal{U}}$ which includes exactly one $(z,y) \in S_{\mathcal{U}}$ for each distinct classification $\left(g_{(z,y)}(h)\right)_{h\in\hat{\mathcal{H}}}$ of $\hat{\mathcal{H}}$ realized by some $(z,y) \in S_{\mathcal{U}}$. In particular, by applying Sauer's lemma Vapnik and Chervonenkis (1971), Sauer (1972) on the dual class $\mathcal{G}$, $|\hat{S}_{\mathcal{U}}| = |S_{\mathcal{U}}|_{\hat{\mathcal{H}}}| \leq \left(\frac{e|\hat{\mathcal{H}}|}{d^*}\right)^{d^*}$, which is at most $m^{nd^*}$. In particular, note that for any $T \in \mathbb{N}$ and $h_1, \dots, h_T \in \hat{\mathcal{H}}$, if $\frac{1}{T}\sum_{t=1}^{T} 1[h_t(x) = y] > \frac{1}{2}$ for every $(z,y) \in \hat{S}_{\mathcal{U}}$, then $\frac{1}{T}\sum_{t=1}^{T} 1[h_t(x) = y] > \frac{1}{2}$ for every $(z,y) \in S_{\mathcal{U}}$ as well, which would further imply $\mathrm{R}_{\mathcal{U}}(\mathrm{MAJ}(h_1, \dots, h_T); S) = 0$. Thus, we have shown that there *exists* a finite discretization $\hat{S}_{\mathcal{U}}$ of $S_{\mathcal{U}}$ where it suffices to find predictors $h_1, \dots, h_T \in \hat{\mathcal{H}}$ that achieve zero loss on $\hat{S}_{\mathcal{U}}$.

It remains to show how to construct the discretization $\hat{S}_{\mathcal{U}}$ using only interactions with the perfect attack oracle $\mathsf{O}_{\mathcal{U}}$. To this end, for each $(x,y) \in S$, initialize $P = \{(x,y)\}$. The robust learner $\mathbb{B}$ constructs a query $(f_P, (x,y))$ where $f_P : \mathcal{X} \to \mathcal{Y}$ is a predictor of the form:

$$f_P(x') = y \text{ if and only if } \left(\exists_{(z,y)\in P}\right) \left(\forall_{h\in\hat{\mathcal{H}}}\right) g_{(z,y)}(h) = g_{(x',y)}(h).$$

By the definition of $f_P$, if there is a perturbation $z' \in \mathcal{U}(x)$ such that the classification pattern $\left(g_{(z',y)}(h)\right)_{h\in\hat{\mathcal{H}}}$ is distinct from the classification pattern $\left(g_{(z,y)}(h)\right)_{h\in\hat{\mathcal{H}}}$ of any of the points $(z,y) \in P$, then $f_P(z') \neq y$, and therefore the oracle $\mathsf{O}_{\mathcal{U}}$ would reveal to the learner perturbation $z'$. Next, the learner adds the point $(z',y)$ to $P$, and repeats the procedure again until $f_P$ is robustly correct on example $(x,y)$. In each oracle query, the learner is forcing the oracle $\mathsf{O}_{\mathcal{U}}$ to reveal perturbations $z \in \mathcal{U}(x)$ with distinct classification patterns that the learner did not see before. Since we know that $\left|\hat{S}_{\mathcal{U}}\right| \leq m^{n\mathrm{vc}^*(\mathbb{A})}$, the learner makes at most $m^{n\mathrm{vc}^*(\mathbb{A})}$ oracle calls to $\mathsf{O}_{\mathcal{U}}$ before $f_P$ is robustly correct on $(x,y)$. This process is repeated for each training example $(x,y) \in S$, and so the total number of oracle calls to $\mathsf{O}_{\mathcal{U}}$ is at most $m^{n\mathrm{vc}^*(\mathbb{A})+1}$.

Oracle Complexity Our robust learner $\mathbb{B}$ makes $\left(\frac{em}{n}\right)^n n\mathrm{lit}(\mathcal{H})$ oracle calls to $\mathsf{O}_{\mathcal{U}}$ to construct $\hat{\mathcal{H}}$ and $m^{n\mathrm{vc}^*(\mathbb{A})+1}$ oracle calls to $\mathsf{O}_{\mathcal{U}}$ to construct $\hat{S}_{\mathcal{U}}$.

Sample Complexity and Robust Generalization We proceed by running the sample compression scheme from Montasser et al. (2019) on the discretized dataset $\hat{S}_{\mathcal{U}}$. In this stage no more oracle queries to $\mathsf{O}_{\mathcal{U}}$ are needed since the learner has already precomputed $\hat{\mathcal{H}}$ and the discretized dataset $\hat{S}_{\mathcal{U}}$. As mentioned above, our goal in this stage is to find predictors $h_1, \dots, h_T \in \hat{\mathcal{H}}$ where the majority-vote $\mathrm{MAJ}(h_1, \dots, h_T)$ achieves zero loss on $\hat{S}_{\mathcal{U}}$. This implies that $\mathrm{MAJ}(h_1, \dots, h_T)$ achieves zero robust loss on $S$, $\mathrm{R}_{\mathcal{U}}(\mathrm{MAJ}(h_1, \dots, h_T); S) = 0$, by properties of $\hat{\mathcal{H}}$ and $\hat{S}_{\mathcal{U}}$. We will next go about finding such a set of $h_t$ predictors.

We run the $\alpha$-Boost algorithm on the discretized dataset $\hat{S}_{\mathcal{U}}$, this time with $\mathbb{A}_{\mathrm{cyc}}$ (CycleRobust (Algorithm E.3)) as the subprocedure. Specifically, on each round of boosting, $\alpha$-Boost computes an empirical distribution $D_t$ over $\hat{S}_{\mathcal{U}}$. We draw $n = O(\mathrm{vc}(\mathbb{A}))$ samples $S'$ from $D_t$, and *project* $S'$ to a dataset $L_t \subset S$ by replacing each perturbation $(z, y) \in S'$ with its corresponding original point $(x, y) \in S$, and then we run $\mathbb{A}_{\mathrm{cyc}}$ on dataset $L_t$ (this is already precomputed since $\mathbb{A}_{\mathrm{cyc}}(L_t) \in \hat{\mathcal{H}}$ by definition of $\hat{\mathcal{H}}$). The projection step is crucial for the proof to work, since we use a *sample compression* argument to argue about *robust* generalization, and the sample compression must be done on the *original* points that appeared in $S$ rather than the perturbations in $\hat{S}_{\mathcal{U}}$.

By classic PAC learning guarantees Vapnik and Chervonenkis (1974), Blumer et al. (1989a), with $n = O(\mathrm{vc}(\mathbb{A}))$, we are guaranteed uniform convergence of 0-1 risk over predictors in $\hat{\mathcal{H}}$. So, for any distribution $D$ over $\mathcal{X} \times \mathcal{Y}$ with $\inf_{h \in \mathcal{H}} \mathrm{err}(h; \mathcal{D}) = 0$, with nonzero probability over $S' \sim \mathcal{D}^n$, every $h' \in \hat{\mathcal{H}}$ satisfying $\mathrm{err}_{S'}(h') = 0$, also has $\mathrm{err}_D(h') < 1/3$. As discussed above, we know that $h_t = \mathbb{A}_{\mathrm{cyc}}(L_t)$ achieves zero robust loss on $L_t$, $\mathrm{R}_{\mathcal{U}}(h_t; L_t) = 0$, which by definition of the projection means that $\mathrm{err}_{S'}(h_t) = 0$, and thus $\mathrm{err}_{D_t}(h_t) < 1/3$. This allows us to use $\mathbb{A}_{\mathrm{cyc}}$ with $\alpha$-Boost to establish a *robust* generalization guarantee. Specifically, Lemma 6.4 implies that running the $\alpha$-Boost algorithm with $\hat{S}_{\mathcal{U}}$ as its dataset for $T = O(\log(|\hat{S}_{\mathcal{U}}|))$ rounds, using $\mathbb{A}_{\mathrm{cyc}}$ to produce the hypotheses $h_t \in \hat{\mathcal{H}}$ for the distributions $D_t$ produced on each round of the algorithm, will produce a sequence of hypotheses $h_1, \dots, h_T \in \hat{\mathcal{H}}$ such that:

$$\forall (z, y) \in \hat{S}_{\mathcal{U}}, \frac{1}{T} \sum_{i=1}^{T} 1[h_i(z) = y] \geq \frac{5}{9}.$$

Specifically, this implies that the majority-vote over hypotheses $h_1, \dots, h_T$ achieves zero *robust* loss on dataset $S$, $\mathrm{R}_{\mathcal{U}}(\mathrm{MAJ}(h_1, \dots, h_T); S) = 0$. Note that each of these classifiers $h_t$ is equal to $\mathbb{A}(L_t, \mathsf{O}_{\mathcal{U}})$ for some $L_t \subseteq S$ with $|L_t| = n$. Thus, the classifier $\mathrm{MAJ}(h_1, \ldots, h_T)$ is representable as the value of an (order-dependent) reconstruction function $\varphi$ with a compression set size

$$nT = O(n \log(|S_{\mathcal{U}}|)).$$

We can further reduce the compression set size by sparsifying the majority-vote. Lemma 6.5 (with $\varepsilon = 1/18, \delta = 1/3$) guarantees that for $N = O(\mathrm{vc}^*(\mathbb{A}))$, the sampled predictors $h_{i_1}, \ldots, h_{i_N} \in \hat{\mathcal{H}}$ satisfy:

$$\forall (z, y) \in \hat{S}_{\mathcal{U}}, \frac{1}{N}\sum_{j=1}^{N} 1[h_{i_j}(z) = y] > \frac{1}{T}\sum_{i=1}^{T} 1[h_i(z) = y] - \frac{1}{18} > \frac{5}{9} - \frac{1}{18} = \frac{1}{2},$$

so that the majority-vote achieves zero robust loss on $S$, $\mathrm{R}_{\mathcal{U}}(\mathrm{MAJ}(h_{i_1}, \ldots, h_{i_N}); S) = 0$. Since again, each $h_{i_j}$ is the result of $\mathbb{A}(L_t, \mathsf{O}_{\mathcal{U}})$ for some $L_t \subseteq S$ with $|L_t| = m_0$, we have that the classifier $\mathrm{MAJ}(h_{i_1}, \ldots, h_{i_N})$ can be represented as the value of an (order-dependent) reconstruction function $\varphi$ with a compression set size $nN = O(\mathrm{vc}(\mathbb{A})\mathrm{vc}^*(\mathbb{A}))$. Lemma A.1 (Montasser et al. (2019)) which extends to the robust loss the classic compression-based generalization guarantees from the 0-1 loss, implies that for $m \geq c\mathrm{vc}(\mathbb{A})\mathrm{vc}(\mathbb{A})^*$ (for an appropriately large numerical constant $c$), with probability at least $1 - \delta$ over $S \sim \mathcal{D}^m$,

$$R_{\mathcal{U}}(\mathrm{MAJ}(h_{i_1}, \ldots, h_{i_N}); \mathcal{D}) \leq O\left(\frac{\mathrm{vc}(\mathbb{A})\mathrm{vc}^*(\mathbb{A})}{m}\log(m) + \frac{1}{m}\log(1/\delta)\right). \qquad \text{(E.1)}$$

Bounding the complexity of $\mathbb{A}$ A result due to (Hanneke et al., 2021, Theorem 3) states that for any class $\mathcal{H}$ of Littlestone dimension $\mathrm{lit}(\mathcal{H})$ and dual VC dimension $d^*$, there is an online learner $\mathbb{A}$ with mistake bound $M(\mathbb{A}, \mathcal{H}) = O(\mathrm{lit}(\mathcal{H}))$ which represents its hypotheses as (unweighted) majority votes of $O(d^*)$ predictors of $\mathcal{H}$. In other words,

$$\mathrm{im}(\mathbb{A}) \subseteq \mathrm{co}^{O(d^*)}(\mathcal{H}) \triangleq \left\{x \mapsto \mathrm{MAJ}(h_1, \ldots, h_{O(d^*)})(x) : h_1, \ldots, h_{O(d^*)} \in \mathcal{H}\right\}.$$

By (Blumer et al., 1989a), the VC dimension of $\mathrm{co}^{O(d^*)}(\mathcal{H})$ is at most $O(dd^* \log d^*)$, and by Lemma 6.3, the dual VC dimension of $\mathrm{co}^{O(d^*)}(\mathcal{H})$ is at most $O(d^* \log d^*)$. Since $\mathrm{im}(\mathbb{A}) \subseteq \mathrm{co}^{O(d^*)}(\mathcal{H})$, this implies that $\mathrm{vc}(\mathbb{A}) = O(dd^* \log d^*)$ and $\mathrm{vc}^*(\mathbb{A}) = O(d^* \log d^*)$. Substituting these upper bounds in Equation E.1, and setting it less than $\varepsilon$ and solving for a sufficient size of $m$ yields the stated sample complexity bound. $\square$

*Proof of Theorem 7.6.* Let $\mathbb{B}$ be the robust learning algorithm (Algorithm E.3) described in Theorem E.2. We will use $\mathbb{B}$ as a *weak* robust learner with fixed parameters $\varepsilon_0 = 1/3$ and $\delta_0 = 1/3$. By the guarantee of Theorem E.2, with fixed sample complexity $m_0 = O(dd^{*2}\log^2 d^*)$, for any distribution $\mathcal{D}$ over $\mathcal{X} \times \mathcal{Y}$ such that $\inf_{h\in\mathcal{H}} \mathrm{R}_{\mathcal{U}}(h;\mathcal{D}) = 0$, with probability at least $1/3$ over $S \sim \mathcal{D}^{m_0}$, $\mathrm{R}_{\mathcal{U}}(\mathbb{B}(S);\mathcal{D}) \leq 1/3$. Furthermore, $\mathbb{B}$ makes at most $O(dd^{*2}\log^2 d^*)^{O(dd^{*2}\log^2 d^*)} + O(dd^{*2}\log^2 d^*)^{O(dd^*\log d^*)}\mathrm{lit}(\mathcal{H}) = \exp\left\{O(dd^{*2}\log^2 d^*)\right\} + \exp\left\{O(d^2 d^{*2}\log^2 d^*)\right\}\mathrm{lit}(\mathcal{H})$ oracle calls to $\mathsf{O}_{\mathcal{U}}$.

We will now boost the confidence and robust error guarantee of the *weak* robust learner $\mathbb{B}$ by running boosting with respect to the *robust* loss (rather than the standard 0-1 loss). Specifically, fix $(\varepsilon, \delta) \in (0,1)$ and a sample size $m(\varepsilon,\delta)$ that will be determined later. Let $S = \{(x_1,y_1),\ldots,(x_m,y_m)\}$ be an iid sample from $\mathcal{D}$. Run the $\alpha$-Boost algorithm on dataset $S$ using $\mathbb{B}$ as the weak robust learner for a number of rounds $L = O(\log m)$. On each round $t$, $\alpha$-Boost computes an empirical distribution $D_t$ over $S$ by applying the following update for each $(x,y) \in S$:

$$D_t(\{(x,y)\}) = \frac{D_{t-1}(\{(x,y)\})}{Z_{t-1}} \times \begin{cases} e^{-2\alpha} & \text{if } \sup_{z\in\mathcal{U}(x)} \mathbb{1}[h_{t-1}(z) \neq y] = 0 \\ 1 & \text{otherwise} \end{cases}$$

where $Z_{t-1}$ is a normalization factor, $\alpha$ is set as in Lemma 6.4, and $h_{t-1}$ is the *weak* robust predictor outputted by $\mathbb{B}$ on round $t-1$ that satisfies $\mathrm{R}_{\mathcal{U}}(h_{t-1};D_{t-1}) \leq 1/3$. Note that computing $D_t$ requires $|S| = m$ oracle calls to $\mathsf{O}_{\mathcal{U}}$. Once $D_t$ is computed, we sample $m_0$ examples from $D_t$ and run *weak* robust learner $\mathbb{B}$ on these examples to produce a hypothesis $h_t$ with robust error guarantee $\mathrm{R}_{\mathcal{U}}(h_t;D_t) \leq 1/3$. This step has failure probability at most $\delta_0 = 1/3$. We will repeat it for at most $\lceil \log(2L/\delta)\rceil$ times, until $\mathbb{B}$ succeeds in finding $h_t$ with robust error guarantee $\mathrm{R}_{\mathcal{U}}(h_t;D_t) \leq 1/3$. By a union bound argument, we are guaranteed that with probability at least $1-\delta/2$, foreach $1 \leq t \leq L$, $\mathrm{R}_{\mathcal{U}}(h_t;D_t) \leq 1/3$. Furthermore, by Lemma 6.4, we are guaranteed that $\mathrm{R}_{\mathcal{U}}(\mathrm{MAJ}(h_1,\ldots,h_L);S) = 0$. Note that each of these classifiers $h_t$ is equal to $\mathbb{B}(S'_t, \mathsf{O}_{\mathcal{U}})$ for some $S'_t \subseteq S$ with $|S'_t| = m_0$. Thus, the classifier $\mathrm{MAJ}(h_1,\ldots,h_L)$ is representable as the value of an (order-dependent) reconstruction function $\varphi$ with a compression set size $m_0 L = m_0 O(\log m)$. Now, invoking Lemma A.1, with probability at least

$1-\delta/2$,

$$\mathrm{R}_{\mathcal{U}}(\mathrm{MAJ}(h_1,\dots,h_L);\mathcal{D}) \le O\left(\frac{m_0\log^2 m}{m} + \frac{\log(2/\delta)}{m}\right),$$

and setting this less than $\varepsilon$ and solving for a sufficient size of $m$ yields the stated sample complexity bound.

Oracle Complexity Observe that we run boosting for $L$ rounds, in each round the *weak* robust learner is invoked at most $\lceil\log(2L/\delta)\rceil$ times. In each of these invocations, $\mathbb{B}$ makes at most $\exp\left\{O(dd^{*2}\log^2 d^*)\right\} + \exp\left\{O(d^2d^{*2}\log^2 d^*)\right\}\mathrm{lit}(\mathcal{H})$ oracle calls to $\mathsf{O}_{\mathcal{U}}$, and an additional $m(\varepsilon,\delta)$ oracle calls to $\mathsf{O}_{\mathcal{U}}$ are made by $\alpha$-Boost to compute the robust error of the $h_t$ hypotheses produced by $\mathbb{B}$. Thus, the total number of calls to $\mathsf{O}_{\mathcal{U}}$ is at most

$$\lceil L\log(2L/\delta)\rceil\left(\exp\left\{O(dd^{*2}\log^2 d^*)\right\} + \exp\left\{O(d^2d^{*2}\log^2 d^*)\right\}\mathrm{lit}(\mathcal{H}) + m(\varepsilon,\delta)\right).$$

□

## E.2 The Perfect Attack Oracle model (Proofs for Agnostic Guarantees)

**Lemma E.3.** *For any class $\mathcal{H}$ with finite cardinality,* Weighted Majority *(Algorithm E.4) guarantees that for any $\mathcal{U}$ and any sequence of examples $(x_1,y_1),\dots,(x_T,y_T)$:*

$$\sum_{t=1}^{T}\sup_{z\in\mathcal{U}(x_t)} 1[\mathrm{MAJ}_{P_{t-1}}(z)\neq y_t] \le a_\eta \min_{h\in\mathcal{H}}\sum_{t=1}^{T}\sup_{z\in\mathcal{U}(x_t)} 1[h(z)\neq y_t] + b_\eta \ln|\mathcal{H}|,$$

*where $a_\eta = \frac{\ln(1/\eta)}{\log(2/(1+\eta))}$ and $b_\eta = \frac{1}{\ln(2/(1+\eta))}$. In particular, setting $1-\eta = \min\{(2\ln|\mathcal{H}|)/T, 1/2\}$ yields*

$$\sum_{t=1}^{T}\sup_{z\in\mathcal{U}(x_t)} 1[\mathrm{MAJ}_{P_{t-1}}(z)\neq y_t] \le 2\mathsf{OPT} + 4\sqrt{\mathsf{OPT}\ln|\mathcal{H}|}.$$

*Furthermore,* Weighted Majority *(Algorithm E.4) makes at most $T$ oracle queries to $\mathsf{O}_{\mathcal{U}}$.*

*Proof.* This proof follows from standard analysis for the Weighted Majority algorithm (see e.g. Schapire (2006), Blum and Monsour (2007)). Let $W_t = \sum_{h\in\mathcal{H}} P_t(h)$. Observe that

**Algorithm E.4:** Weighted Majority

**Input:** paramter $\eta \in [0,1)$, black-box perfect attack oracle $\mathsf{O}_{\mathcal{U}}$, and finite hypothesis class $\mathcal{H}$.

Initialize $P_0$ to be uniform over $\mathcal{H}$, i.e. $\forall h \in \mathcal{H}, P_0(h) = 1$.

**for** $1 \leq t \leq T$ **do**

- Receive $(x_t, y_t)$.
- Certify the robustness of the weighted-majority-vote $\mathrm{MAJ}_{P_{t-1}}$ on $(x_t, y_t)$ by sending the query $(\mathrm{MAJ}_{P_{t-1}}, (x_t, y_t))$ to the perfect attack oracle $\mathsf{O}_{\mathcal{U}}$.
- **if** $\mathrm{MAJ}_{P_{t-1}}$ *is not robustly correct on* $(x_t, y_t)$ **then**
  - Let $z_t$ be the perturbation returned by $\mathsf{O}_{\mathcal{U}}$ where $\mathrm{MAJ}_{P_{t-1}}(z_t) \neq y_t$.
  - Foreach $h \in \mathcal{H}$ such that $h(z_t) \neq y_t$, update $P_t(h) = \eta P_{t-1}(h)$.

**Output:** The weighted-majority-vote $\mathrm{MAJ}_{P_T}$ over $\mathcal{H}$.

`Expert(`*Indices* $i_1 < i_2 < \cdots < i_L$, *and hypothesis class* $\mathcal{H}$`):`

- Initialize $V_1 = \mathcal{H}$.
- **for** $1 \leq t \leq T$ **do**
  - Receive $x_t$.
  - Let $V_t^y = \{h \in V_t : h(x_t) = y\}$ for $y \in \{\pm 1\}$.
  - Let $\tilde{y}_t = \mathrm{argmax}_{y \in \{\pm 1\}} \mathrm{lit}(V_t^y)$ (in case of a tie set $\tilde{y}_t = +1$).
  - **if** $t \in \{i_1, \ldots, i_L\}$ **then**
    - Predict $\hat{y}_t = -\tilde{y}_t$.
  - **else**
    - Predict $\hat{y}_t = \tilde{y}_t$.
  - Update $V_{t+1} = V_t^{\hat{y}_t}$.

on round $t$, if the weighted-majority-vote $\mathrm{MAJ}_{P_{t-1}}$ is not robustly correct on $(x_t, y_t)$, then:

$$W_t = \eta \sum_{h \in \mathcal{H}: h(z_t) \neq y} P_{t-1}(h) + \sum_{h \in \mathcal{H}: h(z_t) = y} P_{t-1}(h) = \eta \sum_{h \in \mathcal{H}: h(z_t) \neq y} P_{t-1}(h) + W_{t-1} - \sum_{h \in \mathcal{H}: h(z_t) \neq y} P_{t-1}(h)$$

$$= W_{t-1} - (1-\eta) \left( \sum_{h \in \mathcal{H}: h(z_t) \neq y} P_{t-1}(h) \right) \leq W_{t-1} - (1-\eta) \frac{1}{2} W_{t-1} = \left( \frac{\eta + 1}{2} \right) W_{t-1},$$

where the last inequality follows from the fact that $\sum_{h \in \mathcal{H}: h(z_t) \neq y} P_{t-1}(h) \geq \sum_{h \in \mathcal{H}: h(z_t) = y} P_{t-1}(h)$.

Denote by $M = \sum_{t=1}^{T} \sup_{z \in \mathcal{U}(x_t)} \mathbb{1}[\mathrm{MAJ}_{P_{t-1}}(z) \neq y_t]$ the number of rounds on which the weighted-majority-vote was not robustly correct during the total $T$ rounds. The above implies that $W_T \leq \left(\frac{\eta+1}{2}\right)^M W_0 = \left(\frac{\eta+1}{2}\right)^M |\mathcal{H}|$. On the other hand, denote by $\mathsf{OPT} = \min_{h\in\mathcal{H}} \sum_{t=1}^{T} \sup_{z\in\mathcal{U}(x)} 1[h(z) \neq y]$ the number of rounds on which the best predictor $h^*$ in $\mathcal{H}$ was not robustly correct. Whenever the weighted-majority-vote is not robustly correct, $h^*$ might make a mistake on $(z_t, y_t)$. It follows that after $T$ rounds, $P_T(h^*) \geq \eta^{\mathsf{OPT}}$. Combining the above inequalities, we get

$$\eta^{\mathsf{OPT}} \leq P_T(h^*) \leq W_T \leq \left(\frac{\eta+1}{2}\right)^M |\mathcal{H}|\,,$$

and solving for $M$ yields

$$M \leq \frac{\ln(1/\eta)}{\ln(2/(1+\eta))}\mathsf{OPT} + \frac{1}{\ln(2/(1+\eta))}\ln|\mathcal{H}|\,.$$

To conclude the proof, observe that for $\eta \in [0,1), \ln(2/(1+\eta)) \geq \frac{1-\eta}{2}$, and $\ln(1/\eta) \leq (1-\eta) + (1-\eta)^2$ for $0 \leq 1-\eta \leq 1/2$. Setting $1-\eta = \min\{(2\ln|\mathcal{H}|)/T, 1/2\}$ yields the desired bound. $\square$

**Lemma E.4.** *For any class $\mathcal{H}$ with finite Littlestone dimension* $\mathrm{lit}(\mathcal{H}) < \infty$ *and integer $T$, let* $\mathsf{Experts}_{\mathcal{H}} = \{\mathsf{Expert}(i_1, \ldots, i_L) : 1 \leq i_1 < \cdots < i_L \leq T, L \leq \mathrm{lit}(\mathcal{H})\}$ *be a set of experts as described in Algorithm E.4. Then, running* Weighted Majority *(Algorithm E.4) with* $\mathsf{Experts}_{\mathcal{H}}$ *guarantees that for any perturbation set $\mathcal{U}$ and any sequence of examples* $(x_1, y_1), \ldots, (x_T, y_T)$,

$$\sum_{t=1}^{T} \sup_{z\in\mathcal{U}(x_t)} 1[\mathrm{MAJ}_{P_{t-1}}(z) \neq y_t] \leq 2\mathsf{OPT} + 4\sqrt{\mathsf{OPT}\ln|\mathsf{Experts}_{\mathcal{H}}|},$$

*where* $\mathsf{OPT} = \min_{h\in\mathcal{H}} \sum_{t=1}^{T} \sup_{z\in\mathcal{U}(x_t)} 1[h(z) \neq y_t]$, *and* $1-\eta = \min\{(2\ln|\mathsf{Experts}_{\mathcal{H}}|)/T, 1/2\}$. *Furthermore,* Weighted Majority *(Algorithm E.4) makes at most $T$ oracle queries to* $\mathsf{O}_{\mathcal{U}}$.

*Proof.* Let $\mathcal{U}$ be an arbitrary adversary, and $(x_1, y_1), \ldots, (x_T, y_T) \in \mathcal{X} \times \mathcal{Y}$ be an arbitrary sequence. Let $h^* \in \mathcal{H}$ be an optimal robust predictor on this sequence, i.e. $\sum_{t=1}^{T} \sup_{z\in\mathcal{U}(x_t)} 1[h^*(z) \neq y_t] = \mathsf{OPT}$. Let $\mathtt{Experts}_{\mathcal{H}} = \{\mathtt{Expert}(i_1, \ldots, i_L) : 1 \leq i_1 < \cdots < i_L \leq T, L \leq \mathrm{lit}(\mathcal{H})\}$ denote the set of experts instantiated that simulate the Standard Optimal Algorithm as described in Algorithm E.4.

Consider running Weighted Majority (Algorithm E.4) with $\text{Experts}_{\mathcal{H}}$ as its finite cardinality set of experts on the sequence $(x_1, y_1), \ldots, (x_T, y_T)$. Consider the set of perturbations returned by the perfect attack oracle $\mathsf{O}_{\mathcal{U}}$ during the rounds on which the weighted-majority-vote was not robustly correct,

$$Q = \left\{ (z_t, y_t) : 1 \le t \le T \wedge \text{MAJ}_{P_{t-1}} \text{ is not robustly correct on } (x_t, y_t) \right\}.$$

By Algorithm E.4, there is a choice of rounds $i_1^* < \cdots < i_L^*$ such that $\texttt{Expert}(i_1^*, \ldots, i_L^*) \in \texttt{Experts}_{\mathcal{H}}$ agrees with the predictions of $h^*$ on this particular sequence $Q$. Observe that the number of mistakes $h^*$ makes on this sequence $M(h^*) := |\{(z, y) \in Q : h^*(z) \neq y\}| \le \mathsf{OPT}$. Thus, the weight of $\texttt{Expert}(i_1^*, \ldots, i_L^*)$ is at least $\eta^{M(h^*)} \ge \eta^{\mathsf{OPT}}$ (since $\eta < 1$). The remainder of the proof follows exactly as in the proof of Theorem 7.7. □

*Proof of Theorem 7.7.* Let $\mathcal{H} \subseteq \mathcal{Y}^{\mathcal{X}}$ be a hypotesis class with finite Littlestone dimension $\text{lit}(\mathcal{H}) < \infty$. Let $\mathbb{B} : (\mathcal{X} \times \mathcal{Y})^* \to \mathcal{Y}^{\mathcal{X}}$ denote the Weighted Majority algorithm running with experts $\text{Experts}_{\mathcal{H}}$ as described in Theorem 7.7. We will apply a standard online-to-batch conversion Cesa-Bianchi et al. (2004) to get the desired result. Specifically, on input dataset $S = \left\{(x_j, y_j)\right\}_{j=1}^m$ that is drawn iid from some unknown distribution $\mathcal{D}$ over $\mathcal{X} \times \mathcal{Y}$, output a uniform distribution over hypotheses $\hat{h}_0, \hat{h}_1, \ldots, \hat{h}_{m-1}$ where $\hat{h}_i = \mathbb{B}(\left\{(x_j, y_j)\right\}_{j=1}^{i-1})$. We are guaranteed that with probability at least $1 - \delta$ over $S \sim \mathcal{D}^m$,

$$\begin{aligned}
\mathop{\mathbb{E}}_{j \sim \text{Unif}\{0,\ldots,m-1\}} \left[ \mathrm{R}_{\mathcal{U}}(\hat{h}_j; \mathcal{D}) \right] &\le \frac{1}{m} \sum_{j=1}^m \sup_{z \in \mathcal{U}(x_j)} 1[\hat{h}_{j-1}(z) \neq y_j] + \sqrt{\frac{2 \ln(1/\delta)}{m}} \\
&\le 2 \min_{h \in \mathcal{H}} \frac{1}{m} \sum_{j=1}^m \sup_{z \in \mathcal{U}(x_j)} 1[h(z) \neq y_j] + 4\sqrt{\frac{\ln |\text{Experts}_{\mathcal{H}}|}{m}} + \sqrt{\frac{2 \ln(1/\delta)}{m}} \\
&\le 2 \min_{h \in \mathcal{H}} \mathrm{R}_{\mathcal{U}}(h; \mathcal{D}) + 4\sqrt{\frac{\ln |\text{Experts}_{\mathcal{H}}|}{m}} + 2\sqrt{\frac{2 \ln(1/\delta)}{m}}.
\end{aligned}$$

This yields a sample complexity bound of $m(\varepsilon, \delta) = O\left(\frac{\ln|\text{Experts}_{\mathcal{H}}| + \ln(1/\delta)}{\varepsilon^2}\right)$. The oracle complexity $T(\varepsilon, \delta) = O(m(\varepsilon, \delta)^2)$ since we invoke learner $\mathbb{B}$ $m$ times on datasets of size at most $m$. □

*Proof of Theorem 7.8.* This proof follows an argument originally made by David et al. (2016)

to reduce agnostic sample compression to realizable sample compression in the non-robust setting, and later adapted by Montasser et al. (2019) for the robust setting. Let $\mathcal{D}$ be an arbitrary distribution over $\mathcal{X} \times \mathcal{Y}$. Fix $\varepsilon, \delta \in (0,1)$ and a sample size $m$ that will be determined later. Let $S = \{(x_1, y_1), \ldots, (x_m, y_m)\}$ be an iid sample from $\mathcal{D}$. Denote by $\tilde{\mathbb{B}}$ the robust learning algorithm in the realizable setting from Theorem 7.6, and denote by $\mathbb{A}_{\text{cyc}}$ CycleRobust (Algorithm E.3) from Theorem 7.5. The proof is broken into two parts.

Finding Maximal Subsequence $S'$ with Zero Robust Loss   We will use $\mathbb{A}_{\text{cyc}}$ to find a maximal subsequence $S' \subseteq S$ on which the robust loss can be zero, i.e. $\inf_{h \in \mathcal{H}} \mathrm{R}_{\mathcal{U}}(h; S') = 0$. This can be done by running $\mathbb{A}_{\text{cyc}}$ on all $2^m$ possible subsequences, with a total oracle complexity of $2^m \mathrm{lit}(\mathcal{H})$.

Agnostic Sample Compression   We now run the boosting algorithm $\tilde{B}$ on $S'$. Theorem 7.6 guarantees that the robust risk of $\hat{h} = \tilde{B}(S', \mathsf{O}_{\mathcal{U}}, \mathbb{A}_{\text{cyc}})$ is zero, $\mathrm{R}_{\mathcal{U}}(\hat{h}; S') = 0$. Since $S'$ is a maximal subsequence on which the robust loss can be zero, this implies that

$$\mathrm{R}_{\mathcal{U}}(\hat{h}; S) \leq \min_{h \in \mathcal{H}} \mathrm{R}_{\mathcal{U}}(h; S).$$

Furthermore, the predictor $\hat{h}$ can be specified using $m_0 O(\log |S'|) \leq m_0 O(\log m)$ points from $S$, which is due to the robust compression guarantee in the proof of Theorem 7.6. Now, we can apply agnostic sample compression generalization guarantees for the robust loss.

Similarly to the realizable case (see Lemma A.1), uniform convergence guarantees for sample compression schemes Graepel et al. (2005) remain valid for the robust loss, by essentially the same argument; the essential argument is the same as in the proof of Lemma A.1 except using Hoeffding's inequality to get concentration of the empirical robust risks for each fixed index sequence, and then a union bound over the possible index sequnces as before. We omit the details for brevity. In particular, denoting $T_m = O(\log m)$, for $m > m_0 T_m$, with probability at least $1 - \delta/2$,

$$\mathrm{R}_{\mathcal{U}}(\hat{h}; \mathcal{D}) \leq \hat{\mathrm{R}}_{\mathcal{U}}(\hat{h}; S) + \sqrt{\frac{m_0 T_m \ln(m) + \ln(2/\delta)}{2m - 2m_0 T_m}}.$$

Let $h^* = \mathrm{argmin}_{h \in \mathcal{H}} \mathrm{R}_{\mathcal{U}}(h; \mathcal{D})$ (supposing the min is realized, for simplicity; else we could take an $h^*$ with very-nearly minimal risk). By Hoeffding's inequality, with probability at least $1 - \delta/2$,

$$\hat{\mathrm{R}}_{\mathcal{U}}(h^*; S) \le \mathrm{R}_{\mathcal{U}}(h^*; \mathcal{D}) + \sqrt{\frac{\ln(2/\delta)}{2m}}.$$

By the union bound, if $m \ge 2m_0 T_m$, with probability at least $1 - \delta$,

$$\begin{aligned}
\mathrm{R}_{\mathcal{U}}(\hat{h}; \mathcal{D}) &\le \min_{h \in \mathcal{H}} \hat{\mathrm{R}}_{\mathcal{U}}(h; S) + \sqrt{\frac{m_0 T_m \ln(m) + \ln(2/\delta)}{m}} \\
&\le \hat{\mathrm{R}}_{\mathcal{U}}(h^*; S) + \sqrt{\frac{m_0 T_m \ln(m) + \ln(2/\delta)}{m}} \\
&\le \mathrm{R}_{\mathcal{U}}(h^*; \mathcal{D}) + 2\sqrt{\frac{m_0 T_m \ln(m) + \ln(2/\delta)}{m}}.
\end{aligned}$$

Since $T_m = O(\log(m))$, the above is at most $\varepsilon$ for an appropriate choice of sample size $m = O\left(\frac{m_0}{\varepsilon^2} \log^2\left(\frac{m_0}{\varepsilon}\right) + \frac{1}{\varepsilon^2} \log\left(\frac{1}{\delta}\right)\right)$. □

## E.3 Proof for Oracle Complexity Lower Bound in the Perfect Attack Oracle model

*Proof of Theorem 7.10.* Let $d = \mathrm{Tdim}(\mathcal{H})$. By definition of the threshold dimension, $\exists P = \{x_1, \dots, x_d\} \subseteq \mathcal{X}$ that is threshold-shattered using $C = \{h_1, \dots, h_d\} \subseteq \mathcal{H}$. Let $\mathcal{D}$ be a uniform distribution over $(x_1, +1)$ and $(x_d, -1)$. Let $\mathbb{B}$ be an arbitrary learner in the Perfect Attack Oracle model. For any $h \in C \setminus \{h_d\}$, let $\mathcal{U}_h : \mathcal{X} \to 2^{\mathcal{X}}$ be defined as:

$$\begin{aligned}
\mathcal{U}_h(x_1) &= \{x \in P : h(x) = +1\}, \\
\mathcal{U}_h(x_d) &= \{x \in P : h(x) = -1\} = P \setminus \mathcal{U}_h(x_1), \\
\mathcal{U}_h(x) &= \{x_0\} \ \forall x \in \mathcal{X} \setminus \{x_1, x_d\},
\end{aligned}$$

where $x_0 \in \mathcal{X} \setminus P$.

For any such $\mathcal{U}_h$, observe that finding a predictor $\hat{h} : \mathcal{X} \to \mathcal{Y}$ that achieves zero robust loss on $\mathcal{D}$, $\mathrm{R}_{\mathcal{U}_h}(\hat{h}; \mathcal{D}) = 0$, is equivalent to figuring out which threshold $h \in C \setminus h_d$ was used to construct $\mathcal{U}_h$, since $\mathrm{R}_{\mathcal{U}_h}(h; \mathcal{D}) = 0$ by definition of $\mathcal{U}_h$, but for any other threshold $h' \in C \setminus h_d$ where $h' \neq h$, $\mathrm{R}_{\mathcal{U}}(h'; \mathcal{D}) \ge 1/2$.

We will pick $h$ uniformly at random from $C \setminus h_d$, and we will show that in expectation over the random draw of $h$, learner $\mathbb{A}$ needs to make at least $\Omega(\log |C \setminus h_d|)$ oracle queries to $\mathsf{O}_{\mathcal{U}_h}$ in order to achieve robust loss zero on $\mathcal{D}$. For ease of presentation, for each $i \in [d-1]$, we will encode $h_i \in C \setminus h_d$ with the binary representation $r(i)$ of integer $i$, for example:

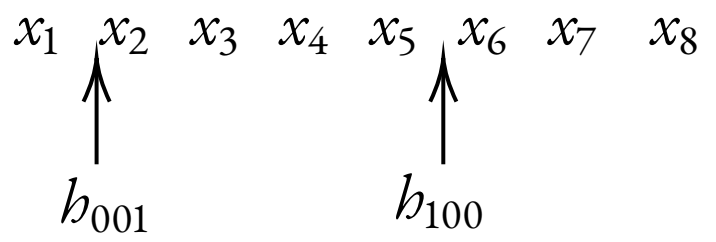

Thus, drawing $h$ uniformly at random from $C \setminus h_d$ is equivalent to drawing a random bit-string $r$ of length $\lceil \log_2 |C \setminus h_d| \rceil$ bits. Next, we will define the behavior of the oracle $\mathsf{O}_{\mathcal{U}_h}$.

**Algorithm E.5:** Perfect Attack Oracle $\mathsf{O}_{\mathcal{U}_h}$

**Input:** A predictor $f : \mathcal{X} \to \mathcal{Y}$ and a labeled example $(x, y)$.

**Output:** Assert that $f$ is robustly correct on $(x, y)$ or return a $z \in \mathcal{U}_h(x)$ such that $f(z) \neq y$.

**if** $x = x_1$ **then**
  Output the first $z \in \mathcal{U}_h(x_1)$ **(to the right of** $x_1$**)** such that $f(z) \neq y$. If no such $z$ exists, assert that $f$ is robustly correct on $(x_1, y)$.
**else if** $x = x_d$ **then**
  Output the first $z \in \mathcal{U}_h(x_d)$ **(to the left of** $x_d$**)** such that $f(z) \neq y$. If no such $z$ exists, assert that $f$ is robustly correct on $(x_d, y)$.
**else**
  Output $x$ if $f(x) \neq y$, otherwise assert that $f$ is robustly correct on $(x, y)$.

Before learning starts, from the perspective of the learner $\mathbb{B}$, the version space $V_0 = \{h_1, \ldots, h_{d-1}\}$, as any of these thresholds could be the true threshold used by $\mathsf{O}_{\mathcal{U}_{h_r}}$ where $r$ was drawn uniformly at random. On each round $t$, learner $\mathbb{B}$ constructs a predictor $h_t : \mathcal{X} \to \mathcal{Y}$ and asks the oracle $\mathcal{O}_{\mathcal{U}_{h_r}}$ with a query $q_t = (h_t, (x_t, y_t))$, and the oracle $\mathsf{O}_{\mathcal{U}_{h_r}}$ responds as described in Algorithm 14. Without loss of generality, we can assume that $(x_t, y_t) = (x_1, +1)$ or $(x_t, y_t) = (x_d, -1)$, since queries concerning other points $x \in \mathcal{X} \setminus \{x_1, x_d\}$ do not reveal helpful information for robustly learning distribution $\mathcal{D}$. Foreach round $t$, the version space $V_t$ describes the set of thresholds that are consistent with the queries constructed by the learner so far, i.e. $\forall i \leq t, \forall h, h' \in V_t, \mathsf{O}_{\mathcal{U}_h}(q_i) = \mathsf{O}_{\mathcal{U}_{h'}}(q_i)$. So, from the perspective of the learner $\mathbb{B}$, any $h \in V_t$ could be the true threshold.

We will show that with each oracle query $q_t$ constructed by the learner $\mathbb{B}$, in expectation over the random draw of $r$, the size of the newly updated version space $|V_t| \geq \frac{1}{4}|V_{t-1}|$. Formally, the expected size of the version space $V_t$ after round $t$ conditioned on query $q_t$ and $V_{t-1}$ is:

$$\mathop{\mathbb{E}}_{r_t}[|V_t|\,|q_t, V_{t-1}] = \Pr_{r_t}[r_t = 0]\,\mathbb{E}[|V_t|\,|q_t, V_{t-1}, r_t] + \Pr_{r_t}[r_t = 1]\,\mathbb{E}[|V_t|\,|q_t, V_{t-1}, r_t],$$

where $r_t$ is the $t^{\text{th}}$ random bit in the random bit string $r$. We need to consider two possible cases depending on the query $q_t = (h_t, (x_t, y_t))$. (Without loss of generality, we are assuming that $h_t \in C \setminus h_d$, as the oracle $\mathsf{O}_{\mathcal{U}_{h_r}}$ will treat it as such by Steps 2 and 4).

If $(x_t, y_t) = (x_1, +1)$, and the $t^{\text{th}}$ bit of $h_t$ is 0, then:

$$\mathop{\mathbb{E}}_{r_t}[|V_t|\,|q_t, V_{t-1}] \geq \Pr_{r_t}[r_t = 1]\,\mathbb{E}[|V_t|\,|q_t, V_{t-1}, r_t] = \frac{1}{2} \cdot \frac{1}{2}|V_{t-1}|.$$

If $(x_t, y_t) = (x_d, -1)$, and the $t^{\text{th}}$ bit of $h_t$ is 1, then:

$$\mathop{\mathbb{E}}_{r_t}[|V_t|\,|q_t, V_{t-1}] \geq \Pr_{r_t}[r_t = 0]\,\mathbb{E}[|V_t|\,|q_t, V_{t-1}, r_t] = \frac{1}{2} \cdot \frac{1}{2}|V_{t-1}|.$$

Therefore, it follows that $\mathbb{E}_{r_t}[|V_t|\,|q_t, V_{t-1}] \geq \frac{1}{4}|V_{t-1}|$. Thus, after $T$ rounds, $\mathbb{E}_r[|V_T|\,|V_0] \geq \frac{1}{4}^T|V_0|$. This implies that there exists a fixed bit-string $r^*$ (or equivalently, an adversary $\mathcal{U}_{h_{r^*}}$) such that for $T \leq \frac{\log|V_0|}{2}$ rounds, $|V_T| \geq 1$. This implies that learner $\mathbb{B}$ needs at least $\frac{\log|V_0|}{2}$ oracle queries to $\mathsf{O}_{\mathcal{U}_{h_{r^*}}}$ in order to robustly learn distribution $\mathcal{D}$. □

## E.4 Proof for Bounding Number of Successful Attacks

*Proof of Theorem 7.14.* Observe that the same lowerbound construction from Theorem 7.10 can be used in the setting of the online model. Specifically, by importing that construction, we get the following: there is a distribution $\mathcal{D}$ over $\mathcal{X} \times \mathcal{Y}$, such that for any learner $\mathbb{B} : (\mathcal{X} \times \mathcal{Y})^* \to \mathcal{Y}^{\mathcal{X}}$, there is a perturbation set $\mathcal{U} : \mathcal{X} \to 2^{\mathcal{X}}$ and an adversary $\mathbb{A} : \mathcal{Y}^{\mathcal{X}} \times (\mathcal{X} \times \mathcal{Y}) \to$

$\mathcal{X}$ (Algorithm 14) such that:

$$M_{\mathcal{U},\mathbb{A}}(\mathbb{B},\mathcal{H};\mathcal{D}) = \sum_{t=1}^{\infty} 1\left[\mathbb{B}(\{(z_i,y_i)\}_{i=1}^{t-1})(z_t) \neq y_t\right] \geq \frac{\log_2(\mathrm{Tdim}(\mathcal{H})-1)}{2}.$$

This is because in the setting of the Perfect Attack Oracle model, learner $\mathbb{B}$ *chooses* which example $(x,y) \in \mathrm{supp}(\mathcal{D})$ to feed into $\mathbb{A}$, and still learner $\mathbb{B}$ makes $\frac{\log_2(\mathrm{Tdim}(\mathcal{H})-1)}{2}$ mistakes before fully robustly learning distribution $\mathcal{D}$. While in this setting, the examples $(x,y)$ that are fed into $\mathbb{A}$ are drawn iid from $\mathcal{D}$, and so its at least as hard as the other setting. □

## E.5 Proof for Robust Generalization to Imperfect Attacks

**Algorithm E.6:** Robust Learner with Imperfect Attack.

**Input:** $S = \{(x_1,y_1),\ldots,(x_m,y_m)\}$, $\varepsilon$, $\delta$, black-box conservative online learner $\mathbb{A}$, black-box attacker $\mathbb{A}$.

Initialize $\hat{h} = \mathbb{A}(\emptyset)$.
**for** $1 \leq i \leq m$ **do**
  Let $z_i = \mathbb{A}(\hat{h},(x_i,y_i))$ be the perturbation returned by the attacker $\mathbb{A}$.
  If $\hat{h}$ is not correct on $(z_i,y_i)$, update $\hat{h}$ by running online learner on $\mathbb{A}$ on $(z_i,y_i)$.
  Break when $\hat{h}$ is correct on a consecutive sequence of perturbations of length $\frac{1}{\varepsilon}\log\left(\frac{\mathrm{lit}(\mathcal{H})}{\delta}\right)$.

**Output:** $\hat{h}$.

*Proof of Theorem 7.18.* Let $\mathcal{H} \subseteq \mathcal{Y}^{\mathcal{X}}$ be an arbitrary hypothesis class, and $\mathbb{A}$ a conservative online learner for $\mathcal{H}$ with mistake bound of $\mathrm{lit}(\mathcal{H})$. Let $\mathcal{U}$ be an arbitrary adversary and $\mathbb{A}$ an arbitrary fixed (but possibly randomized) attack algorithm. Let $\mathcal{D}$ be an arbitrary distribution over $\mathcal{X} \times \mathcal{Y}$ that is robustly realizable,i.e. $\inf_{h\in\mathcal{H}} \mathrm{R}_{\mathcal{U}}(h;\mathcal{D}) = 0$.

Fix $\varepsilon, \delta \in (0,1)$ and a sample size $m = 2\frac{\mathrm{lit}(\mathcal{H})}{\varepsilon}\log\left(\frac{\mathrm{lit}(\mathcal{H})}{\delta}\right)$. Since online learner $\mathbb{A}$ has a mistake bound of $\mathrm{lit}(\mathcal{H})$, Algorithm E.6 will terminate in at most $\frac{\mathrm{lit}(\mathcal{H})}{\varepsilon}\log\left(\frac{\mathrm{lit}(\mathcal{H})}{\delta}\right)$ steps, which is an upperbound on the number of calls to the attack algorithm $\mathbb{A}$. It remains to show that the output of Algorithm E.6, the final predictor $\hat{h}$, will have low error w.r.t. future attacks from $\mathbb{A}$:

$$\mathrm{err}_{\mathbb{A}}(\hat{h}; \mathcal{D}) \triangleq \Pr_{\substack{(x,y)\sim\mathcal{D} \\ \text{randomness of } \mathbb{A}}} \left[ \hat{h}(\mathbb{A}(\hat{h}, (x, y))) \neq y \right].$$

Throughout the runtime of Algorithm E.6, the online learner $\mathbb{A}$ generates a sequence of at most $\mathrm{lit}(\mathcal{H}) + 1$ predictors. There's the initial predictor from Step 1, plus the $\mathrm{lit}(\mathcal{H})$ updated predictors corresponding to potential updates by online learner $\mathbb{A}$. By a union bound over these predictors, the probability that the final predictor $\hat{h}$ has error more than $\varepsilon$

$$\Pr_{S\sim\mathcal{D}^m}\left[\mathrm{err}_{\mathbb{A}}(\hat{h}; \mathcal{D}) > \varepsilon\right] \leq \Pr_{S\sim\mathcal{D}^m}\left[\exists j \in [\mathrm{lit}(\mathcal{H})+1] : \mathrm{err}_{\mathbb{A}}(h_j; \mathcal{D}) > \varepsilon\right] \leq (\mathrm{lit}(\mathcal{H})+1)(1-\varepsilon)^{\frac{1}{\varepsilon}\log\left(\frac{\mathrm{lit}(\mathcal{H})+1}{\delta}\right)} \leq \delta.$$

Therefore, with probability at least $1 - \delta$ over $S \sim \mathcal{D}^m$, Algorithm E.6 outputs a predictor $\hat{h}$ with error $\mathrm{err}_{\mathbb{A}}(\hat{h}; \mathcal{D}) \leq \varepsilon$. □

# F

# Computationally Efficient Robust Learning

## F.1 Auxiliary Proofs for Robust Learning under Random Classification Noise

*Proof of Lemma 8.18.* Based on the definition of the surrogate loss, it suffices to consider three cases:

**Case** $z > \gamma$ :

$$\begin{aligned}
l_1(z) &\stackrel{\text{def}}{=} \eta\varphi(-z) + (1-\eta)\varphi(z) \\
&= \eta(1-\lambda)(1+z/\gamma) + (1-\eta)\lambda(1-z/\gamma) \\
&= (\eta(1-\lambda) - (1-\eta)\lambda)\left(\frac{z}{\gamma}\right) + (1-\eta)\lambda \\
&+ \eta(1-\lambda) \\
&= (\eta-\lambda)\left(\frac{z}{\gamma}\right) + \lambda + \eta - 2\lambda\eta.
\end{aligned}$$

**Case** $-\gamma \leq z \leq \gamma$ :

$$\begin{aligned} l_2(z) &\stackrel{\text{def}}{=} \eta\varphi(-z) + (1-\eta)\varphi(z) \\ &= \eta(1-\lambda)(1+z/\gamma) \\ &+ (1-\eta)(1-\lambda)(1-z/\gamma) \\ &= -(1-2\eta)(1-\lambda)\left(\frac{z}{\gamma}\right) + 1 - \lambda. \end{aligned}$$

**Case** $z < -\gamma$ :

$$\begin{aligned} l_3(z) &\stackrel{\text{def}}{=} \eta\lambda(1+z/\gamma) + (1-\eta)(1-\lambda)(1-z/\gamma) \\ &= (\eta\lambda - (1-\eta)(1-\lambda))\left(\frac{z}{\gamma}\right) + \eta\lambda \\ &+ (1-\eta)(1-\lambda) \\ &= (\eta + \lambda - 1)\left(\frac{z}{\gamma}\right) + 1 - \eta - \lambda + 2\eta\lambda. \end{aligned}$$

Considering the three cases above, we can write

$$\begin{aligned} G(z) = l_1(z) &+ 1\{-\gamma \leq z \leq \gamma\}\left(l_2(z) - l_1(z)\right) \\ &+ 1\{z < -\gamma\}\left(l_3(z) - l_1(z)\right) \; . \end{aligned}$$

Then, we calculate $l_2(z) - l_1(z)$ and $l_3(z) - l_1(z)$ as follows:

$$\begin{aligned}
l_2(z) - l_1(z) &= -(1-2\eta)(1-\lambda)\left(\frac{z}{\gamma}\right) + 1 - \lambda \\
&- (\eta - \lambda)\left(\frac{z}{\gamma}\right) - \lambda - \eta + 2\lambda\eta \\
&= -((1-2\eta)(1-\lambda) + \eta - \lambda)\left(\frac{z}{\gamma}\right) \\
&+ (1-\eta)(1-2\lambda) \\
&= (1-\eta)(1-2\lambda)\left(1 - \frac{z}{\gamma}\right) \text{ , and} \\
l_3(z) - l_1(z) &= (\eta + \lambda - 1)\left(\frac{z}{\gamma}\right) + 1 - \eta - \lambda + 2\eta\lambda \\
&- (\eta - \lambda)\left(\frac{z}{\gamma}\right) - \lambda - \eta + 2\lambda\eta \\
&= (2\lambda - 1)\left(\frac{z}{\gamma}\right) + (1-2\eta)(1-2\lambda) \\
&= (2\lambda - 1)\left(\frac{z}{\gamma} + 2\eta - 1\right).
\end{aligned}$$

Using the above, we have that

$$\begin{aligned}
G(z) = l_1(z) &+ 1\{-\gamma \leq z \leq \gamma\}(1-\eta)(1-2\lambda)\left(1 - \frac{z}{\gamma}\right) \\
&+ 1\{z < -\gamma\}(1-2\lambda)\left(1 - 2\eta - \frac{z}{\gamma}\right).
\end{aligned}$$

□

*Proof of Lemma 8.19.* By Theorem 8.18, and linearity of expectation, we have that

$$\begin{aligned}\mathop{\mathbb{E}}_{\boldsymbol{x}\sim\mathcal{D}_{\boldsymbol{x}}}[G_\lambda^\gamma(\boldsymbol{w},\boldsymbol{x})] &= (\eta-\lambda)\mathop{\mathbb{E}}_{\boldsymbol{x}\sim\mathcal{D}_{\boldsymbol{x}}}\left[\frac{z}{\gamma}\right] + \lambda + \eta - 2\lambda\eta \\ &+ (1-\eta)(1-2\lambda)\mathop{\mathbb{E}}_{\boldsymbol{x}\sim\mathcal{D}_{\boldsymbol{x}}}\left[1\{-\gamma\le z\le\gamma\}\left(1-\frac{z}{\gamma}\right)\right] \\ &+ (1-2\lambda)\mathop{\mathbb{E}}_{\boldsymbol{x}\sim\mathcal{D}_{\boldsymbol{x}}}\left[1\{z<-\gamma\}\left(1-2\eta-\frac{z}{\gamma}\right)\right] \,.\end{aligned}$$

First, observe that for any $\boldsymbol{x}$, $-1\le z = h_{\boldsymbol{w}^*}(\boldsymbol{x})\langle\boldsymbol{w},\boldsymbol{x}\rangle\le 1$ and since $\eta\le\lambda$, we have

$$(\eta-\lambda)\mathop{\mathbb{E}}_{\boldsymbol{x}\sim\mathcal{D}_{\boldsymbol{x}}}\left[\frac{z}{\gamma}\right]\ge\frac{\eta-\lambda}{\gamma}\,.$$

Then we observe that whenever $z<-\gamma, 1-\frac{z}{\gamma}-2\eta>2(1-\eta)>(1-\eta)/2$ and $\lambda\le 1/2$, thus we can bound from below the third term

$$\begin{aligned}(1-2\lambda)\mathop{\mathbb{E}}_{\boldsymbol{x}\sim\mathcal{D}_{\boldsymbol{x}}}\left[1\{z<-\gamma\}\left(1-2\eta-\frac{z}{\gamma}\right)\right]\ge \\ \frac{1}{2}(1-2\lambda)(1-\eta)\mathop{\mathbb{E}}_{\boldsymbol{x}\sim\mathcal{D}_{\boldsymbol{x}}}[1\{z<-\gamma\}]\,.\end{aligned}$$

Next we note that whenever $-\gamma\le z\le\gamma, 1-\frac{z}{\gamma}\ge 0$. This implies that instead of considering $1\{-\gamma\le z\le\gamma\}$, we can relax this and consider the subset $1\left\{-\gamma\le z\le\frac{\gamma}{2}\right\}$, and on this subset $1-\frac{z}{\gamma}\ge 1/2$. Thus, we can bound the second term from below as follows:

$$\begin{aligned}(1-2\lambda)(1-\eta)\mathop{\mathbb{E}}_{\boldsymbol{x}\sim\mathcal{D}_{\boldsymbol{x}}}\left[1\{-\gamma\le z\le\gamma\}\left(1-\frac{z}{\gamma}\right)\right]\ge \\ \frac{1}{2}(1-2\lambda)(1-\eta)\mathop{\mathbb{E}}_{\boldsymbol{x}\sim\mathcal{D}_{\boldsymbol{x}}}\left[1\left\{-\gamma\le z\le\frac{\gamma}{2}\right\}\right]\,.\end{aligned}$$

Combining the above, we obtain

$$\begin{aligned}
&\mathop{\mathbb{E}}_{\boldsymbol{x}\sim\mathcal{D}_{\boldsymbol{x}}}[G_\lambda^\gamma(\boldsymbol{w},\boldsymbol{x})] \geq \frac{\eta-\lambda}{\gamma} \\
&+\frac{1}{2}(1-2\lambda)(1-\eta)\mathop{\mathbb{E}}_{x}\left[1\left\{-\gamma \leq z \leq \frac{\gamma}{2}\right\}+1\{z<-\gamma\}\right] \\
&+\lambda+\eta-2\lambda\eta \\
&\geq \frac{\eta-\lambda}{\gamma}+\frac{1}{2}(1-2\lambda)(1-\eta)\mathop{\mathbb{E}}_{x}\left[1\left\{z \leq \frac{\gamma}{2}\right\}\right] \\
&+\lambda+\eta-2\lambda\eta\ ,
\end{aligned}$$

as desired. □

*Proof of Lemma 8.20.* By definition, we have that $\inf_{\boldsymbol{w}\in\mathbb{R}^d}\mathbb{E}_{\boldsymbol{x}\sim\mathcal{D}_{\boldsymbol{x}}}[G_\lambda^\gamma(\boldsymbol{w},\boldsymbol{x})] \leq \mathbb{E}_{\boldsymbol{x}\sim\mathcal{D}_{\boldsymbol{x}}}[G_\lambda^\gamma(\boldsymbol{w}^*,\boldsymbol{x})]$. By assumption, with probability 1 over $\boldsymbol{x}\sim\mathcal{D}_{\boldsymbol{x}}$, we have $h_{\boldsymbol{w}^*}(\boldsymbol{x})\langle\boldsymbol{w}^*,\boldsymbol{x}\rangle > \gamma$. Thus, by Theorem 8.18, we have

$$\begin{aligned}
\mathop{\mathbb{E}}_{\boldsymbol{x}}[G_\lambda^\gamma(\boldsymbol{w}^*,\boldsymbol{x})] &= (\eta-\lambda)\left(\frac{z}{\gamma}\right)+\lambda+\eta-2\lambda\eta \\
&\leq \eta-\lambda+\lambda+\eta-2\lambda\eta \\
&= 2\eta(1-\lambda)\ ,
\end{aligned}$$

where the last inequality follows from the fact that $\eta<\lambda$ and the fact that $\mathbb{E}_{\boldsymbol{x}}\left[\frac{z}{\gamma}\right]>1$. □

## F.2 Another Approach to Large Margin Learning under Random Classification Noise

We remark that $\gamma$-margin learning of halfspaces has been studied in earlier work, and we have algorithms such as Margin Perceptron Balcan et al. (2008) and SVM. The Margin Perceptron (for $\ell_2$ margin) and other $\ell_p$ margin algorithms have been also implemented in the SQ model Feldman et al. (2017). But no explicit connection has been made to adversarial robustness.

We present here a simple approach to learn $\gamma$-margin halfspaces under random classification noise using only a convex surrogate loss and Stochastic Mirror Descent. The construction of the convex surrogate is based on learning generalized linear models with a suitable link function $u : \mathbb{R} \to \mathbb{R}$. To the best of our knowledge, the result of this section is not explicit in prior work.

**Theorem F.1.** *Let* $\mathcal{X} = \left\{\boldsymbol{x} \in \mathbb{R}^d : \|\boldsymbol{x}\|_p \leq 1\right\}$. *Let* $\mathcal{D}$ *be a distribution over* $\mathcal{X} \times \mathcal{Y}$ *such that there exists a halfspace* $\boldsymbol{w}^* \in \mathbb{R}^d$, $\|\boldsymbol{w}^*\|_q = 1$ *with* $\mathbf{Pr}_{\boldsymbol{x}\sim\mathcal{D}_{\boldsymbol{x}}}\left[\left|\langle \boldsymbol{w}^*, \boldsymbol{x}\rangle\right| > \gamma\right] = 1$ *and* $y$ *is generated by* $h_{\boldsymbol{w}^*}(\boldsymbol{x}) := \operatorname{sign}(\langle \boldsymbol{w}^*, \boldsymbol{x}\rangle)$ *corrupted with random classification noise rate* $\eta < 1/2$. *Then, running Stochastic Mirror Descent on the following convex optimization problem:*

$$\min_{\boldsymbol{w}\in\mathbb{R}^d, \|\boldsymbol{w}\|_q \leq 1} \mathop{\mathbb{E}}_{(\boldsymbol{x},y)\sim\mathcal{D}} \left[\ell(\boldsymbol{w}, (\boldsymbol{x}, y))\right]$$

*where the convex loss function* $\ell$ *is defined in Equation F.1, returns with high probability, a halfspace* $\boldsymbol{w}$ *with* $\gamma/2$*-robust misclassification error* $\mathbb{E}_{(\boldsymbol{x},y)\sim\mathcal{D}}\left[1\{y\langle \boldsymbol{w}, \boldsymbol{x}\rangle \leq \gamma/2\}\right] \leq \eta + \varepsilon$.

We prove Theorem F.1 in the remainder of this section. We will connect our problem to that of solving generalized linear models. We define the link function as follows,

$$u(s) = \begin{cases} \eta & s < -\gamma \\ \frac{1-2\eta}{2\gamma}s + \frac{1}{2} & -\gamma \leq s \leq \gamma \\ 1-\eta & s > \gamma \end{cases}.$$

Observe that $u$ is monotone and $\frac{1-2\eta}{2\gamma}$-Lipschitz.

First, we will relate our loss of interest, which is the $\gamma/2$-margin loss with the squared loss defined in terms of the link function $u$,

**Lemma F.2.** *For any* $\boldsymbol{w} \in \mathbb{R}^d$,

$$\mathop{\mathbb{E}}_{\boldsymbol{x}\sim\mathcal{D}_{\boldsymbol{x}}}\left[1\{h_{\boldsymbol{w}^*}(\boldsymbol{x})\langle \boldsymbol{w}, \boldsymbol{x}\rangle \leq \gamma/2\}\right] \leq \frac{16}{(1-2\eta)^2}\mathop{\mathbb{E}}_{\boldsymbol{x}\sim\mathcal{D}_{\boldsymbol{x}}}\left[\left(u(\langle \boldsymbol{w}, \boldsymbol{x}\rangle) - u(\langle \boldsymbol{w}^*, \boldsymbol{x}\rangle)\right)^2\right].$$

*Proof.* Let $E^+ = \{h_{\boldsymbol{w}^*}(\boldsymbol{x}) = +\}$ and $E^- = \{h_{\boldsymbol{w}^*}(\boldsymbol{x}) = -\}$. Let $U(\boldsymbol{x}) = (u(\langle \boldsymbol{w}, \boldsymbol{x}\rangle) - u(\langle \boldsymbol{w}^*, \boldsymbol{x}\rangle))^2$.

By law of total expectation and definition of the link function $u$, we have

$$\begin{aligned}
&\mathop{\mathbb{E}}_{\boldsymbol{x}\sim\mathcal{D}_{\boldsymbol{x}}}[U(\boldsymbol{x})] = \mathop{\mathbb{E}}_{\boldsymbol{x}\sim\mathcal{D}_{\boldsymbol{x}}}\left[U(\boldsymbol{x})1\{E^+\}\right] + \mathop{\mathbb{E}}_{\boldsymbol{x}\sim\mathcal{D}_{\boldsymbol{x}}}\left[U(\boldsymbol{x})1\{E^-\}\right] \\
&= \mathop{\mathbb{E}}_{\boldsymbol{x}\sim\mathcal{D}_{\boldsymbol{x}}}\left[(u(\langle \boldsymbol{w},\boldsymbol{x}\rangle) - (1-\eta))^2 1\{E^+\}\right] \\
&+ \mathop{\mathbb{E}}_{\boldsymbol{x}\sim\mathcal{D}_{\boldsymbol{x}}}\left[(u(\langle \boldsymbol{w},\boldsymbol{x}\rangle) - \eta)^2 1\{E^-\}\right].
\end{aligned}$$

We will lower bound both terms:

$$\begin{aligned}
&\mathop{\mathbb{E}}_{\boldsymbol{x}\sim\mathcal{D}_{\boldsymbol{x}}}\left[(u(\langle \boldsymbol{w},\boldsymbol{x}\rangle) - (1-\eta))^2 1\{E^+\}\right] \\
&\geq \mathop{\mathbb{E}}_{\boldsymbol{x}\sim\mathcal{D}_{\boldsymbol{x}}}\left[(a - (1-\eta))^2 1\{E^+\}1\{u(\langle \boldsymbol{w},\boldsymbol{x}\rangle) \leq a\}\right] \ , \\
&\mathop{\mathbb{E}}_{\boldsymbol{x}\sim\mathcal{D}_{\boldsymbol{x}}}\left[(u(\langle \boldsymbol{w},\boldsymbol{x}\rangle) - \eta)^2 1\{E^-\}\right] \\
&\geq \mathop{\mathbb{E}}_{\boldsymbol{x}\sim\mathcal{D}_{\boldsymbol{x}}}\left[(b - \eta)^2 1\{E^-\}1\{u(\langle \boldsymbol{w},\boldsymbol{x}\rangle) \geq b\}\right] \ .
\end{aligned}$$

Then, observe that the event $\{\langle \boldsymbol{w},\boldsymbol{x}\rangle \leq \gamma/2\}$ implies the event $\left\{u(\langle \boldsymbol{w},\boldsymbol{x}\rangle) \leq \frac{3-2\eta}{4}\right\}$, and similarly the event $\{\langle \boldsymbol{w},\boldsymbol{x}\rangle \geq -\gamma/2\}$ implies the event $\left\{u(\langle \boldsymbol{w},\boldsymbol{x}\rangle) \geq \frac{2\eta+1}{4}\right\}$. This means that

$$\begin{aligned}
&\mathop{\mathbb{E}}_{\boldsymbol{x}\sim\mathcal{D}_{\boldsymbol{x}}}\left[\left(\frac{3-2\eta}{4} - (1-\eta)\right)^2 1\{E^+\}1\left\{u(\langle \boldsymbol{w},\boldsymbol{x}\rangle) \leq \frac{3-2\eta}{4}\right\}\right] \\
&\geq \left(\frac{3-2\eta}{4} - (1-\eta)\right)^2 \mathop{\mathbb{E}}_{\boldsymbol{x}\sim\mathcal{D}_{\boldsymbol{x}}}\left[1\{E^+\}1\{\langle \boldsymbol{w},\boldsymbol{x}\rangle \leq \gamma/2\}\right] \ , \text{ and} \\
&\mathop{\mathbb{E}}_{\boldsymbol{x}\sim\mathcal{D}_{\boldsymbol{x}}}\left[\left(\frac{2\eta+1}{4} - \eta\right)^2 1\{E^-\}1\left\{u(\langle \boldsymbol{w},\boldsymbol{x}\rangle) \geq \frac{2\eta+1}{4}\right\}\right] \\
&\geq \left(\frac{2\eta+1}{4} - \eta\right)^2 \mathop{\mathbb{E}}_{\boldsymbol{x}\sim\mathcal{D}_{\boldsymbol{x}}}\left[1\{E^-\}1\{\langle \boldsymbol{w},\boldsymbol{x}\rangle \geq -\gamma/2\}\right] \ .
\end{aligned}$$

We combine these observations to conclude the proof,

$$\begin{aligned}
&\mathop{\mathbb{E}}_{\boldsymbol{x}\sim\mathcal{D}_{\boldsymbol{x}}}\left[(u(\langle \boldsymbol{w},\boldsymbol{x}\rangle) - u(\langle \boldsymbol{w}^*,\boldsymbol{x}\rangle))^2\right] \geq \frac{(1-2\eta)^2}{16}\times \\
&\mathop{\mathbb{E}}_{\boldsymbol{x}\sim\mathcal{D}_{\boldsymbol{x}}}\left[1(E^+)1\{\langle \boldsymbol{w},\boldsymbol{x}\rangle \leq \gamma/2\} + 1(E^-)1\{\langle \boldsymbol{w},\boldsymbol{x}\rangle \geq -\gamma/2\}\right] \\
&\geq \frac{(1-2\eta)^2}{16} \mathop{\mathbb{E}}_{\boldsymbol{x}\sim\mathcal{D}_{\boldsymbol{x}}}\left[1\{h_{\boldsymbol{w}^*}(\boldsymbol{x})\langle \boldsymbol{w},\boldsymbol{x}\rangle \leq \gamma/2\}\right].
\end{aligned}$$

□

But note that the squared loss is non-convex and so it may not be easy to optimize. Luckily, we can get a tight upper-bound with the following surrogate loss (see Kanade (2018)):

$$\ell(\boldsymbol{w}, (\boldsymbol{x}, y)) = \int_0^{\langle \boldsymbol{w}, \boldsymbol{x} \rangle} (u(s) - y) ds. \tag{F.1}$$

Note that $\ell(\boldsymbol{w}, (\boldsymbol{x}, y))$ is convex w.r.t $\boldsymbol{w}$ since the Hessian $\nabla^2_{\boldsymbol{w}} \ell(\boldsymbol{w}, (\boldsymbol{x}, y)) = u'(\langle \boldsymbol{w}, \boldsymbol{x} \rangle) \boldsymbol{x}\boldsymbol{x}^T$ is positive semi-definite.

Assuming our labels $y$ have been transformed to $\{0, 1\}$ from $\{\pm 1\}$, observe that $\mathbb{E}[y|x] = u(\langle \boldsymbol{w}^*, \boldsymbol{x} \rangle)$. We now have the following guarantee (see e.g., Cohen, 2014, Kanade, 2018) for $U(\boldsymbol{x}) = (u(\langle \boldsymbol{w}, \boldsymbol{x} \rangle) - u(\langle \boldsymbol{w}^*, \boldsymbol{x} \rangle))^2$:

$$\mathop{\mathbb{E}}_{\boldsymbol{x} \sim \mathcal{D}_{\boldsymbol{x}}} [U(\boldsymbol{x})] \leq \frac{1 - 2\eta}{\gamma} \mathop{\mathbb{E}}_{(\boldsymbol{x}, y) \sim \mathcal{D}} [\ell(\boldsymbol{w}, (\boldsymbol{x}, y)) - \ell(\boldsymbol{w}^*, (\boldsymbol{x}, y))] . \tag{F.2}$$

*Proof of Theorem F.1.* Combining Theorem F.2 and Equation F.2, we get the following guarantee for any $\boldsymbol{w} \in \mathbb{R}^d$,

$$\begin{aligned}
&(1 - 2\eta) \mathop{\mathbb{E}}_{\boldsymbol{x} \sim \mathcal{D}_{\boldsymbol{x}}} [1\{h_{\boldsymbol{w}^*}(\boldsymbol{x}) \langle \boldsymbol{w}, \boldsymbol{x} \rangle \leq \gamma/2\}] \leq \\
&\frac{16}{\gamma} \mathop{\mathbb{E}}_{(\boldsymbol{x}, y) \sim \mathcal{D}} [\ell(\boldsymbol{w}, (\boldsymbol{x}, y)) - \ell(\boldsymbol{w}^*, (\boldsymbol{x}, y))] .
\end{aligned}$$

Thus, running Stochastic Mirror Descent with $\varepsilon' = (\varepsilon\gamma(1 - 2\eta))/16$ and $O(1/\varepsilon'^2)^*$ samples (labels transformed to $\{0, 1\}$), returns with high probability, a halfspace $\boldsymbol{w}$ such that

$$\mathop{\mathbb{E}}_{\boldsymbol{x} \sim \mathcal{D}_{\boldsymbol{x}}} [1\{h_{\boldsymbol{w}^*}(\boldsymbol{x}) \langle \boldsymbol{w}, \boldsymbol{x} \rangle \leq \gamma/2\}] \leq \varepsilon. \tag{F.3}$$

*Note that this sample complexity hides dependence on the Lipschitz and smoothness paramteres of the surrogate loss $\ell$, which may involve dependence on $\gamma$ and $\eta$.

Then, to conclude the proof, observe that

$$\begin{aligned}
&\mathop{\mathbb{E}}_{(\boldsymbol{x},y)\sim\mathcal{D}}[1\{y\langle\boldsymbol{w},\boldsymbol{x}\rangle\leq\gamma/2\}]=\eta\mathop{\mathbb{E}}_{\boldsymbol{x}\sim\mathcal{D}_{\boldsymbol{x}}}[1\{-h_{\boldsymbol{w}^*}(\boldsymbol{x})\langle\boldsymbol{w},\boldsymbol{x}\rangle\leq\gamma/2\}]\\
&+(1-\eta)\mathop{\mathbb{E}}_{\boldsymbol{x}\sim\mathcal{D}_{\boldsymbol{x}}}[1\{h_{\boldsymbol{w}^*}(\boldsymbol{x})\langle\boldsymbol{w},\boldsymbol{x}\rangle\leq\gamma/2\}]\\
&=\eta(1-\mathop{\mathbb{E}}_{\boldsymbol{x}\sim\mathcal{D}_{\boldsymbol{x}}}[1\{h_{\boldsymbol{w}^*}(\boldsymbol{x})\langle\boldsymbol{w},\boldsymbol{x}\rangle\leq-\gamma/2\}])\\
&+(1-\eta)\mathop{\mathbb{E}}_{\boldsymbol{x}\sim\mathcal{D}_{\boldsymbol{x}}}[1\{h_{\boldsymbol{w}^*}(\boldsymbol{x})\langle\boldsymbol{w},\boldsymbol{x}\rangle\leq\gamma/2\}]\\
&\overset{(i)}{\leq}\eta+(1-\eta)\mathop{\mathbb{E}}_{\boldsymbol{x}\sim\mathcal{D}_{\boldsymbol{x}}}[1\{h_{\boldsymbol{w}^*}(\boldsymbol{x})\langle\boldsymbol{w},\boldsymbol{x}\rangle\leq\gamma/2\}]\\
&\overset{(ii)}{\leq}\eta+\varepsilon,
\end{aligned}$$

where $(i)$ follows from that fact that $\mathbb{E}_{\boldsymbol{x}\sim\mathcal{D}_{\boldsymbol{x}}}[1\{h_{\boldsymbol{w}^*}(\boldsymbol{x})\langle\boldsymbol{w},\boldsymbol{x}\rangle\leq-\gamma/2\}]\geq 0$, and $(ii)$ follows from Equation F.3. □

# G Beyond Perturbations: Learning Guarantees with Arbitrary Adversarial Test Examples

## G.1 Rejectron Analysis (Realizable)

In this section, we present the analysis of Rejectron in the realizable case $f \in \mathcal{H}$. Say a classifier $c$ is *consistent* if $c(\boldsymbol{x}) = f(\boldsymbol{x})$ makes 0 training errors. Theorem 10.6 provides transductive guarantees on the empirical error and rejection rates, while Theorem 10.5 provides generalization guarantees that apply to future examples from $P, Q$. Both of these theorems exhibit trade-offs between error and rejection rates. At a high level, their analysis has the following structure:

- Rejectron selects a consistent $h = \mathsf{ERM}(\boldsymbol{x}, f(\boldsymbol{x}))$, since we are in the realizable case.
- Each $c_t$ is a consistent classifier that disagrees with $h|_{S_t}$ on the tests $\tilde{\boldsymbol{x}}$ as much as possible, with $s_t(c_t) = \text{err}_{\tilde{\boldsymbol{x}}}(h|_{S_t}, c_t)$ (since $\text{err}_{\boldsymbol{x}}(h, c_t) = 0$). This follows the facts that $\Lambda > n$, $s_t(h) = 0$, and $s_t(c) < 0$ for any inconsistent $c$. (The algorithm is defined for general $\Lambda < n$ for the agnostic analysis later.)

- Therefore, when the algorithm terminates on iteration $T$, it has empirical test error $\text{err}_{\tilde{\boldsymbol{x}}}(h|_{S_T}, f) \le \varepsilon$ otherwise it could have chosen $c_t = f$.

- The number of iterations $T < 1/\varepsilon$ since on each iteration an additional $\varepsilon$ fraction of $\tilde{\boldsymbol{x}}$ is removed from $S_t$. Lemma 10.4 states this and shows how to use an ERM oracle on an artificial dataset to efficiently find $c_t$.

- All training examples $x_i$ are in $S$ since each $c_t$ and $h$ agree on all $x_i$.

- Transductive error and rejection bounds:

  1. For error, we have already argued that the empirical error $\text{err}_{\tilde{\boldsymbol{x}}} \le \varepsilon$.
  2. For rejection, Lemma G.1 states that it is unlikely that there would be any choice of $h, \boldsymbol{c} = (c_1, \ldots, c_T)$ where the resulting $S(h, \boldsymbol{c}) := \{x \in \mathcal{X} : h(x) = c_1(x) = \ldots = c_T(x)\}$ would contain all training examples but reject (abstain on) many "true" test examples $z_i$ since $\boldsymbol{x}$ and $\boldsymbol{z}$ are identically distributed. The proof uses Sauer's lemma.

- Generalization error and rejection bounds:

  1. For error, Lemma G.3 states that it is unlikely that there is any $h, \boldsymbol{c}$ such that $\text{err}_{\tilde{\boldsymbol{x}}}(h|_{S(h,\boldsymbol{c})}) \le \varepsilon$ yet $\text{err}_Q(h|_{S(h,\boldsymbol{c})}) > 2\varepsilon$.
  2. For rejection rate, Lemma G.4 uses VC bounds to show that it is unlikely that $\text{rej}_P(S(h, \boldsymbol{c})) > \varepsilon$ while $\text{rej}_{\boldsymbol{x}}(S(h, \boldsymbol{c})) = 0$.

  Both proofs use Sauer's lemma.

We next move to the transductive analysis since it is simpler, and it is also used as a stepping stone to the generalization analysis.

### G.1.1 Transductive Guarantees (Realizable)

Note that Rejectron rejects any $x \notin S$, where $S = S(h, \boldsymbol{c})$ is defined by

$$S(h, \boldsymbol{c}) := \{x \in \mathcal{X} : h(x) = c_1(x) = c_2(x) = \ldots = c_T(x)\}. \tag{G.1}$$

In what follows, we prove the transductive analogue of a "generalization" guarantee for arbitrary $h \in \mathcal{H}, \boldsymbol{c} \in \mathcal{H}^T$. This will be useful when proving Theorem 10.6.

**Lemma G.1.** *For any $T, n \in \mathbb{N}$, any $\delta \geq 0$, and $\varepsilon = \frac{1}{n}\left(d(T+1)\log(2n) + \log\frac{1}{\delta}\right)$:*

$$\mathbf{Pr}_{\boldsymbol{x},\boldsymbol{z}\sim P^n}\left[\exists \boldsymbol{c} \in \mathcal{H}^T, h \in \mathcal{H} : \ (\mathrm{rej}_{\boldsymbol{x}}(S(h,\boldsymbol{c})) = 0) \wedge (\mathrm{rej}_{\boldsymbol{z}}(S(h,\boldsymbol{c})) > \varepsilon)\right] \leq \delta.$$

This lemma is proven in Section G.5. Using it, we can show a trade-off between error and rejection rate for the transductive case.

**Theorem G.2.** *For any $n \in \mathbb{N}$, any $\varepsilon, \delta \geq 0$, any $f \in \mathcal{H}$:*

$$\forall \boldsymbol{x}, \tilde{\boldsymbol{x}} \in \mathcal{X}^n : \ \mathrm{err}_{\tilde{\boldsymbol{x}}}(h|_S, f) \leq \varepsilon, \tag{G.2}$$

*where $h|_S = \mathsf{Rejectron}(\boldsymbol{x}, f(\boldsymbol{x}), \tilde{\boldsymbol{x}}, \varepsilon)$, and for any distribution $P$ over $\mathcal{X}$,*

$$\mathbf{Pr}_{\boldsymbol{x},\boldsymbol{z}\sim P^n}\left[\forall \tilde{\boldsymbol{x}} \in \mathcal{X}^n : \ \mathrm{rej}_{\boldsymbol{z}}(S) \leq \frac{1}{n}\left(\frac{2d}{\varepsilon}\log(2n) + \log\frac{1}{\delta}\right)\right] \geq 1 - \delta. \tag{G.3}$$

We note that a natural alternative formalization of Equation (G.3) would be to require that

$$\mathbf{Pr}_{\boldsymbol{x}\sim P^n}\left[\forall \tilde{\boldsymbol{x}} \in \mathcal{X}^n : \ \mathrm{rej}_P(S) \leq \frac{1}{n}\left(\frac{2d}{\varepsilon}\log(2n) + \log\frac{1}{\delta}\right)\right] \geq 1 - \delta.$$

However, the formalization of Equation (G.3) is stronger, as it guarantees that the rejection probability is small, even if the adversary is *"white-box"* and chooses $\tilde{\mathbf{x}}$ after seeing $\mathbf{z}$.

*Proof of Theorem G.2.* We start by proving eq. (G.2). To this end, fix any $n \in \mathbb{N}$, any $\varepsilon > 0$, any $f \in \mathcal{H}$, and any $\boldsymbol{x}, \tilde{\boldsymbol{x}} \in \mathcal{X}^n$. Let $h = \mathsf{ERM}(\boldsymbol{x}, f(\boldsymbol{x}))$. Since we are in the realizable case, this implies that $h$ has zero training error, i.e., $\mathrm{err}_{\boldsymbol{x}}(h, f) = 0$, and hence $s_t(h) = \mathrm{err}_{\tilde{\boldsymbol{x}}}(h|_{S_t}, f)$ for all $t$. Thus, the algorithm cannot terminate on any iteration where $\mathrm{err}_{\tilde{\boldsymbol{x}}}(h|_{S_t}, f) > \varepsilon$ since it can always select $c_t = f \in \mathcal{H}$. This proves Equation (G.2).

It remains to prove eq. (G.3). By Lemma 10.4, $T = \lfloor 1/\varepsilon \rfloor$ is an upper bound on the number of completed iterations of the algorithm. WLOG there are exactly $T$ iterations because if there were actually $T' < T$ iterations, simply "pad" them with $c_{T'+1} = \ldots = c_T = h$ which doesn't change $S$.

We note that the algorithm selects all training examples. This follows from the fact that $\Lambda > n$, together with the fact that $h(x_i) = f(x_i)$ for every $i \in [n]$, where the latter follows from the fact that $f \in \mathcal{H}$ and $h = \mathsf{ERM}(\boldsymbol{x}, f(\boldsymbol{x}))$. By Lemma G.1, with probability $\geq 1 - \delta$ there are no choices $h \in \mathcal{H}, \boldsymbol{c} = (c_1, \ldots, c_T) \in \mathcal{H}^T$ for which $S(h, \boldsymbol{c})$ contains all $x_i$'s but is missing $\geq \varepsilon'$ fraction of $\boldsymbol{z}$ for $\varepsilon' = \frac{1}{n}\left(\frac{2d}{\varepsilon}\log(2n) + \log\frac{1}{\delta}\right)$ since $T + 1 \leq 2/\varepsilon$. □

Theorem 10.6 is a trivial corollary of Theorem G.2.

*Proof of Theorem 10.6.* Recall $\varepsilon^* = \sqrt{\frac{2d}{n}\log 2n} + \frac{1}{n}\log\frac{1}{\delta}$ and $h|_S = \mathsf{Rejectron}(\boldsymbol{x}, f(\boldsymbol{x}), \tilde{\boldsymbol{x}}, \varepsilon^*)$. The proof follows from Theorem G.2 and the fact that:

$$\frac{1}{n}\left(\frac{2d}{\varepsilon^*}\log 2n + \log\frac{1}{\delta}\right) \leq \frac{2d\log 2n}{n\sqrt{\frac{2d}{n}\log 2n}} + \frac{1}{n}\log\frac{1}{\delta} = \varepsilon^*.$$

□

### G.1.2 Generalization Guarantees (Realizable)

Before we state our generalization guarantees, analogous to Lemma G.1 above, we prove that low test error and low training rejection rates imply, with high probability, low generalization error and rejection rates.

**Lemma G.3.** *For any $\delta > 0, \varepsilon \geq \frac{8\ln 8/\delta}{n} + \sqrt{\frac{8d\ln 2n}{n}}$, $T \leq 1/\varepsilon$, any $f, h \in \mathcal{H}$ and any distribution $Q$ over $\mathcal{X}$,*

$$\mathbf{Pr}_{\boldsymbol{z}\sim Q^n}\left[\exists \boldsymbol{c} \in \mathcal{H}^T : \left(\mathrm{err}_Q(h|_{S(h,\boldsymbol{c})}, f) > 2\varepsilon\right) \wedge \left(\mathrm{err}_{\boldsymbol{z}}(h|_{S(h,\boldsymbol{c})}, f) \leq \varepsilon\right)\right] \leq \delta.$$

**Lemma G.4.** *For any $T \geq 1$, any $f \in \mathcal{H}$ and any distribution $P$ over $\mathcal{X}$,*

$$\mathbf{Pr}_{\boldsymbol{x}\sim P^n}\left[\exists h \in \mathcal{H}, \boldsymbol{c} \in \mathcal{H}^T : \left(\mathrm{rej}_P(S(h, \boldsymbol{c})) > \xi\right) \wedge \left(\mathrm{rej}_{\boldsymbol{x}}(S(h, \boldsymbol{c})) = 0\right)\right] \leq \delta,$$

*where* $\xi = \frac{2}{n}(d(T+1)\log(2n) + \log\frac{2}{\delta})$. *Also,*

$$\mathbf{Pr}_{\boldsymbol{x}\sim P^n}\left[\exists h \in \mathcal{H}, \boldsymbol{c} \in \mathcal{H}^T : \left(\mathrm{rej}_P(S(h,\boldsymbol{c})) > 2\alpha\right) \wedge \left(\mathrm{rej}_{\boldsymbol{x}}(S(h,\boldsymbol{c})) \leq \alpha\right)\right] \leq \delta,$$

*for any* $\alpha \geq \frac{8}{n}(d(T+1)\ln(2n) + \ln\frac{8}{\delta})$.

We mention that the first inequality in Lemma G.4 is used to provide generalization guarantees in the realizable setting, whereas the latter inequality is used to provide guarantees in the semi-agnostic setting.

**Theorem G.5.** *For any* $n \in \mathbb{N}$ *and* $\delta > 0$, *any* $\varepsilon \geq \sqrt{\frac{8d\ln 2n}{n}} + \frac{8\ln 8/\delta}{n}$, *any* $f \in \mathcal{H}$ *and any distributions* $P, Q$ *over* $\mathcal{X}$*:*

$$\forall \boldsymbol{x} \in \mathcal{X}^n : \ \mathbf{Pr}_{\tilde{\mathrm{x}}\sim Q^n}\left[\mathrm{err}_Q(h|_S) \leq 2\varepsilon\right] \geq 1-\delta, \tag{G.4}$$

*where* $h|_S := \mathsf{Rejectron}(\boldsymbol{x}, f(\boldsymbol{x}), \tilde{\mathrm{x}}, \varepsilon)$. *Furthermore, for any* $\varepsilon \geq 0$,

$$\mathbf{Pr}_{\boldsymbol{x}\sim P^n}\left[\forall \tilde{\mathrm{x}} \in \mathcal{X}^n : \ \mathrm{rej}_P \leq \frac{2}{n}\left(\frac{2d}{\varepsilon}\log 2n + \log\frac{2}{\delta}\right)\right] \geq 1-\delta. \tag{G.5}$$

*Proof of Theorem G.5.* Let $T = \lfloor 1/\varepsilon \rfloor$ be an upper bound on the number of iterations. We first prove eq. (G.4). Since the ERM algorithm is assumed to be deterministic, the function $h$ is uniquely determined by $\boldsymbol{x}$ and $f$. By Theorem G.2 (Equation (G.2)), the set $S$ has the property that $\mathrm{err}_{\tilde{\mathrm{x}}}(h|_S) \leq \varepsilon$ (with certainty) for all $\boldsymbol{x}, \tilde{\mathrm{x}}$. By Lemma G.3, with probability at most $\delta$ there exists a choice of $h, \boldsymbol{c}$ which would lead to $\mathrm{err}_Q(h|_S) > 2\varepsilon$ and $\mathrm{err}_{\tilde{\mathrm{x}}}(h|_S) \leq \varepsilon$, implying eq. (G.4).

For eq. (G.5), as we argued in the proof of Theorem G.2, the fact that $\Lambda > n$, together with the fact we are in the realizable case (i.e., $\boldsymbol{y} = f(\boldsymbol{x})$), implies that we select all training examples. Because of this and the fact that $T+1 \leq 2/\varepsilon$, Lemma G.4 implies eq. (G.5). $\square$

Theorem 10.5 is a trivial corollary of Theorem G.5.

*Proof of Theorem 10.5.* Recall that $\varepsilon^* = \sqrt{\frac{8d\ln 2n}{n}} + \frac{8\ln 16/\delta}{n}$.

Equation (G.4) implies that $\mathbf{Pr}[\text{err}_Q \leq 2\varepsilon^*] \geq 1 - \delta/2$ and eq. (G.5) implies,

$$\mathbf{Pr}_{\boldsymbol{x}\sim P^n}\left[\forall \boldsymbol{z} \in \mathcal{X}^n : \ \text{rej}_P \leq \frac{2}{n}\left(\frac{2d}{\varepsilon^*}\log 2n + \log\frac{4}{\delta}\right)\right] \geq 1 - \frac{\delta}{2}.$$

Further, note that $\log_2 r \leq 2 \ln r$ for $r \geq 1$ and hence, using $\varepsilon^* > \sqrt{\frac{8d \ln 2n}{n}}$,

$$\frac{2}{n}\left(\frac{2d}{\varepsilon^*}\log 2n + \log\frac{4}{\delta}\right) \leq \frac{8d}{n\varepsilon^*}\ln 2n + \frac{4}{n}\ln\frac{4}{\delta} < \sqrt{\frac{8d \ln 2n}{n}} + \frac{4}{n}\ln\frac{4}{\delta} \leq \varepsilon^*.$$

The proof is completed by the union bound. □

## G.2 Analysis of Urejectron

In this section we present a transductive analysis of URejectron, again in the realizable case. We begin with its computational efficiency.

**Lemma G.6** (URejectron computational efficiency). *For any* $\boldsymbol{x}, \tilde{\boldsymbol{x}} \in \mathcal{X}^n$, $\varepsilon > 0$ *and* $\Lambda \in \mathbb{N}$, URejectron *outputs* $S_{T+1}$ *for* $T \leq \lfloor 1/\varepsilon \rfloor$. *Further, each iteration can be implemented using one call to* $\text{ERM}_{\text{DIS}}$, *as defined in eq.* (10.10), *on at most* $(\Lambda + 1)n$ *examples and* $O(n)$ *evaluations of classifiers in* $\mathcal{H}$.

The proof of this lemma is nearly identical to that of Lemma 10.4.

*Proof of Lemma G.6.* The argument that $T \leq \lfloor 1/\varepsilon \rfloor$ follows for the same reason as before, replacing eq. (10.9) with:

$$|\{i : x_i \in S_t\}| - |\{i : x_i \in S_{t+1}\}| = |\{i : x_i \in S_t \wedge c_t(x_i) \neq c'_t(x_i)\}| = n\,\text{err}_{\tilde{\boldsymbol{x}}}(c_t|_{S_t}, c'_t) \geq n\varepsilon.$$

For efficiency, again all that needs to be stored are the subset of indices $Z_t = \{i \mid \tilde{x}_i \in S_t\}$ and the classifiers $c_1, c'_1, \ldots, c_T, c'_T$ necessary to compute $S$. To implement iteration $t$ using the $\text{ERM}_{\text{DIS}}$ oracle, construct a dataset consisting of each training example, labeled by 0, repeated $\Lambda$ times, and each test example in $\tilde{x}_i \in S_t$, labeled 1, included just once. The accuracy of $\text{dis}_{c,c'}$ on this dataset is easily seen to differ by a constant from $s_t(c, c')$, hence running $\text{ERM}_{\text{DIS}}$ maximizes $s_t$. □

The following Theorem exhibits the trade-off between accuracy and rejections.

**Theorem G.7.** *For any $n \in \mathbb{N}$, any $\varepsilon \geq 0$,*

$$\forall \boldsymbol{x}, \tilde{\boldsymbol{x}} \in \mathcal{X}^n, f \in \mathcal{H} : \text{err}_{\tilde{\boldsymbol{x}}}(h|_S) \leq \varepsilon, \tag{G.6}$$

*where $S = \textsf{URejectron}(\boldsymbol{x}, \tilde{\boldsymbol{x}}, \varepsilon)$ and $h = \textsf{ERM}_{\mathcal{H}}(\boldsymbol{x}, f(\boldsymbol{x}))$. Furthermore, for any $\delta > 0$ and any distribution $P$ over $\mathcal{X}$:*

$$\mathbf{Pr}_{\boldsymbol{x},\boldsymbol{z}\sim P^n}\left[\text{rej}_{\boldsymbol{z}}(S) \leq \frac{1}{n}\left(\frac{2d\log 2n}{\varepsilon} + \log 1/\delta\right)\right] \geq 1-\delta. \tag{G.7}$$

Before we prove Theorem G.7 we provide some generalization bounds that will be used in the proof. To this end, given a family $G$ of classifiers $g : \mathcal{X} \to \{0,1\}$, following Blumer et al. (1989b), define:

$$\Pi_G[2n] := \max_{\boldsymbol{w}\in\mathcal{X}^{2n}} |\{g(\boldsymbol{w}) : g \in G\}|. \tag{G.8}$$

**Lemma G.8** (Transductive train-test bounds)**.** *For any $n \in \mathbb{N}$, any distribution $P$ over a domain $\mathcal{X}$, any set $G$ of classifiers over $\mathcal{X}$, and any $\varepsilon > 0$,*

$$\mathbf{Pr}_{\boldsymbol{x},\boldsymbol{z}\sim P^n}\left[\exists g \in G : \left(\frac{1}{n}\sum_i g(z_i) \geq \varepsilon\right) \wedge \left(\frac{1}{n}\sum_i g(x_i) = 0\right)\right] \leq \Pi_G[2n]2^{-\varepsilon n} \tag{G.9}$$

*and*

$$\mathbf{Pr}_{\boldsymbol{x},\boldsymbol{z}\sim P^n}\left[\exists g \in G : \frac{1}{n}\sum_i g(z_i) \geq \frac{1+\alpha}{n}\sum_i g(x_i) + \varepsilon\right] \leq \Pi_G[2n]e^{-\frac{2\alpha}{(2+\alpha)^2}\varepsilon n}. \tag{G.10}$$

The proof of this lemma is deferred to Section G.5. (Note eq. (G.10) is used for the agnostic analysis later.)

*Proof of Theorem G.7.* We denote for $T \geq 1$ and classifier vectors $\boldsymbol{c}, \boldsymbol{c}' \in \mathcal{H}^T$:

$$\partial_{\boldsymbol{c},\boldsymbol{c}'}(x) \coloneqq \max_{i \in [T]} \mathsf{dis}_{c_i,c_i'}(x) = \begin{cases} 1 & \text{if } c_i(x) \neq c_i'(x) \text{ for some } i \in [T] \\ 0 & \text{otherwise.} \end{cases}$$

$$\Delta_T \coloneqq \left\{ \partial_{\boldsymbol{c},\boldsymbol{c}'} : \boldsymbol{c}, \boldsymbol{c}' \in \mathcal{H}^T \right\}.$$

Thus the output of URejectron is $S_{T+1} = \{x \in \mathcal{X} : \partial_{\boldsymbol{c},\boldsymbol{c}'}(x) = 0\}$ for the vectors $\boldsymbol{c} = (c_1, \ldots, c_T)$ and $\boldsymbol{c}' = (c_1', \ldots, c_T')$ chosen by the algorithm.

Let $T$ be the final iteration of the algorithm so that the output of the algorithm is $S = S_{T+1}$. Note that $\text{err}_{\boldsymbol{x}}(f, h) = 0$, by definition of $\mathsf{ERM}_{\mathcal{H}}$, so $s_{T+1}(f, h) = \text{err}_{\tilde{\boldsymbol{x}}}(h|_S) \leq \varepsilon$ (otherwise the algorithm would have chosen $c = h, c' = f$ instead of halting) which implies eq. (G.6).

By Lemma G.6, WLOG we can take $T = \lfloor 1/\varepsilon \rfloor$ by padding with classifiers $c_t = c_t'$.

We next claim that $x_i \notin S_t$ for all $i \in [n]$, i.e., $\partial_{\boldsymbol{c},\boldsymbol{c}'}(x_i) = 0$. This is because the algorithm is run with $\Lambda = n + 1$, so any disagreement $c_t(x_i) \neq c_t'(x_i)$ would result in a negative score $s_t(c_t, c_t')$. (But a zero score is always possible by choosing $c_t = c_t'$.) Thus we must have the property that $\mathsf{dis}_{c_t',c_t}(x_i) = 0$ and hence $\partial_{\boldsymbol{c},\boldsymbol{c}'}(x_i) = 0$. Now, it is not difficult to see that $\Pi_{\Delta_T}[2n] \leq (2n)^{2d/\varepsilon}$ because, by Sauer's lemma, there are at most $N = (2n)^d$ different labelings of $2n$ examples by classifiers from $\mathcal{H}$, hence there are at most $\binom{N}{2}^T \leq (2n)^{2dT}$ disagreement labelings for $T \leq 1/\varepsilon$ pairs. Thus for $\xi = \frac{1}{n}\left(\frac{2d \log 2n}{\varepsilon} + \log 1/\delta\right)$, by Lemma G.8,

$$\mathbf{Pr}_{\boldsymbol{x},\boldsymbol{z} \sim P^n}\left[\forall g \in \Delta_T \text{ s.t. } \sum_i g(x_i) = 0 : \frac{1}{n}\sum_i g(z_i) \leq \xi\right] \geq 1 - \Pi_{\Delta_T}[2n]2^{-\xi n} \geq 1 - \delta.$$

If this $1 - \delta$ likely event happens, then also $\text{rej}_{\boldsymbol{z}}(S) = \frac{1}{n}\sum_i \partial_{\boldsymbol{c},\boldsymbol{c}'}(z_i) \leq \xi$ for the algorithm choices $\boldsymbol{c}, \boldsymbol{c}'$. □

*Proof of Theorem 10.9.* The proof follows from Theorem G.7 and the fact that,

$$\frac{1}{n}\left(\frac{2d \log 2n}{\varepsilon^*} + \log 1/\delta\right) \leq \frac{2d \log 2n}{n\sqrt{\frac{2d \log 2n}{n}}} + \frac{\log 1/\delta}{n} = \varepsilon^*.$$

□

## G.3 Massart Noise

This section shows that we can PQ learn in the Massart noise model. The Massart model (Massart et al., 2006) is defined with respect to a noise rate $\eta < 1/2$ and function (abusing notation) $\eta : \mathcal{X} \to [0, \eta]$:

**Definition G.9** (Massart Noise Model). *Let $P$ be a distribution on $\mathcal{X}$, $\eta < 1/2$, and $0 \leq \eta(x) \leq \eta$ for all $x \in \mathcal{X}$. The Massart distribution $P_{\eta,f}$ with respect to $f$ over $(x, y) \in \mathcal{X} \times \mathcal{Y}$ is defined as follows: first $x \sim P$ is chosen and then $y = f(x)$ with probability $1 - \eta(x)$ and $y = 1 - f(x)$ with probability $\eta(x)$.*

When clear from context, we omit $f$ and write $P_\eta = P_{\eta,f}$. The following lemma relates the *clean* error rate $\text{err}_P(h, f) = \mathbf{Pr}_P[h(x) \neq f(x)]$ and *noisy* error rate $\text{err}_{P_\eta} = \mathbf{Pr}_{(x,y)\sim P_\eta}[h(x) \neq y]$. Later, we will show how to drive the clean error arbitrarily close to 0 using an ERM.

**Lemma G.10.** *For any classifier $g : \mathcal{X} \to \mathcal{Y}$, any $\eta < 1/2, f \in \mathcal{H}$, and any distribution $P_\eta$ corrupted with Massart noise:*

$$(1 - 2\eta)\,\text{err}_P(g) \leq \text{err}_{P_\eta}(g) - \text{OPT},$$

*where* $\text{OPT} = \min_{h\in\mathcal{H}} \text{err}_{P_\eta}(h) = \mathbb{E}_{x\sim P}[\eta(x)]$.

*Proof.* By definition of the noisy error rate of $g$ under $P_\eta$, observe the following:

$$\begin{aligned}
\text{err}_{P_\eta}(g) &= \mathbf{Pr}_{(x,y)\sim P_\eta}[g(x) \neq y] \\
&= \mathbb{E}_{x\sim P}\left[\eta(x)1\{g(x) = f(x)\} + (1 - \eta(x))1\{g(x) \neq f(x)\}\right] \\
&= \mathbb{E}_{x\sim P}\left[\eta(x)(1 - 1\{g(x) \neq f(x)\}) + (1 - \eta(x))1\{g(x) \neq f(x)\}\right] \\
&= \mathbb{E}_{x\sim P}\left[\eta(x)\right] + \mathbb{E}_{x\sim P}\left[(1 - 2\eta(x))1\{g(x) \neq f(x)\}\right] \\
&= \text{OPT} + \mathbb{E}_{x\sim P}\left[(1 - 2\eta(x))1\{g(x) \neq f(x)\}\right] \\
&\geq \text{OPT} + (1 - 2\eta)\mathbb{E}_{x\sim P}[1\{g(x) \neq f(x)\}] \\
&= \text{OPT} + (1 - 2\eta)\,\text{err}_P(g),
\end{aligned}$$

where the last inequality follows from the fact that $\eta(x) \leq \eta$ for every $x \in \mathcal{X}$. Rearranging the terms concludes the proof. □

The following lemma shows that using an extra $N = \tilde{O}\left(\frac{dn^2}{\delta^2(1-2\eta)^2}\right)$ i.i.d. examples $(\boldsymbol{x}', \boldsymbol{y}') \sim P_\eta^N$, we can "denoise" the $n$ held-out examples $(\boldsymbol{x}, \boldsymbol{y}) \sim P_\eta^n$ with $\hat{h} = \mathsf{ERM}_{\mathcal{H}}(\boldsymbol{x}', \boldsymbol{y}')$, and then run Rejectron on $(\boldsymbol{x}, \hat{h}(\boldsymbol{x}))$. This shows that we can PQ learn $\mathcal{H}$ under Massart noise.

**Lemma G.11** (Massart denoising)**.** *For any $f \in \mathcal{H}$ and any distribution $P$ over $\mathcal{X}$, any $\eta < 1/2$ and $\eta : \mathcal{X} \to [0, \eta]$, let $P_\eta$ be the corresponding Massart distribution over $(x, y)$. For any $n \in \mathbb{N}$, let $(\boldsymbol{x}, \boldsymbol{y}) = (x_1, y_1), \ldots, (x_n, y_n) \sim P_\eta$ be i.i.d. examples sampled from $P_\eta$. Then,*

$$\mathbf{Pr}_{(\boldsymbol{x}', \boldsymbol{y}') \sim P_\eta^N}\left[\mathrm{err}_{\boldsymbol{x}}(\hat{h}, f) = 0\right] \geq 1 - \delta,$$

*where $\hat{h} = \mathsf{ERM}_{\mathcal{H}}(\boldsymbol{x}', \boldsymbol{y}')$ and $N = O\left(\frac{dn^2 + \log(2/\delta)}{\delta^2(1-2\eta)^2}\right)$.*

*Proof.* By agnostic learning guarantees for $\mathsf{ERM}_{\mathcal{H}}$, we have that for any $\varepsilon', \delta > 0$:

$$\mathbf{Pr}_{(\boldsymbol{x}', \boldsymbol{y}') \sim P_\eta^N}\left[\mathrm{err}_{P_\eta}(\hat{h}) \leq \mathrm{OPT} + \varepsilon'\right] \geq 1 - \frac{\delta}{2},$$

where $\hat{h} = \mathsf{ERM}_{\mathcal{H}}(\boldsymbol{x}', \boldsymbol{y}')$ and $N = O(\frac{d + \log(2/\delta)}{\varepsilon'^2})$. By Lemma G.10, choosing $\varepsilon' = \frac{\delta}{2n}(1 - 2\eta)$ guarantees that the clean error rate $\mathrm{err}_P(\hat{h}) \leq \frac{\delta}{2n}$. Since, $(\boldsymbol{x}, \boldsymbol{y}) \sim P_\eta^n$ are independent held-out examples, by a union bound, we get that $\mathrm{err}_{\boldsymbol{x}}(\hat{h}, f) = 0$ with probability $1 - \delta$. □

This yields an easy algorithm and corollary: simply use the $N$ examples $\boldsymbol{x}', \boldsymbol{y}'$ to denoise the $n$ labels for $\boldsymbol{x}$ and then run the Rejectron algorithm.

**Corollary G.12** (PQ guarantees under Massart noise)**.** *For any $n \in \mathbb{N}, \delta > 0, f \in \mathcal{H}$ and distributions $P, Q$ over $\mathcal{X}$, any $\eta < 1/2$ and $\eta : \mathcal{X} \to [0, \eta]$, let $P_\eta$ be the corresponding Massart distribution over $(x, y)$. Then,*

$$\mathbf{Pr}_{(\boldsymbol{x}', \boldsymbol{y}') \sim P_\eta^N, (\boldsymbol{x}, \boldsymbol{y}) \sim P_\eta^n, \tilde{\boldsymbol{x}} \sim Q^n}[\mathrm{err}_Q \leq 2\varepsilon^* \ \wedge \ \mathrm{rej}_P \leq \varepsilon^*] \geq 1 - \delta,$$

*where $\varepsilon^* = \sqrt{8\frac{d \ln 2n}{n} + \frac{8 \ln 32/\delta}{n}}$, $N = O\left(\frac{dn^2 + \log(2/\delta)}{\delta^2(1-2\eta)^2}\right)$, $\hat{h} = \mathsf{ERM}_{\mathcal{H}}(\boldsymbol{x}', \boldsymbol{y}')$, and $h|_S =$ Rejectron$(\boldsymbol{x}, \hat{h}(\boldsymbol{x}), \tilde{\boldsymbol{x}}, \varepsilon^*)$.*

## G.4 Semi-agnostic Analysis

In agnostic learning, the learner is given pairs $(x, y)$ from some unknown distribution $\mu$, and it is assumed that there exists some (unknown) $f \in \mathcal{H}$, $\eta \geq 0$ such that

$$\text{err}_\mu(f) := \mathbf{Pr}_{(x,y)\sim\mu}[y \neq f(x)] \leq \eta.$$

In this work, we consider the case where the test distribution $\tilde{\mu}$ may be (arbitrarily) different from the train distribution $\mu$, yet we require the existence of parameters $\eta, \tilde{\eta} \geq 0$ and an (unknown) $f \in \mathcal{H}$ such that

$$\text{err}_\mu(f) \leq \eta \text{ and } \text{err}_{\tilde{\mu}}(f) \leq \tilde{\eta}.^*$$

Moreover, in this work we assume that $\eta$ and $\tilde{\eta}$ are known. Unfortunately, even with this additional assumption, agnostic learning is challenging when $\mu \neq \tilde{\mu}$ and one cannot achieve guarantees near $\max\{\eta, \tilde{\eta}\}$ as one would hope, as we demonstrate below.

In what follows, we slightly abuse notation and write $(\boldsymbol{x}, \boldsymbol{y}) \sim D^n$ to denote $(x_i, y_i)$ drawn iid from $D$ for $i = 1, 2, \ldots, n$. The definitions of error and rejection with respect to such a distribution are:

$$\text{rej}_D(S) := \mathbf{Pr}_{(x,y)\sim D}[x \notin S]$$
$$\text{err}_D(h|_S) := \mathbf{Pr}_{(x,y)\sim D}[h(x) \neq y \wedge x \in S]$$

We prove the following lower bound.

**Lemma G.13.** *There exists a family of binary classifiers $\mathcal{H}$ of VC dimension 1, such that for any $\eta, \tilde{\eta} \in [0, 1/2]$ and $n \geq 1$, and for any selective classification algorithm $L : \mathcal{X}^n \times \mathcal{Y}^n \times \mathcal{X}^n \to \mathcal{Y}^{\mathcal{X}} \times 2^{\mathcal{X}}$ there exists $\mu, \tilde{\mu}$ over $\mathcal{X} \times \mathcal{Y}$ and $f \in \mathcal{H}$ such that:*

$$\mathbb{E}_{\substack{(\boldsymbol{x},\boldsymbol{y})\sim\mu^n \\ (\tilde{\boldsymbol{x}},\tilde{\boldsymbol{y}})\sim\tilde{\mu}^n}}[\text{err}_{\tilde{\mu}}(h|_S) + \text{rej}_\mu(S)] \geq \max\left\{\sqrt{\eta/8},\ \tilde{\eta}\right\}.$$

*where $h|_S = L(\boldsymbol{x}, \boldsymbol{y}, \boldsymbol{z})$ and where* $\text{err}_\mu(f) \leq \eta$ *and* $\text{err}_{\tilde{\mu}}(f) \leq \tilde{\eta}$.

---

*In Section 10.4, we assumed that $\eta = \tilde{\eta}$ for simplicity, yet here we consider the more general case where $\eta$ and $\tilde{\eta}$ may differ.

The proof is deferred to Section G.6.

We now show that Rejectron can be used to achieve nearly this guarantee. Recall that in the realizable setting, we fixed $\Lambda = n + 1$ in Rejectron. In this semi-agnostic setting, we will set $\Lambda$ as a function of $\eta$, hence our learner requires knowledge of $\eta$ unlike standard agnostic learning when $\mu = \tilde{\mu}$.

**Theorem G.14** (Agnostic generalization). *For any $n \in \mathbb{N}$, any $\delta, \gamma \in (0,1)$, any $\eta, \tilde{\eta} \in [0,1)$, and any distributions $\mu, \tilde{\mu}$ over $\mathcal{X} \times \mathcal{Y}$ such that that $\text{err}_\mu(f) \leq \eta$ and $\text{err}_{\tilde{\mu}}(f) \leq \tilde{\eta}$ simultaneously for some $f \in \mathcal{H}$:*

$$\mathbf{Pr}_{\substack{(\boldsymbol{x},\boldsymbol{y})\sim\mu^n \\ (\tilde{\boldsymbol{x}},\tilde{\boldsymbol{y}})\sim\tilde{\mu}^n}} \left[ \left( \text{err}_{\tilde{\mu}}(h|_S) \leq 2\sqrt{2\eta} + 2\tilde{\eta} + 4\varepsilon^* \right) \wedge \left( \text{rej}_\mu(S) \leq 4\sqrt{2\eta} + 4\varepsilon^* \right) \right] \geq 1 - \delta,$$

*where* $\varepsilon^* = 4\sqrt{\frac{d \ln 2n + \ln 48/\delta}{n}}$, $\Lambda^* = \sqrt{\frac{1}{8\eta + (\varepsilon^*)^2}}$, *and* $h|_S = \mathsf{Rejectron}(\boldsymbol{x}, \boldsymbol{y}, \tilde{\boldsymbol{x}}, \varepsilon^*, \Lambda^*)$.

A few points of comparison are worth making:

- When $\eta = \tilde{\eta} = 0$, one recovers guarantees that are slightly worse than those in the realizable (see Theorem 10.5).

- In standard agnostic learning, where $\mu$ and $\tilde{\mu}$ are identical, and thus $\eta = \tilde{\eta}$, one can set $S = \mathcal{X}$ (i.e., select everything) and ERM guarantees $\text{err}\big(h|_S(\tilde{\boldsymbol{x}}), \tilde{\boldsymbol{y}}\big) \leq \eta + \varepsilon$ w.h.p. for $n$ sufficiently large.

- The above theorem can be used to bound $\text{rej}_{\tilde{\mu}}$ using the following lemma:

**Lemma G.15.** *For any $S \subseteq \mathcal{X}, f, h \in \mathcal{Y}^{\mathcal{X}}$ and distributions $\mu, \tilde{\mu}$ over $\mathcal{X} \times \mathcal{Y}$:*

$$\text{rej}_{\tilde{\mu}}(S) \leq \text{rej}_\mu(S) + |\mu_X - \tilde{\mu}_X|_{\mathsf{TV}} \leq \text{rej}_\mu(S) + |\mu - \tilde{\mu}|_{\mathsf{TV}},$$

*where $\mu_X, \tilde{\mu}_X$ are the marginal distributions of $\mu, \tilde{\mu}$ over $\mathcal{X}$.*

*Proof.* The lemma follows from Lemma 10.2 applied to $P = \mu_X, Q = \tilde{\mu}_X$, and from the fact that the total variation between marginal distributions is no greater than the originals: $|\mu_X - \tilde{\mu}_X|_{\mathsf{TV}} \leq |\mu - \tilde{\mu}|_{\mathsf{TV}}$. □

As before, it will be useful (and easier) to first analyze the transductive case. In this case, it will be useful to further abuse notation and define, for any $\boldsymbol{y}, \boldsymbol{y}' \in \{0, 1, ?\}^n$,

$$\mathrm{err}(\boldsymbol{y}, \boldsymbol{y}') := \frac{1}{n}\left|\{i : y_i = 1 - y_i'\}\right|.$$

Using this, we will show:

**Theorem G.16** (Agnostic transductive). *For any $n \in \mathbb{N}$, $\varepsilon, \delta, \Lambda \geq 0, f \in \mathcal{H}$:*

$$\forall \boldsymbol{x}, \tilde{\boldsymbol{x}} \in \mathcal{X}^n, \boldsymbol{y}, \tilde{\boldsymbol{y}} \in \mathcal{Y}^n : \ \mathrm{err}(h|_S(\tilde{\boldsymbol{x}}), \tilde{\boldsymbol{y}}) \leq \varepsilon + 2\Lambda \cdot \mathrm{err}(f(\boldsymbol{x}), \boldsymbol{y}) + \mathrm{err}(f(\tilde{\boldsymbol{x}}), \tilde{\boldsymbol{y}}), \quad \text{(G.11)}$$

*where $h|_S = \mathsf{Rejectron}(\boldsymbol{x}, \boldsymbol{y}, \tilde{\boldsymbol{x}}, \varepsilon, \Lambda)$. Furthermore,*

$$\mathbf{Pr}_{\boldsymbol{x}, \boldsymbol{z} \sim P^n}\left[\forall \boldsymbol{y} \in \mathcal{Y}^n, \tilde{\boldsymbol{x}} \in \mathcal{X}^n : \ \mathrm{rej}_{\boldsymbol{z}}(S) \leq 2\Lambda^{-1} + \frac{9}{n}\left(\frac{d \ln 2n}{\varepsilon} + \frac{\ln 1/\delta}{2}\right)\right] \geq 1 - \delta. \quad \text{(G.12)}$$

The above bounds suggest the natural choice of $\Lambda = \eta^{-1/2}$ if $\mathrm{err}(f(\boldsymbol{x}), \boldsymbol{y}) \approx \eta$. The following two Lemmas will be used in its proof.

**Lemma G.17.** *For any $n \in \mathbb{N}$, $\varepsilon, \Lambda \geq 0$, $\boldsymbol{x}, \boldsymbol{z} \in \mathcal{X}^n, \boldsymbol{y} \in \mathcal{Y}^n$: $\mathrm{rej}_{\boldsymbol{x}}(S) \leq 1/\Lambda$ where $h|_S = \mathsf{Rejectron}(\boldsymbol{x}, \boldsymbol{y}, \boldsymbol{z}, \varepsilon, \Lambda)$.*

*Proof.* Note that for each iteration $t$ of the algorithm $\mathsf{Rejectron}(\boldsymbol{x}, \boldsymbol{y}, \boldsymbol{z}, \varepsilon, \Lambda)$,

$$\sum_{i \in [n]: z_i \in S_t} |c_t(z_i) - h(z_i)| - \Lambda \sum_{i \in [n]} |c_t(x_i) - h(x_i)| \geq 0,$$

because $c_t$ maximizes the above quantity over $\mathcal{H}$, and that quantity is zero at $c_t = h \in \mathcal{H}$. Also note that $x \notin S$ if and only if $|c_t(x) - h(x)| = 1$ for some $t$. More specifically, for each $i \in [n]$ such that $z_i \notin S$ there exists a *unique* $t \in [T]$ such that $z_i \in S_t$, and yet $z_i \notin S_{t+1}$, where the latter occurs when $|c_t(z_i) - h(z_i)| = 1$. Thus the total number of test and train rejections can be related as follows:

$$n \geq n\, \mathrm{rej}_{\boldsymbol{z}}(S) = \sum_{t \in [T]} \sum_{i \in [n]: z_i \in S_t} |c_t(z_i) - h(z_i)| \geq \sum_{t \in [T]} \Lambda \sum_{i \in [n]} |c_t(x_i) - h(x_i)| \geq n\Lambda\, \mathrm{rej}_{\boldsymbol{x}}(S).$$

Dividing both sides by $n \cdot \Lambda$ gives the lemma. $\square$

The following lemma is proven in Section G.5.

**Lemma G.18.** *For any $T, n \in \mathbb{N}$, any $\delta \geq 0$, and $\varepsilon = \frac{9}{2n}\left(d(T+1)\ln(2n) + \ln\frac{1}{\delta}\right)$:*

$$\mathbf{Pr}_{\boldsymbol{x},\boldsymbol{z}\sim P^n}\left[\exists \boldsymbol{c} \in \mathcal{H}^T, h \in \mathcal{H}: \ \mathrm{rej}_{\boldsymbol{z}}(S(h,\boldsymbol{c})) > 2\,\mathrm{rej}_{\boldsymbol{x}}(S(h,\boldsymbol{c})) + \varepsilon\right] \leq \delta.$$

Using these, we can now prove the transductive agnostic theorem.

*Proof of Theorem G.16.* To prove Equation (G.11), first fix any $\boldsymbol{x}, \tilde{\boldsymbol{x}} \in \mathcal{X}^n, \boldsymbol{y}, \tilde{\boldsymbol{y}} \in \mathcal{Y}^n, f \in \mathcal{H}$. Since $f \in \mathcal{H}$ the output $h = \mathsf{ERM}_{\mathcal{H}}(\boldsymbol{x}, \boldsymbol{y})$ satisfies $\mathrm{err}(h(\boldsymbol{x}), \boldsymbol{y}) \leq \mathrm{err}(f(\boldsymbol{x}), \boldsymbol{y})$. By the triangle inequality, this implies that

$$\mathrm{err}_{\boldsymbol{x}}(h, f) = \frac{1}{n}\sum_{i\in[n]} |h(x_i) - f(x_i)| \leq \frac{1}{n}\sum_{i\in[n]}\left(|h(x_i) - y_i| + |y_i - f(x_i)|\right) \leq 2\,\mathrm{err}(f(\boldsymbol{x}), \boldsymbol{y}). \tag{G.13}$$

Now suppose the algorithm Rejectron terminates on iteration $T+1$ and the output is $h|_S$ for $S = S_{T+1}$. Then by definition, for every $c \in \mathcal{H}$,

$$s_{T+1}(c) = \mathrm{err}_{\tilde{\boldsymbol{x}}}(h|_S, c) - \Lambda \cdot \mathrm{err}_{\boldsymbol{x}}(h, c) \leq \varepsilon,$$

For $c = f$ in particular,

$$\mathrm{err}_{\tilde{\boldsymbol{x}}}(h|_S, f) \leq \Lambda \cdot \mathrm{err}_{\boldsymbol{x}}(h, f) + \varepsilon \leq 2\Lambda \cdot \mathrm{err}(f(\boldsymbol{x}), \boldsymbol{y}) + \varepsilon.$$

Equation (G.11) follows from the above and the fact that

$$\mathrm{err}(h|_S(\tilde{\boldsymbol{x}}), \tilde{\boldsymbol{y}}) \leq \mathrm{err}(h|_{S_T}(\tilde{\boldsymbol{x}}), f(\tilde{\boldsymbol{x}})) + \mathrm{err}(f(\tilde{\boldsymbol{x}}), \tilde{\boldsymbol{y}}).$$

We next prove eq. (G.12). By Lemma G.17, $\mathrm{rej}_{\boldsymbol{x}}(S) \leq 1/\Lambda$ with certainty. So by Lemma G.18 applied to the marginal distribution $P = \mu_X$ over $\mathcal{X}$,

$$\mathbf{Pr}_{\boldsymbol{x},\boldsymbol{z}\sim P^n}\left[\exists h \in \mathcal{H}, \boldsymbol{c} \in \mathcal{H}^T: \ \mathrm{rej}_{\boldsymbol{z}}(S(h,\boldsymbol{c})) > 2\,\mathrm{rej}_{\boldsymbol{x}}(S(h,\boldsymbol{c})) + \xi\right] \leq \delta,$$

for $\xi = \frac{9}{2n}\left(\frac{2d}{\varepsilon}\ln(2n) + \ln\frac{1}{\delta}\right)$ using $T + 1 \leq 2/\varepsilon$. This implies eq. (G.12). $\square$

Returning to the generalization (distributional) agnostic case, the following theorem shows the trade-off between error and rejections:

**Theorem G.19.** *For any $n \in \mathbb{N}$ and $\delta, \Lambda \geq 0$, any $\varepsilon \geq 4\sqrt{\frac{d \ln 2n + \ln 24/\delta}{n}}$, any $f \in \mathcal{H}$, and any distributions $\mu, \tilde{\mu}$ over $\mathcal{X} \times \mathcal{Y}$:*

$$\mathbf{Pr}_{\substack{(\boldsymbol{x},\boldsymbol{y})\sim\mu^n \\ (\tilde{\boldsymbol{x}},\tilde{\boldsymbol{y}})\sim\tilde{\mu}^n}} \left[\text{err}_{\tilde{\mu}}(h|_S) \leq 8\Lambda\, \text{err}_\mu(f) + 2\,\text{err}_{\tilde{\mu}}(f) + \Lambda\varepsilon^2 + 3\varepsilon\right] \geq 1 - \delta, \tag{G.14}$$

*where $h|_S = \mathsf{Rejectron}(\boldsymbol{x}, \boldsymbol{y}, \tilde{\boldsymbol{x}}, \varepsilon, \Lambda)$. Furthermore,*

$$\mathbf{Pr}_{(\boldsymbol{x},\boldsymbol{y})\sim\mu^n} \left[\forall \tilde{\boldsymbol{x}} \in \mathcal{X}^n : \ \text{rej}_\mu(S) \leq \frac{2}{\Lambda} + 2\varepsilon\right] \geq 1 - \delta. \tag{G.15}$$

The proof of this theorem will use the following lemma.

**Lemma G.20.** *For any $h \in \mathcal{H}$, distribution $\mu$ over $\mathcal{X} \times \mathcal{Y}$, and $\varepsilon = \frac{16}{n}\left(dT \ln 2n + \ln \frac{8}{\delta}\right)$,*

$$\mathbf{Pr}_{(\boldsymbol{x},\boldsymbol{y})\sim\mu^n} \left[\forall \boldsymbol{c} \in \mathcal{H}^T : \ \text{err}_\mu(h|_{S(h,\boldsymbol{c})}) \leq \max\left\{2\,\text{err}(h|_{S(h,\boldsymbol{c})}(\boldsymbol{x}), \boldsymbol{y}), \varepsilon\right\}\right] \geq 1 - \delta.$$

The proof of this lemma is deferred to Section G.5.

*Proof of Theorem G.19.* The proof structure follows the proof of Theorem G.5. Fix $f$. We start by proving Equation (G.14). Let $\zeta = \frac{16}{n}\left(\frac{2d}{\varepsilon} \ln 2n + \ln \frac{24}{\delta}\right)$. By Lemma G.20,

$$\forall h \in \mathcal{H}\ \mathbf{Pr}_{(\tilde{\boldsymbol{x}},\tilde{\boldsymbol{y}})\sim\tilde{\mu}^n} \left[\forall \boldsymbol{c} \in \mathcal{H}^T : \ \text{err}_{\tilde{\mu}}(h|_{S(h,\boldsymbol{c})}) \leq \max\left\{2\,\text{err}(h|_{S(h,\boldsymbol{c})}(\tilde{\boldsymbol{x}}), \tilde{\boldsymbol{y}}), \zeta\right\}\right] \geq 1 - \delta/3.$$

Equation (G.11) from Theorem G.16 states that,

$$\forall \boldsymbol{x}, \tilde{\boldsymbol{x}} \in \mathcal{X}^n, \boldsymbol{y}, \tilde{\boldsymbol{y}} \in \mathcal{Y}^n : \ \text{err}(h|_S(\tilde{\boldsymbol{x}}), \tilde{\boldsymbol{y}}) \leq 2\Lambda \cdot \text{err}(f(\boldsymbol{x}), \boldsymbol{y}) + \text{err}(f(\tilde{\boldsymbol{x}}), \tilde{\boldsymbol{y}}) + \varepsilon,$$

with certainty. We next bound $\text{err}(f(\boldsymbol{x}), \boldsymbol{y})$ and $\text{err}(f(\tilde{\mathrm{x}}), \tilde{\boldsymbol{y}})$.

Since $\varepsilon^2/4 \geq \frac{4}{n} \ln \frac{3}{\delta}$, multiplicative Chernoff bounds imply that,

$$\mathbf{Pr}_{(\boldsymbol{x},\boldsymbol{y})\sim\mu^n} \left[\text{err}(f(\boldsymbol{x}), \boldsymbol{y}) \leq 2\,\text{err}_\mu(f) + \frac{\varepsilon^2}{4}\right] \geq 1 - \delta/3.$$

Also, since $\varepsilon/2 \geq \sqrt{\ln(3/\delta)/(2n)}$, additive Chernoff bounds imply that,

$$\mathbf{Pr}_{(\tilde{\boldsymbol{x}},\tilde{\boldsymbol{y}})\sim\tilde{\mu}^n}\left[\mathrm{err}(f(\tilde{\boldsymbol{x}}),\tilde{\boldsymbol{y}}) \leq \mathrm{err}_{\tilde{\mu}}(f) + \frac{\varepsilon}{2}\right] \geq 1 - \delta/3$$

Combining previous four displayed inequalities, which by the union bound all hold with probability $\geq 1-\delta$, gives,

$$\mathbf{Pr}_{\substack{(\boldsymbol{x},\boldsymbol{y})\sim\mu^n\\(\tilde{\boldsymbol{x}},\tilde{\boldsymbol{y}})\sim\tilde{\mu}^n}}\left[\mathrm{err}_{\tilde{\mu}}(h|_S) \leq \max\left\{2\left(2\Lambda(2\,\mathrm{err}_{\mu}(f)+\varepsilon^2/4) + (\mathrm{err}_{\tilde{\mu}}(f)+\varepsilon/2)+\varepsilon\right), \zeta\right\}\right] \geq 1-\delta.$$

Simplifying:

$$\mathbf{Pr}_{\substack{(\boldsymbol{x},\boldsymbol{y})\sim\mu^n\\(\tilde{\boldsymbol{x}},\tilde{\boldsymbol{y}})\sim\tilde{\mu}^n}}\left[\mathrm{err}_{\tilde{\mu}}(h|_S) \leq \max\left\{8\Lambda\,\mathrm{err}_{\mu}(f) + \Lambda\varepsilon^2 + 2\,\mathrm{err}_{\tilde{\mu}}(f) + 3\varepsilon, \zeta\right\}\right] \geq 1-\delta. \tag{G.16}$$

Next, we note that for our requirement of $\varepsilon \geq 4\sqrt{\frac{d\ln 2n+\ln 24/\delta}{n}}$, $\zeta \leq 2\varepsilon$ because:

$$\zeta = \frac{16}{n}\left(\frac{2d}{\varepsilon}\ln 2n + \ln\frac{24}{\delta}\right) \leq \frac{32}{n\varepsilon}\left(d\ln 2n + \ln\frac{24}{\delta}\right) \leq 2\frac{\varepsilon^2}{\varepsilon}.$$

Thus we can remove the maximum from eq. (G.16),

$$\mathbf{Pr}_{\substack{(\boldsymbol{x},\boldsymbol{y})\sim\mu^n\\(\tilde{\boldsymbol{x}},\tilde{\boldsymbol{y}})\sim\tilde{\mu}^n}}\left[\mathrm{err}_{\tilde{\mu}}(h|_S) \leq 8\Lambda\,\mathrm{err}_{\mu}(f) + \Lambda\varepsilon^2 + 2\,\mathrm{err}_{\tilde{\mu}}(f) + 3\varepsilon\right] \geq 1-\delta,$$

which is equivalent to what needed to prove in eq. (G.14).

We next prove eq. (G.15). By Lemma G.17, $\mathrm{rej}_{\boldsymbol{x}}(S) \leq 1/\Lambda$ with certainty. So by Lemma G.22 (Equation (G.24)) with $\gamma = 1/2$,

$$\mathbf{Pr}_{\boldsymbol{x},\boldsymbol{z}\sim P^n}\left[\exists h\in\mathcal{H}, \boldsymbol{c}\in\mathcal{H}^T:\ \mathrm{rej}_{\mu}\big(S(h,\boldsymbol{c})\big) > 2\,\mathrm{rej}_{\boldsymbol{x}}\big(S(h,\boldsymbol{c})\big) + \xi\right] \leq \delta,$$

for $\xi = \frac{16}{n}\left(\frac{2d}{\varepsilon}\ln(2n) + \ln\frac{8}{\delta}\right)$ using $T+1 \leq 2/\varepsilon$. This implies eq. (G.15) using the fact that,

$$\xi = \frac{16}{n}\left(\frac{2d}{\varepsilon}\ln(2n) + \ln\frac{16}{\delta}\right) \leq 2\cdot\frac{16}{n\varepsilon}\left(d\ln(2n) + \ln\frac{16}{\delta}\right) \leq 2\cdot\frac{\varepsilon^2}{\varepsilon} = 2\varepsilon.$$

$\square$

From this theorem, our main agnostic upper-bound follows in a straightforward fashion.

*Proof of Theorem G.14.* Note that for our choice of $\Lambda^* = \sqrt{\frac{1}{8\eta+(\varepsilon^*)^2}}$,

$$\begin{aligned}\left(8\Lambda^* \operatorname{err}_\mu(f) + 2\operatorname{err}_{\tilde{\mu}}(f)\right) + \Lambda^*(\varepsilon^*)^2 + 3\varepsilon^* &\leq \Lambda^*(8\eta + (\varepsilon^*)^2) + 2\tilde{\eta} + 3\varepsilon^* \\ &= \sqrt{8\eta + (\varepsilon^*)^2} + 2\tilde{\eta} + 3\varepsilon^* \\ &\leq 2\sqrt{2\eta} + \varepsilon^* + 2\tilde{\eta} + 3\varepsilon^*,\end{aligned}$$

using the fact that $\sqrt{a+b} \leq \sqrt{a}+\sqrt{b}$. For the chosen $\varepsilon^* = 4\sqrt{\frac{d\ln 2n + \ln 48/\delta}{n}}$, Theorem G.19 implies,

$$\begin{aligned}\mathbf{Pr}_{\substack{(\boldsymbol{x},\boldsymbol{y})\sim\mu^n \\ (\tilde{\boldsymbol{x}},\tilde{\boldsymbol{y}})\sim\tilde{\mu}^n}}\left[\operatorname{err}_{\tilde{\mu}}(h|_S) \leq \left(8\Lambda^* \operatorname{err}_\mu(f) + 2\operatorname{err}_{\tilde{\mu}}(f)\right) + \Lambda^*(\varepsilon^*)^2 + 3\varepsilon^*\right] &\geq 1 - \delta/2 \\ \mathbf{Pr}_{\substack{(\boldsymbol{x},\boldsymbol{y})\sim\mu^n \\ (\tilde{\boldsymbol{x}},\tilde{\boldsymbol{y}})\sim\tilde{\mu}^n}}\left[\operatorname{err}_{\tilde{\mu}}(h|_S) \leq 2\sqrt{2\eta} + 2\tilde{\eta} + 4\varepsilon^*\right] &\geq 1 - \delta/2 \quad \text{(G.17)}\end{aligned}$$

Also note that

$$\frac{2}{\Lambda^*} + 2\varepsilon^* \leq 2\sqrt{8\eta + (\varepsilon^*)^2} + 2\varepsilon^* \leq 4\sqrt{2\eta} + 2\varepsilon^* + 2\varepsilon^* \leq 4\sqrt{2\eta} + 4\varepsilon^*.$$

Theorem G.19 also implies:

$$\begin{aligned}\mathbf{Pr}_{(\boldsymbol{x},\boldsymbol{y})\sim\mu^n}\left[\forall\tilde{\boldsymbol{x}} \in \mathcal{X}^n : \ \operatorname{rej}_\mu(S) \leq \frac{2}{\Lambda^*} + 2\varepsilon^*\right] &\geq 1 - \delta/2 \\ \mathbf{Pr}_{(\boldsymbol{x},\boldsymbol{y})\sim\mu^n}\left[\forall\tilde{\boldsymbol{x}} \in \mathcal{X}^n : \ \operatorname{rej}_\mu(S) \leq 4\sqrt{2\eta} + 4\varepsilon^*\right] &\geq 1 - \delta/2 \\ \mathbf{Pr}_{(\boldsymbol{x},\boldsymbol{y})\sim\mu^n}\left[\forall\tilde{\boldsymbol{x}} \in \mathcal{X}^n : \ \operatorname{rej}_{\tilde{\mu}}(S) \leq 4\sqrt{2\eta} + 4\varepsilon^* + |\mu - \tilde{\mu}|_{\mathsf{TV}}\right] &\geq 1 - \delta/2, \quad \text{(G.18)}\end{aligned}$$

where we have used Lemma G.15 in the last step. The union bound over eq. (G.17) and eq. (G.18) proves the corollary. □

## G.5 Generalization Lemmas

In this section we state auxiliary lemmas that relate the empirical error and rejection rates to generalization error and rejection rates.

To bound generalization, it will be useful to note that the classifiers $h|_S$ output by our algorithm are not too complex. To do this, for any $k \in \mathbb{N}$ and any classifiers $c_1, c_2, \ldots, c_k : \mathcal{X} \to \mathcal{Y}$, define the *disagreement* function that is 1 if any of two disagree on $x$:

$$\mathsf{dis}_{c_1,\ldots,c_k}(x) := \begin{cases} 0 & \text{if } c_1(x) = c_2(x) = \cdots = c_k(x) \\ 1 & \text{otherwise.} \end{cases} \tag{G.19}$$

Also denote by $\bar{f} = 1 - f$ and $\boldsymbol{c} = (c_1, \ldots, c_T) \in \mathcal{H}^T$. In these terms we can write,

$$\mathsf{dis}_{h,\boldsymbol{c}} = \begin{cases} 0 & \text{if } h(x) = c_1(x) = c_2(x) = \cdots = c_T(x) \\ 1 & \text{otherwise} \end{cases}$$

$$\mathsf{dis}_{\bar{f},h,\boldsymbol{c}} = \begin{cases} 1 & \text{if } 1 - f(x) = h(x) = c_1(x) = c_2(x) = \cdots = c_T(x) \\ 0 & \text{otherwise.} \end{cases}$$

Recall the definition of $\Pi_G[2n]$ for a family $G$ of classifiers $g : \mathcal{X} \to \{0, 1\}$:

$$\Pi_G[2n] := \max_{\boldsymbol{w} \in \mathcal{X}^{2n}} |\{g(\boldsymbol{w}) : g \in G\}|.$$

**Lemma G.21** (Complexity of output class). *For any $h \in \mathcal{H}$, let*

$$\mathsf{Dis}_T := \{\mathsf{dis}_{h,c_1,\ldots,c_T} : h, c_1, \ldots, c_T \in \mathcal{H}\} \tag{G.20}$$

$$\mathsf{Dis}_{h,T} := \{\mathsf{dis}_{h,c_1,\ldots,c_T} : c_1, \ldots, c_T \in \mathcal{H}\}, \tag{G.21}$$

$$\mathsf{Dis}_{\bar{f},h,T} := \left\{\mathsf{dis}_{\bar{f},h,c_1,\ldots,c_T} : c_1, \ldots, c_T \in \mathcal{H}\right\}, \tag{G.22}$$

*Then $\Pi_{\mathsf{Dis}_T}[2n] \leq (2n)^{d(T+1)}$, $\Pi_{\mathsf{Dis}_{h,T}}[2n] \leq (2n)^{dT}$, and $\Pi_{\mathsf{Dis}_{\bar{f},h,T}}[2n] \leq (2n)^{dT}$, where $d$ is the VC dimension of $\mathcal{H}$.*

*Proof.* The proof follows trivially from Sauer's lemma, since the number of labelings of $2n$ examples by any $c \in \mathcal{H}$ is at most $(2n)^d$ and there are $T$ choices of $c_1, \ldots, c_T$ and 1 choice of $h$. □

**Lemma G.22** (Generalization bounds using Blumer et al. (1989b)). *For any $n \in \mathbb{N}$, any*

*distribution $P$ over a domain $\mathcal{X}$, any set $G$ of binary classifiers over $\mathcal{X}$, and any $\varepsilon > 0$,*

$$\mathbf{Pr}_{\boldsymbol{z}\sim P^n}\left[\exists g \in G: \ (\mathbb{E}_{x\sim P}[g(x)] > \varepsilon) \wedge \left(\frac{1}{n}\sum_{i\in[n]} g(z_i) = 0\right)\right] \leq 2\Pi_G[2n]2^{-\varepsilon n/2}, \tag{G.23}$$

*and, for any $\gamma \in (0,1)$,*

$$\mathbf{Pr}_{\boldsymbol{z}\sim P^n}\left[\exists g \in G: \mathbb{E}_{x\sim P}[g(x)] > \max\left\{\varepsilon, \frac{1}{1-\gamma}\cdot\frac{1}{n}\sum_{i\in[n]} g(z_i)\right\}\right] \leq 8\Pi_G[2n]e^{-\frac{\gamma^2\varepsilon n}{4}}. \tag{G.24}$$

*Finally, for any distribution $\mu$ over $\mathcal{X}\times\mathcal{Y}$ and any $\gamma \in (0,1)$,*

$$\mathbf{Pr}_{(\boldsymbol{x},\boldsymbol{y})\sim\mu^n}\left[\exists g \in G: \mathrm{err}_\mu(g) > \max\left\{\varepsilon, \frac{1}{1-\gamma}\cdot\frac{1}{n}\sum_{i\in[n]} |g(x_i) - y_i|\right\}\right] \leq 8\Pi_G[2n]e^{-\frac{\gamma^2\varepsilon n}{4}}. \tag{G.25}$$

*Proof.* Simply consider a binary classification problem where the target classifier is the constant 0 function, with training examples $\boldsymbol{z} \sim P^n$. Then the training error rate is $\sum g(z_i)/n$ and the generalization error is $\mathbf{Pr}_P[g(x) = 1]$. By Theorem A2.1 of Blumer et al. (1989b), the probability that any $g \in G$ has 0 training error and test error greater than $\varepsilon$ is at most $\Pi_G[2n]2^{-\varepsilon n/2}$. Similarly eq. (G.24) and (G.25) follow from Theorem A3.1 of Blumer et al. (1989b), noting that the bound holds trivially for all $g$ with $\mathbb{E}[g(x)] \leq \varepsilon$. □

We now prove Lemma G.20, which adapts the last bound above to the agnostic setting.

*Proof of Lemma G.20.* We would like to apply the last inequality of Lemma G.22 with $\gamma = 1/2$, but unfortunately that lemma does not apply to error rates of selective classifiers. First, consider the case where the distribution is "consistent" in that $\mathbf{Pr}_{x,y\sim\mu}[y = \tau(x)]$ for some arbitrary $\tau : \mathcal{X} \to \{0,1\}$. We can consider the modified functions,

$$g_{h,\boldsymbol{c}}(x) = \begin{cases} \tau(x) & \text{if } x \notin S(h,\boldsymbol{c}) \\ h(x) & \text{otherwise.} \end{cases}$$

It follows that $\mathrm{err}_\mu(g_{h,\boldsymbol{c}}) = \mathrm{err}_\mu(h|_S)$. Furthermore, the class $G = \{g_{h,\boldsymbol{c}} : h \in \mathcal{H}, \boldsymbol{c} \in \mathcal{H}^T\}$ satisfies $\Pi_G[2n] \le (2n)^{dT}$ (just as we argued $\Pi_{\mathsf{Dis}_{h,T}}[2n] \le (2n)^{dT}$) because there are $(2n)^d$ different labelings of $c$ on $2n$ elements and thus there are at most $(2n)^{dT}$ choices of $T$ of these for $\boldsymbol{c} \in \mathcal{H}^T$. Thus, applying Lemma G.22 gives the lemma for consistent $\mu$.

The inconsistent case can be reduced to the consistent case by a standard trick. In particular, we will extend $\mathcal{X}$ to $\mathcal{X}' = \mathcal{X} \times \{0,1\}$ by appending a latent (hidden) copy of $y$, call it $b$, to each example $x$. In particular For $c \in \mathcal{H}$, define $c'(x,b) = c(x)$ so that the classifiers cannot depend on $b$. This does not change the VC dimension of the classifiers. However, now, any distribution over $\mu$ can be converted to a consistent distribution $\mu'$ over $\mathcal{X}'$ whose marginal distribution over $\mathcal{X}$ agrees with $\mu$, by making

$$\mu'((x,b),y) = \begin{cases} \mu(x,y) & \text{if } b = y \\ 0 & \text{otherwise.} \end{cases}$$

In other words, $\mathbf{Pr}_{(x,b),y\sim\mu'}[b = y] = 1$. Now, clearly $\mu'$ is consistent. The statement of the lemma applied to $\mu'$ implies the corresponding statement for $\mu$ since the classifiers do not depend on $b$. □

We now prove Lemma G.8.

*Proof of Lemma G.8.* Fix any $n \in \mathbb{N}$, any distribution $P$ over a domain $\mathcal{X}$ and any $\beta \in [n]$. Imagine selecting $\boldsymbol{x}, \boldsymbol{z} \sim P^n$ by first randomly choosing $2n$ random elements $\boldsymbol{w} \sim P^{2n}$ and then randomly dividing these elements into two equal sized sequences $\boldsymbol{x}, \boldsymbol{z}$. Let $\pi(\boldsymbol{w})$ denote the distribution over the $(2n)!$ such divisions $\boldsymbol{x}, \boldsymbol{z} \in \mathcal{X}^n$. For any $g \in G$ and $\boldsymbol{w} \in \mathcal{X}^{2n}$, we claim:

$$\mathbf{Pr}_{(\boldsymbol{x},\boldsymbol{z})\sim\pi(\boldsymbol{w})}\left[\sum_i g(x_i) = 0 \ \wedge\ \sum_i g(z_i) \ge \lceil \varepsilon n \rceil\right] \le 2^{-\lceil \varepsilon n \rceil}.$$

To see this, suppose $s = \sum_i g(w_i) \ge \varepsilon n$ (otherwise the probability above is zero). The probability that all of them are in the test set is at most $2^{-s} \le 2^{-\varepsilon n}$ because the chance that the first rejection is placed in the test set is $1/2$, the second is $(n-1)/(2n-1) < 1/2$, and so forth. The above equation directly implies eq. (G.9) by dividing by $n$.

We now move to eq. (G.10). Consider random variables $A = \sum g(x_i)$ and $B = \sum g(z_i)$. It suffices to show that that $B > (1+\alpha)A + r$ with probability $\le e^{-2\alpha(2+\alpha)^{-2}r}$ for $r = \varepsilon n$.

Note that since $B = s - A$,

$$B \geq (1+\alpha)A + r \iff A \leq \frac{s-r}{2+\alpha}.$$

Hence, it suffices to prove that

$$\mathbf{Pr}\left[A \leq \frac{s-r}{2+\alpha}\right] \leq e^{-\frac{2\alpha}{(2+\alpha)^2}r}. \tag{G.26}$$

Let $\mathcal{D}$ be the Bernoulli distribution on $\{0,1\}$ with mean $\mu = \frac{s}{2n}$. Note that by linearly of expectation, $\mathbb{E}[A] = \mathbb{E}[B] = \mu n$. It is well-known that the probabilities of such an unbalanced split are smaller for sampling without replacement than with replacement (see, e.g., Bardenet et al., 2015). Thus, it suffices to prove Equation (G.26) assuming $A$ was sampled by sampling $n$ iid elements $(A_1, \ldots, A_n) \sim \mathcal{D}^n$, and setting $A = \sum_{i=1}^n A_i$. By the multiplicative Chernoff bound, for every $\rho \in [0,1]$,

$$\mathbf{Pr}\left[A \leq (1-\rho)\mu n\right] \leq e^{-\rho^2 \mu n/2} = e^{-\rho^2 s/4}.$$

In particular, for $\rho = \frac{\alpha s + 2r}{s(2+\alpha)}$, since $1 - \rho = \frac{2s-2r}{s(2+\alpha)}$ and $\mu n = s/2$, this gives:

$$\mathbf{Pr}\left[A \leq \frac{s-r}{2+\alpha}\right] \leq e^{-\frac{(\alpha s+2r)^2}{4(2+\alpha)^2 s}}$$

Hence, it remains to show that the RHS above is at most $\exp\left(-\frac{2\alpha}{(2+\alpha)^2}r\right)$, or equivalently,

$$\frac{(\alpha s + 2r)^2}{4(2+\alpha)^2 s} \geq \frac{2\alpha r}{(2+\alpha)^2}.$$

After multiplying both sides by $4(2+\alpha)^2 s$, the above can be rewritten as $(\alpha s + 2r)^2 \geq 8\alpha s r$, and equivalently as $(\alpha s - 2r)^2 \geq 0$, which indeed always holds. $\square$

We are now ready to prove Lemma G.1.

*Proof of Lemma G.1.* Note that for $\mathsf{dis}_{h,\mathbf{c}}$ defined as in eq. (G.21), $\mathsf{dis}_{h,\mathbf{c}}(x) = 0$ if and only if $x \in S(h, \boldsymbol{c})$. Thus, $\mathsf{rej}_{\boldsymbol{x}}(S(h, \boldsymbol{c})) = 0$ implies that

$$\sum_{i=1}^{n} \mathsf{dis}_{h,\boldsymbol{c}}(x_i) = 0. \tag{G.27}$$

Also note that,

$$\mathsf{rej}_{\boldsymbol{z}}(S(h, \boldsymbol{c})) = \frac{1}{n} \sum_{i=1}^{n} \mathsf{dis}_{h,\boldsymbol{c}}(z_i).$$

Hence, it suffices to show

$$\mathbf{Pr}_{\boldsymbol{x},\boldsymbol{z} \sim P^n} \left[ \exists \boldsymbol{c} \in \mathcal{H}^T, h \in \mathcal{H} : \left( \frac{1}{n} \sum_{i=1}^{n} \mathsf{dis}_{h,\boldsymbol{c}}(x_i) = 0 \right) \wedge \left( \frac{1}{n} \sum_{i=1}^{n} \mathsf{dis}_{h,\boldsymbol{c}}(z_i) > \varepsilon \right) \right] \le \delta \tag{G.28}$$

By Lemma G.8,

$$\mathbf{Pr}_{\boldsymbol{x},\boldsymbol{z} \sim P^n} \left[ \exists \varphi \in \mathsf{Dis}_T : \left( \sum_{i=1}^{n} \varphi(x_i) = 0 \right) \wedge \left( \sum_{i=1}^{n} \varphi(z_i) \ge \varepsilon n \right) \right] \le 2^{-\varepsilon n} \Pi_{\mathsf{Dis}_T}[2n]. \tag{G.29}$$

Lemma G.21 states that $\Pi_{\mathsf{Dis}_T}[2n] \le (2n)^{d(T+1)}$ which combined with our choice of $\varepsilon$, gives:

$$2^{-\varepsilon n} \Pi_{\mathsf{Dis}_T}[2n] \le 2^{-\varepsilon n} (2n)^{d(T+1)} = \delta.$$

Hence, eq. (G.29) implies eq. (G.28) because $\mathsf{dis}_{h,\boldsymbol{c}} \in \Pi_{\mathsf{Dis}_T}$. □

We now prove Lemma G.18.

*Proof of Lemma G.18.* Note that for $\mathsf{dis}_{h,\boldsymbol{c}}$ defined as in eq. (G.21), $\mathsf{dis}_{h,\boldsymbol{c}}(x) = 1$ if and only if Rejectron rejects $x$ when the algorithm's choices are $h \in \mathcal{H}$ and $\boldsymbol{c} \in \mathcal{H}^T$, i.e., $x \notin S$. Thus,

$$\mathsf{rej}_{\boldsymbol{x}}(S) = \frac{1}{n} \sum_{i=1}^{n} \mathsf{dis}_{h,\boldsymbol{c}}(x_i) \text{ and } \mathsf{rej}_{\boldsymbol{z}}(S) = \frac{1}{n} \sum_{i=1}^{n} \mathsf{dis}_{h,\boldsymbol{c}}(z_i).$$

Hence, it suffices to show,

$$\mathbf{Pr}_{\boldsymbol{x},\boldsymbol{z}\sim P^n}\left[\exists \boldsymbol{c} \in \mathcal{H}^T, h \in \mathcal{H} :\ \frac{1}{n}\sum_{i=1}^{n} \mathsf{dis}_{h,\boldsymbol{c}}(z_i) > \frac{2}{n}\sum_{i=1}^{n} \mathsf{dis}_{h,\boldsymbol{c}}(z_i) + \varepsilon\right] \leq \delta \qquad \text{(G.30)}$$

Lemma G.8 (with $\alpha = 1$) implies that:

$$\mathbf{Pr}_{\boldsymbol{x},\boldsymbol{z}\sim P^n}\left[\exists \varphi \in \mathsf{Dis}_T : \left(\sum_i \varphi(x_i) = 0\right) \wedge \left(\sum_i \varphi(z_i) \geq \varepsilon n\right)\right] \leq e^{-\frac{2}{9}\varepsilon n}\Pi_{\mathsf{Dis}_T}[2n].$$

Lemma G.21 states that $\Pi_{\mathsf{Dis}_T}[2n] \leq (2n)^{d(T+1)}$ which combined with our choice of $\varepsilon$, gives:

$$e^{-\frac{2}{9}\varepsilon n}\Pi_{\mathsf{Dis}_T}[2n] \leq e^{-\frac{2}{9}\varepsilon n}(2n)^{d(T+1)} = \delta.$$

Hence, the above implies eq. (G.30) because $\mathsf{dis}_{h,\boldsymbol{c}} \in \Pi_{\mathsf{Dis}_T}$. □

We now prove Lemma G.3.

*Proof of Lemma G.3.* Fix $f, h \in \mathcal{H}$. For every $\boldsymbol{c} \in \mathcal{H}^T$, let $S = S(h, \boldsymbol{c})$ and define:

$$g_{\boldsymbol{c}}(x) := \begin{cases} 1 & \text{if } f(x) \neq h(x) \wedge x \in S \\ 0 & \text{otherwise.} \end{cases} \quad \text{and } G := \{g_{\boldsymbol{c}} :\ \boldsymbol{c} \in \mathcal{H}^T\}$$

So $G$ depends on $h, f$ which we have fixed. Note that $g_{\boldsymbol{c}}(x) = 1$ iff $h|_S(x) = 1 - f(x)$. Hence,

$$\frac{1}{n}\sum_{i\in[n]} g_{\boldsymbol{c}}(\tilde{x}_i) = \mathrm{err}_{\tilde{\boldsymbol{x}}}(h|_S, f).$$

Equation (G.24) of Lemma G.22 (with $\gamma = 1/2$ and substituting $Q$ for $P$ and $\varepsilon' = 2\varepsilon$ for $\varepsilon$) implies that for the entire class of functions $G$:

$$\mathbf{Pr}_{\tilde{\boldsymbol{x}}\sim Q^n}\left[\exists g \in G : \left(\mathbb{E}_{x'\sim Q}[g(x')] > 2\varepsilon\right) \wedge \left(\frac{1}{n}\sum_{i\in[n]} g(\tilde{x}_i) \leq \varepsilon\right)\right] \leq 8\Pi_G[2n]e^{-\varepsilon n/8}.$$

By definition of $G$, the above implies that,

$$\mathbf{Pr}_{\tilde{\boldsymbol{x}}\sim Q^n}\left[\exists \boldsymbol{c}\in\mathcal{H}^T:\ \left(\mathrm{err}_Q(h|_{S(h,\boldsymbol{c})},f)>2\varepsilon\right)\ \wedge\ \left(\mathrm{err}_{\tilde{\boldsymbol{x}}}(h|_{S(h,\boldsymbol{c})},f)\le\varepsilon\right)\right]\le 8\Pi_G[2n]e^{-\varepsilon n/8}.$$

Thus, it remains to prove that

$$8\Pi_G[2n]e^{-\varepsilon n/8}\le\delta.$$

To bound $\Pi_G[2n]$, note that $g_{\boldsymbol{c}}(x)=1-\mathsf{dis}_{\bar{f},h,\boldsymbol{c}}(x)$ and thus $\Pi_G[2n]=\Pi_{\mathsf{Dis}_{\bar{f},h,T}}[2n]$, which is at most $(2n)^{dT}$ by Lemma G.21. Since $T\le 1/\varepsilon$:

$$8(2n)^{dT}e^{-\varepsilon n/8}\le 8(2n)^{d/\varepsilon}e^{-\varepsilon n/8}.$$

Hence it suffices to show that the above is at most $\delta$ for $\varepsilon\ge\frac{8\ln 8/\delta}{n}+\sqrt{\frac{8d\ln 2n}{n}}$, or equivalently that,

$$\varepsilon\frac{n}{8}-\frac{d}{\varepsilon}\ln 2n\ge\ln\frac{8}{\delta}.$$

By multiplying both sides of the equation by $\varepsilon\cdot\frac{8}{n}$ we get

$$\varepsilon^2-\frac{8}{n}d\ln 2n\ge\varepsilon\frac{8}{n}\ln\frac{8}{\delta}.$$

Substituting $c=\frac{8d\ln 2n}{n}$ and $b=\frac{8\ln 8/\delta}{n}$, we have $\varepsilon\ge b+\sqrt{c}$, and what we need to show above is equivalent to:

$$\varepsilon^2-c\ge\varepsilon b$$

or equivalently

$$\varepsilon(\varepsilon-b)\ge c$$

which holds for any $\varepsilon\ge b+\sqrt{c}$ because the LHS above is $\ge(b+\sqrt{c})\sqrt{c}\ge c$. □

We next prove Lemma G.4.

*Proof of Lemma G.4.* Fix any $T\ge 1$ and any $h\in\mathcal{H}$. Consider $\mathsf{dis}_{h,\boldsymbol{c}}$ as defined in eq. (G.19), where $\mathsf{dis}_{h,\boldsymbol{c}}(x)=1$ iff $x\notin S(h,\boldsymbol{c})$ is rejected. Thus,

$$\mathsf{rej}_{\boldsymbol{x}}(S(h,\boldsymbol{c}))=\frac{1}{n}\sum_{i=1}^{n}\mathsf{dis}_{h,\boldsymbol{c}}(x_i)\ \text{ and }\ \mathsf{rej}_P(S(h,\boldsymbol{c}))=\mathbb{E}_{x'\sim P}[\mathsf{dis}_{h,\boldsymbol{c}}(x')].$$

By Lemma G.22 (Equation (G.23)), the probability that any such $\mathsf{dis}_{h,\boldsymbol{c}} \in \mathsf{Dis}_T$ is 0 on all of $\boldsymbol{x}$ but has expectation on $P$ greater than $\xi = \frac{2}{n}(d(T+1)\ln(2n) + \ln \frac{2}{\delta})$ is at most:

$$2\Pi_{\mathsf{Dis}_T}[2n]2^{-\xi n/2} \leq 2(2n)^{d(T+1)}2^{-\xi n/2} = \delta,$$

where the first inequality follows from the fact that $\Pi_{\mathsf{Dis}_T}[2n] \leq (2n)^{d(T+1)}$, which follows from Lemma G.21. Similarly, eq. (G.24) of Lemma G.22 (with $\gamma = 1/2$ and $\varepsilon = 2\alpha$) implies that:

$$\mathbf{Pr}_{\boldsymbol{x}\sim P^n}\left[\exists h, \boldsymbol{c} : (\mathbb{E}_{x'\sim P}[\mathsf{dis}_{h,\boldsymbol{c}}(x')] > 2\alpha) \wedge \left(\frac{1}{n}\sum_{i\in[n]} \mathsf{dis}_{h,\boldsymbol{c}}(x_i) \leq \alpha\right)\right] \leq 8\Pi_{\mathsf{Dis}_T}[2n]e^{-\alpha n/8}.$$

For $\alpha$ as in the lemma, the right hand side above is at most $\delta$. □

## G.6 Proofs of Lower Bounds

We note that, in the lower bound of Theorem 10.7, the distribution $Q$ is fixed, independent of $f$. Since $Q$ is used only for unlabeled test samples, the learning algorithm can gain no information about $Q$ even if it is given a large number $m$ of test samples. In particular, it implies that even if one has $n$ training samples and infinitely many samples from $Q$, one cannot achieve error less than $\Omega(\sqrt{d/n})$. It would be interesting to try to improve the lower-bound to have a specific dependence on $m$ (getting $\Omega(\sqrt{1/n} + 1/m)$ is likely possible using a construction similar to the one below). Also, the lower-bound could be improved if one had fixed distributions $\nu, P, Q$ independent of $n$.

*Proof of Theorem 10.7.* Let $\mathcal{X} = \mathbb{N}$ and $\mathcal{H}$ be the concept class of functions which are 1 on exactly $d$ integers, which can easily be seen to have VC dimension $d$. The distribution $P$ is simply uniform over $[8n] = \{1, 2, \ldots, 8n\}$. Let $k = \sqrt{8dn}$. The distribution $Q$ is uniform over $[k]$. We consider a distribution $\nu$ over functions $f \in \mathcal{H}$ that is uniform over the $\binom{k}{d}$ functions that are 1 on exactly $d$ points in $[k]$. We will show,

$$\mathbb{E}_{f\sim\nu}\left[\mathbb{E}_{\substack{\boldsymbol{x}\sim P^n\\ \tilde{\boldsymbol{x}}\sim Q^n}}\left[\mathrm{rej}_P + \mathrm{err}_Q\right]\right] \geq K\sqrt{\frac{d}{n}}. \quad \text{(G.31)}$$

By the probabilistic method, this would imply the lemma.

The set of training samples is $T = \{x_i : i \in [n]\} \subseteq [8n]$. Say an $j \in [k]$ is "unseen" if it does not occur as a training example, $j \notin T$. WLOG, we may assume that the learner makes the same classification $h|_S$ for each unseen $j \in [k]$ since an asymmetric learner can only be improved by making the (same) optimal decision for each unseen $j \in [k]$, where the optimal decisions are defined to be those that minimize $\mathbb{E}[\text{rej}_P + \text{err}_Q \mid \boldsymbol{x}, f(\boldsymbol{x})]$. (The unlabeled test are irrelevant because $Q$ is fixed.)

Now, let $U \leq k$ be the random variable that is the number of seen $j \in [k]$ and $V \leq d$ be the number that are labeled 1 (which the learner can easily determine).

$$U = |T \cap [k]|$$
$$V = |\{j \in T \cap [k] : f(j) = 1\}|.$$

Note that $\mathbb{E}[U] \leq k/8$ and $\mathbb{E}[V] \leq d/8$ since each $j \in [k]$ is observed with probability $\leq 1/8$ by choice of $P$ (the precise observation probability is $1 - (1 - \frac{1}{8n})^n \leq \frac{1}{8}$). These two inequalities implies that,

$$\mathbb{E}\left[\frac{U}{k} + \frac{V}{d}\right] \leq \frac{1}{8} + \frac{1}{8} = \frac{1}{4}.$$

Thus, by Markov's inequality,

$$\mathbf{Pr}\left[\frac{U}{k} + \frac{V}{d} \leq \frac{1}{2}\right] \geq \frac{1}{2}.$$

This implies that, with probability $\geq 1/2$, both $U \leq k/2$ *and* $V \leq d/2$. Suppose this event happens. Now, consider three cases.

Case 1) if the learner predicts ? on all unseen $j \in [k]$, then

$$\text{rej}_P \geq \frac{k}{2} \cdot \frac{1}{8n} = \sqrt{\frac{d}{32n}}$$

because there are at least $k/2$ unseen $j \in [k]$ and each has probability $\frac{1}{8n}$ under $P$.

Case 2) if the learner predicts 0 on all unseen $j \in [k]$, then

$$\mathrm{err}_Q \geq \frac{d}{2} \cdot \frac{1}{k} = \sqrt{\frac{d}{32n}},$$

because there are at least $d/2$ 1's that are unseen and each has probability $1/k$ under $Q$.

Case 3) if the learner predicts 1 on all unseen $j \in [k]$ then

$$\mathrm{err}_Q \geq \left(\frac{k}{2} - d\right)\frac{1}{k} = \frac{1}{2} - \sqrt{\frac{d}{8n}} \geq \sqrt{\frac{d}{8n}} > \sqrt{\frac{d}{32n}}$$

because there are at least $k/2 - d$ unseen 0's, each with probability $1/k$ under $Q$ (and by assumption $n \geq 2d$ so $\sqrt{d/(8n)} \leq 1/4$). Thus in all three cases, $\mathrm{rej}_P + \mathrm{err}_Q \geq \sqrt{d/(32n)}$. Hence,

$$\mathbb{E}\left[\mathrm{rej}_P + \mathrm{err}_Q \mid U \leq k/2, V \leq d/2\right] \geq \sqrt{\frac{d}{32n}}$$

Since $U \leq k/2, V \leq d/2$ happens with probability $\geq 1/2$, we have that $\mathbb{E}[\mathrm{rej}_P + \mathrm{err}_Q] \geq \frac{1}{2}\sqrt{d/(32n)}$ as required. This establishes eq. (G.31). □

We now prove our agnostic lower bound.

*Proof of Lemma G.13.* Let $\mathcal{X} = \mathbb{N}$ and $\mathcal{H}$ consist of the singleton functions that are 1 at one integer and 0 elsewhere. The VC dimension of $\mathcal{H}$ is easily seen to be 1.

Consider first the case in which $\tilde{\eta} \geq \sqrt{\eta/8}$. In this case, we must construct distributions $\mu, \tilde{\mu}$ and $f \in \mathcal{H}$ such that, $\mathbb{E}[\mathrm{err}_{\tilde{\mu}}(h|_S) + \mathrm{rej}_{\mu}(S)] \geq \tilde{\eta}$. This is trivial: let $\mu$ be arbitrary and $\tilde{\mu}(1,1) = \tilde{\eta}$ and $\tilde{\mu}(1,0) = 1 - \tilde{\eta}$. It is easy to see that no classifier has error less than $\tilde{\eta}$ since $\tilde{\eta} \leq 1/2$.

Thus it suffices to give $\mu, \tilde{\mu}$ and $f \in \mathcal{H}$ such that, $\mathrm{err}_{\tilde{\mu}}(f) = 0$, $\mathrm{err}_{\mu}(f) = \eta$, and,

$$\mathbb{E}_{\substack{(\boldsymbol{x},\boldsymbol{y})\sim\mu^n \\ (\tilde{\boldsymbol{x}},\tilde{\boldsymbol{y}})\sim\tilde{\mu}^n}}[\mathrm{err}_{\tilde{\mu}}(h|_S) + \mathrm{rej}_{\mu}(S)] \geq \sqrt{\eta/8}. \tag{G.32}$$

In particular, we will give a distribution over $f, \mu, \tilde{\mu}$ for which the above holds for the output $h|_S$ of any learning algorithm. By the probabilistic method, this implies that for each learning algorithm, there is at least $f, \mu, \tilde{\mu}$ for which eq. (G.32) holds. To this end, let $k = \lfloor\sqrt{2/\eta}\rfloor$.

Let $\mu$ be the distribution which has $\mu(x, 0) = \eta/2$ for $x \in [k]$ and $\mu(k+1, 0) = 1 - k\eta/2$, so $\mu$ has $y = 0$ with probability 1. Let $f$ be 1 for a uniformly random $x^* \in [k]$ so $\text{err}_\mu(f) = \eta/2$. Let $\tilde{\mu}$ be the distribution where $\tilde{\mu}(x, f(x)) = 1/k$ for $x \in [k]$, so $x$ is uniform over $[k]$ with $\text{err}_{\tilde{\mu}}(f) = 0$.

Now, given the above distribution over $f, \mu, \tilde{\mu}$, there is an optimal learning algorithm that minimizes $\mathbb{E}[\text{err}_{\tilde{\mu}} + \text{rej}_\mu]$. Moreover, notice that the algorithm learns nothing about $\mu$ or $\tilde{\mu}$ from the training data since $\mu$ is fixed as is the distribution over unlabeled examples. Thus the optimal learner, by symmetry, may be taken to make the same classification for all $x \in [k]$. Thus, consider three cases.

- The algorithm predicts $h|_S(x) = ?$ for all $x \in [k]$. In this case,

$$\text{rej}_\mu \geq k\frac{\eta}{2} = \lfloor\sqrt{2/\eta}\rfloor\frac{\eta}{2} \geq \frac{1}{2}\sqrt{\eta/2}$$

  using the fact that $\lfloor r \rfloor \geq r/2$ for $r \geq 1$.

- The algorithm predicts $h|_S(x) = 0$ for all $x \in [k]$. In this case,

$$\text{err}_{\tilde{\mu}} = \frac{1}{k} \geq \sqrt{\eta/2}.$$

- The algorithm predicts $h|_S(x) = 1$ for all $x \in [k]$. In this case, since $\eta \leq 1/2$, $k \geq 2$ and $\text{err}_{\tilde{\mu}} \geq 1/2$.

In all three cases, $\text{err}_{\tilde{\mu}} + \text{rej}_\mu \geq \sqrt{\eta/8}$ proving the lemma. □

We now present the proof of our transductive lower bound.

*Proof of Theorem 10.8.* Just as in the proof of Theorem 10.7, let $\mathcal{X} = \mathbb{N}$ and $\mathcal{H}$ again be the concept class of functions that have exactly $d$ 1's, which has VC dimension $d$. Again, let $P$ be the uniform distribution over $[N]$ for $N = 8n$.

We will construct a distribution $\nu$ over $\mathcal{H}$ and randomized adversary $\mathcal{A}(\boldsymbol{x}, \boldsymbol{z}, f)$ that outputs $\tilde{\boldsymbol{x}} \in \mathcal{X}^n$ such that, for all $L$,

$$\mathbb{E}[\text{rej}_{\boldsymbol{z}} + \text{err}_{\tilde{\boldsymbol{x}}}] \geq \lambda,$$

where $\lambda$ is a lower bound and expectations are over $\boldsymbol{x} \sim P^n, \boldsymbol{z} \sim P^m$ and $f \sim \nu$. By the probabilistic method again, such a guarantee implies that for any learner $L$, there exists some $f \in \mathcal{H}$ and deterministic adversary $\mathcal{A}(\boldsymbol{x}, \boldsymbol{z})$ where the above bound holds for that learner.

We will show two lower bounds that together imply the lemma. The first lower bound will follow from Theorem 10.7 and show that,

$$\mathbb{E}[\text{rej}_{\boldsymbol{z}} + \text{err}_{\tilde{\boldsymbol{x}}}] \geq K\sqrt{d/n},$$

where expectations are over $\boldsymbol{x} \sim P^n, \boldsymbol{z} \sim P^m, f \sim \nu$. Here $K$ is the constant from Theorem 10.7. To get this, the adversary $\mathcal{A}(\boldsymbol{x}, \boldsymbol{z}, f)$ simply ignores the true tests $\boldsymbol{z}$ and selects $\tilde{\boldsymbol{x}} \sim Q^m$. By linearity of expectation, for any learner, $\mathbb{E}[\text{rej}_{\boldsymbol{z}}] = \mathbb{E}[\text{rej}_P]$ and $\mathbb{E}[\text{err}_{\tilde{\boldsymbol{x}}}] = \mathbb{E}[\text{err}_Q]$.

It remains to show a distribution $\nu$ over $\mathcal{H}$ and adversary $A$ such that, for all learners,

$$\mathbb{E}[\text{rej}_{\boldsymbol{z}} + \text{err}_{\tilde{\boldsymbol{x}}}] \geq K\sqrt{d/m}, \tag{G.33}$$

for some constant $K$ and $m < n$ (for $m \geq n$, the previous lower bound subsumes this). Let $\nu$ be the uniform distribution over those $f \in \mathcal{H}$ that have all $d$ 1's in $[N]$, i.e., uniform over $\{f \in \mathcal{H} : \sum_{i \in [N]} f(i) = d\}$.

Let $A := \{x \in [N] : f(x) = 0\}$ and $B := \{x \in \mathbb{N} : f(x) = 1\}$ so $|A| = N - d$ and $|B| = d$.

Let $a = \lfloor \sqrt{md} \rfloor$ and $b = \lceil d/2 \rceil$, and $r = \lfloor m/(a + b) \rfloor$. The adversary will try to construct a dataset $\tilde{\boldsymbol{x}}$ with the following properties:

- $\tilde{\boldsymbol{x}}$ contains exactly $a$ distinct $\tilde{x} \in A$ and each has exactly $r$ copies. (Since $a \leq m < N - d$, this is possible.)
- There are exactly $b$ distinct $\tilde{x} \in B$ and each has exactly $r$ copies.
- The remaining $m - r(a + b)$ examples are all at $\tilde{x} = N + 1$ (these are "easy" as the learner can just label them 0 if it chooses).

We say $x$ is *seen* if $x \in \boldsymbol{x}$ (this notation indicates $x \in \{x_i : i \in [n]\}$ did not occur in the training set) and *unseen* otherwise. Now, we first observe that with probability $\geq 1/8$, the following event $E$ happens: there are at most $d-b$ seen 1's ($x_i \in B$) in the training set and there are at least $a$ distinct unseen 0's in the true test set $\boldsymbol{z}$, i.e.,

$$V_1 := |\{i \in [n] : x_i \in B\}| \leq d-b$$
$$V_0 := |\{z \in A : (z \in \boldsymbol{z}) \wedge (z \notin \boldsymbol{x})\}| \geq a$$

Note that $\mathbb{E}[V_1] = dn/N = d/8$. Markov's inequality guarantees that with probability $\geq 3/4$, $V_1 \leq d/2$ (otherwise $\mathbb{E}[V_1] > d/8$). Since $V_1$ is integer, this means that with probability $\geq 3/4$, $V_1 \leq \lfloor d/2 \rfloor = d-b$. Similarly, for any $i \in A$, the probability that it occurs in $\boldsymbol{z}$ and not in $\boldsymbol{x}$ is,

$$\left(1-\frac{1}{N}\right)^n \left(1-\left(1-\frac{1}{N}\right)^m\right) \geq \left(1-\frac{n}{N}\right)\left(1-e^{-\frac{m}{N}}\right) \geq \frac{7}{8}\cdot\frac{15}{16}\frac{m}{N} \geq 0.8\frac{m}{N},$$

where in the above we have used the fact that $(1-t) \leq e^{-t}$ for $t > 0$ and $1-e^{-t} \geq (15/16)t$ for $t \leq 1/8$. Hence, since $|A| = N-d$,

$$\mathbb{E}[V_0] \geq (N-d)0.8\frac{m}{N} \geq \left(\frac{7}{8}N\right)0.8\frac{m}{N} = 0.7m \geq 0.7a.$$

In particular, Markov's inequality implies that with probability at least 0.4, $V_0 \geq 0.5m$ (otherwise $\mathbb{E}[V_0] < 0.6(0.5m)+0.4m = 0.7m$). Thus, with probability $\geq 1-1/4-0.6 \geq 1/8$.

If this event $E$ does not happen, then the adversary will take all $\tilde{x} = N+1$, making learning easy. However, if $E$ does happen, then there must be at least $a$ unseen 0's in $\boldsymbol{z}$ and $b$ unseen 1's and the the adversary will select $a$ random unseen 0's from $\boldsymbol{z}$ and $b$ random unseen 1's, uniformly at random. It will repeat these examples $r$ times each, add $m - r(a+b)$ copies of $\tilde{x} = N+1$, and permute the $m$ examples.

Now that the adversary and $\nu$ have been specified, we can consider a learner $L$ that minimizes the objective $\mathbb{E}[\text{rej}_{\boldsymbol{z}} + \text{err}_{\tilde{\boldsymbol{x}}}]$. Clearly this learner may reject $N+1 \notin S$ as this cannot increase the objective. Now, by symmetry the learner may also be assumed to make the same classification on all $r(a+b)$ examples $\tilde{x} \in [N]$ as these examples are all unseen and indistinguishable since $B$ is uniformly random.

Case 1) If $h|_S(\tilde{\boldsymbol{x}}_i) = ?$ for all $i$ then

$$\text{rej}_{\boldsymbol{z}} = \frac{a}{m} = \frac{\lfloor \sqrt{md} \rfloor}{m} \geq \frac{\sqrt{md}/2}{m} = \frac{1}{2}\sqrt{\frac{d}{m}},$$

using the fact that $a \geq \sqrt{md}/2$ because $a \geq \sqrt{md}/2$ since $\lfloor t \rfloor \geq t/2$ for $t \geq 1$.

Case 2) If $h|_S(\tilde{\boldsymbol{x}}_i) = 0$ for all $i$ then,

$$\text{err}_{\tilde{\boldsymbol{x}}} = \frac{br}{m} \geq \frac{b\sqrt{m/d}}{4m} = \frac{b}{4\sqrt{md}} \geq \frac{d}{8\sqrt{md}} = \frac{1}{8}\sqrt{\frac{d}{m}}$$

In the above we have used the fact $b \geq d/2$ and that $r \geq \frac{1}{4}\sqrt{m/d}$, which can be verified by noting that:

$$\frac{m}{a+b} \geq \frac{m}{2a} \geq \frac{m}{2\sqrt{md}} = \frac{1}{2}\sqrt{\frac{m}{d}} \geq 1$$

and hence $r \geq \lfloor m/(a+b) \rfloor \geq \frac{1}{2} m/(a+b) \geq \frac{1}{4}\sqrt{m/d}$ again since $\lfloor t \rfloor \geq t/2$ for $t \geq 1$.

Case 3) If $h|_S(\tilde{\boldsymbol{x}}_i) = 0$ for all $i$ then, since $b \leq a$

$$\text{err}_{\tilde{\boldsymbol{x}}} = \frac{b}{a+b} \geq \frac{1}{2}.$$

In all three cases, we have,

$$\text{rej}_{\boldsymbol{z}} + \text{err}_{\tilde{\boldsymbol{x}}} \geq \frac{1}{8}\sqrt{\frac{d}{m}}.$$

Since $E$ happens with probability $\geq 1/2$, we have,

$$\mathbb{E}[\text{rej}_{\boldsymbol{z}} + \text{err}_{\tilde{\boldsymbol{x}}}] \geq \mathbf{Pr}[E]\mathbb{E}[\text{rej}_{\boldsymbol{z}} + \text{err}_{\tilde{\boldsymbol{x}}} \mid E] \geq \frac{1}{8} \cdot \frac{1}{8}\sqrt{\frac{d}{m}}.$$

This is what was required for eq. (G.33). □

## G.7 Tight Bounds Relating Train and Test Rejections

We now move on to tightly relating test and training rejections. As motivation, note that if one knew $P$ and $Q$, it would be natural to take $S^* := \{x \in \mathcal{X} : Q(x) \leq P(x)/\varepsilon\}$ for some $\varepsilon > 0$. For $x \notin S^*$, i.e., $x \in \bar{S}^*$, $P(x) < \varepsilon Q(x)$. This implies that $\text{rej}_P(S^*) = P(\bar{S}^*) < \varepsilon$. It is also straightforward to verify that $\text{err}_Q(h|_{S^*}) \leq \text{err}_P(h)/\varepsilon$. This means that if one can find $h$ of error $\varepsilon^2$ on $P$, e.g., using a PAC-learner, then this gives,

$$\text{rej}_P(S^*) + \text{err}_Q(h|_S^*) \leq 2\varepsilon.$$

This suggests that perhaps we could try to learn $P$ and $Q$ and approximate $S^*$. Unfortunately, this is generally impossible—one cannot even distinguish the case where $P = Q$ from the case where $P$ and $Q$ have disjoint supports with fewer than $\Omega(\sqrt{|\mathcal{X}|})$ examples.[†]

While we cannot learn $S^*$ in general, these sets $S^*$ do give the tightest bounds on $\text{rej}_Q$ in terms of $\text{rej}_P$.

**Lemma G.23.** *For any $S \subseteq \mathcal{X}$ and distributions $P, Q$ over $\mathcal{X}$ and any $\varepsilon \geq 0$ such that* $\text{rej}_P(S) \leq \text{rej}_P(S^*)$,

$$\text{rej}_Q(S) \leq \text{rej}_Q(S^*). \tag{G.34}$$

Note that the $\text{rej}_Q(S) \leq \text{rej}_P(S) + |P - Q|_{\mathsf{TV}}$ bound can be much looser than the bound in the above lemma. For example, $|P - Q|_{\mathsf{TV}} = 0.91$ yet $\text{rej}_Q(S^*) = 0.1$ for $\mathcal{X} = \{0, 1, \ldots, 100\}$, $P$ uniform over $\{1, \ldots, 100\}$, $Q$ uniform over $\{0, 1, \ldots, 9\}$, $\text{rej}_P(S) = 0$, and $\varepsilon = 0.1$ (since $S^* = \{1, 2, \ldots, 100\}$ and only $0 \notin S^*$). One can think of classifying images of a mushroom as "edible" or not based on training data of 100 species of mushrooms, with test data including one new species.

*Proof.* Since $\varepsilon Q(x) - P(x) > 0$ iff $x \notin S^*$,

$$\begin{aligned}
\varepsilon\,\text{rej}_Q(S^*) - \text{rej}_P(S^*) &= \sum_{x \notin S^*} \varepsilon Q(x) - P(x) \\
&\geq \sum_{x \notin S} \varepsilon Q(x) - P(x) = \varepsilon\,\text{rej}_Q(S) - \text{rej}_P(S) \\
\Rightarrow \quad \varepsilon(\text{rej}_Q(S^*) - \text{rej}_Q(S)) &\geq \text{rej}_P(S^*) - \text{rej}_P(S) \geq 0.
\end{aligned}$$

□

[†] To see this, consider the cases where $P = Q$ are both the uniform distribution over $\mathcal{X}$ versus the case where they are each uniform over a random partition of $\mathcal{X}$ into two sets of equal size. By the classic *birthday paradox*, with $O(\sqrt{|\mathcal{X}|})$ samples both cases will likely lead to random disjoint sets of samples.

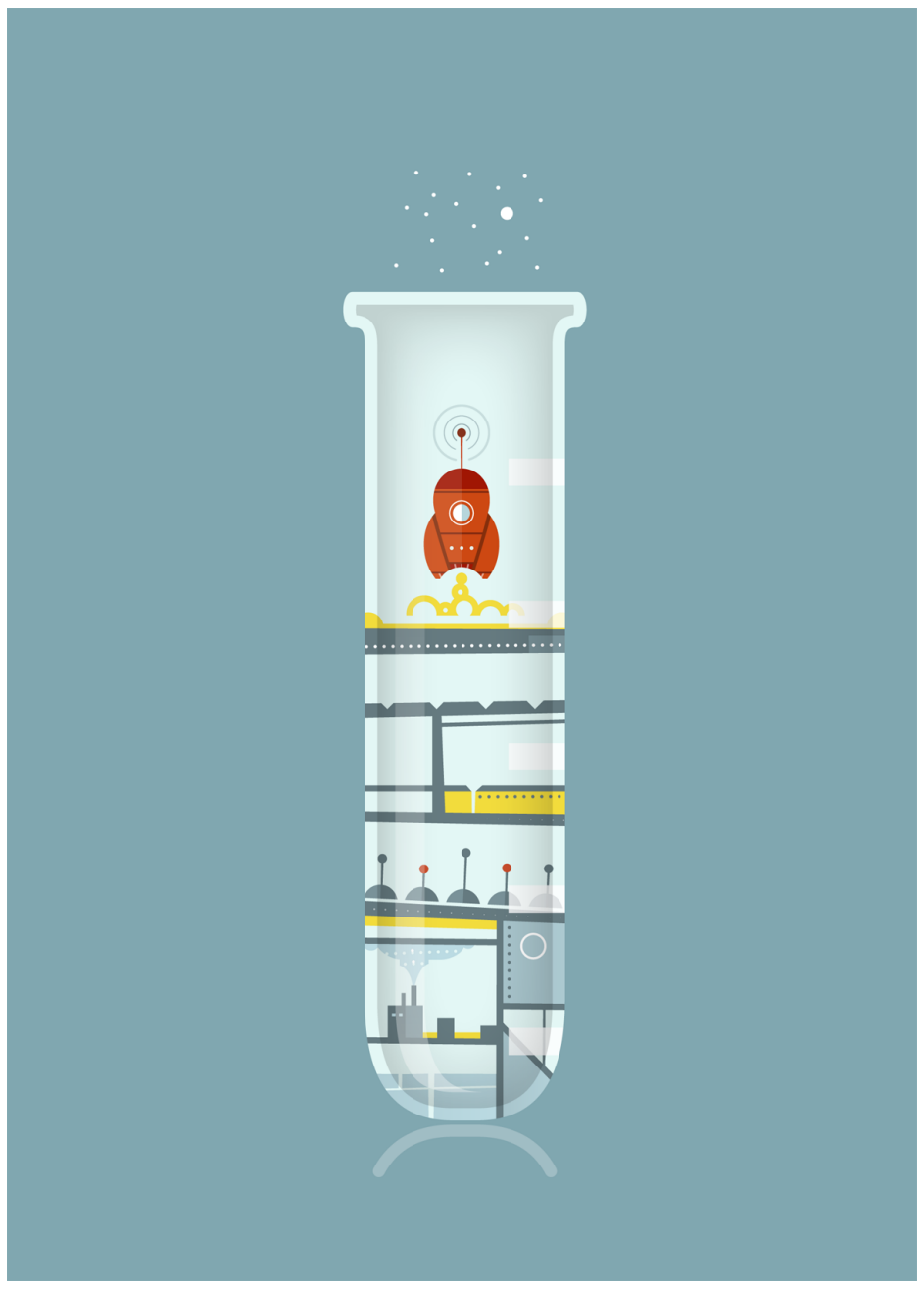

THIS THESIS WAS TYPESET using LaTeX, originally developed by Leslie Lamport and based on Donald Knuth's TeX. The body text is set in 11 point Egenolff-Berner Garamond, a revival of Claude Garamont's humanist typeface. The above illustration, *Science Experiment 02*, was created by Ben Schlitter and released under CC BY-NC-ND 3.0. A template that can be used to format a PhD dissertation with this look & feel has been released under the permissive AGPL license, and can be found online at github.com/suchow/Dissertate or from its lead author, Jordan Suchow, at suchow@post.harvard.edu.